\documentclass{article}

\usepackage[preprint]{main}

\usepackage[utf8]{inputenc} 
\usepackage[T1]{fontenc}    
\usepackage{hyperref}       
\usepackage{url}            
\usepackage{booktabs}       
\usepackage{amsfonts}       
\usepackage{nicefrac}       
\usepackage{microtype}      
\usepackage{xcolor}         

\usepackage{amsmath,amsfonts,bm}

\def\eqref#1{equation~\ref{#1}}

\def\1{\bm{1}}

\DeclareMathAlphabet{\mathsfit}{\encodingdefault}{\sfdefault}{m}{sl}
\SetMathAlphabet{\mathsfit}{bold}{\encodingdefault}{\sfdefault}{bx}{n}

\usepackage{graphicx}
\usepackage{subfigure}
\usepackage{wrapfig}

\usepackage{amsmath} 
\usepackage{algorithm}
\usepackage{algorithmic}

\usepackage{amsmath}
\usepackage{amssymb}
\usepackage{mathtools}
\usepackage{amsthm}
\usepackage{mathrsfs}

\theoremstyle{plain}
\newtheorem{theorem}{Theorem}[section]
\newtheorem{proposition}[theorem]{Proposition}

\newtheorem{corollary}[theorem]{Corollary}
\theoremstyle{definition}
\newtheorem{definition}[theorem]{Definition}

\theoremstyle{remark}
\newtheorem{remark}[theorem]{Remark}

\usepackage{multirow}
\usepackage{placeins}

\title{Geometry-Aware Operator Families for Structured Representation Learning}

\author{
    Zuyuan Zhang\\
    The George Washington University\\
    \texttt{zuyuan.zhang@gwu.edu}\\
    \And
    Fei Xu Yu\\
    The George Washington University\\
    \texttt{fxyu@gwu.edu}\\
    \And
    Tian Lan\\
    The George Washington University\\
    \texttt{tlan@gwu.edu}
}

\begin{document}

\maketitle

\begin{abstract}
The geometry of latent representations governs which components should interact and how information should propagate, making geometry-aware operator design a fundamental ingredient of structured deep representation learning. However, existing neural architectures typically rely on generic operator templates or geometry-specific constructions, creating a need for a unified framework that can derive admissible operators directly from fixed structural information while remaining adaptive to changing contexts. We introduce \emph{Geometry-Induced Operator Families} (GIOF), a general framework that converts fixed geometry into a structured family of propagation operators and dynamically selects an appropriate member of this family according to the current context. GIOF first transforms geometry-derived interaction channels into reusable generator bases, then combines them through a context-dependent selector and adaptive propagation scale, and finally realizes the selected operator through stable continuous-time propagation and a bottleneck residual layer. We establish theoretical guarantees covering parameter compression, identifiability, stability, locality, compositional structure, and oversmoothing behavior, while controlled experiments validate these mechanisms and experiments on PEMS-BAY and METR-LA achieve the lowest mean MAE across all reported regional-outage settings, improving over the strongest retained baseline by 2.4\%--8.8\% at 30\% missing sensors.
\end{abstract}

\section{Introduction}

Deep learning has achieved remarkable success by learning expressive hidden representations through standard computational building blocks,
such as dense linear maps, convolutions, graph message passing, and attention \citep{lecun2015deep,goodfellow2016deep,he2016deep,krizhevsky2012imagenet,vaswani2017attention,kipf2016semi,hamilton2017inductive,velivckovic2017graph,battaglia2018relational}. In these approaches, geometry may be encoded implicitly in learned representations or incorporated externally through regularization and architectural constraints, such as prescribed relational structure or symmetry \citep{gilmer2017neural,cohen2016group}, rather than treated as a primitive object that determines the operators acting on representations during learning.

When the latent space carries geometric structure, such as locality, directionality, hierarchy, obstruction, or compatibility, a natural approach is to construct geometry-conditioned operator families and learn within those families. In contrast to soft regularization or architectural enforcement \citep{belkin2003laplacian,coifman2006diffusion,bronstein2017geometric,bronstein2021geometric,chung1997spectral,grady2010discrete,lim2020hodge,hansen2020sheaf}, this makes geometry an explicit determinant of the admissible operator class. Existing approaches include equivariant constructions \citep{cohen2016group,finzi2020generalizing}, graph- and manifold-based convolutions \citep{monti2017geometric}, edge-conditioned filters \citep{simonovsky2017dynamic}, and parameterizations of hidden-state dynamics \citep{chen2018neural}. These methods provide important geometry-aware mechanisms, but typically specialize the operator form to a particular geometry or architectural construction. Our goal is instead a common finite-dimensional framework in which a prescribed geometry generates the admissible operator family and supports explicit analysis of locality, propagation, coefficient structure, and stability. A broader comparison with related work is given in Appendix~\ref{app:related-work}.

To this end, we introduce \emph{Geometry-Induced Operator Families} (GIOF). Let \(c\in\mathcal C\) denote the context available to a layer and \(H\in\mathbb R^{d\times p_h}\) the latent representation over \(d\) sites. A fixed \emph{geometry descriptor} \(\mathfrak S\) specifies an admissibility relation, pairwise geometric attributes, and a finite family of reusable geometry channels. Learning then uses \(c\) to select coefficients inside the operator family induced by \(\mathfrak S\), rather than generating an unrestricted \(d\times d\) matrix. In the propagation realization, the selector produces a generator \(L(c)\in\mathcal L_{\mathfrak S}\) and a scale \(\tau(c)\ge0\), yielding
$
P(c)=e^{\tau(c)L(c)},
$
which acts on the latent sites directly or on a bottleneck projection of \(H\). Thus geometry determines which transformations are admissible, while context determines which admissible transformation is used for the current representation.

This construction yields guarantees with direct operational meanings. One-step interaction operators have no off-diagonal entries outside the admissibility relation, while \(P(c)\) can transmit information only along admissible directed paths, so depth enlarges the receptive field without creating geometrically forbidden couplings. Because each admissible generator is Metzler and satisfies \(L(c)\mathbf 1=0\), the realized propagator obeys
$
P(c)\ge0,
\quad
P(c)\mathbf1=\mathbf1,
\quad
\|P(c)\|_{\infty\to\infty}=1.
$
For fixed \(c\), propagation therefore does not amplify hidden signals, or the direct feature gradient through \(P(c)\), in the sup norm. When \(L\) and \(\tau\) are differentiable in \(c\), Proposition~\ref{prop:properties-frechet-sensitivity} further gives the first-order context sensitivity
$
\|D_cP(c)[u]\|_{\infty\to\infty}
\le
\tau(c)\|D_cL(c)[u]\|_{\infty\to\infty}
+
\|L(c)\|_{\infty\to\infty}|D_c\tau(c)[u]|
$
for a perturbation direction \(u\). Thus context changes modify propagation through controlled changes of the generator and propagation scale rather than through an unrestricted dense operator. Products of realized propagators retain nonnegativity, constant preservation, and descriptor-level reachability across depth. Reusable geometry channels also compress the operator degrees of freedom and yield identifiable coefficients under the stated independence conditions, while the symmetric case characterizes the complementary oversmoothing and gradient-attenuation regime at large propagation scales. For self-conditioned selectors \(c=c(H)\), additional selector-Jacobian terms appear, and the corresponding nonlinear stability conditions are treated separately in Section~\ref{sec:layers} and Appendix~\ref{app:layers-self-conditioned}.

Together, these results turn latent geometry into an operator-level inductive bias rather than an auxiliary regularizer. Sections~\ref{sec:prelim}--\ref{sec:layers} construct the descriptor-induced families, establish their structural guarantees, and instantiate trainable interaction and propagation layers; Section~\ref{sec:learning} gives an idealized approximation and identifiability interpretation, with task-alignment and additional projection results deferred to Appendix~\ref{app:learning_appendix}.

\section{Deep Representation Spaces and Geometry Descriptors}
\label{sec:prelim}

This section introduces the fixed geometry descriptor from which all later
operator families are constructed. We first represent hidden features as
signals on finitely many latent sites, then specify directed admissibility and
reachability, and finally encode pairwise geometry through a finite channel
dictionary. The resulting masked channel matrices are the descriptor-level
objects used by the operator constructions in Section~\ref{sec:operators}.
Auxiliary descriptor extensions and kernel-level formulations are deferred to
Appendix~\ref{app:prelim-supplement}, while concrete architectural
instantiations are collected separately in
Appendix~\ref{app:canonical-examples-supplement}.

\subsection{Latent sites and representation signals}

\begin{definition}[Deep representation space]
\label{def:prelim-deep-representation-space}
A \emph{deep representation space} is a finite indexed set
\(X=\{z_1,\dots,z_d\}\) of distinct latent sites.
A scalar hidden representation \(h:X\to\mathbb R\) is identified with
\(h=(h_1,\dots,h_d)^\top\in\mathbb R^d\), where \(h_i:=h(z_i)\).
\end{definition}

Vector-valued signals, coordinate realizations, and further interpretation are
deferred to Appendix~\ref{app:prelim-signal-extensions}.
Throughout the paper we use the operator convention
\((Ah)_i=\sum_{j=1}^d A_{ij}h_j\), so \(A_{ij}\) represents influence from
the source site \(z_j\) to the receiving site \(z_i\). Accordingly, the first
argument of every ordered pair \((z_i,z_j)\) is the receiver and the second is
the source.

\begin{definition}[Directed admissibility and reachability]
\label{def:prelim-reachability}
An \emph{admissibility relation} is a binary relation
\(\mathcal E\subseteq X\times X\). We write
\(z_j\to z_i\) whenever \((z_i,z_j)\in\mathcal E\), and its binary mask is
\(M_{ij}:=\mathbf 1_{\mathcal E}(z_i,z_j)\).

We say that \(z_i\) is \emph{reachable from} \(z_j\), written
\(z_j\rightsquigarrow z_i\), if there exist \(\ell\ge1\) and sites
\(z_{k_0},\dots,z_{k_\ell}\) such that
$
z_j=z_{k_0}\to z_{k_1}\to\cdots\to z_{k_\ell}=z_i.
$
We write \(\Delta_X:=\{(z_i,z_i):1\le i\le d\}\) for the diagonal relation.
Reachability always refers to a directed path of positive length; reflexivity is added separately when needed.
\end{definition}

Thus \(\mathcal E\) specifies one-step locality, whereas
\(\rightsquigarrow\) describes the influence that can arise after repeated
composition. Canonical admissibility templates and the order-theoretic specialization are
developed in
Appendices~\ref{app:prelim-admissibility-templates}
and~\ref{app:prelim-order-specialization}.
A controlled learnable extension that preserves a fixed hard support envelope
is discussed separately in
Appendix~\ref{app:prelim-learnable-admissibility}.

\subsection{Pairwise geometry and the geometry descriptor}

\begin{definition}[Geometry descriptor]
\label{def:prelim-geometry-descriptor}
A \emph{geometry descriptor} is a tuple
$
\mathfrak S
=
\bigl(X,\mathcal E,\Psi,\{\phi_r\}_{r=1}^R,\mathcal R_+\bigr),
$
where \(\Psi:X\times X\to\mathbb R^q\) assigns geometric attributes to each
ordered receiver--source pair, and each geometry channel satisfies
$
\phi_r(z_i,z_j)=g_r\!\bigl(\Psi(z_i,z_j)\bigr)
$
for some \(g_r:\mathbb R^q\to\mathbb R\).
Each channel induces the masked channel matrix
$
(\Phi_r)_{ij}
:=
\mathbf 1_{\mathcal E}(z_i,z_j)\phi_r(z_i,z_j).
$
The set \(\mathcal R_+\subseteq\{1,\dots,R\}\) indexes channels satisfying
\((\Phi_r)_{ij}\ge0\) for all \(i\neq j\); coefficients associated with these
channels are constrained to be nonnegative in the operator families below.
\end{definition}

The descriptor therefore separates two roles: \(\mathcal E\) imposes hard
one-step support, while the channel dictionary parameterizes how admissible
pairs interact.

\paragraph{Standing regime.}
Unless stated otherwise, the structural theory assumes a fixed descriptor
\(\mathfrak S\). Context dependence in the later layer constructions acts
through coefficients selected within this fixed descriptor-induced family.
Context-dependent descriptors are discussed separately in
Appendix~\ref{app:prelim-pairwise-attributes}.

\paragraph{Compression requirement.}
The descriptor yields genuine structural compression only when the channel
dictionary is smaller than the ambient edgewise parameterization. In
particular, assigning one independent indicator channel to every admissible
off-diagonal edge recovers the full masked matrix space and removes the
channel-compression advantage. Concrete channel constructions and their correspondence with classical
architectural templates are deferred to
Appendix~\ref{app:canonical-examples-supplement}, where the abstract
descriptor-to-operator construction is instantiated explicitly.

\subsection{Forward link to operator families}

The masked channel matrices \(\{\Phi_r\}_{r=1}^R\) provide the
finite-dimensional realization of the descriptor geometry. Section~\ref{sec:operators}
uses them directly to construct interaction, generator, and propagation
families. The equivalent kernel-level formulation and its algebraic
identifiability properties are deferred to
Appendix~\ref{app:prelim-entrywise-form}.
Because every \(\Phi_r\) is supported on \(\mathcal E\), interaction operators inherit one-step locality
(Proposition~\ref{prop:operators-one-step-support});
the stronger pathwise and compositional consequences are developed in Section~\ref{sec:properties}.

\section{Geometry-Induced Operator Families}
\label{sec:operators}

This section compiles the descriptor geometry of Section~\ref{sec:prelim} into the hierarchy
\(\mathcal A_{\mathfrak S}\to\mathcal L_{\mathfrak S}\to\mathcal P_{\mathfrak S}\):
masked channels define admissible one-step interactions, generator constraints impose nonnegative off-diagonal rates and constant preservation, and exponentiation yields finite propagation.
We keep in the main text only the definitions and support facts needed by the later theory; algebraic identifiability, constrained-generator compilation, exponential identifiability, block lifting, and multiscale variants are deferred to Appendix~\ref{app:operators-supplement}.

\subsection{Interaction families}

Fix a geometry descriptor \(\mathfrak S\) as in Definition~\ref{def:prelim-geometry-descriptor}, with masked channel matrices \(\Phi_1,\dots,\Phi_R\).

\begin{definition}[Geometry-induced interaction family]
\label{def:operators-interaction-family}
The \emph{geometry-induced interaction family} is
$
\mathcal A_{\mathfrak S}
:=
\left\{
\sum_{r=1}^R \theta_r\Phi_r+\operatorname{Diag}(\delta)
\;\middle|\;
\theta\in\mathbb R^R,\ \delta\in\mathbb R^d,\ 
\theta_r\ge0\ \text{for }r\in\mathcal R_+
\right\}.
$
\end{definition}

The channel term controls admissible cross-site coupling, whereas
\(\operatorname{Diag}(\delta)\) supplies sitewise self-action.
Explicit entrywise forms are deferred to
Appendix~\ref{app:operators-matrix-remarks}.
Operator-level coefficient identifiability, including the diagonal-overlap
caveat, is treated in Appendix~\ref{app:operators-algebraic-facts}.

\begin{proposition}[One-step support of geometry-induced interactions]
\label{prop:operators-one-step-support}
For every \(A\in\mathcal A_{\mathfrak S}\) and \(i\neq j\),
$
(z_i,z_j)\notin\mathcal E
\quad\Longrightarrow\quad
A_{ij}=0.
$
\end{proposition}

Thus the interaction family changes coefficients on admissible pairs but does not create new one-step couplings.
We next restrict this family to generators whose exponentials define finite propagation.

\subsection{Generator families}

\begin{definition}[Geometry-induced generator family]
\label{def:operators-generator-family}
The \emph{geometry-induced generator family} associated with \(\mathfrak S\) is
$
\mathcal L_{\mathfrak S}
:=
\Bigl\{
L\in\mathcal A_{\mathfrak S}
\;\Big|\;
L_{ij}\ge 0 \ \text{for all } i\neq j,
\quad
L\mathbf 1=0
\Bigr\},
$
where \(\mathbf 1=(1,\dots,1)^\top\in\mathbb R^d\).
\end{definition}

Hence \(\mathcal L_{\mathfrak S}\) is the constant-preserving Metzler subclass of
\(\mathcal A_{\mathfrak S}\).
Its algebraic structure is deferred to
Appendix~\ref{app:operators-algebraic-facts}, while a sufficient condition
for nontrivial feasibility and the constrained compilation result are given in
Appendix~\ref{app:operators-normalization-compilation}.

\subsection{Propagation families}

For later composition results, fixed matrices
\(U\in\mathbb R^{d\times p}\) and \(V\in\mathbb R^{d\times q}\)
define the constrained subfamily
$
\mathcal L_{\mathfrak S;U,V}
:=
\{L\in\mathcal L_{\mathfrak S}:LU=0,\ V^\top L=0\},
$
where the columns of \(U\) are preserved right modes and
\(V^\top h\) are conserved linear observables.
Its finite polyhedral compilation is deferred to
Appendix~\ref{app:operators-normalization-compilation}.

\begin{definition}[Generator rate and normalized generator shape]
\label{def:operators-normalized-generator}
For \(L\in\mathcal L_{\mathfrak S}\), define
$
\rho(L)
:=
\frac1d\sum_{i=1}^d\sum_{j\ne i}L_{ij}
=
-\frac1d\operatorname{tr}(L),
\quad
\mathcal L_{\mathfrak S}^{(1)}
:=
\{L\in\mathcal L_{\mathfrak S}:\rho(L)=1\},
$
where the trace identity follows from \(L\mathbf1=0\).
\end{definition}

Interpreting \(L\) as the infinitesimal generator of a time-homogeneous
linear evolution,
\(
\frac{d}{dt}h(t)=Lh(t),
\)
the state after propagation magnitude \(\tau\) is uniquely given by
\(
h(\tau)=e^{\tau L}h(0).
\)
Hence the corresponding finite propagation operator is
\(
P_\tau=e^{\tau L},
\)
which motivates the following propagation family.

\begin{definition}[Geometry-induced propagation family]
\label{def:operators-propagation-family}
Assume \(\mathcal L_{\mathfrak S}\neq\{0\}\). Define
$
\mathcal P_{\mathfrak S}(0):=\{I\},
\quad
\mathcal P_{\mathfrak S}(\tau)
:=
\{e^{\tau L}:L\in\mathcal L_{\mathfrak S}^{(1)}\}
\ \ (\tau>0),
\quad
\mathcal P_{\mathfrak S}
:=
\bigcup_{\tau\ge0}\mathcal P_{\mathfrak S}(\tau).
$
Here \(L\) specifies normalized generator shape and \(\tau\) the propagation magnitude.
The exponential form makes the multi-hop structure explicit:
$
e^{\tau L}
=
I+\tau L+\frac{\tau^2}{2!}L^2+\frac{\tau^3}{3!}L^3+\cdots .
$
Because \(L\) encodes local propagation, \(L^k\) represents \(k\) successive
compositions of local interactions and can therefore transmit information
across \(k\)-hop neighborhoods. Thus \(e^{\tau L}\) combines propagation
over all depths, with the coefficient \(\tau^k/k!\) weighting the contribution
of the \(k\)-th propagation order.
\end{definition}

Every nonzero \(G\in\mathcal L_{\mathfrak S}\) has \(\rho(G)>0\) and therefore the unique scale--shape decomposition
\(G=\rho(G)\widehat L\), where
\(\widehat L=G/\rho(G)\in\mathcal L_{\mathfrak S}^{(1)}\).
Normalization removes the scalar ambiguity of a known generator but not possible aliasing under matrix exponentiation; the latter, together with constrained-generator compilation, is discussed in Appendix~\ref{app:operators-normalization-compilation}.
Positivity, constant preservation, quantitative reachability, and composition are developed in Section~\ref{sec:properties}, so they are not restated here in weaker form.
Accordingly, \(\mathcal A_{\mathfrak S}\) governs one-step correction,
\(\mathcal L_{\mathfrak S}\) admissible infinitesimal transport, and
\(\mathcal P_{\mathfrak S}\) finite propagation.
The repeated-composition support statement is recorded in
Appendix~\ref{app:operators-algebraic-facts};
block-valued and multiscale lifts are developed in
Appendices~\ref{app:operators-block-lifting} and~\ref{app:operators-multiscale}.
These families provide the analytical input for the next section.

\section{Structural Properties of Geometry-Induced Operators}
\label{sec:properties}

This section isolates the structural consequences needed later:
coefficient compression, stable finite propagation, spectral attenuation of
nonconstant modes, reachability-preserving locality, persistence under
composition, and restriction relative to the ambient masked matrix space.
Technical statements concerning orientation, general sensitivity,
trainability caveats, and depth-dependent expressivity are deferred to
Appendix~\ref{app:properties-supplement}.

\subsection{Coefficient geometry}

\begin{proposition}[Affine-hull dimension of the interaction family]
\label{prop:properties-affine-hull-dimension}
If
$
\{\Phi_1,\dots,\Phi_R,E_{11},\dots,E_{dd}\}
$
is linearly independent, then
$
\dim\!\bigl(\operatorname{aff}(\mathcal A_{\mathfrak S})\bigr)
=
\dim\!\bigl(\operatorname{span}(\mathcal A_{\mathfrak S})\bigr)
=
R+d.
$
Without linear independence, both dimensions are at most \(R+d\).
\end{proposition}

Thus a single interaction operator replaces free edgewise coefficients by at
most \(R\) reusable channel coefficients plus \(d\) diagonal degrees of
freedom. The same linear-independence condition also guarantees coefficient
identifiability; see
Proposition~\ref{prop:operators-operator-identifiability} and
Remark~\ref{rem:operators-diagonal-overlap}.
Appendix~\ref{app:properties-identifiability-remarks} records the corresponding
affine-hull interpretation. We return to the ambient-space restriction below.

\subsection{Stability and spectral attenuation of propagation}

The generator constraints convert the compressed interaction family into
stable finite-time transport. We first state the operator-level guarantee and
then quantify the attenuation induced by large propagation scales.

\begin{theorem}[Order preservation and sup-norm stability of finite propagation]
\label{thm:properties-order-supnorm-stability}
Let \(L\in\mathcal L_{\mathfrak S}\), \(\tau\ge0\), and
\(P=e^{\tau L}\). Then
$
P\ge0,
\quad
P\mathbf1=\mathbf1,
\quad
\|P\|_{\infty\to\infty}=1.
$
Consequently,
$
h\le g \Longrightarrow Ph\le Pg,
\quad
\|Ph\|_\infty\le\|h\|_\infty
$
for all \(h,g\in\mathbb R^d\).
\end{theorem}

Hence \(e^{\tau L}\) is a stable geometry-aware averaging operator under the
column-signal convention. Spectral stability, Markov orientation, and general
Fr\'echet sensitivity are recorded in
Appendix~\ref{app:properties-dynamic-remarks}.

\begin{theorem}[Oversmoothing of nonconstant modes]
\label{thm:properties-propagation-oversmoothing}
Assume
$
L=L^\top\in\mathcal L_{\mathfrak S},
\quad
\ker(L)=\operatorname{span}\{\mathbf1\}.
$
Let
$
0=\lambda_1>\lambda_2\ge\cdots\ge\lambda_d,
\quad
\gamma:=-\lambda_2>0,
\quad
\overline h:=\frac{\mathbf1^\top h}{d}\mathbf1.
$
Then, for every \(\tau\ge0\),
$
\|e^{\tau L}h-\overline h\|_2
\le
e^{-\gamma\tau}
\|h-\overline h\|_2.
$
\end{theorem}

Thus the same propagation mechanism that is forward stable progressively
suppresses nonconstant modes. The corresponding effect on trainability is
captured next.

\begin{proposition}[Gradient attenuation in the symmetric propagation regime]
\label{prop:properties-propagation-gradient-attenuation}
Under the assumptions of
Theorem~\ref{thm:properties-propagation-oversmoothing}, for \(\tau\ge0\), let
\(P_\tau=e^{\tau L}\). Then, for every \(h\in\mathbb R^d\),
$
\left\|
\frac{\partial}{\partial\tau}P_\tau h
\right\|_2
\le
\|L\|_2e^{-\gamma\tau}
\|h-\overline h\|_2.
$
Moreover, for every symmetric perturbation
\(H\) satisfying \(H\mathbf1=0\),
$
\|D_LP_\tau[H]\|_{2\to2}
\le
\tau e^{-\gamma\tau}\|H\|_2.
$
Hence large propagation scales can produce exponentially small gradients on
nonconstant modes despite stable forward propagation.
\end{proposition}
Sequential propagation and additional conditioning caveats are deferred to
Appendix~\ref{app:properties-dynamic-remarks}.

\subsection{Pathwise locality}

Stability does not remove geometric locality: exponentiation enlarges one-step
support only along admissible paths.

\begin{theorem}[Reachability and quantitative propagation locality]
\label{thm:properties-reachability-preserving-propagation}
Let \(L\in\mathcal L_{\mathfrak S}\), \(\tau>0\), and
$
\mathcal E_L
:=
\{(z_i,z_j):i\ne j,\ L_{ij}>0\}.
$
Choose \(\nu>0\) such that
$
\nu\ge\max_i(-L_{ii}).
$
For \(i\ne j\),
$
(e^{\tau L})_{ij}>0
$
if and only if there exists a directed path from \(z_j\) to \(z_i\) in
\(\mathcal E_L\). If the shortest such path has length \(r\), then
$
(e^{\tau L})_{ij}
\le
\Pr\{\operatorname{Poisson}(\nu\tau)\ge r\}.
$
If no such path exists, the entry is zero.
\end{theorem}

Thus finite propagation creates no off-diagonal coupling outside
descriptor-level reachability, while the Poisson tail quantifies how path
length suppresses finite-time influence. The uniformization identity and its
depth interpretation are recorded in
Appendix~\ref{app:properties-locality-remarks}.

\subsection{Composition and persistent structural guarantees}

Let
$
\mathcal R_{\mathcal E}
:=
\Delta_X
\cup
\left\{
(z_i,z_j):z_j\rightsquigarrow z_i
\right\}
$
be the reflexive transitive closure of the admissibility relation.

\begin{definition}[Composable structural envelope]
\label{def:properties-composable-envelope}
For fixed \(U\) and \(V\), define
$
\mathscr C_{\mathcal E;U,V}
:=
\left\{
P\in\mathbb R^{d\times d}:
P\ge0,\;
P\mathbf1=\mathbf1,\;
PU=U,\;
V^\top P=V^\top,\;
\operatorname{supp}(P)\subseteq\mathcal R_{\mathcal E}
\right\}.
$
\end{definition}

\begin{proposition}[Structural guarantees survive composition]
\label{prop:properties-composable-envelope}
The set \(\mathscr C_{\mathcal E;U,V}\) is nonempty, compact, convex, and closed
under matrix multiplication. Moreover, for every finite sequence
\(Q_1,\ldots,Q_m\in\mathcal L_{\mathfrak S;U,V}\),
$
e^{Q_m}\cdots e^{Q_1}
\in
\mathscr C_{\mathcal E;U,V},
$
with no commutativity assumption.
\end{proposition}

Therefore nonnegative averaging, preserved right modes, conserved left
observables, and exclusion of unreachable couplings persist across propagation
depth. The envelope certifies structural properties rather than exact
realizability: a product of admissible exponentials need not be a single
exponential from the original family.
Appendix~\ref{app:properties-composition-remarks} makes this distinction
explicit, showing exact closure for commuting generators and giving a concrete
counterexample to closure in the general noncommuting case.
For self-conditioned stacks, these guarantees hold pointwise for the selected
propagators and do not by themselves imply nonexpansiveness of the complete
nonlinear input--output map; selector dependence and intervening nonlinear
transformations must be analyzed separately.

\subsection{Expressive restriction relative to the ambient masked space}

Let
$
\mathcal E_{\mathrm{off}}
:=
\{(z_i,z_j)\in\mathcal E:\ i\neq j\},
$
and define the ambient masked matrix space
$
\mathcal M_{\mathcal E}
:=
\Bigl\{
A\in\mathbb R^{d\times d}
\;\Big|\;
A_{ij}=0
\ \text{whenever } i\neq j \text{ and } (z_i,z_j)\notin \mathcal E
\Bigr\}.
$
This is the correct ambient comparison class, because the point of the framework is not sparsity alone but compression from free edgewise parameters to reusable geometric channels; see also Appendix~\ref{app:properties-ambient-remarks}, Remark~\ref{rem:properties-why-compare-with-masked-space}.
The space \(\mathcal M_{\mathcal E}\) is therefore a linear subspace of
\(\mathbb R^{d\times d}\) with
$
\dim(\mathcal M_{\mathcal E})
=
d+|\mathcal E_{\mathrm{off}}|.
$

\begin{theorem}[Strict expressive restriction via channel compression]
\label{thm:properties-strict-expressive-restriction}
If
$
R<|\mathcal E_{\mathrm{off}}|,
$
then
$
\mathcal A_{\mathfrak S}
\subsetneq
\mathcal M_{\mathcal E}.
$
\end{theorem}

Hence channel reuse imposes a genuine single-layer operator restriction beyond
support sparsity.
This is a single-layer operator-space statement: it neither bounds the
functional capacity of unrestricted self-conditioned selectors nor implies a
depth-independent restriction on the algebra generated by repeated layers.
These two qualifications are made precise in
Appendices~\ref{app:properties-ambient-remarks}
and~\ref{app:properties-composition-remarks}, respectively.
The restriction should therefore be interpreted as a structured inductive
bias: the model assumes fewer reusable geometric mechanisms than admissible
directed edges.
These structural properties provide the analytical basis for the geometry-aware layer constructions studied next.

\section{From Geometry to Deep Layers}
\label{sec:layers}

This section turns the operator hierarchy
\(\mathcal A_{\mathfrak S}\to\mathcal L_{\mathfrak S}\to\mathcal P_{\mathfrak S}\)
into trainable network modules.
We retain two core realizations: interaction operators act as structured
residual corrections, while generator operators induce stable latent
propagation through matrix exponentiation.
The experiments instantiate the propagation realization with the residual
bottleneck defined below; the interaction adapter is retained as the
corresponding one-step realization of the same geometry-induced principle.
Balanced transport, multiscale extensions, and implementation details are
deferred to Appendix~\ref{app:layers-supplement}.

\subsection{Geometry-conditioned selectors and layer parameterization}

Let \(\mathcal C\subseteq\mathbb R^{p_c}\) be the context domain and
\(h\in\mathbb R^d\) the representation acted on by the layer.
The context \(c\in\mathcal C\) may coincide with \(h\), be derived from
another representation stream, or include auxiliary information.
Its dimension \(p_c\) is independent of the feature width used by the
matrix-valued realization below.
For interaction layers, we use a continuous selector
$
A:\mathcal C\to\mathcal A_{\mathfrak S},
\quad
A(c)
=
\sum_{r=1}^R\theta_r(c)\Phi_r+\operatorname{Diag}(\delta(c)),
\quad
\theta_r(c)\ge0\ \text{for }r\in\mathcal R_+.
$
Thus learning selects context-dependent operators inside the
descriptor-induced family rather than arbitrary matrices.
Implementation details are deferred to
Appendix~\ref{app:layers-implementation-remarks}.

\subsection{Interaction layers: geometry-aware residual adapters}

\begin{definition}[Geometry-aware residual adapter]
\label{def:layers-geometry-aware-residual-adapter}
The geometry-aware residual adapter is
$
\mathcal R_{\mathfrak S}(c,h)
:=
h+A(c)h,
\quad
A(c)\in\mathcal A_{\mathfrak S}.
$
\end{definition}

\paragraph{Conditional locality.}
For fixed \(c\), Proposition~\ref{prop:operators-one-step-support} gives
$
(z_i,z_j)\notin\mathcal E,\ i\neq j
\quad\Longrightarrow\quad
A_{ij}(c)=0,
$
and hence
\(D_h\mathcal R_{\mathfrak S}(c,h)=I+A(c)\)
has no off-diagonal entries outside \(\mathcal E\).
This is a conditional matrix-level statement: if the selector is
self-conditioned, \(A=A(h)\), then
$
DF(h)[v]=(I+A(h))v+DA(h)[v]h,
$
so matrix support alone does not guarantee functional locality.
A precise locality condition and counterexample are deferred to
Appendix~\ref{app:layers-self-conditioned}.

\subsection{Propagation layers: stable latent evolution}

\paragraph{Constructive generator parameterization.}
The interaction family permits general signed geometry channels, whereas
stable propagation requires nonnegative off-diagonal rates.
For each \(r\in\mathcal R_+\), define the generator-form channel
$
B_r
:=
\Phi_r-\operatorname{Diag}(\Phi_r\mathbf 1),
$
and let
$
\mathcal R_{\mathrm p}
:=
\{\,r\in\mathcal R_+ : B_r\neq 0\,\}.
$
For every \(r\in\mathcal R_{\mathrm p}\), define the rate-normalized
generator basis
$
\widehat B_r
:=
\frac{B_r}{\rho(B_r)}.
$
By construction,
$
B_r\mathbf 1=0,
\quad
(B_r)_{ij}\ge 0
\quad (i\neq j),
\quad
\widehat B_r\in\mathcal L_{\mathfrak S}^{(1)}.
$

A context-dependent selector produces nonnegative mixture weights,
$
\alpha(c)
=
\operatorname{softmax}(f_\theta(c)),
\quad
\alpha_r(c)\ge0,
\quad
\sum_{r\in\mathcal R_{\mathrm p}}\alpha_r(c)=1,
$
and defines the realized generator
$
L(c)
=
\sum_{r\in\mathcal R_{\mathrm p}}
\alpha_r(c)\widehat B_r.
$
Hence
$
L(c)
\in
\operatorname{conv}
\{\widehat B_r:r\in\mathcal R_{\mathrm p}\}
\subseteq
\mathcal L_{\mathfrak S}^{(1)}
$
for every context \(c\).
This parameterization therefore defines a certified tractable subfamily
of the normalized generator family; it need not span all of
\(\mathcal L_{\mathfrak S}^{(1)}\).

Unlike the interaction family
\(
A=\sum_r\theta_r\Phi_r+\operatorname{Diag}(\delta)
\),
the propagation realization has no free diagonal term:
the diagonal is determined by the constant-preservation constraint
\(L(c)\mathbf 1=0\).
Thus \(\Phi_r\) encode admissible pairwise interactions, whereas
\(B_r\) are their generator-form counterparts used for stable finite
propagation.

\begin{definition}[Geometry-aware propagation layer]
\label{def:layers-geometry-aware-propagation-layer}
Assume \(\mathcal L_{\mathfrak S}\neq\{0\}\).
Let \(L(c)\) be the constructive selector above; more generally, the definition
applies to any continuous map
\(
L:\mathcal C\to\mathcal L_{\mathfrak S}^{(1)}
\).
Let
\(
\tau:\mathcal C\to[0,\infty)
\)
be continuous.
The propagation layer is
$
\mathscr P_{\mathfrak S}(c,h)
:=
P(c)h,
\quad
P(c):=e^{\tau(c)L(c)}.
$
\end{definition}

\paragraph{Residual GIOF realization.}
For a matrix-valued hidden state
$
H\in\mathbb R^{d\times p_h},
$
the propagator acts along the \(d\) latent sites and is shared across
feature channels.
To reduce propagation cost, we first project the features to a
\(q\)-dimensional bottleneck,
$
F
=
HW_{\rm in},
\qquad
W_{\rm in}\in\mathbb R^{p_h\times q},
\qquad
F\in\mathbb R^{d\times q},
$
and map the geometry-induced correction back with
\(
W_{\rm out}\in\mathbb R^{q\times p_h}
\).
The resulting GIOF layer is
\begin{equation}
\label{eq:layers-giof-layer}
\operatorname{GIOF}_{\mathfrak S}(c,H)
=
H+
\Bigl[
e^{\tau(c)L(c)}
\bigl(HW_{\rm in}\bigr)
-
HW_{\rm in}
\Bigr]W_{\rm out}.
\end{equation}
Equivalently, with
\(
P(c)=e^{\tau(c)L(c)}
\),
$
\operatorname{GIOF}_{\mathfrak S}(c,H)
=
H+
\bigl[P(c)-I_d\bigr]
HW_{\rm in}W_{\rm out}.
$
Here \(L(c)\) selects the geometry of propagation, while
\(\tau(c)\) controls its propagation scale.
The operator \(P(c)-I_d\) therefore measures the geometry-induced
change relative to the unpropagated bottleneck representation.
When \(q=p_h\) and
\(
W_{\rm in}=W_{\rm out}=I_{p_h}
\),
Eq.~\eqref{eq:layers-giof-layer} reduces to
$
\operatorname{GIOF}_{\mathfrak S}(c,H)
=
P(c)H,
$
which is the columnwise extension of
Definition~\ref{def:layers-geometry-aware-propagation-layer}.
The bottleneck realization used in the experiments is detailed in
Appendix~\ref{app:exp_architecture}.
For every fixed context \(c\), \(P(c)\in\mathcal P_{\mathfrak S}\).
Therefore Theorem~\ref{thm:properties-order-supnorm-stability} and
Theorem~\ref{thm:properties-reachability-preserving-propagation} apply
pointwise:
$
P(c)\ge0,
P(c)\mathbf1=\mathbf1,
P(c)_{ij}=0
\ \text{if } i\neq j \text{ and } z_j\not\rightsquigarrow z_i.
$
Thus propagation preserves nonnegativity and constant signals while
allowing influence only through descriptor-admissible paths.
When the selector is self-conditioned, the preceding guarantees remain
valid pointwise for each realized propagator, but they do not automatically
extend to the complete nonlinear map \(h\mapsto e^{Q(h)}h\).
In particular, global order preservation, nonexpansiveness, and functional
locality require additional control of selector dependence.
Sufficient bi-Lipschitz and ancestor-locality conditions are given in
Appendix~\ref{app:layers-self-conditioned},
Propositions~\ref{prop:layers-self-conditioned-stability}
and~\ref{prop:layers-functional-ancestor-locality}.
Across depth, Proposition~\ref{prop:properties-composable-envelope}
already shows that products of realized propagators remain nonnegative,
row-stochastic, and supported within descriptor-level reachability.
For self-conditioned stacks, this remains a statement about the realized
operator product; the dependence of the selectors on intermediate
representations must still be treated separately.

\paragraph{Structured operator selection.}
Both constructions implement the same architectural principle:
learning selects context-dependent operators from descriptor-induced
families rather than searching over arbitrary matrices in
\(\mathbb R^{d\times d}\).
Residual layers realize the selected operator as a correction, whereas
propagation layers realize it as latent evolution.
These constructions therefore provide the bridge from the operator theory
above to the canonical architectural examples studied next.
Representative local, directional, generator-based, barrier-aware,
multiscale, and topological instantiations are collected in
Appendix~\ref{app:canonical-examples-supplement}.

\section{Approximation and Learning Interpretation}
\label{sec:learning}

Restricting a learned operator to \(\mathcal A_{\mathfrak S}\) introduces a structured inductive bias. To isolate that effect, let \(B\in\mathbb R^{d\times d}\) denote an unrestricted target and define
$
\Pi_{\mathcal A_{\mathfrak S}}(B)
:=
\arg\min_{A\in\mathcal A_{\mathfrak S}}
\|B-A\|_F^2.
$

\begin{theorem}[Best structured approximation]
\label{thm:main_best_structured_approx}
For every \(B\in\mathbb R^{d\times d}\),
\(\Pi_{\mathcal A_{\mathfrak S}}(B)\) exists and is unique.
\end{theorem}

This Frobenius projection is an auxiliary structural reference, not a claim that shared-parameter end-to-end training literally performs metric projection. With non-isotropic observations, the natural objective becomes data weighted; Proposition~\ref{prop:learning-data-weighted-projection} and the coefficient-space quadratic program are given in Appendix~\ref{app:learning-task-alignment} and Appendix~\ref{app:learning_appendix}, respectively.
Operator uniqueness should also be distinguished from recovery of its coefficients from observations. Let \(Q_1,\ldots,Q_k\) denote the fixed geometry-channel and diagonal basis matrices and write
$
A(\alpha)=\sum_{a=1}^k\alpha_aQ_a.
$
For shared probes \(h_1,\ldots,h_N\), define
$
\Gamma_{ab}
=
\frac1N\sum_{n=1}^N
\langle Q_ah_n,Q_bh_n\rangle.
$

\begin{proposition}[Identifiability from shared-operator probes]
\label{prop:learning-probe-identifiability}
For the unconstrained least-squares problem
$
\frac1N\sum_{n=1}^N
\|A(\alpha)h_n-y_n\|_2^2,
$
the coefficient vector is unique if and only if
\(\Gamma\succ0\).
In particular, linear independence of
\(\{Q_a\}_{a=1}^k\)
together with the empirical covariance
$
\frac1N\sum_{n=1}^Nh_nh_n^\top\succ0
$
is sufficient, whereas matrix-basis independence alone is not.
\end{proposition}

Thus algebraic identifiability and observational identifiability are distinct: the former concerns uniqueness of a coefficient representation once the operator is known, whereas the latter additionally requires informative probes. The same projection principle extends pointwise to input-dependent operator fields, while approximation bias, data-weighted recovery, and the spectral criterion for useful propagation are deferred to Appendix~\ref{app:learning-task-alignment}. These results are interpretive statements about the structured hypothesis class and do not imply task-level gains without alignment between the descriptor and the data.

\section{Experiments}
\label{sec:experiments}

\providecommand{\GIOFbox}[2]{%
  \fbox{\parbox[c][#1][c]{%
  \dimexpr\linewidth-2\fboxsep-2\fboxrule\relax}{%
  \centering\footnotesize #2}}}
\providecommand{\GIOFtbd}{\texttt{TBD}}

\begin{figure}[!t]
\centering
\begin{minipage}[t]{0.24\linewidth}
    \centering
    \includegraphics[width=\linewidth]{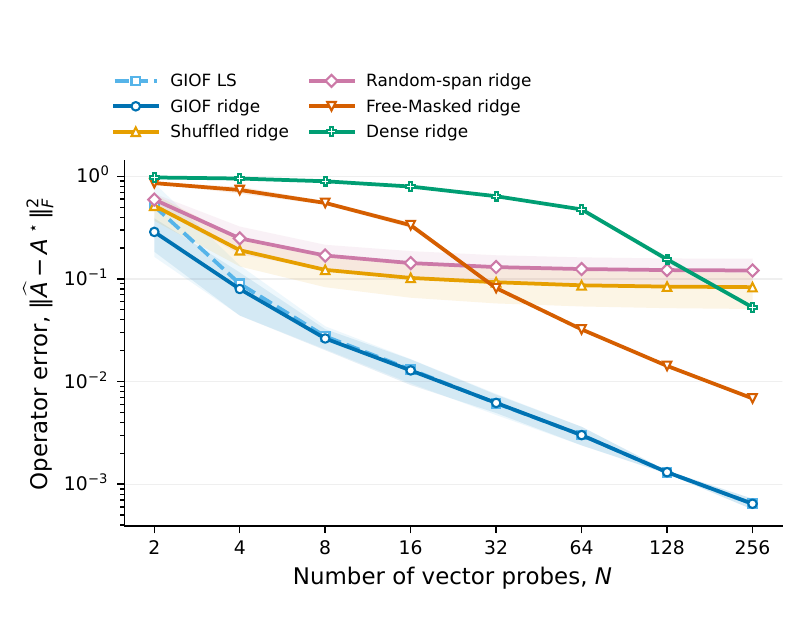}
    \vspace{-1.5mm}
    
    {\scriptsize\textbf{(a)} Probe recovery}
\end{minipage}\hfill
\begin{minipage}[t]{0.24\linewidth}
    \centering
    \includegraphics[width=\linewidth]{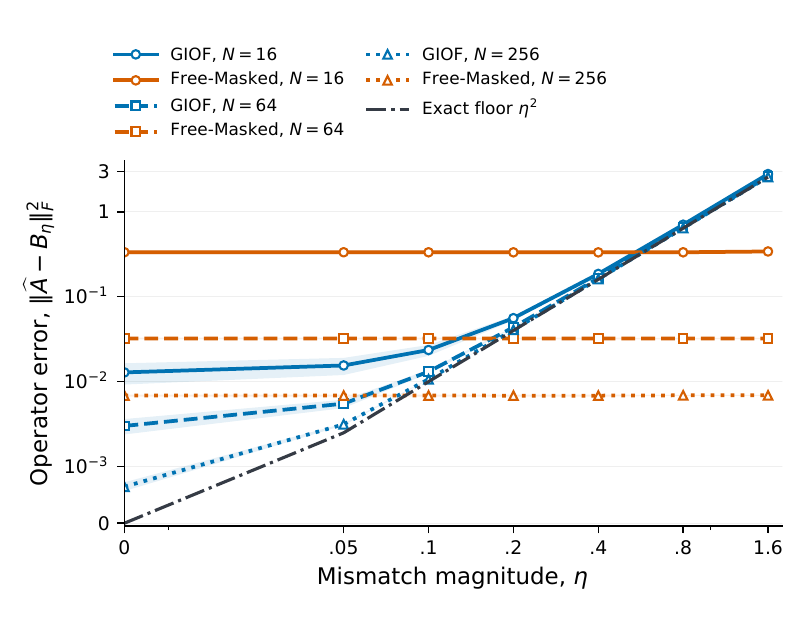}
    \vspace{-1.5mm}
    
    {\scriptsize\textbf{(b)} Mismatch cost}
\end{minipage}\hfill
\begin{minipage}[t]{0.24\linewidth}
    \centering
    \includegraphics[width=\linewidth]{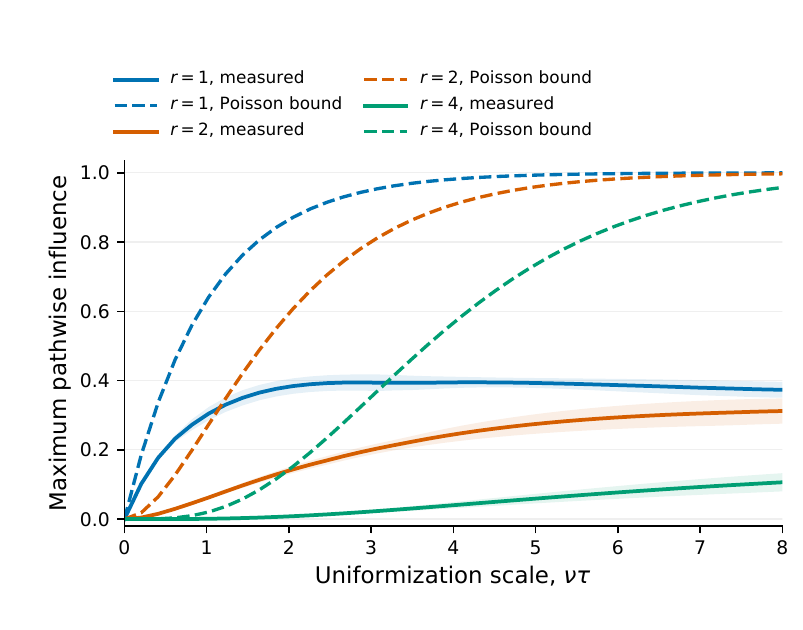}
    \vspace{-1.5mm}
    
    {\scriptsize\textbf{(c)} Pathwise influence}
\end{minipage}\hfill
\begin{minipage}[t]{0.24\linewidth}
    \centering
    \includegraphics[width=\linewidth]{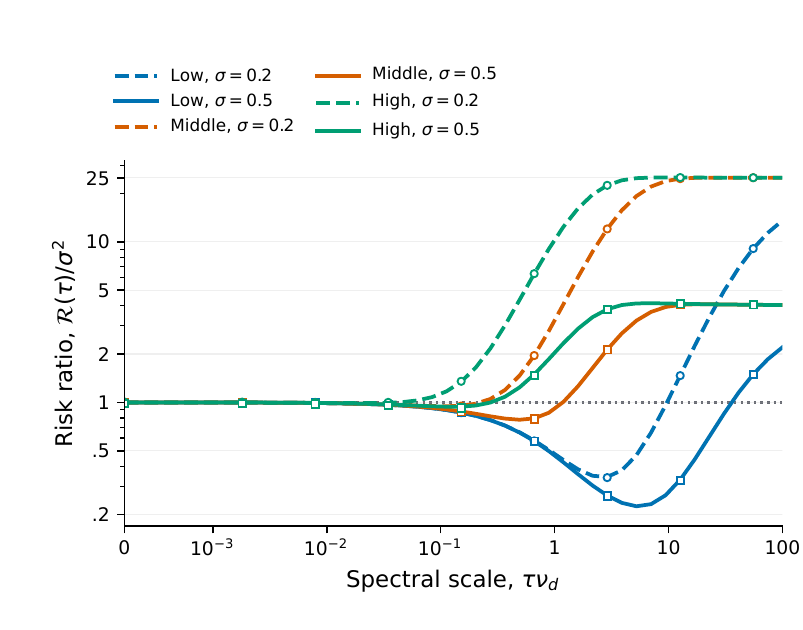}
    \vspace{-1.5mm}
    
    {\scriptsize\textbf{(d)} Spectral risk}
\end{minipage}

\vspace{-1mm}
\caption{
Controlled numerical checks of the theoretical mechanisms.
\textbf{(a)} Structured probe recovery improves rapidly with the number of
vector probes, while less structured estimators require substantially more
observations.
\textbf{(b)} Under matched geometry, the structured estimator approaches the
exact projection floor, whereas increasing mismatch exposes the corresponding
approximation cost.
\textbf{(c)} Positive-rate paths exhibit finite-scale influence controlled by
the corresponding Poisson-tail envelopes, with disconnected leakage remaining
numerically zero.
\textbf{(d)} Propagation reduces risk at intermediate scales for spectrally
aligned signals, while excessive smoothing eventually increases risk.
}
\label{fig:giof_theory}
\end{figure}

We test geometric channel sharing beyond graph sparsity through
controlled operator checks and real sensor reconstruction.
Figure~\ref{fig:giof_theory} reports the completed controlled checks of
probe recovery, mismatch bias, pathwise influence, and useful versus
excessive smoothing. Appendix~\ref{app:exp_details} specifies the full
protocol and reports additional numerical diagnostics.

\paragraph{Real data: reconstruction under regional sensor outages.}
We evaluate whether geometric channel sharing and adaptive multi-hop
propagation improve within-window reconstruction on PEMS-BAY and METR-LA
\citep{li2017diffusion}. Each example contains 24 five-minute observations per
sensor. Regional outages hide the complete window at sensor groups grown along
the fixed sensor graph, so successful reconstruction must use information
outside the missing region. We use 20\% and 100\% of the available training
days, and additionally evaluate independent sensor outages and point
missingness as controls. The five published baselines are GCN, GATv2, APPNP,
MoNet, and ECC. Each is inserted as the spatial module of the same compact
reconstruction network, so the comparison focuses on the spatial operator
rather than unrelated end-to-end architecture differences. All methods use the
same chronological splits, artificial masks, model seeds, and final update
budget. Full data-processing, architecture, and optimization details are given
in Appendix~\ref{app:exp_real_data}--\ref{app:exp_additional}.

\begin{wrapfigure}{r}{0.58\linewidth}
    \vspace{-8pt}
    \centering

    \begin{minipage}[t]{0.48\linewidth}
        \centering
        \includegraphics[width=\linewidth]{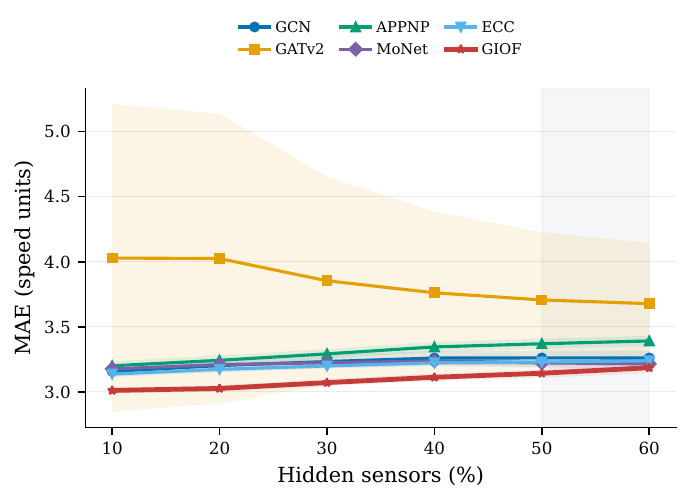}
        \vspace{-1.5mm}

        {\scriptsize\textbf{(a)} PEMS-BAY}
    \end{minipage}
    \hfill
    \begin{minipage}[t]{0.48\linewidth}
        \centering
        \includegraphics[width=\linewidth]{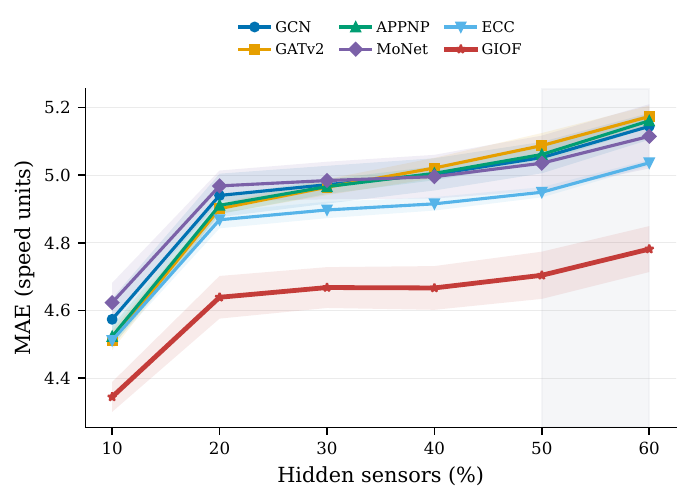}
        \vspace{-1.5mm}

        {\scriptsize\textbf{(b)} METR-LA}
    \end{minipage}

    \vspace{-1mm}
    \caption{Robustness to regional outages with 20\% training days.
    Test MAE is reported under increasing fractions of missing sensors using
    fixed checkpoints. The 60\% missingness setting lies outside the mask rates
    used during training.}
    \label{fig:giof_real_curves}
    \vspace{-8pt}
\end{wrapfigure}

\begin{table}[!t]
\centering
\caption{Regional-outage reconstruction at 30\% missing sensors and measured
computational cost. MAE is in the original speed units; 20\% and 100\% denote
the fraction of available training days. Values first average ten shared mask
realizations within each model seed and then report mean $\pm$ sample standard
deviation. Cost columns use PEMS-BAY with 100\% training
days and average the three final runs. All methods use the same final update
cap; training time includes validation.}
\label{tab:giof_real_main}
\scriptsize
\setlength{\tabcolsep}{3.2pt}
\resizebox{\linewidth}{!}{%
\begin{tabular}{lcccccccc}
\toprule
& \multicolumn{2}{c}{PEMS-BAY MAE}
& \multicolumn{2}{c}{METR-LA MAE}
& \multicolumn{4}{c}{Cost on PEMS-BAY}\\
\cmidrule(lr){2-3}\cmidrule(lr){4-5}\cmidrule(lr){6-9}
Method & 20\% & 100\% & 20\% & 100\%
& Params (K) & Train (min) & GPU (MiB) & Infer (ms)\\
\midrule
GCN   & $3.232\pm0.051$ & $3.208\pm0.020$ & $4.972\pm0.056$ & $5.229\pm0.081$ & 18.98 & 1.04 & 29.95 & 1.02\\
GATv2 & $3.853\pm0.801$ & $3.202\pm0.060$ & $4.964\pm0.024$ & $5.044\pm0.018$ & 19.52 & 1.54 & 51.99 & 1.86\\
APPNP & $3.291\pm0.036$ & $3.280\pm0.043$ & $4.966\pm0.022$ & $5.091\pm0.022$ & 18.43 & 1.56 & 47.58 & 2.20\\
MoNet & $3.223\pm0.027$ & $3.197\pm0.027$ & $4.984\pm0.055$ & $5.234\pm0.122$ & 23.42 & 3.46 & 107.63 & 5.37\\
ECC   & $3.198\pm0.019$ & $3.147\pm0.035$ & $4.897\pm0.024$ & $5.133\pm0.080$ & 36.64 & 1.42 & 49.28 & 1.96\\
\midrule
GIOF  & $\mathbf{3.070\pm0.025}$ & $\mathbf{3.072\pm0.013}$ & $\mathbf{4.668\pm0.060}$ & $\mathbf{4.599\pm0.026}$ & 23.14 & 1.89 & 69.56 & 2.31\\
\bottomrule
\end{tabular}}
\end{table}

At 30\% regional missingness, GIOF obtains the lowest mean MAE in all four
reported dataset/data-budget settings. Relative to the strongest retained
baseline in each column, the mean-MAE reduction ranges from 2.4\% to 8.8\%.
The cost measurements show that this gain is not free: GIOF is slower than the
lighter GCN/GATv2/APPNP/ECC modules, while its training and inference costs are
close to MoNet. We therefore interpret the result as an accuracy--structure
trade-off rather than a universal computational advantage.

\paragraph{Robustness to outage severity.}
Figure~\ref{fig:giof_real_curves} evaluates fixed 20\%-data checkpoints under
increasing regional-outage severity. GIOF achieves the lowest mean MAE across
all tested missing-sensor fractions on both datasets, including the 60\%
setting that lies outside the mask rates used during training. This trend
indicates that the advantage is not confined to the particular corruption
levels seen during optimization. Independent-sensor, point-missingness,
duration, graph-distance, and matched-ablation results are reported in
Appendix~\ref{app:exp_additional}. Predictive improvements remain empirical
observations and are not implied by the operator-level guarantees alone.

\section{Conclusion}

We introduced \emph{Geometry-Induced Operator Families} (GIOF), a framework that uses fixed latent geometry to define structured operator families while allowing context-dependent adaptation within them. GIOF provides a unified connection between geometric structure, stable propagation, locality, and trainable deep layers, together with explicit theoretical guarantees. Controlled and real-data experiments further support these properties and show the practical value of geometry-induced operator design under structured missingness.

\medskip

{
\small
\nocite{*}
\bibliographystyle{unsrtnat}
\bibliography{ref}
}

\appendix
\appendix

\section{Related Work}
\label{app:related-work}

\paragraph{Geometric deep learning and message passing.}
A large body of work incorporates geometry through graph or manifold structure. Spectral constructions derive filters from graph Laplacians and related harmonic structure \citep{chung1997spectral,belkin2003laplacian,coifman2006diffusion,defferrard2016convolutional}, while spatial graph neural networks aggregate information over prescribed neighborhoods \citep{kipf2016semi,hamilton2017inductive,gilmer2017neural,velivckovic2017graph,battaglia2018relational}. MoNet provides a general mixture-model view of local convolution on graphs and manifolds \citep{monti2017geometric}, and edge-conditioned convolution makes filters depend explicitly on edge attributes \citep{simonovsky2017dynamic}. These methods establish effective geometry-aware architectures, but the operator class is typically tied to a particular aggregation rule, kernel, or geometric domain. GIOF instead starts from a finite descriptor containing admissibility and reusable geometry channels, and treats the induced operator family itself as the object to be constructed and analyzed.

\paragraph{Symmetry and equivariance.}
Equivariant networks constrain operators by requiring compatibility with a prescribed group action, as in group-equivariant convolution and its extensions to Lie groups \citep{cohen2016group,finzi2020generalizing}. This is closely related in spirit to GIOF because geometry restricts the hypothesis class by construction rather than through a soft penalty. The distinction is that GIOF does not assume that the latent geometry is generated by a group action: directed locality, barriers, hierarchical relations, and other asymmetric admissibility structures can be encoded directly. Equivariance can therefore appear as a special structured regime, but it is not the primitive organizing assumption.

\paragraph{Dynamic and context-conditioned operators.}
Dynamic filter networks generate input-dependent filters \citep{jia2016dynamic}, edge-conditioned convolution generates filters from pairwise attributes \citep{simonovsky2017dynamic}, and attention uses context-dependent scores to form data-adaptive mixing operators \citep{vaswani2017attention,velivckovic2017graph}. GIOF also permits context-dependent selection, but separates the \emph{selector} from the \emph{admissible operator family}: the context changes coefficients within a descriptor-induced family, while support, channel sharing, and generator constraints are imposed independently of the selector. This separation enables operator-level guarantees to hold pointwise for every realized context. It also clarifies their limitation: if the selector is itself conditioned on the hidden state, derivatives through the selector must be controlled separately from the fixed-context propagator.

\paragraph{Diffusion, continuous-depth models, and higher-order geometry.}
Diffusion-based representation methods motivate propagation through geometry-aware generators \citep{coifman2006diffusion,grady2010discrete}, while neural ODEs model hidden representations through learned continuous-time dynamics \citep{chen2018neural}. GRAND specializes this viewpoint to graph neural diffusion \citep{chamberlain2021grand}, and neural sheaf diffusion uses sheaf structure to address heterophily and oversmoothing \citep{bodnar2022neural}. Higher-order constructions based on simplicial complexes, Hodge Laplacians, and sheaves further enrich the underlying geometry \citep{ebli2020simplicial,lim2020hodge,bodnar2021weisfeiler,roddenberry2021principled,hansen2020sheaf,barbero2022sheaf}. GIOF is complementary to these choices of geometry: its generator family enforces Metzler and constant-preserving constraints before exponentiation, yielding nonnegative, stable finite propagation and reachability guarantees for any descriptor that can be compiled into the finite channel representation.

\paragraph{Positioning of GIOF.}
The contribution is therefore not a new graph convolution, attention rule, diffusion equation, or equivariance principle in isolation. GIOF provides a common finite-dimensional descriptor-to-operator construction in which one can distinguish (i) admissible one-step interactions, (ii) feasible generators, (iii) finite propagation obtained by exponentiation, and (iv) context-dependent selection inside these fixed families. This organization makes it possible to state exact support and reachability properties, quantify the reduction from free edgewise parameters to shared geometry channels, separate algebraic from observational identifiability, and analyze stable propagation and its first-order sensitivity within one framework. The projection results in Appendix~\ref{app:learning_appendix} are likewise an idealized characterization of the structured hypothesis class rather than an equivalence with end-to-end task optimization.

\section{Supplementary Material for Section~\ref{sec:prelim}}
\label{app:prelim-supplement}

This appendix supplements Section~\ref{sec:prelim} with extensions that are useful for interpretation and implementation but are not required for the minimal main-text construction. We follow the dependency structure of the main text: we first clarify the signal space and auxiliary coordinate
realizations, then specialize directed admissibility and reachability, and
finally discuss descriptor variants and the equivalent kernel-level realization. Unless explicitly stated otherwise, the standing regime remains the fixed geometry descriptor \(\mathfrak S\) of
Definition~\ref{def:prelim-geometry-descriptor}; in particular, the hard admissibility relation and realized channel matrices are fixed when the later operator families are constructed.

\subsection{Signal-space extensions}
\label{app:prelim-signal-extensions}

\begin{remark}[Scalar and vector-valued signals]
\label{rem:prelim-scalar-vector-signals}
For conceptual clarity, Section~\ref{sec:prelim} is formulated for scalar
signals \(h\in\mathbb R^d\). If each site carries a \(c\)-dimensional feature
vector, the representation may instead be written as
$
H\in\mathbb R^{d\times c}.
$
A site-level operator \(A\in\mathbb R^{d\times d}\) then acts channelwise as
$
H\longmapsto AH.
$
Under column-wise vectorization, this separable action is equivalently
represented by \(I_c\otimes A\) on \(\mathbb R^{dc}\). Hence all support and
reachability statements at the site level extend immediately to this
separable lift.

More general architectures may also mix feature coordinates during
cross-site transfer. Such mixing requires a genuine block-valued operator
construction rather than only a change of notation; the corresponding
lifting is developed in Appendix~\ref{app:operators-block-lifting}.
Thus the scalar formulation isolates the site-level geometry, while the
block-valued extension supplies the additional feature-space mixing needed
for ordinary multi-channel architectures.
\end{remark}

\begin{remark}[What is and is not assumed]
\label{rem:prelim-what-is-and-is-not-assumed}
The definition of a deep representation space does not assume that \(X\) is a grid, a graph, a group orbit, a manifold discretization, or a partially ordered set.
Those are possible specializations, but none is required at the primitive level.
What matters in the main text is only that \(X\) is a finite indexed site family on which hidden activations can be interpreted as signals.
\end{remark}

\begin{remark}[Auxiliary coordinate realization]
When concrete coordinates are required, we introduce an auxiliary map
$
\iota:X\to\mathbb R^m.
$
This map is not part of the abstract identity of the latent sites:
distinct sites may have identical coordinates under \(\iota\).
It is used only when a particular admissibility rule or geometry channel
requires an ambient coordinate realization.
\end{remark}

\subsection{Admissibility templates}
\label{app:prelim-admissibility-templates}

\begin{remark}[Canonical admissibility templates]
\label{rem:prelim-canonical-admissibility-templates}
Suppose that \(X\) is additionally equipped with an auxiliary coordinate map
$
\iota:X\to\mathbb R^m
$
or with another external structure, depending on the application.
Then several familiar interaction patterns arise as special choices of \(\mathcal E\):
\begin{enumerate}
    \item \emph{Local grid or CNN stencil.}
    If \(r\ge 0\) is a radius, one may set
    \[
    (z_i,z_j)\in\mathcal E_{\mathrm{loc}}
    \quad\Longleftrightarrow\quad
    \|\iota(z_i)-\iota(z_j)\|_\infty\le r.
    \]

    \item \emph{Causal sequence geometry.}
    If the sites are temporally indexed as \(z_1,\dots,z_d\), one may impose
    \[
    (z_i,z_j)\in\mathcal E_{\mathrm{causal}}
    \quad\Longleftrightarrow\quad
    0\le i-j\le r.
    \]

    \item \emph{Graph neighborhood geometry.}
    If the sites form the vertices of a directed graph \(G\), let
    \(\mathcal N^{-}(z_i)\) denote the in-neighborhood of the receiving site
    \(z_i\). Consistently with the receiver--source convention of
    Section~\ref{sec:prelim}, one may set
    \[
    (z_i,z_j)\in\mathcal E_{\mathrm{graph}}
    \quad\Longleftrightarrow\quad
    z_j\in\mathcal N^{-}(z_i),
    \]
    so that an admissible graph edge carries information from \(z_j\) to \(z_i\).
    For an undirected graph, \(\mathcal N^{-}(z_i)\) may simply be replaced by the
    usual neighborhood \(\mathcal N(z_i)\). Self-loops may be included separately
    when desired.

    \item \emph{Windowed or global attention geometry.}
    If the sites are partitioned into windows \(W_1,\dots,W_s\), one may impose
    \[
    (z_i,z_j)\in\mathcal E_{\mathrm{win}}
    \quad\Longleftrightarrow\quad
    z_i\text{ and }z_j\text{ belong to the same window}.
    \]
    Full attention corresponds to \(\mathcal E=X\times X\).

    \item \emph{Multiscale hierarchy.}
    If the sites are organized across scales, one may let \(\mathcal E\) contain both same-scale local edges and cross-scale parent-child edges.
    This yields a geometry appropriate for pyramid, U-Net, or feature-fusion architectures.
\end{enumerate}
The point is that admissibility is not tied to a single architecture.
It is a common language for a broad class of direct-interaction patterns.
\end{remark}

\subsection{Order-theoretic specialization}
\label{app:prelim-order-specialization}

The templates above specify direct admissible interactions. Since the later
operator theory also depends on repeated composition, it is useful to
distinguish this one-step relation from the order structure that may, in
special cases, be induced by its reachability closure.

\begin{remark}[Admissibility versus reachability]
\label{rem:prelim-admissibility-vs-reachability}
The relation \(\mathcal E\) need not be reflexive or transitive.
By contrast, its reflexive-transitive closure is exactly the reachability relation together with identity.
This distinction matters conceptually:
admissibility describes which interactions are \emph{directly} allowed,
whereas reachability describes which influences become possible after repeated composition of such direct interactions.
\end{remark}

The preceding distinction leads to a genuine order structure precisely when
directed cycles do not destroy antisymmetry.

\begin{proposition}[When directed locality does and does not define an order]
\label{prop:prelim-order-specialization}
Let \(\preceq_{\mathcal E}\) denote the reflexive closure of reachability:
\[
z_j\preceq_{\mathcal E} z_i
\quad\Longleftrightarrow\quad
z_j=z_i
\ \text{or}\ 
z_j\rightsquigarrow z_i.
\]
Then:
\begin{enumerate}
    \item if the directed graph \((X,\mathcal E)\) is acyclic, the relation \(\preceq_{\mathcal E}\) is a partial order on \(X\);
    \item if \((X,\mathcal E)\) contains a directed cycle with at least two distinct vertices, then \(\preceq_{\mathcal E}\) fails to be antisymmetric and therefore is not a partial order.
\end{enumerate}
\end{proposition}

This proposition shows that partial order is only a special case.
The primitive notion in the framework is directed locality, not order.

Having clarified the consequences of the directed support relation, we now
turn from admissibility itself to the pairwise geometric information carried
by admissible receiver--source pairs.

\subsection{Pairwise geometric attributes}
\label{app:prelim-pairwise-attributes}

\begin{remark}[Directed versus symmetric attributes]
\label{rem:prelim-directed-vs-symmetric-attributes}
Even when the ambient geometry is symmetric, the attribute map \(\Psi\) need not be symmetric.
In particular,
$
\Psi(z_i,z_j)\neq \Psi(z_j,z_i)
$
is fully allowed and is often desirable in settings with preferred directions, anisotropic transport, multiscale refinement, or causal asymmetry.
\end{remark}

\begin{remark}[Fixed versus context-dependent descriptors]
\label{rem:prelim-static-vs-state-dependent-attributes}
The main structural theory is stated for a fixed descriptor
\(\mathfrak S\). Thus, unless explicitly stated otherwise,
\(\mathcal E\), the realized channel matrices
\(\{\Phi_r\}_{r=1}^R\), and the induced operator families do not vary with
the layer context.

A broader construction may allow the attribute map, channel values, or even
the admissibility relation to depend on a context \(c\), producing a
context-dependent descriptor \(\mathfrak S(c)\). For each fixed \(c\), the
fixed-descriptor structural statements apply to the realized descriptor
\(\mathfrak S(c)\). However, conclusions that require one common
reachability relation, one common coefficient parameterization, or projection
onto one fixed convex family require additional uniform assumptions across
contexts.
\end{remark}

\subsection{Learnable reweighting within a fixed admissibility envelope}
\label{app:prelim-learnable-admissibility}

The preceding discussion allows descriptor components to vary, but exact
locality is easiest to retain when a fixed hard support envelope is preserved.

\begin{remark}[Soft gating inside a fixed support envelope]
\label{rem:prelim-learnable-admissibility}
Let
$
\mathcal E_{\max}\subseteq X\times X
$
be a fixed maximal admissibility relation, and let
$
a_\eta:X\times X\to\mathbb R
$
be a learnable score. Define the soft gate
$
g_\eta(z_i,z_j)
=
\sigma\!\bigl(a_\eta(z_i,z_j)\bigr)
$
and the gated channel
$
\widetilde\phi_{r,\eta}(z_i,z_j)
=
\mathbf 1_{\mathcal E_{\max}}(z_i,z_j)\,
g_\eta(z_i,z_j)\,
\phi_r(z_i,z_j).
$
Then
$
(z_i,z_j)\notin\mathcal E_{\max}
\quad\Longrightarrow\quad
\widetilde\phi_{r,\eta}(z_i,z_j)=0
$
exactly. Consequently, every directed path supported by the gated channels is
contained in the reachability relation induced by
\(\mathcal E_{\max}\).

Because a sigmoid gate is strictly positive for finite logits, this
construction performs soft reweighting rather than exact edge deletion inside
\(\mathcal E_{\max}\). Exact adaptive sparsification would require an
additional hard or sparsity-inducing gating mechanism. If
\(a_\eta\) is further allowed to depend on a context \(c\) or on the current
representation, the resulting object should be interpreted as a
context-dependent descriptor \(\mathfrak S(c)\), with the pointwise versus
uniform distinction described in
Remark~\ref{rem:prelim-static-vs-state-dependent-attributes}.

The hard envelope is essential for the exact structural statement: without
\(\mathbf 1_{\mathcal E_{\max}}\), a generic sigmoid gate is dense and no
nontrivial fixed support or reachability guarantee follows.
\end{remark}

We now return to the standing fixed-descriptor regime and record the
equivalent kernel-level realization of its channel dictionary.

\subsection{Kernel-level formulation and matrix realization}
\label{app:prelim-entrywise-form}

For completeness, the descriptor also admits an equivalent kernel-level
formulation. Define
$
\widetilde{\phi}_r(z_i,z_j)
:=
\mathbf 1_{\mathcal E}(z_i,z_j)\phi_r(z_i,z_j),
$
and let
\[
C_{\mathfrak S}^{\mathrm{ker}}
:=
\left\{
\theta\in\mathbb R^R
\;\middle|\;
\theta_r\ge0 \text{ for } r\in\mathcal R_+
\right\}.
\]
The associated kernel family is
\[
\mathcal K_{\mathfrak S}
:=
\left\{
K_\theta:
K_\theta(z_i,z_j)
=
\sum_{r=1}^R\theta_r\widetilde{\phi}_r(z_i,z_j),
\quad
\theta\in C_{\mathfrak S}^{\mathrm{ker}}
\right\}.
\]

\begin{proposition}[Kernel-level structure and matrix realization]
\label{prop:prelim-matrix-realization-of-kernels}
The family \(\mathcal K_{\mathfrak S}\) is a finite-dimensional closed convex
polyhedral cone. Under the identification
\([K]_{ij}=K(z_i,z_j)\),
$
[K_\theta]
=
\sum_{r=1}^R\theta_r\Phi_r.
$
If the masked channels
\(\{\widetilde{\phi}_r\}_{r=1}^R\) are linearly independent, then the map
\(\theta\mapsto K_\theta\) is injective, and hence the coefficient
representation of every \(K\in\mathcal K_{\mathfrak S}\) is unique.
\end{proposition}

This is a kernel-level identifiability statement: it concerns the coefficients
\(\theta\) before an independent diagonal self-action is introduced.
It should therefore not be confused with identifiability in the interaction
family
\(\mathcal A_{\mathfrak S}\), where the additional term
\(\operatorname{Diag}(\delta)\) may overlap with diagonal components already
contained in the channel matrices. A sufficient condition for uniqueness at
that operator level is the stronger linear-independence condition on
\[
\{\Phi_1,\dots,\Phi_R,E_{11},\dots,E_{dd}\},
\]
stated in Proposition~\ref{prop:operators-operator-identifiability}.

\begin{remark}[Entrywise kernel realization]
\label{rem:prelim-entrywise-form}
Under the receiver--source convention of Section~\ref{sec:prelim}, the matrix
realization of \(K_\theta\in\mathcal K_{\mathfrak S}\) is simply
\[
[K_\theta]_{ij}
=
\mathbf 1_{\mathcal E}(z_i,z_j)
\sum_{r=1}^R
\theta_r\phi_r(z_i,z_j).
\]
Thus every off-support entry vanishes exactly, while all admissible entries are
coupled through the shared coefficient vector \(\theta\). The corresponding
operator family in Section~\ref{sec:operators} augments this kernel-generated
matrix with an independent diagonal self-action.
\end{remark}

\section{Supplementary Material for Section~\ref{sec:operators}}
\label{app:operators-supplement}

This appendix develops the operator-level material deferred from
Section~\ref{sec:operators}. Unless stated otherwise, we retain the fixed
scalar single-scale descriptor \(\mathfrak S\), the receiver--source matrix
convention of Section~\ref{sec:prelim}, and the hierarchy
\[
\mathcal A_{\mathfrak S}
\to
\mathcal L_{\mathfrak S}
\to
\mathcal P_{\mathfrak S}.
\]
We first make the matrix realization and its coefficient structure explicit,
then record support and semigroup facts, next address generator feasibility,
normalization, and exponential aliasing, and finally lift the construction to
vector-valued and multiscale representations. These results supplement the
minimal main-text chain without redefining its basic objects.

\subsection{Matrix-level remarks and explicit forms}
\label{app:operators-matrix-remarks}

\begin{remark}[No new structure is introduced at the matrix level]
\label{rem:operators-no-new-structure}
The matrix viewpoint adds no new geometric degrees of freedom beyond the
masked-channel construction induced by the descriptor in
Section~\ref{sec:prelim} and its kernel-level realization in
Appendix~\ref{app:prelim-entrywise-form}.
It only realizes those admissible pairwise interaction laws as linear maps on
signals \(h\in\mathbb R^d\).
Thus all operator constraints below are imposed on matrices generated by the
fixed descriptor rather than on arbitrary matrices detached from latent
geometry.
\end{remark}

With this scope fixed, the matrix entries can be written directly in terms of
the masked geometry channels and the free diagonal self-action.

\begin{remark}[Entrywise form of an interaction matrix]
\label{rem:operators-entrywise-form}
Every \(A\in\mathcal A_{\mathfrak S}\) has the explicit form
\[
A_{ij}
=
\begin{cases}
\displaystyle \sum_{r=1}^R \theta_r\,\mathbf 1_{\mathcal E}(z_i,z_j)\phi_r(z_i,z_j),
& i\neq j,
\\[0.9em]
\displaystyle \delta_i+\sum_{r=1}^R \theta_r\,\mathbf 1_{\mathcal E}(z_i,z_i)\phi_r(z_i,z_i),
& i=j.
\end{cases}
\]
Equivalently,
\[
\resizebox{\linewidth}{!}{$
A=
\begin{pmatrix}
\delta_1+\sum_r \theta_r \mathbf 1_{\mathcal E}(z_1,z_1)\phi_r(z_1,z_1)
&
\sum_r \theta_r \mathbf 1_{\mathcal E}(z_1,z_2)\phi_r(z_1,z_2)
&
\cdots
&
\sum_r \theta_r \mathbf 1_{\mathcal E}(z_1,z_d)\phi_r(z_1,z_d)
\\[0.6em]
\sum_r \theta_r \mathbf 1_{\mathcal E}(z_2,z_1)\phi_r(z_2,z_1)
&
\delta_2+\sum_r \theta_r \mathbf 1_{\mathcal E}(z_2,z_2)\phi_r(z_2,z_2)
&
\cdots
&
\sum_r \theta_r \mathbf 1_{\mathcal E}(z_2,z_d)\phi_r(z_2,z_d)
\\
\vdots & \vdots & \ddots & \vdots
\\
\sum_r \theta_r \mathbf 1_{\mathcal E}(z_d,z_1)\phi_r(z_d,z_1)
&
\sum_r \theta_r \mathbf 1_{\mathcal E}(z_d,z_2)\phi_r(z_d,z_2)
&
\cdots
&
\delta_d+\sum_r \theta_r \mathbf 1_{\mathcal E}(z_d,z_d)\phi_r(z_d,z_d)
\end{pmatrix}
$}
\]
This explicit form makes clear which entries are forced to vanish by admissibility and which entries remain learnable through channel coefficients and diagonal self-action.
\end{remark}

\begin{remark}[Operator-level realizations of classical layers]
\label{rem:operators-classical-layer-realizations}
The interaction family contains the site-level linear skeleton of several
familiar architectures:
\begin{enumerate}
    \item \emph{Residual convolution.}
    If \(X\) is a grid, \(\mathcal E\) is a finite stencil, and the geometry
    channels are indexed by relative offsets, then \(I+A\) with
    \(A\in\mathcal A_{\mathfrak S}\) gives a geometry-aware residual
    convolutional coupling. Translation sharing is obtained when the channels
    depend only on relative offset.

    \item \emph{Graph message passing.}
    If \(\mathcal E\) is graph adjacency and the channels encode edge type,
    direction, or metric relation, then \(A\in\mathcal A_{\mathfrak S}\)
    defines a graph-structured message-passing operator.

    \item \emph{Attention-style interaction.}
    With global or windowed admissibility, fixed pairwise score channels can be
    represented inside \(\mathcal A_{\mathfrak S}\). State-dependent pair
    scores instead belong to the context-dependent descriptor extension
    discussed in Appendix~\ref{app:prelim-pairwise-attributes}. Standard
    self-attention additionally applies a row-normalization such as softmax, so
    the present family should be interpreted as its geometry-aware linear score
    or mixing skeleton rather than as an exact identification with every
    attention operator.

    \item \emph{Additive multiscale fusion.}
    After the multiscale lifting of Appendix~\ref{app:operators-multiscale},
    sums of transferred operators provide the linear geometric skeleton of
    additive cross-resolution fusion used in pyramid and encoder--decoder
    architectures.
\end{enumerate}
These examples clarify architectural scope; they introduce no additional
operator theory.
\end{remark}

\subsection{Deferred algebraic facts from Section~\ref{sec:operators}}
\label{app:operators-algebraic-facts}

This subsection records algebraic and support properties that are useful for
later arguments but not needed in the shortest main-text chain
\[
\mathcal A_{\mathfrak S}
\to
\mathcal L_{\mathfrak S}
\to
\mathcal P_{\mathfrak S}.
\]
We first describe the interaction cone and its coordinate identifiability,
then pass to the generator cone, whose row-sum constraint determines the
diagonal and whose powers can propagate support only along admissible paths.
We close with the fixed-generator semigroup law.

\begin{proposition}[Basic structure of the interaction family]
\label{prop:operators-interaction-family-structure}
The set \(\mathcal A_{\mathfrak S}\) is a finite-dimensional closed polyhedral convex cone in \(\mathbb R^{d\times d}\).
If \(\mathcal R_+=\varnothing\), then \(\mathcal A_{\mathfrak S}\) is a linear subspace.
More generally, it is generated by the finite collection
$
\{\Phi_r: r\in\mathcal R_+\}
\cup
\{\pm \Phi_r: r\notin\mathcal R_+\}
\cup
\{\pm E_{11},\dots,\pm E_{dd}\},
$
where \(E_{ii}\) denotes the standard diagonal basis matrix.
\end{proposition}

The cone representation above need not provide unique coefficients. The next
statement gives a sufficient condition for uniqueness.

\begin{proposition}[Coefficient identifiability at the operator level]
\label{prop:operators-operator-identifiability}
Assume that the family of matrices
$
\{\Phi_1,\dots,\Phi_R,E_{11},\dots,E_{dd}\}
$
is linearly independent in \(\mathbb R^{d\times d}\).
Then every operator
$
A\in\mathcal A_{\mathfrak S}
$
admits a unique representation
$
A=\sum_{r=1}^R \theta_r \Phi_r + \operatorname{Diag}(\delta),
$
subject to the sign constraints \(\theta_r\ge 0\) for \(r\in\mathcal R_+\).
\end{proposition}

The defining row-sum constraint immediately determines the generator diagonal.

\begin{remark}[Diagonal overlap between channels and free self-action]
\label{rem:operators-diagonal-overlap}
The independence hypothesis in
Proposition~\ref{prop:operators-operator-identifiability} is substantive.
Indeed, both the channel matrices \(\Phi_r\) and
\(\operatorname{Diag}(\delta)\) may contribute to diagonal entries. If one
wants identifiability to be built into the parameterization more directly, one
may either impose
$
\phi_r(z_i,z_i)=0
\qquad
\text{for all }i,r,
$
so that all diagonal self-action is carried by \(\delta\), or absorb each
channel-generated diagonal contribution into the free diagonal term and work
with off-diagonal channel realizations. The paper retains the more general
formulation, so uniqueness should always be understood as conditional on the
stated linear-independence assumption.
\end{remark}

Having fixed the interaction-level algebra, we now restrict to the generator
subclass used by finite propagation.

\begin{proposition}[Basic structure of the generator family]
\label{prop:operators-generator-family-structure}
The set \(\mathcal L_{\mathfrak S}\) is a closed polyhedral convex cone in \(\mathbb R^{d\times d}\).
Equivalently,
$
\mathcal L_{\mathfrak S}
=
\mathcal A_{\mathfrak S}
\cap
\Bigl\{
L\in\mathbb R^{d\times d}
\;\Big|\;
L_{ij}\ge 0 \text{ for } i\neq j,\ L\mathbf 1=0
\Bigr\}.
$
\end{proposition}

\begin{proposition}[Diagonal structure of generators]
\label{prop:operators-generator-diagonal-structure}
Every \(L\in\mathcal L_{\mathfrak S}\) has nonpositive diagonal entries, and for each \(i\),
$
L_{ii}=-\sum_{j\ne i}L_{ij}.
$
In particular, the diagonal of a generator is completely determined by its off-diagonal entries.
\end{proposition}

Membership in \(\mathcal A_{\mathfrak S}\) simultaneously preserves the
one-step admissibility pattern.

\begin{proposition}[Support of generator off-diagonal entries]
\label{prop:operators-generator-support}
If \(L\in\mathcal L_{\mathfrak S}\), then for all \(i\neq j\),
$
L_{ij}\neq 0
\Longrightarrow
(z_i,z_j)\in\mathcal E.
$
Thus, even before exponentiation, a generator already respects the one-step admissibility pattern of the underlying geometry descriptor.
\end{proposition}

Repeated multiplication can enlarge support only by concatenating such
admissible one-step moves.

\begin{proposition}[Pathwise support of powers]
\label{prop:operators-pathwise-support-of-powers}
Let \(L\in\mathcal L_{\mathfrak S}\), and let
$
\bar{\mathcal E}:=\mathcal E\cup \Delta_X,
\qquad
\Delta_X:=\{(z_i,z_i):1\le i\le d\}.
$
Then for every \(n\ge 1\), the support of \(L^n\) is contained in
$
\bar{\mathcal E}^{\circ n}.
$
In particular, if \(i\neq j\) and there is no directed admissible path from \(z_j\) to \(z_i\), that is,
$
z_j \not\rightsquigarrow z_i,
$
then
$
(L^n)_{ij}=0
\qquad
\text{for every } n\ge 1.
$
\end{proposition}

The exponential aggregates these admissible powers. For a fixed generator it
also has the standard semigroup structure.

\begin{proposition}[Semigroup property]
\label{prop:operators-semigroup-property}
Let \(L\in\mathcal L_{\mathfrak S}\).
Then the family
$
\{e^{\tau L}\}_{\tau\ge 0}
$
forms a one-parameter semigroup:
$
e^{0\cdot L}=I,
\qquad
e^{(\tau_1+\tau_2)L}=e^{\tau_1 L}e^{\tau_2 L}
\quad\text{for all }\tau_1,\tau_2\ge 0.
$
\end{proposition}

The preceding semigroup identity concerns repeated propagation with a fixed
generator. Composition across distinct generators is treated in
Appendix~\ref{app:properties-composition-remarks}; in particular, the full
set-valued propagation family is generally not closed under multiplication.
For residual interaction operators, however, the corresponding pathwise
support statement is immediate and useful on its own.

\begin{proposition}[Pathwise support of repeated residual interactions]
\label{prop:operators-depth-stable-geometry}
If \(A^{(1)},\dots,A^{(m)}\in\mathcal A_{\mathfrak S}\), then
\[
\operatorname{supp}\!\left(
(I+A^{(m)})\cdots(I+A^{(1)})
\right)
\subseteq
\bar{\mathcal E}^{\circ m},
\qquad
\bar{\mathcal E}:=\mathcal E\cup\Delta_X.
\]
Thus depth enlarges the receptive field only through concatenations of
admissible one-step interactions.
\end{proposition}

\subsection{Feasibility, constrained compilation, and exponential identifiability}
\label{app:operators-normalization-compilation}

The previous subsection describes intrinsic algebraic structure. We now ask
three constructive questions: when the generator cone is nontrivial, how a
linearly constrained generator family can be compiled into a finite normalized
dictionary, and what normalization does---and does not---identify after matrix
exponentiation.

\begin{proposition}[A sufficient condition for nontrivial generator feasibility]
\label{prop:operators-nontrivial-generator-feasibility}
If some \(r_0\in\mathcal R_+\) satisfies
\((\Phi_{r_0})_{ij}>0\) for at least one \(i\neq j\), then
\[
L^{(r_0)}
:=
\Phi_{r_0}
-
\operatorname{Diag}(\Phi_{r_0}\mathbf1)
\in
\mathcal L_{\mathfrak S}\setminus\{0\}.
\]
\end{proposition}

\begin{proposition}[Finite compilation of feasible generators]
\label{prop:operators-feasible-compilation}
For fixed \(U,V\), the set
\(\mathcal L_{\mathfrak S;U,V}\) is a closed polyhedral convex cone.
Its normalized slice
\[
\mathcal L_{\mathfrak S;U,V}^{(1)}
:=
\{L\in\mathcal L_{\mathfrak S;U,V}:\rho(L)=1\}
\]
is nonempty if and only if
\(\mathcal L_{\mathfrak S;U,V}\neq\{0\}\), and whenever nonempty it is a compact polytope.
Hence there exist vertices \(Z_1,\dots,Z_N\) such that
$
\mathcal L_{\mathfrak S;U,V}^{(1)}
=
\operatorname{conv}\{Z_1,\dots,Z_N\}.
$
Consequently, every nonzero
\(Q\in\mathcal L_{\mathfrak S;U,V}\) can be written as
$
Q
=
\rho(Q)\sum_{s=1}^N\alpha_s Z_s,
\qquad
\alpha\in\Delta_N,
$
where
\(\Delta_N:=\{\alpha\in\mathbb R_+^N:\sum_s\alpha_s=1\}\).
The scale \(\rho(Q)\) and normalized matrix \(Q/\rho(Q)\) are unique,
whereas the mixture coefficients need not be unique.
\end{proposition}

Consequently, nontrivial feasibility of the constrained cone can be checked by
its linear constraints together with \(\rho(L)=1\). Enumerating all vertices
recovers the exact normalized polytope, whereas any smaller feasible dictionary
defines a certified subfamily. A finite-logit softmax reaches only the relative
interior of the simplex; exact boundary selection requires a map into the
closed simplex, such as Euclidean projection. In either case, the mixture
coefficients are identifiable only under additional affine-independence
assumptions on the chosen dictionary.

\begin{remark}[Generator normalization does not remove exponential aliasing]
\label{rem:operators-exponential-aliasing}
The unique scale--shape decomposition of a known generator does not imply identifiability from its exponential.
Let
\[
C=
\begin{pmatrix}
0&1&0\\
0&0&1\\
1&0&0
\end{pmatrix},
\qquad
L_1=C-I,
\qquad
L_2=C^\top-I.
\]
Then \(\rho(L_1)=\rho(L_2)=1\), yet for
\(\tau=2\pi/\sqrt3\),
\(r=e^{-\sqrt3\pi}\), and
\(J=\mathbf1\mathbf1^\top/3\),
\[
e^{\tau L_1}
=
J-r(I-J)
=
e^{\tau L_2},
\qquad
L_1\neq L_2.
\]
For any exactly observed \(P=e^{\tau L}\) with normalized \(L\),
\(\det(P)=e^{-d\tau}\), so
\(\tau=-d^{-1}\log\det(P)\);
the generator shape may nevertheless be nonunique.
If \(L\) is additionally symmetric and \(\tau>0\), then \(P\) is positive definite and
\(L=\tau^{-1}\log(P)\) is uniquely recovered within the symmetric class.
\end{remark}

\subsection{Block-valued lifting}
\label{app:operators-block-lifting}

In realistic deep models, each latent site typically carries a feature vector rather than a scalar.
The scalar construction must therefore be lifted to vector-valued signals.

Let each site carry a \(c\)-dimensional feature vector, so that the total
representation space becomes
$
\mathbb R^d\otimes\mathbb R^c \cong \mathbb R^{dc}.
$
Fix \(M\) feature-mixing matrices
$
G_1,\dots,G_M\in\mathbb R^{c\times c},
$
and an index set
$
\Lambda_+\subseteq \{1,\dots,R\}\times\{1,\dots,M\}
$
whose associated coefficients are constrained to be nonnegative. For compact
notation, write
$
\mathcal G
:=
\bigl(G_1,\dots,G_M,\Lambda_+\bigr)
$
for this fixed block-mixing specification.

\begin{definition}[Block-valued interaction family]
\label{def:operators-block-valued-interaction-family}
Let \(E_{ii}\in\mathbb R^{d\times d}\) denote the \(i\)-th
diagonal matrix unit. The block-valued interaction family is
\[
\mathcal A^{\mathrm{blk}}_{\mathfrak S,\mathcal G}
:=
\left\{
\sum_{r=1}^R\sum_{s=1}^M
\beta_{rs}(\Phi_r\otimes G_s)
+
\sum_{i=1}^d E_{ii}\otimes D_i
\;\middle|\;
\begin{array}{l}
\beta_{rs}\in\mathbb R,\\
\beta_{rs}\ge 0
\text{ for }(r,s)\in\Lambda_+,\\
D_i\in\mathbb R^{c\times c}
\end{array}
\right\}.
\]
\end{definition}

The matrices \(\Phi_r\) specify cross-site geometric coupling,
whereas \(G_s\) specify feature mixing during each transfer.
The blocks \(D_i\) describe sitewise self-action.
Shared self-action is the additional restriction
\(D_1=\cdots=D_d\), rather than part of the unrestricted lift.

\begin{remark}[Common special cases]
\label{rem:operators-block-common-special-cases}
The scalar family is recovered when \(c=1\), \(M=1\), \(G_1=1\),
\(\Lambda_+=\mathcal R_+\times\{1\}\), and \(D_i=\delta_i\).

More generally, choose \(G_1=I_c\),
\(\Lambda_+=\mathcal R_+\times\{1\}\), and
\(D_i=\delta_i I_c\). Then
\[
\sum_r\theta_r(\Phi_r\otimes I_c)
+
\sum_i E_{ii}\otimes(\delta_i I_c)
=
\left(
\sum_r\theta_r\Phi_r+\operatorname{Diag}(\delta)
\right)\otimes I_c.
\]
Thus every scalar interaction \(A\in\mathcal A_{\mathfrak S}\)
has the separable lift \(A\otimes I_c\).

General \(G_s\) and \(D_i\) additionally permit feature mixing.
Any desired site-sharing or equivariance restrictions must be
imposed explicitly on these blocks and their selectors.
\end{remark}

\begin{proposition}[Basic structure of the block-valued interaction family]
\label{prop:operators-block-valued-interaction-structure}
The set \(\mathcal A^{\mathrm{blk}}_{\mathfrak S,\mathcal G}\) is a finite-dimensional closed polyhedral convex cone in \(\mathbb R^{dc\times dc}\).
If \(\Lambda_+=\varnothing\), then it is a linear subspace.
\end{proposition}

At this point one must distinguish two regimes.
If the feature coordinates are generic real-valued hidden features, then \(\mathcal A^{\mathrm{blk}}_{\mathfrak S,\mathcal G}\) is the natural object and no positivity-preserving interpretation should be imposed by default.
If, however, the feature coordinates themselves represent ordered quantities and the standard positive cone on \(\mathbb R^{dc}\) is semantically meaningful, then one may also define a block-valued generator class.

\begin{definition}[Block-valued generator family]
\label{def:operators-block-valued-generator-family}
The \emph{block-valued generator family} is
\[
\mathcal L^{\mathrm{blk}}_{\mathfrak S,\mathcal G}
:=
\Bigl\{
L\in \mathcal A^{\mathrm{blk}}_{\mathfrak S,\mathcal G}
\;\Big|\;
L_{\alpha\beta}\ge 0 \text{ for all } \alpha\neq \beta,
\quad
L\mathbf 1_{dc}=0
\Bigr\},
\]
where \(\mathbf 1_{dc}\) denotes the all-ones vector in \(\mathbb R^{dc}\), and the indices \(\alpha,\beta\) range over the flattened coordinate set \(\{1,\dots,dc\}\).
\end{definition}

\begin{remark}[Block lifting and positivity]
\label{rem:operators-block-lifting-positivity}
Once block lifting has been performed, positivity and conservation are conditions on the full lifted matrix, not merely on its \(d\times d\) site-level skeleton.
Accordingly, Definition~\ref{def:operators-block-valued-generator-family} is appropriate only when the ordered positive cone on the feature space is semantically meaningful.
For arbitrary sign-indefinite hidden features, the block-valued interaction family is usually the relevant object, not the block-valued generator family.
\end{remark}

\begin{remark}[Block lifting and classical architectures]
\label{rem:operators-block-lifting-classical-architectures}
Block lifting is the point at which the framework reaches ordinary multi-channel deep representations.
For example, a width-preserving multi-channel convolution can be represented
by a local site geometry together with offset-dependent feature-mixing blocks.
Likewise, graph neural layers with learnable feature transforms are naturally
represented by graph-structured \(\Phi_r\) tensored with channel-mixing
blocks \(G_s\). Layers that change feature width require the corresponding
rectangular input--output block extension rather than the square lift used
here.
\end{remark}

\subsection{Multiscale variants}
\label{app:operators-multiscale}

The block lift changes the feature space carried by each site, whereas the
multiscale lift changes the resolution of the site space itself; these two
extensions are logically independent. Many deep representations exhibit
coarse-to-fine structure, nested neighborhoods, or hierarchical aggregation,
which motivates the following transferred-operator construction.

Assume that, in addition to the finest latent space \(X^{(0)}=X\), we are given coarser latent spaces
$
X^{(1)},\dots,X^{(L)},
$
with \(d_\ell=|X^{(\ell)}|\), and fixed transfer operators
$
P_\ell:\mathbb R^{d_\ell}\to\mathbb R^{d_0},
\qquad
R_\ell:\mathbb R^{d_0}\to\mathbb R^{d_\ell},
\qquad
\ell=0,\dots,L,
$
where \(P_0=R_0=I_{d_0}\).
Each scale \(X^{(\ell)}\) carries its own geometry descriptor \(\mathfrak S^{(\ell)}\), and hence its own interaction family \(\mathcal A_{\mathfrak S^{(\ell)}}\) and generator family \(\mathcal L_{\mathfrak S^{(\ell)}}\).

\begin{definition}[Multiscale interaction family]
\label{def:operators-multiscale-interaction-family}
The \emph{multiscale interaction family} is
\[
\mathcal A^{\mathrm{ms}}
:=
\left\{
\sum_{\ell=0}^L P_\ell A_\ell R_\ell
\;\middle|\;
A_\ell\in\mathcal A_{\mathfrak S^{(\ell)}}
\right\}.
\]
\end{definition}

This construction means that each scale contributes its own geometry-aware interaction, but that interaction is first computed at the appropriate resolution and then transferred back to the finest space.
The total operator is the sum of these scale-specific contributions.

\begin{definition}[Transferred multiscale generator family]
\label{def:operators-transferred-multiscale-generator-family}
The \emph{transferred multiscale generator family} is
\[
\widehat{\mathcal L}^{\mathrm{ms}}
:=
\left\{
\sum_{\ell=0}^L P_\ell L_\ell R_\ell
\;\middle|\;
L_\ell\in\mathcal L_{\mathfrak S^{(\ell)}}
\right\}.
\]
\end{definition}

The family \(\widehat{\mathcal L}^{\mathrm{ms}}\) is the correct algebraic object to define first.
However, even if every \(L_\ell\) is a generator on its own scale, the transferred sum need not automatically be a generator on the finest space, because transfer operators may alter off-diagonal sign structure.

\begin{proposition}[Constant compatibility of transferred generators]
\label{prop:operators-constant-compatibility-transferred-generators}
Assume that
$
R_\ell \mathbf 1_{d_0}=\mathbf 1_{d_\ell}
\qquad
\text{for every } \ell=0,\dots,L.
$
Then every
$
L\in \widehat{\mathcal L}^{\mathrm{ms}}
$
satisfies
$
L\mathbf 1_{d_0}=0.
$
\end{proposition}

Proposition~\ref{prop:operators-constant-compatibility-transferred-generators} shows that constant preservation transfers cleanly under a natural compatibility condition on the restriction operators.
What may still fail after transfer is the off-diagonal nonnegativity required of a genuine finest-space generator.

\begin{definition}[Actual multiscale generator family]
\label{def:operators-actual-multiscale-generator-family}
The \emph{actual multiscale generator family} is
\[
\mathcal L^{\mathrm{ms}}
:=
\Bigl\{
L\in \widehat{\mathcal L}^{\mathrm{ms}}
\;\Big|\;
L_{ij}\ge 0 \text{ for all } i\neq j,
\quad
L\mathbf 1_{d_0}=0
\Bigr\}.
\]
\end{definition}

The actual generator family is therefore the transferred cone after enforcing
the same Metzler and constant-preservation constraints used in the
single-scale theory. Its basic convex structure is inherited from the
scale-wise cones and the fixed transfer maps.

\begin{proposition}[Basic structure of multiscale families]
\label{prop:operators-basic-structure-multiscale-families}
Assume that the transfer operators \(P_\ell\) and \(R_\ell\) are fixed.
Then the sets
$
\mathcal A^{\mathrm{ms}},
\qquad
\widehat{\mathcal L}^{\mathrm{ms}},
\qquad
\mathcal L^{\mathrm{ms}}
$
are finite-dimensional closed polyhedral convex cones in \(\mathbb R^{d_0\times d_0}\).
More precisely, \(\mathcal A^{\mathrm{ms}}\) and \(\widehat{\mathcal L}^{\mathrm{ms}}\) are linear images of finite-dimensional polyhedral cones, while \(\mathcal L^{\mathrm{ms}}\) is obtained from \(\widehat{\mathcal L}^{\mathrm{ms}}\) by imposing additional linear equations and coordinatewise inequalities.
\end{proposition}

For every nonzero \(L\in\mathcal L^{\mathrm{ms}}\), off-diagonal
nonnegativity and \(L\mathbf 1_{d_0}=0\) imply
\(\rho_{\mathrm{ms}}(L)>0\). Hence whenever
\(\mathcal L^{\mathrm{ms}}\neq\{0\}\), every nonzero multiscale generator
has a unique rate--shape decomposition and the normalized slice below is
nonempty.

\begin{definition}[Multiscale propagation family]
\label{def:operators-multiscale-propagation-family}
For \(L\in\mathcal L^{\mathrm{ms}}\), define
\[
\rho_{\mathrm{ms}}(L)
:=
\frac{1}{d_0}
\sum_{i=1}^{d_0}\sum_{j\neq i}L_{ij}.
\]
If \(\mathcal L^{\mathrm{ms}}\neq\{0\}\), define
$
\mathcal L_{\mathrm{ms}}^{(1)}
:=
\left\{
L\in\mathcal L^{\mathrm{ms}}
\;\middle|\;
\rho_{\mathrm{ms}}(L)=1
\right\}.
$
Set
$
\mathcal P^{\mathrm{ms}}(0):=\{I\},
$
and, for \(\tau>0\),
$
\mathcal P^{\mathrm{ms}}(\tau)
:=
\left\{
e^{\tau L}
\;\middle|\;
L\in\mathcal L_{\mathrm{ms}}^{(1)}
\right\}.
$
Finally,
$
\mathcal P^{\mathrm{ms}}
:=
\bigcup_{\tau\ge0}\mathcal P^{\mathrm{ms}}(\tau).
$
\end{definition}
This is the direct multiscale analogue of
Definition~\ref{def:operators-propagation-family}: the normalized generator
specifies propagation shape on the finest realized space, while \(\tau\)
controls its overall magnitude.

\begin{remark}[Effective geometry in the multiscale case]
\label{rem:operators-effective-geometry-multiscale}
In the single-scale setting, propagation support is controlled directly by reachability in the admissibility graph \((X,\mathcal E)\).
In the multiscale setting, the effective finest-space support is determined not only by the scale-wise admissibility relations but also by the transfer operators \(P_\ell\) and \(R_\ell\).
For that reason, the relevant support analysis in the multiscale case must be carried out at the level of the realized finest-space operator \(L\in\mathcal L^{\mathrm{ms}}\), rather than by reading off reachability from any one scale in isolation.
\end{remark}

\begin{remark}[Multiscale operators as geometric realizations of pyramid architectures]
\label{rem:operators-multiscale-pyramid-architectures}
The multiscale interaction family provides the operator-level template behind architectures that repeatedly exchange information across resolutions.
Fine-scale descriptors model local detail and short-range anisotropy.
Coarse-scale descriptors model long-range organization and large receptive fields.
Transfer operators move these effects across scales.
In this sense, the multiscale family is the natural geometry-induced abstraction of pyramid, encoder-decoder, and U-Net/FPN-style feature interaction.
\end{remark}

\section{Supplementary Material for Section~\ref{sec:properties}}
\label{app:properties-supplement}

This appendix supplements the structural results of
Section~\ref{sec:properties} in the same order as the main-text argument.
We first clarify the coefficient-geometry interpretation of the interaction
family, then record dynamic and spectral consequences of generator
propagation, give a uniformization-based interpretation of pathwise locality,
distinguish fixed-generator and commuting composition from the general
noncommuting case, and finally clarify the scope of the single-layer
ambient-space restriction.

Throughout, we use without restatement the operator-level facts established in
Appendix~\ref{app:operators-supplement}. In particular, the polyhedral
structure of \(\mathcal A_{\mathfrak S}\) and
\(\mathcal L_{\mathfrak S}\), coefficient identifiability, one-step and
pathwise support, and the fixed-generator semigroup law are given by
Propositions~\ref{prop:operators-interaction-family-structure},
\ref{prop:operators-generator-family-structure},
\ref{prop:operators-operator-identifiability},
\ref{prop:operators-one-step-support},
\ref{prop:operators-pathwise-support-of-powers}, and
\ref{prop:operators-semigroup-property}, respectively.
The results below therefore focus only on consequences and qualifications
specific to the structural properties developed in
Section~\ref{sec:properties}.

\subsection{Coefficient-geometry remarks}
\label{app:properties-identifiability-remarks}

The dimension statement in
Proposition~\ref{prop:properties-affine-hull-dimension} concerns the dimension
of the realized operator family, whereas uniqueness of its coefficient
representation is a separate algebraic issue.
The latter is governed by
Proposition~\ref{prop:operators-operator-identifiability}; in particular, the
possible overlap between channel-generated diagonal entries and the free
diagonal term is discussed in
Remark~\ref{rem:operators-diagonal-overlap}.
We therefore only record here the affine-geometric point needed to interpret
the main-text dimension statement.

\begin{remark}[Affine hull versus linear span]
\label{rem:properties-affine-hull-vs-span}
Because \(\mathcal A_{\mathfrak S}\) is a cone containing the origin,
its affine hull is a linear subspace and therefore coincides with its linear
span:
\[
\operatorname{aff}(\mathcal A_{\mathfrak S})
=
\operatorname{span}(\mathcal A_{\mathfrak S}).
\]
Hence the two dimension expressions used in
Proposition~\ref{prop:properties-affine-hull-dimension} are equivalent.
\end{remark}

\subsection{Dynamic and spectral remarks}
\label{app:properties-dynamic-remarks}

Theorem~\ref{thm:properties-order-supnorm-stability} establishes the basic
forward stability of \(e^{\tau L}\).
The following results clarify, in turn, its Markov orientation, sensitivity to
the generator and propagation magnitude, spectral location, behavior under
sequential symmetric propagation, and the distinction between forward
stability and backward conditioning.

\begin{remark}[Operator orientation and mass conservation]
\label{rem:properties-operator-orientation-conservation}
Under the column-signal convention
\[
h\longmapsto e^{\tau L}h,
\]
the condition \(L\mathbf1=0\) implies
$
e^{\tau L}\mathbf1=\mathbf1.
$
Thus \(e^{\tau L}\) preserves constant observables and is naturally
interpreted as a backward Markov operator.
For a column probability vector, the corresponding forward evolution is
$
p(\tau)=e^{\tau L^\top}p(0).
$
Consequently, applying \(e^{\tau L}\) itself to column mass vectors preserves
total mass only under the additional balance condition
$
\mathbf1^\top L=0.
$
\end{remark}

The forward stability theorem also controls first-order perturbations of the
propagator.

\begin{proposition}[Fr\'echet sensitivity of generator propagation]
\label{prop:properties-frechet-sensitivity}
For
\[
\|M\|_{\infty\to\infty}
:=
\max_i\sum_j |M_{ij}|,
\]
let \(L\in\mathcal L_{\mathfrak S}\), \(\tau\ge0\), and
\(H\in\mathbb R^{d\times d}\).
Then
\[
D_L e^{\tau L}[H]
=
\tau\int_0^1
e^{(1-s)\tau L}
H
e^{s\tau L}
\,ds,
\]
and
$
\|D_L e^{\tau L}[H]\|_{\infty\to\infty}
\le
\tau\|H\|_{\infty\to\infty}.
$
Moreover,
$
\partial_\tau e^{\tau L}
=
Le^{\tau L}
=
e^{\tau L}L,
\qquad
\|\partial_\tau e^{\tau L}\|_{\infty\to\infty}
\le
\|L\|_{\infty\to\infty}.
$
\end{proposition}

The same generator constraints also place the spectrum in the closed left
half-plane.

\begin{proposition}[Spectral stability of generators]
\label{prop:properties-spectral-stability}
For every \(L\in\mathcal L_{\mathfrak S}\),
$
\operatorname{Re}(\lambda)\le0
\qquad
\text{for every }\lambda\in\sigma(L),
$
and
$
0\in\sigma(L).
$
\end{proposition}

The preceding conclusions depend essentially on the generator constraints and
should not be transferred to arbitrary residual interaction operators.

\begin{remark}[Residual interactions versus stable propagation]
\label{rem:properties-interaction-vs-propagation}
Membership in \(\mathcal A_{\mathfrak S}\) alone does not imply the positivity,
constant-preservation, spectral, or contraction properties of
Theorem~\ref{thm:properties-order-supnorm-stability}.
Those conclusions are specific to
\(\mathcal L_{\mathfrak S}\) and to propagation operators obtained by
exponentiating its elements.
Thus the two operator families play different roles:
\(\mathcal A_{\mathfrak S}\) describes structured residual correction,
whereas \(\mathcal L_{\mathfrak S}\) describes admissible infinitesimal
transport.
\end{remark}

For symmetric generators, the single-generator oversmoothing estimate extends
multiplicatively to a sequence of possibly different generators.

\begin{proposition}[Sequential symmetric oversmoothing]
\label{prop:properties-sequential-oversmoothing}
Let
\[
L_k=L_k^\top\in\mathcal L_{\mathfrak S},
\qquad
\tau_k\ge0,
\qquad
k=1,\dots,m,
\]
and assume
$
\ker(L_k)=\operatorname{span}\{\mathbf1\}.
$
Define
$
\gamma_k
:=
\min\bigl\{
-\lambda:
\lambda\in\sigma(L_k)\setminus\{0\}
\bigr\}
>0,
\qquad
\overline h
:=
\frac{\mathbf1^\top h}{d}\mathbf1.
$
Then
\[
\left\|
e^{\tau_mL_m}\cdots e^{\tau_1L_1}
(h-\overline h)
\right\|_2
\le
\exp\!\left(
-\sum_{k=1}^m\tau_k\gamma_k
\right)
\|h-\overline h\|_2.
\]
\end{proposition}

Forward contraction should nevertheless be distinguished from conditioning of
the backward map.

\begin{remark}[Forward stability, invertibility, and trainability]
\label{rem:properties-forward-stability-trainability}
Suppose \(L=L^\top\) has eigenvalues
\[
-\lambda_1,\dots,-\lambda_d,
\qquad
0=\lambda_1\le\cdots\le\lambda_d.
\]
Then the singular-value multipliers of \(e^{\tau L}\) are
\(e^{-\tau\lambda_k}\), and therefore
\[
\kappa_2(e^{\tau L})
=
e^{\tau\lambda_d}
=
e^{\tau\lambda_{\max}(-L)}.
\]
Thus forward nonexpansiveness does not imply a well-conditioned inverse or
well-conditioned backward propagation.

At the opposite endpoint,
$
D_L e^{\tau L}[H]\big|_{\tau=0}=0,
$
so exact identity initialization produces zero first-order sensitivity with
respect to generator-shape parameters when the propagation magnitude is
initialized at \(\tau=0\).
Finally, the normalization \(\rho(L)=1\) removes the scalar
generator--magnitude ambiguity, but does not by itself bound local rates,
spectral conditioning, or derivatives of a context-dependent selector.
\end{remark}

\subsection{Locality and finite-propagation interpretation}
\label{app:properties-locality-remarks}

The reachability statement in
Theorem~\ref{thm:properties-reachability-preserving-propagation}
can be understood through uniformization: finite-time propagation is a
nonnegative mixture of discrete path lengths rather than an unrestricted dense
interaction.

\begin{proposition}[Uniformization identity for finite propagation]
\label{prop:properties-uniformization}
Let \(L\in\mathcal L_{\mathfrak S}\), \(\tau>0\), and choose
\(\nu>0\) such that
$
\nu\ge\max_i(-L_{ii}).
$
Define
$
T:=I+\frac{L}{\nu}.
$
Then
\[
e^{\tau L}
=
e^{-\nu\tau}
\sum_{k=0}^{\infty}
\frac{(\nu\tau)^k}{k!}T^k.
\]
Moreover,
$
T\ge0,
\qquad
T\mathbf1=\mathbf1.
$
If
$
\mathcal E_L
:=
\{(z_i,z_j):i\ne j,\ L_{ij}>0\},
$
then the off-diagonal support of \(T\) is contained in
\(\mathcal E_L\), while its diagonal entries represent holding at the current
site.
Hence the exponential is a Poisson mixture of admissible discrete path
lengths.
\end{proposition}

This identity makes explicit why exponentiation enlarges one-step support only
through path concatenation.

\begin{remark}[Why geometry persists across depth]
\label{rem:properties-why-maintain-geometry-across-depth}
Suppose that each layer selects a generator
$
L_k\in\mathcal L_{\mathfrak S}
$
from the same fixed descriptor \(\mathfrak S\).
Each \(L_k\) has off-diagonal support contained in the one-step admissibility
relation \(\mathcal E\), while products and exponentials can create new
couplings only by concatenating such admissible edges.
Consequently, increasing propagation time or composing multiple layers may
enlarge the effective receptive field, but only inside the reflexive
transitive closure of \(\mathcal E\).
Long-range coupling therefore remains geometrically organized rather than
becoming an arbitrary dense interaction.
\end{remark}

The same observation also clarifies the range of feature vectors obtainable
from a propagation operator.

\begin{remark}[Averaging is not unrestricted transport]
\label{rem:properties-propagation-range}
For each site \(i\), define
\[
S_i
:=
\{j:j=i\text{ or }z_j\rightsquigarrow z_i\}.
\]
Let
$
P=e^{\tau L},
\qquad
L\in\mathcal L_{\mathfrak S},
\qquad
H\in\mathbb R^{d\times c}.
$
By positivity, row-stochasticity, and reachability preservation,
\[
(PH)_i
\in
\operatorname{conv}\{H_j:j\in S_i\}.
\]
Thus finite propagation performs reachability-constrained convex aggregation.

Moreover,
\[
\det(e^{\tau L})
=
e^{\tau\operatorname{tr}(L)}
>
0.
\]
Hence the propagation family does not contain every row-stochastic matrix.
For example, the two-site swap matrix has determinant \(-1\) and therefore
cannot equal \(e^{\tau L}\) for any real generator \(L\).
\end{remark}

\subsection{Composition and depth-dependent algebra}
\label{app:properties-composition-remarks}

Proposition~\ref{prop:operators-semigroup-property} gives exact closure when a
single generator is reused.
For distinct generators, it is useful to separate the commuting case, where a
single-generator representation survives, from the general noncommuting case,
where the structural envelope of
Proposition~\ref{prop:properties-composable-envelope} remains valid even
though exact membership in \(\mathcal P_{\mathfrak S}\) may fail.

\begin{proposition}[Composition of commuting propagations]
\label{prop:properties-composition-of-commuting-propagations}
Let
\[
L_1,L_2\in\mathcal L_{\mathfrak S}^{(1)},
\qquad
\tau_1,\tau_2\ge0,
\]
and assume
\[
L_1L_2=L_2L_1.
\]
Then
\[
e^{\tau_1L_1}e^{\tau_2L_2}
=
e^{G},
\qquad
G:=\tau_1L_1+\tau_2L_2
\in\mathcal L_{\mathfrak S}.
\]
If \(\tau_1+\tau_2>0\), then
\[
\rho(G)
=
\tau_1+\tau_2,
\qquad
\widehat L
:=
\frac{G}{\tau_1+\tau_2}
\in
\mathcal L_{\mathfrak S}^{(1)},
\]
and therefore
\[
e^{\tau_1L_1}e^{\tau_2L_2}
=
e^{(\tau_1+\tau_2)\widehat L}
\in
\mathcal P_{\mathfrak S}.
\]
If \(\tau_1=\tau_2=0\), the product is \(I\in\mathcal P_{\mathfrak S}\).
\end{proposition}

Thus commuting propagations are closed after the same
shape--magnitude normalization used in
Definition~\ref{def:operators-propagation-family}.
Without commutativity, however, such a representation is not guaranteed.

\begin{proposition}[The propagation family need not be composition-closed]
\label{prop:properties-full-propagation-not-closed}
There exist geometry descriptors for which
\(\mathcal P_{\mathfrak S}\) is not closed under matrix multiplication.

For example, consider a three-site descriptor whose channel family contains
the off-diagonal indicator channels, so that
\(\mathcal L_{\mathfrak S}\) contains the full three-state generator cone.
Let
\[
L_1
=
\begin{pmatrix}
-1 & 1 & 0\\
0 & 0 & 0\\
0 & 0 & 0
\end{pmatrix},
\qquad
L_2
=
\begin{pmatrix}
0 & 0 & 0\\
0 & -2 & 2\\
0 & 0 & 0
\end{pmatrix},
\qquad
t:=\log 2.
\]
Both \(L_1\) and \(L_2\) belong to
\(\mathcal L_{\mathfrak S}\), and by the scale--shape decomposition
\[
P_1:=e^{tL_1},
\qquad
P_2:=e^{tL_2}
\]
belong to \(\mathcal P_{\mathfrak S}\).
Their product is
\[
P_2P_1
=
\begin{pmatrix}
\frac12 & \frac12 & 0\\
0 & \frac14 & \frac34\\
0 & 0 & 1
\end{pmatrix}.
\]
This matrix has the three distinct positive eigenvalues
\(\frac12,\frac14,1\), and hence its unique real logarithm is
\[
\log(P_2P_1)
=
\begin{pmatrix}
-\log 2 & \log 4 & -\log 2\\
0 & -\log 4 & \log 4\\
0 & 0 & 0
\end{pmatrix}.
\]
The \((1,3)\) entry is negative, so this logarithm is not Metzler and therefore
does not belong to any generator family
\(\mathcal L_{\mathfrak S}\).
Consequently,
$
P_2P_1\notin\mathcal P_{\mathfrak S}.
$
\end{proposition}

The failure of single-exponential closure should not be confused with the
algebraic expressivity obtainable from repeated channel composition.
We next make this distinction explicit.

\begin{proposition}[Single-layer compression need not persist algebraically]
\label{prop:properties-generated-algebra}
Define the effective channel relation
\[
\mathcal E_\Phi
:=
\left\{
(z_i,z_j):
i\ne j,\;
(\Phi_r)_{ij}\ne0
\text{ for some }r
\right\},
\]
and let \(\mathcal R_\Phi\) denote its reflexive transitive closure.
For \(B\in\mathbb R^{d\times d}\), define the site-indexed support
\[
\operatorname{supp}_X(B)
:=
\{(z_i,z_j):B_{ij}\ne0\}.
\]
Then
\[
\operatorname{Alg}_{\mathbb R}
\left(
I,\Phi_1,\ldots,\Phi_R,E_{11},\ldots,E_{dd}
\right)
=
\left\{
B\in\mathbb R^{d\times d}:
\operatorname{supp}_X(B)\subseteq\mathcal R_\Phi
\right\},
\]
where \(\operatorname{Alg}_{\mathbb R}\) denotes the smallest unital real
matrix algebra containing the displayed generators.

Indeed,
\[
E_{ii}\Phi_rE_{jj}
=
(\Phi_r)_{ij}E_{ij},
\]
so every effective one-step edge matrix unit can be isolated, while products
of matrix units generate exactly the pairs connected by admissible paths.
In particular, if the effective channel graph is strongly connected, then
\[
\operatorname{Alg}_{\mathbb R}
\left(
I,\Phi_1,\ldots,\Phi_R,E_{11},\ldots,E_{dd}
\right)
=
\mathbb R^{d\times d}.
\]
This is an algebraic arbitrary-depth statement and does not imply that every
matrix is representable by any fixed-depth architecture.
\end{proposition}

\subsection{Scope of the ambient-space restriction}
\label{app:properties-ambient-remarks}

The strict-inclusion result in
Theorem~\ref{thm:properties-strict-expressive-restriction}
is a statement about the space of realized single-layer matrices.
Two qualifications are useful: the correct ambient comparison is the masked
matrix space rather than the full matrix space, and the dimension of the
selected matrix family does not by itself control the functional capacity of
a context-dependent selector.

\begin{remark}[Why the correct ambient comparison is \(\mathcal M_{\mathcal E}\)]
\label{rem:properties-why-compare-with-masked-space}
The relevant comparison class for
\(\mathcal A_{\mathfrak S}\) is not the unrestricted space
\(\mathbb R^{d\times d}\), but
\[
\mathcal M_{\mathcal E}
=
\Bigl\{
A\in\mathbb R^{d\times d}
\;\Big|\;
A_{ij}=0
\text{ whenever }
i\ne j
\text{ and }
(z_i,z_j)\notin\mathcal E
\Bigr\}.
\]
Every geometry-induced interaction already obeys the same hard support mask.
The distinctive restriction of
\(\mathcal A_{\mathfrak S}\) is therefore not sparsity alone, but the further
compression from independent admissible-edge coefficients to a smaller set of
reusable geometry-channel coefficients.
\end{remark}

The preceding operator-space restriction should not be interpreted as a
capacity bound for unrestricted self-conditioned selectors.

\begin{proposition}[Operator dimension does not bound selector capacity]
\label{prop:properties-unrestricted-diagonal-selectors}
Let
$
K=[1,2]^d.
$
For every continuous map
$
f:K\to\mathbb R^d,
$
there exists a continuous selector
$
A:K\to\mathcal A_{\mathfrak S}
$
using only diagonal matrices such that
$
h+A(h)h=f(h)
\qquad
\text{for every }h\in K.
$
\end{proposition}

A concrete selector is
\[
A(h)
=
\operatorname{Diag}
\left(
\frac{f_1(h)}{h_1}-1,
\dots,
\frac{f_d(h)}{h_d}-1
\right).
\]
Because \(h_i\in[1,2]\), this selector is continuous; because arbitrary
diagonal self-action belongs to \(\mathcal A_{\mathfrak S}\), it is
admissible; and direct substitution gives
\[
h+A(h)h=f(h).
\]

\begin{remark}[Scope of the dimensional restriction]
The dimensional restriction of
\(\mathcal A_{\mathfrak S}\) concerns the family of matrices that can be
selected at a single input, not the statistical or functional complexity of a
self-conditioned map.
Functional expressivity, sample complexity, and generalization additionally
depend on the selector architecture, parameter sharing, regularity, depth, and
the permitted diagonal corrections.
Proposition~\ref{prop:properties-unrestricted-diagonal-selectors} concerns
unrestricted continuous selectors and does not assert universality for any
fixed finite-capacity implementation.
\end{remark}

Together with
Proposition~\ref{prop:properties-generated-algebra}, these observations
separate three distinct notions of expressivity: single-layer operator-space
dimension, arbitrary-depth algebraic closure, and the functional capacity of a
context-dependent selector.

\section{Supplementary Material for Section~\ref{sec:layers}}
\label{app:layers-supplement}

This appendix supplements Section~\ref{sec:layers} by separating four issues
that are deliberately compressed in the main text.
First, we clarify how geometry-conditioned selectors are implemented and
distinguish fixed-context operator guarantees from properties of fully
self-conditioned maps.
Second, we record the additional regularity and locality conditions required
when the selector depends on the representation being transformed.
Third, we develop two extensions of the basic layer construction: balanced
transport, which additionally preserves total mass, and multiscale correction,
which combines geometry-aware interactions across resolutions.
Finally, we summarize the resulting layer implementations algorithmically.
Throughout, the operator-level guarantees inherited from
Sections~\ref{sec:operators} and~\ref{sec:properties} are kept separate from
claims about the complete nonlinear input--output map.

\subsection{Implementation remarks for geometry-conditioned selectors}
\label{app:layers-implementation-remarks}

\begin{remark}[Selector inputs need not equal acted-on signals]
\label{rem:layers-selector-inputs}
For notational clarity, Section~\ref{sec:layers} separates the context variable \(c\in\mathcal C\) from the acted-on signal \(h\in\mathbb R^d\).
Conceptually, however, these roles may coincide.
The self-conditioned case is obtained by taking \(\mathcal C\subseteq\mathbb R^d\) and setting \(c=h\).
More generally, the selector may depend on the current hidden state, on a parallel context stream, on cross-attention features, or on any other side information available to the architecture.
The mathematical structure of the layer depends only on the fact that the realized operator lies in the prescribed geometry-induced family.
\end{remark}

\begin{remark}[Normalized generator selection and scale separation]
\label{rem:layers-generator-selector-implementation}
When propagation is written in the scale--shape form
$
e^{\tau(c)L(c)},
$
the selector should produce a normalized generator shape
$
L(c)\in\mathcal L_{\mathfrak S}^{(1)},
$
so that \(L(c)\) determines propagation geometry while \(\tau(c)\) determines
its overall magnitude.
This is the convention used in Section~\ref{sec:layers}.

There are several valid implementations.
One may directly parameterize coefficients subject to the defining
constraints of \(\mathcal L_{\mathfrak S}^{(1)}\), project a raw matrix onto
\(\mathcal L_{\mathfrak S}^{(1)}\), or form a convex combination of
preconstructed normalized generator basis elements
$
\widehat B_r\in\mathcal L_{\mathfrak S}^{(1)}.
$
The constructive selector introduced in
Section~\ref{sec:layers}, which forms \(L(c)\) as a convex combination of
the normalized generator basis elements \(\widehat B_r\), uses the third route.

By contrast, if a selector produces an unnormalized generator
$
Q(c)\in\mathcal L_{\mathfrak S},
$
then \(Q(c)\) should either be used directly through \(e^{Q(c)}\), or,
whenever \(Q(c)\neq0\), decomposed uniquely as
$
Q(c)
=
\rho(Q(c))\,\widehat L(c),
\qquad
\widehat L(c)\in\mathcal L_{\mathfrak S}^{(1)}.
$
The zero generator corresponds to the identity propagation.

Finally, predicting arbitrary edgewise nonnegative rates on the admissibility
mask and balancing only the diagonal is not sufficient to guarantee
membership in \(\mathcal L_{\mathfrak S}\).
Such a construction satisfies the ambient masked-generator constraints but
may lie outside the compressed geometry-channel family
\(\mathcal A_{\mathfrak S}\).
\end{remark}

The preceding parameterizations also preserve the basic well-posedness of
the layer maps.
Because matrix multiplication and matrix exponentiation are continuous,
continuous selectors \(A(\cdot)\), \(L(\cdot)\), and \(\tau(\cdot)\) induce
jointly continuous maps
$
(c,h)
\longmapsto
h+A(c)h
$
and
$
(c,h)
\longmapsto
e^{\tau(c)L(c)}h.
$
Thus no additional regularity issue arises at the fixed-context layer level.
The situation changes when the selector itself depends on the acted-on
representation, which is treated next.

\subsection{Self-conditioned layers: locality and conditioning}
\label{app:layers-self-conditioned}

The fixed-context guarantees of Section~\ref{sec:layers} concern realized
operators.
When the context is itself a function of the acted-on representation,
and in particular when \(c=h\), two additional questions arise.
First, selector dependence may introduce functional couplings that are not
visible from the support of the realized matrix alone.
Second, variation of the selected propagator may change the conditioning of
the complete nonlinear map.
We therefore treat residual functional locality first, propagation
conditioning second, and ancestor-local functional dependence third.
For propagation it is convenient to absorb scale and shape into the
integrated generator
\[
Q(h):=\tau(h)L(h).
\]

\begin{remark}[Matrix support versus functional dependence]
\label{rem:layers-self-conditioned-residual-locality}
For a differentiable self-conditioned adapter
\[
F(h)=h+A(h)h,
\]
\[
DF(h)[v]=(I+A(h))v+DA(h)[v]h.
\]
Hence support restrictions on \(A(h)\) do not imply the same sparsity for
the full Jacobian.

A sufficient condition for one-step functional locality is that the
\(i\)-th row of \(A(h)\) depend only on
\[
\{h_i\}\cup\{h_j:(z_i,z_j)\in\mathcal E\}.
\]
For example, in dimension two with no admissible off-diagonal edges,
\(A(h)=\operatorname{Diag}(h_2,0)\) is admissible, but
\(F_1(h)=(1+h_2)h_1\), so the output still depends on \(h_2\).
\end{remark}

\begin{proposition}[Conditioning of self-conditioned propagation]
\label{prop:layers-self-conditioned-stability}
Let \(K\subseteq\mathbb R^d\) satisfy
\(\sup_{h\in K}\|h\|_\infty\le M\).
Let
\[
F(h)=e^{Q(h)}h,
\]
where \(Q(h)\) is Metzler, \(Q(h)\mathbf1=0\), and
\[
\|Q(h)\|_{\infty\to\infty}\le b,
\qquad
\|Q(h)-Q(g)\|_{\infty\to\infty}
\le K_Q\|h-g\|_\infty.
\]
Then
\[
\bigl(e^{-b}-MK_Q\bigr)\|h-g\|_\infty
\le
\|F(h)-F(g)\|_\infty
\le
\bigl(1+MK_Q\bigr)\|h-g\|_\infty.
\]
Hence \(F\) is bi-Lipschitz on \(K\) whenever \(MK_Q<e^{-b}\).
\end{proposition}

The preceding proposition controls input--output conditioning, but it does
not determine which input coordinates may influence a given output
coordinate.
Functional locality requires a separate restriction on how the selector
depends on its input.
This motivates the following ancestor-local condition.

\begin{proposition}[Functional locality of adaptive propagation]
\label{prop:layers-functional-ancestor-locality}
For each \(i\), let
\[
S_i:=\{i\}\cup\{j:z_j\rightsquigarrow z_i\}.
\]
Suppose \(Q(h)\in\mathcal L_{\mathfrak S}\) and
\[
h_{S_i}=g_{S_i}
\quad\Longrightarrow\quad
Q(h)_{i,:}=Q(g)_{i,:}.
\]
Then, for \(F(h)=e^{Q(h)}h\),
\[
h_{S_i}=g_{S_i}
\quad\Longrightarrow\quad
F_i(h)=F_i(g).
\]
The same property is preserved under compositions satisfying the same
ancestor-local dependence condition.
\end{proposition}

To see why the rowwise condition is sufficient, note that
\(j\in S_i\) implies \(S_j\subseteq S_i\).
Hence \(h_{S_i}=g_{S_i}\) forces all generator rows that can participate in
a directed path ending at \(i\) to agree for \(h\) and \(g\).
Because every term in the \(i\)-th row of \(Q^k\) is assembled from such
ancestor rows, the same is true for the power series defining \(e^Q\).
Thus the \(i\)-th propagated output can depend only on coordinates in
\(S_i\).

The condition applies to the integrated generator
\(Q(h)=\tau(h)L(h)\), not merely to the support of \(L(h)\).
Hence a globally input-dependent scale or coefficient head need not satisfy
functional locality without additional parameter sharing.

\paragraph{A sufficient selector budget.}
The preceding abstract condition can be made explicit for the normalized
mixture parameterization used in Section~\ref{sec:layers}.
Let
\[
Z_1,\dots,Z_N\in\mathcal L_{\mathfrak S}^{(1)}
\]
be fixed normalized generators and define
\[
\Delta_N
:=
\left\{
\alpha\in\mathbb R_+^N:
\sum_{s=1}^N\alpha_s=1
\right\}.
\]
On the same bounded set \(K\), suppose
\[
Q(h)
=
\tau(h)\sum_{s=1}^N\alpha_s(h)Z_s,
\qquad
\alpha(h)\in\Delta_N,
\qquad
0\le\tau(h)\le\tau_{\max}.
\]
Assume that
\[
|\tau(h)-\tau(g)|
\le
K_\tau\|h-g\|_\infty
\]
and
\[
\|\alpha(h)-\alpha(g)\|_1
\le
K_\alpha\|h-g\|_\infty.
\]
Let
\[
\nu_*
:=
\max_{1\le s\le N}
\max_{1\le i\le d}
\bigl(-(Z_s)_{ii}\bigr).
\]
Since every \(Z_s\) is Metzler and satisfies \(Z_s\mathbf1=0\),
\[
\|Z_s\|_{\infty\to\infty}
=
2\max_i\bigl(-(Z_s)_{ii}\bigr)
\le
2\nu_*.
\]
Consequently,
\[
\|Q(h)\|_{\infty\to\infty}
\le
2\nu_*\tau_{\max},
\]
so one may take
\[
b=2\nu_*\tau_{\max}.
\]
Moreover,
\[
\|Q(h)-Q(g)\|_{\infty\to\infty}
\le
2\nu_*
\bigl(
K_\tau+\tau_{\max}K_\alpha
\bigr)
\|h-g\|_\infty,
\]
and therefore one may take
\[
K_Q
\le
2\nu_*
\bigl(
K_\tau+\tau_{\max}K_\alpha
\bigr).
\]
Proposition~\ref{prop:layers-self-conditioned-stability} then gives the
sufficient bi-Lipschitz condition
\[
2M\nu_*
\bigl(
K_\tau+\tau_{\max}K_\alpha
\bigr)
<
e^{-2\nu_*\tau_{\max}}.
\]
This condition controls input--output conditioning only; it does not imply
nonvanishing parameter gradients, favorable optimization, or convergence of
training.

\subsection{Depth-consistent propagation}
\label{app:layers-propagation-remarks}

The pathwise interpretation of geometry-aware propagation is meaningful only
when subsequent operations respect the distinction between latent-site mixing
and feature-channel mixing.
An unrestricted operator that mixes the latent-site index can introduce
dependencies outside descriptor-level reachability and thereby destroy the
pathwise interpretation.
By contrast, feature-channel transformations applied independently at every
site---for example, right multiplication of
\(H\in\mathbb R^{d\times p_h}\) by \(W_{\rm in}\) or \(W_{\rm out}\)---do
not by themselves create new site-to-site paths.

Accordingly, the residual bottleneck in
Eq.~\eqref{eq:layers-giof-layer} is compatible with the propagation
interpretation: the geometry-aware operator acts along the latent-site
dimension, while \(W_{\rm in}\) and \(W_{\rm out}\) act only along feature
channels.
Across multiple geometry-aware layers, Proposition~\ref{prop:properties-composable-envelope}
then controls the support of the realized propagation product, provided that
the intervening site-to-site operators obey the same descriptor-level
reachability envelope.
For self-conditioned stacks, this remains an operator-level statement; the
dependence of each selector on intermediate representations must additionally
satisfy conditions such as those in
Proposition~\ref{prop:layers-functional-ancestor-locality}.

\begin{remark}[Why one geometry-aware layer is often not enough]
\label{rem:layers-why-one-geometry-layer-not-enough}
If geometry is imposed only at a single layer and then followed by unconstrained dense mixing, the pathwise interpretation is immediately lost.
By contrast, when the same descriptor or a compatible family of descriptors is preserved across depth, long-range feature transport remains mediated by admissible paths.
This is the architectural meaning of geometry-aware propagation:
it is a depth-consistent transport bias, not merely a one-off local regularizer.
\end{remark}

\subsection{Balanced transport layers}
\label{app:layers-balanced-transport}

The propagation family of Section~\ref{sec:layers} preserves nonnegativity
and constant signals, but these properties do not imply conservation of the
sum of the entries of a column-vector signal.
When the latent dynamics are intended to represent redistribution of
nonnegative mass, we therefore impose an additional left-conservation
constraint.

Under the column-vector convention used throughout the paper,
total-mass conservation means
$
\mathbf1^\top y=\mathbf1^\top h.
$
For an infinitesimal generator this requires
$
\mathbf1^\top L=0,
$
in addition to the constant-preservation condition
$
L\mathbf1=0.
$

\begin{definition}[Balanced generator family]
\label{def:layers-balanced-generator-family}
The \emph{balanced generator family} associated with \(\mathfrak S\) is
\[
\mathcal L_{\mathfrak S}^{\mathrm{bal}}
:=
\left\{
L\in\mathcal L_{\mathfrak S}
\;\middle|\;
\mathbf1^\top L=0
\right\}.
\]
\end{definition}

Thus a balanced generator is simultaneously Metzler, constant preserving,
and total-mass preserving.
In the constrained-family notation introduced in
Section~\ref{sec:operators},
$
\mathcal L_{\mathfrak S}^{\mathrm{bal}}
=
\mathcal L_{\mathfrak S;\mathbf1,\mathbf1}.
$

To retain the same scale--shape convention as the main propagation family,
we also normalize nonzero balanced generators.

\begin{definition}[Normalized balanced generator shape]
\label{def:layers-normalized-balanced-generator}
If
\(
\mathcal L_{\mathfrak S}^{\mathrm{bal}}\neq\{0\},
\)
define
\[
\mathcal L_{\mathfrak S}^{\mathrm{bal},(1)}
:=
\left\{
L\in\mathcal L_{\mathfrak S}^{\mathrm{bal}}
\;\middle|\;
\rho(L)=1
\right\}.
\]
\end{definition}

Every nonzero balanced generator has \(\rho(L)>0\), so the normalized
balanced slice is nonempty whenever the balanced family is nontrivial.
This normalization keeps the generator \(L\) responsible for propagation
shape and reserves the scalar \(\tau\) for propagation magnitude, exactly as
in Definition~\ref{def:operators-propagation-family}.

\begin{proposition}[Acyclicity obstructs nontrivial balance]
\label{prop:layers-acyclic-balanced-obstruction}
Suppose that the directed graph induced by the admissible
off-diagonal relation \(\mathcal E_{\mathrm{off}}\) is acyclic.
Then
$
\mathcal L_{\mathfrak S}^{\mathrm{bal}}=\{0\}.
$
Hence every balanced propagation associated with this descriptor
is the identity.

More generally, every positive off-diagonal edge of a balanced
generator belongs to a directed cycle of its positive-rate graph.
\end{proposition}

A directed cycle in the admissibility graph is therefore necessary for
$
\mathcal L_{\mathfrak S}^{\mathrm{bal}}\neq\{0\}.
$
It is not sufficient: because
\(\mathcal A_{\mathfrak S}\) may be a strict compressed subset of the
ambient masked matrix space, the channel-sharing constraints may exclude
every balanced circulation supported on that cycle.
Thus graph-theoretic feasibility and channel-family feasibility are distinct
requirements.

\begin{definition}[Geometry-aware balanced transport layer]
\label{def:layers-balanced-transport-layer}
Assume
$
\mathcal L_{\mathfrak S}^{\mathrm{bal}}\neq\{0\}.
$
Let
$
L:\mathcal C\to
\mathcal L_{\mathfrak S}^{\mathrm{bal},(1)}
$
and
$
\tau:\mathcal C\to[0,\infty)
$
be continuous.
A \emph{geometry-aware balanced transport layer} is the map
$
\mathcal T_{\mathfrak S}:
\mathcal C\times\mathbb R^d
\to
\mathbb R^d
$
defined by
$
\mathcal T_{\mathfrak S}(c,h)
=
T(c)h,
\qquad
T(c)
:=
e^{\tau(c)L(c)}.
$
\end{definition}

The word ``transport'' now has a precise meaning.
The off-diagonal coefficients of \(L(c)\) describe directed transfers between sites, \(L(c)\mathbf 1=0\) preserves uniform levels, and the additional condition \(\mathbf 1^\top L(c)=0\) enforces conservation of total mass.

\begin{proposition}[Mass-conservation interpretation]
\label{prop:layers-mass-conservation-interpretation}
Let \(\mathcal T_{\mathfrak S}\) be a geometry-aware balanced transport
layer and write
$
T(c)=e^{\tau(c)L(c)}.
$
Then, for every \(c\in\mathcal C\),
\begin{enumerate}
    \item \(T(c)_{ij}\ge0\) for all \(i,j\);
    \item \(T(c)\mathbf1=\mathbf1\);
    \item \(\mathbf1^\top T(c)=\mathbf1^\top\);
    \item if \(h\ge0\), then \(T(c)h\ge0\);
    \item for every \(h\in\mathbb R^d\),
    $
    \mathbf1^\top T(c)h
    =
    \mathbf1^\top h.
    $
\end{enumerate}
Hence \(T(c)\) is doubly stochastic.
For nonnegative inputs, the layer therefore redistributes signal mass
without creating negative values or changing total mass.
In particular, if
$
h\ge0,
\qquad
\mathbf1^\top h=1,
$
then
$
\mathcal T_{\mathfrak S}(c,h)
$
remains in the probability simplex.
\end{proposition}

In particular, if \(h\ge0\) and \(\mathbf1^\top h=1\), then the output
remains in the probability simplex.

\begin{remark}[Propagation versus transport]
\label{rem:layers-propagation-vs-transport}
Every balanced transport layer is a propagation layer, but not every propagation layer is a transport layer.
The term ``transport'' should be reserved for settings in which positivity and mass conservation are part of the intended semantic interpretation of the latent dynamics.
\end{remark}

\begin{proposition}[Balanced transport under global average pooling]
\label{prop:layers-global-mean-pooling}
Let
\[
H\in\mathbb R^{d\times p},
\qquad
g(H):=\frac1d\mathbf1^\top H,
\]
and let \(T(H)\in\mathbb R^{d\times d}\) be doubly stochastic for every
\(H\).
Then
\[
g\!\left(T(H)H\right)=g(H),
\]
even when \(T\) depends on \(H\).

More generally, for arbitrary compatible feature maps
$
W_{\rm in}\in\mathbb R^{p\times q},
\qquad
W_{\rm out}\in\mathbb R^{q\times p},
$
the residual bottleneck
\[
\widetilde H
=
H+
\bigl[T(H)-I_d\bigr]
HW_{\rm in}W_{\rm out}
\]
also satisfies
$
g(\widetilde H)=g(H).
$
Consequently, if a balanced transport branch is followed immediately by
global average pooling, its contribution to the pooled representation is
identically zero.
Parameters used only to construct \(T(H)\) therefore receive zero gradient
through that pooled path.
\end{proposition}

\subsubsection{Implementation of balanced transport layers}
\label{app:layers-balanced-transport-implementation}

The balanced layer can be implemented either by selecting directly inside
the normalized balanced family or by projecting a raw shape onto that
family.
Both routes preserve the scale--shape decomposition used by the ordinary
propagation layer.

\paragraph{Direct normalized basis parameterization.}
Let
\[
C_1,\dots,C_m
\in
\mathcal L_{\mathfrak S}^{\mathrm{bal}}\setminus\{0\}
\]
be preconstructed balanced generators and define
\[
\widehat C_r
:=
\frac{C_r}{\rho(C_r)}
\in
\mathcal L_{\mathfrak S}^{\mathrm{bal},(1)}.
\]
A selector may predict simplex weights
\[
\eta(c)
=
\operatorname{softmax}(g_\eta(c)),
\qquad
\eta_r(c)\ge0,
\qquad
\sum_{r=1}^m\eta_r(c)=1,
\]
and set
$
L(c)
=
\sum_{r=1}^m
\eta_r(c)\widehat C_r.
$
Convexity then gives
$
L(c)
\in
\mathcal L_{\mathfrak S}^{\mathrm{bal},(1)}
$
for every context.

\paragraph{Exchange generators as a special case.}
Suppose reciprocal transfer is admissible between sites \(i\) and \(j\).
Define
\[
C^{(ij)}
:=
e_i e_j^\top
+
e_j e_i^\top
-
e_i e_i^\top
-
e_j e_j^\top.
\]
This matrix satisfies
\[
C^{(ij)}\mathbf1=0,
\qquad
\mathbf1^\top C^{(ij)}=0,
\]
and has nonnegative off-diagonal entries.
However, reciprocal admissibility guarantees only support compatibility.
Because \(\mathcal A_{\mathfrak S}\) may be a strict compressed subset of
the ambient masked matrix space, one must additionally verify
$
C^{(ij)}\in\mathcal A_{\mathfrak S}.
$
Whenever this holds,
\[
C^{(ij)}
\in
\mathcal L_{\mathfrak S}^{\mathrm{bal}},
\qquad
\widehat C^{(ij)}
:=
\frac{C^{(ij)}}{\rho(C^{(ij)})}
\in
\mathcal L_{\mathfrak S}^{\mathrm{bal},(1)},
\]
so normalized exchange generators can be included in the direct basis above.

\paragraph{Projection-based parameterization.}
When an explicit balanced basis is inconvenient, predict a raw matrix
$
\widetilde L(c)\in\mathbb R^{d\times d}
$
and define
$
L(c)
=
\Pi_{\mathcal L_{\mathfrak S}^{\mathrm{bal},(1)}}
\bigl(\widetilde L(c)\bigr),
$
where
\[
\Pi_{\mathcal L_{\mathfrak S}^{\mathrm{bal},(1)}}(M)
:=
\arg\min_{
N\in\mathcal L_{\mathfrak S}^{\mathrm{bal},(1)}
}
\|N-M\|_F^2.
\]
Whenever
\(
\mathcal L_{\mathfrak S}^{\mathrm{bal}}\neq\{0\},
\)
the normalized balanced family is a nonempty closed convex set, so this
projection exists and is unique.
The feasible set simultaneously enforces geometry compatibility,
nonnegative off-diagonal rates, both conservation laws, and unit generator
rate.

After selecting \(L(c)\), the propagation magnitude is chosen separately:
\[
\tau(c)\ge0,
\qquad
T(c)=e^{\tau(c)L(c)},
\qquad
y=T(c)h.
\]
For moderate \(d\), the exponential can be evaluated directly.
For large sparse geometries, one may instead evaluate its action on \(h\)
using a Krylov-subspace or related exponential-action method.

\begin{algorithm}[t]
\caption{Geometry-Aware Balanced Transport Layer}
\label{alg:balanced-transport-layer}
\begin{algorithmic}[1]
\REQUIRE Context \(c\in\mathcal C\), input \(h\in\mathbb R^d\),
normalized balanced basis
\(\{\widehat C_r\}_{r=1}^m
\subseteq
\mathcal L_{\mathfrak S}^{\mathrm{bal},(1)}\)
or projection onto
\(\mathcal L_{\mathfrak S}^{\mathrm{bal},(1)}\),
and propagation-scale head \(\tau(\cdot)\)
\IF{using a direct balanced basis}
    \STATE Compute
    \[
    \eta(c)=\operatorname{softmax}(g_\eta(c))
    \]
    \STATE Set
    \[
    L(c)
    =
    \sum_{r=1}^m
    \eta_r(c)\widehat C_r
    \]
\ELSE
    \STATE Predict a raw matrix \(\widetilde L(c)\)
    \STATE Set
    \[
    L(c)
    =
    \Pi_{\mathcal L_{\mathfrak S}^{\mathrm{bal},(1)}}
    \bigl(\widetilde L(c)\bigr)
    \]
\ENDIF
\STATE Compute \(\tau(c)\ge0\)
\STATE Form
\[
T(c)=e^{\tau(c)L(c)}
\]
\STATE Return
\[
y=T(c)h
\]
\end{algorithmic}
\end{algorithm}

\subsection{Multiscale correction layers}
\label{app:layers-multiscale-correction}

The multiscale operator families have already been defined in
Appendix~\ref{app:operators-supplement}.
Here we only record how the multiscale interaction family is realized as a
residual neural layer.

Recall the finest space \(X^{(0)}\), the coarser spaces
\[
X^{(1)},\dots,X^{(L)},
\]
and the fixed transfer maps
\[
P_\ell:\mathbb R^{d_\ell}\to\mathbb R^{d_0},
\qquad
R_\ell:\mathbb R^{d_0}\to\mathbb R^{d_\ell},
\qquad
P_0=R_0=I_{d_0}.
\]
Each scale carries its own descriptor \(\mathfrak S^{(\ell)}\), and
Definition~\ref{def:operators-multiscale-interaction-family} defines
\[
\mathcal A^{\mathrm{ms}}
=
\left\{
\sum_{\ell=0}^L
P_\ell A_\ell R_\ell
\;\middle|\;
A_\ell\in
\mathcal A_{\mathfrak S^{(\ell)}}
\right\}.
\]

\begin{definition}[Geometry-aware multiscale correction layer]
\label{def:layers-multiscale-correction-layer}
A \emph{geometry-aware multiscale correction layer} is a map
\[
\mathcal M_{\mathrm{ms}}:
\mathcal C\times\mathbb R^{d_0}
\to
\mathbb R^{d_0}
\]
of the form
\[
\mathcal M_{\mathrm{ms}}(c,h)
=
h+A^{\mathrm{ms}}(c)h,
\]
where
\[
A^{\mathrm{ms}}(c)
=
\sum_{\ell=0}^L
P_\ell A_\ell(c)R_\ell,
\qquad
A_\ell(c)
\in
\mathcal A_{\mathfrak S^{(\ell)}}.
\]
\end{definition}

Operationally, each scale first restricts the finest representation,
applies a geometry-aware interaction at that resolution, and transfers the
result back to the finest space.
The layer then adds the sum of these scale-specific corrections to the
original representation.
If every selector \(c\mapsto A_\ell(c)\) is continuous, then
\[
(c,h)\longmapsto\mathcal M_{\mathrm{ms}}(c,h)
\]
is jointly continuous.

The correction construction above should be distinguished from multiscale
propagation.
A transferred sum of scale-wise generators need not remain Metzler after
restriction and prolongation.
Therefore propagation across scales must use the
\emph{actual multiscale generator family}
\(\mathcal L^{\mathrm{ms}}\) of
Definition~\ref{def:operators-actual-multiscale-generator-family},
rather than exponentiating an arbitrary transferred generator sum.

The multiscale realization is useful when relevant dependencies occur at
widely separated spatial or structural ranges:
fine scales retain local geometric detail, whereas coarse scales provide
shorter routes for long-range interaction.
The benefit is therefore not that single-scale propagation is incapable of
long-range influence, but that multiscale structure can represent such
influence more directly and with fewer repeated fine-scale propagation
steps.

\subsection{Algorithmic summaries}
\label{app:layers-algorithms}

\begin{algorithm}[t]
\caption{Geometry-Aware Residual Adapter}
\label{alg:residual-adapter}
\begin{algorithmic}[1]
\REQUIRE Context \(c\in\mathcal C\), input representation \(h\in\mathbb R^d\), geometry descriptor \(\mathfrak S\), coefficient heads for \(\theta(c)\) and \(\delta(c)\)
\STATE Compute geometry-channel coefficients \(\theta(c)=(\theta_1(c),\dots,\theta_R(c))\)
\STATE Compute diagonal coefficients \(\delta(c)\in\mathbb R^d\)
\STATE Form the interaction operator
\[
A(c)=\sum_{r=1}^R \theta_r(c)\Phi_r+\operatorname{Diag}(\delta(c))
\]
with \(\theta_r(c)\ge 0\) for \(r\in\mathcal R_+\)
\STATE Return
\[
y=h+A(c)h
\]
\end{algorithmic}
\end{algorithm}

\begin{algorithm}[t]
\caption{Geometry-Aware Propagation Layer}
\label{alg:propagation-layer}
\begin{algorithmic}[1]
\REQUIRE Context \(c\in\mathcal C\), input representation
\(h\in\mathbb R^d\), normalized generator basis
\(\{\widehat B_r\}_{r\in\mathcal R_{\mathrm p}}
\subseteq
\mathcal L_{\mathfrak S}^{(1)}\),
mixture head \(f_\theta\), and propagation-scale head \(\tau(\cdot)\)
\STATE Compute
\[
\alpha(c)
=
\operatorname{softmax}(f_\theta(c))
\]
\STATE Form the normalized generator
\[
L(c)
=
\sum_{r\in\mathcal R_{\mathrm p}}
\alpha_r(c)\widehat B_r
\in
\mathcal L_{\mathfrak S}^{(1)}
\]
\STATE Compute the propagation magnitude
\[
\tau(c)\ge0
\]
\STATE Form
\[
P(c)
=
e^{\tau(c)L(c)}
\]
\STATE Return
\[
y=P(c)h
\]
\end{algorithmic}
\end{algorithm}

Algorithm~\ref{alg:propagation-layer} describes the propagation core.
The matrix-valued residual bottleneck used by GIOF is obtained by inserting
this propagator into Eq.~\eqref{eq:layers-giof-layer}; its experiment-specific
parameterization is given in Appendix~\ref{app:exp_architecture}.

\section{Supplementary Canonical Matrix Examples}
\label{app:canonical-examples-supplement}

This appendix instantiates the abstract descriptor-to-operator hierarchy developed in
Sections~\ref{sec:prelim}--\ref{sec:layers} through a collection of canonical matrix
templates. The purpose is not to introduce new operator families, but to show how
concrete geometric constructions realize the previously defined interaction,
generator, propagation, block-valued, and multiscale families.

Throughout this section, we retain the receiver--source convention
\((Ah)_i=\sum_j A_{ij}h_j\), so that the ordered pair \((z_i,z_j)\) represents
influence from source \(z_j\) to receiver \(z_i\). Whenever different examples use
different admissibility relations or channel dictionaries, we denote the
corresponding descriptors explicitly rather than reusing a single undifferentiated
symbol \(\mathfrak S\).

The examples are organized from scalar single-scale constructions to progressively
richer extensions. We first record a reusable row-balancing map that converts
nonnegative off-diagonal interactions into generators. We then treat local kernels,
convolutional and graph templates, and attention-style normalized mixing; next we
consider directional and barrier-aware propagation; finally, we connect the
multiscale, Hodge, and sheaf constructions to the corresponding extensions already
defined in the preceding appendices.

\subsection{Reusable row-balancing compilation}
\label{app:canonical-examples-row-balancing}

Several examples below start from an interaction matrix with nonnegative
off-diagonal entries and then compile it into a constant-preserving generator.
We isolate this operation once so that the same diagonal correction need not be
redefined separately for local, directional, and barrier-aware kernels.

\begin{definition}[Row-balanced generator map]
\label{def:examples-row-balanced-generator-map}
For any \(K\in\mathbb R^{d\times d}\), define its off-diagonal part by
\[
\operatorname{Off}(K)_{ij}
:=
\begin{cases}
K_{ij}, & i\neq j,\\
0, & i=j,
\end{cases}
\]
and define
\[
\Gamma(K)
:=
\operatorname{Off}(K)
-
\operatorname{Diag}\!\bigl(\operatorname{Off}(K)\mathbf 1\bigr).
\]
Equivalently,
\[
\Gamma(K)_{ij}
=
\begin{cases}
K_{ij}, & i\neq j,\\[0.4em]
-\displaystyle\sum_{j\neq i}K_{ij}, & i=j.
\end{cases}
\]
\end{definition}

The map \(\Gamma\) preserves all off-diagonal transfers and replaces only the
diagonal so that constants are preserved.

\begin{proposition}[Canonical generator property]
\label{prop:examples-canonical-generator-property}
Let \(K\in\mathcal A_{\mathfrak S}\) satisfy
\[
K_{ij}\ge 0
\qquad
\text{for all } i\neq j.
\]
Then
\[
\Gamma(K)\in\mathcal L_{\mathfrak S}.
\]
If \(\Gamma(K)\neq 0\), writing
\[
\widehat L
=
\frac{\Gamma(K)}{\rho(\Gamma(K))}
\in
\mathcal L_{\mathfrak S}^{(1)}
\]
gives
\[
e^{\tau\Gamma(K)}
=
e^{\,\tau\rho(\Gamma(K))\widehat L}
\in
\mathcal P_{\mathfrak S}
\qquad
\text{for every }\tau\ge 0.
\]
If \(\Gamma(K)=0\), the exponential is the identity.
\end{proposition}

The first conclusion follows because
\(\mathcal A_{\mathfrak S}\) contains arbitrary diagonal self-action:
removing or replacing the diagonal therefore does not leave the interaction
family. The resulting matrix has nonnegative off-diagonal entries and zero row
sums by construction.

\begin{proposition}[Symmetric interactions yield balanced generators]
\label{prop:examples-symmetric-yields-balanced}
If, in addition, \(K=K^\top\), then
\[
\mathbf 1^\top \Gamma(K)=0.
\]
Hence
$
\Gamma(K)\in
\mathcal L_{\mathfrak S}^{\mathrm{bal}}.
$
\end{proposition}

Thus row balancing always produces a propagation generator, whereas balanced
transport follows when the underlying off-diagonal interaction is symmetric,
or more generally when the additional left-conservation condition holds.

\subsection{Local kernels and classical deep-learning templates}
\label{app:canonical-examples-local-classical}

\paragraph{Local kernel matrices.}
We begin with the simplest scalar example: interaction strength is determined
only by local proximity in an auxiliary coordinate realization.

Let
\[
X=\{z_1,\dots,z_d\}
\]
and let
$
\iota:X\to\mathbb R^m
$
be an auxiliary coordinate map. Let
\(\mathcal E_{\mathrm{loc}}\subseteq X\times X\) be a fixed admissibility
relation, for example a radius graph or a \(k\)-nearest-neighbor relation.
For a fixed bandwidth \(\varepsilon>0\), define
\[
\Psi_{\mathrm{loc}}(z_i,z_j)
:=
\iota(z_j)-\iota(z_i),
\]
and
\[
\phi_{\mathrm{loc}}(z_i,z_j)
:=
\kappa_\varepsilon(z_i,z_j)
:=
\exp\!\left(
-\frac{\|\iota(z_i)-\iota(z_j)\|^2}{\varepsilon^2}
\right).
\]
The corresponding descriptor is
\[
\mathfrak S_{\mathrm{loc}}
=
\bigl(
X,
\mathcal E_{\mathrm{loc}},
\Psi_{\mathrm{loc}},
\{\phi_{\mathrm{loc}}\},
\{1\}
\bigr),
\]
where the unique channel is placed in the positive-channel set because it is
nonnegative.

\begin{definition}[Local kernel interaction matrix]
\label{def:examples-local-kernel-interaction}
The local kernel interaction matrix is
\[
K^{\mathrm{loc}}_\varepsilon\in\mathbb R^{d\times d},
\qquad
(K^{\mathrm{loc}}_\varepsilon)_{ij}
=
\mathbf 1_{\mathcal E_{\mathrm{loc}}}(z_i,z_j)
\kappa_\varepsilon(z_i,z_j).
\]
\end{definition}

\begin{proposition}[Local kernel matrices as interaction operators]
\label{prop:examples-local-kernel-in-interaction-family}
The matrix \(K^{\mathrm{loc}}_\varepsilon\) is precisely the masked realization
of the unique channel of \(\mathfrak S_{\mathrm{loc}}\). Hence
$
K^{\mathrm{loc}}_\varepsilon
\in
\mathcal A_{\mathfrak S_{\mathrm{loc}}}.
$
\end{proposition}

\begin{definition}[Local kernel generator]
\label{def:examples-local-kernel-generator}
The associated local kernel generator is
$
L^{\mathrm{loc}}_\varepsilon
:=
\Gamma(K^{\mathrm{loc}}_\varepsilon).
$
Equivalently,
\[
(L^{\mathrm{loc}}_\varepsilon)_{ij}
=
\begin{cases}
\mathbf 1_{\mathcal E_{\mathrm{loc}}}(z_i,z_j)
\kappa_\varepsilon(z_i,z_j),
& i\neq j,\\[0.6em]
-\displaystyle\sum_{j\neq i}
\mathbf 1_{\mathcal E_{\mathrm{loc}}}(z_i,z_j)
\kappa_\varepsilon(z_i,z_j),
& i=j.
\end{cases}
\]
\end{definition}

By Proposition~\ref{prop:examples-canonical-generator-property},
$
L^{\mathrm{loc}}_\varepsilon
\in
\mathcal L_{\mathfrak S_{\mathrm{loc}}}.
$
Therefore its exponential defines stable finite propagation generated by local
admissible edges. Importantly, the finite-time operator is not restricted to
one-hop support: by
Theorem~\ref{thm:properties-reachability-preserving-propagation}, influence may
extend to multiple hops, but only along paths in
\(\mathcal E_{\mathrm{loc}}\).

If \(\mathcal E_{\mathrm{loc}}\) is symmetric, then
\(K^{\mathrm{loc}}_\varepsilon\) is symmetric and
Proposition~\ref{prop:examples-symmetric-yields-balanced} additionally gives
$
L^{\mathrm{loc}}_\varepsilon
\in
\mathcal L_{\mathfrak S_{\mathrm{loc}}}^{\mathrm{bal}}.
$

\begin{remark}[Interpretation]
\label{rem:examples-local-kernel-interpretation}
This construction is the canonical proximity-based realization of the framework:
the descriptor determines one-step neighborhood structure, row balancing converts
the resulting positive interaction into a generator, and exponentiation combines
all admissible propagation depths.
\end{remark}

\paragraph{CNN-style stencil matrices.}
Assume that the latent sites carry grid positions
$
p_i\in\mathbb Z^n,
$
and fix a finite offset stencil
$
\mathcal S\subset\mathbb Z^n.
$
For each \(\Delta\in\mathcal S\), define
\[
\phi_\Delta(z_i,z_j)
:=
\mathbf 1\{p_j-p_i=\Delta\},
\]
and let
\[
(z_i,z_j)\in\mathcal E_{\mathrm{cnn}}
\quad\Longleftrightarrow\quad
p_j-p_i\in\mathcal S.
\]
Let \(\mathfrak S_{\mathrm{cnn}}\) denote the descriptor determined by this
admissibility relation and these offset channels.

Because standard convolutional weights may have either sign, we take
$
\mathcal R_+=\varnothing
$
for the unrestricted signed stencil realization. If a particular application
requires nonnegative coefficients on selected offsets, only those channels should
instead be included in \(\mathcal R_+\).

\begin{definition}[CNN-style stencil interaction matrix]
\label{def:examples-cnn-stencil-matrix}
A CNN-style stencil interaction matrix is any matrix
\[
A^{\mathrm{cnn}}
=
\sum_{\Delta\in\mathcal S}
w_\Delta\Phi_\Delta
+
\operatorname{Diag}(\delta),
\qquad
w_\Delta\in\mathbb R,
\quad
\delta\in\mathbb R^d,
\]
where \(\Phi_\Delta\) is the masked channel matrix induced by
\(\phi_\Delta\).
\end{definition}

\begin{proposition}[CNN-style matrices as geometry-induced interactions]
\label{prop:examples-cnn-exact-geometry-induced}
Every matrix \(A^{\mathrm{cnn}}\) above belongs to
$
\mathcal A_{\mathfrak S_{\mathrm{cnn}}}.
$
The unrestricted geometry-induced family allows arbitrary sitewise diagonal
self-action. Exact translation-shared scalar convolution is the tied subfamily
obtained by taking
$
\delta=\delta_0\mathbf 1,
$
or, equivalently, by absorbing the common self-interaction into a shared
zero-offset channel.
\end{proposition}

Exact translation equivariance additionally requires the grid domain and its
boundary convention to respect translations, as on a periodic grid or on an
appropriate interior region. Thus the GIOF interaction family contains the
usual finite-stencil convolutional coupling as a structured subfamily while
also permitting controlled departures from exact translation sharing.

\paragraph{Graph message-passing matrices.}
Let the latent sites be the vertices of a directed or undirected graph, and let
its admissible ordered receiver--source pairs be partitioned into finitely many
edge classes,
\[
\mathcal E_{\mathrm{graph}}
=
\bigsqcup_{t=1}^T\mathcal E_t.
\]
Define
$
\phi_t(z_i,z_j)
:=
\mathbf 1\{(z_i,z_j)\in\mathcal E_t\}.
$
Let \(\mathfrak S_{\mathrm{graph}}\) denote the corresponding descriptor.

For ordinary signed edge-type coefficients we again take
$
\mathcal R_+=\varnothing.
$
If selected edge-type contributions are required to be nonnegative, the
corresponding channel indices may instead be included in \(\mathcal R_+\).

\begin{definition}[Graph message-passing matrix]
\label{def:examples-graph-message-passing-matrix}
A graph message-passing matrix is any matrix
\[
A^{\mathrm{graph}}
=
\sum_{t=1}^T
\theta_t\Phi_t
+
\operatorname{Diag}(\delta),
\qquad
\theta_t\in\mathbb R,
\quad
\delta\in\mathbb R^d,
\]
where \(\Phi_t\) is the masked channel matrix induced by \(\phi_t\).
\end{definition}

\begin{proposition}[Graph message passing as a geometry-induced interaction family]
\label{prop:examples-graph-message-passing-geometry-induced}
Every matrix \(A^{\mathrm{graph}}\) above belongs to
$
\mathcal A_{\mathfrak S_{\mathrm{graph}}}.
$
It represents the site-level aggregation skeleton of typed or directional
message passing. When each site carries a multi-channel feature vector and
messages additionally apply feature transforms, the corresponding exact
realization is obtained through the block-valued lifting of
Appendix~\ref{app:operators-block-lifting}.
\end{proposition}

\paragraph{Attention-style score and mixing matrices.}
Attention differs from the preceding examples because the final mixing matrix
is obtained through a nonlinear row-wise normalization. Accordingly, the
geometry-induced object to specify first is the admissible score structure.

Let
$
\mathcal E_{\mathrm{att}}\subseteq X\times X
$
be a global or windowed admissibility relation. For a context \(c\), let
$
q_i(c),k_j(c)\in\mathbb R^p
$
be query and key vectors, and let \(b_{ij}\) be a geometry-derived bias.

\begin{definition}[Attention-style score matrix]
\label{def:examples-attention-score-matrix}
Define
\[
S^{\mathrm{att}}_{ij}(c)
=
\begin{cases}
\displaystyle
\frac{\langle q_i(c),k_j(c)\rangle}{\sqrt p}
+b_{ij},
&
(z_i,z_j)\in\mathcal E_{\mathrm{att}},
\\[0.8em]
-\infty,
&
(z_i,z_j)\notin\mathcal E_{\mathrm{att}}.
\end{cases}
\]
The associated mixing matrix is
\[
A^{\mathrm{att}}(c)
=
\operatorname{RowSoftmax}
\bigl(S^{\mathrm{att}}(c)\bigr),
\]
where each row is normalized only over admissible source sites.
\end{definition}

\begin{proposition}[Geometry-conditioned attention support]
\label{prop:examples-attention-geometry-aware}
Assume that every receiver has at least one admissible source and that every
score on an admissible pair is finite. Then
\[
A^{\mathrm{att}}(c)\ge 0,
\qquad
A^{\mathrm{att}}(c)\mathbf 1=\mathbf 1,
\]
and
$
A^{\mathrm{att}}_{ij}(c)=0
\qquad
\text{whenever }
(z_i,z_j)\notin\mathcal E_{\mathrm{att}}.
$
Thus, for each fixed context, attention defines a nonnegative row-stochastic
mixing operator supported on the admissibility relation.
\end{proposition}

Two distinctions from the preceding fixed-family examples are essential.
First, the masked score array contains \(-\infty\) and is not itself an element
of the real matrix space \(\mathbb R^{d\times d}\). Second, row-wise softmax is
nonlinear and does not generally preserve the finite linear dictionary span
defining a fixed family \(\mathcal A_{\mathfrak S}\).

Moreover, when \(q_i(c)\) and \(k_j(c)\) vary with the context, the pairwise
score law is context dependent and should be interpreted through the
context-dependent descriptor extension of
Appendix~\ref{app:prelim-pairwise-attributes}. Membership of the resulting
mixing matrix in any particular fixed GIOF interaction family therefore
requires a separate representability argument.

\begin{remark}[Scope of the classical-template examples]
\label{rem:examples-local-classical-summary}
The local-kernel, stencil, and graph examples are direct realizations of
appropriately chosen geometry-induced interaction families. Attention shares
the same admissibility-based geometric organization, but its state-dependent
scores and nonlinear row normalization place it outside a fixed linear
dictionary family in general. This distinction prevents the architectural
analogy from being mistaken for an exact algebraic identification.
\end{remark}

\subsection{Directional and barrier-respecting examples}
\label{app:canonical-examples-drift-generator-barrier}

\paragraph{Anisotropic drift matrices.}
Many latent spaces are not isotropic: information may preferentially move along certain local directions.

Assume that the auxiliary coordinate map \(\iota:X\to\mathbb R^m\) is given, and that each site \(z_i\) is equipped with
\begin{enumerate}
    \item a positive definite matrix \(G_i\in\mathbb R^{m\times m}\), representing a local anisotropic metric;
    \item a drift vector \(b_i\in\mathbb R^m\), representing a preferred local transport direction.
\end{enumerate}

Fix parameters \(\varepsilon>0\) and \(\beta\ge 0\), and define
\[
\kappa^{\mathrm{aniso}}(z_i,z_j)
:=
\exp\!\left(
-\frac{1}{\varepsilon^2}(\iota(z_j)-\iota(z_i))^\top G_i (\iota(z_j)-\iota(z_i))
+\beta\, b_i^\top (\iota(z_j)-\iota(z_i))
\right).
\]

\begin{definition}[Anisotropic drift interaction matrix]
\label{def:examples-anisotropic-drift-interaction}
The \emph{anisotropic drift interaction matrix} is
\[
K^{\mathrm{drift}}\in\mathbb R^{d\times d},
\qquad
(K^{\mathrm{drift}})_{ij}
=
\mathbf 1_{\mathcal E}(z_i,z_j)\,
\kappa^{\mathrm{aniso}}(z_i,z_j).
\]
\end{definition}

This matrix is generally not symmetric.
Its asymmetry is not arbitrary; it is induced by the local drift field \(b_i\) and anisotropic metric \(G_i\).

Let \(\mathfrak S_{\mathrm{drift}}\) denote the descriptor whose admissibility
relation is \(\mathcal E\), whose pairwise attributes include
$
\iota(z_j)-\iota(z_i),\qquad G_i,\qquad b_i,
$
and whose unique channel is
$
\phi_{\mathrm{drift}}(z_i,z_j)
=
\kappa^{\mathrm{aniso}}(z_i,z_j).
$
Because this channel is strictly positive, take its channel index to belong to
the positive-channel set.

\begin{proposition}[Anisotropic drift matrices as geometry-induced operators]
\label{prop:examples-anisotropic-drift-geometry-induced}
The matrix \(K^{\mathrm{drift}}\) belongs to
$
\mathcal A_{\mathfrak S_{\mathrm{drift}}}.
$
Its possible asymmetry is induced by the geometry descriptor, in particular
by the receiver-dependent metric \(G_i\), the drift vector \(b_i\), and any
directionality encoded by the admissibility relation, rather than by
unconstrained matrix parameters.
\end{proposition}

\begin{definition}[Anisotropic drift generator]
\label{def:examples-anisotropic-drift-generator}
The associated drift generator is
$
L^{\mathrm{drift}}
:=
\Gamma(K^{\mathrm{drift}}).
$
\end{definition}

By Proposition~\ref{prop:examples-canonical-generator-property},
$
L^{\mathrm{drift}}
\in
\mathcal L_{\mathfrak S_{\mathrm{drift}}}.
$
Unlike the symmetric local-kernel case, no left-conservation or balanced
transport property follows automatically when
\(K^{\mathrm{drift}}\neq(K^{\mathrm{drift}})^\top\).

\begin{remark}[Interpretation]
\label{rem:examples-anisotropic-drift-interpretation}
This example is appropriate when the latent geometry has preferred directions, such as semantic flow, local orientation, anisotropic connectivity, or directional refinement.
It shows that the framework can model geometry beyond undirected proximity.
\end{remark}

\paragraph{Barrier-masked matrices.}
We next consider a geometry in which admissibility is determined by hard
visibility constraints rather than by distance alone.

Assume that
$
\iota(X)\subseteq\Omega\subset\mathbb R^m
$
and that \(\mathcal O\subset\Omega\) is an obstacle set. Define
\[
\operatorname{Vis}(z_i,z_j)
:=
\mathbf 1
\bigl\{
[\iota(z_i),\iota(z_j)]
\cap
\mathcal O
=
\varnothing
\bigr\},
\]
and
\[
\mathcal E_{\mathrm{bar}}
:=
\{
(z_i,z_j)\in X\times X:
\operatorname{Vis}(z_i,z_j)=1
\}.
\]

For \(\varepsilon>0\), define the barrier-masked interaction
\[
K^{\mathrm{bar}}_{ij}
=
\operatorname{Vis}(z_i,z_j)
\exp\!\left(
-\frac{\|\iota(z_i)-\iota(z_j)\|^2}{\varepsilon^2}
\right).
\]
Let \(\mathfrak S_{\mathrm{bar}}\) denote the corresponding descriptor with
this nonnegative kernel as its geometry channel.

\begin{definition}[Barrier-masked propagation generator]
\label{def:examples-barrier-masked-generator}
The barrier-masked propagation generator is
$
L^{\mathrm{bar}}
:=
\Gamma(K^{\mathrm{bar}}).
$
\end{definition}

By Proposition~\ref{prop:examples-canonical-generator-property},
$
L^{\mathrm{bar}}
\in
\mathcal L_{\mathfrak S_{\mathrm{bar}}}.
$

\begin{proposition}[Barrier-respecting propagation]
\label{prop:examples-barrier-respecting-propagation}
For every \(\tau>0\) and every \(i\neq j\),
$
\bigl(e^{\tau L^{\mathrm{bar}}}\bigr)_{ij}>0
$
if and only if
$
z_j\rightsquigarrow z_i
$
in the directed graph induced by
\(\mathcal E_{\mathrm{bar}}\).
Consequently, finite propagation cannot cross an obstacle except through a
sequence of visibility-admissible elementary transfers.
\end{proposition}

This is the direct specialization of
Theorem~\ref{thm:properties-reachability-preserving-propagation}.
Because line-segment visibility is symmetric and the Gaussian factor is also
symmetric,
$
K^{\mathrm{bar}}
=
(K^{\mathrm{bar}})^\top.
$
Hence Proposition~\ref{prop:examples-symmetric-yields-balanced} gives
$
L^{\mathrm{bar}}
\in
\mathcal L_{\mathfrak S_{\mathrm{bar}}}^{\mathrm{bal}}.
$
Thus this particular barrier construction is not merely stable
barrier-respecting propagation: it also defines balanced transport.

\paragraph{Multiscale coarse-to-fine matrices.}
We now instantiate the multiscale families already defined in
Appendix~\ref{app:operators-multiscale}; no new multiscale family is introduced
here.

Let
\[
X^{(0)}=X,
\qquad
X^{(1)},\dots,X^{(L)}
\]
be the scale hierarchy, with fixed transfer maps
\[
R_\ell:\mathbb R^{d_0}\to\mathbb R^{d_\ell},
\qquad
P_\ell:\mathbb R^{d_\ell}\to\mathbb R^{d_0},
\qquad
P_0=R_0=I_{d_0}.
\]
For any scale-wise choices
$
A_\ell\in\mathcal A_{\mathfrak S^{(\ell)}},
$
the finest-space interaction
$
A^{\mathrm{ms}}
=
\sum_{\ell=0}^L
P_\ell A_\ell R_\ell
$
belongs, by
Definition~\ref{def:operators-multiscale-interaction-family}, to
$
\mathcal A^{\mathrm{ms}}.
$

Likewise, if
$
L_\ell\in\mathcal L_{\mathfrak S^{(\ell)}},
$
then
\[
\widehat L^{\mathrm{ms}}
=
\sum_{\ell=0}^L
P_\ell L_\ell R_\ell
\in
\widehat{\mathcal L}^{\mathrm{ms}}.
\]
This transferred matrix is not automatically a finest-space generator.
If
$
R_\ell\mathbf 1_{d_0}
=
\mathbf 1_{d_\ell}
\qquad
\text{for every }\ell,
$
then Proposition~\ref{prop:operators-constant-compatibility-transferred-generators}
gives
$
\widehat L^{\mathrm{ms}}\mathbf 1_{d_0}=0.
$
Under this compatibility condition, membership in the actual multiscale
generator family reduces to checking the remaining Metzler condition
$
\widehat L^{\mathrm{ms}}_{ij}\ge 0
\qquad
(i\neq j).
$
Only after these finest-space generator conditions hold may the matrix be used
in the multiscale propagation family of
Definition~\ref{def:operators-multiscale-propagation-family}.

\begin{remark}[Interpretation]
\label{rem:examples-multiscale-interpretation}
The fine-scale terms encode short-range geometric detail, whereas coarse-scale
terms provide more direct routes for large-scale interaction. The important
point is that valid generator structure must be checked after transfer to the
finest space; it is not inherited automatically from the scale-wise
generators.
\end{remark}

\paragraph{Hodge-structured edge operators.}
We finally move beyond scalar site-to-site transport to operators induced by
incidence and topological structure.

Let \(\mathcal K\) be a finite cell or simplicial complex with vertex set
\(V\), edge set \(E\), and, when present, face set \(F\). Let
$
B_1\in\mathbb R^{|V|\times|E|}
$
be the vertex--edge incidence matrix and
$
B_2\in\mathbb R^{|E|\times|F|}
$
the edge--face incidence matrix. Define the edge-space Hodge Laplacian by
\[
\Delta^{\mathrm H}_1
:=
B_1^\top B_1+B_2B_2^\top.
\]

\begin{definition}[Hodge correction operator]
\label{def:examples-hodge-correction-matrix}
For \(\alpha>0\), define
$
A^{\mathrm H}
:=
-\alpha\Delta^{\mathrm H}_1.
$
\end{definition}

This operator acts on edge-space signals and regularizes disagreement with the
incidence structure of the underlying complex. It should not be interpreted as
a scalar positivity-preserving transport generator.

\begin{proposition}[Structural meaning and step-size dependence]
\label{prop:examples-hodge-structural-meaning}
The matrix \(\Delta^{\mathrm H}_1\) is symmetric positive semidefinite.
Consequently,
$
A^{\mathrm H}
=
-\alpha\Delta^{\mathrm H}_1
$
is symmetric negative semidefinite and has the same kernel as
\(\Delta^{\mathrm H}_1\).

If
$
\lambda_{\max}(\Delta^{\mathrm H}_1)>0,
$
then the residual operator
$
I-\alpha\Delta^{\mathrm H}_1
$
is nonexpansive in Euclidean norm if and only if
$
0\le\alpha
\le
\frac{2}{\lambda_{\max}(\Delta^{\mathrm H}_1)}.
$
By contrast,
$
e^{-\tau\Delta^{\mathrm H}_1}
$
is nonexpansive for every \(\tau\ge0\), preserves the harmonic subspace, and
converges as \(\tau\to\infty\) to the orthogonal projection onto
\(\ker(\Delta^{\mathrm H}_1)\).
\end{proposition}

\begin{remark}[Relation to the GIOF extensions]
\label{rem:examples-hodge-lifted-operator}
The Hodge construction illustrates how the geometry-induced philosophy can be
extended from pairwise site geometry to incidence-based topological geometry.
However, membership of \(A^{\mathrm H}\) in a particular fixed GIOF
interaction or lifted interaction family is not automatic: it requires the
chosen descriptor and lifted dictionary to represent the corresponding
incidence-induced couplings. Thus the example is a structured topological
realization, not an unconditional family-membership claim.
\end{remark}

\paragraph{Sheaf block operators.}
Let \(\mathcal F\) be a cellular sheaf on a finite graph, with Euclidean inner
products on all stalks. Assume for simplicity that every vertex stalk has the
same dimension,
$
\mathcal F(v)\cong\mathbb R^c.
$
For every incident vertex--edge pair, let
$
R_{ve}:\mathcal F(v)\to\mathcal F(e)
$
be the corresponding restriction map.

For an oriented edge \(e:u\to v\), define the degree-zero coboundary by
\[
(\delta_0^{\mathcal F}h)_e
=
R_{ve}h_v-R_{ue}h_u.
\]
The corresponding sheaf Laplacian is
$
\Delta^{\mathrm{sh}}
:=
(\delta_0^{\mathcal F})^\top
\delta_0^{\mathcal F}.
$
This is the vertex-to-edge restriction convention used in
\citep{barbero2022sheaf}.

\begin{definition}[Sheaf block correction operator]
\label{def:examples-sheaf-block-interaction-matrix}
Let
$
D\in\mathbb R^{c\times c}
$
be a shared within-stalk linear map. For \(\alpha>0\), define
\[
A^{\mathrm{sh}}
:=
-\alpha\Delta^{\mathrm{sh}}
+
I_{|V|}\otimes D.
\]
\end{definition}

Membership of \(A^{\mathrm{sh}}\) in a particular fixed block-valued GIOF
family
$
\mathcal A^{\mathrm{blk}}_{\mathfrak S,\mathcal G}
$
requires the chosen lifted dictionary to represent the sheaf-induced blocks.
Such membership is therefore a representability condition rather than an
automatic consequence of the sheaf construction.

\begin{proposition}[Structural meaning of sheaf block operators]
\label{prop:examples-sheaf-structural-meaning}
The sheaf Laplacian is symmetric positive semidefinite and satisfies
\[
h^\top\Delta^{\mathrm{sh}}h
=
\sum_{e:u\to v}
\|R_{ve}h_v-R_{ue}h_u\|_2^2.
\]
Hence
$
\ker(\Delta^{\mathrm{sh}})
=
\ker(\delta_0^{\mathcal F}),
$
which is the space of globally consistent sections.

Consequently,
$
-\alpha\Delta^{\mathrm{sh}}
$
is symmetric negative semidefinite. No corresponding definiteness statement
holds for arbitrary \(D\). If
$
D=D^\top\preceq0,
$
then
$
A^{\mathrm{sh}}\preceq0
$
and \(A^{\mathrm{sh}}\) is symmetric.

These spectral conclusions do not imply that the flattened matrix is Metzler
or positivity preserving; such properties would require the additional
ordered-feature assumptions of
Definition~\ref{def:operators-block-valued-generator-family}.
\end{proposition}

\begin{remark}[Why the Hodge and sheaf examples matter]
\label{rem:examples-why-hodge-sheaf-matter}
The preceding scalar examples organize information flow primarily through
admissibility and path structure. Hodge and sheaf operators show how the same
design principle can incorporate incidence, orientation, circulation, and
cross-stalk consistency. They should therefore be viewed as structured
topological extensions whose exact membership in a fixed GIOF dictionary must
be verified at the lifted-operator level.
\end{remark}

Taken together, these examples separate three levels of interpretation.
Local kernels, stencil matrices, and graph aggregation give direct
single-scale realizations of descriptor-induced interaction families;
row-balanced local, directional, and barrier-aware kernels instantiate the
generator and propagation constructions; multiscale and topological examples
use the corresponding extensions already defined in the preceding appendices.
Attention-style mixing is deliberately kept separate because context-dependent
scores and nonlinear row normalization need not remain inside a fixed linear
dictionary.

The examples therefore add no new abstract operator assumptions. Their role is
to make explicit how concrete geometric structure is compiled into the
families on which the projection and learning analysis of the next appendix
operates.

\section{Projection and Learning Details for Geometry-Induced Families}
\label{app:learning_appendix}

This appendix supplies the technical results used by the approximation and
learning interpretation in Section~\ref{sec:learning}.
The Frobenius projection considered there is an auxiliary structural reference
problem, not a claim that task-driven training literally performs metric
projection.
We first record the Hilbert-space and coefficient geometry of structured
projection, then treat data-weighted recovery and task alignment, extend the
projection principle to operator fields and residual maps, and finally state
which parts carry over to other closed convex geometry-induced families.
Throughout, matrix spaces are equipped with the Frobenius inner product and
norm.

\subsection{Projection geometry}

Because \(\mathcal A_{\mathfrak S}\) is a nonempty closed convex subset of
the Euclidean space \(\mathbb R^{d\times d}\) by
Proposition~\ref{prop:operators-interaction-family-structure}, its metric
projection is single-valued and satisfies the standard Hilbert-space
projection identities.

\begin{proposition}[Nonexpansiveness of the metric projection]
\label{prop:appendix_projection_nonexpansive}
For all \(B_1,B_2\in\mathbb R^{d\times d}\),
\[
\bigl\|
\Pi_{\mathcal A_{\mathfrak S}}(B_1)
-
\Pi_{\mathcal A_{\mathfrak S}}(B_2)
\bigr\|_F
\le
\|B_1-B_2\|_F.
\]
In particular,
\(\Pi_{\mathcal A_{\mathfrak S}}\) is continuous.
\end{proposition}

Nonexpansiveness controls how the best structured target changes under
perturbations of the unrestricted target.
The corresponding first-order geometry is described by the projection
variational inequality.

\begin{proposition}[Projection variational inequality and unconstrained orthogonality]
\label{prop:appendix_projection_variational}
Let
\[
A^\star
=
\Pi_{\mathcal A_{\mathfrak S}}(B).
\]
Then, for every \(A\in\mathcal A_{\mathfrak S}\),
\[
\langle
B-A^\star,\,
A-A^\star
\rangle_F
\le 0.
\]
If \(\mathcal R_+=\varnothing\), so that
\(\mathcal A_{\mathfrak S}\) is a linear subspace, this strengthens to
$
\langle B-A^\star,\,A\rangle_F
=
0
\qquad
\text{for every }A\in\mathcal A_{\mathfrak S}.
$
Thus, in the unconstrained case, the projection residual is orthogonal to
the entire structured subspace.
\end{proposition}

Expanding
\(B-A=(B-A^\star)-(A-A^\star)\)
turns the preceding variational statement into the approximation
decomposition used in Section~\ref{sec:learning}.

\begin{corollary}[Pythagorean relations for structured projection]
\label{cor:appendix_pythagorean_projection}
Let
$
A^\star
=
\Pi_{\mathcal A_{\mathfrak S}}(B).
$
Then every \(A\in\mathcal A_{\mathfrak S}\) satisfies
\[
\|B-A\|_F^2
\ge
\|B-A^\star\|_F^2
+
\|A-A^\star\|_F^2.
\]
If \(\mathcal R_+=\varnothing\), the inequality becomes the exact identity
\[
\|B-A\|_F^2
=
\|B-A^\star\|_F^2
+
\|A-A^\star\|_F^2.
\]
The first term is the irreducible structured approximation error.
When \(A\) is a learned operator, the second term may be interpreted as
its deviation from the best structured target.
\end{corollary}

\subsection{Coefficient representation and constrained projection}

We now translate the operator-level projection into the coefficient
coordinates introduced in Section~\ref{sec:learning}.
Let \(k=R+d\), with
\[
Q_r=\Phi_r
\quad (1\le r\le R),
\qquad
Q_{R+i}=E_{ii}
\quad (1\le i\le d),
\]
and define the linear synthesis map
\[
T\alpha
:=
\sum_{a=1}^k \alpha_a Q_a,
\qquad
\alpha=(\theta,\delta)\in\mathbb R^{R+d}.
\]
The feasible coefficient set is
\[
\mathcal C_\alpha
:=
\left\{
(\theta,\delta)\in\mathbb R^{R+d}
\;\middle|\;
\theta_r\ge0
\text{ for }r\in\mathcal R_+
\right\},
\]
so that
$
\mathcal A_{\mathfrak S}
=
T(\mathcal C_\alpha).
$

\begin{theorem}[Coefficient form of structured projection]
\label{thm:appendix_coeff_projection}
Define
\[
G_{ab}
:=
\langle Q_a,Q_b\rangle_F,
\qquad
b_a(B)
:=
\langle B,Q_a\rangle_F.
\]
Then every minimizer \(\alpha^\star\) of
\[
\min_{\alpha\in\mathcal C_\alpha}
\left(
\frac12\alpha^\top G\alpha
-
b(B)^\top\alpha
\right)
\]
satisfies
$
T\alpha^\star
=
\Pi_{\mathcal A_{\mathfrak S}}(B).
$
Conversely, every feasible coefficient vector representing the projected
matrix minimizes this quadratic program.
Hence the projected matrix is unique even when its coefficient
representation is not.

If \(\mathcal R_+=\varnothing\), the minimizers are exactly the solutions
of the normal equations
$
G\alpha^\star
=
b(B).
$
If \(Q_1,\ldots,Q_k\) are linearly independent, then
\(G\succ0\), and the minimizing coefficient vector is unique in both the
constrained and unconstrained cases.
\end{theorem}

Indeed,
\[
\frac12\|B-T\alpha\|_F^2
=
\frac12\|B\|_F^2
+
\frac12\alpha^\top G\alpha
-
b(B)^\top\alpha,
\]
so the coefficient problem differs from the Frobenius projection objective
only by the constant \(\frac12\|B\|_F^2\).

When the generating family is nonredundant, the constrained quadratic
program also admits the usual coordinatewise KKT characterization.

\begin{proposition}[KKT characterization of constrained projection]
\label{prop:appendix_kkt_constrained}
Assume that \(Q_1,\dots,Q_k\) are linearly independent.
A feasible vector
\(\alpha^\star\in\mathcal C_\alpha\)
is the unique minimizer if and only if there exist multipliers
$
\lambda_r\ge0,
\qquad
r\in\mathcal R_+,
$
such that
\begin{align*}
G\alpha^\star-b(B)
-
\sum_{r\in\mathcal R_+}\lambda_r e_r
&=0,
&&\text{(stationarity)},\\
\alpha_r^\star
&\ge0,
&&r\in\mathcal R_+,
\qquad\text{(primal feasibility)},\\
\lambda_r
&\ge0,
&&r\in\mathcal R_+,
\qquad\text{(dual feasibility)},\\
\lambda_r\alpha_r^\star
&=0,
&&r\in\mathcal R_+,
\qquad\text{(complementary slackness)},
\end{align*}
where \(e_r\) is the \(r\)-th standard basis vector of
\(\mathbb R^k\).
When \(\mathcal R_+=\varnothing\), stationarity reduces to the normal
equations in Theorem~\ref{thm:appendix_coeff_projection}.
\end{proposition}

Thus operator uniqueness, coefficient identifiability, and active
coefficient constraints are distinct issues.
Metric projection fixes the structured operator uniquely, whereas
basis independence is additionally required to identify a unique
coefficient vector.

\subsection{Data-weighted recovery and task alignment}
\label{app:learning-task-alignment}

The Frobenius metric is a structural reference. When an operator is observed only through its action on data, the representation covariance changes the relevant quadratic form.

\begin{proposition}[Data-weighted operator approximation]
\label{prop:learning-data-weighted-projection}
Let \(h\) have finite second moment and set
$
\Sigma:=\mathbb E[hh^\top].
$
For fixed \(A,B\in\mathbb R^{d\times d}\),
\[
\mathbb E\|(B-A)h\|_2^2
=
\operatorname{tr}\!\bigl(
(B-A)\Sigma(B-A)^\top
\bigr).
\]
If \(\Sigma\succ0\), this objective has a unique minimizer over
\(\mathcal A_{\mathfrak S}\); if \(\Sigma\) is singular, uniqueness need not hold. In coefficient form,
\[
(\Gamma_\Sigma)_{ab}
=
\operatorname{tr}(Q_a\Sigma Q_b^\top),
\qquad
(b_\Sigma)_a
=
\operatorname{tr}(B\Sigma Q_a^\top).
\]
\end{proposition}

For
$
A^\star
=
\Pi_{\mathcal A_{\mathfrak S}}(B),
$
the descriptor-induced irreducible approximation bias is
\[
\operatorname{dist}(B,\mathcal A_{\mathfrak S})
=
\|B-A^\star\|_F.
\]
When
\(\{Q_a\}_{a=1}^k\)
is linearly independent, the coefficient representation of
\(A^\star\)
is unique. Irrespective of coefficient uniqueness,
\(A^\star\)
remains in the prescribed operator family and therefore obeys its support and channel-sign restrictions.
Corollary~\ref{cor:appendix_pythagorean_projection}
gives the corresponding Pythagorean decomposition.

Approximation bias alone does not determine whether finite propagation is useful for a prediction task; this additionally requires alignment between the generator spectrum and the signal--noise decomposition.

\begin{proposition}[A task-alignment criterion for useful propagation]
\label{prop:learning-spectral-alignment}
Let
$
L=L^\top\in\mathcal L_{\mathfrak S}
$
be fixed, with orthonormal eigenbasis
\(\{u_k\}_{k=1}^d\)
satisfying
\[
Lu_k=-\nu_k u_k,
\qquad
\nu_k\ge0.
\]
Suppose
$
h=s+\varepsilon,
\qquad
\mathbb E[\varepsilon]=0,
\qquad
\mathbb E[\varepsilon\varepsilon^\top]=\sigma^2I,
$
where \(s\) is deterministic, and write
$
s_k=u_k^\top s.
$
Then
\[
\mathbb E\|e^{\tau L}h-s\|_2^2
=
\sum_{k=1}^d
(1-e^{-\tau\nu_k})^2s_k^2
+
\sigma^2
\sum_{k=1}^d
e^{-2\tau\nu_k}.
\]
Hence propagation has strictly lower mean-squared error than the unprocessed observation if and only if
\[
\sum_{k=1}^d
(1-e^{-\tau\nu_k})^2s_k^2
<
\sigma^2
\sum_{k=1}^d
(1-e^{-2\tau\nu_k}).
\]
\end{proposition}

The proposition is a fixed-operator denoising criterion: propagation is beneficial only when it suppresses nuisance-dominated modes without removing too much task-relevant signal. It is not a domain-generalization guarantee and does not automatically extend to self-conditioned propagation. Together with Proposition~\ref{prop:learning-probe-identifiability}, it separates structural approximation, observational recoverability, and task alignment.

\subsection{Operator fields and residual approximation}

The preceding results concern a single target matrix.
We next apply the same fixed-set projection pointwise to an
input-dependent target field, which is the idealized setting summarized
in Section~\ref{sec:learning}.

\begin{theorem}[Pointwise projection for operator fields]
\label{thm:appendix_pointwise_projection}
Let \(K\subset\mathbb R^d\) be compact, let \(\mu\) be a probability
measure on \(K\), and assume
$
B
\in
L^2\!\left(
K,\mu;\mathbb R^{d\times d}
\right).
$
Define
\[
\mathcal F_{\mathfrak S}
:=
\left\{
A\in
L^2\!\left(
K,\mu;\mathbb R^{d\times d}
\right)
\;\middle|\;
A(x)\in\mathcal A_{\mathfrak S}
\text{ for }\mu\text{-almost every }x
\right\}.
\]
Then \(\mathcal F_{\mathfrak S}\) is a nonempty closed convex subset of
the Hilbert space
\(L^2(K,\mu;\mathbb R^{d\times d})\), and
\[
\min_{A\in\mathcal F_{\mathfrak S}}
\int_K
\|B(x)-A(x)\|_F^2
\,d\mu(x)
\]
admits a unique minimizer up to \(\mu\)-almost-everywhere equality,
given by
$
A^\star(x)
=
\Pi_{\mathcal A_{\mathfrak S}}(B(x))
\qquad
\text{for }\mu\text{-almost every }x.
$
If \(B\) admits a continuous representative, then \(A^\star\) admits
the continuous representative
$
x
\longmapsto
\Pi_{\mathcal A_{\mathfrak S}}(B(x)).
$
\end{theorem}

The continuity conclusion follows directly from
Proposition~\ref{prop:appendix_projection_nonexpansive}.
Moreover, since
\(0\in\mathcal A_{\mathfrak S}\),
nonexpansiveness gives
\[
\|A^\star(x)\|_F
=
\bigl\|
\Pi_{\mathcal A_{\mathfrak S}}(B(x))
-
\Pi_{\mathcal A_{\mathfrak S}}(0)
\bigr\|_F
\le
\|B(x)\|_F,
\]
which makes the \(L^2\) admissibility of \(A^\star\) explicit.

The field-level operator error immediately controls the induced residual-map
error on bounded representation domains.

\begin{corollary}[Residual approximation bound]
\label{cor:appendix_integrated_residual_bound}
For
\(A,B\in L^2(K,\mu;\mathbb R^{d\times d})\),
define
\[
u_B(x):=B(x)x,
\qquad
u_A(x):=A(x)x,
\qquad
R_K:=\sup_{x\in K}\|x\|_2.
\]
Then, for \(\mu\)-almost every \(x\),
\[
\|u_B(x)-u_A(x)\|_2
\le
\|B(x)-A(x)\|_F\,\|x\|_2.
\]
Consequently,
\[
\int_K
\|u_B(x)-u_A(x)\|_2^2
\,d\mu(x)
\le
R_K^2
\int_K
\|B(x)-A(x)\|_F^2
\,d\mu(x).
\]
If \(A\) and \(B\) are continuous on \(K\), then the pointwise inequality
holds everywhere and also yields
\[
\sup_{x\in K}
\|u_B(x)-u_A(x)\|_2
\le
R_K
\sup_{x\in K}
\|B(x)-A(x)\|_F.
\]
\end{corollary}

\begin{remark}[Relation to practical context-dependent training]
\label{rem:appendix_pointwise_projection_vs_practical_training}
Theorem~\ref{thm:appendix_pointwise_projection} optimizes over the full
field class \(\mathcal F_{\mathfrak S}\), so it is a population-level
structural idealization.
In the trainable layers of Section~\ref{sec:layers}, the selector
$
c\mapsto A(c)
$
is realized by shared parameters.
Consequently, different contexts are coupled through the common
parameterization, and practical optimization need not decouple pointwise.

The notation \(A(x)x\) in
Corollary~\ref{cor:appendix_integrated_residual_bound} corresponds to the
self-conditioned case \(c=h=x\).
For a separate context \(c\) and acted-on representation \(h\), the same
elementary estimate becomes
\[
\|(B(c)-A(c))h\|_2
\le
\|B(c)-A(c)\|_F\,\|h\|_2.
\]
\end{remark}

\subsection{Extension to other geometry-induced families}

The preceding metric-projection argument uses only nonemptiness, closedness,
and convexity of the feasible operator family.
It therefore extends beyond the scalar single-scale interaction family
whenever these properties have already been established.
Coefficient identifiability, by contrast, remains dependent on the chosen
operator basis.

\begin{proposition}[Projection principle for other convex geometry-induced families]
\label{prop:appendix_projection_other_families}
Let
$
\mathcal U
\subseteq
\mathbb R^{N\times N}
$
be any nonempty closed convex geometry-induced family.
Then every
\(B\in\mathbb R^{N\times N}\)
admits a unique metric projection
\[
\Pi_{\mathcal U}(B)
=
\arg\min_{A\in\mathcal U}
\|B-A\|_F^2.
\]
In particular, this applies, in their corresponding ambient matrix spaces,
to
$
\mathcal L_{\mathfrak S},
\qquad
\mathcal L_{\mathfrak S}^{\mathrm{bal}},
\qquad
\mathcal A^{\mathrm{blk}}_{\mathfrak S,\mathcal G},
\qquad
\mathcal L^{\mathrm{blk}}_{\mathfrak S,\mathcal G},
\qquad
\mathcal A^{\mathrm{ms}},
\qquad
\widehat{\mathcal L}^{\mathrm{ms}},
\qquad
\mathcal L^{\mathrm{ms}},
$
with the block-valued generator interpretation restricted to the
ordered-feature setting described in
Definition~\ref{def:operators-block-valued-generator-family}.
\end{proposition}

For \(\mathcal L_{\mathfrak S}\), closed convexity is established in
Proposition~\ref{prop:operators-generator-family-structure}.
For
$
\mathcal L_{\mathfrak S}^{\mathrm{bal}}
=
\mathcal L_{\mathfrak S;\mathbf1,\mathbf1},
$
it follows from
Proposition~\ref{prop:operators-feasible-compilation}.
The block-valued interaction case is covered by
Proposition~\ref{prop:operators-block-valued-interaction-structure},
while
\(\mathcal L^{\mathrm{blk}}_{\mathfrak S,\mathcal G}\)
is obtained by intersecting that closed convex cone with linear equality
constraints and coordinatewise half-spaces.
The multiscale cases follow from
Proposition~\ref{prop:operators-basic-structure-multiscale-families}.
All of these families contain the zero operator and are therefore nonempty.

The same projection principle also applies to nonempty normalized generator
slices.
For example, when
\(\mathcal L_{\mathfrak S}\neq\{0\}\),
$
\mathcal L_{\mathfrak S}^{(1)}
=
\left\{
L\in\mathcal L_{\mathfrak S}
\;\middle|\;
\rho(L)=1
\right\}
$
is closed and convex.
Moreover, because its off-diagonal entries are nonnegative and satisfy
$
\sum_{i=1}^d\sum_{j\neq i}L_{ij}=d,
$
it is bounded and hence compact.
This justifies, at the operator level, the projection-based normalized
generator parameterization discussed in
Remark~\ref{rem:layers-generator-selector-implementation}.

\begin{remark}[Lifted-basis identifiability]
\label{rem:appendix_lifted_identifiability}
Projection-based approximation extends to the preceding closed convex
families without requiring unique coefficients.
Coefficient interpretability does not.
In the block-valued setting, it requires linear independence of the
realized operators
$
\Phi_r\otimes G_s
$
together with a nonredundant basis for the within-site blocks.
In the multiscale setting, it requires linear independence after the
scale-wise basis operators have been transferred to the finest space.
Thus identifiability must be checked in the realized lifted operator space;
it cannot be inferred automatically from scalar single-scale independence.
\end{remark}

\begin{remark}[Why propagation families are treated indirectly]
\label{rem:appendix_propagation_indirect}
The propagation family
\(\mathcal P_{\mathfrak S}\)
is obtained through the nonlinear matrix exponential.
Consequently, the closed-convex projection theorem above does not
automatically apply to
\(\mathcal P_{\mathfrak S}\)
itself.

When a target is naturally specified at the infinitesimal level, one may
approximate or learn its generator inside
\(\mathcal L_{\mathfrak S}\)
and then exponentiate.
In the trainable propagation layer of Section~\ref{sec:layers}, one instead
learns a structured normalized generator \(L(c)\) together with the
propagation magnitude \(\tau(c)\), and uses
$
P(c)
=
e^{\tau(c)L(c)}.
$
Neither construction should be interpreted as asserting that exponentiating
a generator-level metric projection solves a Frobenius projection problem
directly over
\(\mathcal P_{\mathfrak S}\).
\end{remark}

\section{Experimental Details and Additional Results}
\label{app:exp_details}

This section specifies all experiments in Section~\ref{sec:experiments}.
The protocol separates numerical checks of operator statements from
empirical tests of reconstruction quality. 

\subsection{Controlled checks of the theoretical mechanisms}
\label{app:exp_controlled}

\paragraph{Common construction.}
For panels (a,b), sample $d=64$ sites uniformly in $[0,1]^2$ and
symmetrize the eight-nearest-neighbor graph. The off-diagonal dictionary
contains two Gaussian distance kernels, with bandwidths $0.15,0.35$,
each multiplied by $1,\cos\vartheta_{ij},\sin\vartheta_{ij}$, or
$\cos(2\vartheta_{ij})$; $\vartheta_{ij}$ is the angle of the
receiver-minus-source displacement. Use $\mathcal R_+=\varnothing$.
Orthonormalize the eight-channel Frobenius span, append the 64 diagonal
coordinate matrices, and verify rank 72. Regenerate a rank-deficient
draw. Orthonormalization is confined to this unconstrained experiment.
Draw standard Gaussian coefficients and rescale the resulting
$A^\star$ to $\|A^\star\|_F=1$.

\paragraph{(a) Recovery and informative probes.}
Generate $y_n=A^\star h_n+\varepsilon_n$, where
$h_n\sim\mathcal N(0,I)$ and
$\varepsilon_n\sim\mathcal N(0,0.05^2I)$, using nested fitting sets
$N\in\{2,4,8,16,32,64,128,256\}$.
Compare six estimators: GIOF least squares, GIOF ridge,
Shuffled ridge, Random-span ridge, Free-Masked ridge, and Dense ridge.
Shuffled permutes complete channel vectors across admissible edges;
Random-span uses eight independent Gaussian masked matrices.
Both retain the same free diagonal and eight-dimensional off-diagonal
budget. Record their numerical ranks.
Free-Masked learns every admissible matrix entry independently.
Dense ridge learns all entries. Ridge penalizes $\|A\|_F^2$,
with strength selected from $\{10^{-6},10^{-4},10^{-2},1\}$
by generalized cross-validation on the fitting observations only;
no additional labeled probes are supplied. Least squares uses the
minimum-norm solution when needed.

Plot $\|\widehat A-A^\star\|_F^2$ and record the rank and smallest
eigenvalue of the probe Gram matrix $\Gamma$.
The exact check repeats noiseless recovery and uses
$\Sigma=\operatorname{Diag}(2I_{32},0_{32})$ as a nonidentifiable
control: the last 32 diagonal coefficients cannot be observed.
Do not identify a recovery threshold merely with $R+d$, since one
vector probe provides multiple scalar observations.

\paragraph{(b) Compression benefit and mismatch cost.}
Project a random matrix in $\mathcal M_{\mathcal E}$ onto the
Frobenius-orthogonal complement of the structured span and normalize
the nonzero residual to obtain $E_\perp$.
Set $B_\eta=A^\star+\eta E_\perp$ without renormalizing, for
$\eta\in\{0,0.05,0.1,0.2,0.4,0.8,1.6\}$.
Generate observations using $B_\eta$ and the preceding noise model.
Plot GIOF ridge and Free-Masked ridge at $N\in\{16,64,256\}$:
six curves plus the exact structured approximation floor
$\operatorname{dist}(B_\eta,\mathcal A_{\mathfrak S})^2=\eta^2$.
The floor is a projection identity, not a prediction that GIOF wins
at every mismatch. With isotropic probes the squared Frobenius error
equals noise-free prediction risk. For a supplementary anisotropic
check, use a trace-$d$ positive-definite covariance with condition
number $10^2$ and compare against the $\Sigma$-weighted optimum in
Proposition~\ref{prop:learning-data-weighted-projection}.

\paragraph{(c) Reachability and finite propagation.}
Use two disconnected 32-node bidirectional chains with four edge
templates whose admissible entries are independently uniform on
$[0.1,1]$. Mix them with $\operatorname{Dirichlet}(1,1,1,1)$
coefficients, balance rows, and rescale the generator so that the mean
exit rate $d^{-1}\sum_i(-L_{ii})$ equals one.
For $\nu=1.01\max_i(-L_{ii})$, evaluate 40 values of
$\nu\tau\in[0,8]$ with dense float64 reference exponentials.
For shortest positive-rate path lengths $r\in\{1,2,4\}$, plot
the maximum corresponding entry of $e^{\tau L}$ and the bound
$\Pr\{\operatorname{Poisson}(\nu\tau)\ge r\}$.
This gives six curves. Separately record the maximum entry between
disconnected components. Strict positivity uses positive-rate paths,
not merely edges allowed by a mask.

\paragraph{(d) Task alignment and excessive smoothing.}
Use the symmetric negative Laplacian of a connected 64-node cycle,
normalized to $\rho(L)=1$, with $Lu_k=-\nu_k u_k$ and eigenvalues
ordered increasingly in $\nu_k$.
Allocate signal energy $\|s\|_2^2=d$ equally to modes
2--9, 25--32, or 57--64, and cross these settings with
$\sigma\in\{0.2,0.5\}$.
For each of the six settings, compare 256 Gaussian-noise realizations
with the exact per-site risk
\[
\frac1d\left[
\sum_k(1-e^{-\tau\nu_k})^2s_k^2
+\sigma^2\sum_k e^{-2\tau\nu_k}\right].
\]
Use 40 logarithmically spaced values of
$\tau\nu_d\in[10^{-3},10^2]$, with $\tau=0$ evaluated separately.
Plot the ratio to the unprocessed risk $\sigma^2$, adding the
unit-risk reference. Exact curves and empirical markers share colors.
Retain the constant-mode noise contribution and unfavorable scales.
This identity concerns additive isotropic noise and a fixed symmetric
operator; it is not an imputation or generalization theorem.

\paragraph{Additional exact checks.}
At depths $m\in\{1,2,4,8,16,32\}$, compose independently selected,
generally noncommuting propagators on the common disconnected support,
using independent Dirichlet mixtures and scales uniform on $[0.1,1]$.
Record commutator norms and violations of nonnegativity,
$P\mathbf1=\mathbf1$, $\|P\|_{\infty\to\infty}=1$, and exclusion
of unreachable pairs. Test component indicators as preserved right
modes; a separate symmetric construction also tests conserved left
component sums. On the connected symmetric graph, compare contraction,
$\partial_\tau e^{\tau L}h$, and symmetric row-balanced generator
perturbations with the oversmoothing and gradient-attenuation bounds.
For constrained projection, retain only the two nonnegative radial
channels, constrain their coefficients, and record feasibility and
KKT residuals. 

\paragraph{Observed controlled results.}
The first four controlled outcomes are summarized in
Figure~\ref{fig:giof_theory}. The diagnostics below isolate numerical
conditions that are less visible in the main panels: probe-Gram rank and
conditioning, noiseless identifiability, covariance-weighted projection,
noncommuting composition, oversmoothing, and scale/shape sensitivity.
At $N=256$ under matched geometry, the mean squared Frobenius error of
GIOF ridge is $6.45\times10^{-4}$, compared with
$6.86\times10^{-3}$ for Free-Masked ridge and $5.29\times10^{-2}$
for Dense ridge. Under the largest prescribed mismatch
$\eta=1.6$, however, the GIOF error increases to $2.57$, while the
Free-Masked error remains $6.90\times10^{-3}$. Thus the experiment
exhibits the intended statistical-efficiency--mismatch tradeoff rather
than universal dominance of the structured estimator. The automated
validation suite passes all 148 prespecified numerical checks at their
stated tolerances.

\begin{table}[!htbp]
\centering
\small
\caption{Noiseless probe identifiability. Ranks show the range across
seeds; errors are mean $\pm$ sample standard deviation. Isotropic probes
identify the full 72-dimensional dictionary, whereas the singular
covariance leaves 32 diagonal directions unobserved and reduces the rank
to 40.}
\label{tab:giof_property_identifiability}
\resizebox{\linewidth}{!}{%
\begin{tabular}{lcccc}
\toprule
$N$ & Isotropic rank & Singular rank & Isotropic error & Singular error \\
\midrule
2   & 72--72 & 40--40 & $5.03\times10^{-30} \pm 2.3\times10^{-30}$ & $0.386 \pm 0.20$ \\
4   & 72--72 & 40--40 & $4.79\times10^{-30} \pm 1.5\times10^{-30}$ & $0.386 \pm 0.20$ \\
8   & 72--72 & 40--40 & $4.91\times10^{-30} \pm 2.4\times10^{-30}$ & $0.386 \pm 0.20$ \\
16  & 72--72 & 40--40 & $3.45\times10^{-30} \pm 8.8\times10^{-31}$ & $0.386 \pm 0.20$ \\
32  & 72--72 & 40--40 & $3.64\times10^{-30} \pm 1.4\times10^{-30}$ & $0.386 \pm 0.20$ \\
64  & 72--72 & 40--40 & $3.68\times10^{-30} \pm 8.1\times10^{-31}$ & $0.386 \pm 0.20$ \\
128 & 72--72 & 40--40 & $3.63\times10^{-30} \pm 2.5\times10^{-31}$ & $0.386 \pm 0.20$ \\
256 & 72--72 & 40--40 & $4.66\times10^{-30} \pm 1.9\times10^{-30}$ & $0.386 \pm 0.20$ \\
\bottomrule
\end{tabular}}
\end{table}

\begin{table}[!htbp]
\centering
\small
\caption{Maximum numerical residual over all specified seeds and scales.
Bound residuals report only positive violations; zero means that no
violation was observed at the stated precision.}
\label{tab:giof_property_structural}
\resizebox{0.92\linewidth}{!}{%
\begin{tabular}{lccc}
\toprule
Check & Maximum residual & Tolerance & Status \\
\midrule
Mean-exit normalization & $1.11\times10^{-16}$ & $10^{-9}$ & Pass \\
Pathwise Poisson bound & $0$ & $10^{-9}$ & Pass \\
Disconnected leakage & $0$ & $10^{-9}$ & Pass \\
Composition: directed invariants & $1.67\times10^{-15}$ & $10^{-9}$ & Pass \\
Composition: symmetric invariants & $1.33\times10^{-15}$ & $10^{-9}$ & Pass \\
Projection identity & $8.88\times10^{-16}$ & $10^{-9}$ & Pass \\
Weighted normal equations & $8.58\times10^{-16}$ & $10^{-9}$ & Pass \\
Oversmoothing bound & $1.11\times10^{-15}$ & $10^{-9}$ & Pass \\
Scale-gradient bound & $0$ & $10^{-9}$ & Pass \\
Shape-gradient bound & $0$ & $10^{-9}$ & Pass \\
Fr\'echet finite-difference check & $1.22\times10^{-10}$ & $10^{-7}$ & Pass \\
Constrained KKT conditions & $1.87\times10^{-13}$ & $10^{-9}$ & Pass \\
Projection nonexpansiveness & $0$ & $10^{-9}$ & Pass \\
\bottomrule
\end{tabular}}
\end{table}

\begin{figure*}[!t]
\centering
\begin{minipage}[t]{0.245\textwidth}
    \centering
    \includegraphics[width=\linewidth]{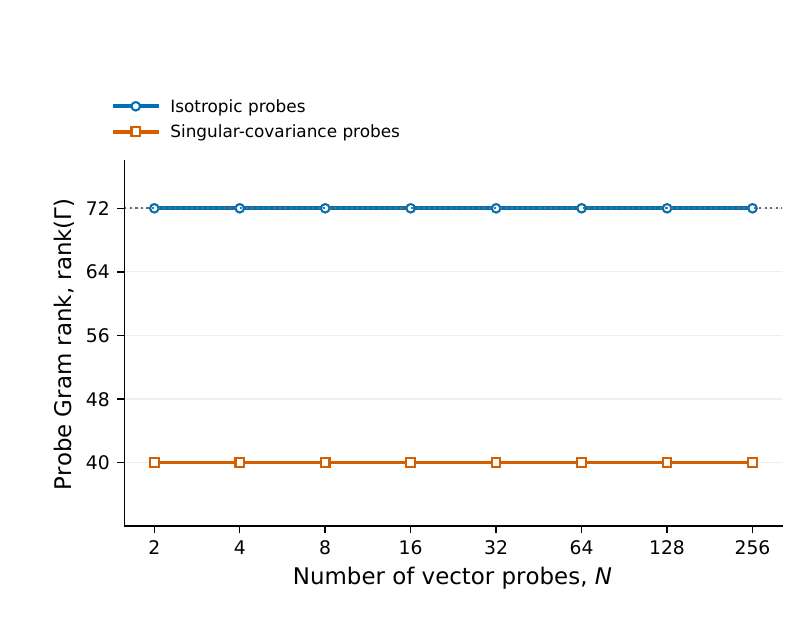}
\end{minipage}\hfill
\begin{minipage}[t]{0.245\textwidth}
    \centering
    \includegraphics[width=\linewidth]{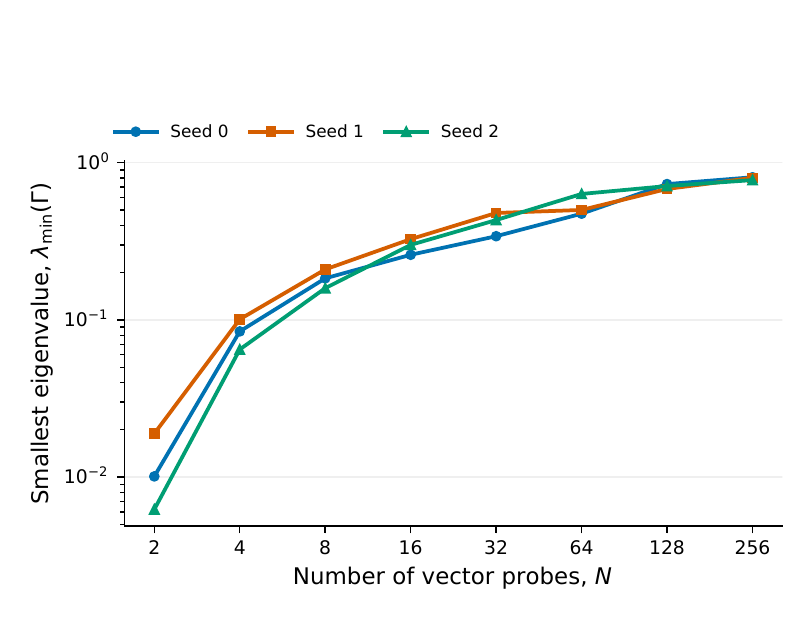}
\end{minipage}\hfill
\begin{minipage}[t]{0.245\textwidth}
    \centering
    \includegraphics[width=\linewidth]{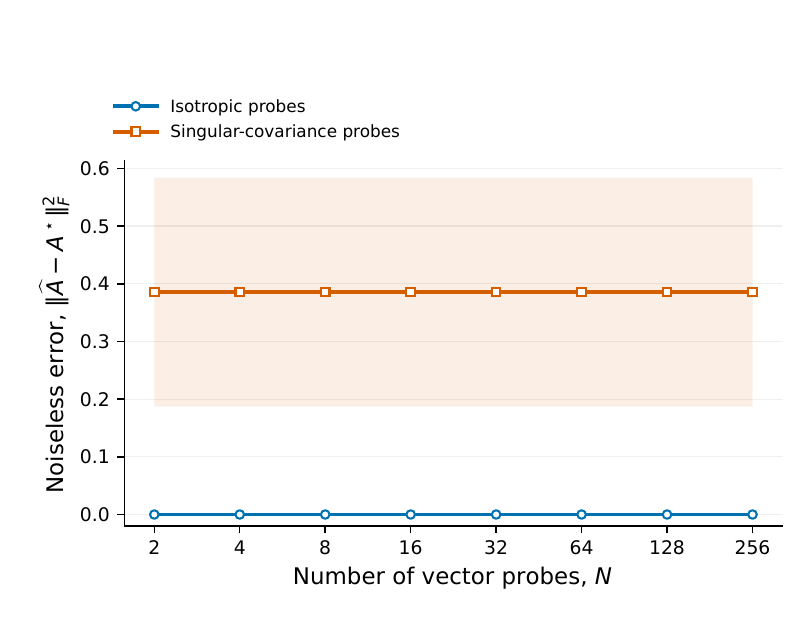}
\end{minipage}\hfill
\begin{minipage}[t]{0.245\textwidth}
    \centering
    \includegraphics[width=\linewidth]{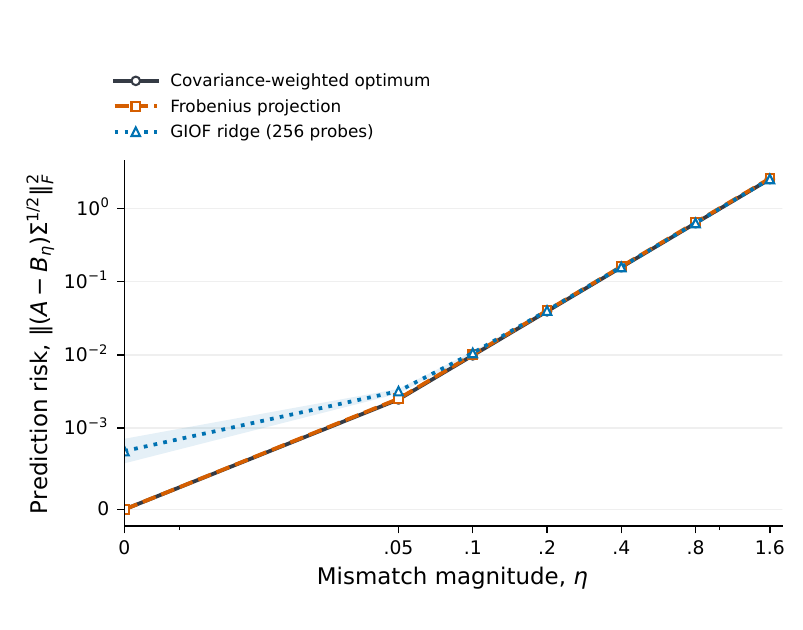}
\end{minipage}

\vspace{-1mm}
\caption{
Additional controlled diagnostics for identifiability and structured projection.
\textbf{(a)} Probe-Gram rank: isotropic probes identify the full structured
dictionary, whereas singular covariance reduces the informative rank.
\textbf{(b)} Probe-Gram conditioning: the smallest positive eigenvalue
quantifies the conditioning of the identifiable probe system.
\textbf{(c)} Noiseless recovery: isotropic probing recovers the structured
operator to numerical precision, whereas singular probing leaves a persistent
unidentifiable component.
\textbf{(d)} Anisotropic projection: under covariance-weighted prediction risk,
the optimal structured approximation need not coincide with the Frobenius
projection.
}
\label{fig:giof_controlled_identifiability}
\end{figure*}

\begin{figure*}[!t]
\centering
\begin{minipage}[t]{0.245\textwidth}
    \centering
    \includegraphics[width=\linewidth]{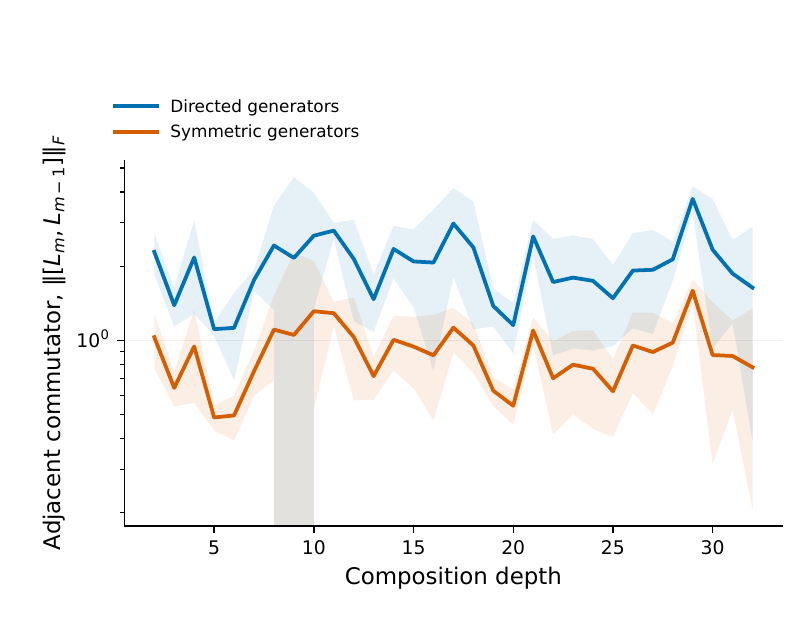}
\end{minipage}\hfill
\begin{minipage}[t]{0.245\textwidth}
    \centering
    \includegraphics[width=\linewidth]{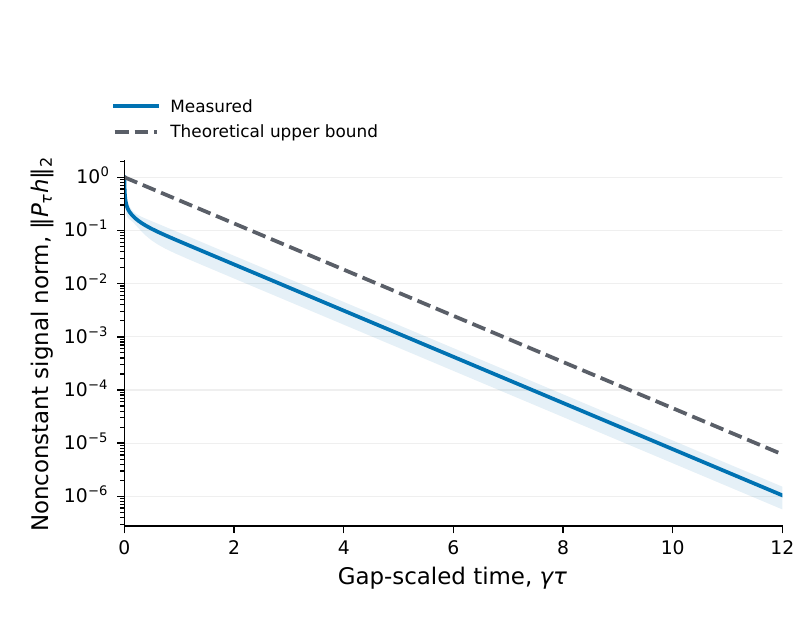}
\end{minipage}\hfill
\begin{minipage}[t]{0.245\textwidth}
    \centering
    \includegraphics[width=\linewidth]{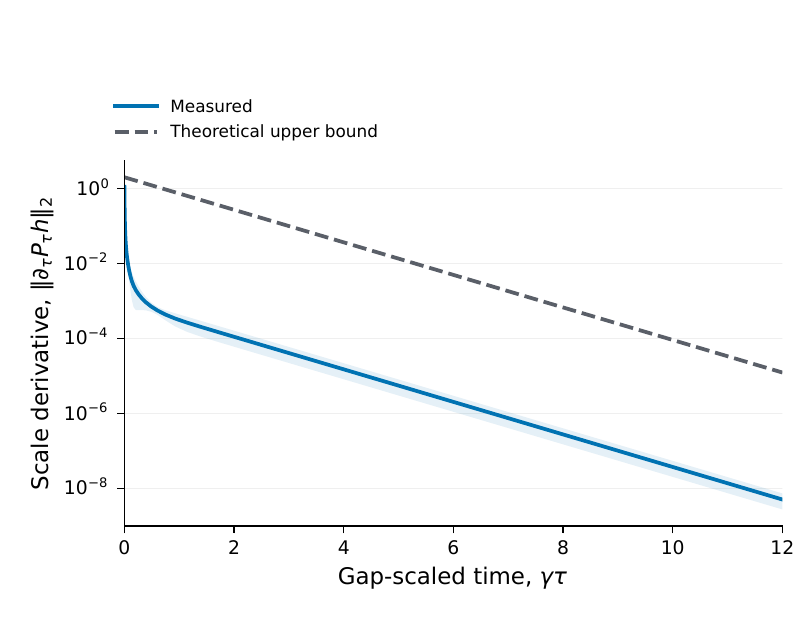}
\end{minipage}\hfill
\begin{minipage}[t]{0.245\textwidth}
    \centering
    \includegraphics[width=\linewidth]{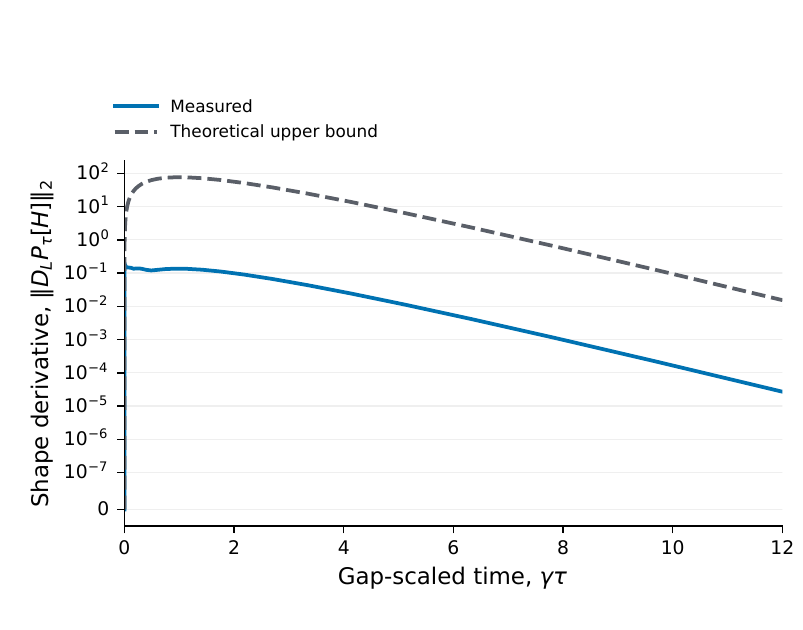}
\end{minipage}

\vspace{-1mm}
\caption{
Additional controlled diagnostics for propagation and sensitivity.
\textbf{(a)} Noncommuting composition: adjacent depth-dependent generators
have nonzero commutators, while the composed propagators retain the required
structural invariants.
\textbf{(b)} Oversmoothing: zero-mean signals contract consistently with the
spectral-gap upper bound.
\textbf{(c)} Scale sensitivity: the propagation derivative with respect to the
diffusion scale remains below the corresponding theoretical bound.
\textbf{(d)} Shape sensitivity: perturbations of the generator obey the
predicted sensitivity bound, with vanishing sensitivity at $\tau=0$.
}
\label{fig:giof_controlled_propagation}
\end{figure*}

\FloatBarrier

\subsection{Real observations, splits, and missingness}
\label{app:exp_real_data}

\paragraph{Task and data.}
We use the public PEMS-BAY and METR-LA traffic-speed releases
\citep{li2017diffusion}, containing 325 and 207 sensors, respectively,
at five-minute resolution. A sample is a real observation window
$Y\in\mathbb R^{d\times T}$ with $T=24$.
Artificial hiding changes availability, not the target values.
This is offline within-window reconstruction: every method may use
available observations on either side of a gap, but receives no
observations outside the window. It is not a forecasting or online
decision-making benchmark.

Split the raw time axis chronologically into 60\%/20\%/20\%
training/validation/test before forming windows. Keep windows within
complete calendar days and split boundaries; use training stride 12
and nonoverlapping validation/test stride 24.
For each weekday, order eligible training days with seed 1729 and
retain the first $\lceil fN_w\rceil$ days, where $N_w$ is the available
count for that weekday and $f\in\{0.2,1\}$.
These nested day sets are identical across methods and model seeds.
Report exact dates and realized day/window counts.

Preserve the original validity mask $O^0$, using the missing-value
conventions of the public benchmark loader accompanying
\citep{cini2021filling}; archive the loader version and its treatment of
stored zeros. Naturally missing values are neither inputs nor scored
targets and are not pre-interpolated. Fit each sensor's normalization
statistics only on valid entries of the selected training days, with
a pooled training-statistic fallback and a $10^{-6}$ standard-deviation
floor. Unselected days contribute no fitted statistics or pretraining.
For an artificial hiding mask $H$, define
\[
O=O^0\odot(1-H),\qquad
\Omega=\{(i,t):O^0_{it}H_{it}=1\}.
\]
Zero-fill unavailable normalized inputs and provide $O$ explicitly.
All losses and evaluation masks exclude naturally unavailable targets.

\paragraph{Three missingness mechanisms.}
Regional outages hide all $T$ readings at $k=\lfloor pd\rfloor$
sensors. Choose a starting sensor uniformly and run breadth-first
search on the fixed undirected support defined below, breaking
same-depth ties with a seeded random ordering. Hide the first $k$
visited sensors. If a connected component is exhausted, restart at a
uniformly chosen unvisited sensor until $k$ sensors have been selected.
Thus each region is connected until a component is exhausted; the
procedure also handles disconnected sensors without deleting them.
Use prefixes of the same traversal when varying $p$, yielding nested
outage sets. Mask construction never uses signal values or model errors.

Independent sensor outages instead choose $k$ sensors uniformly
without replacement and hide their complete windows. Point missingness
hides each originally valid entry independently with probability $p$.
During training, sample the mechanism uniformly from these three cases
and sample $p$ uniformly from $\{0.1,0.3,0.5\}$.
Use the same window order and pre-generated artificial masks for
corresponding updates of every method within a seed.
Choose checkpoints by the equally weighted mean of the nine
mechanism--rate validation MAEs, using fixed validation masks.
All three mechanisms are therefore represented during selection.

Table~\ref{tab:giof_real_main} evaluates regional outages at $p=0.3$.
Figure~\ref{fig:giof_real_curves} uses
$p\in\{0.1,0.2,0.3,0.4,0.5,0.6\}$ and the selected $f=0.2$
checkpoints without retraining. For each condition use ten independent
test-mask realizations shared across methods.
Report independent-sensor and point-missingness results separately;
do not combine the three test tasks into a favorable weighted score.

\paragraph{Metrics and uncertainty.}
Compute MAE and RMSE in original speed units over $\Omega$ only.
Sum absolute/squared errors and target counts across all test windows
before forming a metric. Record target counts and never score copied observed
entries. Report paired per-seed differences against every comparator,
including negative differences. 

\subsection{Shared architecture and geometry-matched alternatives}
\label{app:exp_architecture}

\paragraph{Common reconstruction network.}
For GIOF and the five spatial-module baselines, a shared sensorwise
two-layer GELU encoder of width 64 maps the masked window, its mask,
an eight-dimensional learned sensor embedding, and four calendar
features to $Z\in\mathbb R^{d\times64}$.
The calendar features are sine/cosine encodings of the window-start
time of day and day of week. A shared $64\to64\to T$ GELU decoder
$D$ returns the reconstructed window. Spatial mixing uses
\[
F=ZW_{\rm in},\qquad
\widehat Y=D\!\left(Z+
[\mathcal S(F,c)-F]W_{\rm out}\right),
\quad
W_{\rm in}\in\mathbb R^{64\times16},\quad
W_{\rm out}\in\mathbb R^{16\times64}.
\]
The context $c$ concatenates the node average of $Z$ with the four
calendar features. Initialize $W_{\rm out}=0$ and use identical
initial encoder, embedding, decoder, and bottleneck weights across
common-network methods within each seed. Train all weights jointly.
The nodewise parent sets $\mathcal S(F,c)=F$ and is used only as
an ablation. The decoder performs no further cross-sensor mixing.

\paragraph{Fixed geometry.}
Align the released adjacency and sensor-coordinate files with the
observation-column identifiers. Let $A^{\rm rel}$ be the released
nonnegative affinity and use off-diagonal support
\[
M_{ij}=\mathbf1\{A^{\rm rel}_{ij}+A^{\rm rel}_{ji}>0\},
\quad i\ne j,\qquad M_{ii}=0.
\]
This symmetrized support describes proximity, not physical traffic
direction. All graph-based methods receive this support; weighted
methods use $\max\{A^{\rm rel}_{ij},A^{\rm rel}_{ji}\}$.
The released latitude/longitude coordinates determine haversine
distance $D_{ij}$, and $\bar D_{ij}$ is this distance divided by the
median positive distance on the support. Let $n_j=\sum_i M_{ij}$.
Construct eight nonnegative raw channels
\[
(\widetilde\Phi_{b,s})_{ij}
=M_{ij}\exp\!\left[-\frac{(\bar D_{ij}-\mu_b)^2}{2a^2}\right]
(1+n_j)^{-s},\qquad
\mu=(0,0.5,1,2),\quad a=0.5,\quad s\in\{0,1\}.
\]
Normalize every nonempty row to obtain $\Phi_r$; isolated rows stay
zero. Compute normalization stably in the log domain. Include the
distance, endpoint degrees, and fixed row-normalization factors in
$\Psi$, so each realized channel remains a function of a fixed
descriptor. Edge-conditioned alternatives receive the same distance
and degree information. Record the realized dictionary rank;
do not assume coefficient identifiability when it is rank deficient.
No signal correlations, validation labels, or test observations define
or tune the support or dictionary.

\paragraph{GIOF realization.}
Use the propagation bottleneck of Eq.~\eqref{eq:layers-giof-layer}, with
\[
B_r=\Phi_r-\operatorname{Diag}(\Phi_r\mathbf1),\qquad
\widehat B_r=B_r/\rho(B_r),\qquad
L(c)=\sum_{r=1}^R\alpha_r(c)\widehat B_r,
\]
\[
\alpha(c)=\operatorname{softmax}(f_\theta(c)),\qquad
\tau(c)=\tau_{\max}\operatorname{sigmoid}(f_\tau(c)),\qquad
\mathcal S(F,c)=e^{\tau(c)L(c)}F.
\]
Exclude identically zero generators. The two selectors each have one
width-32 GELU hidden layer; initialize a uniform mixture and
$\tau=0.1$. Select $\tau_{\max}\in\{2,4\}$ using the budget below.
Here $\rho$ is the mean exit rate from
Definition~\ref{def:operators-normalized-generator}, not a spectral
radius. The realized propagator is nonnegative and constant preserving.
Its guarantees do not automatically apply to the full residual network
or its self-conditioned input--output Jacobian. Row normalization also
need not produce a symmetric generator, so the symmetric spectral
identities are not asserted for this real-data implementation.

\paragraph{Controls for channel sharing.}
GeoMLP predicts an edge logit $g_\psi(c,\Psi_{ij})$ with a shared
width-32 MLP. FreeEdge augments it by
$a_{ij}^{\top}v(c)+b_{ij}$, where $v(c)\in\mathbb R^8$ and
$(a_{ij},b_{ij})$ are independently learned for each admissible edge.
For both, apply masked row-softmax to obtain $W(c)$, with zero
diagonal and zero isolated rows, and set
\[
G(c)=W(c)-\operatorname{Diag}(W(c)\mathbf1),\qquad
L(c)=G(c)/\rho(G(c)).
\]
Use exactly the same context, bottleneck, scale selector, and
exponential-action routine as GIOF. These controls retain the support
and generator constraints while relaxing fixed-dictionary sharing;
they are purpose-built ablations, not additional published baselines.
Report their actual parameter counts rather than claiming parameter
equality.

\subsection{Baselines, training budget, and numerical implementation}
\label{app:exp_baselines}
\label{app:complexity}

\paragraph{Five published baselines.}
GCN \citep{kipf2016semi} tests normalized neighborhood aggregation;
GATv2 \citep{brody2021attentive} tests feature-dependent edge attention;
APPNP \citep{gasteiger2018predict} tests multi-hop propagation with
teleportation; MoNet \citep{monti2017geometric} tests learned geometric
kernels; and ECC \citep{simonovsky2017dynamic} tests filters generated
from edge attributes. We insert each spatial module into the common
network above, retaining its native feature transforms and normalization.
Thus these names denote geometry-matched module substitutions rather than
unchanged published end-to-end models. GCN, GATv2, MoNet, and ECC use
width 16 and validation-selected depth two or four, with GELU between
layers. GATv2 uses four heads of width four; MoNet uses eight kernels;
ECC uses a width-32 edge MLP producing $16\times16$ filters. APPNP uses
ten propagation steps and validation selects teleportation probability
$0.05$ or $0.2$. Native self-connections are permitted in addition to
the common off-diagonal support.

\paragraph{Optimization and selection budget.}
Use Adam, weight decay $10^{-4}$, effective batch size 32, and
gradient clipping at norm five. All compared models optimize hidden-target MAE under the common reconstruction
objective. Never supply hidden targets as encoder inputs. Use float32 and no
mixed precision for the primary comparison.

For each method and dataset, run six validation-only pilot trials on
the 20\%-day subset: learning rates
$\{3\times10^{-4},10^{-3},3\times10^{-3}\}$ crossed with the
two method-specific settings above. Each pilot uses seed 17 and
500 updates. Freeze the selected configuration for both data budgets; pilot checkpoints are not
reused. GIOF receives the same six-trial budget as every baseline.

Validate every 100 updates and use a final cap of 1,600 optimizer updates.
Select the checkpoint with the lowest validation criterion among the recorded
checkpoints and flag runs that reach the cap while still improving as budget
limited. Use microbatches of eight with four-step accumulation for every method.
Record actual updates, stopping reasons, and validation traces; equal update
budgets do not imply equal wall-clock cost. Because the update cap is fixed, the
20\%/100\% data comparison changes data availability while matching optimization
steps, not the number of passes through the training set.

\paragraph{Sparse propagation.}
Let $d_{\rm a}$ be the number of nonisolated rows. All normalized
channels and row-softmax controls have common active-row exit rate
$\nu=d/d_{\rm a}$. Set $T_L=I+L/\nu$ and
\[
w_k=e^{-\nu\tau}\frac{(\nu\tau)^k}{k!},\qquad
q_K=1-\sum_{k=0}^K w_k,\qquad
\widetilde P_KF=\sum_{k=0}^K w_kT_L^kF+q_KF.
\]
The tail assigned to identity preserves nonnegativity and constants
in exact arithmetic, with
\[
\|e^{\tau L}-\widetilde P_K\|_{\infty\to\infty}\le2q_K.
\]
Before training choose one fixed $K$ per dataset satisfying
$\Pr\{\operatorname{Poisson}(4\nu)>K\}\le5\times10^{-7}$.
Evaluate Poisson weights/tails stably, record floating-point residuals,
and use the same $K$ for GIOF, GeoMLP, and FreeEdge.
The finite action retains reachability exclusion but need not assign
positive weight to every reachable pair. Compare actions and selector
gradients with dense float64 exponentials on 32 saved validation
contexts. These implementation checks do not replace the completed
controlled property experiments.

For $m=d+|\mathcal E_{\rm off}|$ and bottleneck width $C=16$,
GIOF's propagation-core cost per window is $O(Rm+KmC)$.
Fixed sparse channels require $O(Rm)$ storage; ordinary backpropagation
through the recurrence additionally stores $O(KdC)$ activations per
window, reducible by checkpointing. Encoder, decoder, selectors, and
batch size contribute additional costs. No dense $d\times d$
exponential is formed during training.

\paragraph{Measured computational cost.}
Use one shared GPU, an otherwise idle device, identical precision,
and the same data-loader worker count. Record GPU/CPU models, RAM,
framework, CUDA, and sparse-kernel versions. In
Table~\ref{tab:giof_real_main}, Params is the total trainable parameter
count in thousands, including encoder, decoder, embeddings, and
selectors; fixed graph buffers are recorded separately in the supplement.
Train is elapsed time from the first training batch until stopping,
including data loading and validation, averaged across final seeds.
Report time to the selected checkpoint and pilot-search time separately;
neither is substituted for full-run time.

GPU is the maximum allocated device memory during a training update,
including model, gradients, optimizer state, graph buffers, and
activations, with microbatch eight and effective batch 32.
Reset the peak counter after optimizer-state initialization and measure
100 complete updates after 20 warm-up updates; also record peak
reserved memory separately. Infer is batch-one latency for a complete
24-step window on the same device, including the encoder, spatial
module, and decoder. Preload inputs and exclude host-to-device
transfer, perform 20 warm-up passes, and report the median of 100 synchronized inference
passes per seed, then average seed medians.
Record latency percentiles and batch-32 throughput in the supplement.
Preprocessing, compilation, and tuning costs are reported separately.
Analytical operation counts are not used as substitutes for timing.

\subsection{Supplementary results and interpretation}
\label{app:exp_additional}

\paragraph{Reuse of trained models.}
Only the two main training-day budgets require the full six-method
comparison. Reuse those checkpoints for the entire severity sweep,
independent sensor outages, point missingness, and missing intervals
of length $\ell\in\{1,3,6,12,18,24\}$ within a window.
For the duration diagnostic, choose 30\% of sensors uniformly and
one uniformly positioned contiguous interval per chosen sensor.
Archive all per-seed MAEs, RMSEs, target counts, predictions, and masks.
Figure~\ref{fig:giof_real_curves}(c,d) uses the regional-outage MAE
at each logged validation checkpoint. 

\begin{figure}[!t]
\centering
\begin{minipage}[t]{0.49\linewidth}
    \centering
    \includegraphics[width=\linewidth]{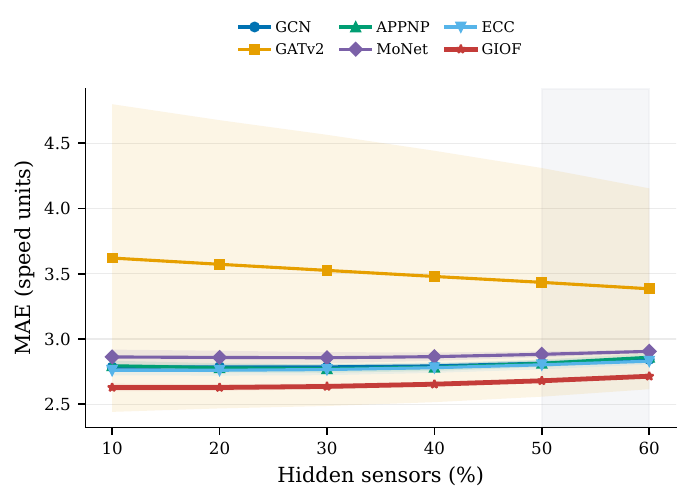}
    \vspace{-1.5mm}

    {\scriptsize\textbf{(a)} PEMS-BAY: independent sensor outages}
\end{minipage}\hfill
\begin{minipage}[t]{0.49\linewidth}
    \centering
    \includegraphics[width=\linewidth]{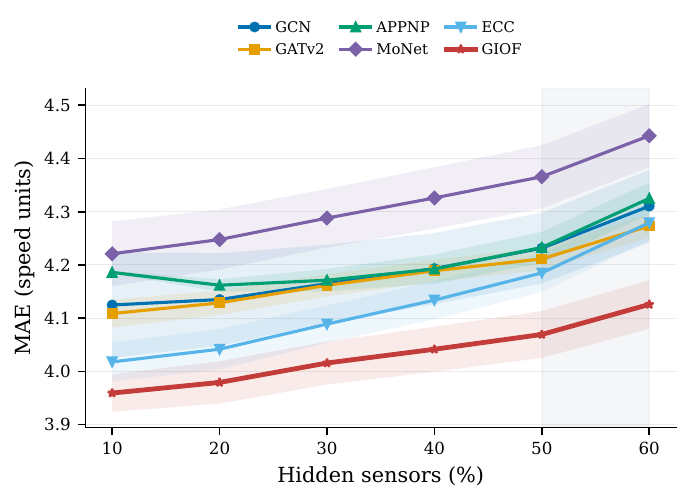}
    \vspace{-1.5mm}

    {\scriptsize\textbf{(b)} METR-LA: independent sensor outages}
\end{minipage}

\vspace{1mm}

\begin{minipage}[t]{0.49\linewidth}
    \centering
    \includegraphics[width=\linewidth]{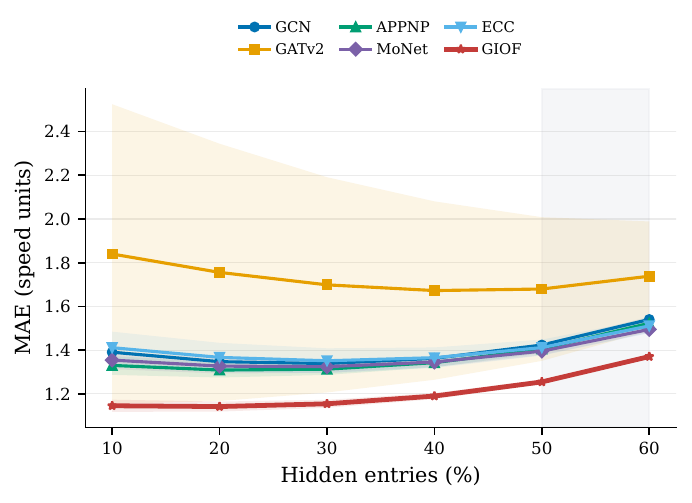}
    \vspace{-1.5mm}

    {\scriptsize\textbf{(c)} PEMS-BAY: point missingness}
\end{minipage}\hfill
\begin{minipage}[t]{0.49\linewidth}
    \centering
    \includegraphics[width=\linewidth]{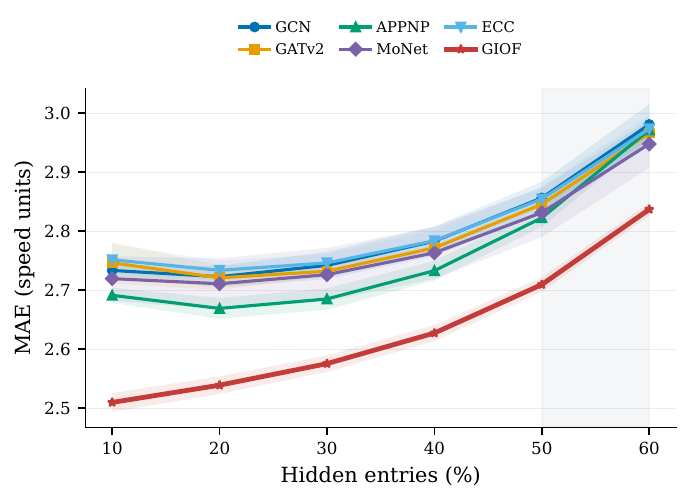}
    \vspace{-1.5mm}

    {\scriptsize\textbf{(d)} METR-LA: point missingness}
\end{minipage}

\caption{Additional missingness controls using the 20\%-training-day
checkpoints. \textbf{(a,b)} Independent sensor outages.
\textbf{(c,d)} Independent point missingness. Curves report test MAE and
aggregate seeds after averaging shared mask realizations within
each seed.}
\label{fig:giof_real_controls}
\end{figure}

\begin{figure}[!t]
\centering
\begin{minipage}[t]{0.49\linewidth}
    \centering
    \includegraphics[width=\linewidth]{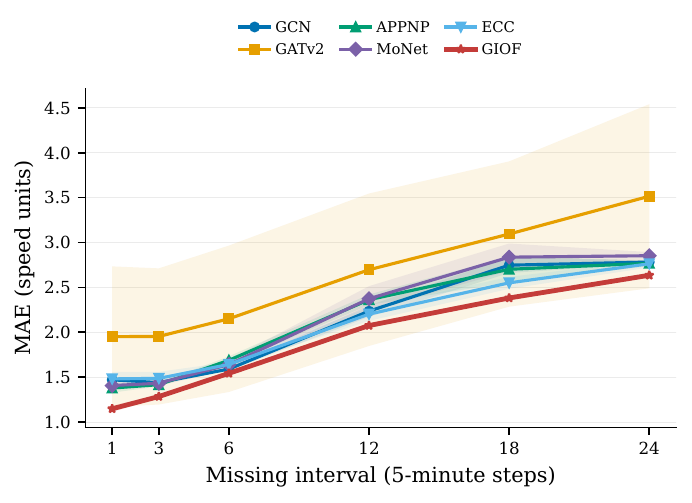}
    \vspace{-1.5mm}

    {\scriptsize\textbf{(a)} PEMS-BAY: outage duration}
\end{minipage}\hfill
\begin{minipage}[t]{0.49\linewidth}
    \centering
    \includegraphics[width=\linewidth]{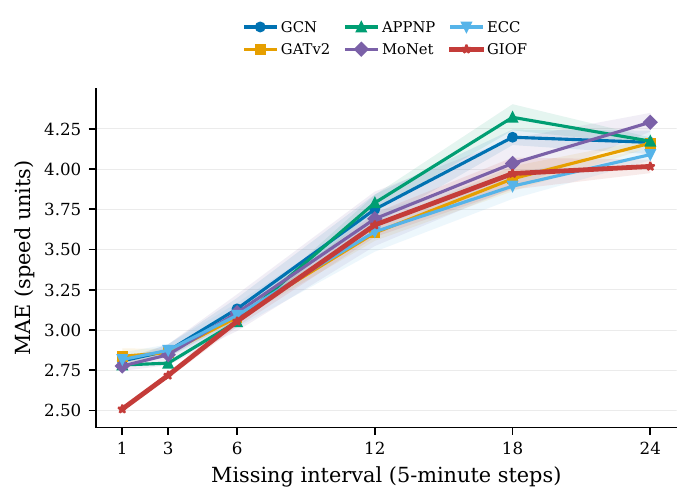}
    \vspace{-1.5mm}

    {\scriptsize\textbf{(b)} METR-LA: outage duration}
\end{minipage}

\vspace{1mm}

\begin{minipage}[t]{0.49\linewidth}
    \centering
    \includegraphics[width=\linewidth]{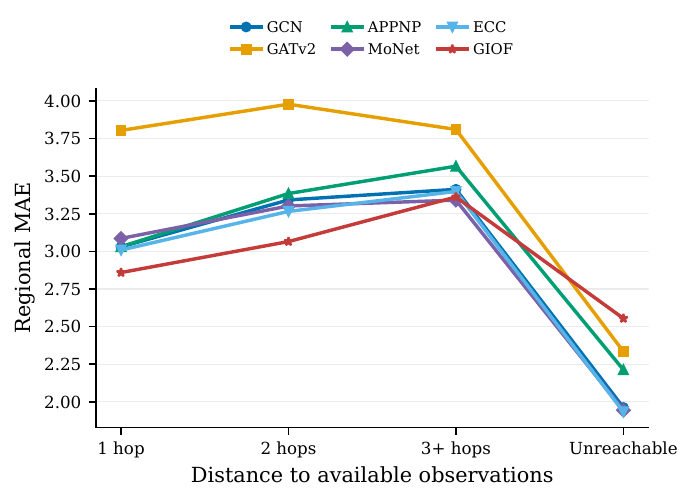}
    \vspace{-1.5mm}

    {\scriptsize\textbf{(c)} PEMS-BAY: graph distance}
\end{minipage}\hfill
\begin{minipage}[t]{0.49\linewidth}
    \centering
    \includegraphics[width=\linewidth]{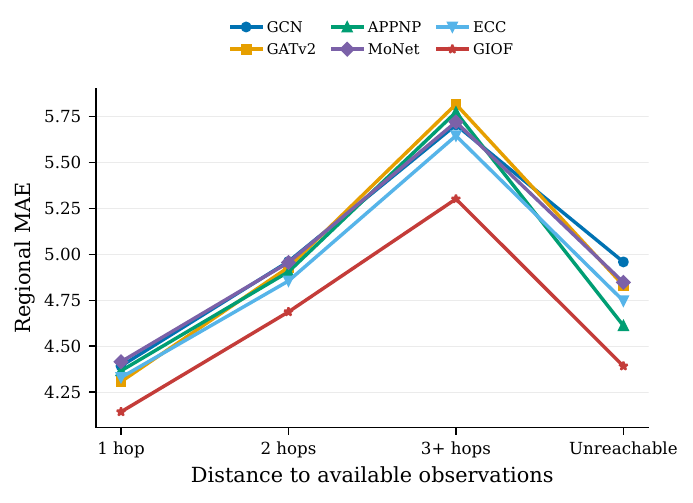}
    \vspace{-1.5mm}

    {\scriptsize\textbf{(d)} METR-LA: graph distance}
\end{minipage}

\caption{Diagnostics for structured missingness.
\textbf{(a,b)} Reconstruction MAE as the contiguous missing interval grows.
\textbf{(c,d)} Regional-outage error stratified by graph distance from
available sensors. These evaluations reuse trained checkpoints and require no
additional model fitting.}
\label{fig:giof_real_diagnostics}
\end{figure}

\paragraph{Matched ablations on real observations.}
Restrict additional training to PEMS-BAY with 20\% training days.
No-spatial removes the spatial branch. Static learns a single constant
mixture and scale. Shuffled permutes entire eight-dimensional raw
channel vectors across admissible ordered edges with seed 1729, then
renormalizes rows and recompiles generators; the permutation is fixed
before training. It preserves support and the coefficient budget while
disrupting the correspondence with geometry, and is not merely a
renaming of channels. One-step uses
\[
\mathcal S_{\rm one}(F,c)
=\bigl[(1-\beta(c))I+\beta(c)T_{L(c)}\bigr]F,
\qquad \beta(c)=\operatorname{sigmoid}(f_\beta(c)),
\]
with the same dictionary and mixture selector, retaining nonnegative
constant-preserving one-hop mixing. GeoMLP and FreeEdge are defined
above. Retrain each ablation with seeds and the same stopping
rule. For each, use six pilot trials given by the three learning rates
crossed with weight decay $\{0,10^{-4}\}$; retain GIOF's selected
$\tau_{\max}$ and numerical tolerance for the propagation controls.

\begin{table}[!t]
\centering
\scriptsize
\setlength{\tabcolsep}{3.2pt}
\caption{Real-data ablations on PEMS-BAY with 20\% training days at 30\%
missingness. Reg.: regional outages; Ind.: independent sensor outages; Point:
independent point missingness. Values report mean $\pm$ sample standard
deviation over seeds.}
\label{tab:giof_ablations}
\resizebox{\linewidth}{!}{%
\begin{tabular}{lccccc}
\toprule
Variant & Reg. MAE & Ind. MAE & Point MAE & Reg. RMSE & Params (K)\\
\midrule
No-spatial & $3.696\pm0.002$ & $3.566\pm0.003$ & $1.280\pm0.053$ & $6.886\pm0.018$ & 16.38\\
Static     & $3.483\pm0.003$ & $2.891\pm0.006$ & $1.307\pm0.037$ & $6.444\pm0.005$ & 18.44\\
Shuffled   & $3.501\pm0.013$ & $2.944\pm0.015$ & $1.291\pm0.029$ & $6.458\pm0.027$ & 23.14\\
One-step   & $3.534\pm0.010$ & $2.885\pm0.032$ & $1.306\pm0.042$ & $6.562\pm0.023$ & 23.14\\
GeoMLP     & $3.390\pm0.007$ & $2.870\pm0.005$ & $1.292\pm0.023$ & $6.263\pm0.018$ & 23.27\\
FreeEdge   & $3.384\pm0.138$ & $2.774\pm0.117$ & $1.364\pm0.055$ & $6.770\pm0.793$ & 61.24\\
GIOF       & $\mathbf{3.070\pm0.025}$ & $\mathbf{2.636\pm0.022}$ & $\mathbf{1.154\pm0.018}$ & $\mathbf{5.670\pm0.042}$ & 23.14\\
\bottomrule
\end{tabular}}
\end{table}

\paragraph{Connection to the structural claims.}
GeoMLP and FreeEdge compare fixed-channel reuse with more flexible
edge parameterization under the same support and propagation solver.
Static tests context-dependent selection; Shuffled tests geometry
alignment; One-step tests propagation range. No-spatial measures the
benefit of adding the spatial branch but alone does not isolate
geometry from added capacity. These comparisons concern task-level
consequences of the structured bias, rather than proofs of the
Frobenius projection or spectral denoising identities.

For regional masks, stratify hidden-entry errors by graph distance to
the nearest sensor with at least one available within-window reading:
one hop, two hops, at least three hops, or no reachable observed sensor.
Report all bin counts and retain the unreachable group in the overall
metric. This diagnostic uses the existing predictions and does not
require retraining. Interpret it alongside the One-step and APPNP
comparisons; it is not by itself a test of functional Jacobian locality.
For a qualitative example, choose the first chronological test window
with scored targets under mask realization zero before inspecting
predictions. Keep unfavorable examples and full-data outcomes.
Any claimed accuracy or efficiency advantage must be supported by the
reported measurements; no universal baseline dominance is assumed.

Archive dataset hashes, sensor-ID alignment, dates, graph buffers,
dictionary rank, all masks and selected days, trial configurations,
source commits, checkpoints, numerical validation errors, and the raw
arrays underlying every plot and table. Also release training,
validation, timing, memory, and stopping logs so that accuracy--cost
tradeoffs can be reconstructed independently.

\FloatBarrier

\section{Proofs}
\label{app:proofs}

\subsection{Proof of Proposition~\ref{prop:operators-one-step-support}}

\begin{proof}
Let \(A\in\mathcal A_{\mathfrak S}\) be arbitrary, and fix two distinct indices
\(i,j\in\{1,\dots,d\}\) with \(i\neq j\). By
Definition~\ref{def:operators-interaction-family}, membership of \(A\) in
\(\mathcal A_{\mathfrak S}\) means that there exist coefficients
\(\theta=(\theta_1,\dots,\theta_R)\in\mathbb R^R\) and
\(\delta=(\delta_1,\dots,\delta_d)\in\mathbb R^d\), with
\(\theta_r\ge 0\) for every \(r\in\mathcal R_+\), such that
$
A
=
\sum_{r=1}^R \theta_r\Phi_r
+
\operatorname{Diag}(\delta).
$
Taking the \((i,j)\)-entry of this identity gives
$
A_{ij}
=
\sum_{r=1}^R \theta_r(\Phi_r)_{ij}
+
\bigl(\operatorname{Diag}(\delta)\bigr)_{ij}.
$
Since \(i\neq j\), the matrix \(\operatorname{Diag}(\delta)\) has no
contribution at the \((i,j)\)-entry. Explicitly,
$
\bigl(\operatorname{Diag}(\delta)\bigr)_{ij}=0.
$
Hence
$
A_{ij}
=
\sum_{r=1}^R \theta_r(\Phi_r)_{ij}.
$
By Definition~\ref{def:prelim-geometry-descriptor}, each masked channel
matrix \(\Phi_r\) is defined entrywise by
$
(\Phi_r)_{ij}
=
\mathbf 1_{\mathcal E}(z_i,z_j)\,
\phi_r(z_i,z_j).
$
Now assume that
$
(z_i,z_j)\notin\mathcal E.
$
By the definition of the indicator function of the admissibility relation,
this implies
$
\mathbf 1_{\mathcal E}(z_i,z_j)=0.
$
Therefore, for every \(r\in\{1,\dots,R\}\),
$
(\Phi_r)_{ij}
=
\mathbf 1_{\mathcal E}(z_i,z_j)\,
\phi_r(z_i,z_j)
=
0\cdot\phi_r(z_i,z_j)
=
0.
$
Substituting these identities into the expression for \(A_{ij}\), we obtain
\[
A_{ij}
=
\sum_{r=1}^R \theta_r(\Phi_r)_{ij}
=
\sum_{r=1}^R \theta_r\cdot 0
=
0.
\]
Thus, for every \(A\in\mathcal A_{\mathfrak S}\) and every pair of distinct
indices \(i\neq j\),
$
(z_i,z_j)\notin\mathcal E
\quad\Longrightarrow\quad
A_{ij}=0,
$
as claimed. Notice that the sign restrictions
\(\theta_r\ge 0\) for \(r\in\mathcal R_+\) are not needed for this support
argument: the conclusion follows solely from the masking factor
\(\mathbf 1_{\mathcal E}(z_i,z_j)\) and from the fact that the free
self-action term \(\operatorname{Diag}(\delta)\) is diagonal.
\end{proof}

\subsection{Proof of Proposition~\ref{prop:properties-affine-hull-dimension}}

\begin{proof}
Let
\[
V
:=
\operatorname{span}
\bigl\{
\Phi_1,\ldots,\Phi_R,E_{11},\ldots,E_{dd}
\bigr\}
\subseteq
\mathbb R^{d\times d}.
\]
We first identify the linear span of
\(\mathcal A_{\mathfrak S}\) exactly. By
Definition~\ref{def:operators-interaction-family}, every
\(A\in\mathcal A_{\mathfrak S}\) admits a representation
\[
A
=
\sum_{r=1}^R\theta_r\Phi_r
+
\operatorname{Diag}(\delta),
\]
where
\(\theta\in\mathbb R^R\),
\(\delta\in\mathbb R^d\), and
\(\theta_r\ge 0\) whenever \(r\in\mathcal R_+\).
Writing
\(\delta=(\delta_1,\ldots,\delta_d)^\top\), the diagonal term can be expanded
in the standard diagonal matrix units as
$
\operatorname{Diag}(\delta)
=
\sum_{i=1}^d \delta_i E_{ii}.
$
Consequently,
\[
A
=
\sum_{r=1}^R\theta_r\Phi_r
+
\sum_{i=1}^d\delta_iE_{ii},
\]
and hence every element of \(\mathcal A_{\mathfrak S}\) belongs to \(V\).
Therefore
$
\mathcal A_{\mathfrak S}\subseteq V.
$
Since \(V\) is itself a linear subspace, taking linear spans on the left-hand
side yields
$
\operatorname{span}(\mathcal A_{\mathfrak S})
\subseteq
V.
$

We now prove the reverse inclusion. Fix
\(r\in\{1,\ldots,R\}\), and choose a coefficient vector
\(\theta^{(r)}\in\mathbb R^R\) by
\[
\theta^{(r)}_s
=
\begin{cases}
1, & s=r,\\
0, & s\neq r.
\end{cases}
\]
For every \(s\in\mathcal R_+\), one has
\(\theta^{(r)}_s\in\{0,1\}\), and therefore
$
\theta^{(r)}_s\ge 0.
$
Thus \(\theta^{(r)}\) satisfies all sign constraints in
Definition~\ref{def:operators-interaction-family}. Taking
\(\delta=0\) gives
\[
\sum_{s=1}^R\theta^{(r)}_s\Phi_s
+
\operatorname{Diag}(0)
=
\Phi_r,
\]
so
$
\Phi_r\in\mathcal A_{\mathfrak S}
\qquad
\text{for every }r=1,\ldots,R.
$
Similarly, fix \(i\in\{1,\ldots,d\}\). Let \(e_i\in\mathbb R^d\) denote the
\(i\)-th standard basis vector. Taking
\(\theta=0\) and \(\delta=e_i\) satisfies all coefficient constraints and
gives
$
\sum_{r=1}^R0\cdot\Phi_r
+
\operatorname{Diag}(e_i)
=
E_{ii}.
$
Hence
$
E_{ii}\in\mathcal A_{\mathfrak S}
\qquad
\text{for every }i=1,\ldots,d.
$
It follows that every generator of \(V\) belongs to
\(\mathcal A_{\mathfrak S}\), and therefore in particular belongs to
\(\operatorname{span}(\mathcal A_{\mathfrak S})\). Thus
\[
V
=
\operatorname{span}
\bigl\{
\Phi_1,\ldots,\Phi_R,E_{11},\ldots,E_{dd}
\bigr\}
\subseteq
\operatorname{span}(\mathcal A_{\mathfrak S}).
\]
Combining the two inclusions gives the exact identity
\[
\operatorname{span}(\mathcal A_{\mathfrak S})
=
\operatorname{span}
\bigl\{
\Phi_1,\ldots,\Phi_R,E_{11},\ldots,E_{dd}
\bigr\}.
\]

It remains to compare the affine hull with the linear span. Setting
\(\theta=0\) and \(\delta=0\) in
Definition~\ref{def:operators-interaction-family} shows that
$
0\in\mathcal A_{\mathfrak S}.
$
Every affine combination is, in particular, a linear combination, so one
always has
$
\operatorname{aff}(\mathcal A_{\mathfrak S})
\subseteq
\operatorname{span}(\mathcal A_{\mathfrak S}).
$
For the converse inclusion, let
\(B\in\operatorname{span}(\mathcal A_{\mathfrak S})\) be arbitrary. By the
definition of linear span, there exist
\(A^{(1)},\ldots,A^{(m)}\in\mathcal A_{\mathfrak S}\) and scalars
\(\alpha_1,\ldots,\alpha_m\in\mathbb R\) such that
$
B
=
\sum_{k=1}^m\alpha_k A^{(k)}.
$
Define
\[
\alpha_0
:=
1-\sum_{k=1}^m\alpha_k.
\]
Because \(0\in\mathcal A_{\mathfrak S}\), we may rewrite \(B\) as
$
B
=
\alpha_0\,0
+
\sum_{k=1}^m\alpha_k A^{(k)}.
$
The coefficients in this representation satisfy
\[
\alpha_0+\sum_{k=1}^m\alpha_k
=
1-\sum_{k=1}^m\alpha_k+\sum_{k=1}^m\alpha_k
=
1.
\]
Hence the preceding representation is an affine combination of elements of
\(\mathcal A_{\mathfrak S}\), and therefore
$
B\in\operatorname{aff}(\mathcal A_{\mathfrak S}).
$
Since \(B\) was arbitrary,
$
\operatorname{span}(\mathcal A_{\mathfrak S})
\subseteq
\operatorname{aff}(\mathcal A_{\mathfrak S}).
$
Together with the opposite inclusion, this proves
\[
\operatorname{aff}(\mathcal A_{\mathfrak S})
=
\operatorname{span}(\mathcal A_{\mathfrak S})
=
\operatorname{span}
\bigl\{
\Phi_1,\ldots,\Phi_R,E_{11},\ldots,E_{dd}
\bigr\}.
\]

If
$
\{\Phi_1,\ldots,\Phi_R,E_{11},\ldots,E_{dd}\}
$
is linearly independent, then it consists of exactly \(R+d\) linearly
independent matrices. It is therefore a basis of the common space above, and
hence
\[
\dim\!\bigl(\operatorname{aff}(\mathcal A_{\mathfrak S})\bigr)
=
\dim\!\bigl(\operatorname{span}(\mathcal A_{\mathfrak S})\bigr)
=
R+d.
\]
If linear independence is not assumed, the same common space is still spanned
by the \(R+d\) matrices
\(\Phi_1,\ldots,\Phi_R,E_{11},\ldots,E_{dd}\). A vector space spanned by
\(R+d\) vectors has dimension at most \(R+d\), so
\[
\dim\!\bigl(\operatorname{aff}(\mathcal A_{\mathfrak S})\bigr)
=
\dim\!\bigl(\operatorname{span}(\mathcal A_{\mathfrak S})\bigr)
\le
R+d.
\]
This proves both assertions.
\end{proof}

\subsection{Proof of Theorem~\ref{thm:properties-order-supnorm-stability}}

\begin{proof}
Let \(L\in\mathcal L_{\mathfrak S}\) and let \(\tau\ge0\). By
Definition~\ref{def:operators-generator-family}, the matrix \(L\) satisfies
$
L_{ij}\ge0
\qquad
\text{for every }i\neq j,
$
and
$
L\mathbf1=0.
$
The latter identity means that the entries in every row of \(L\) sum to zero. Thus, for each \(i\in\{1,\ldots,d\}\),
\[
0
=
(L\mathbf1)_i
=
\sum_{j=1}^d L_{ij}
=
L_{ii}+\sum_{j\neq i}L_{ij},
\]
and therefore
$
L_{ii}
=
-\sum_{j\neq i}L_{ij}.
$
Since every off-diagonal entry \(L_{ij}\) with \(j\neq i\) is nonnegative, it follows that
\[
-L_{ii}
=
\sum_{j\neq i}L_{ij}
\ge0,
\qquad
L_{ii}\le0.
\]

Choose any number
$
\nu>\max_{1\le i\le d}(-L_{ii}).
$
Such a \(\nu\) exists because the maximum is finite. Define
\[
T
:=
I+\frac{1}{\nu}L.
\]
We first verify directly that \(T\) is entrywise nonnegative. If \(i\neq j\), then
$
T_{ij}
=
\frac{L_{ij}}{\nu}
\ge0.
$
For a diagonal entry,
\[
T_{ii}
=
1+\frac{L_{ii}}{\nu}
=
1-\frac{-L_{ii}}{\nu}.
\]
By the choice of \(\nu\),
$
0\le -L_{ii}<\nu,
$
and hence
\[
T_{ii}
=
1-\frac{-L_{ii}}{\nu}
>0.
\]
Consequently,
$
T\ge0
$
entrywise. Moreover,
\[
T\mathbf1
=
\left(I+\frac{1}{\nu}L\right)\mathbf1
=
I\mathbf1+\frac{1}{\nu}L\mathbf1
=
\mathbf1+\frac{1}{\nu}\,0
=
\mathbf1.
\]
Thus \(T\) is a nonnegative matrix whose row sums are all equal to one.

From the definition of \(T\),
$
T-I=\frac{1}{\nu}L,
$
so
$
L=\nu(T-I).
$
Therefore
\[
P
=
e^{\tau L}
=
e^{\nu\tau(T-I)}.
\]
Since \(T\) commutes with the identity matrix \(I\), the matrices
\(\nu\tau T\) and \(-\nu\tau I\) commute. Hence the exponential of their sum factors as
\[
e^{\nu\tau(T-I)}
=
e^{\nu\tau T-\nu\tau I}
=
e^{-\nu\tau I}e^{\nu\tau T}.
\]
Because
$
e^{-\nu\tau I}
=
e^{-\nu\tau}I,
$
we obtain
\[
P
=
e^{-\nu\tau}e^{\nu\tau T}.
\]
Expanding the second exponential in its absolutely convergent matrix power series gives
\[
e^{\nu\tau T}
=
\sum_{k=0}^{\infty}
\frac{(\nu\tau T)^k}{k!}
=
\sum_{k=0}^{\infty}
\frac{(\nu\tau)^k}{k!}T^k.
\]
Thus
\[
P
=
e^{-\nu\tau}
\sum_{k=0}^{\infty}
\frac{(\nu\tau)^k}{k!}T^k.
\]

We now verify the sign of every term in this expansion. Since \(T\ge0\), one has
$
T^0=I\ge0
\qquad\text{and}\qquad
T^1=T\ge0.
$
If \(T^k\ge0\) for some \(k\ge1\), then for every pair \(i,j\),
\[
(T^{k+1})_{ij}
=
(T^kT)_{ij}
=
\sum_{\ell=1}^d
(T^k)_{i\ell}T_{\ell j}.
\]
Every factor in every summand is nonnegative, so
$
(T^{k+1})_{ij}\ge0.
$
By induction,
$
T^k\ge0
\qquad
\text{for every }k\ge0.
$
In addition, because \(\nu>0\) and \(\tau\ge0\),
\[
e^{-\nu\tau}>0,
\qquad
\frac{(\nu\tau)^k}{k!}\ge0
\qquad
\text{for every }k\ge0.
\]
Hence every matrix in the series
$
e^{-\nu\tau}
\frac{(\nu\tau)^k}{k!}T^k
$
is entrywise nonnegative. Taking the convergent entrywise sum therefore yields
\[
P=e^{\tau L}\ge0.
\]

We next prove preservation of the constant vector. Since
$
T\mathbf1=\mathbf1,
$
repeated application of \(T\) gives
$
T^2\mathbf1
=
T(T\mathbf1)
=
T\mathbf1
=
\mathbf1.
$
Proceeding inductively,
$
T^k\mathbf1=\mathbf1
\qquad
\text{for every }k\ge0,
$
where \(T^0\mathbf1=I\mathbf1=\mathbf1\). Applying the preceding exponential expansion to \(\mathbf1\) therefore gives
\[
P\mathbf1
=
e^{-\nu\tau}
\sum_{k=0}^{\infty}
\frac{(\nu\tau)^k}{k!}T^k\mathbf1
=
e^{-\nu\tau}
\sum_{k=0}^{\infty}
\frac{(\nu\tau)^k}{k!}\mathbf1.
\]
Factoring out the common vector \(\mathbf1\),
$
P\mathbf1
=
e^{-\nu\tau}
\left(
\sum_{k=0}^{\infty}
\frac{(\nu\tau)^k}{k!}
\right)\mathbf1.
$
The scalar series is the ordinary exponential series,
$
\sum_{k=0}^{\infty}
\frac{(\nu\tau)^k}{k!}
=
e^{\nu\tau}.
$
Consequently,
\[
P\mathbf1
=
e^{-\nu\tau}e^{\nu\tau}\mathbf1
=
\mathbf1.
\]
Thus \(P\) is entrywise nonnegative and every row of \(P\) sums to one.

We now establish the induced sup-norm bound. Let
\(h=(h_1,\ldots,h_d)^\top\in\mathbb R^d\). For each \(i\),
$
(Ph)_i
=
\sum_{j=1}^dP_{ij}h_j.
$
Using the triangle inequality,
\[
|(Ph)_i|
=
\left|
\sum_{j=1}^dP_{ij}h_j
\right|
\le
\sum_{j=1}^d
|P_{ij}h_j|.
\]
Since \(P_{ij}\ge0\),
$
|P_{ij}h_j|
=
P_{ij}|h_j|,
$
and therefore
\[
|(Ph)_i|
\le
\sum_{j=1}^dP_{ij}|h_j|.
\]
By the definition of the sup norm,
$
|h_j|\le\|h\|_\infty
\qquad
\text{for every }j,
$
so
\[
|(Ph)_i|
\le
\sum_{j=1}^dP_{ij}\|h\|_\infty
=
\|h\|_\infty
\sum_{j=1}^dP_{ij}.
\]
Because \(P\mathbf1=\mathbf1\),
$
\sum_{j=1}^dP_{ij}=1
\qquad
\text{for every }i.
$
Hence
$
|(Ph)_i|
\le
\|h\|_\infty
\qquad
\text{for every }i.
$
Taking the maximum over \(i\) gives
\[
\|Ph\|_\infty
=
\max_i |(Ph)_i|
\le
\|h\|_\infty.
\]
It follows that
$
\|P\|_{\infty\to\infty}
\le1.
$
To obtain the reverse inequality, take \(h=\mathbf1\). Since
\(P\mathbf1=\mathbf1\),
\[
\frac{\|P\mathbf1\|_\infty}
{\|\mathbf1\|_\infty}
=
\frac{\|\mathbf1\|_\infty}
{\|\mathbf1\|_\infty}
=
1.
\]
By the definition of the induced operator norm,
\[
\|P\|_{\infty\to\infty}
\ge
\frac{\|P\mathbf1\|_\infty}
{\|\mathbf1\|_\infty}
=
1.
\]
Combining the upper and lower bounds yields
$
\|P\|_{\infty\to\infty}=1.
$

It remains to verify order preservation. Let \(h,g\in\mathbb R^d\) satisfy
$
h\le g
$
componentwise. Then
$
g-h\ge0.
$
Since \(P\ge0\), for each \(i\),
\[
\bigl(P(g-h)\bigr)_i
=
\sum_{j=1}^d
P_{ij}(g_j-h_j)
\ge0,
\]
because every factor \(P_{ij}\) and every difference \(g_j-h_j\) is
nonnegative. Hence
$
P(g-h)\ge0.
$
By linearity,
\[
P(g-h)=Pg-Ph,
\]
and therefore
\[
Pg-Ph\ge0,
\]
which is equivalent to
$
Ph\le Pg.
$
Together with the already established estimate
$
\|Ph\|_\infty\le\|h\|_\infty,
$
this proves all of the stated conclusions.
\end{proof}

\subsection{Proof of Theorem~\ref{thm:properties-propagation-oversmoothing}}

\begin{proof}
Fix \(\tau\ge0\). Since
\(L\in\mathcal L_{\mathfrak S}\), Definition~\ref{def:operators-generator-family}
gives
$
L_{ij}\ge0
\qquad
\text{for every } i\neq j,
$
together with
$
L\mathbf1=0.
$
The latter identity implies that every row of \(L\) sums to zero. Hence, for
each \(i\in\{1,\ldots,d\}\),
\[
0
=
(L\mathbf1)_i
=
\sum_{j=1}^dL_{ij}
=
L_{ii}+\sum_{j\neq i}L_{ij},
\]
and therefore
$
L_{ii}
=
-\sum_{j\neq i}L_{ij}.
$
Because \(L=L^\top\), the matrix \(L\) is real symmetric. For an arbitrary
vector \(x=(x_1,\ldots,x_d)^\top\in\mathbb R^d\), we may therefore write its
quadratic form as
\[
x^\top Lx
=
\sum_{i=1}^d L_{ii}x_i^2
+
2\sum_{1\le i<j\le d}L_{ij}x_ix_j.
\]
Substituting the preceding expression for each diagonal entry gives
\[
x^\top Lx
=
-\sum_{i=1}^d
\left(
\sum_{j\neq i}L_{ij}
\right)x_i^2
+
2\sum_{1\le i<j\le d}L_{ij}x_ix_j.
\]
Using the symmetry \(L_{ij}=L_{ji}\), every unordered pair
\(\{i,j\}\) with \(i<j\) contributes once through \(x_i^2\) and once through
\(x_j^2\) in the first double sum. Thus
\[
-\sum_{i=1}^d
\left(
\sum_{j\neq i}L_{ij}
\right)x_i^2
=
-\sum_{1\le i<j\le d}
L_{ij}(x_i^2+x_j^2),
\]
and consequently
\[
\begin{aligned}
x^\top Lx
&=
-\sum_{1\le i<j\le d}
L_{ij}(x_i^2+x_j^2)
+
2\sum_{1\le i<j\le d}L_{ij}x_ix_j\\
&=
-\sum_{1\le i<j\le d}
L_{ij}
\bigl(x_i^2-2x_ix_j+x_j^2\bigr)\\
&=
-\sum_{1\le i<j\le d}
L_{ij}(x_i-x_j)^2.
\end{aligned}
\]
Since \(L_{ij}\ge0\) for \(i\neq j\), every term in the last sum is
nonpositive after multiplication by the leading minus sign. Hence
$
x^\top Lx\le0
\qquad
\text{for every }x\in\mathbb R^d.
$
Therefore \(L\) is negative semidefinite. In particular, because \(L\) is
real symmetric, all of its eigenvalues are real and nonpositive.

Moreover,
$
L\mathbf1=0,
$
so \(0\) is an eigenvalue of \(L\) with eigenvector \(\mathbf1\). The
assumption
$
\ker(L)=\operatorname{span}\{\mathbf1\}
$
shows that this zero eigenspace is one-dimensional. Hence \(0\) is a simple
eigenvalue, while every other eigenvalue is strictly negative. This is
consistent with the ordering in the statement,
\[
0=\lambda_1>\lambda_2\ge\cdots\ge\lambda_d,
\]
and therefore
$
\gamma=-\lambda_2>0.
$

By the spectral theorem for real symmetric matrices, there exists an
orthonormal basis
$
u_1,\ldots,u_d
$
of \(\mathbb R^d\) such that
$
Lu_k=\lambda_k u_k,
\qquad
k=1,\ldots,d.
$
Since the eigenspace associated with \(\lambda_1=0\) is precisely
\(\operatorname{span}\{\mathbf1\}\), we may choose
\[
u_1
=
\frac{\mathbf1}{\|\mathbf1\|_2}.
\]
Because
$
\|\mathbf1\|_2^2
=
\mathbf1^\top\mathbf1
=
d,
$
we have
$
u_1
=
\frac1{\sqrt d}\mathbf1.
$

Expand an arbitrary \(h\in\mathbb R^d\) in this orthonormal eigenbasis:
\[
h
=
\sum_{k=1}^d\alpha_k u_k,
\qquad
\alpha_k
=
u_k^\top h.
\]
For the first coefficient,
\[
\alpha_1
=
u_1^\top h
=
\left(
\frac1{\sqrt d}\mathbf1
\right)^\top h
=
\frac{\mathbf1^\top h}{\sqrt d}.
\]
It follows that the component of \(h\) in the constant direction is
\[
\alpha_1u_1
=
\frac{\mathbf1^\top h}{\sqrt d}
\frac{\mathbf1}{\sqrt d}
=
\frac{\mathbf1^\top h}{d}\mathbf1
=
\overline h.
\]
Therefore
$
h
=
\overline h
+
\sum_{k=2}^d\alpha_k u_k,
$
and hence
$
h-\overline h
=
\sum_{k=2}^d\alpha_k u_k.
$
Because \(u_2,\ldots,u_d\) are orthonormal,
\[
\|h-\overline h\|_2^2
=
\left\|
\sum_{k=2}^d\alpha_k u_k
\right\|_2^2
=
\sum_{k=2}^d\alpha_k^2.
\]

We next compute the action of the matrix exponential on these eigenmodes.
For every \(k\) and every integer \(m\ge0\),
$
L^m u_k
=
\lambda_k^m u_k.
$
Indeed, this is immediate for \(m=0\), while for \(m\ge1\) it follows
repeatedly from \(Lu_k=\lambda_k u_k\). Using the power-series definition of
the matrix exponential,
$
e^{\tau L}
=
\sum_{m=0}^{\infty}
\frac{\tau^mL^m}{m!},
$
we therefore obtain
\[
\begin{aligned}
e^{\tau L}u_k
&=
\sum_{m=0}^{\infty}
\frac{\tau^m}{m!}L^m u_k\\
&=
\sum_{m=0}^{\infty}
\frac{\tau^m}{m!}\lambda_k^m u_k\\
&=
\left(
\sum_{m=0}^{\infty}
\frac{(\tau\lambda_k)^m}{m!}
\right)u_k\\
&=
e^{\tau\lambda_k}u_k.
\end{aligned}
\]
Applying this identity to the eigen-expansion of \(h\) gives
\[
\begin{aligned}
e^{\tau L}h
&=
e^{\tau L}
\left(
\sum_{k=1}^d\alpha_k u_k
\right)\\
&=
\sum_{k=1}^d
\alpha_k e^{\tau L}u_k\\
&=
\sum_{k=1}^d
\alpha_k e^{\tau\lambda_k}u_k.
\end{aligned}
\]
Since \(\lambda_1=0\),
$
e^{\tau\lambda_1}=1,
$
and since \(\alpha_1u_1=\overline h\), this becomes
\[
e^{\tau L}h
=
\overline h
+
\sum_{k=2}^d
\alpha_k e^{\tau\lambda_k}u_k.
\]
Consequently,
$
e^{\tau L}h-\overline h
=
\sum_{k=2}^d
\alpha_k e^{\tau\lambda_k}u_k.
$

Using orthonormality once more,
\[
\begin{aligned}
\|e^{\tau L}h-\overline h\|_2^2
&=
\left\|
\sum_{k=2}^d
\alpha_k e^{\tau\lambda_k}u_k
\right\|_2^2\\
&=
\sum_{k=2}^d
\alpha_k^2 e^{2\tau\lambda_k}.
\end{aligned}
\]
For every \(k\ge2\), the ordering of the eigenvalues gives
$
\lambda_k\le\lambda_2.
$
Since
$
\lambda_2=-\gamma
$
and \(\tau\ge0\), multiplication by \(2\tau\) preserves the inequality:
\[
2\tau\lambda_k
\le
2\tau\lambda_2
=
-2\gamma\tau.
\]
The scalar exponential is increasing, so
$
e^{2\tau\lambda_k}
\le
e^{-2\gamma\tau}
\qquad
\text{for every } k\ge2.
$
Substituting this bound into the preceding norm identity yields
\[
\begin{aligned}
\|e^{\tau L}h-\overline h\|_2^2
&=
\sum_{k=2}^d
\alpha_k^2 e^{2\tau\lambda_k}\\
&\le
\sum_{k=2}^d
\alpha_k^2 e^{-2\gamma\tau}\\
&=
e^{-2\gamma\tau}
\sum_{k=2}^d\alpha_k^2.
\end{aligned}
\]
Using
$
\sum_{k=2}^d\alpha_k^2
=
\|h-\overline h\|_2^2,
$
we arrive at
\[
\|e^{\tau L}h-\overline h\|_2^2
\le
e^{-2\gamma\tau}
\|h-\overline h\|_2^2.
\]
Both sides are nonnegative, so taking square roots gives
\[
\|e^{\tau L}h-\overline h\|_2
\le
e^{-\gamma\tau}
\|h-\overline h\|_2.
\]
This is the claimed exponential attenuation of all nonconstant modes.
\end{proof}

\subsection{Proof of Proposition~\ref{prop:properties-propagation-gradient-attenuation}}

\begin{proof}
Throughout the proof, let \(\tau\ge0\). Under the assumptions of
Theorem~\ref{thm:properties-propagation-oversmoothing}, the generator \(L\)
is real symmetric,
$
L=L^\top,
$
satisfies
$
L\mathbf1=0,
$
and has
$
\ker(L)=\operatorname{span}\{\mathbf1\}.
$
Its eigenvalues may therefore be ordered as
\[
0=\lambda_1>\lambda_2\ge\cdots\ge\lambda_d,
\qquad
\gamma=-\lambda_2>0.
\]
As established in the proof of
Theorem~\ref{thm:properties-propagation-oversmoothing}, every vector
\(v\in\mathbf1^\perp\) satisfies
$
\|e^{tL}v\|_2
\le
e^{-\gamma t}\|v\|_2
\qquad
\text{for every }t\ge0.
$
Indeed, \(v\in\mathbf1^\perp\) has no component in the zero eigenspace of
\(L\), so only the eigenmodes associated with
\(\lambda_2,\ldots,\lambda_d\) remain.

We first consider differentiation with respect to the propagation magnitude.
By the power-series definition of the matrix exponential,
\[
P_\tau
=
e^{\tau L}
=
\sum_{m=0}^{\infty}\frac{\tau^mL^m}{m!}.
\]
Since this series and its termwise derivative converge absolutely on every
bounded interval of \(\tau\), we may differentiate term by term:
\[
\begin{aligned}
\frac{\partial}{\partial\tau}P_\tau
&=
\sum_{m=1}^{\infty}
\frac{m\tau^{m-1}L^m}{m!}\\
&=
\sum_{m=1}^{\infty}
\frac{\tau^{m-1}L^m}{(m-1)!}.
\end{aligned}
\]
Writing \(q=m-1\), this becomes
\[
\begin{aligned}
\frac{\partial}{\partial\tau}P_\tau
&=
\sum_{q=0}^{\infty}
\frac{\tau^qL^{q+1}}{q!}\\
&=
L
\sum_{q=0}^{\infty}
\frac{\tau^qL^q}{q!}\\
&=
Le^{\tau L}.
\end{aligned}
\]
Because every power of \(L\) commutes with \(L\), the same calculation also
gives
\[
\frac{\partial}{\partial\tau}P_\tau
=
e^{\tau L}L.
\]
Hence, for every \(h\in\mathbb R^d\),
$
\frac{\partial}{\partial\tau}P_\tau h
=
Le^{\tau L}h.
$

Recall that
\[
\overline h
=
\frac{\mathbf1^\top h}{d}\mathbf1.
\]
Since \(L\mathbf1=0\),
$
L\overline h
=
\frac{\mathbf1^\top h}{d}L\mathbf1
=
0.
$
Moreover, because \(L\) commutes with \(e^{\tau L}\),
$
Le^{\tau L}\overline h
=
e^{\tau L}L\overline h
=
0.
$
We may therefore subtract the constant component of \(h\) without changing
the derivative:
\[
\begin{aligned}
\frac{\partial}{\partial\tau}P_\tau h
&=
Le^{\tau L}h\\
&=
Le^{\tau L}
\bigl((h-\overline h)+\overline h\bigr)\\
&=
Le^{\tau L}(h-\overline h)
+
Le^{\tau L}\overline h\\
&=
Le^{\tau L}(h-\overline h).
\end{aligned}
\]
By construction,
\[
\mathbf1^\top(h-\overline h)
=
\mathbf1^\top h
-
\frac{\mathbf1^\top h}{d}\mathbf1^\top\mathbf1
=
\mathbf1^\top h
-
\frac{\mathbf1^\top h}{d}d
=
0,
\]
so
$
h-\overline h\in\mathbf1^\perp.
$
Using the definition of the induced Euclidean operator norm,
$
\|Lv\|_2\le\|L\|_2\|v\|_2,
$
we obtain
\[
\left\|
\frac{\partial}{\partial\tau}P_\tau h
\right\|_2
=
\|Le^{\tau L}(h-\overline h)\|_2
\le
\|L\|_2
\|e^{\tau L}(h-\overline h)\|_2.
\]
Since \(h-\overline h\in\mathbf1^\perp\), the nonconstant-mode contraction
from Theorem~\ref{thm:properties-propagation-oversmoothing} yields
\[
\|e^{\tau L}(h-\overline h)\|_2
\le
e^{-\gamma\tau}
\|h-\overline h\|_2.
\]
Consequently,
$
\left\|
\frac{\partial}{\partial\tau}P_\tau h
\right\|_2
\le
\|L\|_2e^{-\gamma\tau}
\|h-\overline h\|_2,
$
which proves the first assertion.

We now turn to the derivative with respect to the generator. To make the
dependence on \(L\) explicit, consider the matrix-valued map
$
\mathcal E_\tau(M):=e^{\tau M}.
$
We first derive its Fr\'echet derivative at \(M=L\) in the direction \(H\).
For every positive integer \(m\), the directional derivative of the matrix
power \(M^m\) at \(L\) is
\[
D_L(M^m)[H]
=
\sum_{q=0}^{m-1}
L^{m-1-q}HL^q.
\]
To see this directly, expand
$
(L+\varepsilon H)^m.
$
The terms containing no occurrence of \(H\) give \(L^m\). The terms
containing exactly one occurrence of \(\varepsilon H\) are
$
\varepsilon
\sum_{q=0}^{m-1}
L^{m-1-q}HL^q,
$
where \(q\) counts the number of factors \(L\) to the right of \(H\). Every
remaining term contains at least two factors \(\varepsilon H\) and is
therefore of order \(O(\varepsilon^2)\) as \(\varepsilon\to0\). Hence
\[
(L+\varepsilon H)^m
=
L^m
+
\varepsilon
\sum_{q=0}^{m-1}
L^{m-1-q}HL^q
+
O(\varepsilon^2),
\]
which gives the stated derivative.

Applying this identity termwise to
$
e^{\tau L}
=
\sum_{m=0}^{\infty}
\frac{\tau^mL^m}{m!},
$
and using absolute convergence of the exponential series in finite
dimensions, we obtain
\[
D_LP_\tau[H]
=
\sum_{m=1}^{\infty}
\frac{\tau^m}{m!}
\sum_{q=0}^{m-1}
L^{m-1-q}HL^q.
\]
We now rewrite this series in an integral form that is better suited to the
spectral estimate. Expanding both exponentials in
\[
\tau\int_0^1
e^{(1-s)\tau L}
H
e^{s\tau L}
\,ds
\]
gives
\[
\begin{aligned}
&\tau\int_0^1
\left(
\sum_{a=0}^{\infty}
\frac{((1-s)\tau)^aL^a}{a!}
\right)
H
\left(
\sum_{b=0}^{\infty}
\frac{(s\tau)^bL^b}{b!}
\right)
\,ds\\
&\qquad=
\tau
\sum_{a=0}^{\infty}
\sum_{b=0}^{\infty}
\frac{\tau^{a+b}}{a!\,b!}
L^aHL^b
\int_0^1
(1-s)^a s^b\,ds.
\end{aligned}
\]
The interchange of the sums and the integral is justified because the two
matrix exponential series converge absolutely and uniformly for
\(s\in[0,1]\). For nonnegative integers \(a,b\), the beta-integral identity
gives
\[
\int_0^1
(1-s)^a s^b\,ds
=
\frac{a!\,b!}{(a+b+1)!}.
\]
Therefore
\[
\begin{aligned}
\tau\int_0^1
e^{(1-s)\tau L}
H
e^{s\tau L}
\,ds
&=
\sum_{a=0}^{\infty}
\sum_{b=0}^{\infty}
\frac{\tau^{a+b+1}}{(a+b+1)!}
L^aHL^b.
\end{aligned}
\]
Setting
$
m=a+b+1
\qquad\text{and}\qquad
q=b
$
shows that \(a=m-1-q\), with \(q=0,\ldots,m-1\). Thus the last expression is
exactly
\[
\sum_{m=1}^{\infty}
\frac{\tau^m}{m!}
\sum_{q=0}^{m-1}
L^{m-1-q}HL^q.
\]
Comparing with the preceding expression for the Fr\'echet derivative yields
\[
D_LP_\tau[H]
=
\tau\int_0^1
e^{(1-s)\tau L}
H
e^{s\tau L}
\,ds.
\]

We now exploit the symmetry assumptions. Let
\(x\in\mathbb R^d\) be arbitrary and define
$
\overline x
:=
\frac{\mathbf1^\top x}{d}\mathbf1,
\qquad
x_\perp
:=
x-\overline x.
$
Then
$
\mathbf1^\top x_\perp=0,
$
so$
x_\perp\in\mathbf1^\perp.
$
Moreover, \(\overline x\) and \(x_\perp\) are orthogonal because
\(\overline x\in\operatorname{span}\{\mathbf1\}\) and
\(x_\perp\in\mathbf1^\perp\). Hence
$
\|x\|_2^2
=
\|\overline x\|_2^2
+
\|x_\perp\|_2^2,
$
and in particular
$
\|x_\perp\|_2\le\|x\|_2.
$

Since \(L\mathbf1=0\),
$
e^{s\tau L}\mathbf1=\mathbf1
\qquad
\text{for every }s\in[0,1].
$
Thus
$
e^{s\tau L}\overline x
=
\overline x.
$
Because \(H\mathbf1=0\), every vector proportional to \(\mathbf1\) lies in
the kernel of \(H\), and therefore
$
He^{s\tau L}\overline x
=
H\overline x
=
0.
$
It follows that
$
He^{s\tau L}x
=
He^{s\tau L}x_\perp.
$
Consequently,
\[
D_LP_\tau[H]x
=
\tau\int_0^1
e^{(1-s)\tau L}
H
e^{s\tau L}x_\perp
\,ds.
\]

Because \(L=L^\top\) and \(L\mathbf1=0\),
$
\mathbf1^\top L
=
(L\mathbf1)^\top
=
0.
$
Hence
$
\mathbf1^\top e^{tL}
=
\mathbf1^\top
\qquad
\text{for every }t\ge0,
$
and therefore \(e^{tL}\) leaves \(\mathbf1^\perp\) invariant. In particular,
since \(x_\perp\in\mathbf1^\perp\),
$
e^{s\tau L}x_\perp\in\mathbf1^\perp.
$
The perturbation \(H\) is also symmetric. Thus
$
\mathbf1^\top H
=
(H\mathbf1)^\top
=
0.
$
It follows that, for every \(y\in\mathbb R^d\),
$
\mathbf1^\top Hy
=
0,
$
so
$
Hy\in\mathbf1^\perp.
$
In particular,
$
H e^{s\tau L}x_\perp
\in
\mathbf1^\perp.
$

We can therefore apply the nonconstant-mode contraction estimate to both
propagation factors appearing in the integrand. First,
\[
\|e^{s\tau L}x_\perp\|_2
\le
e^{-s\gamma\tau}\|x_\perp\|_2.
\]
By the definition of the matrix \(2\)-norm,
$
\|Hy\|_2
\le
\|H\|_2\|y\|_2,
$
so
\[
\begin{aligned}
\|H e^{s\tau L}x_\perp\|_2
&\le
\|H\|_2
\|e^{s\tau L}x_\perp\|_2\\
&\le
\|H\|_2
e^{-s\gamma\tau}
\|x_\perp\|_2.
\end{aligned}
\]
Since
\(H e^{s\tau L}x_\perp\in\mathbf1^\perp\), applying the same spectral
contraction for propagation time \((1-s)\tau\ge0\) gives
\[
\begin{aligned}
&
\left\|
e^{(1-s)\tau L}
H e^{s\tau L}x_\perp
\right\|_2\\
&\qquad\le
e^{-(1-s)\gamma\tau}
\left\|
H e^{s\tau L}x_\perp
\right\|_2\\
&\qquad\le
e^{-(1-s)\gamma\tau}
\|H\|_2
e^{-s\gamma\tau}
\|x_\perp\|_2.
\end{aligned}
\]
The two exponential factors combine exactly:
\[
e^{-(1-s)\gamma\tau}
e^{-s\gamma\tau}
=
e^{-((1-s)+s)\gamma\tau}
=
e^{-\gamma\tau}.
\]
Hence, for every \(s\in[0,1]\),
$
\left\|
e^{(1-s)\tau L}
H e^{s\tau L}x_\perp
\right\|_2
\le
e^{-\gamma\tau}
\|H\|_2
\|x_\perp\|_2.
$

Using the integral representation of the Fr\'echet derivative and the triangle
inequality for vector-valued integrals,
\[
\begin{aligned}
\|D_LP_\tau[H]x\|_2
&=
\left\|
\tau\int_0^1
e^{(1-s)\tau L}
H e^{s\tau L}x_\perp
\,ds
\right\|_2\\
&\le
\tau\int_0^1
\left\|
e^{(1-s)\tau L}
H e^{s\tau L}x_\perp
\right\|_2
\,ds\\
&\le
\tau\int_0^1
e^{-\gamma\tau}
\|H\|_2
\|x_\perp\|_2
\,ds\\
&=
\tau e^{-\gamma\tau}
\|H\|_2
\|x_\perp\|_2.
\end{aligned}
\]
Since
$
\|x_\perp\|_2\le\|x\|_2,
$
we conclude that
$
\|D_LP_\tau[H]x\|_2
\le
\tau e^{-\gamma\tau}
\|H\|_2
\|x\|_2.
$
Taking the supremum over all nonzero
\(x\in\mathbb R^d\) yields
\[
\begin{aligned}
\|D_LP_\tau[H]\|_{2\to2}
&=
\sup_{x\neq0}
\frac{\|D_LP_\tau[H]x\|_2}{\|x\|_2}\\
&\le
\tau e^{-\gamma\tau}\|H\|_2.
\end{aligned}
\]

Thus the derivative with respect to the propagation magnitude is suppressed
on the nonconstant component of the signal by the factor
\(e^{-\gamma\tau}\), while symmetric generator perturbations preserving the
constant mode are suppressed by
\(\tau e^{-\gamma\tau}\). This establishes both claimed gradient attenuation
bounds.
\end{proof}

\subsection{Proof of Theorem~\ref{thm:properties-reachability-preserving-propagation}}

\begin{proof}
Let \(L\in\mathcal L_{\mathfrak S}\), let \(\tau>0\), and choose
\(\nu>0\) such that
$
\nu\ge \max_{1\le i\le d}(-L_{ii}).
$
By Definition~\ref{def:operators-generator-family},
$
L_{ij}\ge0
\qquad
\text{for every }i\neq j,
$
and
$
L\mathbf1=0.
$
The latter identity implies, for every \(i\),
\[
0
=
(L\mathbf1)_i
=
\sum_{j=1}^dL_{ij}
=
L_{ii}+\sum_{j\neq i}L_{ij}.
\]
Hence
$
L_{ii}
=
-\sum_{j\neq i}L_{ij}
\le0.
$
Define
$
T
:=
I+\frac{L}{\nu}.
$
If \(i\neq j\), then
\[
T_{ij}
=
\frac{L_{ij}}{\nu}
\ge0.
\]
For a diagonal entry,
\[
T_{ii}
=
1+\frac{L_{ii}}{\nu}
=
1-\frac{-L_{ii}}{\nu}.
\]
Since
$
-L_{ii}
\le
\max_{\ell}(-L_{\ell\ell})
\le
\nu,
$
we have
$
T_{ii}\ge0.
$
Therefore
$
T\ge0
$
entrywise. Moreover,
\[
T\mathbf1
=
\left(I+\frac{L}{\nu}\right)\mathbf1
=
\mathbf1+\frac{1}{\nu}L\mathbf1
=
\mathbf1.
\]
Thus every row of \(T\) sums to one. In particular, \(T\) is a nonnegative
row-stochastic matrix.

From the definition of \(T\),
$
T-I=\frac{L}{\nu},
$
and hence
$
L=\nu(T-I).
$
Because \(T\) commutes with the identity matrix,
\[
\begin{aligned}
e^{\tau L}
&=
e^{\nu\tau(T-I)}\\
&=
e^{\nu\tau T-\nu\tau I}\\
&=
e^{-\nu\tau I}e^{\nu\tau T}\\
&=
e^{-\nu\tau}
\sum_{k=0}^{\infty}
\frac{(\nu\tau)^k}{k!}T^k.
\end{aligned}
\]
Consequently, for every \(i,j\),
\[
(e^{\tau L})_{ij}
=
e^{-\nu\tau}
\sum_{k=0}^{\infty}
\frac{(\nu\tau)^k}{k!}(T^k)_{ij}.
\]
All coefficients in this series are nonnegative, and every matrix
\(T^k\) is entrywise nonnegative because \(T\ge0\).

We now relate the entries of the powers of \(T\) to directed paths in
\(\mathcal E_L\). Recall that
$
\mathcal E_L
=
\{(z_a,z_b):a\neq b,\ L_{ab}>0\}.
$
Under the standing receiver--source convention,
\((z_a,z_b)\in\mathcal E_L\) means that there is a directed edge from
the source \(z_b\) to the receiver \(z_a\). For distinct indices
\(a\neq b\),
$
T_{ab}
=
\frac{L_{ab}}{\nu},
$
and therefore
\[
T_{ab}>0
\quad\Longleftrightarrow\quad
L_{ab}>0
\quad\Longleftrightarrow\quad
(z_a,z_b)\in\mathcal E_L.
\]
Thus every positive off-diagonal entry of \(T\) represents exactly one
directed edge of \(\mathcal E_L\), whereas a positive diagonal entry
\(T_{aa}\) represents only a stay at the same site.

For \(k\ge1\), repeated matrix multiplication gives
\[
(T^k)_{ij}
=
\sum_{a_1,\ldots,a_{k-1}=1}^d
T_{i a_{k-1}}
T_{a_{k-1}a_{k-2}}
\cdots
T_{a_2a_1}
T_{a_1j}.
\]
Every summand is nonnegative. Consequently,
$
(T^k)_{ij}>0
$
if and only if at least one sequence
\[
j=a_0,a_1,\ldots,a_{k-1},a_k=i
\]
satisfies
$
T_{a_\ell a_{\ell-1}}>0
\qquad
\text{for every }\ell=1,\ldots,k.
$
Such a sequence is a length-\(k\) walk for the matrix \(T\). Whenever
\(a_\ell\neq a_{\ell-1}\), the positivity
$
T_{a_\ell a_{\ell-1}}>0
$
implies
$
(z_{a_\ell},z_{a_{\ell-1}})\in\mathcal E_L,
$
so this nontrivial move is a directed edge
$
z_{a_{\ell-1}}\longrightarrow z_{a_\ell}
$
in \(\mathcal E_L\). Whenever
\(a_\ell=a_{\ell-1}\), the corresponding factor is diagonal and represents
only a holding move.

Suppose first that
$
(T^k)_{ij}>0
$
for some \(k\ge1\), with \(i\neq j\). Choose a sequence
$
j=a_0,a_1,\ldots,a_k=i
$
whose corresponding product is strictly positive. Since \(i\neq j\), not
all consecutive indices can be equal. Delete from this sequence every
holding move for which
$
a_\ell=a_{\ell-1}.
$
The remaining sequence starts at \(j\), ends at \(i\), has positive length,
and every consecutive pair in it corresponds to a positive off-diagonal
entry of \(T\). Hence every such pair belongs to \(\mathcal E_L\).
Therefore the reduced sequence is a directed path from \(z_j\) to \(z_i\)
in \(\mathcal E_L\). We have proved that
\[
(T^k)_{ij}>0
\quad\Longrightarrow\quad
\text{there exists a directed path from \(z_j\) to \(z_i\) in
\(\mathcal E_L\)}.
\]

Conversely, suppose that there exists a directed path from \(z_j\) to
\(z_i\) in \(\mathcal E_L\). Write one such path as
\[
z_j=z_{a_0}
\longrightarrow
z_{a_1}
\longrightarrow
\cdots
\longrightarrow
z_{a_r}=z_i,
\]
where \(r\ge1\). By the definition of \(\mathcal E_L\),
$
L_{a_\ell a_{\ell-1}}>0
\qquad
\text{for }\ell=1,\ldots,r.
$
Since \(\nu>0\),
$
T_{a_\ell a_{\ell-1}}
=
\frac{L_{a_\ell a_{\ell-1}}}{\nu}
>
0
\qquad
\text{for }\ell=1,\ldots,r.
$
Hence
$
\prod_{\ell=1}^r
T_{a_\ell a_{\ell-1}}
>
0.
$
This product appears as one of the nonnegative summands in the matrix-product
expansion
\[
(T^r)_{ij}
=
\sum_{b_1,\ldots,b_{r-1}=1}^d
T_{i b_{r-1}}
\cdots
T_{b_1j}.
\]
Therefore
$
(T^r)_{ij}
\ge
\prod_{\ell=1}^r
T_{a_\ell a_{\ell-1}}
>
0.
$
Using the uniformization series,
\[
(e^{\tau L})_{ij}
=
e^{-\nu\tau}
\sum_{k=0}^{\infty}
\frac{(\nu\tau)^k}{k!}(T^k)_{ij}.
\]
Every term is nonnegative, while the term corresponding to \(k=r\) is
strictly positive because
$
e^{-\nu\tau}>0,
\qquad
\frac{(\nu\tau)^r}{r!}>0,
\qquad
(T^r)_{ij}>0,
$
where we used \(\nu>0\) and \(\tau>0\). Thus
$
(e^{\tau L})_{ij}>0.
$

We have therefore shown, for \(i\neq j\),
\[
(e^{\tau L})_{ij}>0
\quad\Longleftrightarrow\quad
\text{there exists a directed path from \(z_j\) to \(z_i\) in
\(\mathcal E_L\)}.
\]

Assume now that such a path exists and that its shortest possible length is
\(r\). We claim that
$
(T^k)_{ij}=0
\qquad
\text{for every }k<r.
$
Indeed, if \((T^k)_{ij}>0\) for some \(k<r\), the argument above would
produce, after deleting holding moves, a directed path from \(z_j\) to
\(z_i\) in \(\mathcal E_L\) whose length is at most \(k\). This would give a
path of length strictly smaller than \(r\), contradicting the definition of
\(r\). Hence
$
(T^k)_{ij}=0
\qquad
\text{for }0\le k<r.
$
Here the case \(k=0\) also satisfies the identity because \(i\neq j\) and
$
(T^0)_{ij}
=
I_{ij}
=
0.
$

Since \(T\mathbf1=\mathbf1\), induction gives
$
T^k\mathbf1=\mathbf1
\qquad
\text{for every }k\ge0.
$
Because \(T^k\ge0\), every row of \(T^k\) consists of nonnegative numbers
summing to one. In particular,
$
0\le (T^k)_{ij}\le1
\qquad
\text{for every }i,j,k.
$
Using the vanishing of all terms with \(k<r\), we obtain
\[
\begin{aligned}
(e^{\tau L})_{ij}
&=
e^{-\nu\tau}
\sum_{k=r}^{\infty}
\frac{(\nu\tau)^k}{k!}(T^k)_{ij}\\
&\le
e^{-\nu\tau}
\sum_{k=r}^{\infty}
\frac{(\nu\tau)^k}{k!}.
\end{aligned}
\]
If
$
N\sim\operatorname{Poisson}(\nu\tau),
$
then
\[
\Pr\{N=k\}
=
e^{-\nu\tau}
\frac{(\nu\tau)^k}{k!},
\qquad
k=0,1,2,\ldots,
\]
and therefore
\[
e^{-\nu\tau}
\sum_{k=r}^{\infty}
\frac{(\nu\tau)^k}{k!}
=
\Pr\{N\ge r\}.
\]
Thus
$
(e^{\tau L})_{ij}
\le
\Pr\{\operatorname{Poisson}(\nu\tau)\ge r\}.
$

Finally, suppose that no directed path from \(z_j\) to \(z_i\) exists in
\(\mathcal E_L\). If there were some \(k\ge1\) such that
$
(T^k)_{ij}>0,
$
the walk argument above would produce such a directed path, a contradiction.
Therefore
$
(T^k)_{ij}=0
\qquad
\text{for every }k\ge1.
$
Since \(i\neq j\),
$
(T^0)_{ij}=I_{ij}=0
$
as well. Substituting into the uniformization expansion gives
\[
(e^{\tau L})_{ij}
=
e^{-\nu\tau}
\sum_{k=0}^{\infty}
\frac{(\nu\tau)^k}{k!}(T^k)_{ij}
=
0.
\]
This proves the reachability characterization, the Poisson-tail bound, and
the zero-entry statement.
\end{proof}

\subsection{Proof of Proposition~\ref{prop:properties-composable-envelope}}

\begin{proof}
Recall that
$
\mathcal R_{\mathcal E}
=
\Delta_X
\cup
\left\{
(z_i,z_j):z_j\rightsquigarrow z_i
\right\}
$
is the reflexive transitive closure of the admissibility relation
\(\mathcal E\), and that
\[
\mathscr C_{\mathcal E;U,V}
=
\left\{
P\in\mathbb R^{d\times d}:
P\ge0,\;
P\mathbf1=\mathbf1,\;
PU=U,\;
V^\top P=V^\top,\;
\operatorname{supp}(P)\subseteq\mathcal R_{\mathcal E}
\right\}.
\]
We first verify that this set is nonempty. Consider the identity matrix
\(I\in\mathbb R^{d\times d}\). Clearly,
$
I\ge0
$
entrywise and
$
I\mathbf1=\mathbf1.
$
Moreover,
$
IU=U
$
and
$
V^\top I=V^\top.
$
The support of the identity consists only of diagonal pairs:
\[
\operatorname{supp}(I)
=
\Delta_X.
\]
Since \(\mathcal R_{\mathcal E}\) is reflexive by definition,
$
\Delta_X\subseteq\mathcal R_{\mathcal E}.
$
Hence
$
\operatorname{supp}(I)
\subseteq
\mathcal R_{\mathcal E},
$
and therefore
$
I\in\mathscr C_{\mathcal E;U,V}.
$
Thus \(\mathscr C_{\mathcal E;U,V}\) is nonempty.

We next establish closedness. The condition \(P\ge0\) consists of the finitely
many inequalities
$
P_{ij}\ge0,
\qquad
1\le i,j\le d,
$
each of which defines a closed half-space in the finite-dimensional vector
space \(\mathbb R^{d\times d}\). The conditions
$
P\mathbf1=\mathbf1,
\qquad
PU=U,
\qquad
V^\top P=V^\top
$
are systems of linear equalities and hence define closed affine subsets of
\(\mathbb R^{d\times d}\). Finally,
\[
\operatorname{supp}(P)
\subseteq
\mathcal R_{\mathcal E}
\]
is equivalent to requiring
$
P_{ij}=0
\qquad
\text{whenever }
(z_i,z_j)\notin\mathcal R_{\mathcal E}.
$
Because \(X=\{z_1,\ldots,z_d\}\) is finite, this is again a finite collection
of linear equality constraints. It follows that
\(\mathscr C_{\mathcal E;U,V}\), being the intersection of finitely many
closed sets, is closed.

The same defining conditions also imply boundedness. Let
\(P\in\mathscr C_{\mathcal E;U,V}\). Since
$
P\ge0
$
and
$
P\mathbf1=\mathbf1,
$
for every row index \(i\),
$
\sum_{j=1}^d P_{ij}=1.
$
Every summand in this equality is nonnegative, so
$
0\le P_{ij}\le1
\qquad
\text{for all }i,j.
$
In particular,
$
P_{ij}^2\le P_{ij}.
$
Therefore
\[
\begin{aligned}
\|P\|_F^2
&=
\sum_{i=1}^d\sum_{j=1}^d P_{ij}^2\\
&\le
\sum_{i=1}^d\sum_{j=1}^d P_{ij}\\
&=
\sum_{i=1}^d 1\\
&=
d.
\end{aligned}
\]
Hence
$
\|P\|_F\le\sqrt d
\qquad
\text{for every }
P\in\mathscr C_{\mathcal E;U,V},
$
so the set is bounded. Since
\(\mathbb R^{d\times d}\) is finite-dimensional, the Heine--Borel theorem
implies that every closed and bounded subset is compact. Consequently,
$
\mathscr C_{\mathcal E;U,V}
$
is compact.

We now verify convexity. Let
$
P,Q\in\mathscr C_{\mathcal E;U,V}
$
and let
$
\alpha\in[0,1].
$
Define
\[
R
:=
\alpha P+(1-\alpha)Q.
\]
Since \(P\ge0\), \(Q\ge0\), and both scalar coefficients are nonnegative,
$
R\ge0.
$
Furthermore,
\[
\begin{aligned}
R\mathbf1
&=
\alpha P\mathbf1
+
(1-\alpha)Q\mathbf1\\
&=
\alpha\mathbf1+(1-\alpha)\mathbf1\\
&=
\mathbf1.
\end{aligned}
\]
Similarly,
\[
\begin{aligned}
RU
&=
\alpha PU+(1-\alpha)QU\\
&=
\alpha U+(1-\alpha)U\\
&=
U,
\end{aligned}
\]
and
\[
\begin{aligned}
V^\top R
&=
\alpha V^\top P
+
(1-\alpha)V^\top Q\\
&=
\alpha V^\top+(1-\alpha)V^\top\\
&=
V^\top.
\end{aligned}
\]
Now take any pair
$
(z_i,z_j)\notin\mathcal R_{\mathcal E}.
$
Because
$
\operatorname{supp}(P)
\subseteq
\mathcal R_{\mathcal E}
\qquad\text{and}\qquad
\operatorname{supp}(Q)
\subseteq
\mathcal R_{\mathcal E},
$
we have
$
P_{ij}=0
\qquad\text{and}\qquad
Q_{ij}=0.
$
Therefore
\[
R_{ij}
=
\alpha P_{ij}
+
(1-\alpha)Q_{ij}
=
0.
\]
Thus
$
\operatorname{supp}(R)
\subseteq
\mathcal R_{\mathcal E}.
$
We conclude that
$
R\in\mathscr C_{\mathcal E;U,V},
$
and hence \(\mathscr C_{\mathcal E;U,V}\) is convex.

We next prove closure under matrix multiplication. Let
$
P,Q\in\mathscr C_{\mathcal E;U,V}
$
and set
$
R:=PQ.
$
For every \(i,j\),
\[
R_{ij}
=
(PQ)_{ij}
=
\sum_{k=1}^dP_{ik}Q_{kj}.
\]
Since
$
P_{ik}\ge0
\qquad\text{and}\qquad
Q_{kj}\ge0
$
for every \(i,j,k\), every summand in the preceding expression is
nonnegative. Hence
$
R_{ij}\ge0
\qquad
\text{for all }i,j,
$
and therefore
$
R\ge0.
$
The constant vector is preserved because
\[
\begin{aligned}
R\mathbf1
&=
PQ\mathbf1\\
&=
P(Q\mathbf1)\\
&=
P\mathbf1\\
&=
\mathbf1.
\end{aligned}
\]
The right modes are preserved because
\[
\begin{aligned}
RU
&=
PQU\\
&=
P(QU)\\
&=
PU\\
&=
U.
\end{aligned}
\]
Likewise, the left observables are preserved:
\[
\begin{aligned}
V^\top R
&=
V^\top PQ\\
&=
(V^\top P)Q\\
&=
V^\top Q\\
&=
V^\top.
\end{aligned}
\]

It remains to check the support of the product. Suppose that
$
R_{ij}>0.
$
Since
$
R_{ij}
=
\sum_{k=1}^dP_{ik}Q_{kj}
$
is a finite sum of nonnegative terms, the strict inequality \(R_{ij}>0\)
implies that there exists at least one index
\(k\in\{1,\ldots,d\}\) such that
$
P_{ik}Q_{kj}>0.
$
Because both factors are nonnegative, this implies simultaneously
$
P_{ik}>0
\qquad\text{and}\qquad
Q_{kj}>0.
$
Hence
$
(z_i,z_k)\in\operatorname{supp}(P)
\subseteq
\mathcal R_{\mathcal E}
$
and
$
(z_k,z_j)\in\operatorname{supp}(Q)
\subseteq
\mathcal R_{\mathcal E}.
$
Under the receiver--source convention, the first relation says that either
\(z_k=z_i\) or \(z_k\rightsquigarrow z_i\), while the second says that
either \(z_j=z_k\) or \(z_j\rightsquigarrow z_k\). Since
\(\mathcal R_{\mathcal E}\) is reflexive and transitive, composing these two
relations gives
$
(z_i,z_j)\in\mathcal R_{\mathcal E}.
$
We have therefore shown that
$
R_{ij}>0
\quad\Longrightarrow\quad
(z_i,z_j)\in\mathcal R_{\mathcal E}.
$
Since \(R\ge0\), every nonzero entry of \(R\) is strictly positive, and thus
$
\operatorname{supp}(R)
\subseteq
\mathcal R_{\mathcal E}.
$
Consequently,
$
PQ\in\mathscr C_{\mathcal E;U,V}.
$
Thus \(\mathscr C_{\mathcal E;U,V}\) is closed under matrix multiplication.

We finally consider a finite sequence
\[
Q_1,\ldots,Q_m
\in
\mathcal L_{\mathfrak S;U,V}.
\]
Fix \(\ell\in\{1,\ldots,m\}\). By the definition of the constrained
generator family,
$
Q_\ell\in\mathcal L_{\mathfrak S},
\qquad
Q_\ell U=0,
\qquad
V^\top Q_\ell=0.
$
Applying
Theorem~\ref{thm:properties-order-supnorm-stability}
to \(L=Q_\ell\) and propagation magnitude \(\tau=1\) gives
$
e^{Q_\ell}\ge0
$
and
$
e^{Q_\ell}\mathbf1=\mathbf1.
$

The constraint \(Q_\ell U=0\) implies, for every integer \(n\ge1\),
$
Q_\ell^nU
=
Q_\ell^{n-1}(Q_\ell U)
=
0.
$
Using the matrix-exponential series,
$
e^{Q_\ell}
=
I+
\sum_{n=1}^{\infty}
\frac{Q_\ell^n}{n!},
$
we therefore obtain
\[
\begin{aligned}
e^{Q_\ell}U
&=
\left(
I+
\sum_{n=1}^{\infty}
\frac{Q_\ell^n}{n!}
\right)U\\
&=
U+
\sum_{n=1}^{\infty}
\frac{Q_\ell^nU}{n!}\\
&=
U.
\end{aligned}
\]
Similarly, since
$
V^\top Q_\ell=0,
$
we have for every \(n\ge1\),
$
V^\top Q_\ell^n
=
(V^\top Q_\ell)Q_\ell^{n-1}
=
0,
$
and consequently
\[
\begin{aligned}
V^\top e^{Q_\ell}
&=
V^\top
\left(
I+
\sum_{n=1}^{\infty}
\frac{Q_\ell^n}{n!}
\right)\\
&=
V^\top+
\sum_{n=1}^{\infty}
\frac{V^\top Q_\ell^n}{n!}\\
&=
V^\top.
\end{aligned}
\]

It remains to prove the required support restriction for \(e^{Q_\ell}\).
Define the positive off-diagonal edge relation associated with \(Q_\ell\) by
$
\mathcal E_{Q_\ell}
:=
\left\{
(z_a,z_b):
a\neq b,\;
(Q_\ell)_{ab}>0
\right\}.
$
Since
$
Q_\ell\in\mathcal L_{\mathfrak S}
\subseteq
\mathcal A_{\mathfrak S},
$
Proposition~\ref{prop:operators-one-step-support} implies that, for
\(a\neq b\),
\[
(z_a,z_b)\notin\mathcal E
\quad\Longrightarrow\quad
(Q_\ell)_{ab}=0.
\]
Taking the contrapositive of this implication gives
\[
(Q_\ell)_{ab}\neq0
\quad\Longrightarrow\quad
(z_a,z_b)\in\mathcal E.
\]
In particular,
$
(Q_\ell)_{ab}>0
\quad\Longrightarrow\quad
(z_a,z_b)\in\mathcal E,
$
and hence
$
\mathcal E_{Q_\ell}
\subseteq
\mathcal E.
$

Now let \(i\neq j\) and suppose that
$
(e^{Q_\ell})_{ij}>0.
$
Applying
Theorem~\ref{thm:properties-reachability-preserving-propagation}
with \(L=Q_\ell\) and \(\tau=1\), there exists a directed path
\[
z_j
=
z_{a_0}
\longrightarrow
z_{a_1}
\longrightarrow
\cdots
\longrightarrow
z_{a_r}
=
z_i
\]
whose every directed edge belongs to
\(\mathcal E_{Q_\ell}\). Since
$
\mathcal E_{Q_\ell}
\subseteq
\mathcal E,
$
the same sequence is also a directed path using edges of
\(\mathcal E\). Therefore
$
z_j\rightsquigarrow z_i,
$
which means precisely that
$
(z_i,z_j)\in\mathcal R_{\mathcal E}.
$
If \(i=j\), then automatically
$
(z_i,z_i)\in\Delta_X
\subseteq
\mathcal R_{\mathcal E}.
$
Because \(e^{Q_\ell}\ge0\), every entry belonging to its support is strictly
positive, and the preceding argument therefore proves
$
\operatorname{supp}(e^{Q_\ell})
\subseteq
\mathcal R_{\mathcal E}.
$
Combining all of the established properties yields
$
e^{Q_\ell}
\in
\mathscr C_{\mathcal E;U,V}
\qquad
\text{for every }\ell=1,\ldots,m.
$

Set
$
P_\ell:=e^{Q_\ell}.
$
We have just proved that
$
P_\ell\in\mathscr C_{\mathcal E;U,V}
\qquad
\text{for every }\ell.
$
Since the envelope is closed under matrix multiplication,
$
P_2P_1
\in
\mathscr C_{\mathcal E;U,V}.
$
Multiplying on the left by \(P_3\in\mathscr C_{\mathcal E;U,V}\) and using
the same closure property gives
$
P_3P_2P_1
\in
\mathscr C_{\mathcal E;U,V}.
$
Continuing inductively through the finite sequence yields
$
P_mP_{m-1}\cdots P_1
\in
\mathscr C_{\mathcal E;U,V}.
$
Substituting \(P_\ell=e^{Q_\ell}\), we conclude that
\[
e^{Q_m}\cdots e^{Q_1}
\in
\mathscr C_{\mathcal E;U,V}.
\]
At no point in this argument was any identity of the form
$
Q_kQ_\ell=Q_\ell Q_k
$
used. The conclusion follows from the structural properties of the individual
exponentials and the multiplicative closure of
\(\mathscr C_{\mathcal E;U,V}\), rather than from rewriting the product as a
single exponential. Hence no commutativity assumption is required.
\end{proof}

\subsection{Proof of Theorem~\ref{thm:properties-strict-expressive-restriction}}

\begin{proof}
We first verify the inclusion
$
\mathcal A_{\mathfrak S}
\subseteq
\mathcal M_{\mathcal E}.
$
Let
$
A\in\mathcal A_{\mathfrak S}
$
be arbitrary. By the definition of the ambient masked matrix space,
membership in \(\mathcal M_{\mathcal E}\) requires precisely that
$
A_{ij}=0
\qquad
\text{whenever }
i\neq j
\text{ and }
(z_i,z_j)\notin\mathcal E.
$
But this is exactly the one-step support property established in
Proposition~\ref{prop:operators-one-step-support}. Therefore every
\(A\in\mathcal A_{\mathfrak S}\) satisfies the defining support constraint of
\(\mathcal M_{\mathcal E}\), and hence
$
A\in\mathcal M_{\mathcal E}.
$
Since \(A\) was arbitrary,
$
\mathcal A_{\mathfrak S}
\subseteq
\mathcal M_{\mathcal E}.
$

We now determine the dimension of the ambient comparison space explicitly.
Recall that
$
\mathcal E_{\mathrm{off}}
=
\{(z_i,z_j)\in\mathcal E:\ i\neq j\}.
$
For \(1\le i,j\le d\), let \(E_{ij}\in\mathbb R^{d\times d}\) denote the
standard matrix unit having a \(1\) in position \((i,j)\) and zeros
elsewhere. Consider the collection
\[
\mathcal B_{\mathcal E}
:=
\{E_{11},\ldots,E_{dd}\}
\cup
\{E_{ij}:(z_i,z_j)\in\mathcal E_{\mathrm{off}}\}.
\]
We claim that \(\mathcal B_{\mathcal E}\) is a basis of
\(\mathcal M_{\mathcal E}\).

To see that it spans \(\mathcal M_{\mathcal E}\), let
$
B\in\mathcal M_{\mathcal E}.
$
By the definition of \(\mathcal M_{\mathcal E}\),
$
B_{ij}=0
\qquad
\text{whenever }
i\neq j
\text{ and }
(z_i,z_j)\notin\mathcal E.
$
Consequently, the standard matrix-unit expansion
$
B
=
\sum_{i=1}^d\sum_{j=1}^d B_{ij}E_{ij}
$
contains no nonzero off-diagonal term except those indexed by
\(\mathcal E_{\mathrm{off}}\). Hence it reduces to
\[
B
=
\sum_{i=1}^d B_{ii}E_{ii}
+
\sum_{(z_i,z_j)\in\mathcal E_{\mathrm{off}}}
B_{ij}E_{ij}.
\]
Thus every \(B\in\mathcal M_{\mathcal E}\) belongs to the span of
\(\mathcal B_{\mathcal E}\), and therefore
$
\mathcal M_{\mathcal E}
\subseteq
\operatorname{span}(\mathcal B_{\mathcal E}).
$
Conversely, every diagonal matrix unit \(E_{ii}\) belongs to
\(\mathcal M_{\mathcal E}\), because the definition of
\(\mathcal M_{\mathcal E}\) imposes no restriction on diagonal entries.
Moreover, whenever
$
(z_i,z_j)\in\mathcal E_{\mathrm{off}},
$
the matrix unit \(E_{ij}\) also belongs to \(\mathcal M_{\mathcal E}\),
because its only nonzero off-diagonal entry occurs at an admissible pair.
Since \(\mathcal M_{\mathcal E}\) is closed under linear combinations, this
gives
$
\operatorname{span}(\mathcal B_{\mathcal E})
\subseteq
\mathcal M_{\mathcal E}.
$
Combining the two inclusions yields
$
\mathcal M_{\mathcal E}
=
\operatorname{span}(\mathcal B_{\mathcal E}).
$

It remains to check linear independence. Suppose that
\[
\sum_{i=1}^d a_iE_{ii}
+
\sum_{(z_i,z_j)\in\mathcal E_{\mathrm{off}}}
b_{ij}E_{ij}
=
0.
\]
Looking at the \((i,i)\)-entry of this matrix identity gives
$
a_i=0
\qquad
\text{for every }i=1,\ldots,d.
$
Likewise, for every
\((z_i,z_j)\in\mathcal E_{\mathrm{off}}\), looking at the
\((i,j)\)-entry gives
$
b_{ij}=0.
$
Thus all coefficients vanish, so
\(\mathcal B_{\mathcal E}\) is linearly independent. It is therefore a
basis of \(\mathcal M_{\mathcal E}\).

The first part of \(\mathcal B_{\mathcal E}\) contains exactly \(d\)
matrices,
$
E_{11},\ldots,E_{dd},
$
and the second part contains exactly one matrix unit for each element of
\(\mathcal E_{\mathrm{off}}\). Hence
$
|\mathcal B_{\mathcal E}|
=
d+|\mathcal E_{\mathrm{off}}|,
$
and therefore
\[
\dim(\mathcal M_{\mathcal E})
=
d+|\mathcal E_{\mathrm{off}}|.
\]

We now compare this ambient dimension with the number of degrees of freedom
available to the geometry-induced interaction family. By
Proposition~\ref{prop:properties-affine-hull-dimension}, without requiring
any linear-independence assumption on the channel matrices,
\[
\dim\!\bigl(
\operatorname{span}(\mathcal A_{\mathfrak S})
\bigr)
\le
R+d.
\]
Under the hypothesis of the present theorem,
$
R<|\mathcal E_{\mathrm{off}}|.
$
Adding \(d\) to both sides gives
$
R+d
<
|\mathcal E_{\mathrm{off}}|+d.
$
Combining this strict inequality with the preceding dimension bound yields
\[
\dim\!\bigl(
\operatorname{span}(\mathcal A_{\mathfrak S})
\bigr)
\le
R+d
<
d+|\mathcal E_{\mathrm{off}}|
=
\dim(\mathcal M_{\mathcal E}).
\]
In particular,
$
\dim\!\bigl(
\operatorname{span}(\mathcal A_{\mathfrak S})
\bigr)
<
\dim(\mathcal M_{\mathcal E}).
$

Suppose, for contradiction, that the previously established inclusion were
not strict. Since
$
\mathcal A_{\mathfrak S}
\subseteq
\mathcal M_{\mathcal E},
$
failure of strictness would mean
$
\mathcal A_{\mathfrak S}
=
\mathcal M_{\mathcal E}.
$
Taking linear spans of both sides would give
$
\operatorname{span}(\mathcal A_{\mathfrak S})
=
\operatorname{span}(\mathcal M_{\mathcal E}).
$
The set \(\mathcal M_{\mathcal E}\) is already a linear subspace of
\(\mathbb R^{d\times d}\), so
$
\operatorname{span}(\mathcal M_{\mathcal E})
=
\mathcal M_{\mathcal E}.
$
Consequently, the assumed equality would imply
$
\operatorname{span}(\mathcal A_{\mathfrak S})
=
\mathcal M_{\mathcal E},
$
and hence
$
\dim\!\bigl(
\operatorname{span}(\mathcal A_{\mathfrak S})
\bigr)
=
\dim(\mathcal M_{\mathcal E}).
$
This contradicts the strict dimension inequality
\[
\dim\!\bigl(
\operatorname{span}(\mathcal A_{\mathfrak S})
\bigr)
<
\dim(\mathcal M_{\mathcal E})
\]
derived above. Therefore equality is impossible, and the inclusion must be
proper:
$
\mathcal A_{\mathfrak S}
\subsetneq
\mathcal M_{\mathcal E}.
$
Thus, whenever
$
R<|\mathcal E_{\mathrm{off}}|,
$
the reuse of \(R\) geometry channels imposes a genuine single-layer
expressive restriction beyond the hard admissibility mask alone.
\end{proof}

\subsection{Proof of Theorem~\ref{thm:main_best_structured_approx}}

\begin{proof}
Fix an arbitrary target matrix
$
B\in\mathbb R^{d\times d}.
$
We first verify directly the structural properties of
\(\mathcal A_{\mathfrak S}\) that are needed for the projection argument.
By Definition~\ref{def:operators-interaction-family},
\[
\mathcal A_{\mathfrak S}
=
\left\{
\sum_{r=1}^R\theta_r\Phi_r+\operatorname{Diag}(\delta)
\;\middle|\;
\theta\in\mathbb R^R,\;
\delta\in\mathbb R^d,\;
\theta_r\ge0
\text{ for }r\in\mathcal R_+
\right\}.
\]
Taking
$
\theta=0,
\qquad
\delta=0,
$
gives
$
0\in\mathcal A_{\mathfrak S},
$
so the feasible set is nonempty.

It is also convex. Indeed, let
$
A^{(1)},A^{(2)}
\in
\mathcal A_{\mathfrak S}.
$
Then there exist coefficient vectors
\(\theta^{(1)},\theta^{(2)}\in\mathbb R^R\) and
\(\delta^{(1)},\delta^{(2)}\in\mathbb R^d\) such that
\[
A^{(\ell)}
=
\sum_{r=1}^R
\theta_r^{(\ell)}\Phi_r
+
\operatorname{Diag}(\delta^{(\ell)}),
\qquad
\ell\in\{1,2\},
\]
with
$
\theta_r^{(\ell)}\ge0
\qquad
\text{for every }
r\in\mathcal R_+.
$
Let \(t\in[0,1]\). Then
\[
\begin{aligned}
tA^{(1)}+(1-t)A^{(2)}
&=
\sum_{r=1}^R
\left(
t\theta_r^{(1)}
+
(1-t)\theta_r^{(2)}
\right)\Phi_r\\
&\qquad+
\operatorname{Diag}
\left(
t\delta^{(1)}
+
(1-t)\delta^{(2)}
\right).
\end{aligned}
\]
For every \(r\in\mathcal R_+\),
$
t\theta_r^{(1)}
+
(1-t)\theta_r^{(2)}
\ge0,
$
because both coefficients are nonnegative and
\(t,1-t\ge0\). Therefore
\[
tA^{(1)}+(1-t)A^{(2)}
\in
\mathcal A_{\mathfrak S}.
\]
Hence \(\mathcal A_{\mathfrak S}\) is convex.

For completeness, we also establish closedness directly from the finite
coefficient representation. Let \(E_{ii}\) denote the standard matrix unit
with a single \(1\) in position \((i,i)\). Since
$
\operatorname{Diag}(\delta)
=
\sum_{i=1}^d\delta_iE_{ii},
$
an unrestricted scalar coefficient can be decomposed into its positive and
negative parts. In particular, for \(r\notin\mathcal R_+\),
\[
\theta_r
=
\theta_r^+-\theta_r^-,
\qquad
\theta_r^+
:=
\max\{\theta_r,0\},
\qquad
\theta_r^-
:=
\max\{-\theta_r,0\},
\]
with
$
\theta_r^+\ge0,
\qquad
\theta_r^-\ge0.
$
Likewise,
$
\delta_i
=
\delta_i^+-\delta_i^-,
\qquad
\delta_i^\pm\ge0.
$
It follows that \(\mathcal A_{\mathfrak S}\) is exactly the conic hull of
the finite collection
\[
\mathcal G
:=
\{\Phi_r:r\in\mathcal R_+\}
\cup
\{\Phi_r,-\Phi_r:r\notin\mathcal R_+\}
\cup
\{E_{ii},-E_{ii}:1\le i\le d\}.
\]
Indeed, every admissible coefficient representation can be rewritten as a
nonnegative linear combination of matrices in \(\mathcal G\), and conversely
every nonnegative linear combination of matrices in \(\mathcal G\) produces
an element of \(\mathcal A_{\mathfrak S}\). Thus, after enumerating the
elements of \(\mathcal G\) as
$
G_1,\ldots,G_N,
$
we have
\[
\mathcal A_{\mathfrak S}
=
\left\{
\sum_{a=1}^N c_aG_a:
c_a\ge0
\right\}.
\]

We now prove directly that this finitely generated cone is closed. Let
$
A_n\in\mathcal A_{\mathfrak S},
\qquad
A_n\longrightarrow A
$
in Frobenius norm. For every \(n\), choose a representation
\[
A_n
=
\sum_{a=1}^N c_a^{(n)}G_a,
\qquad
c_a^{(n)}\ge0,
\]
having the smallest possible number of strictly positive coefficients among
all conic representations of \(A_n\). We claim that the generators
corresponding to the strictly positive coefficients in such a representation
are linearly independent.

Suppose otherwise. Let
$
I_n
:=
\{a:c_a^{(n)}>0\},
$
and suppose that
\(\{G_a:a\in I_n\}\) is linearly dependent. Then there exist real numbers
\(\beta_a\), not all zero, such that
$
\sum_{a\in I_n}\beta_aG_a=0.
$
After multiplying all \(\beta_a\) by \(-1\) if necessary, we may assume that
\(\beta_a>0\) for at least one \(a\in I_n\). Define
$
t_n
:=
\min_{\substack{a\in I_n\\ \beta_a>0}}
\frac{c_a^{(n)}}{\beta_a}.
$
Every quantity entering this minimum is strictly positive, so
$
t_n>0.
$
Now define modified coefficients by
\[
\widetilde c_a^{(n)}
:=
c_a^{(n)}-t_n\beta_a
\qquad
(a\in I_n),
\]
and retain coefficient \(0\) outside \(I_n\). If \(\beta_a>0\), then by the
definition of \(t_n\),
\[
\widetilde c_a^{(n)}
=
c_a^{(n)}-t_n\beta_a
\ge0.
\]
If \(\beta_a\le0\), then
$
\widetilde c_a^{(n)}
=
c_a^{(n)}-t_n\beta_a
\ge
c_a^{(n)}
>0.
$
Moreover, for at least one index achieving the minimum in the definition of
\(t_n\),
$
\widetilde c_a^{(n)}=0.
$
On the other hand,
\[
\begin{aligned}
\sum_{a\in I_n}
\widetilde c_a^{(n)}G_a
&=
\sum_{a\in I_n}
\left(
c_a^{(n)}-t_n\beta_a
\right)G_a\\
&=
\sum_{a\in I_n}
c_a^{(n)}G_a
-
t_n
\sum_{a\in I_n}
\beta_aG_a\\
&=
A_n.
\end{aligned}
\]
Thus we have obtained another conic representation of \(A_n\) with strictly
fewer positive coefficients, contradicting the minimality of the original
representation. Hence the active generators in every chosen minimal
representation are linearly independent.

There are only finitely many subsets of
\(\{1,\ldots,N\}\). Therefore, after passing to a subsequence, which we do
without changing notation, we may assume that the active index set is a fixed
set
$
I\subseteq\{1,\ldots,N\}
$
for every \(n\). Hence
\[
A_n
=
\sum_{a\in I}c_a^{(n)}G_a,
\qquad
c_a^{(n)}>0,
\]
and the matrices
$
\{G_a:a\in I\}
$
are linearly independent.

Consider the linear map
$
T_I:\mathbb R^{|I|}
\longrightarrow
\mathbb R^{d\times d},
\qquad
T_I(c)
=
\sum_{a\in I}c_aG_a.
$
Linear independence of the generators implies that \(T_I\) is injective.
Therefore \(T_I\) is a linear isomorphism from
\(\mathbb R^{|I|}\) onto its image, and its inverse on that finite-dimensional
image is continuous. Consequently, there exists a constant \(C_I>0\) such
that
$
\|c\|_2
\le
C_I\|T_I(c)\|_F
\qquad
\text{for every }c\in\mathbb R^{|I|}.
$
Since
$
A_n\longrightarrow A,
$
the sequence \((A_n)\) is bounded in Frobenius norm. Hence
\[
\left\|
(c_a^{(n)})_{a\in I}
\right\|_2
\le
C_I\|A_n\|_F
\]
shows that the coefficient sequence is also bounded. By finite-dimensional
compactness, it has a convergent subsequence. Passing to this subsequence,
write
$
c_a^{(n)}
\longrightarrow
c_a
\qquad
\text{for every }a\in I.
$
Since every \(c_a^{(n)}\ge0\),
$
c_a\ge0.
$
Taking limits in
\[
A_n
=
\sum_{a\in I}c_a^{(n)}G_a
\]
gives
$
A
=
\sum_{a\in I}c_aG_a.
$
This is a nonnegative linear combination of generators in \(\mathcal G\), so
$
A\in\mathcal A_{\mathfrak S}.
$
Therefore \(\mathcal A_{\mathfrak S}\) is closed.

We now prove existence of a best approximation. Define
\[
\mu
:=
\inf_{A\in\mathcal A_{\mathfrak S}}
\|B-A\|_F^2.
\]
Because squared norms are nonnegative,
$
\mu\ge0.
$
Since
$
0\in\mathcal A_{\mathfrak S},
$
we also have
\[
\mu
\le
\|B-0\|_F^2
=
\|B\|_F^2.
\]
Thus \(\mu\) is finite. By the definition of an infimum, there exists a
sequence
$
A_n\in\mathcal A_{\mathfrak S}
$
such that
$
\|B-A_n\|_F^2
\longrightarrow
\mu.
$
In particular, for all sufficiently large \(n\),
\[
\|B-A_n\|_F^2
\le
\mu+1.
\]
Hence
\[
\|B-A_n\|_F
\le
\sqrt{\mu+1}.
\]
The triangle inequality gives
\[
\begin{aligned}
\|A_n\|_F
&=
\|A_n-B+B\|_F\\
&\le
\|A_n-B\|_F+\|B\|_F\\
&\le
\sqrt{\mu+1}+\|B\|_F.
\end{aligned}
\]
Therefore the minimizing sequence is bounded in
\(\mathbb R^{d\times d}\). Since this space is finite-dimensional,
Bolzano--Weierstrass yields a convergent subsequence$
A_{n_k}
\longrightarrow
A^\star
$
for some
\(A^\star\in\mathbb R^{d\times d}\). Because
\(\mathcal A_{\mathfrak S}\) is closed and every
\(A_{n_k}\in\mathcal A_{\mathfrak S}\),
$
A^\star\in\mathcal A_{\mathfrak S}.
$
The map
$
A\longmapsto\|B-A\|_F^2
$
is continuous, and hence
\[
\begin{aligned}
\|B-A^\star\|_F^2
&=
\lim_{k\to\infty}
\|B-A_{n_k}\|_F^2\\
&=
\mu.
\end{aligned}
\]
Thus the infimum is attained, so a minimizer exists.

It remains to prove uniqueness. Suppose, toward a contradiction, that there
are two distinct minimizers
$
A_1^\star,A_2^\star
\in
\mathcal A_{\mathfrak S},
\qquad
A_1^\star\neq A_2^\star.
$
Since both attain the minimum,
$
\|B-A_1^\star\|_F^2
=
\mu
\qquad\text{and}\qquad
\|B-A_2^\star\|_F^2
=
\mu.
$
Because \(\mathcal A_{\mathfrak S}\) is convex, their midpoint
$
A_{\mathrm{mid}}
:=
\frac{A_1^\star+A_2^\star}{2}
$
also belongs to \(\mathcal A_{\mathfrak S}\).

Set
$
X:=B-A_1^\star,
\qquad
Y:=B-A_2^\star.
$
Then
$
B-A_{\mathrm{mid}}
=
\frac{X+Y}{2}.
$
Using the Frobenius inner product,
$
\|M\|_F^2=\langle M,M\rangle_F,
$
we compute
\[
\begin{aligned}
\left\|
B-A_{\mathrm{mid}}
\right\|_F^2
&=
\left\|
\frac{X+Y}{2}
\right\|_F^2\\
&=
\frac14
\left(
\|X\|_F^2
+
2\langle X,Y\rangle_F
+
\|Y\|_F^2
\right).
\end{aligned}
\]
Also,
\[
\begin{aligned}
\|X-Y\|_F^2
&=
\|X\|_F^2
-
2\langle X,Y\rangle_F
+
\|Y\|_F^2,
\end{aligned}
\]
so
$
2\langle X,Y\rangle_F
=
\|X\|_F^2+\|Y\|_F^2-\|X-Y\|_F^2.
$
Substituting this identity gives
\[
\left\|
B-A_{\mathrm{mid}}
\right\|_F^2
=
\frac12\|X\|_F^2
+
\frac12\|Y\|_F^2
-
\frac14\|X-Y\|_F^2.
\]
Since
$
X-Y
=
(B-A_1^\star)-(B-A_2^\star)
=
A_2^\star-A_1^\star,
$
we obtain
\[
\begin{aligned}
\left\|
B-\frac{A_1^\star+A_2^\star}{2}
\right\|_F^2
&=
\frac12
\|B-A_1^\star\|_F^2
+
\frac12
\|B-A_2^\star\|_F^2\\
&\qquad
-
\frac14
\|A_1^\star-A_2^\star\|_F^2\\
&=
\mu
-
\frac14
\|A_1^\star-A_2^\star\|_F^2.
\end{aligned}
\]
Because
$
A_1^\star\neq A_2^\star,
$
we have
$
\|A_1^\star-A_2^\star\|_F^2>0.
$
Therefore
$
\left\|
B-\frac{A_1^\star+A_2^\star}{2}
\right\|_F^2
<
\mu.
$
But the midpoint belongs to
\(\mathcal A_{\mathfrak S}\), so this contradicts the definition of
\(\mu\) as the infimum of the objective over
\(\mathcal A_{\mathfrak S}\). Hence two distinct minimizers cannot exist.

We have therefore shown that, for every
\(B\in\mathbb R^{d\times d}\), there exists exactly one
\(A^\star\in\mathcal A_{\mathfrak S}\) satisfying
$
\|B-A^\star\|_F^2
=
\min_{A\in\mathcal A_{\mathfrak S}}
\|B-A\|_F^2.
$
Consequently,
$
\Pi_{\mathcal A_{\mathfrak S}}(B)
=
A^\star
$
exists and is unique.
\end{proof}

\subsection{Proof of Proposition~\ref{prop:learning-data-weighted-projection}}

\begin{proof}
Let
$
M:=B-A.
$
Since \(h\) has finite second moment, every entry of
$
\Sigma=\mathbb E[hh^\top]
$
is finite. Moreover,
$
\Sigma^\top
=
\mathbb E[(hh^\top)^\top]
=
\mathbb E[hh^\top]
=
\Sigma,
$
and for every \(x\in\mathbb R^d\),
$
x^\top\Sigma x
=
x^\top\mathbb E[hh^\top]x
=
\mathbb E\!\left[x^\top hh^\top x\right]
=
\mathbb E\!\left[(h^\top x)^2\right]
\ge0.
$
Thus \(\Sigma\) is symmetric positive semidefinite.

For the first identity, observe that
$
\|Mh\|_2^2
=
(Mh)^\top(Mh)
=
h^\top M^\top Mh.
$
The scalar on the right may be written as its own trace:
\[
h^\top M^\top Mh
=
\operatorname{tr}\!\left(h^\top M^\top Mh\right).
\]
Using cyclic invariance of the trace,
$
\operatorname{tr}\!\left(h^\top M^\top Mh\right)
=
\operatorname{tr}\!\left(Mhh^\top M^\top\right).
$
Therefore
$
\|Mh\|_2^2
=
\operatorname{tr}\!\left(Mhh^\top M^\top\right).
$
Taking expectations and using linearity of both expectation and trace gives
\[
\begin{aligned}
\mathbb E\|Mh\|_2^2
&=
\mathbb E\!\left[
\operatorname{tr}\!\left(Mhh^\top M^\top\right)
\right]\\
&=
\operatorname{tr}\!\left(
M\,\mathbb E[hh^\top]\,M^\top
\right)\\
&=
\operatorname{tr}\!\left(
M\Sigma M^\top
\right).
\end{aligned}
\]
Substituting \(M=B-A\) yields
$
\mathbb E\|(B-A)h\|_2^2
=
\operatorname{tr}\!\bigl(
(B-A)\Sigma(B-A)^\top
\bigr).
$

We next consider the case
$
\Sigma\succ0.
$
Let
$
\lambda_{\min}
:=
\lambda_{\min}(\Sigma)>0.
$
For an arbitrary matrix \(C\in\mathbb R^{d\times d}\), write its \(i\)-th
row as \(c_i^\top\), where \(c_i\in\mathbb R^d\). The \(i\)-th diagonal
entry of \(C\Sigma C^\top\) is
$
(C\Sigma C^\top)_{ii}
=
c_i^\top\Sigma c_i.
$
Hence
$
\operatorname{tr}(C\Sigma C^\top)
=
\sum_{i=1}^d c_i^\top\Sigma c_i.
$
Since \(\Sigma\succ0\),
$
c_i^\top\Sigma c_i
\ge
\lambda_{\min}\|c_i\|_2^2
$
for every \(i\). Therefore
\[
\begin{aligned}
\operatorname{tr}(C\Sigma C^\top)
&=
\sum_{i=1}^d c_i^\top\Sigma c_i\\
&\ge
\lambda_{\min}
\sum_{i=1}^d\|c_i\|_2^2\\
&=
\lambda_{\min}\|C\|_F^2.
\end{aligned}
\]
In particular,
$
\operatorname{tr}(C\Sigma C^\top)=0
\quad\Longrightarrow\quad
C=0.
$

Define
$
F_\Sigma(A)
:=
\operatorname{tr}\!\bigl(
(B-A)\Sigma(B-A)^\top
\bigr).
$
By Proposition~\ref{prop:operators-interaction-family-structure},
\(\mathcal A_{\mathfrak S}\) is a nonempty closed convex subset of
\(\mathbb R^{d\times d}\). We first prove that \(F_\Sigma\) attains its
minimum on this set. Let
$
\mu
:=
\inf_{A\in\mathcal A_{\mathfrak S}}F_\Sigma(A).
$
Because \(F_\Sigma(A)\ge0\),
$
\mu\ge0.
$
Also,
$
0\in\mathcal A_{\mathfrak S},
$
so
$
\mu
\le
F_\Sigma(0)
=
\operatorname{tr}(B\Sigma B^\top)
<
\infty.
$
Choose a minimizing sequence
$
A_n\in\mathcal A_{\mathfrak S}
$
such that
$
F_\Sigma(A_n)\longrightarrow\mu.
$
For all sufficiently large \(n\),
$
F_\Sigma(A_n)\le\mu+1.
$
Applying the lower bound above with
$
C=B-A_n
$
gives
$
\lambda_{\min}\|B-A_n\|_F^2
\le
F_\Sigma(A_n)
\le
\mu+1.
$
Consequently,
\[
\|B-A_n\|_F
\le
\sqrt{\frac{\mu+1}{\lambda_{\min}}}.
\]
The triangle inequality then yields
\[
\begin{aligned}
\|A_n\|_F
&\le
\|A_n-B\|_F+\|B\|_F\\
&\le
\sqrt{\frac{\mu+1}{\lambda_{\min}}}
+
\|B\|_F.
\end{aligned}
\]
Thus the minimizing sequence is bounded. Since
\(\mathbb R^{d\times d}\) is finite-dimensional, there exists a convergent
subsequence, which we denote by
$
A_{n_j}\longrightarrow A^\star.
$
Closedness of \(\mathcal A_{\mathfrak S}\) gives
$
A^\star\in\mathcal A_{\mathfrak S}.
$
Because \(F_\Sigma\) is a polynomial, and hence continuous, in the entries of
\(A\),
\[
\begin{aligned}
F_\Sigma(A^\star)
&=
\lim_{j\to\infty}F_\Sigma(A_{n_j})\\
&=
\mu.
\end{aligned}
\]
Therefore a minimizing matrix exists.

We now prove that this minimizing matrix is unique. Let
$
A_1,A_2\in\mathbb R^{d\times d},
\qquad
A_1\neq A_2,
$
and let \(t\in(0,1)\). Set
$
X:=B-A_1,
\qquad
Y:=B-A_2.
$
Then
$
B-\bigl(tA_1+(1-t)A_2\bigr)
=
tX+(1-t)Y.
$
Expanding the corresponding weighted quadratic form gives
\[
\begin{aligned}
&
\operatorname{tr}\!\left(
[tX+(1-t)Y]\Sigma[tX+(1-t)Y]^\top
\right)\\
&\quad=
t^2\operatorname{tr}(X\Sigma X^\top)
+
(1-t)^2\operatorname{tr}(Y\Sigma Y^\top)\\
&\qquad
+
t(1-t)\operatorname{tr}(X\Sigma Y^\top)
+
t(1-t)\operatorname{tr}(Y\Sigma X^\top).
\end{aligned}
\]
Since \(\Sigma^\top=\Sigma\),
\[
\begin{aligned}
\operatorname{tr}(Y\Sigma X^\top)
&=
\operatorname{tr}\!\left(
(Y\Sigma X^\top)^\top
\right)\\
&=
\operatorname{tr}(X\Sigma Y^\top).
\end{aligned}
\]
Therefore
\[
\begin{aligned}
&
F_\Sigma\bigl(tA_1+(1-t)A_2\bigr)\\
&\quad=
t^2\operatorname{tr}(X\Sigma X^\top)
+
(1-t)^2\operatorname{tr}(Y\Sigma Y^\top)
+
2t(1-t)\operatorname{tr}(X\Sigma Y^\top).
\end{aligned}
\]
On the other hand,
\[
\begin{aligned}
&(X-Y)\Sigma(X-Y)^\top\\
&\quad=
X\Sigma X^\top
-
X\Sigma Y^\top
-
Y\Sigma X^\top
+
Y\Sigma Y^\top,
\end{aligned}
\]
and hence, after taking traces,
\[
\begin{aligned}
\operatorname{tr}\!\left(
(X-Y)\Sigma(X-Y)^\top
\right)
&=
\operatorname{tr}(X\Sigma X^\top)
+
\operatorname{tr}(Y\Sigma Y^\top)\\
&\qquad
-
2\operatorname{tr}(X\Sigma Y^\top).
\end{aligned}
\]
Rearranging gives
$
2\operatorname{tr}(X\Sigma Y^\top)
=
\operatorname{tr}(X\Sigma X^\top)
+
\operatorname{tr}(Y\Sigma Y^\top)
-
\operatorname{tr}\!\left(
(X-Y)\Sigma(X-Y)^\top
\right).
$
Substituting this expression into the previous expansion yields
\[
\begin{aligned}
F_\Sigma\bigl(tA_1+(1-t)A_2\bigr)
&=
tF_\Sigma(A_1)
+
(1-t)F_\Sigma(A_2)\\
&\qquad
-
t(1-t)
\operatorname{tr}\!\left(
(X-Y)\Sigma(X-Y)^\top
\right).
\end{aligned}
\]
Since
$
X-Y
=
(B-A_1)-(B-A_2)
=
A_2-A_1,
$
we obtain
\[
\begin{aligned}
F_\Sigma\bigl(tA_1+(1-t)A_2\bigr)
&=
tF_\Sigma(A_1)
+
(1-t)F_\Sigma(A_2)\\
&\qquad
-
t(1-t)
\operatorname{tr}\!\left(
(A_1-A_2)\Sigma(A_1-A_2)^\top
\right).
\end{aligned}
\]
Because \(A_1-A_2\neq0\) and \(\Sigma\succ0\), the positive-definiteness
estimate proved above gives
\[
\operatorname{tr}\!\left(
(A_1-A_2)\Sigma(A_1-A_2)^\top
\right)
\ge
\lambda_{\min}
\|A_1-A_2\|_F^2
>
0.
\]
Since \(t(1-t)>0\),
$
F_\Sigma\bigl(tA_1+(1-t)A_2\bigr)
<
tF_\Sigma(A_1)
+
(1-t)F_\Sigma(A_2).
$
Thus \(F_\Sigma\) is strictly convex as a function of the matrix \(A\).

Suppose now that two distinct matrices
$
A_1^\star,A_2^\star
\in
\mathcal A_{\mathfrak S}
$
both minimize \(F_\Sigma\), with common minimum value \(\mu\). Convexity of
\(\mathcal A_{\mathfrak S}\) implies that
$
\frac{A_1^\star+A_2^\star}{2}
\in
\mathcal A_{\mathfrak S}.
$
Strict convexity gives
\[
\begin{aligned}
F_\Sigma\!\left(
\frac{A_1^\star+A_2^\star}{2}
\right)
&<
\frac12F_\Sigma(A_1^\star)
+
\frac12F_\Sigma(A_2^\star)\\
&=
\mu,
\end{aligned}
\]
contradicting minimality. Hence, when
\(\Sigma\succ0\), the minimizing matrix in
\(\mathcal A_{\mathfrak S}\) is unique.

If \(\Sigma\) is singular, the preceding strict-convexity argument can fail,
because
$
\operatorname{tr}(C\Sigma C^\top)
$
may vanish for a nonzero matrix \(C\). Thus uniqueness cannot be guaranteed
in general. A concrete example is obtained by taking
$
h=0
\qquad
\text{almost surely}.
$
Then
$
\Sigma
=
\mathbb E[hh^\top]
=
0,
$
which is singular, and for every
\(A\in\mathcal A_{\mathfrak S}\),
$
\mathbb E\|(B-A)h\|_2^2
=
0.
$
In particular, both
$
0\in\mathcal A_{\mathfrak S}
$
and
$
E_{11}\in\mathcal A_{\mathfrak S},
$
the latter being obtained by setting all channel coefficients to zero and
choosing the free diagonal coefficient vector
\(\delta=e_1\). Since
$
0\neq E_{11},
$
the minimizing matrix is not unique in this singular example. This proves
that singularity of \(\Sigma\) may destroy uniqueness, although it does not
assert that nonuniqueness must occur for every singular second-moment
matrix.

It remains to derive the coefficient form. As fixed in the main text, let
$
k=R+d
$
and write
\[
A(\alpha)
=
\sum_{a=1}^k\alpha_aQ_a,
\]
where
$
Q_r=\Phi_r
\quad
(1\le r\le R),
\qquad
Q_{R+i}=E_{ii}
\quad
(1\le i\le d).
$
Expanding the data-weighted objective gives
\[
\begin{aligned}
F_\Sigma(A)
&=
\operatorname{tr}\!\left(
(B-A)\Sigma(B-A)^\top
\right)\\
&=
\operatorname{tr}(B\Sigma B^\top)
-
\operatorname{tr}(B\Sigma A^\top)
-
\operatorname{tr}(A\Sigma B^\top)
+
\operatorname{tr}(A\Sigma A^\top).
\end{aligned}
\]
Because \(\Sigma^\top=\Sigma\),
\[
\begin{aligned}
\operatorname{tr}(A\Sigma B^\top)
&=
\operatorname{tr}\!\left(
(A\Sigma B^\top)^\top
\right)\\
&=
\operatorname{tr}(B\Sigma A^\top).
\end{aligned}
\]
Hence
$
F_\Sigma(A)
=
\operatorname{tr}(B\Sigma B^\top)
-
2\operatorname{tr}(B\Sigma A^\top)
+
\operatorname{tr}(A\Sigma A^\top).
$
Substituting
$
A=\sum_{a=1}^k\alpha_aQ_a
$
into the linear term gives
\[
\begin{aligned}
\operatorname{tr}(B\Sigma A^\top)
&=
\operatorname{tr}\!\left(
B\Sigma
\left(
\sum_{a=1}^k\alpha_aQ_a
\right)^\top
\right)\\
&=
\operatorname{tr}\!\left(
B\Sigma
\sum_{a=1}^k\alpha_aQ_a^\top
\right)\\
&=
\sum_{a=1}^k
\alpha_a
\operatorname{tr}(B\Sigma Q_a^\top).
\end{aligned}
\]
Therefore, defining
$
(b_\Sigma)_a
:=
\operatorname{tr}(B\Sigma Q_a^\top),
$
we have
$
\operatorname{tr}(B\Sigma A^\top)
=
b_\Sigma^\top\alpha.
$

For the quadratic term,
\[
\begin{aligned}
\operatorname{tr}(A\Sigma A^\top)
&=
\operatorname{tr}\!\left[
\left(
\sum_{a=1}^k\alpha_aQ_a
\right)
\Sigma
\left(
\sum_{b=1}^k\alpha_bQ_b
\right)^\top
\right]\\
&=
\operatorname{tr}\!\left[
\left(
\sum_{a=1}^k\alpha_aQ_a
\right)
\Sigma
\left(
\sum_{b=1}^k\alpha_bQ_b^\top
\right)
\right]\\
&=
\sum_{a=1}^k
\sum_{b=1}^k
\alpha_a\alpha_b
\operatorname{tr}(Q_a\Sigma Q_b^\top).
\end{aligned}
\]
Thus, defining
$
(\Gamma_\Sigma)_{ab}
:=
\operatorname{tr}(Q_a\Sigma Q_b^\top),
$
we obtain
$
\operatorname{tr}(A\Sigma A^\top)
=
\alpha^\top\Gamma_\Sigma\alpha.
$
Consequently,
\[
F_\Sigma(A(\alpha))
=
\operatorname{tr}(B\Sigma B^\top)
-
2b_\Sigma^\top\alpha
+
\alpha^\top\Gamma_\Sigma\alpha.
\]
The first term is independent of \(\alpha\), so minimizing the original
objective is equivalently represented, up to multiplication by the positive
constant \(1/2\) and addition of an irrelevant constant, by the quadratic
objective
$
\frac12\alpha^\top\Gamma_\Sigma\alpha
-
b_\Sigma^\top\alpha
$
over the prescribed coefficient-feasible set. In particular, the quadratic
and linear coefficients are exactly
$
(\Gamma_\Sigma)_{ab}
=
\operatorname{tr}(Q_a\Sigma Q_b^\top),
\qquad
(b_\Sigma)_a
=
\operatorname{tr}(B\Sigma Q_a^\top).
$
The uniqueness established above concerns the realized minimizing matrix
\(A^\star\). If the matrices \(Q_a\) are linearly dependent, distinct
coefficient vectors may still represent that same unique matrix.
\end{proof}

\subsection{Proof of Proposition~\ref{prop:learning-probe-identifiability}}

\begin{proof}
Because the model is unconstrained, the coefficient vector ranges over all of
\(\mathbb R^k\). For any coefficient vector
\(c=(c_1,\ldots,c_k)^\top\in\mathbb R^k\), define
$
A(c)
:=
\sum_{a=1}^k c_aQ_a.
$
We first record the precise relation between the probe Gram matrix
\(\Gamma\), the empirical second-moment matrix
\(\widehat\Sigma\), and the action of the synthesized operator
\(A(c)\) on the observed probes.

For every pair \(a,b\),
\[
\begin{aligned}
\frac1N\sum_{n=1}^N
\langle Q_ah_n,Q_bh_n\rangle
&=
\frac1N\sum_{n=1}^N
(Q_ah_n)^\top(Q_bh_n)\\
&=
\frac1N\sum_{n=1}^N
h_n^\top Q_a^\top Q_bh_n.
\end{aligned}
\]
Each scalar on the right may be written as a trace. Using cyclic invariance
of the trace,
\[
\begin{aligned}
h_n^\top Q_a^\top Q_bh_n
&=
\operatorname{tr}
\left(
h_n^\top Q_a^\top Q_bh_n
\right)\\
&=
\operatorname{tr}
\left(
Q_bh_nh_n^\top Q_a^\top
\right).
\end{aligned}
\]
Equivalently, since
$
\operatorname{tr}
\left(
Q_bh_nh_n^\top Q_a^\top
\right)
=
\operatorname{tr}
\left(
Q_ah_nh_n^\top Q_b^\top
\right),
$
the symmetry of the scalar inner product gives
\[
\begin{aligned}
\Gamma_{ab}
&=
\frac1N\sum_{n=1}^N
\langle Q_ah_n,Q_bh_n\rangle\\
&=
\operatorname{tr}
\left(
Q_a
\left(
\frac1N\sum_{n=1}^Nh_nh_n^\top
\right)
Q_b^\top
\right)\\
&=
\operatorname{tr}
\left(
Q_a\widehat\Sigma Q_b^\top
\right).
\end{aligned}
\]
Thus the two expressions for \(\Gamma_{ab}\) in the statement agree.

Now let \(c\in\mathbb R^k\) be arbitrary. Expanding its quadratic form with
respect to \(\Gamma\) gives
\[
\begin{aligned}
c^\top\Gamma c
&=
\sum_{a=1}^k\sum_{b=1}^k
c_ac_b\Gamma_{ab}\\
&=
\frac1N
\sum_{n=1}^N
\sum_{a=1}^k\sum_{b=1}^k
c_ac_b
\langle Q_ah_n,Q_bh_n\rangle.
\end{aligned}
\]
By bilinearity of the Euclidean inner product,
\[
\begin{aligned}
c^\top\Gamma c
&=
\frac1N
\sum_{n=1}^N
\left\langle
\sum_{a=1}^kc_aQ_ah_n,\,
\sum_{b=1}^kc_bQ_bh_n
\right\rangle\\
&=
\frac1N
\sum_{n=1}^N
\left\|
\sum_{a=1}^kc_aQ_ah_n
\right\|_2^2\\
&=
\frac1N
\sum_{n=1}^N
\|A(c)h_n\|_2^2.
\end{aligned}
\]
Every term in the last sum is nonnegative. Therefore
$
c^\top\Gamma c\ge0
\qquad
\text{for every }c\in\mathbb R^k,
$
so
$
\Gamma\succeq0.
$
Moreover, because a finite sum of nonnegative numbers is zero if and only if
each summand is zero,
$
c^\top\Gamma c=0
\quad\Longleftrightarrow\quad
A(c)h_n=0
\quad
\text{for every }n=1,\ldots,N.
$
Since \(\Gamma\) is symmetric positive semidefinite, this is equivalently
$
c\in\ker(\Gamma)
\quad\Longleftrightarrow\quad
A(c)h_n=0
\quad
\text{for every }n.
$

We next expand the least-squares objective. For every probe \(h_n\),
$
A(\alpha)h_n
=
\sum_{a=1}^k\alpha_aQ_ah_n.
$
Hence
\[
\begin{aligned}
\|A(\alpha)h_n-y_n\|_2^2
&=
\langle
A(\alpha)h_n-y_n,\,
A(\alpha)h_n-y_n
\rangle\\
&=
\|A(\alpha)h_n\|_2^2
-
2\langle A(\alpha)h_n,y_n\rangle
+
\|y_n\|_2^2.
\end{aligned}
\]
Averaging over \(n\) yields
\[
\begin{aligned}
J(\alpha)
&=
\frac1N\sum_{n=1}^N
\|A(\alpha)h_n\|_2^2\\
&\qquad
-
\frac{2}{N}\sum_{n=1}^N
\langle A(\alpha)h_n,y_n\rangle
+
\frac1N\sum_{n=1}^N\|y_n\|_2^2.
\end{aligned}
\]
The first term is exactly
$
\alpha^\top\Gamma\alpha.
$
Define
$
g\in\mathbb R^k,
\qquad
g_a
:=
\frac1N
\sum_{n=1}^N
\langle Q_ah_n,y_n\rangle,
$
and define the constant
$
C_y
:=
\frac1N\sum_{n=1}^N\|y_n\|_2^2.
$
Then
\[
\begin{aligned}
\frac1N\sum_{n=1}^N
\langle A(\alpha)h_n,y_n\rangle
&=
\frac1N\sum_{n=1}^N
\left\langle
\sum_{a=1}^k\alpha_aQ_ah_n,
y_n
\right\rangle\\
&=
\sum_{a=1}^k
\alpha_a
\left(
\frac1N
\sum_{n=1}^N
\langle Q_ah_n,y_n\rangle
\right)\\
&=
g^\top\alpha.
\end{aligned}
\]
Consequently,
$
J(\alpha)
=
\alpha^\top\Gamma\alpha
-
2g^\top\alpha
+
C_y.
$
Since \(\Gamma\) is symmetric,
$
\nabla J(\alpha)
=
2\Gamma\alpha-2g.
$
Thus every stationary point satisfies the normal equations
$
\Gamma\alpha=g.
$

We now verify that these normal equations always have at least one solution.
Let
$
z\in\ker(\Gamma).
$
From the Gram identity proved above,
$
A(z)h_n=0
\qquad
\text{for every }n.
$
Therefore
\[
\begin{aligned}
z^\top g
&=
\sum_{a=1}^kz_ag_a\\
&=
\frac1N
\sum_{n=1}^N
\sum_{a=1}^k
z_a
\langle Q_ah_n,y_n\rangle\\
&=
\frac1N
\sum_{n=1}^N
\left\langle
\sum_{a=1}^kz_aQ_ah_n,\,
y_n
\right\rangle\\
&=
\frac1N
\sum_{n=1}^N
\langle A(z)h_n,y_n\rangle\\
&=
0.
\end{aligned}
\]
Hence
$
g\perp\ker(\Gamma).
$
Because \(\Gamma\) is a real symmetric matrix,
$
\operatorname{range}(\Gamma)
=
\ker(\Gamma)^\perp.
$
Indeed,
\(\operatorname{range}(\Gamma)\subseteq\ker(\Gamma)^\perp\) follows from
symmetry, and both spaces have dimension
$
k-\dim\ker(\Gamma).
$
Therefore
$
g\in\operatorname{range}(\Gamma),
$
so there exists at least one vector
\(\alpha^\star\in\mathbb R^k\) satisfying
$
\Gamma\alpha^\star=g.
$

For any \(u\in\mathbb R^k\), substitute
\(\alpha=\alpha^\star+u\) into the objective:
\[
\begin{aligned}
J(\alpha^\star+u)
&=
(\alpha^\star+u)^\top
\Gamma
(\alpha^\star+u)
-
2g^\top(\alpha^\star+u)
+
C_y\\
&=
(\alpha^\star)^\top\Gamma\alpha^\star
+
2u^\top\Gamma\alpha^\star
+
u^\top\Gamma u\\
&\qquad
-
2g^\top\alpha^\star
-
2g^\top u
+
C_y.
\end{aligned}
\]
Since
$
\Gamma\alpha^\star=g,
$
the cross terms cancel:
$
2u^\top\Gamma\alpha^\star-2g^\top u
=
2u^\top g-2g^\top u
=
0.
$
Therefore
$
J(\alpha^\star+u)
=
J(\alpha^\star)
+
u^\top\Gamma u.
$
Since
$
\Gamma\succeq0,
$
we have
$
u^\top\Gamma u\ge0,
$
and thus
$
J(\alpha^\star+u)
\ge
J(\alpha^\star)
\qquad
\text{for every }u\in\mathbb R^k.
$
Hence every solution of the normal equations is a global least-squares
minimizer.

Suppose first that
$
\Gamma\succ0.
$
Then
$
u^\top\Gamma u>0
\qquad
\text{for every }u\neq0.
$
Consequently,
$
J(\alpha^\star+u)
>
J(\alpha^\star)
\qquad
\text{whenever }u\neq0.
$
Thus no coefficient vector other than \(\alpha^\star\) can attain the same
least-squares value. Equivalently, the least-squares coefficient vector is
unique. One may also see this directly from the normal equations: positive
definiteness implies that \(\Gamma\) is invertible, and therefore
$
\alpha^\star
=
\Gamma^{-1}g
$
is the unique solution.

Conversely, suppose that \(\Gamma\) is not positive definite. Since
\(\Gamma\succeq0\), failure of positive definiteness means that
\(\Gamma\) is singular. Hence there exists
$
z\in\mathbb R^k,
\qquad
z\neq0,
$
such that
$
\Gamma z=0.
$
Let \(\alpha^\star\) be any solution of
$
\Gamma\alpha^\star=g,
$
whose existence was established above. For every \(t\in\mathbb R\),
$
\Gamma(\alpha^\star+t z)
=
\Gamma\alpha^\star+t\Gamma z
=
g.
$
Thus every vector
$
\alpha^\star+t z
$
is also a solution of the normal equations and hence a global minimizer.
Because \(z\neq0\), choosing, for example, \(t=0\) and \(t=1\) produces two
distinct minimizing coefficient vectors. Therefore the least-squares
coefficient vector is not unique. We have proved the equivalence
\[
\text{the least-squares coefficient vector is unique}
\quad\Longleftrightarrow\quad
\Gamma\succ0.
\]

The same conclusion has a direct observational interpretation. If
\(z\in\ker(\Gamma)\), then
\[
0
=
z^\top\Gamma z
=
\frac1N\sum_{n=1}^N
\|A(z)h_n\|_2^2,
\]
and hence
$
A(z)h_n=0
\qquad
\text{for every }n.
$
Therefore, for every coefficient vector \(\alpha\) and every scalar \(t\),
\[
\begin{aligned}
A(\alpha+t z)h_n
&=
A(\alpha)h_n+tA(z)h_n\\
&=
A(\alpha)h_n.
\end{aligned}
\]
Thus a nonzero vector in \(\ker(\Gamma)\) is exactly a coefficient direction
that is invisible to all of the available probes.

We finally prove the stated sufficient condition. Assume that
$
Q_1,\ldots,Q_k
$
are linearly independent and that
$
\widehat\Sigma\succ0.
$
Let
$
c\in\mathbb R^k,
\qquad
c\neq0.
$
By linear independence of the matrices \(Q_a\),
$
A(c)
=
\sum_{a=1}^kc_aQ_a
\neq0.
$
Using the trace representation of \(\Gamma\),
\[
\begin{aligned}
c^\top\Gamma c
&=
\sum_{a=1}^k\sum_{b=1}^k
c_ac_b
\operatorname{tr}
\left(
Q_a\widehat\Sigma Q_b^\top
\right)\\
&=
\operatorname{tr}
\left[
\left(
\sum_{a=1}^kc_aQ_a
\right)
\widehat\Sigma
\left(
\sum_{b=1}^kc_bQ_b
\right)^\top
\right]\\
&=
\operatorname{tr}
\left(
A(c)\widehat\Sigma A(c)^\top
\right).
\end{aligned}
\]
Let
$
\lambda_{\min}
:=
\lambda_{\min}(\widehat\Sigma).
$
Since
\(\widehat\Sigma\succ0\),
$
\lambda_{\min}>0.
$
Writing the rows of \(A(c)\) as
\(r_1^\top,\ldots,r_d^\top\), we obtain
\[
\begin{aligned}
\operatorname{tr}
\left(
A(c)\widehat\Sigma A(c)^\top
\right)
&=
\sum_{i=1}^d
r_i^\top\widehat\Sigma r_i\\
&\ge
\lambda_{\min}
\sum_{i=1}^d
\|r_i\|_2^2\\
&=
\lambda_{\min}
\|A(c)\|_F^2.
\end{aligned}
\]
Since \(A(c)\neq0\),
$
\|A(c)\|_F^2>0,
$
and therefore
\[
c^\top\Gamma c
\ge
\lambda_{\min}
\|A(c)\|_F^2
>
0.
\]
This holds for every nonzero
\(c\in\mathbb R^k\), so
$
\Gamma\succ0.
$
By the equivalence already proved, the least-squares coefficient vector is
therefore unique.

Matrix linear independence by itself does not imply this observational
condition. For example, take \(d=2\),
$
Q_1=E_{11},
\qquad
Q_2=E_{12},
$
which are linearly independent matrices, and suppose that every probe is
$
h_n=e_1.
$
Then
$
Q_1h_n=e_1,
\qquad
Q_2h_n=0
$
for every \(n\). Consequently,
$
\Gamma
=
\begin{pmatrix}
1&0\\
0&0
\end{pmatrix},
$
which is singular. The coefficient multiplying \(Q_2\) is completely
invisible to these probes, even though \(Q_1\) and \(Q_2\) are linearly
independent as matrices. Hence matrix-basis independence alone does not
guarantee coefficient identifiability from operator actions.
\end{proof}

\subsection{Proof of Proposition~\ref{prop:learning-spectral-alignment}}

\begin{proof}
Since \(L=L^\top\), the assumed eigenvectors
$
u_1,\ldots,u_d
$
form an orthonormal basis of \(\mathbb R^d\), and for every
\(k\in\{1,\ldots,d\}\),
$
Lu_k=-\nu_k u_k,
\qquad
\nu_k\ge0.
$
We first determine the action of the propagation operator on each spectral
mode. By the power-series definition of the matrix exponential,
$
e^{\tau L}
=
\sum_{m=0}^{\infty}
\frac{\tau^mL^m}{m!}.
$
From
$
Lu_k=-\nu_k u_k
$
it follows recursively that
$
L^m u_k
=
(-\nu_k)^m u_k
\qquad
\text{for every }m\ge0.
$
Therefore
\[
\begin{aligned}
e^{\tau L}u_k
&=
\sum_{m=0}^{\infty}
\frac{\tau^m}{m!}L^m u_k\\
&=
\sum_{m=0}^{\infty}
\frac{\tau^m(-\nu_k)^m}{m!}u_k\\
&=
\left(
\sum_{m=0}^{\infty}
\frac{(-\tau\nu_k)^m}{m!}
\right)u_k\\
&=
e^{-\tau\nu_k}u_k.
\end{aligned}
\]

Expand the deterministic signal \(s\) in this orthonormal eigenbasis:
$
s
=
\sum_{k=1}^d s_k u_k,
\qquad
s_k=u_k^\top s.
$
Likewise, define the random spectral coefficients of the noise by
$
\varepsilon_k
:=
u_k^\top\varepsilon.
$
Because \(\{u_k\}_{k=1}^d\) is an orthonormal basis,
$
\varepsilon
=
\sum_{k=1}^d\varepsilon_k u_k,
$
and hence
$
h
=
s+\varepsilon
=
\sum_{k=1}^d
(s_k+\varepsilon_k)u_k.
$

The assumptions on the noise imply precise moment identities for the
spectral coefficients. Since
$
\mathbb E[\varepsilon]=0,
$
we have
$
\mathbb E[\varepsilon_k]
=
\mathbb E[u_k^\top\varepsilon]
=
u_k^\top\mathbb E[\varepsilon]
=
0.
$
Furthermore, for any \(k,\ell\),
\[
\begin{aligned}
\mathbb E[\varepsilon_k\varepsilon_\ell]
&=
\mathbb E[
(u_k^\top\varepsilon)
(u_\ell^\top\varepsilon)
]\\
&=
\mathbb E[
u_k^\top
\varepsilon\varepsilon^\top
u_\ell
]\\
&=
u_k^\top
\mathbb E[
\varepsilon\varepsilon^\top
]
u_\ell\\
&=
u_k^\top
(\sigma^2I)
u_\ell\\
&=
\sigma^2u_k^\top u_\ell.
\end{aligned}
\]
By orthonormality,
$
u_k^\top u_\ell
=
\begin{cases}
1, & k=\ell,\\
0, & k\neq\ell,
\end{cases}
$
and therefore
$
\mathbb E[\varepsilon_k\varepsilon_\ell]
=
\sigma^2\delta_{k\ell}.
$
In particular,
$
\mathbb E[\varepsilon_k^2]
=
\sigma^2
\qquad
\text{for every }k.
$

Applying \(e^{\tau L}\) to \(h\) and using its diagonal action in the
eigenbasis gives
\[
\begin{aligned}
e^{\tau L}h
&=
e^{\tau L}
\sum_{k=1}^d
(s_k+\varepsilon_k)u_k\\
&=
\sum_{k=1}^d
(s_k+\varepsilon_k)e^{\tau L}u_k\\
&=
\sum_{k=1}^d
e^{-\tau\nu_k}
(s_k+\varepsilon_k)u_k.
\end{aligned}
\]
Subtracting
$
s=\sum_{k=1}^d s_k u_k
$
yields
\[
\begin{aligned}
e^{\tau L}h-s
&=
\sum_{k=1}^d
\left[
e^{-\tau\nu_k}(s_k+\varepsilon_k)-s_k
\right]u_k\\
&=
\sum_{k=1}^d
\left[
(e^{-\tau\nu_k}-1)s_k
+
e^{-\tau\nu_k}\varepsilon_k
\right]u_k.
\end{aligned}
\]
Equivalently,
\[
e^{\tau L}h-s
=
\sum_{k=1}^d
\left[
-(1-e^{-\tau\nu_k})s_k
+
e^{-\tau\nu_k}\varepsilon_k
\right]u_k.
\]

Since the eigenvectors are orthonormal, Parseval's identity gives
\[
\begin{aligned}
\|e^{\tau L}h-s\|_2^2
&=
\sum_{k=1}^d
\left[
(e^{-\tau\nu_k}-1)s_k
+
e^{-\tau\nu_k}\varepsilon_k
\right]^2.
\end{aligned}
\]
Expanding the square in the \(k\)-th summand,
\[
\begin{aligned}
&
\left[
(e^{-\tau\nu_k}-1)s_k
+
e^{-\tau\nu_k}\varepsilon_k
\right]^2\\
&\quad=
(e^{-\tau\nu_k}-1)^2s_k^2
+
2(e^{-\tau\nu_k}-1)
e^{-\tau\nu_k}
s_k\varepsilon_k
+
e^{-2\tau\nu_k}\varepsilon_k^2.
\end{aligned}
\]
Taking expectations and recalling that \(s_k\), \(\nu_k\), and \(\tau\) are
deterministic gives
\[
\begin{aligned}
&
\mathbb E
\left[
(e^{-\tau\nu_k}-1)s_k
+
e^{-\tau\nu_k}\varepsilon_k
\right]^2\\
&\quad=
(e^{-\tau\nu_k}-1)^2s_k^2\\
&\qquad
+
2(e^{-\tau\nu_k}-1)
e^{-\tau\nu_k}
s_k
\mathbb E[\varepsilon_k]\\
&\qquad
+
e^{-2\tau\nu_k}
\mathbb E[\varepsilon_k^2].
\end{aligned}
\]
Using
$
\mathbb E[\varepsilon_k]=0
$
and
$
\mathbb E[\varepsilon_k^2]=\sigma^2,
$
the mixed signal--noise term vanishes and we obtain
\[
\begin{aligned}
&
\mathbb E
\left[
(e^{-\tau\nu_k}-1)s_k
+
e^{-\tau\nu_k}\varepsilon_k
\right]^2\\
&\quad=
(e^{-\tau\nu_k}-1)^2s_k^2
+
\sigma^2e^{-2\tau\nu_k}.
\end{aligned}
\]
Since
$
(e^{-\tau\nu_k}-1)^2
=
(1-e^{-\tau\nu_k})^2,
$
summing over all spectral modes gives
\[
\begin{aligned}
\mathbb E\|e^{\tau L}h-s\|_2^2
&=
\sum_{k=1}^d
(1-e^{-\tau\nu_k})^2s_k^2
+
\sigma^2
\sum_{k=1}^d
e^{-2\tau\nu_k}.
\end{aligned}
\]
This proves the stated mean-squared-error identity.

It remains to compare this risk with that of the unprocessed observation.
Since
$
h=s+\varepsilon,
$
we have
$
h-s=\varepsilon.
$
Therefore
$
\mathbb E\|h-s\|_2^2
=
\mathbb E\|\varepsilon\|_2^2.
$
Using
$
\|\varepsilon\|_2^2
=
\operatorname{tr}
(\varepsilon\varepsilon^\top),
$
linearity of expectation gives
\[
\begin{aligned}
\mathbb E\|\varepsilon\|_2^2
&=
\operatorname{tr}
\left(
\mathbb E[
\varepsilon\varepsilon^\top
]
\right)\\
&=
\operatorname{tr}(\sigma^2I)\\
&=
d\sigma^2.
\end{aligned}
\]
Equivalently, since there are \(d\) modes,
$
\mathbb E\|h-s\|_2^2
=
\sigma^2
\sum_{k=1}^d1.
$

Propagation has strictly lower mean-squared error than the unprocessed
observation precisely when
$
\mathbb E\|e^{\tau L}h-s\|_2^2
<
\mathbb E\|h-s\|_2^2.
$
Substituting the two expressions just derived, this is equivalent to
$
\sum_{k=1}^d
(1-e^{-\tau\nu_k})^2s_k^2
+
\sigma^2
\sum_{k=1}^d
e^{-2\tau\nu_k}
<
\sigma^2
\sum_{k=1}^d1.
$
Subtracting
$
\sigma^2
\sum_{k=1}^d
e^{-2\tau\nu_k}
$
from both sides gives
$
\sum_{k=1}^d
(1-e^{-\tau\nu_k})^2s_k^2
<
\sigma^2
\sum_{k=1}^d
\left(
1-e^{-2\tau\nu_k}
\right).
$
Thus propagation improves the mean-squared error if and only if the spectral
bias introduced by attenuating the signal is strictly smaller than the
amount of noise variance removed by the same propagation. This is exactly
the claimed task-alignment criterion.
\end{proof}

\subsection{Proof of Proposition~\ref{prop:prelim-order-specialization}}

\begin{proof}
Recall that the relation \(\preceq_{\mathcal E}\) is defined by
$
z_j\preceq_{\mathcal E}z_i
\quad\Longleftrightarrow\quad
z_j=z_i
\ \text{or}\
z_j\rightsquigarrow z_i,
$
where
$
z_j\rightsquigarrow z_i
$
means that there exists a finite directed path from \(z_j\) to \(z_i\)
whose every edge belongs to \(\mathcal E\).

Assume first that the directed graph \((X,\mathcal E)\) is acyclic. We
verify the three defining properties of a partial order.

For every \(z_i\in X\), one has
$
z_i=z_i.
$
Therefore, directly from the definition of \(\preceq_{\mathcal E}\),
$
z_i\preceq_{\mathcal E}z_i.
$
Hence \(\preceq_{\mathcal E}\) is reflexive.

We next prove transitivity. Let \(z_a,z_b,z_c\in X\) satisfy
$
z_a\preceq_{\mathcal E}z_b
\qquad\text{and}\qquad
z_b\preceq_{\mathcal E}z_c.
$
By definition, the first relation means that either
$
z_a=z_b
$
or there exists a directed path
\[
z_a
=
z_{p_0}
\to
z_{p_1}
\to
\cdots
\to
z_{p_r}
=
z_b
\]
whose every edge belongs to \(\mathcal E\). Likewise, the second relation
means that either
$
z_b=z_c
$
or there exists a directed path
\[
z_b
=
z_{q_0}
\to
z_{q_1}
\to
\cdots
\to
z_{q_s}
=
z_c
\]
whose every edge belongs to \(\mathcal E\).

If
$
z_a=z_b
\qquad\text{and}\qquad
z_b=z_c,
$
then
$
z_a=z_c,
$
and hence
$
z_a\preceq_{\mathcal E}z_c.
$
If
$
z_a=z_b
$
while
$
z_b\rightsquigarrow z_c,
$
then replacing the initial vertex \(z_b\) of the latter path by the equal
vertex \(z_a\) gives
$
z_a\rightsquigarrow z_c,
$
so again
$
z_a\preceq_{\mathcal E}z_c.
$
Similarly, if
$
z_a\rightsquigarrow z_b
$
and
$
z_b=z_c,
$
then
$
z_a\rightsquigarrow z_c,
$
and therefore
$
z_a\preceq_{\mathcal E}z_c.
$

It remains to consider the case in which both relations are realized by
nontrivial directed paths. Suppose
\[
z_a
=
z_{p_0}
\to
z_{p_1}
\to
\cdots
\to
z_{p_r}
=
z_b
\]
and
$
z_b
=
z_{q_0}
\to
z_{q_1}
\to
\cdots
\to
z_{q_s}
=
z_c.
$
Since
$
z_{p_r}=z_b=z_{q_0},
$
the two paths may be concatenated to obtain the directed walk
\[
z_a
=
z_{p_0}
\to
z_{p_1}
\to
\cdots
\to
z_{p_r}
=
z_{q_0}
\to
z_{q_1}
\to
\cdots
\to
z_{q_s}
=
z_c.
\]
Every edge of this concatenated walk belongs to \(\mathcal E\). If the walk
contains repeated intermediate vertices, one may remove the portion between
two consecutive occurrences of any repeated vertex. Repeating this deletion
finitely many times produces a directed path with the same initial and
terminal vertices. Hence there exists a directed path from \(z_a\) to
\(z_c\), that is,
$
z_a\rightsquigarrow z_c.
$
Consequently,
$
z_a\preceq_{\mathcal E}z_c.
$
Thus \(\preceq_{\mathcal E}\) is transitive.

We now establish antisymmetry, which is the point at which acyclicity is
essential. Suppose
$
z_a\preceq_{\mathcal E}z_b
\qquad\text{and}\qquad
z_b\preceq_{\mathcal E}z_a.
$
We must show that
$
z_a=z_b.
$
Assume, toward a contradiction, that
$
z_a\neq z_b.
$
Because the vertices are distinct, neither of the two relations above can
hold through the equality alternative in the definition of
\(\preceq_{\mathcal E}\). Therefore
$
z_a\rightsquigarrow z_b
\qquad\text{and}\qquad
z_b\rightsquigarrow z_a.
$
Hence there exist directed paths
\[
z_a
=
z_{p_0}
\to
z_{p_1}
\to
\cdots
\to
z_{p_r}
=
z_b
\]
and
$
z_b
=
z_{q_0}
\to
z_{q_1}
\to
\cdots
\to
z_{q_s}
=
z_a,
$
with
$
r\ge1,
\qquad
s\ge1.
$
Concatenating them produces a nontrivial directed closed walk
\[
z_a
=
z_{p_0}
\to
\cdots
\to
z_{p_r}
=
z_b
=
z_{q_0}
\to
\cdots
\to
z_{q_s}
=
z_a.
\]
This closed walk has positive length, since
$
r+s\ge2.
$

We briefly justify why such a finite directed closed walk necessarily
contains a directed cycle. Write the vertices of the closed walk as
$
v_0,v_1,\ldots,v_m,
\qquad
v_m=v_0,
\qquad
m\ge1.
$
Among all pairs of indices
$
0\le p<q\le m
$
satisfying
$
v_p=v_q,
$
choose one for which \(q-p\) is minimal. Then
\[
v_p\to v_{p+1}\to\cdots\to v_q=v_p
\]
is a directed closed subwalk. By minimality of \(q-p\), the intermediate
vertices
$
v_p,v_{p+1},\ldots,v_{q-1}
$
are pairwise distinct; otherwise a repeated pair of intermediate vertices
would determine a shorter directed closed subwalk. Hence
$
v_p\to v_{p+1}\to\cdots\to v_{q-1}\to v_p
$
is a directed cycle.

Applying this observation to the closed walk obtained from the two
reachability paths shows that \((X,\mathcal E)\) contains a directed cycle.
This contradicts the assumption that \((X,\mathcal E)\) is acyclic.
Therefore the assumption
$
z_a\neq z_b
$
is impossible, and we must have
$
z_a=z_b.
$
Thus \(\preceq_{\mathcal E}\) is antisymmetric.

We have shown that, when \((X,\mathcal E)\) is acyclic,
\(\preceq_{\mathcal E}\) is reflexive, transitive, and antisymmetric.
Therefore it is a partial order on \(X\).

Conversely, suppose that \((X,\mathcal E)\) contains a directed cycle with
at least two distinct vertices. Write such a cycle as
\[
z_{i_0}
\to
z_{i_1}
\to
\cdots
\to
z_{i_{m-1}}
\to
z_{i_m},
\qquad
z_{i_m}=z_{i_0},
\]
with
$
m\ge2,
$
and with at least two distinct vertices among
$
z_{i_0},\ldots,z_{i_{m-1}}.
$
Choose two distinct vertices occurring on the cycle, and denote them by
\(z_a\) and \(z_b\). After cyclically relabeling the cycle if necessary, we
may write it in the form
\[
z_a
\to
\cdots
\to
z_b
\to
\cdots
\to
z_a.
\]
The first segment of the cycle is a directed path from \(z_a\) to \(z_b\),
so
$
z_a\rightsquigarrow z_b.
$
Hence
$
z_a\preceq_{\mathcal E}z_b.
$
The remaining segment of the same cycle is a directed path from \(z_b\) back
to \(z_a\), so
$
z_b\rightsquigarrow z_a,
$
and therefore
$
z_b\preceq_{\mathcal E}z_a.
$
By construction,
$
z_a\neq z_b.
$
Thus we have two distinct elements satisfying
$
z_a\preceq_{\mathcal E}z_b
\qquad\text{and}\qquad
z_b\preceq_{\mathcal E}z_a.
$
This violates antisymmetry. Consequently,
\(\preceq_{\mathcal E}\) is not a partial order whenever the admissibility
graph contains a directed cycle involving at least two distinct vertices.
\end{proof}

\subsection{Proof of Proposition~\ref{prop:prelim-matrix-realization-of-kernels}}

\begin{proof}
Recall that
$
C_{\mathfrak S}^{\mathrm{ker}}
=
\left\{
\theta\in\mathbb R^R
\;\middle|\;
\theta_r\ge0
\text{ for every }r\in\mathcal R_+
\right\},
$
and that, for
\(\theta\in C_{\mathfrak S}^{\mathrm{ker}}\),
the associated admissible kernel is
\[
K_\theta(z_i,z_j)
=
\sum_{r=1}^R
\theta_r\widetilde{\phi}_r(z_i,z_j).
\]
Hence
$
\mathcal K_{\mathfrak S}
=
\left\{
\sum_{r=1}^R
\theta_r\widetilde{\phi}_r
\;\middle|\;
\theta\in C_{\mathfrak S}^{\mathrm{ker}}
\right\}.
$

Since
$
X=\{z_1,\ldots,z_d\}
$
is finite, the vector space of all real-valued functions on
\(X\times X\) is finite-dimensional. More precisely, evaluation on the
ordered pairs
$
(z_i,z_j),
\qquad
1\le i,j\le d,
$
identifies this function space with
\(\mathbb R^{d^2}\). In particular,
\(\mathcal K_{\mathfrak S}\) is contained in the finite-dimensional
linear subspace
$
\operatorname{span}
\{
\widetilde{\phi}_1,\ldots,\widetilde{\phi}_R
\},
$
and therefore
$
\dim
\operatorname{span}(\mathcal K_{\mathfrak S})
\le R.
$
This establishes the asserted finite-dimensionality.

We next verify the conic and convex structure directly. Let
$
K_\theta,K_\eta
\in
\mathcal K_{\mathfrak S},
$
where
$
\theta,\eta
\in
C_{\mathfrak S}^{\mathrm{ker}}.
$
Let \(a,b\ge0\). Then
\[
aK_\theta+bK_\eta
=
\sum_{r=1}^R
(a\theta_r+b\eta_r)
\widetilde{\phi}_r.
\]
For every
\(r\in\mathcal R_+\),
$
\theta_r\ge0
\qquad\text{and}\qquad
\eta_r\ge0,
$
and hence
$
a\theta_r+b\eta_r\ge0.
$
Thus
$
a\theta+b\eta
\in
C_{\mathfrak S}^{\mathrm{ker}},
$
which implies
$
aK_\theta+bK_\eta
\in
\mathcal K_{\mathfrak S}.
$
Therefore
\(\mathcal K_{\mathfrak S}\) is a convex cone.

We now make its polyhedral structure explicit. For every index
\(r\notin\mathcal R_+\), the coefficient \(\theta_r\) is unrestricted.
Write its positive and negative parts as
\[
\theta_r^+
:=
\max\{\theta_r,0\},
\qquad
\theta_r^-
:=
\max\{-\theta_r,0\}.
\]
Then
$
\theta_r^+\ge0,
\qquad
\theta_r^-\ge0,
$
and
$
\theta_r
=
\theta_r^+-\theta_r^-.
$
Consequently, every kernel in
\(\mathcal K_{\mathfrak S}\) may be written as
\[
K_\theta
=
\sum_{r\in\mathcal R_+}
\theta_r\widetilde{\phi}_r
+
\sum_{r\notin\mathcal R_+}
\theta_r^+\widetilde{\phi}_r
+
\sum_{r\notin\mathcal R_+}
\theta_r^-(-\widetilde{\phi}_r),
\]
where all coefficients appearing on the right-hand side are nonnegative.

Conversely, every nonnegative linear combination of the functions
\[
\mathcal G
:=
\{
\widetilde{\phi}_r:
r\in\mathcal R_+
\}
\cup
\{
\widetilde{\phi}_r,-\widetilde{\phi}_r:
r\notin\mathcal R_+
\}
\]
corresponds to an admissible coefficient vector in
\(C_{\mathfrak S}^{\mathrm{ker}}\). Hence
$
\mathcal K_{\mathfrak S}
=
\operatorname{cone}(\mathcal G).
$
The set \(\mathcal G\) is finite, containing at most \(2R\) functions.
Thus \(\mathcal K_{\mathfrak S}\) is a finitely generated convex cone.
By the finite-dimensional Minkowski--Weyl theorem, a finitely generated
convex cone is a polyhedral cone. Therefore
$
\mathcal K_{\mathfrak S}
$
is polyhedral.

For completeness, we also verify closedness directly, rather than relying
only on the general polyhedral-cone theorem. Enumerate the finite generating
set \(\mathcal G\) as
$
G_1,\ldots,G_M.
$
Then
\[
\mathcal K_{\mathfrak S}
=
\left\{
\sum_{\ell=1}^M
c_\ell G_\ell
\;\middle|\;
c_\ell\ge0
\right\}.
\]
Let
$
K_n\in\mathcal K_{\mathfrak S}
$
be a sequence satisfying
$
K_n\longrightarrow K
$
in the finite-dimensional function space on \(X\times X\). For every \(n\),
choose a conic representation
$
K_n
=
\sum_{\ell=1}^M
c_\ell^{(n)}G_\ell,
\qquad
c_\ell^{(n)}\ge0,
$
having the smallest possible number of strictly positive coefficients among
all such representations of \(K_n\).

We claim that the active generators in this representation are linearly
independent. Let
$
I_n
:=
\{
\ell:
c_\ell^{(n)}>0
\}.
$
Suppose, to the contrary, that
$
\{G_\ell:\ell\in I_n\}
$
is linearly dependent. Then there exist real numbers
\(\beta_\ell\), not all zero, such that$
\sum_{\ell\in I_n}
\beta_\ell G_\ell
=
0.
$
Multiplying all \(\beta_\ell\) by \(-1\), if necessary, we may assume that
at least one of them is strictly positive. Define
$
t
:=
\min_{\substack{\ell\in I_n\\ \beta_\ell>0}}
\frac{c_\ell^{(n)}}{\beta_\ell}.
$
Since every coefficient \(c_\ell^{(n)}\) with
\(\ell\in I_n\) is strictly positive,
$
t>0.
$
Set
\[
\widetilde c_\ell^{(n)}
:=
c_\ell^{(n)}-t\beta_\ell
\qquad
(\ell\in I_n).
\]
If \(\beta_\ell>0\), the definition of \(t\) gives
$
\widetilde c_\ell^{(n)}
\ge0.
$
If \(\beta_\ell\le0\), then
$
\widetilde c_\ell^{(n)}
=
c_\ell^{(n)}-t\beta_\ell
\ge
c_\ell^{(n)}
>0.
$
Moreover, for at least one index attaining the minimum in the definition of
\(t\),
$
\widetilde c_\ell^{(n)}=0.
$
At the same time,
\[
\begin{aligned}
\sum_{\ell\in I_n}
\widetilde c_\ell^{(n)}G_\ell
&=
\sum_{\ell\in I_n}
c_\ell^{(n)}G_\ell
-
t
\sum_{\ell\in I_n}
\beta_\ell G_\ell\\
&=
K_n.
\end{aligned}
\]
We have therefore produced a conic representation of \(K_n\) with strictly
fewer positive coefficients, contradicting the chosen minimality.
Thus the active generators must be linearly independent.

Since there are only finitely many subsets of
\(\{1,\ldots,M\}\), after passing to a subsequence we may assume that the
active set is a fixed subset
$
I\subseteq\{1,\ldots,M\}
$
for every \(n\) in the subsequence. Hence
$
K_n
=
\sum_{\ell\in I}
c_\ell^{(n)}G_\ell,
$
where
$
\{G_\ell:\ell\in I\}
$
is linearly independent.

Consider the linear map
\[
T_I:\mathbb R^{|I|}
\longrightarrow
\mathbb R^{X\times X},
\qquad
T_I(c)
=
\sum_{\ell\in I}c_\ell G_\ell.
\]
Linear independence of the \(G_\ell\) implies that \(T_I\) is injective.
Because both its domain and its image are finite-dimensional, the inverse
map
$
T_I^{-1}:
\operatorname{range}(T_I)
\longrightarrow
\mathbb R^{|I|}
$
is continuous. Consequently, there exists a constant \(C_I>0\) such that
$
\|c\|_2
\le
C_I
\|T_I(c)\|
$
for every \(c\in\mathbb R^{|I|}\), where any fixed norm on the
finite-dimensional function space may be used.

Since
$
K_n\longrightarrow K,
$
the sequence \((K_n)\) is bounded. Therefore
$
\left\|
(c_\ell^{(n)})_{\ell\in I}
\right\|_2
$
is also bounded. Passing to a further subsequence, we may assume that
$
c_\ell^{(n)}
\longrightarrow
c_\ell
\qquad
\text{for every }\ell\in I.
$
Because every
\(c_\ell^{(n)}\ge0\),
$
c_\ell\ge0.
$
Taking limits in
$
K_n
=
\sum_{\ell\in I}
c_\ell^{(n)}G_\ell
$
gives
$
K
=
\sum_{\ell\in I}
c_\ell G_\ell.
$
Thus
$
K\in\mathcal K_{\mathfrak S}.
$
Therefore \(\mathcal K_{\mathfrak S}\) is closed. We have now established
that it is a finite-dimensional closed convex polyhedral cone.

We next verify the matrix realization. By
Definition~\ref{def:prelim-geometry-descriptor} and the definition of
\(\widetilde{\phi}_r\) above,
$
(\Phi_r)_{ij}
=
\widetilde{\phi}_r(z_i,z_j)
\qquad
\text{for every }r,i,j.
$
For an admissible coefficient vector
\(\theta\in C_{\mathfrak S}^{\mathrm{ker}}\), the matrix associated with
\(K_\theta\) is defined entrywise by
$
[K_\theta]_{ij}
=
K_\theta(z_i,z_j).
$
Using the definition of \(K_\theta\),
\[
\begin{aligned}
[K_\theta]_{ij}
&=
K_\theta(z_i,z_j)\\
&=
\sum_{r=1}^R
\theta_r
\widetilde{\phi}_r(z_i,z_j)\\
&=
\sum_{r=1}^R
\theta_r
(\Phi_r)_{ij}\\
&=
\left(
\sum_{r=1}^R
\theta_r\Phi_r
\right)_{ij}.
\end{aligned}
\]
This identity holds for every pair
\(1\le i,j\le d\). Two matrices are equal if and only if all of their
entries are equal, and therefore
$
[K_\theta]
=
\sum_{r=1}^R
\theta_r\Phi_r.
$

The identification
$
K\longmapsto[K],
\qquad
[K]_{ij}=K(z_i,z_j),
$
is itself one-to-one. Indeed, if two kernels \(K\) and \(K'\) satisfy
$
[K]=[K'],
$
then
$
K(z_i,z_j)
=
[K]_{ij}
=
[K']_{ij}
=
K'(z_i,z_j)
$
for every ordered pair
\((z_i,z_j)\in X\times X\). Since these ordered pairs exhaust the domain
\(X\times X\),
$
K=K'.
$
Thus the matrix realization loses no kernel-level information.

Finally, assume that the masked channels
$
\widetilde{\phi}_1,\ldots,\widetilde{\phi}_R
$
are linearly independent as functions on \(X\times X\). Suppose two
admissible coefficient vectors
$
\theta,\eta
\in
C_{\mathfrak S}^{\mathrm{ker}}
$
produce the same kernel:
$
K_\theta=K_\eta.
$
Subtracting the two representations gives
$
0
=
K_\theta-K_\eta
=
\sum_{r=1}^R
(\theta_r-\eta_r)
\widetilde{\phi}_r.
$
Linear independence of the masked channels implies that every coefficient in
this linear combination must vanish:
$
\theta_r-\eta_r=0
\qquad
\text{for every }r=1,\ldots,R.
$
Hence
$
\theta_r=\eta_r
\qquad
\text{for every }r,
$
and therefore
$
\theta=\eta.
$
Thus the map
$
\theta\longmapsto K_\theta
$
is injective. Since every kernel in
\(\mathcal K_{\mathfrak S}\) has at least one admissible coefficient
representation by definition, injectivity implies that this representation
is unique for every
\(K\in\mathcal K_{\mathfrak S}\).
\end{proof}

\subsection{Proof of Proposition~\ref{prop:operators-interaction-family-structure}}

\begin{proof}
By Definition~\ref{def:operators-interaction-family},
\[
\mathcal A_{\mathfrak S}
=
\left\{
\sum_{r=1}^R\theta_r\Phi_r
+
\operatorname{Diag}(\delta)
\;\middle|\;
\theta\in\mathbb R^R,\;
\delta\in\mathbb R^d,\;
\theta_r\ge0
\text{ for }r\in\mathcal R_+
\right\}.
\]
Write
$
\delta=(\delta_1,\ldots,\delta_d)^\top.
$
Since \(E_{ii}\) denotes the standard matrix unit with a single \(1\) in the
\((i,i)\)-entry, we have
$
\operatorname{Diag}(\delta)
=
\sum_{i=1}^d\delta_iE_{ii}.
$
Hence every \(A\in\mathcal A_{\mathfrak S}\) can be written as
$
A
=
\sum_{r=1}^R\theta_r\Phi_r
+
\sum_{i=1}^d\delta_iE_{ii}.
$

We first identify the family exactly as a finitely generated cone. For every
index \(r\notin\mathcal R_+\), the coefficient \(\theta_r\) is unrestricted.
Define its positive and negative parts by
\[
\theta_r^+
:=
\max\{\theta_r,0\},
\qquad
\theta_r^-
:=
\max\{-\theta_r,0\}.
\]
Then
$
\theta_r^+\ge0,
\qquad
\theta_r^-\ge0,
\qquad
\theta_r
=
\theta_r^+-\theta_r^-.
$
Likewise, for every diagonal coefficient \(\delta_i\), define
\[
\delta_i^+
:=
\max\{\delta_i,0\},
\qquad
\delta_i^-
:=
\max\{-\delta_i,0\},
\]
so that
$
\delta_i^\pm\ge0
\qquad\text{and}\qquad
\delta_i
=
\delta_i^+-\delta_i^-.
$
Substituting these decompositions into the representation of \(A\) gives
\[
\begin{aligned}
A
&=
\sum_{r\in\mathcal R_+}
\theta_r\Phi_r\\
&\quad+
\sum_{r\notin\mathcal R_+}
\theta_r^+\Phi_r
+
\sum_{r\notin\mathcal R_+}
\theta_r^-(-\Phi_r)\\
&\quad+
\sum_{i=1}^d
\delta_i^+E_{ii}
+
\sum_{i=1}^d
\delta_i^-(-E_{ii}).
\end{aligned}
\]
Every coefficient appearing in this expression is nonnegative. Therefore
every matrix in \(\mathcal A_{\mathfrak S}\) belongs to the conic hull of
the finite collection
\[
\mathcal G
:=
\{\Phi_r:r\in\mathcal R_+\}
\cup
\{\Phi_r,-\Phi_r:r\notin\mathcal R_+\}
\cup
\{E_{11},-E_{11},\ldots,E_{dd},-E_{dd}\}.
\]
Thus
$
\mathcal A_{\mathfrak S}
\subseteq
\operatorname{cone}(\mathcal G).
$

We now prove the reverse inclusion. Let
\(A\in\operatorname{cone}(\mathcal G)\). Then there exist nonnegative
numbers
$
a_r\ge0
\quad (r\in\mathcal R_+),
$
nonnegative numbers
$
b_r^+,b_r^-\ge0
\quad (r\notin\mathcal R_+),
$
and nonnegative numbers
$
c_i^+,c_i^-\ge0
\quad (1\le i\le d)
$
such that
\[
\begin{aligned}
A
&=
\sum_{r\in\mathcal R_+}
a_r\Phi_r\\
&\quad+
\sum_{r\notin\mathcal R_+}
\bigl(
b_r^+\Phi_r+b_r^-(-\Phi_r)
\bigr)\\
&\quad+
\sum_{i=1}^d
\bigl(
c_i^+E_{ii}+c_i^-(-E_{ii})
\bigr).
\end{aligned}
\]
Define
$
\theta_r
:=
a_r
\qquad
(r\in\mathcal R_+),
$
and
$
\theta_r
:=
b_r^+-b_r^-
\qquad
(r\notin\mathcal R_+).
$
Also define
$
\delta_i
:=
c_i^+-c_i^-,
\qquad
i=1,\ldots,d.
$
Then
$
\theta_r\ge0
\qquad
\text{for every }r\in\mathcal R_+,
$
while the coefficients indexed by
\(r\notin\mathcal R_+\) and the diagonal coefficients \(\delta_i\) are
allowed to have arbitrary sign. Recombining the preceding expression gives
\[
A
=
\sum_{r=1}^R
\theta_r\Phi_r
+
\sum_{i=1}^d\delta_iE_{ii}
=
\sum_{r=1}^R
\theta_r\Phi_r
+
\operatorname{Diag}(\delta).
\]
Hence
$
A\in\mathcal A_{\mathfrak S}.
$
Therefore
$
\mathcal A_{\mathfrak S}
=
\operatorname{cone}(\mathcal G).
$

This identity immediately gives the cone property. Indeed, let
$
A,B\in\mathcal A_{\mathfrak S}
$
and let
$
\alpha,\beta\ge0.
$
Since both \(A\) and \(B\) are nonnegative linear combinations of the
generators in \(\mathcal G\), their combination
$
\alpha A+\beta B
$
is again a nonnegative linear combination of the same generators. Hence
$
\alpha A+\beta B
\in
\mathcal A_{\mathfrak S}.
$
Thus \(\mathcal A_{\mathfrak S}\) is a convex cone. In particular, taking
\(\beta=1-\alpha\) with \(0\le\alpha\le1\) shows explicitly that it is
convex.

The same representation also makes finite-dimensionality transparent.
Every generator in \(\mathcal G\) belongs to
\[
\operatorname{span}
\{
\Phi_1,\ldots,\Phi_R,
E_{11},\ldots,E_{dd}
\}.
\]
Therefore
$
\operatorname{span}(\mathcal A_{\mathfrak S})
\subseteq
\operatorname{span}
\{
\Phi_1,\ldots,\Phi_R,
E_{11},\ldots,E_{dd}
\}.
$
Consequently,
$
\dim
\operatorname{span}(\mathcal A_{\mathfrak S})
\le
R+d
<
\infty.
$
In particular, \(\mathcal A_{\mathfrak S}\) is a finite-dimensional cone in
\(\mathbb R^{d\times d}\).

Because \(\mathcal G\) is finite, the equality
$
\mathcal A_{\mathfrak S}
=
\operatorname{cone}(\mathcal G)
$
shows that \(\mathcal A_{\mathfrak S}\) is a finitely generated convex cone.
By the finite-dimensional Minkowski--Weyl theorem, finitely generated convex
cones are exactly polyhedral convex cones. Hence
$
\mathcal A_{\mathfrak S}
$
is polyhedral.

For completeness, we now verify closedness directly. Enumerate the finite
generating collection as
$
\mathcal G
=
\{G_1,\ldots,G_M\}.
$
Thus
\[
\mathcal A_{\mathfrak S}
=
\left\{
\sum_{\ell=1}^M c_\ell G_\ell
\;\middle|\;
c_\ell\ge0
\right\}.
\]
Let
$
A_n\in\mathcal A_{\mathfrak S}
$
be a sequence converging in Frobenius norm to some matrix
$
A\in\mathbb R^{d\times d},
\qquad
A_n\longrightarrow A.
$
For every \(n\), choose among all conic representations of \(A_n\) one with
the smallest possible number of strictly positive coefficients. Thus write
$
A_n
=
\sum_{\ell=1}^M
c_\ell^{(n)}G_\ell,
\qquad
c_\ell^{(n)}\ge0,
$
and define the active index set
$
I_n
:=
\{\ell:c_\ell^{(n)}>0\}.
$

We claim that
$
\{G_\ell:\ell\in I_n\}
$
is linearly independent. Suppose otherwise. Then there exist scalars
\(\beta_\ell\), not all zero, such that
$
\sum_{\ell\in I_n}
\beta_\ell G_\ell
=
0.
$
Replacing all \(\beta_\ell\) by their negatives if necessary, we may assume
that at least one \(\beta_\ell\) is strictly positive. Define
\[
t
:=
\min_{\substack{\ell\in I_n\\ \beta_\ell>0}}
\frac{c_\ell^{(n)}}{\beta_\ell}.
\]
Every numerator in this minimum is positive, so
$
t>0.
$
Now set
$
\widetilde c_\ell^{(n)}
:=
c_\ell^{(n)}-t\beta_\ell
\qquad
(\ell\in I_n).
$
If \(\beta_\ell>0\), then the definition of \(t\) gives
$
t
\le
\frac{c_\ell^{(n)}}{\beta_\ell},
$
and hence
$
\widetilde c_\ell^{(n)}
=
c_\ell^{(n)}-t\beta_\ell
\ge0.
$
If \(\beta_\ell\le0\), then
$
\widetilde c_\ell^{(n)}
=
c_\ell^{(n)}-t\beta_\ell
\ge
c_\ell^{(n)}
>0.
$
Furthermore, for at least one index attaining the minimum in the definition
of \(t\),
$
\widetilde c_\ell^{(n)}=0.
$
At the same time,
\[
\begin{aligned}
\sum_{\ell\in I_n}
\widetilde c_\ell^{(n)}G_\ell
&=
\sum_{\ell\in I_n}
\left(
c_\ell^{(n)}-t\beta_\ell
\right)G_\ell\\
&=
\sum_{\ell\in I_n}
c_\ell^{(n)}G_\ell
-
t
\sum_{\ell\in I_n}
\beta_\ell G_\ell\\
&=
A_n.
\end{aligned}
\]
Thus \(A_n\) would admit another conic representation with strictly fewer
positive coefficients than the representation we selected, which is a
contradiction. Therefore the active generators are linearly independent.

There are only finitely many subsets of
\(\{1,\ldots,M\}\). Hence, after passing to a subsequence, we may assume
that
$
I_n=I
$
for one fixed subset \(I\subseteq\{1,\ldots,M\}\) and every \(n\) in the
subsequence. We may therefore write
$
A_n
=
\sum_{\ell\in I}
c_\ell^{(n)}G_\ell,
$
where the family
$
\{G_\ell:\ell\in I\}
$
is linearly independent.

Consider the linear map
$
T_I:
\mathbb R^{|I|}
\longrightarrow
\mathbb R^{d\times d},
\qquad
T_I(c)
=
\sum_{\ell\in I}
c_\ell G_\ell.
$
Because the matrices \(G_\ell\), \(\ell\in I\), are linearly independent,
\(T_I\) is injective. It is therefore a linear isomorphism from
\(\mathbb R^{|I|}\) onto its image
$
\operatorname{range}(T_I).
$
Both spaces are finite-dimensional, so the inverse
$
T_I^{-1}:
\operatorname{range}(T_I)
\longrightarrow
\mathbb R^{|I|}
$
is continuous. Consequently, there exists a constant \(C_I>0\) such that
$
\|c\|_2
\le
C_I\|T_I(c)\|_F
\qquad
\text{for every }c\in\mathbb R^{|I|}.
$
Applying this to
$
c^{(n)}
=
(c_\ell^{(n)})_{\ell\in I}
$
gives
$
\|c^{(n)}\|_2
\le
C_I\|A_n\|_F.
$
Since
$
A_n\longrightarrow A,
$
the sequence \((A_n)\) is bounded, and therefore the coefficient sequence
\((c^{(n)})\) is bounded as well. By the Bolzano--Weierstrass theorem in
finite-dimensional Euclidean space, after passing to a further subsequence
we may assume that
$
c_\ell^{(n)}
\longrightarrow
c_\ell
\qquad
\text{for every }\ell\in I.
$
Because
$
c_\ell^{(n)}\ge0
$
for every \(n\), taking the limit gives
$
c_\ell\ge0.
$
Passing to the limit in
$
A_n
=
\sum_{\ell\in I}
c_\ell^{(n)}G_\ell
$
yields
$
A
=
\sum_{\ell\in I}
c_\ell G_\ell.
$
The coefficients on the right are nonnegative, so
\[
A\in
\operatorname{cone}(\mathcal G)
=
\mathcal A_{\mathfrak S}.
\]
Thus every convergent sequence in
\(\mathcal A_{\mathfrak S}\) has its limit in
\(\mathcal A_{\mathfrak S}\). Since
\(\mathbb R^{d\times d}\) is a finite-dimensional metric space, sequential
closedness is equivalent to closedness. Therefore
$
\mathcal A_{\mathfrak S}
$
is closed.

It remains to consider the special case
$
\mathcal R_+=\varnothing.
$
Then no channel coefficient is subject to a sign restriction. Hence, by
Definition~\ref{def:operators-interaction-family},
\[
\mathcal A_{\mathfrak S}
=
\left\{
\sum_{r=1}^R
\theta_r\Phi_r
+
\sum_{i=1}^d
\delta_iE_{ii}
\;\middle|\;
\theta_r,\delta_i\in\mathbb R
\right\}.
\]
The right-hand side is precisely the linear span
\[
\mathcal A_{\mathfrak S}
=
\operatorname{span}
\{
\Phi_1,\ldots,\Phi_R,
E_{11},\ldots,E_{dd}
\}.
\]
Therefore, whenever
\(\mathcal R_+=\varnothing\),
\(\mathcal A_{\mathfrak S}\) is a linear subspace of
\(\mathbb R^{d\times d}\).

Combining the preceding conclusions, \(\mathcal A_{\mathfrak S}\) is a
finite-dimensional closed polyhedral convex cone, it is generated by
$
\{\Phi_r:r\in\mathcal R_+\}
\cup
\{\pm\Phi_r:r\notin\mathcal R_+\}
\cup
\{\pm E_{11},\ldots,\pm E_{dd}\},
$
and in the absence of sign-constrained channels it reduces to a linear
subspace.
\end{proof}

\subsection{Proof of Proposition~\ref{prop:operators-operator-identifiability}}

\begin{proof}
Let \(A\in\mathcal A_{\mathfrak S}\) be arbitrary. By
Definition~\ref{def:operators-interaction-family}, membership in
\(\mathcal A_{\mathfrak S}\) means precisely that there exist coefficients
\[
\theta=(\theta_1,\ldots,\theta_R)^\top\in\mathbb R^R
\qquad\text{and}\qquad
\delta=(\delta_1,\ldots,\delta_d)^\top\in\mathbb R^d
\]
such that
$
\theta_r\ge0
\qquad
\text{for every }r\in\mathcal R_+,
$
and
$
A
=
\sum_{r=1}^R\theta_r\Phi_r
+
\operatorname{Diag}(\delta).
$
Thus the existence of at least one representation satisfying the required
sign constraints follows directly from the definition of the interaction
family. It remains to prove that this representation is unique.

Suppose, therefore, that the same operator \(A\) admits two representations
of the prescribed form. Thus assume that
$
A
=
\sum_{r=1}^R\theta_r\Phi_r
+
\operatorname{Diag}(\delta)
$
and also
$
A
=
\sum_{r=1}^R\widetilde{\theta}_r\Phi_r
+
\operatorname{Diag}(\widetilde{\delta}),
$
where
$
\theta,\widetilde{\theta}\in\mathbb R^R,
\qquad
\delta,\widetilde{\delta}\in\mathbb R^d,
$
and both channel coefficient vectors satisfy the feasibility conditions
$
\theta_r\ge0
\qquad\text{and}\qquad
\widetilde{\theta}_r\ge0
\qquad
\text{for every }r\in\mathcal R_+.
$
Because the two expressions represent the same matrix \(A\), subtracting the
second equality from the first gives
\[
0
=
\sum_{r=1}^R
\theta_r\Phi_r
-
\sum_{r=1}^R
\widetilde{\theta}_r\Phi_r
+
\operatorname{Diag}(\delta)
-
\operatorname{Diag}(\widetilde{\delta}).
\]
Collecting the channel coefficients yields
$
0
=
\sum_{r=1}^R
(\theta_r-\widetilde{\theta}_r)\Phi_r
+
\operatorname{Diag}(\delta)
-
\operatorname{Diag}(\widetilde{\delta}).
$
Since the diagonal map is linear,
$
\operatorname{Diag}(\delta)
-
\operatorname{Diag}(\widetilde{\delta})
=
\operatorname{Diag}(\delta-\widetilde{\delta}),
$
and hence
$
0
=
\sum_{r=1}^R
(\theta_r-\widetilde{\theta}_r)\Phi_r
+
\operatorname{Diag}(\delta-\widetilde{\delta}).
$
Writing
$
\delta-\widetilde{\delta}
=
\bigl(
\delta_1-\widetilde{\delta}_1,
\ldots,
\delta_d-\widetilde{\delta}_d
\bigr)^\top,
$
and recalling that \(E_{ii}\) denotes the standard matrix unit whose only
nonzero entry is a \(1\) in position \((i,i)\), we have
$
\operatorname{Diag}(\delta-\widetilde{\delta})
=
\sum_{i=1}^d
(\delta_i-\widetilde{\delta}_i)E_{ii}.
$
Therefore the equality above becomes
$
\sum_{r=1}^R
(\theta_r-\widetilde{\theta}_r)\Phi_r
+
\sum_{i=1}^d
(\delta_i-\widetilde{\delta}_i)E_{ii}
=
0.
$
This is a linear relation among the matrices
$
\Phi_1,\ldots,\Phi_R,E_{11},\ldots,E_{dd}.
$
By the hypothesis of the proposition, this family is linearly independent in
\(\mathbb R^{d\times d}\). By the definition of linear independence, the only
linear combination of these matrices that equals the zero matrix is the
trivial linear combination. Consequently, every coefficient in the preceding
relation must vanish. Hence
$
\theta_r-\widetilde{\theta}_r=0
\qquad
\text{for every }r=1,\ldots,R,
$
and
$
\delta_i-\widetilde{\delta}_i=0
\qquad
\text{for every }i=1,\ldots,d.
$
Equivalently,
$
\theta_r=\widetilde{\theta}_r
\qquad
\text{for every }r=1,\ldots,R,
$
and
$
\delta_i=\widetilde{\delta}_i
\qquad
\text{for every }i=1,\ldots,d.
$
Thus
$
\theta=\widetilde{\theta}
\qquad\text{and}\qquad
\delta=\widetilde{\delta}.
$
Therefore two distinct admissible coefficient pairs cannot represent the same
operator \(A\). Since existence was already guaranteed by
\(A\in\mathcal A_{\mathfrak S}\), every
\(A\in\mathcal A_{\mathfrak S}\) admits exactly one representation
$
A
=
\sum_{r=1}^R\theta_r\Phi_r
+
\operatorname{Diag}(\delta)
$
with
$
\theta_r\ge0
\qquad
\text{for }r\in\mathcal R_+.
$
In fact, the linear-independence assumption proves the slightly stronger
statement that the coefficient map
$
(\theta,\delta)
\longmapsto
\sum_{r=1}^R\theta_r\Phi_r+\operatorname{Diag}(\delta)
$
is injective on all of \(\mathbb R^R\times\mathbb R^d\); the sign constraints
only restrict the admissible parameter set and are not needed for the
uniqueness argument itself.
\end{proof}

\subsection{Proof of Proposition~\ref{prop:operators-generator-family-structure}}

\begin{proof}
Define the auxiliary set
\[
\mathcal C_{\mathrm{gen}}
:=
\Bigl\{
L\in\mathbb R^{d\times d}
\;\Big|\;
L_{ij}\ge0 \text{ for every } i\neq j,
\quad
L\mathbf 1=0
\Bigr\}.
\]
We first record explicitly the geometric structure of this constraint set.
Since
$
\mathbf 1=(1,\ldots,1)^\top,
$
the \(i\)-th component of \(L\mathbf 1\) is
$
(L\mathbf 1)_i
=
\sum_{j=1}^d L_{ij}.
$
Consequently,
$
L\mathbf 1=0
$
is equivalent to the system of \(d\) homogeneous linear equations
$
\sum_{j=1}^d L_{ij}=0,
\qquad
i=1,\ldots,d.
$
Therefore
\[
\mathcal C_{\mathrm{gen}}
=
\left\{
L\in\mathbb R^{d\times d}
\;\middle|\;
L_{ij}\ge0
\text{ for }i\neq j,
\quad
\sum_{j=1}^dL_{ij}=0
\text{ for }i=1,\ldots,d
\right\}.
\]
Each equality in the preceding description can, if desired, be written as a
pair of homogeneous linear inequalities:
$
\sum_{j=1}^dL_{ij}=0
\quad\Longleftrightarrow\quad
\sum_{j=1}^dL_{ij}\ge0
\quad\text{and}\quad
-\sum_{j=1}^dL_{ij}\ge0.
$
Hence
\[
\begin{aligned}
\mathcal C_{\mathrm{gen}}
&=
\bigcap_{\substack{1\le i,j\le d\\ i\neq j}}
\left\{
L\in\mathbb R^{d\times d}:L_{ij}\ge0
\right\}
\\
&\qquad\cap
\bigcap_{i=1}^d
\left\{
L\in\mathbb R^{d\times d}:
\sum_{j=1}^dL_{ij}\ge0
\right\}
\\
&\qquad\cap
\bigcap_{i=1}^d
\left\{
L\in\mathbb R^{d\times d}:
-\sum_{j=1}^dL_{ij}\ge0
\right\}.
\end{aligned}
\]
Every set appearing on the right-hand side is a homogeneous linear
half-space in the finite-dimensional vector space
\(\mathbb R^{d\times d}\). There are only finitely many such half-spaces:
there are \(d(d-1)\) inequalities enforcing off-diagonal nonnegativity and
\(2d\) inequalities encoding the \(d\) zero-row-sum equations. It follows
directly that \(\mathcal C_{\mathrm{gen}}\) is polyhedral.

The same representation also makes closedness explicit. For fixed
\(i\neq j\), the coordinate map
$
L\longmapsto L_{ij}
$
is a linear, hence continuous, functional on
\(\mathbb R^{d\times d}\). Thus
$
\left\{
L:L_{ij}\ge0
\right\}
$
is closed. Likewise, for every fixed \(i\), the row-sum map
$
L
\longmapsto
\sum_{j=1}^dL_{ij}
$
is linear and continuous, so its zero level set
$
\left\{
L:
\sum_{j=1}^dL_{ij}=0
\right\}
$
is closed. Since \(\mathcal C_{\mathrm{gen}}\) is the intersection of
finitely many such closed sets, \(\mathcal C_{\mathrm{gen}}\) is closed.

We next verify directly that \(\mathcal C_{\mathrm{gen}}\) is a convex
cone. Let
$
L,M\in\mathcal C_{\mathrm{gen}}
$
and let
$
\alpha,\beta\ge0.
$
For every pair \(i\neq j\), membership of \(L\) and \(M\) in
\(\mathcal C_{\mathrm{gen}}\) gives
$
L_{ij}\ge0
\qquad\text{and}\qquad
M_{ij}\ge0.
$
Therefore
\[
(\alpha L+\beta M)_{ij}
=
\alpha L_{ij}+\beta M_{ij}
\ge0.
\]
Moreover,
$
L\mathbf 1=0
\qquad\text{and}\qquad
M\mathbf 1=0,
$
and hence, by linearity of matrix multiplication,
\[
\begin{aligned}
(\alpha L+\beta M)\mathbf 1
&=
\alpha L\mathbf 1+\beta M\mathbf 1\\
&=
\alpha\,0+\beta\,0\\
&=
0.
\end{aligned}
\]
Thus
$
\alpha L+\beta M
\in
\mathcal C_{\mathrm{gen}}
\qquad
\text{for every }\alpha,\beta\ge0.
$
This proves that \(\mathcal C_{\mathrm{gen}}\) is a convex cone. In
particular, taking \(\alpha=t\) and \(\beta=1-t\) for
\(t\in[0,1]\) gives
$
tL+(1-t)M\in\mathcal C_{\mathrm{gen}},
$
which verifies convexity directly.

We now identify the generator family with the intersection stated in the
proposition. By Definition~\ref{def:operators-generator-family},
$
L\in\mathcal L_{\mathfrak S}
$
if and only if all three conditions
$
L\in\mathcal A_{\mathfrak S},
\qquad
L_{ij}\ge0
\quad
\text{for every }i\neq j,
\qquad
L\mathbf 1=0
$
hold simultaneously. The last two conditions are precisely the statement
that
$
L\in\mathcal C_{\mathrm{gen}}.
$
Hence every \(L\in\mathcal L_{\mathfrak S}\) satisfies
$
L\in
\mathcal A_{\mathfrak S}
\cap
\mathcal C_{\mathrm{gen}},
$
and therefore
$
\mathcal L_{\mathfrak S}
\subseteq
\mathcal A_{\mathfrak S}
\cap
\mathcal C_{\mathrm{gen}}.
$
Conversely, suppose that
$
L\in
\mathcal A_{\mathfrak S}
\cap
\mathcal C_{\mathrm{gen}}.
$
Then
$
L\in\mathcal A_{\mathfrak S},
$
while membership in \(\mathcal C_{\mathrm{gen}}\) gives
$
L_{ij}\ge0
\qquad
\text{for every }i\neq j
$
and
$
L\mathbf 1=0.
$
These are exactly the defining conditions for
\(L\in\mathcal L_{\mathfrak S}\). Thus
$
\mathcal A_{\mathfrak S}
\cap
\mathcal C_{\mathrm{gen}}
\subseteq
\mathcal L_{\mathfrak S}.
$
Combining the two inclusions yields
$
\mathcal L_{\mathfrak S}
=
\mathcal A_{\mathfrak S}
\cap
\mathcal C_{\mathrm{gen}},
$
that is,
\[
\mathcal L_{\mathfrak S}
=
\mathcal A_{\mathfrak S}
\cap
\Bigl\{
L\in\mathbb R^{d\times d}
\;\Big|\;
L_{ij}\ge0 \text{ for }i\neq j,
\quad
L\mathbf 1=0
\Bigr\}.
\]

It remains to verify the asserted structural properties of this intersection.
By Proposition~\ref{prop:operators-interaction-family-structure},
\(\mathcal A_{\mathfrak S}\) is a closed polyhedral convex cone in
\(\mathbb R^{d\times d}\). We have just shown that
\(\mathcal C_{\mathrm{gen}}\) is also a closed polyhedral convex cone.
Therefore their intersection is closed, because the intersection of closed
sets is closed:
$
\mathcal L_{\mathfrak S}
=
\mathcal A_{\mathfrak S}
\cap
\mathcal C_{\mathrm{gen}}
$
is closed.

The cone property can likewise be checked without invoking any additional
result. Let
$
L,M\in\mathcal L_{\mathfrak S}
$
and let
$
\alpha,\beta\ge0.
$
The equality
$
\mathcal L_{\mathfrak S}
=
\mathcal A_{\mathfrak S}
\cap
\mathcal C_{\mathrm{gen}}
$
implies that
$
L,M\in\mathcal A_{\mathfrak S}
\qquad\text{and}\qquad
L,M\in\mathcal C_{\mathrm{gen}}.
$
Since both
\(\mathcal A_{\mathfrak S}\) and
\(\mathcal C_{\mathrm{gen}}\) are convex cones,
$
\alpha L+\beta M
\in\mathcal A_{\mathfrak S}
$
and simultaneously
$
\alpha L+\beta M
\in\mathcal C_{\mathrm{gen}}.
$
Consequently,
\[
\alpha L+\beta M
\in
\mathcal A_{\mathfrak S}
\cap
\mathcal C_{\mathrm{gen}}
=
\mathcal L_{\mathfrak S}.
\]
Thus \(\mathcal L_{\mathfrak S}\) is a convex cone.

Finally, we verify the polyhedral claim explicitly. Since
\(\mathcal A_{\mathfrak S}\) is a polyhedral cone, there exist finitely many
linear functionals
$
\ell_1,\ldots,\ell_m:
\mathbb R^{d\times d}\to\mathbb R
$
such that
\[
\mathcal A_{\mathfrak S}
=
\bigcap_{q=1}^m
\left\{
L\in\mathbb R^{d\times d}:
\ell_q(L)\ge0
\right\}.
\]
Any linear equalities that occur in a polyhedral representation may be
included in this form by replacing an equality
\(\ell(L)=0\) with the two inequalities
\(\ell(L)\ge0\) and \(-\ell(L)\ge0\).
Combining this finite representation with the explicit representation of
\(\mathcal C_{\mathrm{gen}}\) above gives
\[
\begin{aligned}
\mathcal L_{\mathfrak S}
&=
\bigcap_{q=1}^m
\left\{
L:
\ell_q(L)\ge0
\right\}
\\
&\qquad\cap
\bigcap_{\substack{1\le i,j\le d\\ i\neq j}}
\left\{
L:
L_{ij}\ge0
\right\}
\\
&\qquad\cap
\bigcap_{i=1}^d
\left\{
L:
\sum_{j=1}^dL_{ij}\ge0
\right\}
\\
&\qquad\cap
\bigcap_{i=1}^d
\left\{
L:
-\sum_{j=1}^dL_{ij}\ge0
\right\}.
\end{aligned}
\]
Thus \(\mathcal L_{\mathfrak S}\) is described by finitely many homogeneous
linear inequalities in \(\mathbb R^{d\times d}\), and hence is a
polyhedral cone. Together with the closedness and convexity established
above, this proves that
$
\mathcal L_{\mathfrak S}
$
is a closed polyhedral convex cone in
\(\mathbb R^{d\times d}\), with the claimed intersection representation.
\end{proof}

\subsection{Proof of Proposition~\ref{prop:operators-generator-diagonal-structure}}

\begin{proof}
Let \(L\in\mathcal L_{\mathfrak S}\) be arbitrary. By
Definition~\ref{def:operators-generator-family}, every such \(L\) satisfies
$
L\mathbf 1=0,
$
where
$
\mathbf 1=(1,\ldots,1)^\top\in\mathbb R^d.
$
Fix an arbitrary index \(i\in\{1,\ldots,d\}\). The \(i\)-th component of the
vector \(L\mathbf 1\) is
$
(L\mathbf 1)_i
=
\sum_{j=1}^d L_{ij}\mathbf 1_j.
$
Since every component of \(\mathbf 1\) is equal to \(1\), this reduces to
$
(L\mathbf 1)_i
=
\sum_{j=1}^d L_{ij}.
$
On the other hand, \(L\mathbf 1=0\), so its \(i\)-th component must vanish.
Therefore
$
\sum_{j=1}^d L_{ij}=0.
$
Separating the diagonal term \(j=i\) from the remaining terms gives
\[
L_{ii}
+
\sum_{\substack{1\le j\le d\\ j\ne i}}L_{ij}
=
0.
\]
Equivalently,
$
L_{ii}
=
-
\sum_{\substack{1\le j\le d\\ j\ne i}}L_{ij}
=
-\sum_{j\ne i}L_{ij}.
$
Since the index \(i\) was arbitrary, the identity
$
L_{ii}
=
-\sum_{j\ne i}L_{ij}
$
holds for every \(i=1,\ldots,d\).

It remains to determine the sign of the diagonal entries. Again by
Definition~\ref{def:operators-generator-family}, \(L\) is Metzler in the
sense that
$
L_{ij}\ge0
\qquad
\text{whenever }i\ne j.
$
Consequently, for each fixed \(i\),
$
\sum_{j\ne i}L_{ij}\ge0.
$
Multiplying this inequality by \(-1\) reverses its sign and yields
$
-\sum_{j\ne i}L_{ij}\le0.
$
Using the identity established above, we therefore obtain
$
L_{ii}
=
-\sum_{j\ne i}L_{ij}
\le0.
$
Thus every diagonal entry of \(L\) is nonpositive.

The same identity also proves that the diagonal is completely determined by
the off-diagonal entries. Indeed, once the collection
$
\{L_{ij}:i\ne j\}
$
is fixed, the \(i\)-th diagonal entry has no remaining freedom, because it
must satisfy
$
L_{ii}
=
-\sum_{j\ne i}L_{ij}.
$
More explicitly, suppose that \(L,\widetilde L\in\mathcal L_{\mathfrak S}\)
have identical off-diagonal entries, so that
$
L_{ij}=\widetilde L_{ij}
\qquad
\text{for every }i\ne j.
$
Applying the preceding identity to both matrices gives, for every
\(i=1,\ldots,d\),
$
L_{ii}
=
-\sum_{j\ne i}L_{ij}
$
and
$
\widetilde L_{ii}
=
-\sum_{j\ne i}\widetilde L_{ij}.
$
Because the corresponding off-diagonal entries agree,
$
\sum_{j\ne i}L_{ij}
=
\sum_{j\ne i}\widetilde L_{ij},
$
and hence
\[
L_{ii}
=
-\sum_{j\ne i}L_{ij}
=
-\sum_{j\ne i}\widetilde L_{ij}
=
\widetilde L_{ii}.
\]
Thus equality of all off-diagonal entries forces equality of all diagonal
entries. In particular, within the generator family
\(\mathcal L_{\mathfrak S}\), the diagonal entries are uniquely determined by
the off-diagonal entries, as claimed.
\end{proof}

\subsection{Proof of Proposition~\ref{prop:operators-generator-support}}

\begin{proof}
Let \(L\in\mathcal L_{\mathfrak S}\) be arbitrary, and fix two distinct
indices
$
i,j\in\{1,\ldots,d\},
\qquad
i\neq j.
$
By Definition~\ref{def:operators-generator-family}, the generator family is
defined as a subclass of the interaction family:
\[
\mathcal L_{\mathfrak S}
=
\Bigl\{
L\in\mathcal A_{\mathfrak S}
\;\Big|\;
L_{ij}\ge0 \text{ for all } i\neq j,
\quad
L\mathbf 1=0
\Bigr\}.
\]
Hence the assumption
$
L\in\mathcal L_{\mathfrak S}
$
implies, in particular, that
$
L\in\mathcal A_{\mathfrak S}.
$

Proposition~\ref{prop:operators-one-step-support} applies to every operator
in \(\mathcal A_{\mathfrak S}\). Applying that proposition to the present
matrix \(L\), we obtain, for the fixed pair \(i\neq j\),
$
(z_i,z_j)\notin\mathcal E
\quad\Longrightarrow\quad
L_{ij}=0.
$
We now prove the implication stated in the proposition. Suppose that
$
L_{ij}\neq0.
$
Assume, for contradiction, that the corresponding ordered receiver--source
pair is not admissible, namely,
$
(z_i,z_j)\notin\mathcal E.
$
Since \(L\in\mathcal A_{\mathfrak S}\) and \(i\neq j\),
Proposition~\ref{prop:operators-one-step-support} then gives
$
L_{ij}=0.
$
This contradicts the assumption
$
L_{ij}\neq0.
$
Therefore the supposition
$
(z_i,z_j)\notin\mathcal E
$
is impossible, and we must have
$
(z_i,z_j)\in\mathcal E.
$
Since \(L\in\mathcal L_{\mathfrak S}\) and the distinct indices
\(i,j\) were arbitrary, it follows that
\[
L_{ij}\neq0
\quad\Longrightarrow\quad
(z_i,z_j)\in\mathcal E
\qquad
\text{for every }i\neq j.
\]

Equivalently, the off-diagonal support of every generator satisfies
\[
\bigl\{
(z_i,z_j)\in X\times X:
i\neq j,\ L_{ij}\neq0
\bigr\}
\subseteq
\mathcal E.
\]
Thus passing from the interaction family
\(\mathcal A_{\mathfrak S}\) to the generator family
\(\mathcal L_{\mathfrak S}\) cannot introduce any new off-diagonal
one-step coupling outside the admissibility relation. The additional
generator conditions
$
L_{ij}\ge0
\qquad (i\neq j)
$
and
$
L\mathbf 1=0
$
restrict the values of the admissible entries and determine the compatible
diagonal structure, but they do not enlarge the off-diagonal support.
Consequently, every \(L\in\mathcal L_{\mathfrak S}\) already respects the
one-step admissibility pattern encoded by the geometry descriptor before
any matrix exponential is taken.
\end{proof}

\subsection{Proof of Proposition~\ref{prop:operators-pathwise-support-of-powers}}

\begin{proof}
Let \(L\in\mathcal L_{\mathfrak S}\) be fixed, and recall that
$
\bar{\mathcal E}
=
\mathcal E\cup\Delta_X,
\qquad
\Delta_X
=
\{(z_i,z_i):1\le i\le d\}.
$
Throughout this proof, the support of a matrix \(B\in\mathbb R^{d\times d}\)
is understood in the receiver--source convention as
\[
\operatorname{supp}(B)
:=
\bigl\{
(z_i,z_j)\in X\times X:
B_{ij}\neq0
\bigr\}.
\]
We also make explicit the meaning of the relational power appearing in the
statement. For a relation \(\mathcal R\subseteq X\times X\), write
$
\mathcal R^{\circ 1}:=\mathcal R,
$
and, for \(n\ge2\), let \(\mathcal R^{\circ n}\) denote its \(n\)-fold
relational composition. Under the receiver--source ordering used throughout
the paper, this means that
$
(z_i,z_j)\in\mathcal R^{\circ n}
$
if and only if there exist indices
$
a_0,a_1,\ldots,a_n\in\{1,\ldots,d\}
$
with
$
a_0=j,
\qquad
a_n=i,
$
such that
$
(z_{a_\ell},z_{a_{\ell-1}})
\in
\mathcal R
\qquad
\text{for every }\ell=1,\ldots,n.
$
Indeed, for \(n=1\) this is simply the definition of
\(\mathcal R^{\circ1}\). If the characterization holds for \(n\), then
membership of \((z_i,z_j)\) in
\(\mathcal R^{\circ(n+1)}
=
\mathcal R\circ\mathcal R^{\circ n}\)
means that there exists an intermediate site \(z_k\) such that
$
(z_i,z_k)\in\mathcal R
$
and
$
(z_k,z_j)\in\mathcal R^{\circ n}.
$
Applying the characterization to the second relation and appending the final
relation \((z_i,z_k)\in\mathcal R\) produces a chain of \(n+1\) consecutive
relations. Conversely, any such chain can be split at its final intermediate
site, giving membership in
\(\mathcal R\circ\mathcal R^{\circ n}\).
Thus the stated chain characterization holds for every \(n\ge1\).

We first observe that
$
\operatorname{supp}(L)
\subseteq
\bar{\mathcal E}.
$
To see this, let \(i,j\in\{1,\ldots,d\}\) and suppose that
$
L_{ij}\neq0.
$
If \(i=j\), then by definition
$
(z_i,z_j)
=
(z_i,z_i)
\in
\Delta_X
\subseteq
\bar{\mathcal E}.
$
If \(i\neq j\), Proposition~\ref{prop:operators-generator-support} gives
$
L_{ij}\neq0
\quad\Longrightarrow\quad
(z_i,z_j)\in\mathcal E,
$
and therefore again
$
(z_i,z_j)\in\bar{\mathcal E}.
$
Hence every nonzero entry of \(L\) is supported on
\(\bar{\mathcal E}\), which proves
$
\operatorname{supp}(L)
\subseteq
\bar{\mathcal E}
=
\bar{\mathcal E}^{\circ1}.
$

Now let \(n\ge2\) be arbitrary, and fix indices
\(i,j\in\{1,\ldots,d\}\). Repeated matrix multiplication gives the
entrywise expansion
\[
(L^n)_{ij}
=
\sum_{a_1=1}^d
\cdots
\sum_{a_{n-1}=1}^d
L_{i a_{n-1}}
L_{a_{n-1}a_{n-2}}
\cdots
L_{a_2a_1}
L_{a_1j}.
\]
Equivalently, after setting
$
a_0:=j
\qquad\text{and}\qquad
a_n:=i,
$
we may write
\[
(L^n)_{ij}
=
\sum_{a_1,\ldots,a_{n-1}=1}^d
\prod_{\ell=1}^n
L_{a_\ell a_{\ell-1}}.
\]
Suppose that
$
(L^n)_{ij}\neq0.
$
The preceding expression is a finite sum of real numbers. If every summand
were equal to zero, then the entire sum would equal zero. Since the sum is
nonzero, there must therefore exist at least one choice of intermediate
indices
$
a_1,\ldots,a_{n-1}
$
for which
$
\prod_{\ell=1}^n
L_{a_\ell a_{\ell-1}}
\neq0.
$
A finite product of real numbers is nonzero if and only if each of its
factors is nonzero. Consequently,
$
L_{a_\ell a_{\ell-1}}\neq0
\qquad
\text{for every }\ell=1,\ldots,n.
$

Fix any \(\ell\in\{1,\ldots,n\}\). If
$
a_\ell=a_{\ell-1},
$
then
\[
(z_{a_\ell},z_{a_{\ell-1}})
=
(z_{a_\ell},z_{a_\ell})
\in
\Delta_X
\subseteq
\bar{\mathcal E}.
\]
If instead
$
a_\ell\neq a_{\ell-1},
$
then
$
L_{a_\ell a_{\ell-1}}\neq0,
$
and Proposition~\ref{prop:operators-generator-support} implies
$
(z_{a_\ell},z_{a_{\ell-1}})
\in
\mathcal E
\subseteq
\bar{\mathcal E}.
$
Thus, in either case,
$
(z_{a_\ell},z_{a_{\ell-1}})
\in
\bar{\mathcal E}
\qquad
\text{for every }\ell=1,\ldots,n.
$
Together with
$
a_0=j
\qquad\text{and}\qquad
a_n=i,
$
this gives the chain
\[
z_j=z_{a_0},
\quad
z_{a_1},
\quad
\ldots,
\quad
z_{a_{n-1}},
\quad
z_{a_n}=z_i,
\]
such that every successive receiver--source pair belongs to
\(\bar{\mathcal E}\):
\[
(z_{a_1},z_{a_0}),
(z_{a_2},z_{a_1}),
\ldots,
(z_{a_n},z_{a_{n-1}})
\in
\bar{\mathcal E}.
\]
By the chain characterization of \(n\)-fold relational composition given
above, it follows that
$
(z_i,z_j)
=
(z_{a_n},z_{a_0})
\in
\bar{\mathcal E}^{\circ n}.
$
We have therefore proved that
$
(L^n)_{ij}\neq0
\quad\Longrightarrow\quad
(z_i,z_j)\in
\bar{\mathcal E}^{\circ n}.
$
Since \(i\) and \(j\) were arbitrary,
$
\operatorname{supp}(L^n)
\subseteq
\bar{\mathcal E}^{\circ n}.
$
Combining this with the already established case \(n=1\), the inclusion holds
for every integer \(n\ge1\).

It remains to prove the stated reachability consequence. Fix distinct
indices \(i\neq j\), and suppose that
$
z_j\not\rightsquigarrow z_i.
$
Assume, toward a contradiction, that there exists some \(n\ge1\) such that
$
(L^n)_{ij}\neq0.
$
By the support inclusion just established,
$
(z_i,z_j)
\in
\bar{\mathcal E}^{\circ n}.
$
Hence there exist indices
$
a_0,a_1,\ldots,a_n
$
with
$
a_0=j,
\qquad
a_n=i,
$
and
$
(z_{a_\ell},z_{a_{\ell-1}})
\in
\bar{\mathcal E}
\qquad
\text{for every }\ell=1,\ldots,n.
$
Each transition in this chain is of one of two kinds. If
$
a_\ell=a_{\ell-1},
$
then it is a diagonal holding move belonging to \(\Delta_X\). If
$
a_\ell\neq a_{\ell-1},
$
then the pair
$
(z_{a_\ell},z_{a_{\ell-1}})
$
cannot belong to \(\Delta_X\), because \(\Delta_X\) contains only diagonal
pairs. Since it belongs to
$
\bar{\mathcal E}
=
\mathcal E\cup\Delta_X,
$
it must therefore satisfy
$
(z_{a_\ell},z_{a_{\ell-1}})
\in
\mathcal E.
$
Equivalently,
$
z_{a_{\ell-1}}
\longrightarrow
z_{a_\ell}.
$

Remove from the sequence
\[
a_0,a_1,\ldots,a_n
\]
all consecutive repetitions arising from indices satisfying
\(a_\ell=a_{\ell-1}\). This produces a sequence
\[
b_0,b_1,\ldots,b_m
\]
with
$
b_0=j,
\qquad
b_m=i,
$
such that
$
b_{r-1}\neq b_r
\qquad
\text{for every }r=1,\ldots,m,
$
and every transition retained from the original chain satisfies
$
(z_{b_r},z_{b_{r-1}})
\in
\mathcal E.
$
Because \(i\neq j\), the reduced sequence cannot have length zero. Hence
$
m\ge1.
$
Therefore
\[
z_j=z_{b_0}
\longrightarrow
z_{b_1}
\longrightarrow
\cdots
\longrightarrow
z_{b_m}=z_i
\]
is a directed admissible path of positive length. By
Definition~\ref{def:prelim-reachability}, this implies
$
z_j\rightsquigarrow z_i,
$
contradicting the assumption
$
z_j\not\rightsquigarrow z_i.
$
Thus no such \(n\) can exist. Consequently,
$
(L^n)_{ij}=0
\qquad
\text{for every }n\ge1,
$
whenever \(i\neq j\) and
\(z_j\not\rightsquigarrow z_i\).
\end{proof}

\subsection{Proof of Proposition~\ref{prop:operators-semigroup-property}}

\begin{proof}
Let \(L\in\mathcal L_{\mathfrak S}\) be fixed. We prove the two identities
directly from the power-series definition of the matrix exponential. Recall
that, for every \(\tau\ge0\),
\[
e^{\tau L}
=
\sum_{n=0}^{\infty}\frac{(\tau L)^n}{n!}
=
\sum_{n=0}^{\infty}\frac{\tau^nL^n}{n!},
\]
where the zeroth power is understood as
$
L^0=I.
$

We first verify that the series manipulations used below are justified.
Fix any submultiplicative matrix norm
$
\|\cdot\|
$
on \(\mathbb R^{d\times d}\), so that
$
\|AB\|
\le
\|A\|\,\|B\|
$
for all \(A,B\in\mathbb R^{d\times d}\). By repeated application of
submultiplicativity,
$
\|L^n\|
\le
\|L\|^n
\qquad
\text{for every }n\ge0.
$
Therefore, for every \(\tau\ge0\),
\[
\sum_{n=0}^{\infty}
\left\|
\frac{\tau^nL^n}{n!}
\right\|
=
\sum_{n=0}^{\infty}
\frac{\tau^n\|L^n\|}{n!}
\le
\sum_{n=0}^{\infty}
\frac{\tau^n\|L\|^n}{n!}.
\]
The scalar series on the right is the ordinary exponential series,
$
\sum_{n=0}^{\infty}
\frac{\tau^n\|L\|^n}{n!}
=
e^{\tau\|L\|}
<
\infty.
$
Hence the matrix-exponential series converges absolutely for every
\(\tau\ge0\).

We first consider the parameter value \(\tau=0\). Using the power-series
definition,
$
e^{0\cdot L}
=
\sum_{n=0}^{\infty}
\frac{0^nL^n}{n!}.
$
The zeroth-order term is
$
\frac{0^0L^0}{0!}=I,
$
where, equivalently, one may regard the zeroth term in the matrix
exponential series as \(I\) by definition. For every \(n\ge1\),
$
(0L)^n=0,
$
and therefore all positive-order terms vanish. Consequently,
$
e^{0\cdot L}
=
I.
$

Now let
$
\tau_1,\tau_2\ge0.
$
By the power-series definition,
$
e^{\tau_1L}
=
\sum_{m=0}^{\infty}
\frac{\tau_1^mL^m}{m!}
$
and
$
e^{\tau_2L}
=
\sum_{n=0}^{\infty}
\frac{\tau_2^nL^n}{n!}.
$
Before multiplying these two series, observe that the corresponding double
series is absolutely convergent. Indeed,
\[
\begin{aligned}
\sum_{m=0}^{\infty}
\sum_{n=0}^{\infty}
\left\|
\frac{\tau_1^mL^m}{m!}
\frac{\tau_2^nL^n}{n!}
\right\|
&\le
\sum_{m=0}^{\infty}
\sum_{n=0}^{\infty}
\frac{\tau_1^m\tau_2^n}{m!n!}
\|L^m\|\,\|L^n\| \\
&\le
\sum_{m=0}^{\infty}
\sum_{n=0}^{\infty}
\frac{\tau_1^m\tau_2^n}{m!n!}
\|L\|^{m+n}.
\end{aligned}
\]
Since all terms in the last expression are nonnegative, the double sum
factors as
\[
\begin{aligned}
\sum_{m=0}^{\infty}
\sum_{n=0}^{\infty}
\frac{\tau_1^m\tau_2^n}{m!n!}
\|L\|^{m+n}
&=
\left(
\sum_{m=0}^{\infty}
\frac{(\tau_1\|L\|)^m}{m!}
\right)
\left(
\sum_{n=0}^{\infty}
\frac{(\tau_2\|L\|)^n}{n!}
\right) \\
&=
e^{\tau_1\|L\|}
e^{\tau_2\|L\|} \\
&=
e^{(\tau_1+\tau_2)\|L\|} \\
&<
\infty.
\end{aligned}
\]
Thus the product of the two matrix-exponential series may be expanded by the
Cauchy product, and the resulting absolutely convergent double series may be
rearranged according to the total degree \(m+n\).

We therefore obtain
\[
\begin{aligned}
e^{\tau_1L}e^{\tau_2L}
&=
\left(
\sum_{m=0}^{\infty}
\frac{\tau_1^mL^m}{m!}
\right)
\left(
\sum_{n=0}^{\infty}
\frac{\tau_2^nL^n}{n!}
\right) \\
&=
\sum_{m=0}^{\infty}
\sum_{n=0}^{\infty}
\frac{\tau_1^m\tau_2^n}{m!n!}
L^mL^n.
\end{aligned}
\]
Because both factors are powers of the same fixed matrix \(L\),
$
L^mL^n
=
L^{m+n}
\qquad
\text{for every }m,n\ge0.
$
Hence
\[
e^{\tau_1L}e^{\tau_2L}
=
\sum_{m=0}^{\infty}
\sum_{n=0}^{\infty}
\frac{\tau_1^m\tau_2^n}{m!n!}
L^{m+n}.
\]
We now group together all terms having the same total degree
$
k=m+n.
$
For a fixed \(k\ge0\), the possible pairs \((m,n)\) are
$
(m,n)
=
(0,k),(1,k-1),\ldots,(k,0).
$
Therefore,
\[
e^{\tau_1L}e^{\tau_2L}
=
\sum_{k=0}^{\infty}
\left(
\sum_{m=0}^{k}
\frac{\tau_1^m\tau_2^{k-m}}
{m!(k-m)!}
\right)
L^k.
\]
For each fixed \(k\), use
$
\binom{k}{m}
=
\frac{k!}{m!(k-m)!}.
$
It follows that
$
\frac{1}{m!(k-m)!}
=
\frac{1}{k!}\binom{k}{m},
$
and hence
\[
\begin{aligned}
\sum_{m=0}^{k}
\frac{\tau_1^m\tau_2^{k-m}}
{m!(k-m)!}
&=
\frac{1}{k!}
\sum_{m=0}^{k}
\binom{k}{m}
\tau_1^m\tau_2^{k-m}.
\end{aligned}
\]
By the scalar binomial theorem,
$
\sum_{m=0}^{k}
\binom{k}{m}
\tau_1^m\tau_2^{k-m}
=
(\tau_1+\tau_2)^k.
$
Consequently,
$
\sum_{m=0}^{k}
\frac{\tau_1^m\tau_2^{k-m}}
{m!(k-m)!}
=
\frac{(\tau_1+\tau_2)^k}{k!}.
$
Substituting this coefficient identity into the previous expansion gives
\[
\begin{aligned}
e^{\tau_1L}e^{\tau_2L}
&=
\sum_{k=0}^{\infty}
\frac{(\tau_1+\tau_2)^k}{k!}L^k \\
&=
\sum_{k=0}^{\infty}
\frac{\bigl((\tau_1+\tau_2)L\bigr)^k}{k!} \\
&=
e^{(\tau_1+\tau_2)L}.
\end{aligned}
\]
Thus, for all
$
\tau_1,\tau_2\ge0,
$
we have established
$
e^{(\tau_1+\tau_2)L}
=
e^{\tau_1L}e^{\tau_2L}.
$

Since
$
\tau_1+\tau_2\ge0
$
whenever
$
\tau_1,\tau_2\ge0,
$
the product of any two members of the family
$
\{e^{\tau L}\}_{\tau\ge0}
$
is again a member of the same family. Matrix multiplication is associative,
and the member corresponding to \(\tau=0\) is the identity matrix \(I\).
Therefore
$
\{e^{\tau L}\}_{\tau\ge0}
$
forms a one-parameter semigroup, with
$
e^{0\cdot L}=I
$
and
$
e^{(\tau_1+\tau_2)L}
=
e^{\tau_1L}e^{\tau_2L}
\qquad
\text{for all }\tau_1,\tau_2\ge0.
$
\end{proof}

\subsection{Proof of Proposition~\ref{prop:operators-depth-stable-geometry}}

\begin{proof}
Let
$
A^{(1)},\ldots,A^{(m)}
\in
\mathcal A_{\mathfrak S},
$
and define
$
B^{(\ell)}
:=
I+A^{(\ell)},
\qquad
\ell=1,\ldots,m.
$
We use the site-indexed support convention
\[
\operatorname{supp}(B)
:=
\bigl\{
(z_i,z_j)\in X\times X:
B_{ij}\neq0
\bigr\}.
\]
Recall also that
$
\bar{\mathcal E}
=
\mathcal E\cup\Delta_X,
\qquad
\Delta_X
=
\{(z_i,z_i):1\le i\le d\}.
$
We first verify that each individual residual factor
\(B^{(\ell)}\) is supported on \(\bar{\mathcal E}\). Fix
\(\ell\in\{1,\ldots,m\}\), and suppose that
$
B^{(\ell)}_{ij}\neq0
$
for some \(i,j\in\{1,\ldots,d\}\). If \(i=j\), then
\[
(z_i,z_j)
=
(z_i,z_i)
\in
\Delta_X
\subseteq
\bar{\mathcal E}.
\]
If \(i\neq j\), then the corresponding off-diagonal entry of the identity
matrix vanishes:
$
I_{ij}=0.
$
Consequently,
\[
B^{(\ell)}_{ij}
=
(I+A^{(\ell)})_{ij}
=
I_{ij}+A^{(\ell)}_{ij}
=
A^{(\ell)}_{ij}.
\]
Since \(B^{(\ell)}_{ij}\neq0\), we obtain
$
A^{(\ell)}_{ij}\neq0.
$
Because
$
A^{(\ell)}
\in
\mathcal A_{\mathfrak S},
$
Proposition~\ref{prop:operators-one-step-support} gives
$
(z_i,z_j)\notin\mathcal E
\quad\Longrightarrow\quad
A^{(\ell)}_{ij}=0.
$
Taking the contrapositive yields
$
A^{(\ell)}_{ij}\neq0
\quad\Longrightarrow\quad
(z_i,z_j)\in\mathcal E.
$
Hence, in the present case,
$
(z_i,z_j)
\in
\mathcal E
\subseteq
\bar{\mathcal E}.
$
Thus both the diagonal and off-diagonal cases give
$
B^{(\ell)}_{ij}\neq0
\quad\Longrightarrow\quad
(z_i,z_j)\in\bar{\mathcal E}.
$
Since \(i,j\) and \(\ell\) were arbitrary,
$
\operatorname{supp}(B^{(\ell)})
\subseteq
\bar{\mathcal E}
\qquad
\text{for every }\ell=1,\ldots,m.
$

We now consider the complete depth-\(m\) residual product
\[
P_m
:=
B^{(m)}B^{(m-1)}\cdots B^{(1)}
=
(I+A^{(m)})\cdots(I+A^{(1)}).
\]
For \(m=1\), the preceding support inclusion immediately gives
$
\operatorname{supp}(P_1)
=
\operatorname{supp}(B^{(1)})
\subseteq
\bar{\mathcal E}
=
\bar{\mathcal E}^{\circ1}.
$
Suppose now that \(m\ge2\), and fix arbitrary indices
$
i,j\in\{1,\ldots,d\}.
$
Repeated application of the definition of matrix multiplication gives
\[
(P_m)_{ij}
=
\sum_{a_{m-1}=1}^d
\sum_{a_{m-2}=1}^d
\cdots
\sum_{a_1=1}^d
B^{(m)}_{i a_{m-1}}
B^{(m-1)}_{a_{m-1}a_{m-2}}
\cdots
B^{(2)}_{a_2a_1}
B^{(1)}_{a_1j}.
\]
Introduce the endpoint notation
$
a_0:=j,
\qquad
a_m:=i.
$
Then the preceding expression can be written more compactly as
\[
(P_m)_{ij}
=
\sum_{a_1,\ldots,a_{m-1}=1}^d
\prod_{\ell=1}^m
B^{(\ell)}_{a_\ell a_{\ell-1}}.
\]

Assume that
$
(P_m)_{ij}\neq0.
$
The right-hand side is a finite sum. If every summand in this finite sum
were zero, then the sum itself would be zero. Since the sum is nonzero,
there must exist at least one choice of intermediate indices
$
a_1,\ldots,a_{m-1}
\in
\{1,\ldots,d\}
$
such that
$
\prod_{\ell=1}^m
B^{(\ell)}_{a_\ell a_{\ell-1}}
\neq0.
$
A finite product of real numbers can be nonzero only if every factor in the
product is nonzero. Therefore
$
B^{(\ell)}_{a_\ell a_{\ell-1}}
\neq0
\qquad
\text{for every }\ell=1,\ldots,m.
$
Applying the single-factor support property proved above to each
\(B^{(\ell)}\) gives
$
(z_{a_\ell},z_{a_{\ell-1}})
\in
\bar{\mathcal E}
\qquad
\text{for every }\ell=1,\ldots,m.
$
Thus we have obtained the sequence
\[
a_0=j,\;
a_1,\;
\ldots,\;
a_{m-1},\;
a_m=i
\]
for which
\[
(z_{a_1},z_{a_0}),
\,
(z_{a_2},z_{a_1}),
\,
\ldots,
\,
(z_{a_m},z_{a_{m-1}})
\in
\bar{\mathcal E}.
\]
Under the standing receiver--source convention, this is precisely a chain of
\(m\) consecutive relations in \(\bar{\mathcal E}\), beginning at the
source \(z_j=z_{a_0}\) and ending at the receiver
\(z_i=z_{a_m}\). By the definition of the \(m\)-fold relational
composition,
$
(z_{a_m},z_{a_0})
\in
\bar{\mathcal E}^{\circ m}.
$
Since
$
a_m=i
\qquad\text{and}\qquad
a_0=j,
$
we conclude that
$
(z_i,z_j)
\in
\bar{\mathcal E}^{\circ m}.
$
We have therefore proved the implication
$
(P_m)_{ij}\neq0
\quad\Longrightarrow\quad
(z_i,z_j)
\in
\bar{\mathcal E}^{\circ m}.
$
Because \(i\) and \(j\) were arbitrary,
$
\operatorname{supp}(P_m)
\subseteq
\bar{\mathcal E}^{\circ m}.
$
Substituting the definition of \(P_m\) yields
\[
\operatorname{supp}\!\left(
(I+A^{(m)})\cdots(I+A^{(1)})
\right)
\subseteq
\bar{\mathcal E}^{\circ m}.
\]

Finally, the role of the diagonal relation in
\(\bar{\mathcal E}=\mathcal E\cup\Delta_X\) can be made explicit.
For each layer \(\ell\), a nonzero factor
$
B^{(\ell)}_{a_\ell a_{\ell-1}}
$
with
$
a_\ell=a_{\ell-1}
$
corresponds to a diagonal self-action and therefore to a holding move in
\(\Delta_X\). Whenever
$
a_\ell\neq a_{\ell-1},
$
the identity matrix contributes nothing to that entry, and the nonzero
factor must arise from \(A^{(\ell)}\); hence the corresponding transition
belongs to \(\mathcal E\). Therefore every nonzero contribution to the
depth-\(m\) product is obtained by concatenating admissible one-step
interactions, possibly interspersed with diagonal holding moves. Removing
the holding moves leaves a directed admissible path containing at most
\(m\) genuine cross-site interactions. Thus increasing depth can enlarge
the receptive field only through successive compositions of the
one-step geometry already encoded by \(\mathcal E\), and cannot introduce
a cross-site interaction unsupported by such an admissible concatenation.
\end{proof}

\subsection{Proof of Proposition~\ref{prop:operators-nontrivial-generator-feasibility}}

\begin{proof}
Let \(r_0\in\mathcal R_+\) satisfy the hypothesis of the proposition, so that
there exist distinct indices \(i_0,j_0\in\{1,\ldots,d\}\) for which
$
(\Phi_{r_0})_{i_0j_0}>0.
$
Consider the matrix
$
L^{(r_0)}
:=
\Phi_{r_0}
-
\operatorname{Diag}(\Phi_{r_0}\mathbf1).
$
We verify directly that this matrix belongs to
\(\mathcal L_{\mathfrak S}\) and is nonzero.

We first check that \(L^{(r_0)}\) belongs to the interaction family
\(\mathcal A_{\mathfrak S}\). Define a coefficient vector
\(\theta^{(r_0)}\in\mathbb R^R\) by
\[
\theta^{(r_0)}_r
:=
\begin{cases}
1, & r=r_0,\\
0, & r\neq r_0,
\end{cases}
\]
and define
$
\delta^{(r_0)}
:=
-\Phi_{r_0}\mathbf1
\in\mathbb R^d.
$
For every \(r\in\mathcal R_+\), one has
$
\theta^{(r_0)}_r\ge0.
$
Indeed, if \(r=r_0\), then
$
\theta^{(r_0)}_{r_0}=1>0,
$
whereas if \(r\neq r_0\), then
$
\theta^{(r_0)}_r=0.
$
Thus the sign constraints required in
Definition~\ref{def:operators-interaction-family} are satisfied. Moreover,
\[
\begin{aligned}
\sum_{r=1}^R
\theta^{(r_0)}_r\Phi_r
+
\operatorname{Diag}(\delta^{(r_0)})
&=
1\cdot\Phi_{r_0}
+
\sum_{\substack{r=1\\r\neq r_0}}^R
0\cdot\Phi_r
+
\operatorname{Diag}(-\Phi_{r_0}\mathbf1)
\\
&=
\Phi_{r_0}
-
\operatorname{Diag}(\Phi_{r_0}\mathbf1)
\\
&=
L^{(r_0)}.
\end{aligned}
\]
Therefore,
$
L^{(r_0)}
\in
\mathcal A_{\mathfrak S}.
$

We next verify the off-diagonal nonnegativity required for membership in the
generator family. Let \(i,j\in\{1,\ldots,d\}\) with \(i\neq j\). Since
\(\operatorname{Diag}(\Phi_{r_0}\mathbf1)\) is a diagonal matrix,
$
\bigl(
\operatorname{Diag}(\Phi_{r_0}\mathbf1)
\bigr)_{ij}
=
0
\qquad
\text{whenever }i\neq j.
$
Consequently,
\[
\begin{aligned}
L^{(r_0)}_{ij}
&=
(\Phi_{r_0})_{ij}
-
\bigl(
\operatorname{Diag}(\Phi_{r_0}\mathbf1)
\bigr)_{ij}
\\
&=
(\Phi_{r_0})_{ij}.
\end{aligned}
\]
Because \(r_0\in\mathcal R_+\), the definition of the positive channel set in
Definition~\ref{def:prelim-geometry-descriptor} gives
$
(\Phi_{r_0})_{ij}\ge0
\qquad
\text{for every }i\neq j.
$
Hence
$
L^{(r_0)}_{ij}\ge0
\qquad
\text{for every }i\neq j.
$

It remains to verify the zero-row-sum condition. Set, temporarily,
$
v:=\Phi_{r_0}\mathbf1\in\mathbb R^d.
$
By definition of the diagonal matrix associated with \(v\),
\[
\operatorname{Diag}(v)_{ij}
=
\begin{cases}
v_i, & i=j,\\
0, & i\neq j.
\end{cases}
\]
Therefore, for each \(i\in\{1,\ldots,d\}\),
\[
\begin{aligned}
\bigl(\operatorname{Diag}(v)\mathbf1\bigr)_i
&=
\sum_{j=1}^d
\operatorname{Diag}(v)_{ij}\mathbf1_j
\\
&=
\sum_{j=1}^d
\operatorname{Diag}(v)_{ij}
\\
&=
\operatorname{Diag}(v)_{ii}
\\
&=
v_i.
\end{aligned}
\]
Since this holds for every coordinate \(i\),
$
\operatorname{Diag}(v)\mathbf1=v.
$
Substituting
$
v=\Phi_{r_0}\mathbf1
$
gives
$
\operatorname{Diag}(\Phi_{r_0}\mathbf1)\mathbf1
=
\Phi_{r_0}\mathbf1.
$
It follows that
\[
\begin{aligned}
L^{(r_0)}\mathbf1
&=
\left(
\Phi_{r_0}
-
\operatorname{Diag}(\Phi_{r_0}\mathbf1)
\right)\mathbf1
\\
&=
\Phi_{r_0}\mathbf1
-
\operatorname{Diag}(\Phi_{r_0}\mathbf1)\mathbf1
\\
&=
\Phi_{r_0}\mathbf1
-
\Phi_{r_0}\mathbf1
\\
&=
0.
\end{aligned}
\]
Equivalently, the same calculation can be seen row by row. For every
\(i\in\{1,\ldots,d\}\),
\[
\begin{aligned}
\sum_{j=1}^d L^{(r_0)}_{ij}
&=
\sum_{j=1}^d(\Phi_{r_0})_{ij}
-
(\Phi_{r_0}\mathbf1)_i
\\
&=
\sum_{j=1}^d(\Phi_{r_0})_{ij}
-
\sum_{j=1}^d(\Phi_{r_0})_{ij}
\\
&=
0.
\end{aligned}
\]
Thus \(L^{(r_0)}\) is constant preserving.

We have now established all defining properties of the generator family:
$
L^{(r_0)}\in\mathcal A_{\mathfrak S},
$
$
L^{(r_0)}_{ij}\ge0
\qquad
\text{for every }i\neq j,
$
and
$
L^{(r_0)}\mathbf1=0.
$
Therefore, by Definition~\ref{def:operators-generator-family},
$
L^{(r_0)}
\in
\mathcal L_{\mathfrak S}.
$

For completeness, the diagonal entries generated by this construction are
also consistent with the generator structure established previously. Indeed,
for each \(i\),
\[
\begin{aligned}
L^{(r_0)}_{ii}
&=
(\Phi_{r_0})_{ii}
-
(\Phi_{r_0}\mathbf1)_i
\\
&=
(\Phi_{r_0})_{ii}
-
\sum_{j=1}^d(\Phi_{r_0})_{ij}
\\
&=
-\sum_{j\neq i}(\Phi_{r_0})_{ij}.
\end{aligned}
\]
Since \(r_0\in\mathcal R_+\),
$
(\Phi_{r_0})_{ij}\ge0
\qquad
\text{for }j\neq i,
$
and hence
$
L^{(r_0)}_{ii}
=
-\sum_{j\neq i}(\Phi_{r_0})_{ij}
\le0,
$
in agreement with
Proposition~\ref{prop:operators-generator-diagonal-structure}.

It remains only to show that the constructed generator is nonzero. By
hypothesis, there exist \(i_0\neq j_0\) such that
$
(\Phi_{r_0})_{i_0j_0}>0.
$
Since the diagonal correction
\(\operatorname{Diag}(\Phi_{r_0}\mathbf1)\) has zero off-diagonal entries,
\[
\begin{aligned}
L^{(r_0)}_{i_0j_0}
&=
(\Phi_{r_0})_{i_0j_0}
-
\bigl(
\operatorname{Diag}(\Phi_{r_0}\mathbf1)
\bigr)_{i_0j_0}
\\
&=
(\Phi_{r_0})_{i_0j_0}
\\
&>
0.
\end{aligned}
\]
Thus \(L^{(r_0)}\) contains at least one strictly positive entry and therefore
cannot be the zero matrix:
$
L^{(r_0)}\neq0.
$
Combining this fact with
\(L^{(r_0)}\in\mathcal L_{\mathfrak S}\) yields
$
L^{(r_0)}
\in
\mathcal L_{\mathfrak S}\setminus\{0\},
$
as claimed.
\end{proof}

\subsection{Proof of Proposition~\ref{prop:operators-feasible-compilation}}

\begin{proof}
Fix
$
U\in\mathbb R^{d\times p},
\qquad
V\in\mathbb R^{d\times q}.
$
Recall that the constrained generator family is
\[
\mathcal L_{\mathfrak S;U,V}
=
\left\{
L\in\mathcal L_{\mathfrak S}
\;\middle|\;
LU=0,\quad V^\top L=0
\right\}.
\]
We first verify its geometric structure. Consider the linear maps
$
\mathcal T_U:\mathbb R^{d\times d}\to\mathbb R^{d\times p},
\qquad
\mathcal T_U(L):=LU,
$
and
$
\mathcal S_V:\mathbb R^{d\times d}\to\mathbb R^{q\times d},
\qquad
\mathcal S_V(L):=V^\top L.
$
Both maps are linear. Indeed, for arbitrary
\(L_1,L_2\in\mathbb R^{d\times d}\) and
\(a,b\in\mathbb R\),
\[
\mathcal T_U(aL_1+bL_2)
=
(aL_1+bL_2)U
=
aL_1U+bL_2U
=
a\mathcal T_U(L_1)+b\mathcal T_U(L_2),
\]
and similarly,
\[
\mathcal S_V(aL_1+bL_2)
=
V^\top(aL_1+bL_2)
=
aV^\top L_1+bV^\top L_2
=
a\mathcal S_V(L_1)+b\mathcal S_V(L_2).
\]
Hence
$
\{L:LU=0\}
=
\ker(\mathcal T_U)
$
and
$
\{L:V^\top L=0\}
=
\ker(\mathcal S_V)
$
are linear subspaces of \(\mathbb R^{d\times d}\). In particular, they are
closed convex cones.

More explicitly, the condition \(LU=0\) consists of the finitely many
homogeneous linear equations
$
\sum_{k=1}^d L_{ik}U_{ka}=0,
\qquad
i=1,\ldots,d,\quad a=1,\ldots,p,
$
while \(V^\top L=0\) consists of the finitely many homogeneous linear
equations
$
\sum_{k=1}^d V_{kb}L_{kj}=0,
\qquad
b=1,\ldots,q,\quad j=1,\ldots,d.
$
By Proposition~\ref{prop:operators-generator-family-structure},
\(\mathcal L_{\mathfrak S}\) is already a closed polyhedral convex cone.
Therefore
\[
\mathcal L_{\mathfrak S;U,V}
=
\mathcal L_{\mathfrak S}
\cap
\ker(\mathcal T_U)
\cap
\ker(\mathcal S_V)
\]
is obtained from a polyhedral cone by imposing only finitely many additional
homogeneous linear equalities. It follows that
$
\mathcal L_{\mathfrak S;U,V}
$
is again a polyhedral cone. Since it is an intersection of closed sets, it is
closed; and since each of the three sets in the intersection is convex, it is
convex. Thus
$
\mathcal L_{\mathfrak S;U,V}
$
is a closed polyhedral convex cone.

We next establish the basic positivity property of the normalization
functional on this cone. Let
$
L\in\mathcal L_{\mathfrak S;U,V}.
$
Since
$
\mathcal L_{\mathfrak S;U,V}
\subseteq
\mathcal L_{\mathfrak S},
$
Definition~\ref{def:operators-normalized-generator} gives
$
\rho(L)
=
\frac1d
\sum_{i=1}^d\sum_{j\ne i}L_{ij}.
$
Every generator satisfies
$
L_{ij}\ge0
\qquad
\text{for }i\neq j,
$
so every term appearing in the sum defining \(\rho(L)\) is nonnegative.
Consequently,
$
\rho(L)\ge0.
$
We claim that equality can occur only when \(L=0\). Suppose that
$
\rho(L)=0.
$
Then
\[
0
=
d\,\rho(L)
=
\sum_{i=1}^d\sum_{j\ne i}L_{ij}.
\]
The right-hand side is a finite sum of nonnegative real numbers. A finite
sum of nonnegative numbers can be zero only if every summand is zero.
Therefore
$
L_{ij}=0
\qquad
\text{for every }i\neq j.
$
By Proposition~\ref{prop:operators-generator-diagonal-structure}, every
generator also satisfies
$
L_{ii}
=
-\sum_{j\ne i}L_{ij}.
$
Since all off-diagonal entries have just been shown to vanish,
$
L_{ii}
=
-\sum_{j\ne i}0
=
0
\qquad
\text{for every }i.
$
Thus every entry of \(L\) is zero, and hence
$
L=0.
$
We have therefore proved
$
L\in\mathcal L_{\mathfrak S;U,V}\setminus\{0\}
\quad\Longrightarrow\quad
\rho(L)>0.
$

We can now characterize nonemptiness of the normalized slice. Suppose first
that
$
\mathcal L_{\mathfrak S;U,V}^{(1)}
\neq\varnothing.
$
Then there exists some
$
L\in\mathcal L_{\mathfrak S;U,V}
$
such that
$
\rho(L)=1.
$
In particular,
$
L\neq0,
$
because the zero matrix satisfies
$
\rho(0)=0.
$
Hence
$
\mathcal L_{\mathfrak S;U,V}\neq\{0\}.
$

Conversely, suppose that
$
\mathcal L_{\mathfrak S;U,V}\neq\{0\}.
$
Choose
$
Q\in\mathcal L_{\mathfrak S;U,V}\setminus\{0\}.
$
By the positivity result above,
$
\rho(Q)>0.
$
Because \(\mathcal L_{\mathfrak S;U,V}\) is a cone and
\(1/\rho(Q)>0\),
$
\widehat Q
:=
\frac{Q}{\rho(Q)}
\in
\mathcal L_{\mathfrak S;U,V}.
$
Moreover, the definition of \(\rho\) immediately gives positive
homogeneity:
$
\rho(cL)
=
c\,\rho(L)
\qquad
\text{for every }c\ge0
\text{ and }L\in\mathcal L_{\mathfrak S}.
$
Indeed,
\[
\begin{aligned}
\rho(cL)
&=
\frac1d
\sum_{i=1}^d\sum_{j\ne i}(cL_{ij})
\\
&=
c\,
\frac1d
\sum_{i=1}^d\sum_{j\ne i}L_{ij}
\\
&=
c\,\rho(L).
\end{aligned}
\]
Therefore
$
\rho(\widehat Q)
=
\rho\!\left(\frac{Q}{\rho(Q)}\right)
=
\frac{1}{\rho(Q)}\rho(Q)
=
1.
$
Thus
$
\widehat Q
\in
\mathcal L_{\mathfrak S;U,V}^{(1)},
$
and hence
$
\mathcal L_{\mathfrak S;U,V}^{(1)}
\neq\varnothing.
$
We have proved the equivalence
\[
\mathcal L_{\mathfrak S;U,V}^{(1)}
\neq\varnothing
\quad\Longleftrightarrow\quad
\mathcal L_{\mathfrak S;U,V}\neq\{0\}.
\]

Assume from now on that this normalized slice is nonempty. Since
\(\mathcal L_{\mathfrak S;U,V}\) is a polyhedral set and the condition
$
\rho(L)=1
$
is an affine linear equality, the set
\[
\mathcal L_{\mathfrak S;U,V}^{(1)}
=
\mathcal L_{\mathfrak S;U,V}
\cap
\{L:\rho(L)=1\}
\]
is a polyhedron. It is also closed, because both
\(\mathcal L_{\mathfrak S;U,V}\) and the affine hyperplane
$
\{L:\rho(L)=1\}
$
are closed. To conclude compactness, it remains to prove boundedness.

Let
$
L\in\mathcal L_{\mathfrak S;U,V}^{(1)}.
$
The normalization condition gives
$
1
=
\rho(L)
=
\frac1d
\sum_{i=1}^d\sum_{j\ne i}L_{ij}.
$
Multiplying by \(d\),
$
\sum_{i=1}^d\sum_{j\ne i}L_{ij}
=
d.
$
Every summand on the left is nonnegative. Hence no individual summand can
exceed the total sum. Therefore, for every \(i\neq j\),
$
0
\le
L_{ij}
\le
d.
$
The diagonal entries are controlled by the zero-row-sum condition. By
Proposition~\ref{prop:operators-generator-diagonal-structure},
$
L_{ii}
=
-\sum_{j\ne i}L_{ij}.
$
Since the terms in this sum are nonnegative,
$
L_{ii}\le0.
$
Moreover,
$
\sum_{j\ne i}L_{ij}
\le
\sum_{k=1}^d\sum_{\ell\ne k}L_{k\ell}
=
d,
$
and therefore
$
-d
\le
L_{ii}
\le
0.
$
Combining the diagonal and off-diagonal estimates gives
$
|L_{ij}|
\le
d
\qquad
\text{for every }i,j.
$
Thus, for example, the Frobenius norm satisfies
\[
\begin{aligned}
\|L\|_F^2
&=
\sum_{i=1}^d\sum_{j=1}^d L_{ij}^2
\\
&\le
\sum_{i=1}^d\sum_{j=1}^d d^2
\\
&=
d^4,
\end{aligned}
\]
and hence
$
\|L\|_F\le d^2.
$
This bound is uniform over
\(\mathcal L_{\mathfrak S;U,V}^{(1)}\). Therefore the normalized slice is
bounded.

Since \(\mathbb R^{d\times d}\) is finite dimensional, closedness and
boundedness imply compactness. Hence
$
\mathcal L_{\mathfrak S;U,V}^{(1)}
$
is a nonempty compact polyhedron, and therefore a polytope.

A bounded polyhedron in a finite-dimensional vector space has only finitely
many vertices and is equal to the convex hull of those vertices. Denote the
vertices of
\(\mathcal L_{\mathfrak S;U,V}^{(1)}\) by
$
Z_1,\ldots,Z_N.
$
Then
$
Z_s
\in
\mathcal L_{\mathfrak S;U,V}^{(1)}
\qquad
\text{for every }s=1,\ldots,N,
$
and
$
\mathcal L_{\mathfrak S;U,V}^{(1)}
=
\operatorname{conv}\{Z_1,\ldots,Z_N\}.
$
Equivalently, every
$
Z\in
\mathcal L_{\mathfrak S;U,V}^{(1)}
$
admits coefficients
$
\alpha_1,\ldots,\alpha_N\ge0,
\qquad
\sum_{s=1}^N\alpha_s=1,
$
such that
$
Z
=
\sum_{s=1}^N\alpha_s Z_s.
$

Now let
$
Q\in
\mathcal L_{\mathfrak S;U,V}\setminus\{0\}
$
be arbitrary. As proved above,
$
\rho(Q)>0.
$
Hence its normalized version
$
\widehat Q
:=
\frac{Q}{\rho(Q)}
$
belongs to the normalized polytope:
$
\widehat Q
\in
\mathcal L_{\mathfrak S;U,V}^{(1)}.
$
Therefore there exists
\[
\alpha=(\alpha_1,\ldots,\alpha_N)^\top
\in\Delta_N
\]
such that
$
\widehat Q
=
\sum_{s=1}^N\alpha_sZ_s.
$
Multiplying both sides by \(\rho(Q)\) gives
\[
Q
=
\rho(Q)\widehat Q
=
\rho(Q)
\sum_{s=1}^N\alpha_sZ_s,
\]
which is the required finite compilation.

It remains to establish the stated uniqueness properties. The number
\(\rho(Q)\) is determined directly by \(Q\) through
Definition~\ref{def:operators-normalized-generator}, so its value is unique.
More intrinsically, suppose that the same nonzero \(Q\) admits a
rate--shape decomposition
$
Q=tZ,
$
where
$
t>0,
\qquad
Z\in\mathcal L_{\mathfrak S;U,V}^{(1)}.
$
Since
$
\rho(Z)=1,
$
positive homogeneity of \(\rho\) gives
$
\rho(Q)
=
\rho(tZ)
=
t\rho(Z)
=
t.
$
Thus necessarily
$
t=\rho(Q).
$
Substituting this identity into
$
Q=tZ
$
gives
$
Z
=
\frac{Q}{t}
=
\frac{Q}{\rho(Q)}.
$
Therefore both the scale
$
\rho(Q)
$
and the normalized matrix
$
\frac{Q}{\rho(Q)}
$
are uniquely determined by \(Q\).

In contrast, uniqueness of the coefficient vector
\(\alpha\in\Delta_N\) is not implied by the hypotheses. Indeed, uniqueness
of barycentric coordinates with respect to the entire vertex set would
require the vertices
$
Z_1,\ldots,Z_N
$
to be affinely independent. No such assumption is imposed here. To see
explicitly what can happen, suppose that the vertices are affinely
dependent. Then there exists a nonzero vector
$
c=(c_1,\ldots,c_N)^\top\in\mathbb R^N
$
such that
$
\sum_{s=1}^N c_s=0
$
and
$
\sum_{s=1}^N c_sZ_s=0.
$
Consider the strictly positive simplex point
$
\bar\alpha_s:=\frac1N,
\qquad
s=1,\ldots,N.
$
Because every \(\bar\alpha_s>0\), for sufficiently small
\(\varepsilon>0\) both vectors
$
\alpha^{+}
:=
\bar\alpha+\varepsilon c
$
and
$
\alpha^{-}
:=
\bar\alpha-\varepsilon c
$
remain componentwise nonnegative. Moreover,
\[
\sum_{s=1}^N\alpha_s^{+}
=
\sum_{s=1}^N\bar\alpha_s
+
\varepsilon\sum_{s=1}^Nc_s
=
1,
\]
and similarly
$
\sum_{s=1}^N\alpha_s^{-}=1.
$
Thus
$
\alpha^{+},\alpha^{-}\in\Delta_N.
$
Since \(c\neq0\),
$
\alpha^{+}\neq\alpha^{-},
$
while
\[
\begin{aligned}
\sum_{s=1}^N\alpha_s^{+}Z_s
&=
\sum_{s=1}^N\bar\alpha_sZ_s
+
\varepsilon\sum_{s=1}^Nc_sZ_s
\\
&=
\sum_{s=1}^N\bar\alpha_sZ_s,
\end{aligned}
\]
and likewise
$
\sum_{s=1}^N\alpha_s^{-}Z_s
=
\sum_{s=1}^N\bar\alpha_sZ_s.
$
Hence distinct simplex coefficients can represent the same normalized
generator whenever the vertex set is affinely dependent. Therefore the
vertex mixture coefficients are guaranteed to exist but need not be unique,
whereas the rate \(\rho(Q)\) and normalized generator
\(Q/\rho(Q)\) are unique.
\end{proof}

\subsection{Proof of Proposition~\ref{prop:operators-block-valued-interaction-structure}}

\begin{proof}
Let \(F_{ab}\in\mathbb R^{c\times c}\), \(1\le a,b\le c\), denote the
standard matrix unit whose \((a,b)\)-entry is equal to \(1\) and whose
remaining entries are equal to \(0\). Then
$
\{F_{ab}:1\le a,b\le c\}
$
is a basis of \(\mathbb R^{c\times c}\). In particular, every block
\(D_i\in\mathbb R^{c\times c}\) appearing in
Definition~\ref{def:operators-block-valued-interaction-family} admits the
expansion
$
D_i
=
\sum_{a=1}^c\sum_{b=1}^c
d_{i,ab}F_{ab}
$
for uniquely determined coefficients
$
d_{i,ab}\in\mathbb R.
$
Consequently,
\[
\begin{aligned}
\sum_{i=1}^d E_{ii}\otimes D_i
&=
\sum_{i=1}^d
E_{ii}\otimes
\left(
\sum_{a=1}^c\sum_{b=1}^c
d_{i,ab}F_{ab}
\right)\\
&=
\sum_{i=1}^d
\sum_{a=1}^c
\sum_{b=1}^c
d_{i,ab}
\bigl(E_{ii}\otimes F_{ab}\bigr),
\end{aligned}
\]
where the last equality follows from bilinearity of the Kronecker product.

We now construct a finite generating collection for
\(\mathcal A^{\mathrm{blk}}_{\mathfrak S,\mathcal G}\). Define
\[
\begin{aligned}
\mathcal H
:={}&
\left\{
\Phi_r\otimes G_s:
(r,s)\in\Lambda_+
\right\}\\
&\cup
\left\{
\pm(\Phi_r\otimes G_s):
(r,s)\notin\Lambda_+
\right\}\\
&\cup
\left\{
\pm(E_{ii}\otimes F_{ab}):
1\le i\le d,\;
1\le a,b\le c
\right\}.
\end{aligned}
\]
This collection is finite. Indeed, there are only \(RM\) possible pairs
\((r,s)\), and there are only \(dc^2\) triples \((i,a,b)\). Thus
\(\mathcal H\) contains finitely many matrices in
\(\mathbb R^{dc\times dc}\).

We claim that
$
\mathcal A^{\mathrm{blk}}_{\mathfrak S,\mathcal G}
=
\operatorname{cone}(\mathcal H).
$
We prove the two inclusions separately. Let
$
A\in
\mathcal A^{\mathrm{blk}}_{\mathfrak S,\mathcal G}.
$
By
Definition~\ref{def:operators-block-valued-interaction-family},
there exist coefficients
$
\beta_{rs}\in\mathbb R,
$
satisfying
$
\beta_{rs}\ge0
\qquad
\text{for every }(r,s)\in\Lambda_+,
$
and blocks
$
D_i\in\mathbb R^{c\times c}
$
such that
\[
A
=
\sum_{r=1}^R\sum_{s=1}^M
\beta_{rs}(\Phi_r\otimes G_s)
+
\sum_{i=1}^dE_{ii}\otimes D_i.
\]
For every pair
$
(r,s)\notin\Lambda_+,
$
the coefficient \(\beta_{rs}\) is unrestricted in sign. Write its positive
and negative parts as
$
\beta_{rs}^{+}
:=
\max\{\beta_{rs},0\},
\qquad
\beta_{rs}^{-}
:=
\max\{-\beta_{rs},0\}.
$
Then
$
\beta_{rs}^{+}\ge0,
\qquad
\beta_{rs}^{-}\ge0,
$
and
$
\beta_{rs}
=
\beta_{rs}^{+}-\beta_{rs}^{-}.
$
Therefore,
\[
\begin{aligned}
\beta_{rs}(\Phi_r\otimes G_s)
&=
\beta_{rs}^{+}(\Phi_r\otimes G_s)
-
\beta_{rs}^{-}(\Phi_r\otimes G_s)\\
&=
\beta_{rs}^{+}(\Phi_r\otimes G_s)
+
\beta_{rs}^{-}\bigl(-(\Phi_r\otimes G_s)\bigr).
\end{aligned}
\]
Both matrices
$
\Phi_r\otimes G_s
\qquad\text{and}\qquad
-(\Phi_r\otimes G_s)
$
belong to \(\mathcal H\) whenever
\((r,s)\notin\Lambda_+\), and the coefficients multiplying them above are
nonnegative.

Similarly, expand each block \(D_i\) as
$
D_i
=
\sum_{a=1}^c\sum_{b=1}^c
d_{i,ab}F_{ab}.
$
For each scalar coefficient \(d_{i,ab}\in\mathbb R\), define
\[
d_{i,ab}^{+}
:=
\max\{d_{i,ab},0\},
\qquad
d_{i,ab}^{-}
:=
\max\{-d_{i,ab},0\}.
\]
Then
$
d_{i,ab}^{+},d_{i,ab}^{-}\ge0
$
and
$
d_{i,ab}
=
d_{i,ab}^{+}-d_{i,ab}^{-}.
$
Hence
\[
\begin{aligned}
d_{i,ab}(E_{ii}\otimes F_{ab})
&=
d_{i,ab}^{+}(E_{ii}\otimes F_{ab})
+
d_{i,ab}^{-}\bigl(-(E_{ii}\otimes F_{ab})\bigr).
\end{aligned}
\]
Again, both matrices occurring on the right belong to
\(\mathcal H\).

Combining these decompositions with the coefficients already constrained to
be nonnegative on \(\Lambda_+\), we obtain
\[
\begin{aligned}
A
={}&
\sum_{(r,s)\in\Lambda_+}
\beta_{rs}(\Phi_r\otimes G_s)
\\
&+
\sum_{(r,s)\notin\Lambda_+}
\left[
\beta_{rs}^{+}(\Phi_r\otimes G_s)
+
\beta_{rs}^{-}
\bigl(-(\Phi_r\otimes G_s)\bigr)
\right]
\\
&+
\sum_{i=1}^d
\sum_{a=1}^c
\sum_{b=1}^c
\left[
d_{i,ab}^{+}(E_{ii}\otimes F_{ab})
+
d_{i,ab}^{-}
\bigl(-(E_{ii}\otimes F_{ab})\bigr)
\right].
\end{aligned}
\]
Every coefficient in this representation is nonnegative, and every matrix
being multiplied by such a coefficient belongs to \(\mathcal H\).
Therefore
$
A\in\operatorname{cone}(\mathcal H).
$
This proves
$
\mathcal A^{\mathrm{blk}}_{\mathfrak S,\mathcal G}
\subseteq
\operatorname{cone}(\mathcal H).
$

Conversely, let
$
A\in\operatorname{cone}(\mathcal H).
$
By definition of conic hull, \(A\) can be written as a nonnegative linear
combination of matrices in \(\mathcal H\). Thus there exist coefficients
$
a_{rs}\ge0
\qquad
((r,s)\in\Lambda_+),
$
coefficients
$
b_{rs}^{+},b_{rs}^{-}\ge0
\qquad
((r,s)\notin\Lambda_+),
$
and coefficients
$
c_{i,ab}^{+},c_{i,ab}^{-}\ge0
$
such that
\[
\begin{aligned}
A
={}&
\sum_{(r,s)\in\Lambda_+}
a_{rs}(\Phi_r\otimes G_s)
\\
&+
\sum_{(r,s)\notin\Lambda_+}
\left[
b_{rs}^{+}(\Phi_r\otimes G_s)
+
b_{rs}^{-}
\bigl(-(\Phi_r\otimes G_s)\bigr)
\right]
\\
&+
\sum_{i=1}^d
\sum_{a=1}^c
\sum_{b=1}^c
\left[
c_{i,ab}^{+}(E_{ii}\otimes F_{ab})
+
c_{i,ab}^{-}
\bigl(-(E_{ii}\otimes F_{ab})\bigr)
\right].
\end{aligned}
\]
Define
$
\beta_{rs}
:=
a_{rs}
\qquad
\text{for }(r,s)\in\Lambda_+,
$
and
$
\beta_{rs}
:=
b_{rs}^{+}-b_{rs}^{-}
\qquad
\text{for }(r,s)\notin\Lambda_+.
$
Then
$
\beta_{rs}\ge0
\qquad
\text{whenever }(r,s)\in\Lambda_+,
$
while the remaining coefficients \(\beta_{rs}\) are unrestricted real
numbers, exactly as required by
Definition~\ref{def:operators-block-valued-interaction-family}.

For each \(i=1,\ldots,d\), define
\[
D_i
:=
\sum_{a=1}^c\sum_{b=1}^c
\bigl(c_{i,ab}^{+}-c_{i,ab}^{-}\bigr)F_{ab}.
\]
Since each coefficient
$
c_{i,ab}^{+}-c_{i,ab}^{-}
$
is a real number, this defines an arbitrary admissible matrix
$
D_i\in\mathbb R^{c\times c}.
$
Using these definitions, the preceding conic representation recombines as
\[
A
=
\sum_{r=1}^R\sum_{s=1}^M
\beta_{rs}(\Phi_r\otimes G_s)
+
\sum_{i=1}^dE_{ii}\otimes D_i.
\]
Together with
$
\beta_{rs}\ge0
\qquad
\text{for }(r,s)\in\Lambda_+,
$
this is precisely the defining form of an element of
\(\mathcal A^{\mathrm{blk}}_{\mathfrak S,\mathcal G}\). Hence
$
A
\in
\mathcal A^{\mathrm{blk}}_{\mathfrak S,\mathcal G}.
$
Therefore,
$
\operatorname{cone}(\mathcal H)
\subseteq
\mathcal A^{\mathrm{blk}}_{\mathfrak S,\mathcal G}.
$
Combining the two inclusions yields
$
\mathcal A^{\mathrm{blk}}_{\mathfrak S,\mathcal G}
=
\operatorname{cone}(\mathcal H).
$

This representation immediately proves the cone and convexity properties.
Indeed, let
$
A,B
\in
\mathcal A^{\mathrm{blk}}_{\mathfrak S,\mathcal G},
$
and let
$
\lambda,\mu\ge0.
$
Because both \(A\) and \(B\) are nonnegative linear combinations of the
same finite generating collection \(\mathcal H\), the matrix
$
\lambda A+\mu B
$
is again a nonnegative linear combination of matrices in \(\mathcal H\).
Therefore
\[
\lambda A+\mu B
\in
\mathcal A^{\mathrm{blk}}_{\mathfrak S,\mathcal G}.
\]
Thus
$
\mathcal A^{\mathrm{blk}}_{\mathfrak S,\mathcal G}
$
is a convex cone.

Its finite-dimensionality can also be seen directly from the defining
representation. Every element belongs to
\[
\operatorname{span}
\left(
\{\Phi_r\otimes G_s:
1\le r\le R,\ 1\le s\le M\}
\cup
\{E_{ii}\otimes F_{ab}:
1\le i\le d,\ 1\le a,b\le c\}
\right).
\]
Consequently,
\[
\begin{aligned}
\dim
\operatorname{span}
\left(
\mathcal A^{\mathrm{blk}}_{\mathfrak S,\mathcal G}
\right)
&\le
RM+dc^2\\
&<
\infty.
\end{aligned}
\]
No linear-independence assumption is needed for this bound; dependencies
among the displayed matrices can only reduce the dimension. In particular,
the family is finite dimensional as a subset of the finite-dimensional
ambient space
$
\mathbb R^{dc\times dc}.
$

Since \(\mathcal H\) is finite and
$
\mathcal A^{\mathrm{blk}}_{\mathfrak S,\mathcal G}
=
\operatorname{cone}(\mathcal H),
$
the block-valued interaction family is a finitely generated convex cone.
By the finite-dimensional Minkowski--Weyl theorem, a finitely generated
convex cone is a polyhedral convex cone. Hence
$
\mathcal A^{\mathrm{blk}}_{\mathfrak S,\mathcal G}
$
is polyhedral. Equivalently, there exist finitely many linear functionals
\[
\ell_1,\ldots,\ell_K:
\mathbb R^{dc\times dc}\to\mathbb R
\]
such that
\[
\mathcal A^{\mathrm{blk}}_{\mathfrak S,\mathcal G}
=
\bigcap_{k=1}^K
\left\{
A\in\mathbb R^{dc\times dc}:
\ell_k(A)\ge0
\right\}.
\]
Each linear functional on the finite-dimensional space
\(\mathbb R^{dc\times dc}\) is continuous, and therefore every set
$
\left\{
A:
\ell_k(A)\ge0
\right\}
$
is closed. Since the intersection above is finite,
$
\mathcal A^{\mathrm{blk}}_{\mathfrak S,\mathcal G}
$
is closed. We have therefore established that it is a finite-dimensional
closed polyhedral convex cone in
\(\mathbb R^{dc\times dc}\).

It remains to consider the case
$
\Lambda_+=\varnothing.
$
In this case no coefficient \(\beta_{rs}\) is subject to a sign constraint.
Definition~\ref{def:operators-block-valued-interaction-family} therefore
reduces to
\[
\mathcal A^{\mathrm{blk}}_{\mathfrak S,\mathcal G}
=
\left\{
\sum_{r=1}^R\sum_{s=1}^M
\beta_{rs}(\Phi_r\otimes G_s)
+
\sum_{i=1}^dE_{ii}\otimes D_i
\;\middle|\;
\beta_{rs}\in\mathbb R,\;
D_i\in\mathbb R^{c\times c}
\right\}.
\]
Expanding every \(D_i\) in the basis \(\{F_{ab}\}\) gives
\[
\begin{aligned}
\mathcal A^{\mathrm{blk}}_{\mathfrak S,\mathcal G}
=
\operatorname{span}
\Bigl(
&
\{\Phi_r\otimes G_s:
1\le r\le R,\ 1\le s\le M\}
\\
&\cup
\{E_{ii}\otimes F_{ab}:
1\le i\le d,\ 1\le a,b\le c\}
\Bigr).
\end{aligned}
\]
The right-hand side is a linear span and hence is a linear subspace of
\(\mathbb R^{dc\times dc}\). Therefore, whenever
$
\Lambda_+=\varnothing,
$
the block-valued interaction family
\(\mathcal A^{\mathrm{blk}}_{\mathfrak S,\mathcal G}\) is a linear
subspace, as claimed.
\end{proof}

\subsection{Proof of Proposition~\ref{prop:operators-constant-compatibility-transferred-generators}}

\begin{proof}
Let
$
L\in\widehat{\mathcal L}^{\mathrm{ms}}
$
be arbitrary. By
Definition~\ref{def:operators-transferred-multiscale-generator-family},
there exist scale-wise generators
$
L_\ell\in\mathcal L_{\mathfrak S^{(\ell)}},
\qquad
\ell=0,\ldots,L,
$
such that
$
L
=
\sum_{\ell=0}^L
P_\ell L_\ell R_\ell.
$
Here
$
R_\ell:\mathbb R^{d_0}\to\mathbb R^{d_\ell},
\qquad
L_\ell:\mathbb R^{d_\ell}\to\mathbb R^{d_\ell},
\qquad
P_\ell:\mathbb R^{d_\ell}\to\mathbb R^{d_0},
$
so each composition
$
P_\ell L_\ell R_\ell
$
is a well-defined linear operator from
\(\mathbb R^{d_0}\) to \(\mathbb R^{d_0}\).

We evaluate the transferred operator on the finest-scale constant vector
\(\mathbf 1_{d_0}\). By linearity of matrix multiplication with respect to
finite sums,
\[
\begin{aligned}
L\mathbf 1_{d_0}
&=
\left(
\sum_{\ell=0}^L
P_\ell L_\ell R_\ell
\right)
\mathbf 1_{d_0}
\\
&=
\sum_{\ell=0}^L
P_\ell L_\ell
\left(
R_\ell\mathbf 1_{d_0}
\right).
\end{aligned}
\]
By the compatibility assumption of the proposition,
$
R_\ell\mathbf 1_{d_0}
=
\mathbf 1_{d_\ell}
\qquad
\text{for every }\ell=0,\ldots,L.
$
Substituting this identity into the preceding expression gives
$
L\mathbf 1_{d_0}
=
\sum_{\ell=0}^L
P_\ell L_\ell
\mathbf 1_{d_\ell}.
$

For each fixed
\(\ell\in\{0,\ldots,L\}\), we have
$
L_\ell
\in
\mathcal L_{\mathfrak S^{(\ell)}}.
$
By Definition~\ref{def:operators-generator-family}, applied to the geometry
descriptor \(\mathfrak S^{(\ell)}\) on the scale
\(X^{(\ell)}\) of cardinality \(d_\ell\), every such generator satisfies the
constant-preservation condition
$
L_\ell\mathbf 1_{d_\ell}
=
0_{d_\ell},
$
where \(0_{d_\ell}\in\mathbb R^{d_\ell}\) denotes the zero vector.
Therefore, for every \(\ell\),
$
P_\ell L_\ell\mathbf 1_{d_\ell}
=
P_\ell 0_{d_\ell}.
$
Since \(P_\ell\) is linear,
$
P_\ell 0_{d_\ell}
=
0_{d_0}.
$
Consequently, every term in the transferred sum vanishes, and hence
\[
\begin{aligned}
L\mathbf 1_{d_0}
&=
\sum_{\ell=0}^L
P_\ell L_\ell\mathbf 1_{d_\ell}
\\
&=
\sum_{\ell=0}^L
P_\ell 0_{d_\ell}
\\
&=
\sum_{\ell=0}^L
0_{d_0}
\\
&=
0_{d_0}.
\end{aligned}
\]
Thus
$
L\mathbf 1_{d_0}=0.
$

Since \(L\in\widehat{\mathcal L}^{\mathrm{ms}}\) was arbitrary, every
transferred multiscale generator satisfies the finest-scale
constant-preservation condition under the stated compatibility assumption.

Notice that no additional condition on the prolongation operators
\(P_\ell\) is required for this conclusion. The compatibility relation
$
R_\ell\mathbf 1_{d_0}
=
\mathbf 1_{d_\ell}
$
first transfers the finest-scale constant vector to the constant vector on
scale \(\ell\); the scale-wise generator then annihilates that vector:
$
L_\ell
\left(
R_\ell\mathbf 1_{d_0}
\right)
=
L_\ell\mathbf 1_{d_\ell}
=
0_{d_\ell}.
$
The subsequent application of \(P_\ell\) therefore acts only on the zero
vector and cannot change the conclusion. Hence constant preservation is
inherited by the entire transferred sum.
\end{proof}

\subsection{Proof of Proposition~\ref{prop:operators-basic-structure-multiscale-families}}

\begin{proof}
Let
$
\mathbb V
:=
\prod_{\ell=0}^L
\mathbb R^{d_\ell\times d_\ell}.
$
Since the number of scales is finite and every factor
\(\mathbb R^{d_\ell\times d_\ell}\) is finite dimensional, the product space
\(\mathbb V\) is itself finite dimensional, with
$
\dim(\mathbb V)
=
\sum_{\ell=0}^L d_\ell^2
<
\infty.
$
An element of \(\mathbb V\) will be denoted by
$
\mathbf M
=
(M_0,\ldots,M_L),
\qquad
M_\ell\in\mathbb R^{d_\ell\times d_\ell}.
$

Consider first the scale-wise interaction cones. For every
\(\ell=0,\ldots,L\), Proposition~\ref{prop:operators-interaction-family-structure},
applied to the geometry descriptor \(\mathfrak S^{(\ell)}\), shows that
$
\mathcal A_{\mathfrak S^{(\ell)}}
\subseteq
\mathbb R^{d_\ell\times d_\ell}
$
is a finite-dimensional closed polyhedral convex cone. In fact, the same
proposition gives finite generation. Hence, for every \(\ell\), there exists
a finite collection
\[
\Gamma_\ell^{A}
=
\left\{
G_{\ell,1}^{A},
\ldots,
G_{\ell,N_\ell^{A}}^{A}
\right\}
\subseteq
\mathbb R^{d_\ell\times d_\ell}
\]
such that
$
\mathcal A_{\mathfrak S^{(\ell)}}
=
\operatorname{cone}\!\left(\Gamma_\ell^{A}\right).
$
Thus every
$
A_\ell\in\mathcal A_{\mathfrak S^{(\ell)}}
$
admits a representation
\[
A_\ell
=
\sum_{s=1}^{N_\ell^{A}}
\alpha_{\ell,s}G_{\ell,s}^{A},
\qquad
\alpha_{\ell,s}\ge0.
\]

Define the product cone
$
\mathcal C_A
:=
\prod_{\ell=0}^L
\mathcal A_{\mathfrak S^{(\ell)}}
\subseteq
\mathbb V.
$
For every scale \(\ell\), let
$
\iota_\ell:
\mathbb R^{d_\ell\times d_\ell}
\longrightarrow
\mathbb V
$
denote the canonical coordinate embedding, namely,
$
\iota_\ell(M)
=
(0,\ldots,0,M,0,\ldots,0),
$
where \(M\) occupies the \(\ell\)-th component. We claim that
$
\mathcal C_A
=
\operatorname{cone}
\left(
\bigcup_{\ell=0}^L
\iota_\ell(\Gamma_\ell^{A})
\right).
$
Indeed, let
$
(A_0,\ldots,A_L)
\in
\mathcal C_A.
$
For every \(\ell\),
$
A_\ell
=
\sum_{s=1}^{N_\ell^{A}}
\alpha_{\ell,s}G_{\ell,s}^{A},
\qquad
\alpha_{\ell,s}\ge0.
$
Therefore
\[
\begin{aligned}
(A_0,\ldots,A_L)
&=
\sum_{\ell=0}^L
\iota_\ell(A_\ell)
\\
&=
\sum_{\ell=0}^L
\iota_\ell
\left(
\sum_{s=1}^{N_\ell^{A}}
\alpha_{\ell,s}G_{\ell,s}^{A}
\right)
\\
&=
\sum_{\ell=0}^L
\sum_{s=1}^{N_\ell^{A}}
\alpha_{\ell,s}
\iota_\ell(G_{\ell,s}^{A}).
\end{aligned}
\]
Every coefficient in the last expression is nonnegative, so the tuple lies in
the conic hull on the right. Conversely, every generator
\(\iota_\ell(G_{\ell,s}^{A})\) belongs to \(\mathcal C_A\), and
\(\mathcal C_A\) is a cone. Hence every nonnegative linear combination of
such generators also belongs to \(\mathcal C_A\), proving the claimed
equality. Since both the number of scales and every
\(N_\ell^{A}\) are finite, \(\mathcal C_A\) is a finitely generated cone.
By the finite-dimensional Minkowski--Weyl theorem, it is therefore a
polyhedral convex cone, and in particular it is closed.

Now define
$
\mathcal T:
\mathbb V
\longrightarrow
\mathbb R^{d_0\times d_0}
$
by
$
\mathcal T(M_0,\ldots,M_L)
:=
\sum_{\ell=0}^L
P_\ell M_\ell R_\ell.
$
The assumption that \(P_\ell\) and \(R_\ell\) are fixed is essential here:
with these matrices fixed, \(\mathcal T\) is a linear map. Indeed, if
\[
\mathbf M=(M_0,\ldots,M_L),
\qquad
\mathbf N=(N_0,\ldots,N_L),
\]
and \(a,b\in\mathbb R\), then
\[
\begin{aligned}
\mathcal T(a\mathbf M+b\mathbf N)
&=
\sum_{\ell=0}^L
P_\ell(aM_\ell+bN_\ell)R_\ell
\\
&=
a\sum_{\ell=0}^L
P_\ell M_\ell R_\ell
+
b\sum_{\ell=0}^L
P_\ell N_\ell R_\ell
\\
&=
a\,\mathcal T(\mathbf M)
+
b\,\mathcal T(\mathbf N).
\end{aligned}
\]
By Definition~\ref{def:operators-multiscale-interaction-family},
$
\mathcal A^{\mathrm{ms}}
=
\mathcal T(\mathcal C_A).
$

We next make the structure of this linear image explicit. Using the finite
generating representation of \(\mathcal C_A\),
\[
\begin{aligned}
\mathcal T(\mathcal C_A)
&=
\mathcal T
\left[
\operatorname{cone}
\left(
\bigcup_{\ell=0}^L
\iota_\ell(\Gamma_\ell^{A})
\right)
\right].
\end{aligned}
\]
If
$
\mathbf M
=
\sum_{\ell=0}^L
\sum_{s=1}^{N_\ell^{A}}
\alpha_{\ell,s}
\iota_\ell(G_{\ell,s}^{A}),
\qquad
\alpha_{\ell,s}\ge0,
$
then linearity gives
$
\mathcal T(\mathbf M)
=
\sum_{\ell=0}^L
\sum_{s=1}^{N_\ell^{A}}
\alpha_{\ell,s}
\mathcal T\!\left(
\iota_\ell(G_{\ell,s}^{A})
\right).
$
For an embedded single-scale matrix,
$
\mathcal T\!\left(
\iota_\ell(G_{\ell,s}^{A})
\right)
=
P_\ell G_{\ell,s}^{A}R_\ell.
$
It follows that
\[
\mathcal A^{\mathrm{ms}}
=
\operatorname{cone}
\left\{
P_\ell G_{\ell,s}^{A}R_\ell
\;\middle|\;
0\le\ell\le L,\;
1\le s\le N_\ell^{A}
\right\}.
\]
This is a conic hull of finitely many matrices in
\(\mathbb R^{d_0\times d_0}\). Hence
\(\mathcal A^{\mathrm{ms}}\) is a finitely generated convex cone.
By the finite-dimensional Minkowski--Weyl theorem, every finitely generated
cone is polyhedral, and every polyhedral cone in a finite-dimensional vector
space is closed. Consequently,
$
\mathcal A^{\mathrm{ms}}
$
is a finite-dimensional closed polyhedral convex cone in
\(\mathbb R^{d_0\times d_0}\). This also proves explicitly that it is the
linear image, under the fixed map \(\mathcal T\), of the finite-dimensional
polyhedral cone \(\mathcal C_A\).

We now repeat the same argument for the transferred generator family.
For every \(\ell\), Proposition~\ref{prop:operators-generator-family-structure}
shows that
$
\mathcal L_{\mathfrak S^{(\ell)}}
\subseteq
\mathbb R^{d_\ell\times d_\ell}
$
is a closed polyhedral convex cone. Since its ambient space is finite
dimensional, the Minkowski--Weyl theorem implies that this cone is finitely
generated. Thus there exists a finite collection
\[
\Gamma_\ell^{L}
=
\left\{
G_{\ell,1}^{L},
\ldots,
G_{\ell,N_\ell^{L}}^{L}
\right\}
\subseteq
\mathbb R^{d_\ell\times d_\ell}
\]
such that
$
\mathcal L_{\mathfrak S^{(\ell)}}
=
\operatorname{cone}\!\left(\Gamma_\ell^{L}\right).
$
If a scale-wise generator cone is the zero cone, the same notation remains
valid by taking a finite generating collection containing only the zero
matrix.

Define
$
\mathcal C_L
:=
\prod_{\ell=0}^L
\mathcal L_{\mathfrak S^{(\ell)}}
\subseteq
\mathbb V.
$
Exactly as above,
\[
\mathcal C_L
=
\operatorname{cone}
\left(
\bigcup_{\ell=0}^L
\iota_\ell(\Gamma_\ell^{L})
\right).
\]
Indeed, every tuple
$
(L_0,\ldots,L_L)
\in
\mathcal C_L
$
can be written componentwise as
$
L_\ell
=
\sum_{s=1}^{N_\ell^{L}}
\beta_{\ell,s}G_{\ell,s}^{L},
\qquad
\beta_{\ell,s}\ge0,
$
and hence
$
(L_0,\ldots,L_L)
=
\sum_{\ell=0}^L
\sum_{s=1}^{N_\ell^{L}}
\beta_{\ell,s}
\iota_\ell(G_{\ell,s}^{L}).
$
Thus \(\mathcal C_L\) is a finite-dimensional polyhedral convex cone.

By
Definition~\ref{def:operators-transferred-multiscale-generator-family},
$
\widehat{\mathcal L}^{\mathrm{ms}}
=
\mathcal T(\mathcal C_L).
$
Using the finite generating representation of \(\mathcal C_L\) and the
linearity of \(\mathcal T\), we obtain
\[
\widehat{\mathcal L}^{\mathrm{ms}}
=
\operatorname{cone}
\left\{
P_\ell G_{\ell,s}^{L}R_\ell
\;\middle|\;
0\le\ell\le L,\;
1\le s\le N_\ell^{L}
\right\}.
\]
Indeed, every element of the right-hand side is the image under
\(\mathcal T\) of a nonnegative linear combination of embedded scale-wise
generators, while every element of \(\mathcal T(\mathcal C_L)\) has such a
representation. Hence the two sets coincide.

The generating collection above is finite. Therefore
$
\widehat{\mathcal L}^{\mathrm{ms}}
$
is a finitely generated convex cone, and the finite-dimensional
Minkowski--Weyl theorem again implies that it is polyhedral and closed.
Thus
$
\widehat{\mathcal L}^{\mathrm{ms}}
$
is a finite-dimensional closed polyhedral convex cone in
\(\mathbb R^{d_0\times d_0}\), and it is explicitly the linear image of the
finite-dimensional polyhedral cone \(\mathcal C_L\).

It remains to establish the corresponding statement for the actual
multiscale generator family. Introduce
\[
\mathcal C_{\mathrm{gen}}^{(0)}
:=
\left\{
M\in\mathbb R^{d_0\times d_0}
\;\middle|\;
M_{ij}\ge0\text{ for every }i\neq j,
\quad
M\mathbf1_{d_0}=0
\right\}.
\]
By
Definition~\ref{def:operators-actual-multiscale-generator-family},
$
\mathcal L^{\mathrm{ms}}
=
\widehat{\mathcal L}^{\mathrm{ms}}
\cap
\mathcal C_{\mathrm{gen}}^{(0)}.
$
We now make explicit that
\(\mathcal C_{\mathrm{gen}}^{(0)}\) is a closed polyhedral convex cone.
For every \(i\neq j\), the condition
$
M_{ij}\ge0
$
is a homogeneous linear inequality in the entries of \(M\). There are
exactly
$
d_0(d_0-1)
$
such off-diagonal inequalities. Moreover,
$
M\mathbf1_{d_0}=0
$
is equivalent, row by row, to
$
\sum_{j=1}^{d_0}M_{ij}=0,
\qquad
i=1,\ldots,d_0.
$
Thus it consists of \(d_0\) homogeneous linear equations. If one wishes to
write the entire set using inequalities only, each equation may be replaced
by the two inequalities
$
\sum_{j=1}^{d_0}M_{ij}\ge0
\qquad\text{and}\qquad
-\sum_{j=1}^{d_0}M_{ij}\ge0.
$
Consequently,
\[
\begin{aligned}
\mathcal C_{\mathrm{gen}}^{(0)}
&=
\bigcap_{\substack{1\le i,j\le d_0\\i\neq j}}
\left\{
M:M_{ij}\ge0
\right\}
\\
&\qquad\cap
\bigcap_{i=1}^{d_0}
\left\{
M:
\sum_{j=1}^{d_0}M_{ij}=0
\right\},
\end{aligned}
\]
so it is described by finitely many homogeneous linear equations and
inequalities. Hence it is a polyhedral cone.

For completeness, its convex-cone property follows directly as well. If
$
M,N\in\mathcal C_{\mathrm{gen}}^{(0)}
$
and \(a,b\ge0\), then for every \(i\neq j\),
$
(aM+bN)_{ij}
=
aM_{ij}+bN_{ij}
\ge0,
$
while
\[
\begin{aligned}
(aM+bN)\mathbf1_{d_0}
&=
aM\mathbf1_{d_0}
+
bN\mathbf1_{d_0}
\\
&=
0.
\end{aligned}
\]
Therefore
$
aM+bN
\in
\mathcal C_{\mathrm{gen}}^{(0)}.
$
Each defining equality or inequality is the inverse image of a closed subset
of \(\mathbb R\) under a continuous linear functional, so
\(\mathcal C_{\mathrm{gen}}^{(0)}\) is also closed.

Since both
$
\widehat{\mathcal L}^{\mathrm{ms}}
$
and
$
\mathcal C_{\mathrm{gen}}^{(0)}
$
are closed polyhedral convex cones, their intersection is again a closed
convex cone. It is also polyhedral, because one obtains a finite
half-space/equality description of the intersection simply by adjoining the
finite system of constraints defining
\(\mathcal C_{\mathrm{gen}}^{(0)}\) to any finite polyhedral description of
\(\widehat{\mathcal L}^{\mathrm{ms}}\). Hence
$
\mathcal L^{\mathrm{ms}}
=
\widehat{\mathcal L}^{\mathrm{ms}}
\cap
\mathcal C_{\mathrm{gen}}^{(0)}
$
is a finite-dimensional closed polyhedral convex cone in
\(\mathbb R^{d_0\times d_0}\).

Thus
$
\mathcal A^{\mathrm{ms}}
\quad\text{and}\quad
\widehat{\mathcal L}^{\mathrm{ms}}
$
are linear images, under the fixed transfer map
\[
\mathcal T(M_0,\ldots,M_L)
=
\sum_{\ell=0}^L P_\ell M_\ell R_\ell,
\]
of finite-dimensional polyhedral cones, while
$
\mathcal L^{\mathrm{ms}}
$
is obtained from
\(\widehat{\mathcal L}^{\mathrm{ms}}\) by imposing precisely the additional
coordinatewise inequalities
$
L_{ij}\ge0
\qquad
(i\neq j)
$
and the linear equations
$
L\mathbf1_{d_0}=0.
$
This proves all of the asserted structural claims.
\end{proof}

\subsection{Proof of Proposition~\ref{prop:properties-frechet-sensitivity}}

\begin{proof}
Throughout the proof, write
\[
\|M\|
:=
\|M\|_{\infty\to\infty}
=
\max_{1\le i\le d}\sum_{j=1}^d|M_{ij}|.
\]
We regard the map
$
M\longmapsto e^{\tau M}
$
as a map on the ambient finite-dimensional vector space
\(\mathbb R^{d\times d}\). Thus the perturbation direction
\(H\in\mathbb R^{d\times d}\) need not itself satisfy the generator
constraints.

We first record the norm properties and exponential estimates that will be
used below. For arbitrary matrices \(A,B\in\mathbb R^{d\times d}\), the
\(i\)-th absolute row sum of \(AB\) satisfies
\[
\begin{aligned}
\sum_{j=1}^d |(AB)_{ij}|
&=
\sum_{j=1}^d
\left|
\sum_{k=1}^d A_{ik}B_{kj}
\right|
\\
&\le
\sum_{j=1}^d
\sum_{k=1}^d
|A_{ik}|\,|B_{kj}|
\\
&=
\sum_{k=1}^d
|A_{ik}|
\sum_{j=1}^d|B_{kj}|
\\
&\le
\sum_{k=1}^d
|A_{ik}|\,
\|B\|
\\
&=
\|B\|
\sum_{k=1}^d|A_{ik}|.
\end{aligned}
\]
Taking the maximum over \(i\) gives
$
\|AB\|
\le
\|A\|\,\|B\|.
$
Thus \(\|\cdot\|_{\infty\to\infty}\) is submultiplicative. In particular,
by induction,
$
\|B^n\|
\le
\|B\|^n
\qquad
\text{for every }n\ge0.
$
Using the power-series definition of the matrix exponential,
$
e^B
=
\sum_{n=0}^{\infty}\frac{B^n}{n!},
$
we therefore have
\[
\begin{aligned}
\|e^B\|
&\le
\sum_{n=0}^{\infty}
\frac{\|B^n\|}{n!}
\\
&\le
\sum_{n=0}^{\infty}
\frac{\|B\|^n}{n!}
\\
&=
e^{\|B\|}.
\end{aligned}
\]
This also shows absolute convergence of the matrix exponential in the norm
used throughout the proposition.

We next record the derivative of a matrix exponential with respect to a
scalar parameter, since this identity will also be used to establish the
Fr\'echet formula. Let \(B\in\mathbb R^{d\times d}\) be fixed. For
\(t\in\mathbb R\),
$
e^{tB}
=
\sum_{n=0}^{\infty}
\frac{t^nB^n}{n!}.
$
On every bounded interval \(|t|\le T\), the derivative series is dominated
in norm by
\[
\sum_{n=1}^{\infty}
\frac{nT^{n-1}\|B\|^n}{n!}
=
\|B\|
\sum_{n=1}^{\infty}
\frac{(T\|B\|)^{n-1}}{(n-1)!}
=
\|B\|e^{T\|B\|},
\]
and hence converges uniformly. Termwise differentiation is therefore
justified, giving
\[
\begin{aligned}
\frac{d}{dt}e^{tB}
&=
\sum_{n=1}^{\infty}
\frac{nt^{\,n-1}B^n}{n!}
\\
&=
\sum_{n=1}^{\infty}
\frac{t^{\,n-1}B^n}{(n-1)!}.
\end{aligned}
\]
Setting \(m=n-1\), we obtain
\[
\begin{aligned}
\frac{d}{dt}e^{tB}
&=
\sum_{m=0}^{\infty}
\frac{t^mB^{m+1}}{m!}
\\
&=
B
\sum_{m=0}^{\infty}
\frac{t^mB^m}{m!}
\\
&=
Be^{tB}.
\end{aligned}
\]
Since every power \(B^m\) commutes with \(B\), the same series may also be
factored on the right:
$
\frac{d}{dt}e^{tB}
=
e^{tB}B.
$
Hence
\[
\frac{d}{dt}e^{tB}
=
Be^{tB}
=
e^{tB}B.
\]

We now derive an exact perturbation identity for the matrix exponential.
Let \(A,E\in\mathbb R^{d\times d}\) be arbitrary and define, for
\(s\in[0,1]\),
$
\Psi(s)
:=
e^{(1-s)(A+E)}e^{sA}.
$
Using the scalar differentiation formula established above and the product
rule,
\[
\begin{aligned}
\Psi'(s)
&=
-e^{(1-s)(A+E)}
(A+E)e^{sA}
+
e^{(1-s)(A+E)}
Ae^{sA}.
\end{aligned}
\]
Here \(A+E\) commutes with
\(e^{(1-s)(A+E)}\), so the first differentiated factor may indeed be
written with \(A+E\) immediately to the right of its exponential. Combining
the two terms gives
\[
\begin{aligned}
\Psi'(s)
&=
e^{(1-s)(A+E)}
\bigl(-(A+E)+A\bigr)
e^{sA}
\\
&=
-
e^{(1-s)(A+E)}
E
e^{sA}.
\end{aligned}
\]
Integrating from \(0\) to \(1\) yields
\[
\Psi(1)-\Psi(0)
=
-
\int_0^1
e^{(1-s)(A+E)}
E
e^{sA}
\,ds.
\]
Since
$
\Psi(1)=e^A
\qquad\text{and}\qquad
\Psi(0)=e^{A+E},
$
we obtain the exact Duhamel identity
\[
e^{A+E}-e^A
=
\int_0^1
e^{(1-s)(A+E)}
E
e^{sA}
\,ds.
\]

We apply this identity to the map
$
\mathcal E_\tau(M)
:=
e^{\tau M}.
$
Fix \(L\in\mathcal L_{\mathfrak S}\) and let
\(K\in\mathbb R^{d\times d}\) be an arbitrary increment. Taking
$
A=\tau L,
\qquad
E=\tau K,
$
gives
\[
e^{\tau(L+K)}-e^{\tau L}
=
\tau
\int_0^1
e^{(1-s)\tau(L+K)}
K
e^{s\tau L}
\,ds.
\]
Motivated by the first-order part of this identity, define the linear map
\[
\mathcal D_{L,\tau}[K]
:=
\tau
\int_0^1
e^{(1-s)\tau L}
K
e^{s\tau L}
\,ds.
\]
The dependence on \(K\) is linear because both matrix multiplication by
fixed matrices and integration are linear. Subtracting this candidate
linear term from the exact difference gives the remainder
\[
\begin{aligned}
\mathcal R_{L,\tau}(K)
&:=
e^{\tau(L+K)}
-
e^{\tau L}
-
\mathcal D_{L,\tau}[K]
\\
&=
\tau
\int_0^1
\left(
e^{(1-s)\tau(L+K)}
-
e^{(1-s)\tau L}
\right)
K
e^{s\tau L}
\,ds.
\end{aligned}
\]
We now verify that this remainder is of second order in \(K\). For a fixed
\(s\in[0,1]\), apply the same Duhamel identity with
$
A_s=(1-s)\tau L,
\qquad
E_s=(1-s)\tau K.
$
Then
\[
\begin{aligned}
&e^{(1-s)\tau(L+K)}
-
e^{(1-s)\tau L}
\\
&\qquad=
(1-s)\tau
\int_0^1
e^{(1-r)(1-s)\tau(L+K)}
K
e^{r(1-s)\tau L}
\,dr.
\end{aligned}
\]
Using submultiplicativity and
\(\|e^B\|\le e^{\|B\|}\), we obtain
\[
\begin{aligned}
&
\left\|
e^{(1-s)\tau(L+K)}
-
e^{(1-s)\tau L}
\right\|
\\
&\qquad\le
(1-s)\tau\|K\|
\int_0^1
\left\|
e^{(1-r)(1-s)\tau(L+K)}
\right\|
\left\|
e^{r(1-s)\tau L}
\right\|
\,dr
\\
&\qquad\le
(1-s)\tau\|K\|
\int_0^1
\exp\!\left(
(1-r)(1-s)\tau\|L+K\|
\right)
\exp\!\left(
r(1-s)\tau\|L\|
\right)
\,dr.
\end{aligned}
\]
Since
$
\|L+K\|
\le
\|L\|+\|K\|,
$
the exponent in the integrand satisfies
\[
\begin{aligned}
&(1-r)(1-s)\tau\|L+K\|
+
r(1-s)\tau\|L\|
\\
&\qquad\le
(1-s)\tau
\left[
(1-r)(\|L\|+\|K\|)
+r\|L\|
\right]
\\
&\qquad=
(1-s)\tau
\left[
\|L\|+(1-r)\|K\|
\right]
\\
&\qquad\le
\tau(\|L\|+\|K\|).
\end{aligned}
\]
Consequently,
$
\left\|
e^{(1-s)\tau(L+K)}
-
e^{(1-s)\tau L}
\right\|
\le
(1-s)\tau
e^{\tau(\|L\|+\|K\|)}
\|K\|.
$
Returning to the remainder and using
$
\|e^{s\tau L}\|
\le
e^{s\tau\|L\|}
\le
e^{\tau\|L\|},
$
we obtain
\[
\begin{aligned}
\|\mathcal R_{L,\tau}(K)\|
&\le
\tau
\int_0^1
\left\|
e^{(1-s)\tau(L+K)}
-
e^{(1-s)\tau L}
\right\|
\|K\|
\|e^{s\tau L}\|
\,ds
\\
&\le
\tau
\int_0^1
(1-s)\tau
e^{\tau(\|L\|+\|K\|)}
\|K\|^2
e^{\tau\|L\|}
\,ds
\\
&=
\tau^2
e^{\tau(2\|L\|+\|K\|)}
\|K\|^2
\int_0^1(1-s)\,ds
\\
&=
\frac{\tau^2}{2}
e^{\tau(2\|L\|+\|K\|)}
\|K\|^2.
\end{aligned}
\]
In particular, whenever \(\|K\|\le1\),
$
\|\mathcal R_{L,\tau}(K)\|
\le
\frac{\tau^2}{2}
e^{\tau(2\|L\|+1)}
\|K\|^2.
$
Therefore
\[
\frac{
\|\mathcal R_{L,\tau}(K)\|
}{
\|K\|
}
\le
\frac{\tau^2}{2}
e^{\tau(2\|L\|+1)}
\|K\|
\longrightarrow
0
\qquad
\text{as }\|K\|\to0.
\]
This is precisely the Fr\'echet differentiability condition. Hence
\[
D_L e^{\tau L}[K]
=
\tau
\int_0^1
e^{(1-s)\tau L}
K
e^{s\tau L}
\,ds.
\]
Evaluating this bounded linear map in the direction \(H\) gives
\[
D_L e^{\tau L}[H]
=
\tau
\int_0^1
e^{(1-s)\tau L}
H
e^{s\tau L}
\,ds.
\]

We now use the generator structure to obtain the sharper norm estimate stated
in the proposition. Since
$
L\in\mathcal L_{\mathfrak S}
$
and both
$
(1-s)\tau\ge0
\qquad\text{and}\qquad
s\tau\ge0
$
for every \(s\in[0,1]\), Theorem~\ref{thm:properties-order-supnorm-stability}
gives
$
\left\|
e^{(1-s)\tau L}
\right\|
=
1
$
and
$
\left\|
e^{s\tau L}
\right\|
=
1.
$
Using the triangle inequality for the finite-dimensional matrix integral and
submultiplicativity of the induced norm,
\[
\begin{aligned}
\|D_L e^{\tau L}[H]\|
&=
\left\|
\tau
\int_0^1
e^{(1-s)\tau L}
H
e^{s\tau L}
\,ds
\right\|
\\
&\le
\tau
\int_0^1
\left\|
e^{(1-s)\tau L}
H
e^{s\tau L}
\right\|
\,ds
\\
&\le
\tau
\int_0^1
\left\|
e^{(1-s)\tau L}
\right\|
\|H\|
\left\|
e^{s\tau L}
\right\|
\,ds
\\
&=
\tau
\int_0^1
\|H\|
\,ds
\\
&=
\tau\|H\|.
\end{aligned}
\]
Restoring the full norm notation yields
$
\|D_L e^{\tau L}[H]\|_{\infty\to\infty}
\le
\tau
\|H\|_{\infty\to\infty}.
$

It remains to establish the derivative with respect to the propagation
magnitude. Applying the scalar differentiation identity proved above with
\(B=L\) and \(t=\tau\) gives
$
\partial_\tau e^{\tau L}
=
Le^{\tau L}.
$
Since \(e^{\tau L}\) is represented by a convergent power series in \(L\),
every term in that series commutes with \(L\). Hence
$
Le^{\tau L}
=
e^{\tau L}L,
$
and therefore
$
\partial_\tau e^{\tau L}
=
Le^{\tau L}
=
e^{\tau L}L.
$
Finally, Theorem~\ref{thm:properties-order-supnorm-stability} again gives
$
\|e^{\tau L}\|_{\infty\to\infty}=1.
$
Using submultiplicativity,
\[
\begin{aligned}
\|\partial_\tau e^{\tau L}\|_{\infty\to\infty}
&=
\|Le^{\tau L}\|_{\infty\to\infty}
\\
&\le
\|L\|_{\infty\to\infty}
\|e^{\tau L}\|_{\infty\to\infty}
\\
&=
\|L\|_{\infty\to\infty}.
\end{aligned}
\]
This proves all of the stated sensitivity identities and bounds.
\end{proof}

\subsection{Proof of Proposition~\ref{prop:properties-spectral-stability}}

\begin{proof}
Let
$
L\in\mathcal L_{\mathfrak S}
$
be arbitrary. By Definition~\ref{def:operators-generator-family}, \(L\) is a
real matrix satisfying
$
L_{ij}\ge0
\qquad
\text{for every }i\neq j,
$
and
$
L\mathbf1=0.
$
Moreover, by
Proposition~\ref{prop:operators-generator-diagonal-structure}, the zero-row-sum
condition implies
$
L_{ii}
=
-\sum_{j\neq i}L_{ij}
\qquad
\text{for every }i=1,\ldots,d.
$

Let
$
\lambda\in\sigma(L)
$
be arbitrary. Although \(L\) has real entries, its spectrum is understood over
\(\mathbb C\), so there exists a nonzero vector
$
v=(v_1,\ldots,v_d)^\top\in\mathbb C^d
$
such that
$
Lv=\lambda v.
$
Since \(v\neq0\), at least one component of \(v\) is nonzero. Choose an index
\(i_\ast\in\{1,\ldots,d\}\) satisfying
$
|v_{i_\ast}|
=
\max_{1\le k\le d}|v_k|.
$
Because \(v\neq0\), this maximum is strictly positive, and therefore
$
v_{i_\ast}\neq0.
$
For every \(j\in\{1,\ldots,d\}\), the maximality of
\(|v_{i_\ast}|\) gives
$
|v_j|
\le
|v_{i_\ast}|,
$
and hence
\[
\left|
\frac{v_j}{v_{i_\ast}}
\right|
=
\frac{|v_j|}{|v_{i_\ast}|}
\le
1.
\]

We now examine the \(i_\ast\)-th component of the eigenvalue equation
$
Lv=\lambda v.
$
Its left-hand side is
$
(Lv)_{i_\ast}
=
\sum_{j=1}^d
L_{i_\ast j}v_j,
$
while its right-hand side is
$
(\lambda v)_{i_\ast}
=
\lambda v_{i_\ast}.
$
Thus
$
\lambda v_{i_\ast}
=
\sum_{j=1}^d
L_{i_\ast j}v_j.
$
Separating the diagonal term from the off-diagonal terms gives
\[
\lambda v_{i_\ast}
=
L_{i_\ast i_\ast}v_{i_\ast}
+
\sum_{j\neq i_\ast}
L_{i_\ast j}v_j.
\]
By
Proposition~\ref{prop:operators-generator-diagonal-structure},
$
L_{i_\ast i_\ast}
=
-\sum_{j\neq i_\ast}
L_{i_\ast j}.
$
Substituting this identity into the previous equation yields
\[
\lambda v_{i_\ast}
=
-
\left(
\sum_{j\neq i_\ast}
L_{i_\ast j}
\right)
v_{i_\ast}
+
\sum_{j\neq i_\ast}
L_{i_\ast j}v_j.
\]
Distributing the first sum gives
$
\lambda v_{i_\ast}
=
\sum_{j\neq i_\ast}
\left(
-L_{i_\ast j}v_{i_\ast}
+
L_{i_\ast j}v_j
\right).
$
Factoring \(L_{i_\ast j}\) from every summand,
\[
\lambda v_{i_\ast}
=
\sum_{j\neq i_\ast}
L_{i_\ast j}
\left(
v_j-v_{i_\ast}
\right).
\]
Since \(v_{i_\ast}\neq0\), we may divide both sides by
\(v_{i_\ast}\). We obtain
$
\lambda
=
\sum_{j\neq i_\ast}
L_{i_\ast j}
\left(
\frac{v_j}{v_{i_\ast}}-1
\right).
$
Taking real parts and using the fact that the coefficients
\(L_{i_\ast j}\) are real gives
\[
\operatorname{Re}(\lambda)
=
\sum_{j\neq i_\ast}
L_{i_\ast j}
\operatorname{Re}
\left(
\frac{v_j}{v_{i_\ast}}-1
\right).
\]
Since \(1\) is real,
$
\operatorname{Re}
\left(
\frac{v_j}{v_{i_\ast}}-1
\right)
=
\operatorname{Re}
\left(
\frac{v_j}{v_{i_\ast}}
\right)-1.
$
Therefore
\[
\operatorname{Re}(\lambda)
=
\sum_{j\neq i_\ast}
L_{i_\ast j}
\left[
\operatorname{Re}
\left(
\frac{v_j}{v_{i_\ast}}
\right)-1
\right].
\]

For any complex number \(z\), one has
$
\operatorname{Re}(z)\le |z|.
$
Indeed, writing
$
z=x+\mathrm{i}y,
$
we have
$
|z|^2
=
x^2+y^2
\ge
x^2,
$
so
$
|\operatorname{Re}(z)|
=
|x|
\le
|z|,
$
which in particular implies
$
\operatorname{Re}(z)\le|z|.
$
Applying this elementary inequality with
$
z=\frac{v_j}{v_{i_\ast}}
$
and using the maximality of \(|v_{i_\ast}|\), we obtain
\[
\operatorname{Re}
\left(
\frac{v_j}{v_{i_\ast}}
\right)
\le
\left|
\frac{v_j}{v_{i_\ast}}
\right|
\le
1.
\]
Hence
$
\operatorname{Re}
\left(
\frac{v_j}{v_{i_\ast}}
\right)-1
\le
0
\qquad
\text{for every }j\neq i_\ast.
$
On the other hand, because
\(L\in\mathcal L_{\mathfrak S}\),
$
L_{i_\ast j}\ge0
\qquad
\text{for every }j\neq i_\ast.
$
It follows term by term that
$
L_{i_\ast j}
\left[
\operatorname{Re}
\left(
\frac{v_j}{v_{i_\ast}}
\right)-1
\right]
\le
0
\qquad
\text{for every }j\neq i_\ast.
$
Summing these nonpositive quantities gives
\[
\operatorname{Re}(\lambda)
=
\sum_{j\neq i_\ast}
L_{i_\ast j}
\left[
\operatorname{Re}
\left(
\frac{v_j}{v_{i_\ast}}
\right)-1
\right]
\le
0.
\]
Since the eigenvalue
\(\lambda\in\sigma(L)\) was arbitrary, we conclude that
$
\operatorname{Re}(\lambda)\le0
\qquad
\text{for every }\lambda\in\sigma(L).
$

It remains to show that the origin belongs to the spectrum. Again by
Definition~\ref{def:operators-generator-family},
$
L\mathbf1=0.
$
The vector
$
\mathbf1=(1,\ldots,1)^\top
$
is nonzero, so the preceding identity can be written as
$
L\mathbf1
=
0\,\mathbf1.
$
Thus \(\mathbf1\) is a nonzero right eigenvector of \(L\) associated with
the eigenvalue \(0\). Consequently,
$
0\in\sigma(L).
$

Therefore every eigenvalue of \(L\) lies in the closed left half-plane, and
the origin is always an eigenvalue, as claimed.
\end{proof}

\subsection{Proof of Proposition~\ref{prop:properties-sequential-oversmoothing}}

\begin{proof}
Let
$
L_k=L_k^\top\in\mathcal L_{\mathfrak S},
\qquad
\tau_k\ge0,
\qquad
k=1,\ldots,m,
$
and assume
$
\ker(L_k)=\operatorname{span}\{\mathbf1\}
\qquad
\text{for every }k.
$
We first identify the common subspace on which each propagation factor
contracts. Define
\[
\mathbf1^\perp
:=
\left\{
x\in\mathbb R^d:
\mathbf1^\top x=0
\right\}.
\]
For the vector appearing in the statement,
$
\overline h
=
\frac{\mathbf1^\top h}{d}\mathbf1.
$
Since
$
\mathbf1^\top\mathbf1=d,
$
we have
\[
\begin{aligned}
\mathbf1^\top(h-\overline h)
&=
\mathbf1^\top h
-
\mathbf1^\top
\left(
\frac{\mathbf1^\top h}{d}\mathbf1
\right)
\\
&=
\mathbf1^\top h
-
\frac{\mathbf1^\top h}{d}
\mathbf1^\top\mathbf1
\\
&=
\mathbf1^\top h
-
\frac{\mathbf1^\top h}{d}d
\\
&=
0.
\end{aligned}
\]
Therefore
$
h-\overline h\in\mathbf1^\perp.
$

Fix now an arbitrary
$
k\in\{1,\ldots,m\}.
$
Because \(L_k\) is real symmetric, the spectral theorem provides an
orthonormal basis
$
u_{k,1},\ldots,u_{k,d}
$
of \(\mathbb R^d\) consisting of eigenvectors of \(L_k\). Denote the
corresponding eigenvalues by
$
\lambda_{k,1},\ldots,\lambda_{k,d}.
$
Thus
\[
L_ku_{k,r}
=
\lambda_{k,r}u_{k,r},
\qquad
r=1,\ldots,d.
\]

Since
$
L_k\in\mathcal L_{\mathfrak S},
$
the generator condition gives
$
L_k\mathbf1=0.
$
Hence \(0\) is an eigenvalue of \(L_k\) with eigenvector \(\mathbf1\).
Moreover, the assumption
$
\ker(L_k)=\operatorname{span}\{\mathbf1\}
$
shows that the zero eigenspace is one-dimensional. We may therefore choose
\[
u_{k,1}
=
\frac{\mathbf1}{\|\mathbf1\|_2}
=
\frac1{\sqrt d}\mathbf1,
\qquad
\lambda_{k,1}=0.
\]
Because \(L_k\) is symmetric, all of its eigenvalues are real. In addition,
Proposition~\ref{prop:properties-spectral-stability} gives
$
\lambda_{k,r}\le0
\qquad
\text{for every }r.
$
For \(r\ge2\), the eigenvalue \(\lambda_{k,r}\) cannot be zero, because the
zero eigenspace is exactly
\(\operatorname{span}\{\mathbf1\}\). Hence
$
\lambda_{k,r}<0
\qquad
\text{for every }r=2,\ldots,d.
$
By the definition
\[
\gamma_k
=
\min\bigl\{
-\lambda:
\lambda\in\sigma(L_k)\setminus\{0\}
\bigr\},
\]
we consequently have
$
\gamma_k>0.
$
Furthermore, for every nonzero eigenvalue \(\lambda_{k,r}\),
$
-\lambda_{k,r}
\ge
\gamma_k,
$
which is equivalent to
$
\lambda_{k,r}
\le
-\gamma_k,
\qquad
r=2,\ldots,d.
$

We now establish the contraction of the \(k\)-th propagation operator on
\(\mathbf1^\perp\). Let
$
x\in\mathbf1^\perp
$
be arbitrary. Expand \(x\) in the orthonormal eigenbasis of \(L_k\):
$
x
=
\sum_{r=1}^d
\alpha_r u_{k,r},
\qquad
\alpha_r
=
u_{k,r}^\top x.
$
For the coefficient in the constant direction,
\[
\begin{aligned}
\alpha_1
&=
u_{k,1}^\top x
\\
&=
\left(
\frac1{\sqrt d}\mathbf1
\right)^\top x
\\
&=
\frac1{\sqrt d}\mathbf1^\top x
\\
&=
0,
\end{aligned}
\]
because \(x\in\mathbf1^\perp\). Therefore
$
x
=
\sum_{r=2}^d
\alpha_r u_{k,r}.
$
By orthonormality,
$
\|x\|_2^2
=
\sum_{r=2}^d\alpha_r^2.
$

For every eigenvector \(u_{k,r}\), the power-series definition of the
matrix exponential gives
$
e^{\tau_kL_k}u_{k,r}
=
e^{\tau_k\lambda_{k,r}}u_{k,r}.
$
Indeed,
\[
\begin{aligned}
e^{\tau_kL_k}u_{k,r}
&=
\sum_{q=0}^\infty
\frac{\tau_k^qL_k^q}{q!}u_{k,r}
\\
&=
\sum_{q=0}^\infty
\frac{\tau_k^q\lambda_{k,r}^q}{q!}u_{k,r}
\\
&=
e^{\tau_k\lambda_{k,r}}u_{k,r}.
\end{aligned}
\]
Consequently,
\[
\begin{aligned}
e^{\tau_kL_k}x
&=
e^{\tau_kL_k}
\left(
\sum_{r=2}^d
\alpha_r u_{k,r}
\right)
\\
&=
\sum_{r=2}^d
\alpha_r
e^{\tau_kL_k}u_{k,r}
\\
&=
\sum_{r=2}^d
\alpha_r
e^{\tau_k\lambda_{k,r}}
u_{k,r}.
\end{aligned}
\]
This expansion contains no component in the direction \(u_{k,1}\). Since
$
u_{k,1}
=
\frac1{\sqrt d}\mathbf1,
$
it follows immediately that
$
e^{\tau_kL_k}x
\in
\mathbf1^\perp.
$
Thus the subspace \(\mathbf1^\perp\) is invariant under every
\(e^{\tau_kL_k}\).

Using orthonormality again,
\[
\begin{aligned}
\|e^{\tau_kL_k}x\|_2^2
&=
\left\|
\sum_{r=2}^d
\alpha_r
e^{\tau_k\lambda_{k,r}}
u_{k,r}
\right\|_2^2
\\
&=
\sum_{r=2}^d
\alpha_r^2
e^{2\tau_k\lambda_{k,r}}.
\end{aligned}
\]
For every \(r\ge2\),
$
\lambda_{k,r}
\le
-\gamma_k.
$
Since \(\tau_k\ge0\),
$
2\tau_k\lambda_{k,r}
\le
-2\tau_k\gamma_k.
$
The scalar exponential function is increasing, and hence
$
e^{2\tau_k\lambda_{k,r}}
\le
e^{-2\tau_k\gamma_k}.
$
Substituting this inequality into the norm expansion gives
\[
\begin{aligned}
\|e^{\tau_kL_k}x\|_2^2
&=
\sum_{r=2}^d
\alpha_r^2
e^{2\tau_k\lambda_{k,r}}
\\
&\le
\sum_{r=2}^d
\alpha_r^2
e^{-2\tau_k\gamma_k}
\\
&=
e^{-2\tau_k\gamma_k}
\sum_{r=2}^d
\alpha_r^2
\\
&=
e^{-2\tau_k\gamma_k}
\|x\|_2^2.
\end{aligned}
\]
Both sides are nonnegative, so taking square roots yields
$
\|e^{\tau_kL_k}x\|_2
\le
e^{-\tau_k\gamma_k}
\|x\|_2
\qquad
\text{for every }x\in\mathbf1^\perp.
$
Thus, for each \(k\), the restriction of \(e^{\tau_kL_k}\) to the common
nonconstant subspace satisfies
$
\left\|
e^{\tau_kL_k}\big|_{\mathbf1^\perp}
\right\|_{2\to2}
\le
e^{-\tau_k\gamma_k}.
$

We now apply these bounds sequentially. Define
$
x_0
:=
h-\overline h
$
and, recursively,
$
x_k
:=
e^{\tau_kL_k}x_{k-1},
\qquad
k=1,\ldots,m.
$
As shown above,
$
x_0\in\mathbf1^\perp.
$
Because each factor \(e^{\tau_kL_k}\) maps
\(\mathbf1^\perp\) into itself, an immediate induction gives
$
x_k\in\mathbf1^\perp
\qquad
\text{for every }k=0,\ldots,m.
$
Therefore the preceding contraction estimate may be applied at every stage.
For \(k=1\),
\[
\|x_1\|_2
=
\|e^{\tau_1L_1}x_0\|_2
\le
e^{-\tau_1\gamma_1}
\|x_0\|_2.
\]
For the next stage,
\[
\begin{aligned}
\|x_2\|_2
&=
\|e^{\tau_2L_2}x_1\|_2
\\
&\le
e^{-\tau_2\gamma_2}
\|x_1\|_2
\\
&\le
e^{-\tau_2\gamma_2}
e^{-\tau_1\gamma_1}
\|x_0\|_2.
\end{aligned}
\]
Continuing the same argument through all \(m\) factors gives
\[
\|x_m\|_2
\le
\left(
\prod_{k=1}^m
e^{-\tau_k\gamma_k}
\right)
\|x_0\|_2.
\]
Using the elementary identity
\[
\prod_{k=1}^m
e^{-\tau_k\gamma_k}
=
\exp\left(
-\sum_{k=1}^m\tau_k\gamma_k
\right),
\]
we obtain
$
\|x_m\|_2
\le
\exp\left(
-\sum_{k=1}^m\tau_k\gamma_k
\right)
\|x_0\|_2.
$
Finally, by the recursive definition,
\[
x_m
=
e^{\tau_mL_m}
\cdots
e^{\tau_2L_2}
e^{\tau_1L_1}
(h-\overline h),
\]
while
$
x_0=h-\overline h.
$
Substituting these identities gives
\[
\left\|
e^{\tau_mL_m}\cdots e^{\tau_1L_1}
(h-\overline h)
\right\|_2
\le
\exp\left(
-\sum_{k=1}^m\tau_k\gamma_k
\right)
\|h-\overline h\|_2,
\]
which is the claimed sequential oversmoothing estimate.

No commutativity assumption between the matrices
\(L_1,\ldots,L_m\) is required: the argument uses only the fact that every
factor preserves the same subspace \(\mathbf1^\perp\) and contracts that
subspace by its own spectral factor
\(e^{-\tau_k\gamma_k}\).
\end{proof}

\subsection{Proof of Proposition~\ref{prop:properties-uniformization}}

\begin{proof}
Let
$
L\in\mathcal L_{\mathfrak S},
\qquad
\tau>0,
$
and choose
$
\nu>0
$
such that
$
\nu\ge\max_{1\le i\le d}(-L_{ii}).
$
Define
$
T
:=
I+\frac{L}{\nu}.
$
We first verify the structural properties of \(T\). By
Definition~\ref{def:operators-generator-family},
$
L_{ij}\ge0
\qquad
\text{for every }i\neq j,
$
and
$
L\mathbf1=0.
$
The latter condition implies, for each \(i\in\{1,\ldots,d\}\),
\[
0
=
(L\mathbf1)_i
=
\sum_{j=1}^dL_{ij}.
\]
Separating the diagonal term gives
$
0
=
L_{ii}
+
\sum_{j\neq i}L_{ij},
$
and therefore
$
L_{ii}
=
-\sum_{j\neq i}L_{ij}.
$
In particular,
$
L_{ii}\le0
$
and
$
-L_{ii}
=
\sum_{j\neq i}L_{ij}
\ge0.
$
This is also exactly the diagonal identity established in
Proposition~\ref{prop:operators-generator-diagonal-structure}.

Now fix \(i\neq j\). Since the identity matrix has zero off-diagonal
entries,
$
I_{ij}=0,
$
and hence
\[
T_{ij}
=
I_{ij}+\frac{L_{ij}}{\nu}
=
\frac{L_{ij}}{\nu}.
\]
Because
$
L_{ij}\ge0
\qquad\text{and}\qquad
\nu>0,
$
we obtain
$
T_{ij}\ge0
\qquad
\text{for every }i\neq j.
$
For a diagonal entry,
\[
T_{ii}
=
1+\frac{L_{ii}}{\nu}
=
1-\frac{-L_{ii}}{\nu}.
\]
By the choice of \(\nu\),
\[
-L_{ii}
\le
\max_{1\le r\le d}(-L_{rr})
\le
\nu,
\]
so
$
0
\le
\frac{-L_{ii}}{\nu}
\le
1.
$
Therefore
$
T_{ii}
=
1-\frac{-L_{ii}}{\nu}
\ge0.
$
Thus every entry of \(T\) is nonnegative:
$
T\ge0.
$

The row-sum property follows directly from the generator constraint. Indeed,
\[
\begin{aligned}
T\mathbf1
&=
\left(
I+\frac{L}{\nu}
\right)\mathbf1
\\
&=
I\mathbf1
+
\frac1\nu L\mathbf1
\\
&=
\mathbf1
+
\frac1\nu\,0
\\
&=
\mathbf1.
\end{aligned}
\]
Hence every row of \(T\) sums to one. Together with \(T\ge0\), this shows
that \(T\) is a nonnegative row-stochastic matrix.

We next derive the uniformization identity. From the definition of \(T\),
$
T-I
=
\frac{L}{\nu},
$
and therefore
$
L
=
\nu(T-I).
$
Consequently,
\[
\tau L
=
\nu\tau T-\nu\tau I.
\]
The matrices
$
\nu\tau T
\qquad\text{and}\qquad
-\nu\tau I
$
commute, because the identity matrix commutes with every matrix. We briefly
justify the corresponding exponential factorization directly from the
power-series definition. If \(A\) and \(B\) are commuting square matrices,
then, using any submultiplicative matrix norm,
\[
\sum_{r=0}^{\infty}
\sum_{s=0}^{\infty}
\left\|
\frac{A^rB^s}{r!\,s!}
\right\|
\le
\left(
\sum_{r=0}^{\infty}\frac{\|A\|^r}{r!}
\right)
\left(
\sum_{s=0}^{\infty}\frac{\|B\|^s}{s!}
\right)
=
e^{\|A\|}e^{\|B\|}
<
\infty.
\]
Thus the double series is absolutely convergent and may be rearranged
according to the total degree \(n=r+s\). Hence
\[
\begin{aligned}
e^Ae^B
&=
\left(
\sum_{r=0}^{\infty}\frac{A^r}{r!}
\right)
\left(
\sum_{s=0}^{\infty}\frac{B^s}{s!}
\right)
\\
&=
\sum_{r=0}^{\infty}
\sum_{s=0}^{\infty}
\frac{A^rB^s}{r!\,s!}
\\
&=
\sum_{n=0}^{\infty}
\sum_{r=0}^{n}
\frac{A^rB^{\,n-r}}{r!(n-r)!}
\\
&=
\sum_{n=0}^{\infty}
\frac1{n!}
\sum_{r=0}^{n}
\binom{n}{r}
A^rB^{\,n-r}.
\end{aligned}
\]
Because \(A\) and \(B\) commute, the matrix binomial identity applies:
\[
(A+B)^n
=
\sum_{r=0}^{n}
\binom{n}{r}
A^rB^{\,n-r}.
\]
Therefore
$
e^Ae^B
=
\sum_{n=0}^{\infty}
\frac{(A+B)^n}{n!}
=
e^{A+B}.
$
Applying this identity with
$
A=\nu\tau T,
B=-\nu\tau I,
$
gives
\[
\begin{aligned}
e^{\tau L}
&=
e^{\nu\tau T-\nu\tau I}
\\
&=
e^{\nu\tau T}e^{-\nu\tau I}.
\end{aligned}
\]
Since
$
(-\nu\tau I)^n
=
(-\nu\tau)^nI,
$
the exponential of the scalar multiple of the identity satisfies
\[
\begin{aligned}
e^{-\nu\tau I}
&=
\sum_{n=0}^{\infty}
\frac{(-\nu\tau I)^n}{n!}
\\
&=
\left(
\sum_{n=0}^{\infty}
\frac{(-\nu\tau)^n}{n!}
\right)I
\\
&=
e^{-\nu\tau}I.
\end{aligned}
\]
Likewise,
$
e^{\nu\tau T}
=
\sum_{k=0}^{\infty}
\frac{(\nu\tau T)^k}{k!}
=
\sum_{k=0}^{\infty}
\frac{(\nu\tau)^k}{k!}T^k.
$
Substituting these two identities yields
\[
\begin{aligned}
e^{\tau L}
&=
\left(
\sum_{k=0}^{\infty}
\frac{(\nu\tau)^k}{k!}T^k
\right)
e^{-\nu\tau}I
\\
&=
e^{-\nu\tau}
\sum_{k=0}^{\infty}
\frac{(\nu\tau)^k}{k!}T^k.
\end{aligned}
\]
This proves the asserted uniformization formula.

The coefficients in this expansion have a precise probabilistic
normalization. Define
$
p_k
:=
e^{-\nu\tau}
\frac{(\nu\tau)^k}{k!},
\qquad
k=0,1,2,\ldots.
$
Because
$
\nu>0
\qquad\text{and}\qquad
\tau>0,
$
we have
$
p_k\ge0.
$
Moreover,
\[
\begin{aligned}
\sum_{k=0}^{\infty}p_k
&=
e^{-\nu\tau}
\sum_{k=0}^{\infty}
\frac{(\nu\tau)^k}{k!}
\\
&=
e^{-\nu\tau}e^{\nu\tau}
\\
&=
1.
\end{aligned}
\]
Thus
$
\{p_k\}_{k\ge0}
$
is exactly the probability mass function of a Poisson random variable with
mean \(\nu\tau\).

It is also useful to observe that each \(T^k\) remains a nonnegative
row-stochastic matrix. Since \(T\ge0\), repeated matrix multiplication gives
$
T^k\ge0
\qquad
\text{for every }k\ge0.
$
Furthermore,
$
T\mathbf1=\mathbf1,
$
so
\[
T^2\mathbf1
=
T(T\mathbf1)
=
T\mathbf1
=
\mathbf1.
\]
Repeating the same calculation gives
$
T^k\mathbf1
=
\mathbf1
\qquad
\text{for every }k\ge0,
$
where \(T^0=I\). Thus the uniformization identity expresses
\(e^{\tau L}\) as a countable convex mixture of the discrete propagation
operators \(T^k\).

We now identify the one-step support of \(T\). Recall that
\[
\mathcal E_L
:=
\left\{
(z_i,z_j):
i\neq j,\ L_{ij}>0
\right\}.
\]
For distinct indices \(i\neq j\),
$
T_{ij}
=
\frac{L_{ij}}{\nu}.
$
Because \(\nu>0\),
$
T_{ij}>0
\quad\Longleftrightarrow\quad
L_{ij}>0.
$
Therefore
$
T_{ij}>0
\quad\Longleftrightarrow\quad
(z_i,z_j)\in\mathcal E_L,
\qquad
i\neq j.
$
In particular, the off-diagonal support of \(T\) is contained in
\(\mathcal E_L\); in fact, with the present definition of
\(\mathcal E_L\), it agrees with \(\mathcal E_L\) exactly.

Moreover, Proposition~\ref{prop:operators-generator-support} shows that
every nonzero off-diagonal entry of \(L\) respects the descriptor-level
one-step geometry:
\[
L_{ij}\neq0,\quad i\neq j
\quad\Longrightarrow\quad
(z_i,z_j)\in\mathcal E.
\]
Since
$
L_{ij}>0
\quad\Longrightarrow\quad
L_{ij}\neq0,
$
we obtain
$
\mathcal E_L
\subseteq
\mathcal E.
$
Thus every genuine cross-site move represented by \(T\) is an admissible
one-step interaction of the underlying geometry descriptor.

The diagonal entries have a different interpretation. From the zero-row-sum
identity,
$
-L_{ii}
=
\sum_{j\neq i}L_{ij},
$
and hence
\[
T_{ii}
=
1-\frac{-L_{ii}}{\nu}
=
1-\frac1\nu
\sum_{j\neq i}L_{ij}.
\]
Since
$
0\le -L_{ii}\le\nu,
$
we have
$
0\le T_{ii}\le1.
$
A diagonal factor \(T_{ii}\) therefore leaves the site index unchanged and
represents a holding step at \(z_i\). If \(T_{ii}=0\), no such holding step
has positive weight at that site; if \(T_{ii}>0\), the corresponding
self-loop has positive weight.

Finally, we make the discrete path interpretation of the powers \(T^k\)
explicit. Let \(k\ge1\). Repeated matrix multiplication gives, for
\(i,j\in\{1,\ldots,d\}\),
\[
(T^k)_{ij}
=
\sum_{a_1,\ldots,a_{k-1}=1}^d
T_{i a_{k-1}}
T_{a_{k-1}a_{k-2}}
\cdots
T_{a_2a_1}
T_{a_1j}.
\]
Writing
$
a_0:=j,
\qquad
a_k:=i,
$
this becomes
\[
(T^k)_{ij}
=
\sum_{a_1,\ldots,a_{k-1}=1}^d
\prod_{\ell=1}^k
T_{a_\ell a_{\ell-1}}.
\]
Every factor in every product is nonnegative. Hence every summand is
nonnegative. A summand can be strictly positive only if
$
T_{a_\ell a_{\ell-1}}>0
\qquad
\text{for every }\ell=1,\ldots,k.
$
For a given \(\ell\), if
$
a_\ell\neq a_{\ell-1},
$
then
$
T_{a_\ell a_{\ell-1}}>0
$
implies
$
(z_{a_\ell},z_{a_{\ell-1}})
\in
\mathcal E_L
\subseteq
\mathcal E.
$
Thus this factor represents an admissible cross-site move from
\(z_{a_{\ell-1}}\) to \(z_{a_\ell}\) under the standing
receiver--source convention. If instead
$
a_\ell=a_{\ell-1},
$
then the factor is diagonal and represents a holding step at the same site.
Consequently, every nonzero contribution to \((T^k)_{ij}\) corresponds to a
length-\(k\) discrete walk from \(z_j\) to \(z_i\) whose nontrivial moves
follow edges of \(\mathcal E_L\), while the remaining steps are diagonal
holds.

Combining this path expansion with
$
e^{\tau L}
=
\sum_{k=0}^{\infty}p_kT^k,
\qquad
p_k
=
e^{-\nu\tau}\frac{(\nu\tau)^k}{k!},
$
shows that finite propagation is a Poisson mixture over discrete walk
lengths \(k\). Every genuine cross-site transition appearing in those walks
is supported on \(\mathcal E_L\subseteq\mathcal E\), and diagonal factors
only allow the walk to remain at its current site. Hence the exponential is
a Poisson mixture of admissible discrete path lengths, as claimed.
\end{proof}

\subsection{Proof of Proposition~\ref{prop:properties-composition-of-commuting-propagations}}

\begin{proof}
Let
$
L_1,L_2\in\mathcal L_{\mathfrak S}^{(1)},
\qquad
\tau_1,\tau_2\ge0,
$
and suppose that
$
L_1L_2=L_2L_1.
$
Define
$
G
:=
\tau_1L_1+\tau_2L_2.
$
We first verify that \(G\) is again an admissible generator. Since
$
\mathcal L_{\mathfrak S}^{(1)}
\subseteq
\mathcal L_{\mathfrak S},
$
we have
$
L_1,L_2\in\mathcal L_{\mathfrak S}.
$
By Proposition~\ref{prop:operators-generator-family-structure},
\(\mathcal L_{\mathfrak S}\) is a convex cone. In particular, it is closed
under nonnegative linear combinations. Because
$
\tau_1\ge0
\qquad\text{and}\qquad
\tau_2\ge0,
$
it follows immediately that
\[
G
=
\tau_1L_1+\tau_2L_2
\in
\mathcal L_{\mathfrak S}.
\]
Equivalently, this conclusion may be checked directly from the generator
constraints. Since \(L_1,L_2\in\mathcal L_{\mathfrak S}\), for every
\(i\neq j\),
$
(L_1)_{ij}\ge0
\qquad\text{and}\qquad
(L_2)_{ij}\ge0,
$
and hence
$
G_{ij}
=
\tau_1(L_1)_{ij}
+
\tau_2(L_2)_{ij}
\ge0.
$
Moreover,
\[
\begin{aligned}
G\mathbf1
&=
(\tau_1L_1+\tau_2L_2)\mathbf1
\\
&=
\tau_1L_1\mathbf1
+
\tau_2L_2\mathbf1
\\
&=
0.
\end{aligned}
\]
Since \(\mathcal A_{\mathfrak S}\) is itself a cone and
\(L_1,L_2\in\mathcal A_{\mathfrak S}\), one also has
\(G\in\mathcal A_{\mathfrak S}\), so all defining conditions of
\(\mathcal L_{\mathfrak S}\) are satisfied.

We next prove the exponential composition identity. Set
$
A:=\tau_1L_1,
\qquad
B:=\tau_2L_2.
$
The assumed commutativity of \(L_1\) and \(L_2\) implies
\[
\begin{aligned}
AB
&=
(\tau_1L_1)(\tau_2L_2)
\\
&=
\tau_1\tau_2L_1L_2
\\
&=
\tau_1\tau_2L_2L_1
\\
&=
(\tau_2L_2)(\tau_1L_1)
\\
&=
BA.
\end{aligned}
\]
Thus \(A\) and \(B\) commute.

To justify the multiplication and rearrangement of the matrix exponential
series, fix any submultiplicative matrix norm \(\|\cdot\|\). Then
$
\|A^mB^n\|
\le
\|A\|^m\|B\|^n,
$
and therefore
\[
\begin{aligned}
\sum_{m=0}^{\infty}
\sum_{n=0}^{\infty}
\left\|
\frac{A^mB^n}{m!\,n!}
\right\|
&\le
\sum_{m=0}^{\infty}
\sum_{n=0}^{\infty}
\frac{\|A\|^m\|B\|^n}{m!\,n!}
\\
&=
\left(
\sum_{m=0}^{\infty}
\frac{\|A\|^m}{m!}
\right)
\left(
\sum_{n=0}^{\infty}
\frac{\|B\|^n}{n!}
\right)
\\
&=
e^{\|A\|}
e^{\|B\|}
\\
&<
\infty.
\end{aligned}
\]
Hence the resulting double series is absolutely convergent, and its terms
may be rearranged according to the total degree \(k=m+n\). Using the
power-series definition of the matrix exponential,
\[
\begin{aligned}
e^Ae^B
&=
\left(
\sum_{m=0}^{\infty}\frac{A^m}{m!}
\right)
\left(
\sum_{n=0}^{\infty}\frac{B^n}{n!}
\right)
\\
&=
\sum_{m=0}^{\infty}
\sum_{n=0}^{\infty}
\frac{A^mB^n}{m!\,n!}
\\
&=
\sum_{k=0}^{\infty}
\sum_{m=0}^{k}
\frac{A^mB^{k-m}}{m!(k-m)!}.
\end{aligned}
\]
For each fixed \(k\),
$
\frac{1}{m!(k-m)!}
=
\frac1{k!}\binom{k}{m},
$
so
\[
e^Ae^B
=
\sum_{k=0}^{\infty}
\frac1{k!}
\sum_{m=0}^{k}
\binom{k}{m}
A^mB^{k-m}.
\]
Because \(A\) and \(B\) commute, the usual binomial identity remains valid:
\[
(A+B)^k
=
\sum_{m=0}^{k}
\binom{k}{m}
A^mB^{k-m}.
\]
For completeness, this identity follows by induction. It is immediate for
\(k=0\). If it holds for some \(k\), then
\[
\begin{aligned}
(A+B)^{k+1}
&=
(A+B)^k(A+B)
\\
&=
\sum_{m=0}^{k}
\binom{k}{m}
A^mB^{k-m}A
+
\sum_{m=0}^{k}
\binom{k}{m}
A^mB^{k-m}B.
\end{aligned}
\]
Since \(AB=BA\), every power of \(A\) commutes with every power of \(B\), so
$
B^{k-m}A
=
AB^{k-m}.
$
Consequently,
\[
\begin{aligned}
(A+B)^{k+1}
&=
\sum_{m=0}^{k}
\binom{k}{m}
A^{m+1}B^{k-m}
+
\sum_{m=0}^{k}
\binom{k}{m}
A^mB^{k+1-m}.
\end{aligned}
\]
Reindexing the first sum and combining equal powers using Pascal's identity
$
\binom{k}{m-1}+\binom{k}{m}
=
\binom{k+1}{m}
$
gives
$
(A+B)^{k+1}
=
\sum_{m=0}^{k+1}
\binom{k+1}{m}
A^mB^{k+1-m},
$
which proves the commuting matrix binomial formula for every \(k\ge0\).

Substituting this identity into the exponential product gives
\[
\begin{aligned}
e^Ae^B
&=
\sum_{k=0}^{\infty}
\frac{(A+B)^k}{k!}
\\
&=
e^{A+B}.
\end{aligned}
\]
Returning to the definitions of \(A\) and \(B\),
$
A+B
=
\tau_1L_1+\tau_2L_2
=
G.
$
Therefore
$
e^{\tau_1L_1}e^{\tau_2L_2}
=
e^G.
$

We now determine the generator rate of \(G\). Since
$
L_1,L_2\in\mathcal L_{\mathfrak S}^{(1)},
$
Definition~\ref{def:operators-normalized-generator} gives
$
\rho(L_1)=1
\qquad\text{and}\qquad
\rho(L_2)=1.
$
Using the off-diagonal definition of \(\rho\),
$
\rho(M)
=
\frac1d
\sum_{i=1}^d
\sum_{j\neq i}M_{ij},
$
we compute
\[
\begin{aligned}
\rho(G)
&=
\frac1d
\sum_{i=1}^d
\sum_{j\neq i}
G_{ij}
\\
&=
\frac1d
\sum_{i=1}^d
\sum_{j\neq i}
\left[
\tau_1(L_1)_{ij}
+
\tau_2(L_2)_{ij}
\right]
\\
&=
\tau_1
\frac1d
\sum_{i=1}^d
\sum_{j\neq i}
(L_1)_{ij}
+
\tau_2
\frac1d
\sum_{i=1}^d
\sum_{j\neq i}
(L_2)_{ij}
\\
&=
\tau_1\rho(L_1)
+
\tau_2\rho(L_2)
\\
&=
\tau_1+\tau_2.
\end{aligned}
\]
Thus
$
\rho(G)
=
\tau_1+\tau_2.
$

Suppose now that
$
\tau_1+\tau_2>0.
$
Define
$
\widehat L
:=
\frac{G}{\tau_1+\tau_2}.
$
Using the definition of \(G\), this matrix can also be written as
\[
\widehat L
=
\frac{\tau_1}{\tau_1+\tau_2}L_1
+
\frac{\tau_2}{\tau_1+\tau_2}L_2.
\]
The two coefficients satisfy
$
\frac{\tau_1}{\tau_1+\tau_2}\ge0,
\qquad
\frac{\tau_2}{\tau_1+\tau_2}\ge0,
$
and
$
\frac{\tau_1}{\tau_1+\tau_2}
+
\frac{\tau_2}{\tau_1+\tau_2}
=
1.
$
Thus \(\widehat L\) is a convex combination of \(L_1\) and \(L_2\).
Since \(\mathcal L_{\mathfrak S}\) is a convex cone,
$
\widehat L
\in
\mathcal L_{\mathfrak S}.
$
Alternatively, because \(G\in\mathcal L_{\mathfrak S}\) and
$
\frac1{\tau_1+\tau_2}>0,
$
the cone property immediately gives the same conclusion.

By positive homogeneity of \(\rho\),
\[
\begin{aligned}
\rho(\widehat L)
&=
\rho\left(
\frac{G}{\tau_1+\tau_2}
\right)
\\
&=
\frac{1}{\tau_1+\tau_2}
\rho(G)
\\
&=
\frac{\tau_1+\tau_2}{\tau_1+\tau_2}
\\
&=
1.
\end{aligned}
\]
Therefore
$
\widehat L
\in
\mathcal L_{\mathfrak S}^{(1)}.
$
Moreover,
$
G
=
(\tau_1+\tau_2)\widehat L,
$
so the exponential identity already established becomes
\[
\begin{aligned}
e^{\tau_1L_1}e^{\tau_2L_2}
&=
e^G
\\
&=
e^{(\tau_1+\tau_2)\widehat L}.
\end{aligned}
\]
Because
$
\tau_1+\tau_2>0
\qquad\text{and}\qquad
\widehat L\in\mathcal L_{\mathfrak S}^{(1)},
$
Definition~\ref{def:operators-propagation-family} implies
\[
e^{(\tau_1+\tau_2)\widehat L}
\in
\mathcal P_{\mathfrak S}(\tau_1+\tau_2)
\subseteq
\mathcal P_{\mathfrak S}.
\]
Hence
$
e^{\tau_1L_1}e^{\tau_2L_2}
\in
\mathcal P_{\mathfrak S}
$
whenever
$
\tau_1+\tau_2>0.
$

It remains only to consider the zero-magnitude case. Since
$
\tau_1,\tau_2\ge0,
$
the equality
$
\tau_1+\tau_2=0
$
is equivalent to
$
\tau_1=\tau_2=0.
$
Then
$
e^{\tau_1L_1}
=
e^{0\cdot L_1}
=
I
$
and
$
e^{\tau_2L_2}
=
e^{0\cdot L_2}
=
I.
$
Therefore
$
e^{\tau_1L_1}e^{\tau_2L_2}
=
I.
$
By Definition~\ref{def:operators-propagation-family},
$
\mathcal P_{\mathfrak S}(0)
=
\{I\},
$
so
$
I\in\mathcal P_{\mathfrak S}.
$
This completes the proof.
\end{proof}

\subsection{Proof of Proposition~\ref{prop:properties-full-propagation-not-closed}}

\begin{proof}
We construct the stated three-state counterexample and verify each part
explicitly. Consider a geometry descriptor on three sites whose admissible
relation contains every ordered off-diagonal pair and whose channel family
contains the six off-diagonal indicator matrices
$
E_{ij},
\qquad
1\le i,j\le3,\quad i\neq j.
$
For instance, these indicator channels may all be included among the
positivity-constrained channels. Since
\(\operatorname{Diag}(\delta)\) supplies arbitrary diagonal entries, every
three-state matrix with nonnegative off-diagonal entries can be represented
inside the interaction family. Consequently, imposing the zero-row-sum
condition shows that
\(\mathcal L_{\mathfrak S}\) contains the full three-state generator cone.

Consider
\[
L_1
=
\begin{pmatrix}
-1 & 1 & 0\\
0 & 0 & 0\\
0 & 0 & 0
\end{pmatrix},
\qquad
L_2
=
\begin{pmatrix}
0 & 0 & 0\\
0 & -2 & 2\\
0 & 0 & 0
\end{pmatrix}.
\]
Both matrices have nonnegative off-diagonal entries. Their row sums are
zero:
$
L_1\mathbf1
=
\begin{pmatrix}
-1+1+0\\
0\\
0
\end{pmatrix}
=
0,
$
and
$
L_2\mathbf1
=
\begin{pmatrix}
0\\
-2+2\\
0
\end{pmatrix}
=
0.
$
Hence
$
L_1,L_2\in\mathcal L_{\mathfrak S}.
$
They are both nonzero generators. Since \(d=3\), their generator rates are
\[
\begin{aligned}
\rho(L_1)
&=
\frac13
\sum_{i=1}^3\sum_{j\neq i}(L_1)_{ij}
=
\frac13,
\\
\rho(L_2)
&=
\frac13
\sum_{i=1}^3\sum_{j\neq i}(L_2)_{ij}
=
\frac23.
\end{aligned}
\]
Therefore their normalized shapes are
$
\widehat L_1
=
\frac{L_1}{\rho(L_1)}
=
3L_1
\in
\mathcal L_{\mathfrak S}^{(1)}
$
and
$
\widehat L_2
=
\frac{L_2}{\rho(L_2)}
=
\frac32L_2
\in
\mathcal L_{\mathfrak S}^{(1)}.
$
Let
$
t:=\log2>0.
$
Then
$
tL_1
=
\frac{t}{3}\widehat L_1
$
and
$
tL_2
=
\frac{2t}{3}\widehat L_2.
$
Thus, by Definition~\ref{def:operators-propagation-family},
$
P_1:=e^{tL_1}
=
e^{(t/3)\widehat L_1}
\in
\mathcal P_{\mathfrak S}(t/3)
\subseteq
\mathcal P_{\mathfrak S},
$
and similarly
\[
P_2:=e^{tL_2}
=
e^{(2t/3)\widehat L_2}
\in
\mathcal P_{\mathfrak S}(2t/3)
\subseteq
\mathcal P_{\mathfrak S}.
\]

We now compute these two propagators explicitly. Direct multiplication gives
\[
L_1^2
=
\begin{pmatrix}
1 & -1 & 0\\
0 & 0 & 0\\
0 & 0 & 0
\end{pmatrix}
=
-L_1.
\]
It follows inductively that
$
L_1^n
=
(-1)^{n-1}L_1
\qquad
\text{for every }n\ge1.
$
Therefore
\[
\begin{aligned}
e^{tL_1}
&=
I+
\sum_{n=1}^{\infty}
\frac{t^nL_1^n}{n!}
\\
&=
I+
\left(
\sum_{n=1}^{\infty}
\frac{(-1)^{n-1}t^n}{n!}
\right)L_1.
\end{aligned}
\]
Since
$
e^{-t}
=
1+
\sum_{n=1}^{\infty}
\frac{(-1)^nt^n}{n!},
$
we have
$
\sum_{n=1}^{\infty}
\frac{(-1)^{n-1}t^n}{n!}
=
1-e^{-t}.
$
Thus
$
e^{tL_1}
=
I+(1-e^{-t})L_1.
$
Because
$
t=\log2,
\qquad
e^{-t}
=
\frac12,
$
this gives
$
P_1
=
e^{tL_1}
=
\begin{pmatrix}
\frac12 & \frac12 & 0\\
0 & 1 & 0\\
0 & 0 & 1
\end{pmatrix}.
$

Likewise,
$
L_2^2
=
\begin{pmatrix}
0 & 0 & 0\\
0 & 4 & -4\\
0 & 0 & 0
\end{pmatrix}
=
-2L_2.
$
Hence
$
L_2^n
=
(-2)^{n-1}L_2
\qquad
\text{for every }n\ge1.
$
Using the exponential series,
\[
\begin{aligned}
e^{tL_2}
&=
I+
\sum_{n=1}^{\infty}
\frac{t^n(-2)^{n-1}}{n!}L_2
\\
&=
I+
\frac12
\left(
\sum_{n=1}^{\infty}
\frac{(-1)^{n-1}(2t)^n}{n!}
\right)L_2
\\
&=
I+
\frac{1-e^{-2t}}{2}L_2.
\end{aligned}
\]
Since
$
e^{-2t}
=
e^{-2\log2}
=
\frac14,
$
we obtain
$
P_2
=
e^{tL_2}
=
\begin{pmatrix}
1 & 0 & 0\\
0 & \frac14 & \frac34\\
0 & 0 & 1
\end{pmatrix}.
$
Multiplying the two propagators in the order stated in the proposition gives
\[
\begin{aligned}
P_2P_1
&=
\begin{pmatrix}
1 & 0 & 0\\
0 & \frac14 & \frac34\\
0 & 0 & 1
\end{pmatrix}
\begin{pmatrix}
\frac12 & \frac12 & 0\\
0 & 1 & 0\\
0 & 0 & 1
\end{pmatrix}
\\
&=
\begin{pmatrix}
\frac12 & \frac12 & 0\\
0 & \frac14 & \frac34\\
0 & 0 & 1
\end{pmatrix}.
\end{aligned}
\]
For brevity, write
$
M:=P_2P_1.
$

Since \(M\) is upper triangular, its eigenvalues are its diagonal entries:
$
\sigma(M)
=
\left\{
\frac12,\frac14,1
\right\}.
$
These three eigenvalues are real, positive, and pairwise distinct. Explicit
corresponding eigenvectors are
\[
v_1
=
\begin{pmatrix}
1\\0\\0
\end{pmatrix},
\qquad
v_2
=
\begin{pmatrix}
-2\\1\\0
\end{pmatrix},
\qquad
v_3
=
\begin{pmatrix}
1\\1\\1
\end{pmatrix}.
\]
Indeed,
$
Mv_1
=
\frac12v_1,
\qquad
Mv_2
=
\frac14v_2,
\qquad
Mv_3
=
v_3.
$
The vectors \(v_1,v_2,v_3\) are linearly independent. Define
\[
S
:=
\begin{pmatrix}
1 & -2 & 1\\
0 & 1 & 1\\
0 & 0 & 1
\end{pmatrix},
\]
whose columns are \(v_1,v_2,v_3\). Then
\[
S^{-1}
=
\begin{pmatrix}
1 & 2 & -3\\
0 & 1 & -1\\
0 & 0 & 1
\end{pmatrix},
\]
and the diagonalization of \(M\) is
\[
M
=
S
\begin{pmatrix}
\frac12 & 0 & 0\\
0 & \frac14 & 0\\
0 & 0 & 1
\end{pmatrix}
S^{-1}.
\]

Define
\[
X
:=
S
\begin{pmatrix}
-\log2 & 0 & 0\\
0 & -\log4 & 0\\
0 & 0 & 0
\end{pmatrix}
S^{-1}.
\]
Since similarity transformations commute with matrix powers,
$
(SDS^{-1})^n
=
SD^nS^{-1}
$
for every \(n\ge0\). Hence the matrix exponential also respects similarity:
$
e^{SDS^{-1}}
=
Se^DS^{-1}.
$
Applying this identity to the preceding diagonal matrix gives
\[
\begin{aligned}
e^X
&=
S
\begin{pmatrix}
e^{-\log2} & 0 & 0\\
0 & e^{-\log4} & 0\\
0 & 0 & 1
\end{pmatrix}
S^{-1}
\\
&=
S
\begin{pmatrix}
\frac12 & 0 & 0\\
0 & \frac14 & 0\\
0 & 0 & 1
\end{pmatrix}
S^{-1}
\\
&=
M.
\end{aligned}
\]
Thus \(X\) is a real logarithm of \(M\). Direct multiplication yields
\[
\begin{aligned}
X
&=
\begin{pmatrix}
1 & -2 & 1\\
0 & 1 & 1\\
0 & 0 & 1
\end{pmatrix}
\begin{pmatrix}
-\log2 & 0 & 0\\
0 & -\log4 & 0\\
0 & 0 & 0
\end{pmatrix}
\begin{pmatrix}
1 & 2 & -3\\
0 & 1 & -1\\
0 & 0 & 1
\end{pmatrix}
\\
&=
\begin{pmatrix}
-\log2 & \log4 & -\log2\\
0 & -\log4 & \log4\\
0 & 0 & 0
\end{pmatrix}.
\end{aligned}
\]

We now prove that this is not merely one real logarithm of \(M\), but its
unique real logarithm. Let
$
Y\in\mathbb R^{3\times3}
$
be any real matrix satisfying
$
e^Y=M.
$
Every matrix commutes with its own exponential, because
$
e^Y
=
\sum_{n=0}^{\infty}\frac{Y^n}{n!}
$
is a convergent power series in \(Y\). Hence
$
Ye^Y=e^YY,
$
and therefore
$
YM=MY.
$

Let \(v_r\) be any one of the three eigenvectors above, with corresponding
eigenvalue
$
\lambda_r\in
\left\{
\frac12,\frac14,1
\right\}.
$
Using \(YM=MY\),
\[
\begin{aligned}
M(Yv_r)
&=
Y(Mv_r)
\\
&=
Y(\lambda_rv_r)
\\
&=
\lambda_rYv_r.
\end{aligned}
\]
Thus \(Yv_r\) belongs to the \(\lambda_r\)-eigenspace of \(M\). Because all
three eigenvalues of \(M\) are distinct, each corresponding eigenspace is
one-dimensional. Hence there exists a scalar \(\mu_r\) such that
$
Yv_r=\mu_rv_r.
$
Both \(Y\) and \(v_r\) are real, so
$
\mu_r\in\mathbb R.
$
Applying \(e^Y\) to \(v_r\) now gives
\[
\begin{aligned}
Mv_r
&=
e^Yv_r
\\
&=
\sum_{n=0}^{\infty}
\frac{Y^nv_r}{n!}
\\
&=
\sum_{n=0}^{\infty}
\frac{\mu_r^n}{n!}v_r
\\
&=
e^{\mu_r}v_r.
\end{aligned}
\]
On the other hand,
$
Mv_r=\lambda_rv_r.
$
Since \(v_r\neq0\),
$
e^{\mu_r}
=
\lambda_r.
$
Because
$
\mu_r\in\mathbb R
\qquad\text{and}\qquad
\lambda_r>0,
$
the real scalar exponential is injective, and therefore
$
\mu_r
=
\log\lambda_r.
$
Consequently,
$
Yv_1
=
-\log2\,v_1,
$
$
Yv_2
=
-\log4\,v_2,
$
and
$
Yv_3
=
0.
$
Since \(v_1,v_2,v_3\) form a basis of \(\mathbb R^3\), these three identities
determine \(Y\) uniquely. Hence
\[
Y
=
S
\begin{pmatrix}
-\log2 & 0 & 0\\
0 & -\log4 & 0\\
0 & 0 & 0
\end{pmatrix}
S^{-1}
=
X.
\]
Therefore \(M\) has exactly one real logarithm, namely
\[
\log M
=
X
=
\begin{pmatrix}
-\log2 & \log4 & -\log2\\
0 & -\log4 & \log4\\
0 & 0 & 0
\end{pmatrix}.
\]

The entry in position \((1,3)\) is
$
X_{13}
=
-\log2
<
0.
$
Since \(1\neq3\), this is an off-diagonal entry. Hence \(X\) is not a
Metzler matrix. Every element of every generator family
\(\mathcal L_{\mathfrak S}\) must have nonnegative off-diagonal entries by
Definition~\ref{def:operators-generator-family}. Therefore
$
X\notin\mathcal L_{\mathfrak S}.
$

It remains to show that this excludes \(M\) from the propagation family,
rather than merely excluding the particular logarithm displayed above.
Assume, toward a contradiction, that
$
M
\in
\mathcal P_{\mathfrak S}.
$
Since
$
M_{11}
=
\frac12
\neq1,
$
we have
$
M\neq I.
$
Thus \(M\notin\mathcal P_{\mathfrak S}(0)\), because
$
\mathcal P_{\mathfrak S}(0)=\{I\}.
$
By Definition~\ref{def:operators-propagation-family}, there must therefore
exist
$
\tau>0
$
and
$
\widehat L
\in
\mathcal L_{\mathfrak S}^{(1)}
$
such that
$
M
=
e^{\tau\widehat L}.
$
Set
$
Y:=\tau\widehat L.
$
Then \(Y\) is real and
$
e^Y=M.
$
Moreover, because
$
\widehat L\in\mathcal L_{\mathfrak S}
\qquad\text{and}\qquad
\tau>0,
$
the cone property of
\(\mathcal L_{\mathfrak S}\), established in
Proposition~\ref{prop:operators-generator-family-structure}, gives
$
Y=\tau\widehat L
\in
\mathcal L_{\mathfrak S}.
$
In particular, \(Y\) must be Metzler.

However, the uniqueness of the real logarithm proved above forces
$
Y
=
X
=
\log M.
$
But \(X\) has the strictly negative off-diagonal entry
$
X_{13}
=
-\log2<0,
$
so
$
X\notin\mathcal L_{\mathfrak S}.
$
This is a contradiction. Hence
$
M
=
P_2P_1
\notin
\mathcal P_{\mathfrak S}.
$

We have thus exhibited two matrices
$
P_1,P_2\in\mathcal P_{\mathfrak S}
$
whose product does not belong to
\(\mathcal P_{\mathfrak S}\). Therefore there exist geometry descriptors for
which the propagation family is not closed under matrix multiplication.
\end{proof}

\subsection{Proof of Proposition~\ref{prop:properties-generated-algebra}}

\begin{proof}
For brevity, denote
$
\mathscr A_\Phi
:=
\operatorname{Alg}_{\mathbb R}
\left(
I,\Phi_1,\ldots,\Phi_R,E_{11},\ldots,E_{dd}
\right),
$
and define
$
\mathscr B_\Phi
:=
\left\{
B\in\mathbb R^{d\times d}:
\operatorname{supp}_X(B)\subseteq\mathcal R_\Phi
\right\}.
$
We prove that
$
\mathscr A_\Phi=\mathscr B_\Phi
$
by establishing both inclusions.

We first verify that \(\mathscr B_\Phi\) is itself a unital real matrix
algebra. Let \(B,C\in\mathscr B_\Phi\) and let
\(\alpha,\beta\in\mathbb R\). Suppose that
$
(z_i,z_j)\notin\mathcal R_\Phi.
$
Since
$
\operatorname{supp}_X(B)
\subseteq
\mathcal R_\Phi
\qquad\text{and}\qquad
\operatorname{supp}_X(C)
\subseteq
\mathcal R_\Phi,
$
we necessarily have
$
B_{ij}=0
\qquad\text{and}\qquad
C_{ij}=0.
$
Consequently,
\[
(\alpha B+\beta C)_{ij}
=
\alpha B_{ij}+\beta C_{ij}
=
0.
\]
Thus
$
\operatorname{supp}_X(\alpha B+\beta C)
\subseteq
\mathcal R_\Phi,
$
and therefore
$
\alpha B+\beta C\in\mathscr B_\Phi.
$
Hence \(\mathscr B_\Phi\) is a real vector subspace of
\(\mathbb R^{d\times d}\).

Because \(\mathcal R_\Phi\) is the reflexive transitive closure of
\(\mathcal E_\Phi\), it is reflexive. Therefore
$
(z_i,z_i)\in\mathcal R_\Phi
\qquad
\text{for every }i=1,\ldots,d.
$
The identity matrix has support
$
\operatorname{supp}_X(I)
=
\{(z_i,z_i):1\le i\le d\},
$
and hence
$
\operatorname{supp}_X(I)
\subseteq
\mathcal R_\Phi.
$
It follows that
$
I\in\mathscr B_\Phi.
$

We next prove closure under matrix multiplication. Let
$
B,C\in\mathscr B_\Phi,
$
and fix \(i,j\in\{1,\ldots,d\}\). Suppose that
$
(BC)_{ij}\neq0.
$
By the definition of matrix multiplication,
$
(BC)_{ij}
=
\sum_{k=1}^d B_{ik}C_{kj}.
$
This is a finite sum. If every summand were zero, then the sum itself would
be zero. Since
$
(BC)_{ij}\neq0,
$
there must exist at least one
$
k\in\{1,\ldots,d\}
$
such that
$
B_{ik}C_{kj}\neq0.
$
A product of two real numbers is nonzero only if both factors are nonzero,
so
$
B_{ik}\neq0
\qquad\text{and}\qquad
C_{kj}\neq0.
$
Because \(B,C\in\mathscr B_\Phi\), these two nonzero entries imply
$
(z_i,z_k)\in\mathcal R_\Phi
$
and
$
(z_k,z_j)\in\mathcal R_\Phi.
$
The relation \(\mathcal R_\Phi\) is transitive, since it is a reflexive
transitive closure. Therefore
$
(z_i,z_k)\in\mathcal R_\Phi
\quad\text{and}\quad
(z_k,z_j)\in\mathcal R_\Phi
$
imply
$
(z_i,z_j)\in\mathcal R_\Phi.
$
We have thus shown that
\[
(BC)_{ij}\neq0
\quad\Longrightarrow\quad
(z_i,z_j)\in\mathcal R_\Phi.
\]
Since \(i,j\) were arbitrary,
$
\operatorname{supp}_X(BC)
\subseteq
\mathcal R_\Phi,
$
and hence
$
BC\in\mathscr B_\Phi.
$
Thus \(\mathscr B_\Phi\) is a unital real matrix algebra.

We now check that this algebra contains every generator used to define
\(\mathscr A_\Phi\). We have already shown that
$
I\in\mathscr B_\Phi.
$
For each \(i\),
\[
\operatorname{supp}_X(E_{ii})
=
\{(z_i,z_i)\}
\subseteq
\mathcal R_\Phi,
\]
so
$
E_{ii}\in\mathscr B_\Phi.
$
Now fix a channel index \(r\in\{1,\ldots,R\}\). Suppose that
$
(\Phi_r)_{ij}\neq0.
$
If \(i=j\), then
$
(z_i,z_j)
=
(z_i,z_i)
\in
\mathcal R_\Phi
$
by reflexivity. If \(i\neq j\), then the definition of the effective channel
relation gives
$
(z_i,z_j)\in\mathcal E_\Phi,
$
because the channel \(\Phi_r\) itself has a nonzero entry at \((i,j)\).
Since
$
\mathcal E_\Phi
\subseteq
\mathcal R_\Phi,
$
we again obtain
$
(z_i,z_j)\in\mathcal R_\Phi.
$
Hence every nonzero entry of \(\Phi_r\) is indexed by a pair in
\(\mathcal R_\Phi\), so
$
\operatorname{supp}_X(\Phi_r)
\subseteq
\mathcal R_\Phi.
$
Therefore
$
\Phi_r\in\mathscr B_\Phi
\qquad
\text{for every }r=1,\ldots,R.
$

Thus \(\mathscr B_\Phi\) is a unital real matrix algebra containing
\[
I,\Phi_1,\ldots,\Phi_R,E_{11},\ldots,E_{dd}.
\]
By the defining minimality of the generated algebra
\(\mathscr A_\Phi\), it follows that
$
\mathscr A_\Phi
\subseteq
\mathscr B_\Phi.
$

We now prove the reverse inclusion. We first show that every matrix unit
\(E_{ij}\) whose site pair belongs to \(\mathcal R_\Phi\) lies in
\(\mathscr A_\Phi\). Let
$
(z_i,z_j)\in\mathcal R_\Phi.
$
If \(i=j\), then
$
E_{ij}=E_{ii},
$
which is one of the defining generators of \(\mathscr A_\Phi\). Hence
$
E_{ii}\in\mathscr A_\Phi.
$

Assume now that
$
i\neq j.
$
Since \(\mathcal R_\Phi\) is the reflexive transitive closure of
\(\mathcal E_\Phi\), membership of the distinct pair
\((z_i,z_j)\) in \(\mathcal R_\Phi\) means that there exist an integer
\(m\ge1\) and indices
\[
a_0,a_1,\ldots,a_m\in\{1,\ldots,d\}
\]
such that
$
a_0=j,
\qquad
a_m=i,
$
and
$
(z_{a_\ell},z_{a_{\ell-1}})
\in
\mathcal E_\Phi
\qquad
\text{for every }\ell=1,\ldots,m.
$
Under the standing receiver--source convention, this is a directed effective
channel path
\[
z_j=z_{a_0}
\longrightarrow
z_{a_1}
\longrightarrow
\cdots
\longrightarrow
z_{a_m}=z_i.
\]

Fix
$
\ell\in\{1,\ldots,m\}.
$
Since
$
(z_{a_\ell},z_{a_{\ell-1}})
\in
\mathcal E_\Phi,
$
the definition of \(\mathcal E_\Phi\) guarantees that there exists at least
one channel index
$
r_\ell\in\{1,\ldots,R\}
$
such that
$
(\Phi_{r_\ell})_{a_\ell a_{\ell-1}}
\neq0.
$
We claim that the corresponding one-step matrix unit
$
E_{a_\ell a_{\ell-1}}
$
belongs to \(\mathscr A_\Phi\). Since
$
E_{a_\ell a_\ell},
\qquad
\Phi_{r_\ell},
\qquad
E_{a_{\ell-1}a_{\ell-1}}
$
all belong to \(\mathscr A_\Phi\), and \(\mathscr A_\Phi\) is closed under
matrix multiplication,
$
E_{a_\ell a_\ell}
\Phi_{r_\ell}
E_{a_{\ell-1}a_{\ell-1}}
\in
\mathscr A_\Phi.
$
We now compute this product entrywise. For arbitrary
\(p,q\in\{1,\ldots,d\}\),
\[
\begin{aligned}
\left(
E_{a_\ell a_\ell}
\Phi_{r_\ell}
E_{a_{\ell-1}a_{\ell-1}}
\right)_{pq}
&=
\sum_{u=1}^d
\sum_{v=1}^d
(E_{a_\ell a_\ell})_{pu}
(\Phi_{r_\ell})_{uv}
(E_{a_{\ell-1}a_{\ell-1}})_{vq}.
\end{aligned}
\]
The matrix \(E_{a_\ell a_\ell}\) has only one nonzero entry, so the first
factor can be nonzero only when
$
p=u=a_\ell.
$
Likewise, the final factor can be nonzero only when
$
v=q=a_{\ell-1}.
$
Hence every term in the double sum vanishes except the one with
$
u=a_\ell,
\qquad
v=a_{\ell-1},
$
and therefore
\[
\left(
E_{a_\ell a_\ell}
\Phi_{r_\ell}
E_{a_{\ell-1}a_{\ell-1}}
\right)_{pq}
=
\begin{cases}
(\Phi_{r_\ell})_{a_\ell a_{\ell-1}},
&
p=a_\ell,\ q=a_{\ell-1},
\\
0,
&
\text{otherwise}.
\end{cases}
\]
Equivalently,
$
E_{a_\ell a_\ell}
\Phi_{r_\ell}
E_{a_{\ell-1}a_{\ell-1}}
=
(\Phi_{r_\ell})_{a_\ell a_{\ell-1}}
E_{a_\ell a_{\ell-1}}.
$
Since
$
(\Phi_{r_\ell})_{a_\ell a_{\ell-1}}
\neq0,
$
its reciprocal is a well-defined real scalar. A real matrix algebra is closed
under real scalar multiplication, so
\[
\begin{aligned}
E_{a_\ell a_{\ell-1}}
&=
\frac{1}{
(\Phi_{r_\ell})_{a_\ell a_{\ell-1}}
}
E_{a_\ell a_\ell}
\Phi_{r_\ell}
E_{a_{\ell-1}a_{\ell-1}}
\\
&\in
\mathscr A_\Phi.
\end{aligned}
\]
Thus every effective one-step edge along the chosen path yields its
corresponding matrix unit inside the generated algebra.

We now multiply these matrix units along the path. Recall that for standard
matrix units,
\[
E_{ab}E_{cd}
=
\begin{cases}
E_{ad},& b=c,\\
0,& b\neq c.
\end{cases}
\]
Indeed, for arbitrary \(p,q\),
\[
\begin{aligned}
(E_{ab}E_{cd})_{pq}
&=
\sum_{s=1}^d
(E_{ab})_{ps}(E_{cd})_{sq}.
\end{aligned}
\]
The first factor is nonzero only when
$
p=a,\qquad s=b,
$
while the second is nonzero only when
$
s=c,\qquad q=d.
$
Thus a nonzero contribution exists exactly when \(b=c\), in which case the
only nonzero entry of the product occurs at \((a,d)\). This proves the stated
matrix-unit multiplication identity.

Applying this identity successively to the path gives
\[
\begin{aligned}
&
E_{a_m a_{m-1}}
E_{a_{m-1}a_{m-2}}
\cdots
E_{a_2a_1}
E_{a_1a_0}
\\
&\qquad=
E_{a_m a_0}.
\end{aligned}
\]
Since
$
a_m=i
\qquad\text{and}\qquad
a_0=j,
$
we obtain
$
E_{a_m a_0}
=
E_{ij}.
$
Every factor on the left belongs to \(\mathscr A_\Phi\), and
\(\mathscr A_\Phi\) is closed under multiplication. Hence
$
E_{ij}\in\mathscr A_\Phi.
$
We have therefore established
\[
(z_i,z_j)\in\mathcal R_\Phi
\quad\Longrightarrow\quad
E_{ij}\in\mathscr A_\Phi
\]
for every \(i,j\).

Now let
$
B\in\mathscr B_\Phi.
$
By definition,
$
\operatorname{supp}_X(B)
\subseteq
\mathcal R_\Phi.
$
Every matrix admits its standard matrix-unit expansion
$
B
=
\sum_{i=1}^d
\sum_{j=1}^d
B_{ij}E_{ij}.
$
If
$
(z_i,z_j)\notin\mathcal R_\Phi,
$
then the support condition forces
$
B_{ij}=0.
$
Hence the expansion reduces to
$
B
=
\sum_{\substack{1\le i,j\le d\\
(z_i,z_j)\in\mathcal R_\Phi}}
B_{ij}E_{ij}.
$
For every pair occurring in this sum, we have already proved that
$
E_{ij}\in\mathscr A_\Phi.
$
Since \(\mathscr A_\Phi\) is a real vector space, every finite real linear
combination of these matrix units also belongs to \(\mathscr A_\Phi\).
Therefore
$
B\in\mathscr A_\Phi.
$
Since \(B\in\mathscr B_\Phi\) was arbitrary,
$
\mathscr B_\Phi
\subseteq
\mathscr A_\Phi.
$
Combining the two inclusions gives
\[
\operatorname{Alg}_{\mathbb R}
\left(
I,\Phi_1,\ldots,\Phi_R,E_{11},\ldots,E_{dd}
\right)
=
\left\{
B\in\mathbb R^{d\times d}:
\operatorname{supp}_X(B)\subseteq\mathcal R_\Phi
\right\}.
\]

Finally, suppose that the directed graph induced by
\(\mathcal E_\Phi\) is strongly connected. Then for every ordered pair of
sites \(z_j,z_i\), either \(i=j\), in which case the pair belongs to the
reflexive closure, or there exists a directed path in
\(\mathcal E_\Phi\) from \(z_j\) to \(z_i\). Therefore
$
\mathcal R_\Phi
=
X\times X.
$
The support restriction on the right-hand side of the preceding equality is
then vacuous:
\[
\left\{
B\in\mathbb R^{d\times d}:
\operatorname{supp}_X(B)\subseteq\mathcal R_\Phi
\right\}
=
\mathbb R^{d\times d}.
\]
Consequently,
$
\operatorname{Alg}_{\mathbb R}
\left(
I,\Phi_1,\ldots,\Phi_R,E_{11},\ldots,E_{dd}
\right)
=
\mathbb R^{d\times d}.
$

The argument is algebraic: a matrix unit associated with a pair at path
length \(m\) is obtained through a product of \(m\) isolated one-step matrix
units, followed by arbitrary real linear combinations. Thus the conclusion
concerns the unbounded finite algebra generated by repeated compositions.
It does not assert that a prescribed fixed number of architectural layers
already realizes every matrix in this algebra.
\end{proof}

\subsection{Proof of Proposition~\ref{prop:properties-unrestricted-diagonal-selectors}}

\begin{proof}
Let
$
f:K\to\mathbb R^d,
\qquad
K=[1,2]^d,
$
be an arbitrary continuous map. Write
$
f(h)
=
\bigl(
f_1(h),\ldots,f_d(h)
\bigr)^\top
$
for
$
h=(h_1,\ldots,h_d)^\top\in K.
$
Since
$
h_i\in[1,2]
\qquad
\text{for every }i=1,\ldots,d
$
and every \(h\in K\), we have in particular
$
h_i\ge1>0.
$
Thus division by \(h_i\) is well defined at every point of \(K\).

Define
$
\delta:K\to\mathbb R^d
$
coordinatewise by
$
\delta_i(h)
:=
\frac{f_i(h)-h_i}{h_i},
\qquad
i=1,\ldots,d.
$
Equivalently,
$
\delta_i(h)
=
\frac{f_i(h)}{h_i}-1.
$
We then define the matrix-valued selector
\[
A(h)
:=
\operatorname{Diag}(\delta(h))
=
\operatorname{Diag}
\left(
\frac{f_1(h)-h_1}{h_1},
\ldots,
\frac{f_d(h)-h_d}{h_d}
\right).
\]
By construction, \(A(h)\) is diagonal for every \(h\in K\).

We first verify that this selector is continuous. For each fixed
\(i\in\{1,\ldots,d\}\), the coordinate projection
$
h\longmapsto h_i
$
is continuous on \(K\), and by assumption
$
h\longmapsto f_i(h)
$
is also continuous. Therefore
$
h\longmapsto f_i(h)-h_i
$
is continuous. Moreover, because
$
h_i\ge1
\qquad
\text{for every }h\in K,
$
the denominator \(h_i\) never vanishes on \(K\). Hence the quotient
$
h\longmapsto
\frac{f_i(h)-h_i}{h_i}
$
is continuous on \(K\). Thus every coordinate function
\(\delta_i\) is continuous, and consequently
$
\delta:K\to\mathbb R^d
$
is continuous.

The diagonal embedding
\[
\operatorname{Diag}:\mathbb R^d\to\mathbb R^{d\times d},
\qquad
v\longmapsto\operatorname{Diag}(v),
\]
is linear and therefore continuous. Since
$
A
=
\operatorname{Diag}\circ\delta,
$
it follows that
$
A:K\to\mathbb R^{d\times d}
$
is continuous. More explicitly, if \(h^{(n)}\to h\) in \(K\), then
$
\delta_i(h^{(n)})
\longrightarrow
\delta_i(h)
\qquad
\text{for every }i,
$
and therefore, for example in the Frobenius norm,
\[
\begin{aligned}
\|A(h^{(n)})-A(h)\|_F^2
&=
\sum_{i=1}^d
\left|
\delta_i(h^{(n)})-\delta_i(h)
\right|^2
\\
&\longrightarrow
0.
\end{aligned}
\]
Hence the required matrix-valued selector is indeed continuous.

We next verify that
$
A(h)\in\mathcal A_{\mathfrak S}
\qquad
\text{for every }h\in K.
$
Recall from Definition~\ref{def:operators-interaction-family} that
\[
\mathcal A_{\mathfrak S}
=
\left\{
\sum_{r=1}^R
\theta_r\Phi_r
+
\operatorname{Diag}(\delta)
\;\middle|\;
\theta\in\mathbb R^R,\;
\delta\in\mathbb R^d,\;
\theta_r\ge0
\text{ for }r\in\mathcal R_+
\right\}.
\]
For the matrix \(A(h)\), choose
$
\theta_r=0
\qquad
\text{for every }r=1,\ldots,R,
$
and choose the diagonal coefficient vector to be
$
\delta=\delta(h).
$
For every constrained channel index
\(r\in\mathcal R_+\),
$
\theta_r=0\ge0,
$
so all sign constraints are satisfied. Moreover,
\[
\begin{aligned}
\sum_{r=1}^R
\theta_r\Phi_r
+
\operatorname{Diag}(\delta(h))
&=
\sum_{r=1}^R
0\cdot\Phi_r
+
\operatorname{Diag}(\delta(h))
\\
&=
\operatorname{Diag}(\delta(h))
\\
&=
A(h).
\end{aligned}
\]
Therefore
$
A(h)\in\mathcal A_{\mathfrak S}
\qquad
\text{for every }h\in K.
$
In particular, the construction uses no off-diagonal channel term at all;
the entire selector is realized through the unrestricted diagonal
self-action already present in
\(\mathcal A_{\mathfrak S}\).

It remains to verify the claimed identity. Since \(A(h)\) is diagonal, the
\(i\)-th coordinate of \(A(h)h\) is
$
(A(h)h)_i
=
A(h)_{ii}h_i.
$
By the definition of \(A(h)\),
$
A(h)_{ii}
=
\delta_i(h)
=
\frac{f_i(h)-h_i}{h_i}.
$
Hence
\[
\begin{aligned}
(A(h)h)_i
&=
\frac{f_i(h)-h_i}{h_i}\,h_i
\\
&=
f_i(h)-h_i,
\end{aligned}
\]
where cancellation of \(h_i\) is valid because
$
h_i\ge1>0.
$
It follows that
\[
\begin{aligned}
\bigl(h+A(h)h\bigr)_i
&=
h_i+(A(h)h)_i
\\
&=
h_i+f_i(h)-h_i
\\
&=
f_i(h).
\end{aligned}
\]
This equality holds for every
$
i=1,\ldots,d.
$
Therefore the two vectors agree coordinatewise:
$
h+A(h)h
=
f(h)
\qquad
\text{for every }h\in K.
$

Thus every continuous map
$
f:K\to\mathbb R^d
$
can be represented exactly by a continuous self-conditioned selector taking
values in \(\mathcal A_{\mathfrak S}\), even when the selector is restricted
to use diagonal matrices only. This proves the proposition.
\end{proof}

\subsection{Proof of Proposition~\ref{prop:layers-self-conditioned-stability}}

\begin{proof}
Let \(h,g\in K\) be arbitrary, and for brevity write
$
Q_h:=Q(h),
\qquad
Q_g:=Q(g),
$
and
$
\|\cdot\|
:=
\|\cdot\|_{\infty\to\infty}
$
for matrix norms throughout the proof. We first establish the basic
properties of the exponentials generated by \(Q_h\) and \(Q_g\).

Let \(Q\in\mathbb R^{d\times d}\) be any Metzler matrix satisfying
$
Q\mathbf1=0.
$
We claim that, for every \(t\ge0\),
$
e^{tQ}\ge0,
\qquad
e^{tQ}\mathbf1=\mathbf1,
\qquad
\|e^{tQ}\|_{\infty\to\infty}=1.
$
Choose a scalar \(a>0\) sufficiently large that
$
a\ge
\max_{1\le i\le d}(-Q_{ii}).
$
Because \(Q\) is Metzler,
$
Q_{ij}\ge0
\qquad
\text{for }i\neq j.
$
Hence the matrix
$
B:=Q+aI
$
is entrywise nonnegative. Indeed, for \(i\neq j\),
$
B_{ij}=Q_{ij}\ge0,
$
whereas, for every \(i\),
$
B_{ii}=Q_{ii}+a\ge0.
$
Therefore
$
B\ge0.
$
Since a product of entrywise nonnegative matrices is entrywise
nonnegative,
$
B^n\ge0
\qquad
\text{for every }n\ge0.
$
The power-series definition of the exponential consequently gives
$
e^{tB}
=
\sum_{n=0}^{\infty}
\frac{t^nB^n}{n!}
\ge0
\qquad
\text{for every }t\ge0.
$
Because
$
Q=B-aI
$
and \(B\) commutes with \(I\),
\[
e^{tQ}
=
e^{t(B-aI)}
=
e^{-at}e^{tB}.
\]
The scalar factor \(e^{-at}\) is strictly positive, and hence
$
e^{tQ}\ge0.
$

We next use the zero-row-sum property. Since
$
Q\mathbf1=0,
$
we have
$
Q^n\mathbf1=0
\qquad
\text{for every }n\ge1.
$
Indeed, the case \(n=1\) is the assumed identity, and if
\(Q^n\mathbf1=0\), then
\[
Q^{n+1}\mathbf1
=
Q(Q^n\mathbf1)
=
Q0
=
0.
\]
Therefore
\[
\begin{aligned}
e^{tQ}\mathbf1
&=
\left(
I+
\sum_{n=1}^{\infty}
\frac{t^nQ^n}{n!}
\right)\mathbf1
\\
&=
\mathbf1
+
\sum_{n=1}^{\infty}
\frac{t^n}{n!}Q^n\mathbf1
\\
&=
\mathbf1.
\end{aligned}
\]
Thus \(e^{tQ}\) is entrywise nonnegative and every row sums to one.

For a nonnegative matrix \(P\) satisfying
$
P\mathbf1=\mathbf1,
$
its induced \(\ell_\infty\) operator norm is
\[
\begin{aligned}
\|P\|_{\infty\to\infty}
&=
\max_i\sum_j|P_{ij}|
\\
&=
\max_i\sum_jP_{ij}
\\
&=
1.
\end{aligned}
\]
Consequently,
$
\|e^{tQ}\|_{\infty\to\infty}
=
1
\qquad
\text{for every }t\ge0.
$
Applying this observation to both \(Q_h\) and \(Q_g\), we obtain
$
\|e^{tQ_h}\|_{\infty\to\infty}
=
\|e^{tQ_g}\|_{\infty\to\infty}
=
1
\qquad
\text{for every }t\ge0.
$

We now compare the two exponentials. For arbitrary square matrices
\(A,B\), define
$
\Psi(s)
:=
e^{(1-s)A}e^{sB},
\qquad
s\in[0,1].
$
Differentiating with respect to \(s\) gives
\[
\begin{aligned}
\Psi'(s)
&=
-e^{(1-s)A}A e^{sB}
+
e^{(1-s)A}B e^{sB}
\\
&=
e^{(1-s)A}(B-A)e^{sB}.
\end{aligned}
\]
Here we used the fact that \(A\) commutes with its own exponential.
Integrating over \(s\in[0,1]\) yields
\[
\begin{aligned}
e^B-e^A
&=
\Psi(1)-\Psi(0)
\\
&=
\int_0^1
e^{(1-s)A}(B-A)e^{sB}
\,ds.
\end{aligned}
\]
Equivalently,
\[
e^A-e^B
=
\int_0^1
e^{(1-s)A}(A-B)e^{sB}
\,ds.
\]
Taking
$
A=Q_h,
\qquad
B=Q_g,
$
we obtain
\[
e^{Q_h}-e^{Q_g}
=
\int_0^1
e^{(1-s)Q_h}
(Q_h-Q_g)
e^{sQ_g}
\,ds.
\]
For every \(s\in[0,1]\),
$
(1-s)Q_h
$
and
$
sQ_g
$
are again Metzler matrices with zero row sums. Hence, by the property
proved above,
$
\|e^{(1-s)Q_h}\|_{\infty\to\infty}
=
1
$
and
$
\|e^{sQ_g}\|_{\infty\to\infty}
=
1.
$
Using the triangle inequality for the matrix integral and
submultiplicativity of the induced norm,
\[
\begin{aligned}
\|e^{Q_h}-e^{Q_g}\|
&\le
\int_0^1
\left\|
e^{(1-s)Q_h}
(Q_h-Q_g)
e^{sQ_g}
\right\|
\,ds
\\
&\le
\int_0^1
\|e^{(1-s)Q_h}\|
\|Q_h-Q_g\|
\|e^{sQ_g}\|
\,ds
\\
&=
\int_0^1
\|Q_h-Q_g\|
\,ds
\\
&=
\|Q_h-Q_g\|.
\end{aligned}
\]
The assumed Lipschitz continuity of the selector therefore gives
$
\|e^{Q_h}-e^{Q_g}\|
\le
K_Q\|h-g\|_\infty.
$

We can now estimate the nonlinear map
$
F(h)=e^{Q(h)}h.
$
Adding and subtracting \(e^{Q_h}g\) gives the exact decomposition
\[
\begin{aligned}
F(h)-F(g)
&=
e^{Q_h}h-e^{Q_g}g
\\
&=
e^{Q_h}h-e^{Q_h}g
+
e^{Q_h}g-e^{Q_g}g
\\
&=
e^{Q_h}(h-g)
+
\left(
e^{Q_h}-e^{Q_g}
\right)g.
\end{aligned}
\]
Using the ordinary triangle inequality,
\[
\begin{aligned}
\|F(h)-F(g)\|_\infty
&\le
\|e^{Q_h}(h-g)\|_\infty
+
\|
(e^{Q_h}-e^{Q_g})g
\|_\infty
\\
&\le
\|e^{Q_h}\|
\|h-g\|_\infty
+
\|e^{Q_h}-e^{Q_g}\|
\|g\|_\infty.
\end{aligned}
\]
Since
$
\|e^{Q_h}\|=1,
$
the selector estimate gives
$
\|e^{Q_h}-e^{Q_g}\|
\le
K_Q\|h-g\|_\infty,
$
and the assumption on \(K\) gives
$
\|g\|_\infty\le M.
$
Therefore
\[
\begin{aligned}
\|F(h)-F(g)\|_\infty
&\le
\|h-g\|_\infty
+
K_Q\|h-g\|_\infty\,M
\\
&=
(1+MK_Q)\|h-g\|_\infty.
\end{aligned}
\]
This proves the upper bound.

For the lower bound, we first establish a uniform lower singular-type
estimate for \(e^{Q_h}\) in the induced \(\ell_\infty\) norm. The matrix
\(e^{Q_h}\) is invertible, with inverse
$
(e^{Q_h})^{-1}
=
e^{-Q_h}.
$
Indeed, \(Q_h\) and \(-Q_h\) commute, so
$
e^{-Q_h}e^{Q_h}
=
e^0
=
I.
$
For any matrix \(C\), the power-series definition and
submultiplicativity give
\[
\begin{aligned}
\|e^C\|
&=
\left\|
\sum_{n=0}^{\infty}
\frac{C^n}{n!}
\right\|
\\
&\le
\sum_{n=0}^{\infty}
\frac{\|C^n\|}{n!}
\\
&\le
\sum_{n=0}^{\infty}
\frac{\|C\|^n}{n!}
\\
&=
e^{\|C\|}.
\end{aligned}
\]
Applying this to
$
C=-Q_h
$
gives
$
\|e^{-Q_h}\|
\le
e^{\|-Q_h\|}
=
e^{\|Q_h\|}.
$
By assumption,
$
\|Q_h\|
\le
b,
$
and therefore
$
\|e^{-Q_h}\|
\le
e^b.
$

Now let \(x\in\mathbb R^d\) be arbitrary. Since
$
x
=
e^{-Q_h}e^{Q_h}x,
$
submultiplicativity gives
\[
\begin{aligned}
\|x\|_\infty
&=
\|e^{-Q_h}e^{Q_h}x\|_\infty
\\
&\le
\|e^{-Q_h}\|
\|e^{Q_h}x\|_\infty
\\
&\le
e^b
\|e^{Q_h}x\|_\infty.
\end{aligned}
\]
Dividing by \(e^b>0\) yields
$
\|e^{Q_h}x\|_\infty
\ge
e^{-b}\|x\|_\infty.
$
Applying this estimate to
$
x=h-g
$
gives
$
\|e^{Q_h}(h-g)\|_\infty
\ge
e^{-b}\|h-g\|_\infty.
$

Returning to the exact decomposition of \(F(h)-F(g)\), the reverse triangle
inequality gives
\[
\begin{aligned}
\|F(h)-F(g)\|_\infty
&=
\left\|
e^{Q_h}(h-g)
+
(e^{Q_h}-e^{Q_g})g
\right\|_\infty
\\
&\ge
\|e^{Q_h}(h-g)\|_\infty
-
\|(e^{Q_h}-e^{Q_g})g\|_\infty
\\
&\ge
e^{-b}\|h-g\|_\infty
-
\|e^{Q_h}-e^{Q_g}\|
\|g\|_\infty
\\
&\ge
e^{-b}\|h-g\|_\infty
-
K_Q\|h-g\|_\infty\,M
\\
&=
(e^{-b}-MK_Q)
\|h-g\|_\infty.
\end{aligned}
\]
Together with the upper estimate, this proves
\[
\bigl(e^{-b}-MK_Q\bigr)\|h-g\|_\infty
\le
\|F(h)-F(g)\|_\infty
\le
\bigl(1+MK_Q\bigr)\|h-g\|_\infty
\]
for every \(h,g\in K\).

Finally, suppose that
$
MK_Q<e^{-b}.
$
Then
$
c
:=
e^{-b}-MK_Q
>
0.
$
The lower estimate becomes
\[
\|F(h)-F(g)\|_\infty
\ge
c\|h-g\|_\infty.
\]
In particular, if
$
F(h)=F(g),
$
then
$
0
\ge
c\|h-g\|_\infty,
$
and since \(c>0\), this forces
$
h=g.
$
Thus \(F\) is injective on \(K\). Moreover, for
$
y_1=F(h),
\qquad
y_2=F(g)
$
in \(F(K)\), the same lower bound gives
\[
\begin{aligned}
\|F^{-1}(y_1)-F^{-1}(y_2)\|_\infty
&=
\|h-g\|_\infty
\\
&\le
\frac{1}{e^{-b}-MK_Q}
\|y_1-y_2\|_\infty.
\end{aligned}
\]
Hence the inverse
$
F^{-1}:F(K)\to K
$
is Lipschitz, while the upper estimate shows that \(F\) itself is Lipschitz
with constant \(1+MK_Q\). Therefore \(F\) is bi-Lipschitz on \(K\)
whenever
$
MK_Q<e^{-b}.
$
\end{proof}

\subsection{Proof of Proposition~\ref{prop:layers-functional-ancestor-locality}}

\begin{proof}
For each \(i\in\{1,\ldots,d\}\), recall that
$
S_i
=
\{i\}
\cup
\{j:z_j\rightsquigarrow z_i\}.
$
We first record a basic nesting property of these ancestor sets that will be
used repeatedly. We claim that
$
j\in S_i
\quad\Longrightarrow\quad
S_j\subseteq S_i.
$
Indeed, if \(j=i\), then the conclusion is immediate because
$
S_j=S_i.
$
Suppose instead that \(j\neq i\). Since \(j\in S_i\), the definition of
\(S_i\) gives
$
z_j\rightsquigarrow z_i.
$
Now let
$
k\in S_j.
$
If \(k=j\), then
$
z_k=z_j\rightsquigarrow z_i,
$
so \(k\in S_i\). If \(k\neq j\), then membership \(k\in S_j\) implies
$
z_k\rightsquigarrow z_j.
$
By Definition~\ref{def:prelim-reachability}, there exist directed admissible
paths of positive length from \(z_k\) to \(z_j\) and from \(z_j\) to
\(z_i\). Concatenating these two directed paths gives a directed admissible
path from \(z_k\) to \(z_i\), and hence
$
z_k\rightsquigarrow z_i.
$
Thus again
$
k\in S_i.
$
Since \(k\in S_j\) was arbitrary, we have proved
$
S_j\subseteq S_i
\qquad
\text{whenever }j\in S_i.
$

Now fix an index \(i\), and let \(h,g\in\mathbb R^d\) satisfy
$
h_{S_i}=g_{S_i}.
$
We will prove
$
F_i(h)=F_i(g),
\qquad
F(x)=e^{Q(x)}x.
$
For brevity, write
$
Q_h:=Q(h),
\qquad
Q_g:=Q(g).
$
By assumption,
$
Q_h,Q_g\in\mathcal L_{\mathfrak S}.
$

We first show that the rows of \(Q_h\) and \(Q_g\) indexed by ancestors of
\(i\) coincide. Let
$
k\in S_i.
$
By the nesting property established above,
$
S_k\subseteq S_i.
$
Since
$
h_{S_i}=g_{S_i},
$
the two vectors agree in particular on every coordinate indexed by \(S_k\).
Hence
$
h_{S_k}=g_{S_k}.
$
Applying the assumed ancestor-local dependence condition to row \(k\) gives
$
Q(h)_{k,:}
=
Q(g)_{k,:}.
$
Equivalently,
$
(Q_h)_{k,:}
=
(Q_g)_{k,:}
\qquad
\text{for every }k\in S_i.
$

We also need the support structure of these rows. Since
$
Q_h,Q_g\in\mathcal L_{\mathfrak S},
$
Proposition~\ref{prop:operators-generator-support} implies that, for
\(a\neq b\),
$
(Q_h)_{ab}\neq0
\quad\Longrightarrow\quad
z_b\rightsquigarrow z_a
$
after one admissible step, and in particular
$
b\in S_a.
$
The same statement holds for \(Q_g\). More explicitly, if
$
(Q_h)_{ab}\neq0,
$
then either \(a=b\), in which case \(b=a\in S_a\), or \(a\neq b\), in which
case Proposition~\ref{prop:operators-generator-support} gives
$
(z_a,z_b)\in\mathcal E,
$
that is,
$
z_b\to z_a.
$
A one-step admissible edge is a directed path of positive length, so
$
z_b\rightsquigarrow z_a,
$
and again
$
b\in S_a.
$
Consequently,
$
(Q_h)_{ab}\neq0
\quad\Longrightarrow\quad
b\in S_a,
$
and likewise
$
(Q_g)_{ab}\neq0
\quad\Longrightarrow\quad
b\in S_a.
$
Thus each row \(a\) of either generator is supported only on coordinates
belonging to \(S_a\).

We now prove, for every integer \(n\ge0\), the two assertions
$
(Q_h^n)_{ij}=0
\qquad
\text{whenever }j\notin S_i,
$
and
$
(Q_h^n)_{i,:}
=
(Q_g^n)_{i,:}.
$
We proceed by induction on \(n\).

For \(n=0\),
$
Q_h^0=Q_g^0=I.
$
The \(i\)-th row of \(I\) has only one nonzero entry, namely the diagonal
entry at column \(i\). Since
$
i\in S_i,
$
we have
$
(I)_{ij}=0
\qquad
\text{for }j\notin S_i.
$
Moreover,
$
(Q_h^0)_{i,:}
=
I_{i,:}
=
(Q_g^0)_{i,:}.
$
Thus both assertions hold for \(n=0\).

Assume now that both assertions hold for some \(n\ge0\). For arbitrary
\(j\in\{1,\ldots,d\}\),
$
(Q_h^{n+1})_{ij}
=
\sum_{k=1}^d
(Q_h^n)_{ik}(Q_h)_{kj}.
$
By the induction hypothesis,
$
(Q_h^n)_{ik}=0
\qquad
\text{whenever }k\notin S_i.
$
Therefore all terms with \(k\notin S_i\) vanish, and the sum reduces to
$
(Q_h^{n+1})_{ij}
=
\sum_{k\in S_i}
(Q_h^n)_{ik}(Q_h)_{kj}.
$
Now suppose that
$
j\notin S_i.
$
For every \(k\in S_i\), the nesting property gives
$
S_k\subseteq S_i.
$
Hence
$
j\notin S_i
\quad\Longrightarrow\quad
j\notin S_k.
$
But row \(k\) of \(Q_h\) is supported only on \(S_k\). Therefore
$
(Q_h)_{kj}=0
\qquad
\text{for every }k\in S_i.
$
It follows that
$
(Q_h^{n+1})_{ij}
=
\sum_{k\in S_i}
(Q_h^n)_{ik}\,0
=
0.
$
Thus the support assertion holds at level \(n+1\). The same reasoning also
applies to \(Q_g\), so
$
(Q_g^{n+1})_{ij}=0
\qquad
\text{for }j\notin S_i.
$

We next compare the \(i\)-th rows of the two powers. For every \(j\),
\[
(Q_h^{n+1})_{ij}
=
\sum_{k\in S_i}
(Q_h^n)_{ik}(Q_h)_{kj}.
\]
By the induction hypothesis,
$
(Q_h^n)_{ik}
=
(Q_g^n)_{ik}
\qquad
\text{for every }k.
$
Moreover, as proved above,
$
(Q_h)_{k,:}
=
(Q_g)_{k,:}
\qquad
\text{for every }k\in S_i.
$
Hence, for every \(k\in S_i\) and every \(j\),
$
(Q_h)_{kj}
=
(Q_g)_{kj}.
$
Substituting these two equalities gives
\[
\begin{aligned}
(Q_h^{n+1})_{ij}
&=
\sum_{k\in S_i}
(Q_h^n)_{ik}(Q_h)_{kj}
\\
&=
\sum_{k\in S_i}
(Q_g^n)_{ik}(Q_g)_{kj}.
\end{aligned}
\]
Since the \(i\)-th row of \(Q_g^n\) also vanishes outside \(S_i\), the last
sum may be extended over all \(k\):
\[
\sum_{k\in S_i}
(Q_g^n)_{ik}(Q_g)_{kj}
=
\sum_{k=1}^d
(Q_g^n)_{ik}(Q_g)_{kj}.
\]
Therefore
$
(Q_h^{n+1})_{ij}
=
(Q_g^{n+1})_{ij}.
$
Since \(j\) was arbitrary,
$
(Q_h^{n+1})_{i,:}
=
(Q_g^{n+1})_{i,:}.
$
The induction is complete. Thus, for every \(n\ge0\),
$
(Q_h^n)_{ij}=0
\qquad
\text{for }j\notin S_i,
$
and
$
(Q_h^n)_{i,:}
=
(Q_g^n)_{i,:}.
$

We now pass from powers to the matrix exponential. The matrix exponential
series
$
e^{Q_h}
=
\sum_{n=0}^{\infty}
\frac{Q_h^n}{n!}
$
converges absolutely in any matrix norm, since
\[
\sum_{n=0}^{\infty}
\frac{\|Q_h^n\|}{n!}
\le
\sum_{n=0}^{\infty}
\frac{\|Q_h\|^n}{n!}
=
e^{\|Q_h\|}
<
\infty.
\]
The same holds for \(e^{Q_g}\). Hence the entries may be compared
term-by-term. For \(j\notin S_i\),
\[
\begin{aligned}
(e^{Q_h})_{ij}
&=
\sum_{n=0}^{\infty}
\frac{(Q_h^n)_{ij}}{n!}
\\
&=
0.
\end{aligned}
\]
Likewise,
$
(e^{Q_g})_{ij}=0
\qquad
\text{for }j\notin S_i.
$
Furthermore, for arbitrary \(j\),
\[
\begin{aligned}
(e^{Q_h})_{ij}
&=
\sum_{n=0}^{\infty}
\frac{(Q_h^n)_{ij}}{n!}
\\
&=
\sum_{n=0}^{\infty}
\frac{(Q_g^n)_{ij}}{n!}
\\
&=
(e^{Q_g})_{ij}.
\end{aligned}
\]
Thus
$
(e^{Q_h})_{i,:}
=
(e^{Q_g})_{i,:},
$
and this common row is supported entirely on \(S_i\).

We can now compare the \(i\)-th propagated outputs. By definition,
\[
F_i(h)
=
(e^{Q_h}h)_i
=
\sum_{j=1}^d
(e^{Q_h})_{ij}h_j.
\]
Since the \(i\)-th row of \(e^{Q_h}\) vanishes outside \(S_i\),
$
F_i(h)
=
\sum_{j\in S_i}
(e^{Q_h})_{ij}h_j.
$
Similarly,
$
F_i(g)
=
\sum_{j\in S_i}
(e^{Q_g})_{ij}g_j.
$
For every \(j\in S_i\), the assumption
$
h_{S_i}=g_{S_i}
$
gives
$
h_j=g_j,
$
and we have already established
$
(e^{Q_h})_{ij}
=
(e^{Q_g})_{ij}.
$
Therefore
\[
\begin{aligned}
F_i(h)
&=
\sum_{j\in S_i}
(e^{Q_h})_{ij}h_j
\\
&=
\sum_{j\in S_i}
(e^{Q_g})_{ij}g_j
\\
&=
F_i(g).
\end{aligned}
\]
This proves
$
h_{S_i}=g_{S_i}
\quad\Longrightarrow\quad
F_i(h)=F_i(g).
$

It remains to verify preservation under composition. Consider a finite
sequence of adaptive propagation maps
$
F^{(\ell)}(x)
=
e^{Q_\ell(x)}x,
\qquad
\ell=1,\ldots,m,
$
where, for every \(\ell\),
$
Q_\ell(x)\in\mathcal L_{\mathfrak S}
$
and the same ancestor-local row dependence condition holds:
$
x_{S_j}=y_{S_j}
\quad\Longrightarrow\quad
Q_\ell(x)_{j,:}
=
Q_\ell(y)_{j,:}
\qquad
\text{for every }j.
$
By the single-layer result just proved, each \(F^{(\ell)}\) satisfies
\[
x_{S_j}=y_{S_j}
\quad\Longrightarrow\quad
F^{(\ell)}_j(x)=F^{(\ell)}_j(y).
\]

We claim that each such layer preserves equality on the entire ancestor set
\(S_i\). Suppose
$
x_{S_i}=y_{S_i}.
$
Let
$
j\in S_i.
$
Since
$
S_j\subseteq S_i,
$
we have
$
x_{S_j}=y_{S_j}.
$
Applying the single-coordinate locality property at site \(j\) gives
$
F^{(\ell)}_j(x)
=
F^{(\ell)}_j(y).
$
Because this holds for every \(j\in S_i\),
$
F^{(\ell)}(x)_{S_i}
=
F^{(\ell)}(y)_{S_i}.
$

Now define recursively
$
x^{(0)}:=h,
\qquad
y^{(0)}:=g,
$
and
\[
x^{(\ell)}
:=
F^{(\ell)}(x^{(\ell-1)}),
\qquad
y^{(\ell)}
:=
F^{(\ell)}(y^{(\ell-1)}),
\qquad
\ell=1,\ldots,m.
\]
If
$
h_{S_i}=g_{S_i},
$
then
$
x^{(0)}_{S_i}
=
y^{(0)}_{S_i}.
$
The setwise preservation property just proved implies successively that
$
x^{(1)}_{S_i}
=
y^{(1)}_{S_i},
$
then
$
x^{(2)}_{S_i}
=
y^{(2)}_{S_i},
$
and, continuing inductively,
$
x^{(\ell)}_{S_i}
=
y^{(\ell)}_{S_i}
\qquad
\text{for every }\ell=0,\ldots,m.
$
In particular,
$
x^{(m)}_i
=
y^{(m)}_i.
$
Equivalently,
\[
\bigl(
F^{(m)}\circ\cdots\circ F^{(1)}
\bigr)_i(h)
=
\bigl(
F^{(m)}\circ\cdots\circ F^{(1)}
\bigr)_i(g).
\]
Thus the same ancestor-local functional dependence property is preserved
under arbitrary finite compositions of layers satisfying the same
ancestor-local condition.
\end{proof}

\subsection{Proof of Proposition~\ref{prop:layers-acyclic-balanced-obstruction}}

\begin{proof}
Let
$
L\in\mathcal L_{\mathfrak S}^{\mathrm{bal}}.
$
By Definition~\ref{def:layers-balanced-generator-family},
$
L\in\mathcal L_{\mathfrak S}
$
and
$
\mathbf1^\top L=0.
$
Since \(L\in\mathcal L_{\mathfrak S}\), Definition~\ref{def:operators-generator-family}
also gives
$
L_{ij}\ge0
\qquad
\text{for every }i\neq j,
$
and
$
L\mathbf1=0.
$
Thus \(L\) has nonnegative off-diagonal entries and satisfies both the
right- and left-conservation conditions.

We first prove the more general statement in the proposition, namely that
every positive off-diagonal edge of \(L\) belongs to a directed cycle of
the positive-rate graph of \(L\). Define the positive off-diagonal relation
\[
\mathcal E_L^{+}
:=
\left\{
(z_a,z_b):
a\neq b,\;
L_{ab}>0
\right\}.
\]
Under the receiver--source convention used throughout the paper,
$
(z_a,z_b)\in\mathcal E_L^{+}
$
represents a directed positive-rate edge
$
z_b\longrightarrow z_a.
$

Before considering a particular edge, we record the local balance identity
implied by the two conservation conditions. Fix
\(r\in\{1,\ldots,d\}\). From
$
L\mathbf1=0
$
we obtain
\[
0
=
(L\mathbf1)_r
=
\sum_{s=1}^dL_{rs}
=
L_{rr}
+
\sum_{s\neq r}L_{rs}.
\]
Hence
$
\sum_{s\neq r}L_{rs}
=
-L_{rr}.
$
On the other hand,
$
\mathbf1^\top L=0
$
implies, by looking at the \(r\)-th component of the row vector
\(\mathbf1^\top L\),
\[
0
=
(\mathbf1^\top L)_r
=
\sum_{s=1}^dL_{sr}
=
L_{rr}
+
\sum_{s\neq r}L_{sr},
\]
and therefore
$
\sum_{s\neq r}L_{sr}
=
-L_{rr}.
$
Combining the two identities gives
$
\sum_{s\neq r}L_{rs}
=
\sum_{s\neq r}L_{sr}.
$
With the receiver--source convention, the left-hand side is the total
positive off-diagonal rate entering \(z_r\), whereas the right-hand side is
the total positive off-diagonal rate leaving \(z_r\). Thus every vertex of
the positive-rate graph has equal total incoming and outgoing rate.

Now fix an arbitrary positive off-diagonal entry
$
L_{ij}>0,
\qquad
i\neq j.
$
This entry represents the directed positive-rate edge
$
z_j\longrightarrow z_i.
$
Let \(S\subseteq\{1,\ldots,d\}\) be the set of indices reachable from
\(i\) by directed positive-rate paths, including \(i\) itself. More
precisely,
\[
S
:=
\{i\}
\cup
\left\{
a:
\begin{array}{l}
\text{there exist }m\ge1\text{ and indices }
a_0,\ldots,a_m\\
\text{with }a_0=i,\ a_m=a,\text{ and }
L_{a_\ell a_{\ell-1}}>0\\
\text{for every }\ell=1,\ldots,m
\end{array}
\right\}.
\]
We claim that
$
j\in S.
$
Suppose, toward a contradiction, that
$
j\notin S.
$

By the definition of \(S\), no positive-rate edge can leave a source in
\(S\) and enter a vertex outside \(S\). Indeed, if
$
b\in S,
\qquad
a\notin S,
$
and
$
L_{ab}>0,
$
then there is a positive-rate edge
$
z_b\longrightarrow z_a.
$
Since \(z_b\) is reachable from \(z_i\), appending this edge would make
\(z_a\) reachable from \(z_i\), contradicting \(a\notin S\). Therefore
$
L_{ab}=0
\qquad
\text{whenever }b\in S,\ a\notin S.
$
Consequently, the total positive rate leaving \(S\) is zero:
$
\sum_{b\in S}
\sum_{a\notin S}
L_{ab}
=
0.
$

We now sum the vertexwise balance identity over all \(r\in S\). For each
\(r\in S\),
$
\sum_{s\neq r}L_{rs}
=
\sum_{s\neq r}L_{sr}.
$
Summing these equalities gives
$
\sum_{r\in S}
\sum_{s\neq r}L_{rs}
=
\sum_{r\in S}
\sum_{s\neq r}L_{sr}.
$
Split each side according to whether the second index lies inside or outside
\(S\). On the left,
\[
\begin{aligned}
\sum_{r\in S}
\sum_{s\neq r}L_{rs}
&=
\sum_{r\in S}
\sum_{\substack{s\in S\\s\neq r}}
L_{rs}
+
\sum_{r\in S}
\sum_{s\notin S}
L_{rs}.
\end{aligned}
\]
On the right,
\[
\begin{aligned}
\sum_{r\in S}
\sum_{s\neq r}L_{sr}
&=
\sum_{r\in S}
\sum_{\substack{s\in S\\s\neq r}}
L_{sr}
+
\sum_{r\in S}
\sum_{s\notin S}
L_{sr}.
\end{aligned}
\]
The two internal sums are equal, because each of them is simply the sum of
\(L_{ab}\) over all ordered pairs of distinct indices \(a,b\in S\):
$
\sum_{r\in S}
\sum_{\substack{s\in S\\s\neq r}}
L_{rs}
=
\sum_{r\in S}
\sum_{\substack{s\in S\\s\neq r}}
L_{sr}.
$
Cancelling these common internal contributions yields
$
\sum_{r\in S}
\sum_{s\notin S}
L_{rs}
=
\sum_{r\in S}
\sum_{s\notin S}
L_{sr}.
$
Equivalently, after renaming dummy indices,
$
\sum_{\substack{a\in S\\ b\notin S}}
L_{ab}
=
\sum_{\substack{a\notin S\\ b\in S}}
L_{ab}.
$
The left-hand side is the total positive rate entering \(S\) from its
complement, because \(b\notin S\) is the source and \(a\in S\) is the
receiver. The right-hand side is the total positive rate leaving \(S\),
because \(b\in S\) is the source and \(a\notin S\) is the receiver.

We have already proved that the latter quantity is zero:
$
\sum_{\substack{a\notin S\\ b\in S}}
L_{ab}
=
0.
$
Hence balance across the cut forces
$
\sum_{\substack{a\in S\\ b\notin S}}
L_{ab}
=
0.
$
However,
$
i\in S
$
by construction, while under the contradiction hypothesis
$
j\notin S.
$
The fixed positive edge satisfies
$
L_{ij}>0,
$
so \(L_{ij}\) is one of the terms in the preceding inflow sum. Since all
off-diagonal entries are nonnegative,
$
\sum_{\substack{a\in S\\ b\notin S}}
L_{ab}
\ge
L_{ij}
>
0,
$
contradicting the equality to zero. Therefore the assumption
\(j\notin S\) is impossible, and we conclude that
$
j\in S.
$

Thus there exists a directed positive-rate path from \(z_i\) to \(z_j\).
Among all such paths, choose one with the smallest number of edges:
\[
z_i=z_{a_0}
\longrightarrow
z_{a_1}
\longrightarrow
\cdots
\longrightarrow
z_{a_m}=z_j,
\]
with
$
L_{a_\ell a_{\ell-1}}>0
\qquad
\text{for }\ell=1,\ldots,m.
$
A shortest directed path cannot repeat a vertex: if some vertex occurred
twice, the segment between the two occurrences could be removed to produce
a shorter directed path with the same endpoints. Hence the vertices
$
z_{a_0},\ldots,z_{a_m}
$
are distinct. Appending the original positive edge
$
z_j\longrightarrow z_i
$
corresponding to \(L_{ij}>0\) produces
\[
z_i=z_{a_0}
\longrightarrow
z_{a_1}
\longrightarrow
\cdots
\longrightarrow
z_{a_m}=z_j
\longrightarrow
z_i,
\]
which is a directed cycle in the positive-rate graph and contains the
original edge \(z_j\to z_i\). Since the choice of the positive
off-diagonal entry \(L_{ij}\) was arbitrary, every positive off-diagonal
edge of a balanced generator belongs to a directed cycle of its
positive-rate graph.

We now impose the acyclicity assumption in the proposition. Suppose that the
directed graph induced by
$
\mathcal E_{\mathrm{off}}
=
\{(z_a,z_b)\in\mathcal E:a\neq b\}
$
is acyclic. Let
$
L\in\mathcal L_{\mathfrak S}^{\mathrm{bal}}.
$
For every \(a\neq b\), Proposition~\ref{prop:operators-generator-support}
gives
\[
L_{ab}\neq0
\quad\Longrightarrow\quad
(z_a,z_b)\in\mathcal E.
\]
Since \(a\neq b\), this implies
\[
L_{ab}>0
\quad\Longrightarrow\quad
(z_a,z_b)\in\mathcal E_{\mathrm{off}}.
\]
Hence the positive-rate graph of \(L\) is a directed subgraph of the
admissible off-diagonal graph.

If there existed any pair \(a\neq b\) with
$
L_{ab}>0,
$
the general result proved above would place this edge on a directed cycle of
the positive-rate graph. That cycle would also be a directed cycle in the
larger graph induced by \(\mathcal E_{\mathrm{off}}\), contradicting the
assumed acyclicity. Therefore no positive off-diagonal entry can exist:
$
L_{ab}\le0
\qquad
\text{for every }a\neq b.
$
But \(L\in\mathcal L_{\mathfrak S}\) is Metzler, so simultaneously
$
L_{ab}\ge0
\qquad
\text{for every }a\neq b.
$
Combining the two inequalities gives
$
L_{ab}=0
\qquad
\text{for every }a\neq b.
$

Finally, the generator condition
$
L\mathbf1=0
$
determines the diagonal entries. For every \(a\),
$
0
=
(L\mathbf1)_a
=
L_{aa}
+
\sum_{b\neq a}L_{ab}.
$
All off-diagonal entries have just been shown to vanish, so
$
L_{aa}
=
-\sum_{b\neq a}L_{ab}
=
0.
$
Thus every entry of \(L\) is zero:
$
L=0.
$
Since the zero matrix satisfies all defining conditions of the balanced
generator family,
$
0\in\mathcal L_{\mathfrak S}^{\mathrm{bal}},
$
and therefore
$
\mathcal L_{\mathfrak S}^{\mathrm{bal}}
=
\{0\}.
$

Consequently, any finite propagation obtained by exponentiating a balanced
generator from this descriptor is necessarily
$
e^{\tau L}
=
e^{\tau 0}
=
I
\qquad
\text{for every }\tau\ge0.
$
Hence every balanced propagation available under an acyclic admissible
off-diagonal relation is the identity.
\end{proof}

\subsection{Proof of Proposition~\ref{prop:layers-mass-conservation-interpretation}}

\begin{proof}
Fix an arbitrary
$
c\in\mathcal C,
$
and write
$
L:=L(c),
\qquad
\tau:=\tau(c),
\qquad
T:=T(c)=e^{\tau L}.
$
By Definition~\ref{def:layers-balanced-transport-layer},
$
L(c)\in
\mathcal L_{\mathfrak S}^{\mathrm{bal},(1)}
$
for every \(c\in\mathcal C\), and therefore
$
L
\in
\mathcal L_{\mathfrak S}^{\mathrm{bal}}
\subseteq
\mathcal L_{\mathfrak S}.
$
Consequently, \(L\) satisfies all generator constraints. In particular,
$
L_{ij}\ge0
\qquad
\text{for every }i\neq j,
$
and
$
L\mathbf1=0.
$
Moreover, because \(L\) belongs to the balanced generator family,
Definition~\ref{def:layers-balanced-generator-family} also gives
$
\mathbf1^\top L=0.
$
Thus \(L\) satisfies both right and left conservation.

Since
$
L\in\mathcal L_{\mathfrak S}
\qquad\text{and}\qquad
\tau\ge0,
$
Theorem~\ref{thm:properties-order-supnorm-stability}, applied to
$
T=e^{\tau L},
$
gives
$
T\ge0
$
entrywise and
$
T\mathbf1=\mathbf1.
$
Therefore
$
T_{ij}\ge0
\qquad
\text{for all }i,j,
$
which proves the first asserted property, and
$
T\mathbf1=\mathbf1,
$
which proves the second. Written componentwise, the latter identity is
$
\sum_{j=1}^d T_{ij}=1
\qquad
\text{for every }i=1,\ldots,d,
$
so every row of \(T\) sums to one.

We next establish the corresponding left-conservation identity. Since
$
\mathbf1^\top L=0,
$
we claim that
$
\mathbf1^\top L^n=0
\qquad
\text{for every integer }n\ge1.
$
For \(n=1\), this is precisely the balanced-generator condition. If the
identity holds for some \(n\ge1\), then
\[
\mathbf1^\top L^{n+1}
=
(\mathbf1^\top L^n)L
=
0^\top L
=
0^\top.
\]
Thus the claim follows by induction.

Using the absolutely convergent power-series representation of the matrix
exponential,
$
T
=
e^{\tau L}
=
I+
\sum_{n=1}^{\infty}
\frac{\tau^nL^n}{n!},
$
we may multiply from the left by \(\mathbf1^\top\) to obtain
\[
\begin{aligned}
\mathbf1^\top T
&=
\mathbf1^\top
\left(
I+
\sum_{n=1}^{\infty}
\frac{\tau^nL^n}{n!}
\right)
\\
&=
\mathbf1^\top I
+
\sum_{n=1}^{\infty}
\frac{\tau^n}{n!}
\mathbf1^\top L^n.
\end{aligned}
\]
Since
$
\mathbf1^\top I=\mathbf1^\top
$
and
$
\mathbf1^\top L^n=0
\qquad
\text{for every }n\ge1,
$
the entire positive-order part vanishes. Hence
$
\mathbf1^\top T
=
\mathbf1^\top.
$
Equivalently, for each column index \(j\),
$
\sum_{i=1}^dT_{ij}=1.
$
Thus every column of \(T\) also sums to one.

Combining
$
T\ge0,
\qquad
T\mathbf1=\mathbf1,
\qquad
\mathbf1^\top T=\mathbf1^\top,
$
shows that \(T\) is a nonnegative matrix whose rows and columns all sum to
one. Therefore
$
T
$
is doubly stochastic.

We now verify positivity preservation. Let
$
h\in\mathbb R^d
$
satisfy
$
h\ge0
$
componentwise. For every \(i\in\{1,\ldots,d\}\),
$
(Th)_i
=
\sum_{j=1}^dT_{ij}h_j.
$
Each factor in every summand is nonnegative:
$
T_{ij}\ge0
\qquad\text{and}\qquad
h_j\ge0.
$
Therefore
$
T_{ij}h_j\ge0
\qquad
\text{for every }i,j,
$
and hence
$
(Th)_i
=
\sum_{j=1}^dT_{ij}h_j
\ge0.
$
Since this holds for every \(i\),
$
Th\ge0.
$
Thus the balanced transport propagation preserves the nonnegative orthant.

For an arbitrary
$
h\in\mathbb R^d,
$
the left-conservation identity gives
\[
\begin{aligned}
\mathbf1^\top Th
&=
(\mathbf1^\top T)h
\\
&=
\mathbf1^\top h.
\end{aligned}
\]
Therefore the sum of the coordinates is preserved exactly under the
propagation:
$
\sum_{i=1}^d(Th)_i
=
\sum_{i=1}^dh_i.
$
In particular, when \(h\ge0\), the action of \(T\) can only redistribute
the existing nonnegative mass among sites; it cannot create negative values
and cannot alter the total mass.

Finally, suppose that
$
h\ge0
\qquad\text{and}\qquad
\mathbf1^\top h=1.
$
By Definition~\ref{def:layers-balanced-transport-layer},
$
\mathcal T_{\mathfrak S}(c,h)
=
T(c)h
=
Th.
$
The positivity-preservation result gives
$
\mathcal T_{\mathfrak S}(c,h)
=
Th
\ge0.
$
At the same time, mass conservation gives
\[
\begin{aligned}
\mathbf1^\top
\mathcal T_{\mathfrak S}(c,h)
&=
\mathbf1^\top Th
\\
&=
\mathbf1^\top h
\\
&=
1.
\end{aligned}
\]
Hence the output has nonnegative coordinates whose sum is one. Therefore
$
\mathcal T_{\mathfrak S}(c,h)
$
remains in the probability simplex.

Since \(c\in\mathcal C\) was arbitrary, all of the stated properties hold
for every context \(c\).
\end{proof}

\subsection{Proof of Proposition~\ref{prop:layers-global-mean-pooling}}

\begin{proof}
Let
$
H\in\mathbb R^{d\times p}
$
be arbitrary, and write
$
T:=T(H)\in\mathbb R^{d\times d}.
$
By assumption, \(T(H)\) is doubly stochastic for every \(H\). In particular,
its columns sum to one, which is equivalently expressed as
$
\mathbf1^\top T
=
\mathbf1^\top.
$
Although \(T\) may depend on \(H\), this identity holds pointwise for the
matrix \(T(H)\) associated with the particular input \(H\). No independence
between \(T\) and \(H\) is required for the algebraic argument below.

Recall that
$
g(H)
=
\frac1d\mathbf1^\top H.
$
Since
$
T(H)H\in\mathbb R^{d\times p},
$
applying the same pooling map gives
\[
g(T(H)H)
=
\frac1d
\mathbf1^\top
T(H)H.
\]
By associativity of matrix multiplication,
$
\mathbf1^\top T(H)H
=
\bigl(\mathbf1^\top T(H)\bigr)H.
$
Using the column-sum identity
$
\mathbf1^\top T(H)
=
\mathbf1^\top,
$
we therefore obtain
\[
\begin{aligned}
g(T(H)H)
&=
\frac1d
\bigl(\mathbf1^\top T(H)\bigr)H
\\
&=
\frac1d
\mathbf1^\top H
\\
&=
g(H).
\end{aligned}
\]
Thus
$
g(T(H)H)=g(H)
$
for every \(H\), even when the transport matrix is selected adaptively as a
function of \(H\).

Equivalently, subtracting \(g(H)\) from both sides gives
$
g(T(H)H)-g(H)=0.
$
By linearity of \(g\),
$
g(T(H)H-H)=0,
$
and hence
$
g\!\left(
[T(H)-I_d]H
\right)
=
0.
$
Thus the part of the transported representation that differs from the
identity branch is annihilated exactly by global average pooling.

We next consider the residual bottleneck
\[
\widetilde H
=
H+
[T(H)-I_d]
HW_{\rm in}W_{\rm out},
\]
where
$
W_{\rm in}\in\mathbb R^{p\times q},
\qquad
W_{\rm out}\in\mathbb R^{q\times p}.
$
The dimensions are compatible because
$
H W_{\rm in}
\in
\mathbb R^{d\times q},
$
$
H W_{\rm in}W_{\rm out}
\in
\mathbb R^{d\times p},
$
and therefore
$
[T(H)-I_d]
H W_{\rm in}W_{\rm out}
\in
\mathbb R^{d\times p}.
$
Applying \(g\) to \(\widetilde H\) and using its linearity gives
\[
\begin{aligned}
g(\widetilde H)
&=
\frac1d
\mathbf1^\top
\left[
H+
[T(H)-I_d]
H W_{\rm in}W_{\rm out}
\right]
\\
&=
\frac1d\mathbf1^\top H
+
\frac1d
\mathbf1^\top
[T(H)-I_d]
H W_{\rm in}W_{\rm out}.
\end{aligned}
\]
Since
$
\mathbf1^\top T(H)
=
\mathbf1^\top,
$
we have
\[
\begin{aligned}
\mathbf1^\top[T(H)-I_d]
&=
\mathbf1^\top T(H)
-
\mathbf1^\top I_d
\\
&=
\mathbf1^\top
-
\mathbf1^\top
\\
&=
0^\top.
\end{aligned}
\]
Consequently,
\[
\begin{aligned}
\mathbf1^\top
[T(H)-I_d]
H W_{\rm in}W_{\rm out}
&=
\left(
\mathbf1^\top[T(H)-I_d]
\right)
H W_{\rm in}W_{\rm out}
\\
&=
0^\top
H W_{\rm in}W_{\rm out}
\\
&=
0^\top.
\end{aligned}
\]
It follows that the entire residual correction disappears under global
average pooling:
$
g\!\left(
[T(H)-I_d]
H W_{\rm in}W_{\rm out}
\right)
=
0.
$
Therefore
\[
\begin{aligned}
g(\widetilde H)
&=
\frac1d\mathbf1^\top H
\\
&=
g(H).
\end{aligned}
\]
This proves the second asserted identity.

The preceding calculation also makes precise the statement about a balanced
transport branch followed immediately by global average pooling. The pooled
transport-induced correction is identically zero:
$
g\!\left(
T(H)H-H
\right)
=
0,
$
and, for the bottleneck realization,
$
g\!\left(
[T(H)-I_d]
H W_{\rm in}W_{\rm out}
\right)
=
0.
$
Thus global average pooling cannot observe any redistribution performed by
the balanced transport operator itself; it observes only the total
site-average, which double stochasticity preserves exactly.

Finally, consider parameters \(\vartheta\) that are used only to construct
the transport matrix, and write the corresponding adaptive matrix as
$
T_\vartheta(H).
$
Assume that \(H\), \(W_{\rm in}\), and \(W_{\rm out}\) do not otherwise
depend on \(\vartheta\) along the pooled path under consideration. Since
\(T_\vartheta(H)\) is doubly stochastic for every admissible
\(\vartheta\),
$
\mathbf1^\top T_\vartheta(H)
=
\mathbf1^\top
$
for every \(\vartheta\). Hence the pooled bottleneck correction
\[
C(\vartheta)
:=
g\!\left(
[T_\vartheta(H)-I_d]
H W_{\rm in}W_{\rm out}
\right)
\]
satisfies
$
C(\vartheta)=0
$
for every admissible \(\vartheta\). Therefore, wherever the parameterization
is differentiable,
$
D_\vartheta C(\vartheta)=0.
$
Equivalently, every partial derivative with respect to a component
\(\vartheta_a\) satisfies
$
\frac{\partial C(\vartheta)}
{\partial\vartheta_a}
=
0.
$

The same conclusion holds for the direct transported representation because
$
g(T_\vartheta(H)H)
=
g(H)
$
is independent of \(\vartheta\). Thus
$
D_\vartheta
g(T_\vartheta(H)H)
=
0
$
whenever the derivative exists.

More generally, suppose that a scalar loss receives information from these
parameters only through such a pooled representation. Writing this loss as
$
\mathcal J(\vartheta)
=
\ell\bigl(g(\widetilde H_\vartheta)\bigr),
$
the identity
$
g(\widetilde H_\vartheta)
=
g(H)
$
shows that the argument of \(\ell\) is constant with respect to
\(\vartheta\). Hence, wherever the relevant derivatives exist, the chain
rule gives
\[
D_\vartheta\mathcal J(\vartheta)
=
D\ell\bigl(g(H)\bigr)
\,D_\vartheta g(\widetilde H_\vartheta)
=
0.
\]
Therefore parameters used only to construct \(T(H)\) receive zero gradient
through this immediately pooled path, as claimed.
\end{proof}

\subsection{Proof of Proposition~\ref{prop:examples-canonical-generator-property}}

\begin{proof}
Let
$
K\in\mathcal A_{\mathfrak S}
$
satisfy
$
K_{ij}\ge0
\qquad
\text{for every }i\neq j.
$
Recall from Definition~\ref{def:examples-row-balanced-generator-map} that
\[
\Gamma(K)
=
\operatorname{Off}(K)
-
\operatorname{Diag}\!\bigl(
\operatorname{Off}(K)\mathbf1
\bigr).
\]
We first rewrite this expression in a form that makes its relation to the
interaction family explicit.

For every pair of distinct indices \(i\neq j\), the definition of
\(\operatorname{Off}(K)\) gives
$
\Gamma(K)_{ij}
=
\operatorname{Off}(K)_{ij}
=
K_{ij}.
$
For a diagonal entry, we have
$
\operatorname{Off}(K)_{ii}=0,
$
and therefore
\[
\begin{aligned}
\Gamma(K)_{ii}
&=
-
\bigl(
\operatorname{Off}(K)\mathbf1
\bigr)_i
\\
&=
-
\sum_{j=1}^d
\operatorname{Off}(K)_{ij}
\\
&=
-
\sum_{j\neq i}K_{ij}.
\end{aligned}
\]
On the other hand,
$
(K\mathbf1)_i
=
\sum_{j=1}^dK_{ij}
=
K_{ii}
+
\sum_{j\neq i}K_{ij},
$
so
\[
\begin{aligned}
K_{ii}-(K\mathbf1)_i
&=
K_{ii}
-
\left(
K_{ii}+\sum_{j\neq i}K_{ij}
\right)
\\
&=
-\sum_{j\neq i}K_{ij}
\\
&=
\Gamma(K)_{ii}.
\end{aligned}
\]
For \(i\neq j\),
$
\bigl(
K-\operatorname{Diag}(K\mathbf1)
\bigr)_{ij}
=
K_{ij}
=
\Gamma(K)_{ij},
$
while the preceding calculation shows equality of the diagonal entries as
well. Hence
$
\Gamma(K)
=
K-\operatorname{Diag}(K\mathbf1).
$

We now show that the row-balancing operation does not leave the interaction
family. Since
$
K\in\mathcal A_{\mathfrak S},
$
Definition~\ref{def:operators-interaction-family} guarantees the existence
of coefficients
$
\theta=(\theta_1,\ldots,\theta_R)^\top\in\mathbb R^R
$
and
$
\delta\in\mathbb R^d
$
such that
\[
K
=
\sum_{r=1}^R\theta_r\Phi_r
+
\operatorname{Diag}(\delta),
\]
with
$
\theta_r\ge0
\qquad
\text{for every }r\in\mathcal R_+.
$
Using the identity derived above,
\[
\begin{aligned}
\Gamma(K)
&=
K-\operatorname{Diag}(K\mathbf1)
\\
&=
\sum_{r=1}^R\theta_r\Phi_r
+
\operatorname{Diag}(\delta)
-
\operatorname{Diag}(K\mathbf1)
\\
&=
\sum_{r=1}^R\theta_r\Phi_r
+
\operatorname{Diag}
\bigl(
\delta-K\mathbf1
\bigr).
\end{aligned}
\]
The channel coefficients in this representation are exactly the same
coefficients \(\theta_r\) as those of \(K\). In particular,
$
\theta_r\ge0
\qquad
\text{for every }r\in\mathcal R_+.
$
Moreover,
$
\delta-K\mathbf1\in\mathbb R^d
$
is an admissible diagonal coefficient vector because diagonal self-action is
unrestricted in \(\mathcal A_{\mathfrak S}\). Therefore
$
\Gamma(K)\in\mathcal A_{\mathfrak S}.
$

It remains to verify the two additional generator constraints. For every
\(i\neq j\), we have already shown that
$
\Gamma(K)_{ij}=K_{ij}.
$
By the hypothesis of the proposition,
$
K_{ij}\ge0
\qquad
\text{for every }i\neq j,
$
and hence
$
\Gamma(K)_{ij}\ge0
\qquad
\text{for every }i\neq j.
$
Thus \(\Gamma(K)\) has nonnegative off-diagonal entries.

We next verify constant preservation. Using
$
\Gamma(K)
=
K-\operatorname{Diag}(K\mathbf1),
$
we obtain
\[
\begin{aligned}
\Gamma(K)\mathbf1
&=
\left[
K-\operatorname{Diag}(K\mathbf1)
\right]\mathbf1
\\
&=
K\mathbf1
-
\operatorname{Diag}(K\mathbf1)\mathbf1.
\end{aligned}
\]
For an arbitrary vector \(v\in\mathbb R^d\),
$
\operatorname{Diag}(v)\mathbf1=v,
$
because its \(i\)-th component is
\[
\begin{aligned}
\bigl(
\operatorname{Diag}(v)\mathbf1
\bigr)_i
&=
\sum_{j=1}^d
\operatorname{Diag}(v)_{ij}
\\
&=
v_i.
\end{aligned}
\]
Applying this identity with
$
v=K\mathbf1
$
gives
$
\operatorname{Diag}(K\mathbf1)\mathbf1
=
K\mathbf1.
$
Consequently,
\[
\begin{aligned}
\Gamma(K)\mathbf1
&=
K\mathbf1-K\mathbf1
\\
&=
0.
\end{aligned}
\]
We have therefore established
$
\Gamma(K)\in\mathcal A_{\mathfrak S},
$
$
\Gamma(K)_{ij}\ge0
\qquad
\text{for every }i\neq j,
$
and
$
\Gamma(K)\mathbf1=0.
$
By Definition~\ref{def:operators-generator-family}, these are exactly the
conditions required for membership in the geometry-induced generator
family. Hence
$
\Gamma(K)\in\mathcal L_{\mathfrak S}.
$

Suppose now that
$
\Gamma(K)\neq0.
$
We first show explicitly that its generator rate is strictly positive. Since
\(\Gamma(K)\in\mathcal L_{\mathfrak S}\),
$
\Gamma(K)_{ij}\ge0
\qquad
\text{for every }i\neq j.
$
Assume, toward a contradiction, that every off-diagonal entry were zero:
$
\Gamma(K)_{ij}=0
\qquad
\text{for every }i\neq j.
$
Since
$
\Gamma(K)\mathbf1=0,
$
the \(i\)-th row satisfies
\[
0
=
\sum_{j=1}^d\Gamma(K)_{ij}
=
\Gamma(K)_{ii}
+
\sum_{j\neq i}\Gamma(K)_{ij}.
\]
Under the assumed vanishing of all off-diagonal entries, this reduces to
$
\Gamma(K)_{ii}=0
\qquad
\text{for every }i.
$
Thus every entry of \(\Gamma(K)\) would vanish, giving
$
\Gamma(K)=0,
$
contrary to the present assumption. Hence there must exist at least one
pair \(i\neq j\) such that
$
\Gamma(K)_{ij}>0.
$

By Definition~\ref{def:operators-normalized-generator},
\[
\rho(\Gamma(K))
=
\frac1d
\sum_{i=1}^d
\sum_{j\neq i}
\Gamma(K)_{ij}.
\]
All terms in this sum are nonnegative, and at least one of them is strictly
positive. Therefore
$
\rho(\Gamma(K))>0.
$
The normalized matrix
$
\widehat L
:=
\frac{\Gamma(K)}
{\rho(\Gamma(K))}
$
is consequently well defined.

Since
$
\Gamma(K)\in\mathcal L_{\mathfrak S}
$
and
$
\frac1{\rho(\Gamma(K))}>0,
$
the cone property of \(\mathcal L_{\mathfrak S}\), established in
Proposition~\ref{prop:operators-generator-family-structure}, gives
$
\widehat L
\in
\mathcal L_{\mathfrak S}.
$
Moreover, using the definition of \(\rho\),
\[
\begin{aligned}
\rho(\widehat L)
&=
\frac1d
\sum_{i=1}^d
\sum_{j\neq i}
\widehat L_{ij}
\\
&=
\frac1d
\sum_{i=1}^d
\sum_{j\neq i}
\frac{\Gamma(K)_{ij}}
{\rho(\Gamma(K))}
\\
&=
\frac1{\rho(\Gamma(K))}
\left[
\frac1d
\sum_{i=1}^d
\sum_{j\neq i}
\Gamma(K)_{ij}
\right]
\\
&=
\frac{\rho(\Gamma(K))}
{\rho(\Gamma(K))}
\\
&=
1.
\end{aligned}
\]
Therefore
$
\widehat L
\in
\mathcal L_{\mathfrak S}^{(1)}.
$
Equivalently,
$
\Gamma(K)
=
\rho(\Gamma(K))\widehat L
$
is the scale--shape decomposition of the canonical generator.

Let now
$
\tau\ge0.
$
Using the preceding decomposition,
\[
\begin{aligned}
e^{\tau\Gamma(K)}
&=
e^{
\tau
\left[
\rho(\Gamma(K))\widehat L
\right]
}
\\
&=
e^{
\left[
\tau\rho(\Gamma(K))
\right]\widehat L
}.
\end{aligned}
\]
If
$
\tau>0,
$
then
$
\tau\rho(\Gamma(K))>0
$
because
$
\rho(\Gamma(K))>0.
$
Since
$
\widehat L
\in
\mathcal L_{\mathfrak S}^{(1)},
$
Definition~\ref{def:operators-propagation-family} gives
\[
e^{
\left[
\tau\rho(\Gamma(K))
\right]\widehat L
}
\in
\mathcal P_{\mathfrak S}
\bigl(
\tau\rho(\Gamma(K))
\bigr)
\subseteq
\mathcal P_{\mathfrak S}.
\]
Thus
$
e^{\tau\Gamma(K)}
\in
\mathcal P_{\mathfrak S}
\qquad
\text{for every }\tau>0.
$
When
$
\tau=0,
$
we instead have
$
e^{\tau\Gamma(K)}
=
e^0
=
I.
$
By Definition~\ref{def:operators-propagation-family},
$
\mathcal P_{\mathfrak S}(0)=\{I\},
$
and therefore
$
e^{0\cdot\Gamma(K)}
=
I
\in
\mathcal P_{\mathfrak S}.
$
Hence, whenever \(\Gamma(K)\neq0\),
$
e^{\tau\Gamma(K)}
=
e^{\,\tau\rho(\Gamma(K))\widehat L}
\in
\mathcal P_{\mathfrak S}
\qquad
\text{for every }\tau\ge0.
$

Finally, suppose that
$
\Gamma(K)=0.
$
Then, for every \(\tau\ge0\),
$
\tau\Gamma(K)=0,
$
and the power-series definition of the exponential gives
\[
\begin{aligned}
e^{\tau\Gamma(K)}
&=
e^0
\\
&=
I.
\end{aligned}
\]
Thus the degenerate row-balanced interaction produces only the identity
propagation, completing the proof.
\end{proof}

\subsection{Proof of Proposition~\ref{prop:examples-symmetric-yields-balanced}}

\begin{proof}
Assume the hypotheses of
Proposition~\ref{prop:examples-canonical-generator-property} and, in
addition, suppose that
$
K=K^\top.
$
By Proposition~\ref{prop:examples-canonical-generator-property}, the
row-balanced matrix
$
\Gamma(K)
$
already belongs to the generator family:
$
\Gamma(K)\in\mathcal L_{\mathfrak S}.$
In particular,
$
\Gamma(K)\mathbf1=0.
$
Thus it remains only to verify the additional left-conservation condition
required for membership in
\(\mathcal L_{\mathfrak S}^{\mathrm{bal}}\).

Recall from
Definition~\ref{def:examples-row-balanced-generator-map} that
\[
\Gamma(K)
=
\operatorname{Off}(K)
-
\operatorname{Diag}\!\bigl(
\operatorname{Off}(K)\mathbf1
\bigr).
\]
We first show that \(\operatorname{Off}(K)\) is symmetric. For arbitrary
indices \(i,j\), if \(i=j\), then by definition
$
\operatorname{Off}(K)_{ii}=0
=
\operatorname{Off}(K)_{ii}.
$
If \(i\neq j\), then
$
\operatorname{Off}(K)_{ij}
=
K_{ij}.
$
Since
$
K=K^\top,
$
we have
$
K_{ij}=K_{ji},
$
and hence
\[
\operatorname{Off}(K)_{ij}
=
K_{ij}
=
K_{ji}
=
\operatorname{Off}(K)_{ji}.
\]
Therefore
$
\operatorname{Off}(K)^\top
=
\operatorname{Off}(K).
$

Now define
$
v
:=
\operatorname{Off}(K)\mathbf1
\in\mathbb R^d.
$
The matrix
$
\operatorname{Diag}(v)
$
is diagonal, and every real diagonal matrix is symmetric. Indeed,
\[
\operatorname{Diag}(v)_{ij}
=
\begin{cases}
v_i, & i=j,\\
0, & i\neq j,
\end{cases}
\]
so
$
\operatorname{Diag}(v)^\top
=
\operatorname{Diag}(v).
$
Consequently,
\[
\begin{aligned}
\Gamma(K)^\top
&=
\left[
\operatorname{Off}(K)
-
\operatorname{Diag}\!\bigl(
\operatorname{Off}(K)\mathbf1
\bigr)
\right]^\top
\\
&=
\operatorname{Off}(K)^\top
-
\operatorname{Diag}\!\bigl(
\operatorname{Off}(K)\mathbf1
\bigr)^\top
\\
&=
\operatorname{Off}(K)
-
\operatorname{Diag}\!\bigl(
\operatorname{Off}(K)\mathbf1
\bigr)
\\
&=
\Gamma(K).
\end{aligned}
\]
Thus the row-balanced generator is itself symmetric:
$
\Gamma(K)=\Gamma(K)^\top.
$

We now combine this symmetry with the zero-row-sum property already obtained
from
\(\Gamma(K)\in\mathcal L_{\mathfrak S}\). Since
$
\Gamma(K)\mathbf1=0,
$
taking transposes gives
$
\bigl(\Gamma(K)\mathbf1\bigr)^\top
=
0^\top.
$
Using
$
(AB)^\top=B^\top A^\top,
$
the left-hand side becomes
$
\bigl(\Gamma(K)\mathbf1\bigr)^\top
=
\mathbf1^\top\Gamma(K)^\top.
$
Since \(\Gamma(K)\) is symmetric,
$
\Gamma(K)^\top=\Gamma(K),
$
and therefore
$
\mathbf1^\top\Gamma(K)
=
0^\top.
$
Equivalently,
$
\mathbf1^\top\Gamma(K)=0.
$

The same conclusion can also be seen directly at the level of row and
column sums. Symmetry gives
$
\Gamma(K)_{ij}
=
\Gamma(K)_{ji},
$
and hence, for every column \(j\),
\[
\begin{aligned}
\sum_{i=1}^d\Gamma(K)_{ij}
&=
\sum_{i=1}^d\Gamma(K)_{ji}
\\
&=
\bigl(\Gamma(K)\mathbf1\bigr)_j
\\
&=
0.
\end{aligned}
\]
Thus every column sum vanishes whenever every row sum vanishes.

We have therefore established both
$
\Gamma(K)\in\mathcal L_{\mathfrak S}
$
and
$
\mathbf1^\top\Gamma(K)=0.
$
By Definition~\ref{def:layers-balanced-generator-family},
$
\mathcal L_{\mathfrak S}^{\mathrm{bal}}
=
\left\{
L\in\mathcal L_{\mathfrak S}
\;\middle|\;
\mathbf1^\top L=0
\right\}.
$
It follows immediately that
$
\Gamma(K)
\in
\mathcal L_{\mathfrak S}^{\mathrm{bal}},
$
as claimed.
\end{proof}

\subsection{Proof of Proposition~\ref{prop:examples-local-kernel-in-interaction-family}}

\begin{proof}
Recall that the local geometry descriptor is
$
\mathfrak S_{\mathrm{loc}}
=
\bigl(
X,
\mathcal E_{\mathrm{loc}},
\Psi_{\mathrm{loc}},
\{\phi_{\mathrm{loc}}\},
\{1\}
\bigr),
$
with the unique geometry channel
$
\phi_{\mathrm{loc}}(z_i,z_j)
=
\kappa_\varepsilon(z_i,z_j)
=
\exp\!\left(
-\frac{
\|\iota(z_i)-\iota(z_j)\|^2
}{
\varepsilon^2
}
\right).
$
Let
$
\Phi_{\mathrm{loc}}\in\mathbb R^{d\times d}
$
denote the masked channel matrix induced by this unique channel. By
Definition~\ref{def:prelim-geometry-descriptor}, the masked realization of a
geometry channel is obtained by multiplying the channel value by the
indicator of the admissibility relation. Hence, for every pair
\(i,j\in\{1,\ldots,d\}\),
$
(\Phi_{\mathrm{loc}})_{ij}
=
\mathbf1_{\mathcal E_{\mathrm{loc}}}(z_i,z_j)
\phi_{\mathrm{loc}}(z_i,z_j).
$
Substituting the definition of the local channel gives
\[
\begin{aligned}
(\Phi_{\mathrm{loc}})_{ij}
&=
\mathbf1_{\mathcal E_{\mathrm{loc}}}(z_i,z_j)
\kappa_\varepsilon(z_i,z_j)
\\
&=
\mathbf1_{\mathcal E_{\mathrm{loc}}}(z_i,z_j)
\exp\!\left(
-\frac{
\|\iota(z_i)-\iota(z_j)\|^2
}{
\varepsilon^2
}
\right).
\end{aligned}
\]
On the other hand, by
Definition~\ref{def:examples-local-kernel-interaction},
the local kernel interaction matrix is defined entrywise by
$
(K^{\mathrm{loc}}_\varepsilon)_{ij}
=
\mathbf1_{\mathcal E_{\mathrm{loc}}}(z_i,z_j)
\kappa_\varepsilon(z_i,z_j).
$
Therefore, for every \(i,j\),
$
(K^{\mathrm{loc}}_\varepsilon)_{ij}
=
(\Phi_{\mathrm{loc}})_{ij}.
$
Equality of all matrix entries implies
$
K^{\mathrm{loc}}_\varepsilon
=
\Phi_{\mathrm{loc}}.
$
Thus \(K^{\mathrm{loc}}_\varepsilon\) is exactly, rather than merely
support-equivalent to, the masked realization of the unique channel of
\(\mathfrak S_{\mathrm{loc}}\).

It remains to verify explicitly that this matrix belongs to the corresponding
interaction family. Since \(\mathfrak S_{\mathrm{loc}}\) contains only one
geometry channel, Definition~\ref{def:operators-interaction-family} gives
\[
\mathcal A_{\mathfrak S_{\mathrm{loc}}}
=
\left\{
\theta_1\Phi_{\mathrm{loc}}
+
\operatorname{Diag}(\delta)
\;\middle|\;
\theta_1\in\mathbb R,\;
\delta\in\mathbb R^d,\;
\theta_1\ge0
\right\},
\]
because the positive-channel index set of
\(\mathfrak S_{\mathrm{loc}}\) is
$
\mathcal R_+=\{1\}.
$
The placement of this channel in the positive-channel set is consistent with
its sign, since
$
\kappa_\varepsilon(z_i,z_j)
=
\exp\!\left(
-\frac{
\|\iota(z_i)-\iota(z_j)\|^2
}{
\varepsilon^2
}
\right)
>
0,
$
and therefore
$
(\Phi_{\mathrm{loc}})_{ij}
=
\mathbf1_{\mathcal E_{\mathrm{loc}}}(z_i,z_j)
\kappa_\varepsilon(z_i,z_j)
\ge0
$
for every \(i,j\), in particular for every \(i\neq j\).

Now choose
$
\theta_1:=1
$
and
$
\delta:=0\in\mathbb R^d.
$
The required coefficient constraint is satisfied because
$
\theta_1=1\ge0.
$
With this choice,
\[
\begin{aligned}
\theta_1\Phi_{\mathrm{loc}}
+
\operatorname{Diag}(\delta)
&=
1\cdot\Phi_{\mathrm{loc}}
+
\operatorname{Diag}(0)
\\
&=
\Phi_{\mathrm{loc}}
\\
&=
K^{\mathrm{loc}}_\varepsilon.
\end{aligned}
\]
Hence \(K^{\mathrm{loc}}_\varepsilon\) admits a valid representation in the
geometry-induced interaction family, and therefore
$
K^{\mathrm{loc}}_\varepsilon
\in
\mathcal A_{\mathfrak S_{\mathrm{loc}}},
$
as claimed.
\end{proof}

\subsection{Proof of Proposition~\ref{prop:examples-cnn-exact-geometry-induced}}

\begin{proof}
Recall the CNN-style descriptor introduced above. The latent sites carry grid
positions
$
p_i\in\mathbb Z^n,
$
a finite stencil
$
\mathcal S\subset\mathbb Z^n
$
is fixed, and, for each offset
\(\Delta\in\mathcal S\), the corresponding geometry channel is
$
\phi_\Delta(z_i,z_j)
=
\mathbf1\{p_j-p_i=\Delta\}.
$
The admissibility relation is
$
(z_i,z_j)\in\mathcal E_{\mathrm{cnn}}
\quad\Longleftrightarrow\quad
p_j-p_i\in\mathcal S.
$
Let \(\Phi_\Delta\) denote the masked channel matrix induced by
\(\phi_\Delta\). By
Definition~\ref{def:prelim-geometry-descriptor},
$
(\Phi_\Delta)_{ij}
=
\mathbf1_{\mathcal E_{\mathrm{cnn}}}(z_i,z_j)
\phi_\Delta(z_i,z_j).
$
Substituting the definitions of the admissibility relation and the offset
channel gives
\[
(\Phi_\Delta)_{ij}
=
\mathbf1\{p_j-p_i\in\mathcal S\}
\mathbf1\{p_j-p_i=\Delta\}.
\]
Because
$
\Delta\in\mathcal S,
$
the event
$
p_j-p_i=\Delta
$
already implies
$
p_j-p_i\in\mathcal S.
$
Hence
\[
\mathbf1\{p_j-p_i\in\mathcal S\}
\mathbf1\{p_j-p_i=\Delta\}
=
\mathbf1\{p_j-p_i=\Delta\},
\]
and therefore
$
(\Phi_\Delta)_{ij}
=
\mathbf1\{p_j-p_i=\Delta\}.
$
Thus \(\Phi_\Delta\) is exactly the indicator matrix of the ordered site
pairs having relative offset \(\Delta\). In particular, if two ordered
receiver--source pairs satisfy
$
p_j-p_i
=
p_{j'}-p_{i'}
=
\Delta,
$
then
$
(\Phi_\Delta)_{ij}
=
(\Phi_\Delta)_{i'j'}
=
1,
$
so the same scalar coefficient multiplying \(\Phi_\Delta\) is shared across
all occurrences of the same relative offset.

Now let \(A^{\mathrm{cnn}}\) be any CNN-style stencil interaction matrix in
the sense of
Definition~\ref{def:examples-cnn-stencil-matrix}. Thus
\[
A^{\mathrm{cnn}}
=
\sum_{\Delta\in\mathcal S}
w_\Delta\Phi_\Delta
+
\operatorname{Diag}(\delta),
\]
where
$
w_\Delta\in\mathbb R
\qquad
\text{for every }\Delta\in\mathcal S,
$
and
$
\delta\in\mathbb R^d.
$
For the unrestricted signed stencil realization, the descriptor was defined
with
$
\mathcal R_+=\varnothing.
$
Consequently, Definition~\ref{def:operators-interaction-family} imposes no
sign constraint on any channel coefficient. More precisely, after indexing
the finite collection of offsets in \(\mathcal S\) as the channel
dictionary of \(\mathfrak S_{\mathrm{cnn}}\), its interaction family is
\[
\mathcal A_{\mathfrak S_{\mathrm{cnn}}}
=
\left\{
\sum_{\Delta\in\mathcal S}
\theta_\Delta\Phi_\Delta
+
\operatorname{Diag}(\eta)
\;\middle|\;
\theta_\Delta\in\mathbb R,\;
\eta\in\mathbb R^d
\right\}.
\]
Taking
$
\theta_\Delta:=w_\Delta
\qquad
\text{for every }\Delta\in\mathcal S
$
and
$
\eta:=\delta
$
gives
\[
\begin{aligned}
\sum_{\Delta\in\mathcal S}
\theta_\Delta\Phi_\Delta
+
\operatorname{Diag}(\eta)
&=
\sum_{\Delta\in\mathcal S}
w_\Delta\Phi_\Delta
+
\operatorname{Diag}(\delta)
\\
&=
A^{\mathrm{cnn}}.
\end{aligned}
\]
Thus \(A^{\mathrm{cnn}}\) has exactly the defining form of an element of the
geometry-induced interaction family, and hence
$
A^{\mathrm{cnn}}
\in
\mathcal A_{\mathfrak S_{\mathrm{cnn}}}.
$

It is useful to make the entrywise form of this inclusion explicit. For
\(i\neq j\),
\[
\begin{aligned}
A^{\mathrm{cnn}}_{ij}
&=
\sum_{\Delta\in\mathcal S}
w_\Delta(\Phi_\Delta)_{ij}
\\
&=
\sum_{\Delta\in\mathcal S}
w_\Delta
\mathbf1\{p_j-p_i=\Delta\}.
\end{aligned}
\]
For a fixed pair \((i,j)\), at most one offset
\(\Delta\in\mathcal S\) can satisfy
$
p_j-p_i=\Delta.
$
Therefore, whenever
$
p_j-p_i\in\mathcal S,
$
we have
$
A^{\mathrm{cnn}}_{ij}
=
w_{p_j-p_i},
$
whereas if
$
p_j-p_i\notin\mathcal S,
$
then
$
A^{\mathrm{cnn}}_{ij}=0.
$
Thus the cross-site interaction weight depends only on the relative grid
offset and not on the absolute site indices. This is precisely the
translation-shared finite-stencil coupling encoded by the channel
dictionary.

The diagonal component of the geometry-induced family is more general.
Indeed,
\[
\operatorname{Diag}(\delta)
=
\begin{pmatrix}
\delta_1 & & 0\\
& \ddots & \\
0 & & \delta_d
\end{pmatrix},
\]
and Definition~\ref{def:operators-interaction-family} places no tying
constraint on the coordinates of \(\delta\). Hence the full
geometry-induced interaction family permits site-dependent diagonal
self-action:
$
\delta_i
\neq
\delta_j
$
is allowed for distinct sites \(i\) and \(j\).

Exact translation-shared scalar convolution is recovered by imposing the
additional tying condition
$
\delta
=
\delta_0\mathbf1
$
for some
$
\delta_0\in\mathbb R.
$
Under this condition,
$
\operatorname{Diag}(\delta)
=
\operatorname{Diag}(\delta_0\mathbf1)
=
\delta_0 I_d.
$
Hence
\[
A^{\mathrm{cnn}}
=
\sum_{\Delta\in\mathcal S}
w_\Delta\Phi_\Delta
+
\delta_0I_d,
\]
so both the cross-site stencil coefficients and the self-interaction
coefficient are shared across grid locations.

The same tied self-interaction may equivalently be represented as a
zero-offset stencil channel. To make this statement precise, suppose that
the zero offset
$
0\in\mathbb Z^n
$
is included in the offset dictionary, either because
\(0\in\mathcal S\) already or because the descriptor is augmented by this
single channel. Define
$
\phi_0(z_i,z_j)
=
\mathbf1\{p_j-p_i=0\}.
$
For the usual grid indexing in which distinct latent sites have distinct
grid positions, the equality
$
p_j-p_i=0
$
holds if and only if
$
i=j.
$
Its masked channel matrix therefore satisfies
$
(\Phi_0)_{ij}
=
\mathbf1\{i=j\},
$
and hence
$
\Phi_0=I_d.
$
Consequently,
$
\delta_0I_d
=
\delta_0\Phi_0.
$
Thus
$
\sum_{\Delta\in\mathcal S}
w_\Delta\Phi_\Delta
+
\delta_0I_d
$
may equivalently be written by absorbing the common diagonal coefficient
into the shared zero-offset weight. For example, when the zero-offset
channel is listed separately from the nonzero stencil offsets,
\[
A^{\mathrm{cnn}}
=
\delta_0\Phi_0
+
\sum_{\substack{\Delta\in\mathcal S\\\Delta\neq0}}
w_\Delta\Phi_\Delta.
\]
The coefficient \(\delta_0\) is then simply the convolutional weight
associated with zero displacement.

Therefore the unrestricted family
\(\mathcal A_{\mathfrak S_{\mathrm{cnn}}}\) contains all CNN-style stencil
matrices defined above, while exact translation-shared scalar convolution
is obtained by the additional diagonal tying
$
\delta=\delta_0\mathbf1,
$
equivalently by representing that common self-action through a shared
zero-offset channel.
\end{proof}

\subsection{Proof of Proposition~\ref{prop:examples-graph-message-passing-geometry-induced}}

\begin{proof}
Recall that the admissible ordered receiver--source relation of the graph
descriptor is partitioned into finitely many pairwise disjoint edge classes,
$
\mathcal E_{\mathrm{graph}}
=
\bigsqcup_{t=1}^T\mathcal E_t,
$
and that the channel associated with the \(t\)-th edge class is
$
\phi_t(z_i,z_j)
=
\mathbf1\{(z_i,z_j)\in\mathcal E_t\}.
$
Let \(\Phi_t\in\mathbb R^{d\times d}\) denote the corresponding masked
channel matrix. By Definition~\ref{def:prelim-geometry-descriptor}, its
entries are
$
(\Phi_t)_{ij}
=
\mathbf1_{\mathcal E_{\mathrm{graph}}}(z_i,z_j)
\phi_t(z_i,z_j).
$
Substituting the definition of \(\phi_t\) gives
\[
(\Phi_t)_{ij}
=
\mathbf1_{\mathcal E_{\mathrm{graph}}}(z_i,z_j)
\mathbf1\{(z_i,z_j)\in\mathcal E_t\}.
\]
Since
$
\mathcal E_t\subseteq\mathcal E_{\mathrm{graph}},
$
membership in \(\mathcal E_t\) already implies membership in
\(\mathcal E_{\mathrm{graph}}\). Hence, for every pair \(i,j\),
\[
\mathbf1_{\mathcal E_{\mathrm{graph}}}(z_i,z_j)
\mathbf1\{(z_i,z_j)\in\mathcal E_t\}
=
\mathbf1\{(z_i,z_j)\in\mathcal E_t\},
\]
and therefore
$
(\Phi_t)_{ij}
=
\mathbf1\{(z_i,z_j)\in\mathcal E_t\}.
$
Thus \(\Phi_t\) is exactly the indicator matrix of the \(t\)-th ordered
edge class; the masking operation introduces no additional coefficient and
does not alter any edge belonging to that class.

Now let \(A^{\mathrm{graph}}\) be any matrix in
Definition~\ref{def:examples-graph-message-passing-matrix}. By definition,
$
A^{\mathrm{graph}}
=
\sum_{t=1}^T
\theta_t\Phi_t
+
\operatorname{Diag}(\delta),
\qquad
\theta_t\in\mathbb R,
\qquad
\delta\in\mathbb R^d.
$
For the ordinary signed graph-message-passing descriptor considered here,
the positive-channel index set is
$
\mathcal R_+=\varnothing.
$
Consequently, Definition~\ref{def:operators-interaction-family} specializes
to
\[
\mathcal A_{\mathfrak S_{\mathrm{graph}}}
=
\left\{
\sum_{t=1}^T
\vartheta_t\Phi_t
+
\operatorname{Diag}(\eta)
\;\middle|\;
\vartheta_t\in\mathbb R,\;
\eta\in\mathbb R^d
\right\}.
\]
Indeed, because \(\mathcal R_+\) is empty, there is no sign constraint on
any channel coefficient. Choosing
$
\vartheta_t:=\theta_t,
\qquad
t=1,\ldots,T,
$
and
$
\eta:=\delta
$
gives
\[
\begin{aligned}
\sum_{t=1}^T
\vartheta_t\Phi_t
+
\operatorname{Diag}(\eta)
&=
\sum_{t=1}^T
\theta_t\Phi_t
+
\operatorname{Diag}(\delta)
\\
&=
A^{\mathrm{graph}}.
\end{aligned}
\]
Hence \(A^{\mathrm{graph}}\) has precisely the defining form of a
geometry-induced interaction operator, and therefore
$
A^{\mathrm{graph}}
\in
\mathcal A_{\mathfrak S_{\mathrm{graph}}}.
$

The entrywise form makes the message-passing interpretation explicit.
Consider first a pair of distinct sites \(i\neq j\). Since the diagonal
self-action contributes nothing to an off-diagonal entry,
$
A^{\mathrm{graph}}_{ij}
=
\sum_{t=1}^T
\theta_t(\Phi_t)_{ij}.
$
Using the explicit channel formula derived above,
\[
A^{\mathrm{graph}}_{ij}
=
\sum_{t=1}^T
\theta_t
\mathbf1\{(z_i,z_j)\in\mathcal E_t\}.
\]
Because
$
\mathcal E_{\mathrm{graph}}
=
\bigsqcup_{t=1}^T\mathcal E_t
$
is a disjoint partition, an admissible ordered pair belongs to exactly one
edge class. Hence, if
$
(z_i,z_j)\in\mathcal E_t,
$
then
$
(\Phi_t)_{ij}=1
$
and
$
(\Phi_s)_{ij}=0
\qquad
\text{for every }s\neq t.
$
It follows that
$
A^{\mathrm{graph}}_{ij}
=
\theta_t.
$
Thus all ordered edges belonging to the same class \(\mathcal E_t\) share
the same scalar interaction coefficient \(\theta_t\). On the other hand,
if
$
(z_i,z_j)\notin\mathcal E_{\mathrm{graph}},
$
then the pair belongs to none of the classes \(\mathcal E_t\), so
$
(\Phi_t)_{ij}=0
\qquad
\text{for every }t,
$
and therefore
$
A^{\mathrm{graph}}_{ij}=0.
$
Accordingly, the off-diagonal action of \(A^{\mathrm{graph}}\) is supported
exactly on the admissible graph relation, with edge-class sharing determined
by the partition
\(\{\mathcal E_t\}_{t=1}^T\).

The use of ordered receiver--source pairs also accounts directly for
directionality. An entry
$
(\Phi_t)_{ij}=1
$
means that the ordered pair
$
(z_i,z_j)
$
belongs to \(\mathcal E_t\), so under the receiver--source convention the
corresponding interaction transfers information from source \(z_j\) to
receiver \(z_i\). There is no requirement that the reversed pair
$
(z_j,z_i)
$
belong to the same class, or even that it be admissible. Thus a directed
graph is represented without any modification of the construction.
For an undirected graph, the descriptor may instead contain both ordered
orientations of each undirected edge, with their classes tied or separated
according to the desired message-passing architecture.

The diagonal term has the complementary interpretation
$
\operatorname{Diag}(\delta)
=
\sum_{i=1}^d\delta_iE_{ii}.
$
It acts independently at each site and therefore represents sitewise
self-action rather than cross-site message transfer. Since
\(\delta\in\mathbb R^d\) is unrestricted in
\(\mathcal A_{\mathfrak S_{\mathrm{graph}}}\), the geometry-induced family
allows arbitrary sitewise scalar self-action in addition to the typed or
directional edge aggregation described above. Hence
\(A^{\mathrm{graph}}\) is exactly the site-level aggregation skeleton of
the corresponding scalar graph message-passing operator.

For completeness, consider now the multi-channel situation in which every
site carries a feature vector
$
x_i\in\mathbb R^c.
$
Stack the site features into
\[
x
=
\begin{pmatrix}
x_1\\
\vdots\\
x_d
\end{pmatrix}
\in\mathbb R^{dc}.
\]
Suppose that a message sent along an edge of class \(t\) is additionally
transformed by a feature-mixing matrix
$
W_t\in\mathbb R^{c\times c}.
$
The corresponding lifted cross-site operator is
$
\mathcal M_{\mathrm{cross}}
:=
\sum_{t=1}^T
\Phi_t\otimes W_t.
$
Its \((i,j)\)-block is
\[
\begin{aligned}
(\mathcal M_{\mathrm{cross}})_{ij}
&=
\sum_{t=1}^T
(\Phi_t)_{ij}W_t
\\
&=
\sum_{t=1}^T
\mathbf1\{(z_i,z_j)\in\mathcal E_t\}W_t.
\end{aligned}
\]
If
$
(z_i,z_j)\in\mathcal E_t,
$
the disjointness of the edge classes gives
$
(\mathcal M_{\mathrm{cross}})_{ij}
=
W_t.
$
If
$
(z_i,z_j)\notin\mathcal E_{\mathrm{graph}},
$
then
$
(\mathcal M_{\mathrm{cross}})_{ij}
=
0.
$
Thus the block operator applies exactly the feature transform associated
with the type or direction of the corresponding graph edge.

Allowing arbitrary sitewise feature self-actions
$
D_i\in\mathbb R^{c\times c}
$
gives the full lifted matrix
\[
\mathcal M
=
\sum_{t=1}^T
\Phi_t\otimes W_t
+
\sum_{i=1}^d
E_{ii}\otimes D_i.
\]
This matrix is an explicit special case of
Definition~\ref{def:operators-block-valued-interaction-family}. To see this
directly, take
$
M=T,
\qquad
G_s:=W_s,
\qquad
s=1,\ldots,T,
$
and, for the ordinary sign-indefinite hidden-feature setting, take
$
\Lambda_+=\varnothing.
$
Choose the block-valued interaction coefficients
$
\beta_{ts}
:=
\begin{cases}
1, & t=s,\\
0, & t\neq s.
\end{cases}
$
Then
\[
\begin{aligned}
\sum_{t=1}^T\sum_{s=1}^T
\beta_{ts}
(\Phi_t\otimes G_s)
&=
\sum_{t=1}^T
\Phi_t\otimes G_t
\\
&=
\sum_{t=1}^T
\Phi_t\otimes W_t.
\end{aligned}
\]
Consequently,
\[
\begin{aligned}
&
\sum_{t=1}^T\sum_{s=1}^T
\beta_{ts}
(\Phi_t\otimes G_s)
+
\sum_{i=1}^dE_{ii}\otimes D_i
\\
&\qquad=
\sum_{t=1}^T
\Phi_t\otimes W_t
+
\sum_{i=1}^dE_{ii}\otimes D_i
\\
&\qquad=
\mathcal M,
\end{aligned}
\]
so
$
\mathcal M
\in
\mathcal A^{\mathrm{blk}}_{\mathfrak S_{\mathrm{graph}},\mathcal G}.
$
Hence the scalar matrices \(A^{\mathrm{graph}}\) describe exactly the
site-level typed or directional aggregation pattern, while the
block-valued lifting of
Appendix~\ref{app:operators-block-lifting} supplies the corresponding
feature transforms and sitewise feature self-actions. This proves both
claims of the proposition.
\end{proof}

\subsection{Proof of Proposition~\ref{prop:examples-attention-geometry-aware}}

\begin{proof}
Fix an arbitrary context
$
c.
$
For each receiver index \(i\in\{1,\ldots,d\}\), define its set of admissible
source indices by
$
\mathcal N_i^{\mathrm{att}}
:=
\left\{
j\in\{1,\ldots,d\}:
(z_i,z_j)\in\mathcal E_{\mathrm{att}}
\right\}.
$
By assumption, every receiver has at least one admissible source. Hence
$
\mathcal N_i^{\mathrm{att}}\neq\varnothing
\qquad
\text{for every }i=1,\ldots,d.
$

For every
$
j\in\mathcal N_i^{\mathrm{att}},
$
Definition~\ref{def:examples-attention-score-matrix} gives
\[
S^{\mathrm{att}}_{ij}(c)
=
\frac{
\langle q_i(c),k_j(c)\rangle
}{
\sqrt p
}
+
b_{ij}.
\]
The hypothesis of the proposition states that this score is finite on every
admissible pair. Therefore
$
S^{\mathrm{att}}_{ij}(c)\in\mathbb R
\qquad
\text{whenever }
j\in\mathcal N_i^{\mathrm{att}}.
$
Consequently,
$
\exp\!\left(
S^{\mathrm{att}}_{ij}(c)
\right)
>0
\qquad
\text{for every }
j\in\mathcal N_i^{\mathrm{att}}.
$

For each receiver \(i\), define the corresponding row-normalization
constant
$
Z_i(c)
:=
\sum_{k\in\mathcal N_i^{\mathrm{att}}}
\exp\!\left(
S^{\mathrm{att}}_{ik}(c)
\right).
$
The index set
\(\mathcal N_i^{\mathrm{att}}\) is nonempty and finite, and every term in
this sum is a finite strictly positive real number. It follows that
$
0<Z_i(c)<\infty.
$
Thus the row-wise normalization is well defined for every receiver.

Because
$
A^{\mathrm{att}}(c)
=
\operatorname{RowSoftmax}
\bigl(
S^{\mathrm{att}}(c)
\bigr)
$
is, by
Definition~\ref{def:examples-attention-score-matrix}, normalized only over
admissible source sites, its entries can be written explicitly as
\[
A^{\mathrm{att}}_{ij}(c)
=
\begin{cases}
\displaystyle
\frac{
\exp\!\left(
S^{\mathrm{att}}_{ij}(c)
\right)
}{
Z_i(c)
},
&
j\in\mathcal N_i^{\mathrm{att}},
\\[1.2em]
0,
&
j\notin\mathcal N_i^{\mathrm{att}}.
\end{cases}
\]
Equivalently,
\[
A^{\mathrm{att}}_{ij}(c)
=
\begin{cases}
\displaystyle
\frac{
\exp\!\left(
S^{\mathrm{att}}_{ij}(c)
\right)
}{
\sum_{k:\,(z_i,z_k)\in\mathcal E_{\mathrm{att}}}
\exp\!\left(
S^{\mathrm{att}}_{ik}(c)
\right)
},
&
(z_i,z_j)\in\mathcal E_{\mathrm{att}},
\\[1.2em]
0,
&
(z_i,z_j)\notin\mathcal E_{\mathrm{att}}.
\end{cases}
\]

We first verify entrywise nonnegativity. If
$
(z_i,z_j)\notin\mathcal E_{\mathrm{att}},
$
then, by the preceding formula,
$
A^{\mathrm{att}}_{ij}(c)=0.
$
If instead
$
(z_i,z_j)\in\mathcal E_{\mathrm{att}},
$
then
$
\exp\!\left(
S^{\mathrm{att}}_{ij}(c)
\right)>0
$
and
$
Z_i(c)>0.
$
Therefore
$
A^{\mathrm{att}}_{ij}(c)
=
\frac{
\exp\!\left(
S^{\mathrm{att}}_{ij}(c)
\right)
}{
Z_i(c)
}
>0.
$
Hence, in both cases,
$
A^{\mathrm{att}}_{ij}(c)\ge0.
$
Since \(i,j\) were arbitrary,
$
A^{\mathrm{att}}(c)\ge0
$
entrywise.

The same explicit formula proves the support statement. Whenever
$
(z_i,z_j)\notin\mathcal E_{\mathrm{att}},
$
we have, by construction,
$
A^{\mathrm{att}}_{ij}(c)=0.
$
Thus
$
\operatorname{supp}_X
\bigl(
A^{\mathrm{att}}(c)
\bigr)
\subseteq
\mathcal E_{\mathrm{att}},
$
up to the precise site-indexed support convention used for the admissibility
relation. In particular, the row-wise softmax normalization cannot introduce
a nonzero interaction on an inadmissible receiver--source pair.

It remains to verify the row-sum identity. Fix
\(i\in\{1,\ldots,d\}\). Since all inadmissible entries in row \(i\) are
zero,
$
\sum_{j=1}^d
A^{\mathrm{att}}_{ij}(c)
=
\sum_{j\in\mathcal N_i^{\mathrm{att}}}
A^{\mathrm{att}}_{ij}(c).
$
Substituting the normalized expression gives
\[
\begin{aligned}
\sum_{j=1}^d
A^{\mathrm{att}}_{ij}(c)
&=
\sum_{j\in\mathcal N_i^{\mathrm{att}}}
\frac{
\exp\!\left(
S^{\mathrm{att}}_{ij}(c)
\right)
}{
Z_i(c)
}
\\
&=
\frac{1}{Z_i(c)}
\sum_{j\in\mathcal N_i^{\mathrm{att}}}
\exp\!\left(
S^{\mathrm{att}}_{ij}(c)
\right)
\\
&=
\frac{Z_i(c)}{Z_i(c)}
\\
&=
1.
\end{aligned}
\]
Because this holds for every receiver \(i\), every row of
\(A^{\mathrm{att}}(c)\) sums to one. Therefore
$
A^{\mathrm{att}}(c)\mathbf1
=
\mathbf1.
$
Indeed, for each \(i\),
\[
\begin{aligned}
\bigl(
A^{\mathrm{att}}(c)\mathbf1
\bigr)_i
&=
\sum_{j=1}^d
A^{\mathrm{att}}_{ij}(c)\mathbf1_j
\\
&=
\sum_{j=1}^d
A^{\mathrm{att}}_{ij}(c)
\\
&=
1.
\end{aligned}
\]

We have therefore established, for the arbitrary fixed context \(c\),
$
A^{\mathrm{att}}(c)\ge0,
$
$
A^{\mathrm{att}}(c)\mathbf1=\mathbf1,
$
and
$
A^{\mathrm{att}}_{ij}(c)=0
\qquad
\text{whenever }
(z_i,z_j)\notin\mathcal E_{\mathrm{att}}.
$
Thus \(A^{\mathrm{att}}(c)\) is a nonnegative row-stochastic matrix whose
cross-site mixing is confined to the prescribed admissibility relation.

Since the argument is pointwise in \(c\), no constancy of the query or key
vectors across contexts is required. The quantities
$
q_i(c),
\qquad
k_j(c),
\qquad
S^{\mathrm{att}}_{ij}(c),
\qquad
A^{\mathrm{att}}_{ij}(c)
$
may all vary with the context, provided that every receiver retains at least
one admissible source and every admissible score remains finite. The
support and row-stochasticity conclusions therefore hold for every such
context \(c\).
\end{proof}

\subsection{Proof of Proposition~\ref{prop:examples-anisotropic-drift-geometry-induced}}

\begin{proof}
Recall from
Definition~\ref{def:examples-anisotropic-drift-interaction} that
$
K^{\mathrm{drift}}
\in\mathbb R^{d\times d}
$
is defined entrywise by
$
(K^{\mathrm{drift}})_{ij}
=
\mathbf1_{\mathcal E}(z_i,z_j)\,
\kappa^{\mathrm{aniso}}(z_i,z_j),
$
where
\[
\kappa^{\mathrm{aniso}}(z_i,z_j)
=
\exp\!\left(
-\frac{1}{\varepsilon^2}
(\iota(z_j)-\iota(z_i))^\top
G_i
(\iota(z_j)-\iota(z_i))
+
\beta\,
b_i^\top
(\iota(z_j)-\iota(z_i))
\right).
\]
The descriptor
\(\mathfrak S_{\mathrm{drift}}\) has admissibility relation
\(\mathcal E\) and a single geometry channel
$
\phi_{\mathrm{drift}}(z_i,z_j)
=
\kappa^{\mathrm{aniso}}(z_i,z_j).
$
Let
$
\Phi_{\mathrm{drift}}
\in\mathbb R^{d\times d}
$
denote the masked channel matrix induced by this channel. By
Definition~\ref{def:prelim-geometry-descriptor},
$
(\Phi_{\mathrm{drift}})_{ij}
=
\mathbf1_{\mathcal E}(z_i,z_j)\,
\phi_{\mathrm{drift}}(z_i,z_j).
$
Substituting the definition of
\(\phi_{\mathrm{drift}}\) gives
\[
\begin{aligned}
(\Phi_{\mathrm{drift}})_{ij}
&=
\mathbf1_{\mathcal E}(z_i,z_j)\,
\kappa^{\mathrm{aniso}}(z_i,z_j)
\\
&=
(K^{\mathrm{drift}})_{ij}.
\end{aligned}
\]
Since this equality holds for every pair
\(i,j\in\{1,\ldots,d\}\), the two matrices coincide:
$
K^{\mathrm{drift}}
=
\Phi_{\mathrm{drift}}.
$
Thus \(K^{\mathrm{drift}}\) is exactly the masked realization of the unique
channel of \(\mathfrak S_{\mathrm{drift}}\).

We next verify explicitly that this masked channel is admissible in the
corresponding interaction family. For every ordered pair
\((z_i,z_j)\), the exponential defining
\(\kappa^{\mathrm{aniso}}\) is strictly positive:
$
\kappa^{\mathrm{aniso}}(z_i,z_j)>0.
$
Indeed, its exponent is a finite real scalar because
$
\varepsilon>0,
\qquad
G_i\in\mathbb R^{m\times m},
\qquad
b_i\in\mathbb R^m,
\qquad
\iota(z_j)-\iota(z_i)\in\mathbb R^m,
$
and the exponential of a finite real number is strictly positive.
Consequently,
$
(\Phi_{\mathrm{drift}})_{ij}
=
\mathbf1_{\mathcal E}(z_i,z_j)
\kappa^{\mathrm{aniso}}(z_i,z_j)
\ge0
$
for every \(i,j\), and in particular
$
(\Phi_{\mathrm{drift}})_{ij}\ge0
\qquad
\text{for every }i\neq j.
$
This is precisely why the unique channel index of
\(\mathfrak S_{\mathrm{drift}}\) may be placed in the positive-channel set
\(\mathcal R_+\).

Since there is only one channel, Definition~\ref{def:operators-interaction-family}
specializes to matrices of the form
$
\theta_{\mathrm{drift}}\Phi_{\mathrm{drift}}
+
\operatorname{Diag}(\delta),
$
where
$
\delta\in\mathbb R^d
$
is arbitrary and the positive-channel constraint requires
$
\theta_{\mathrm{drift}}\ge0.
$
Choose
$
\theta_{\mathrm{drift}}:=1
$
and
$
\delta:=0.
$
The required sign constraint is satisfied because
$
\theta_{\mathrm{drift}}
=
1
\ge0.
$
With this choice,
\[
\begin{aligned}
\theta_{\mathrm{drift}}\Phi_{\mathrm{drift}}
+
\operatorname{Diag}(\delta)
&=
1\cdot\Phi_{\mathrm{drift}}
+
\operatorname{Diag}(0)
\\
&=
\Phi_{\mathrm{drift}}
\\
&=
K^{\mathrm{drift}}.
\end{aligned}
\]
Therefore \(K^{\mathrm{drift}}\) has a valid representation in the
geometry-induced interaction family, and hence
$
K^{\mathrm{drift}}
\in
\mathcal A_{\mathfrak S_{\mathrm{drift}}}.
$

We finally make precise the structural origin of the possible asymmetry.
Fix two distinct indices \(i\neq j\), and define
$
d_{ij}
:=
\iota(z_j)-\iota(z_i).
$
Then
$
\iota(z_i)-\iota(z_j)
=
-d_{ij}.
$
Suppose first that both ordered pairs
$
(z_i,z_j)
\qquad\text{and}\qquad
(z_j,z_i)
$
are admissible, so that the mask does not eliminate either direction. The
forward channel value is
$
\kappa^{\mathrm{aniso}}(z_i,z_j)
=
\exp\!\left(
-\frac{1}{\varepsilon^2}
d_{ij}^\top G_i d_{ij}
+
\beta b_i^\top d_{ij}
\right).
$
For the reversed ordered pair, the receiver is now \(z_j\), so the metric
and drift used by the channel are \(G_j\) and \(b_j\). Hence
\[
\begin{aligned}
\kappa^{\mathrm{aniso}}(z_j,z_i)
&=
\exp\!\left(
-\frac{1}{\varepsilon^2}
(-d_{ij})^\top G_j(-d_{ij})
+
\beta b_j^\top(-d_{ij})
\right)
\\
&=
\exp\!\left(
-\frac{1}{\varepsilon^2}
d_{ij}^\top G_jd_{ij}
-
\beta b_j^\top d_{ij}
\right).
\end{aligned}
\]
Taking logarithms, which is legitimate because both quantities are strictly
positive, gives
$
\log
\kappa^{\mathrm{aniso}}(z_i,z_j)
=
-\frac{1}{\varepsilon^2}
d_{ij}^\top G_i d_{ij}
+
\beta b_i^\top d_{ij}
$
and
$
\log
\kappa^{\mathrm{aniso}}(z_j,z_i)
=
-\frac{1}{\varepsilon^2}
d_{ij}^\top G_jd_{ij}
-
\beta b_j^\top d_{ij}.
$
Subtracting the two identities yields
\[
\begin{aligned}
&
\log
\kappa^{\mathrm{aniso}}(z_i,z_j)
-
\log
\kappa^{\mathrm{aniso}}(z_j,z_i)
\\
&\qquad=
-\frac{1}{\varepsilon^2}
d_{ij}^\top
(G_i-G_j)
d_{ij}
+
\beta
(b_i+b_j)^\top d_{ij}.
\end{aligned}
\]
Thus even when both directions are admissible, the forward and reverse
interaction strengths need not agree. Their difference is completely
determined by the descriptor-level geometric data: the receiver-dependent
metrics \(G_i,G_j\), the receiver-dependent drift vectors \(b_i,b_j\), the
relative displacement \(d_{ij}\), and the fixed parameters
\(\varepsilon,\beta\). In particular, no independent unconstrained scalar
parameter is introduced for the matrix entry
\((K^{\mathrm{drift}})_{ij}\).

The same point can be expressed directly at the matrix level. For every
admissible ordered pair,
$
(K^{\mathrm{drift}})_{ij}
=
\exp\!\left(
-\frac{1}{\varepsilon^2}
d_{ij}^\top G_i d_{ij}
+
\beta b_i^\top d_{ij}
\right).
$
Once
$
\mathcal E,\qquad
\iota,\qquad
\{G_i\}_{i=1}^d,\qquad
\{b_i\}_{i=1}^d,\qquad
\varepsilon,\qquad
\beta
$
are fixed, every matrix entry is fixed as well. There is no family of
independently tunable pairwise coefficients
\(\alpha_{ij}\) whose values could generate arbitrary matrix asymmetry.
Rather, the entries are tied through the same anisotropic drift formula.

If the admissibility relation itself is directed, the mask can provide an
additional geometry-induced source of asymmetry. Indeed, it may happen that
$
(z_i,z_j)\in\mathcal E
\qquad\text{but}\qquad
(z_j,z_i)\notin\mathcal E.
$
Then
$
(K^{\mathrm{drift}})_{ij}
=
\kappa^{\mathrm{aniso}}(z_i,z_j)
>
0,
$
whereas
$
(K^{\mathrm{drift}})_{ji}
=
0.
$
This asymmetry is again imposed by the geometry descriptor, now through its
directed admissibility relation, rather than through unconstrained matrix
parameters. In particular, when the admissibility relation is symmetric,
any remaining asymmetry of the channel values is generated entirely by the
receiver-dependent anisotropic metrics and drift vectors described above.

Therefore \(K^{\mathrm{drift}}\) is a geometry-induced interaction operator,
and its possible directional asymmetry is structurally prescribed by the
geometry descriptor---in particular by the receiver-dependent \(G_i\) and
\(b_i\), together with any directionality already encoded by
\(\mathcal E\)---rather than by arbitrary free matrix entries.
\end{proof}

\subsection{Proof of Proposition~\ref{prop:examples-barrier-respecting-propagation}}

\begin{proof}
Recall that the barrier admissibility relation is
$
\mathcal E_{\mathrm{bar}}
=
\left\{
(z_i,z_j)\in X\times X:
\operatorname{Vis}(z_i,z_j)=1
\right\},
$
where
$
\operatorname{Vis}(z_i,z_j)
=
\mathbf1
\left\{
[\iota(z_i),\iota(z_j)]
\cap
\mathcal O
=
\varnothing
\right\}.
$
The barrier-masked interaction matrix is defined by
$
K^{\mathrm{bar}}_{ij}
=
\operatorname{Vis}(z_i,z_j)
\exp\!\left(
-\frac{
\|\iota(z_i)-\iota(z_j)\|^2
}{
\varepsilon^2
}
\right),
$
and
Definition~\ref{def:examples-barrier-masked-generator} sets
$
L^{\mathrm{bar}}
=
\Gamma(K^{\mathrm{bar}}).
$
Since
$
K^{\mathrm{bar}}_{ij}\ge0
\qquad
\text{for every }i\neq j,
$
Proposition~\ref{prop:examples-canonical-generator-property} applies and
gives
$
L^{\mathrm{bar}}
\in
\mathcal L_{\mathfrak S_{\mathrm{bar}}}.
$
Thus \(L^{\mathrm{bar}}\) is an admissible propagation generator for the
barrier descriptor.

We first identify its positive off-diagonal rate graph exactly. By
Definition~\ref{def:examples-row-balanced-generator-map},
$
\Gamma(K)
=
\operatorname{Off}(K)
-
\operatorname{Diag}
\bigl(
\operatorname{Off}(K)\mathbf1
\bigr).
$
The second term in this expression is diagonal. Therefore, for every pair of
distinct indices \(i\neq j\),
$
\Gamma(K)_{ij}
=
\operatorname{Off}(K)_{ij}
=
K_{ij}.
$
Applying this identity to \(K^{\mathrm{bar}}\) yields
$
L^{\mathrm{bar}}_{ij}
=
K^{\mathrm{bar}}_{ij}
\qquad
\text{for every }i\neq j.
$
Hence
\[
L^{\mathrm{bar}}_{ij}
=
\operatorname{Vis}(z_i,z_j)
\exp\!\left(
-\frac{
\|\iota(z_i)-\iota(z_j)\|^2
}{
\varepsilon^2
}
\right)
\qquad
(i\neq j).
\]

The Gaussian factor is strictly positive for every pair of sites. Indeed,
$
\|\iota(z_i)-\iota(z_j)\|^2
\ge0
$
and
$
\varepsilon>0,
$
so its exponent is a finite real number and therefore
$
\exp\!\left(
-\frac{
\|\iota(z_i)-\iota(z_j)\|^2
}{
\varepsilon^2
}
\right)
>0.
$
Consequently, for \(i\neq j\),
$
L^{\mathrm{bar}}_{ij}>0
$
if and only if
$
\operatorname{Vis}(z_i,z_j)=1.
$
By the definition of
\(\mathcal E_{\mathrm{bar}}\), this is equivalent to
$
(z_i,z_j)\in\mathcal E_{\mathrm{bar}}.
$
Thus, for every \(i\neq j\),
$
L^{\mathrm{bar}}_{ij}>0
\quad\Longleftrightarrow\quad
(z_i,z_j)\in\mathcal E_{\mathrm{bar}}.
$

Introduce the positive-rate relation associated with
\(L^{\mathrm{bar}}\),
$
\mathcal E_{L^{\mathrm{bar}}}
:=
\left\{
(z_a,z_b):
a\neq b,\;
L^{\mathrm{bar}}_{ab}>0
\right\}.
$
The preceding equivalence shows that
\[
\mathcal E_{L^{\mathrm{bar}}}
=
\left\{
(z_a,z_b)\in\mathcal E_{\mathrm{bar}}:
a\neq b
\right\}.
\]
In other words, the positive-rate graph of the generator is exactly the
off-diagonal part of the barrier-visibility graph.

We next verify that the two relations induce the same directed reachability
between distinct sites. Fix
$
i\neq j.
$
Suppose first that there exists a directed path from \(z_j\) to \(z_i\) in
\(\mathcal E_{L^{\mathrm{bar}}}\). Every edge of
\(\mathcal E_{L^{\mathrm{bar}}}\) belongs to
\(\mathcal E_{\mathrm{bar}}\), so the same sequence of sites is also a
directed path in \(\mathcal E_{\mathrm{bar}}\). Therefore
$
z_j\rightsquigarrow z_i
\quad\text{in }\mathcal E_{L^{\mathrm{bar}}}
$
implies
$
z_j\rightsquigarrow z_i
\quad\text{in }\mathcal E_{\mathrm{bar}}.
$

Conversely, suppose that
$
z_j\rightsquigarrow z_i
$
in the directed graph induced by
\(\mathcal E_{\mathrm{bar}}\). Then there exist an integer
\(r\ge1\) and indices
$
a_0,a_1,\ldots,a_r
$
such that
$
a_0=j,
\qquad
a_r=i,
$
and
$
(z_{a_\ell},z_{a_{\ell-1}})
\in
\mathcal E_{\mathrm{bar}}
\qquad
\text{for every }\ell=1,\ldots,r.
$
If some consecutive indices coincide,
$
a_\ell=a_{\ell-1},
$
then that step is merely a self-loop and can be deleted without changing the
initial site or the terminal site. Repeating this deletion produces a
sequence
$
b_0,b_1,\ldots,b_m
$
with
$
b_0=j,
\qquad
b_m=i,
$
such that
$
b_\ell\neq b_{\ell-1}
\qquad
\text{for every }\ell=1,\ldots,m,
$
and
$
(z_{b_\ell},z_{b_{\ell-1}})
\in
\mathcal E_{\mathrm{bar}}
\qquad
\text{for every }\ell.
$
Because \(i\neq j\), the reduced path has positive length:
$
m\ge1.
$
For every remaining edge, the two indices are distinct, and the previously
proved equality of edge relations gives
$
(z_{b_\ell},z_{b_{\ell-1}})
\in
\mathcal E_{L^{\mathrm{bar}}}.
$
Hence
\[
z_j=z_{b_0}
\longrightarrow
z_{b_1}
\longrightarrow
\cdots
\longrightarrow
z_{b_m}=z_i
\]
is a directed path in the positive-rate graph of
\(L^{\mathrm{bar}}\). Therefore
$
z_j\rightsquigarrow z_i
\quad\text{in }\mathcal E_{\mathrm{bar}}
$
implies
$
z_j\rightsquigarrow z_i
\quad\text{in }\mathcal E_{L^{\mathrm{bar}}}.
$
We have thus established, for \(i\neq j\),
\[
z_j\rightsquigarrow z_i
\text{ in }\mathcal E_{\mathrm{bar}}
\quad\Longleftrightarrow\quad
z_j\rightsquigarrow z_i
\text{ in }\mathcal E_{L^{\mathrm{bar}}}.
\]

Now fix
$
\tau>0.
$
Since
$
L^{\mathrm{bar}}
\in
\mathcal L_{\mathfrak S_{\mathrm{bar}}},
$
Theorem~\ref{thm:properties-reachability-preserving-propagation} applies.
For every \(i\neq j\), it gives
$
\bigl(e^{\tau L^{\mathrm{bar}}}\bigr)_{ij}>0
$
if and only if there exists a directed path from \(z_j\) to \(z_i\) in the
positive-rate relation
\(\mathcal E_{L^{\mathrm{bar}}}\). Combining this with the reachability
equivalence established above yields
\[
\bigl(e^{\tau L^{\mathrm{bar}}}\bigr)_{ij}>0
\quad\Longleftrightarrow\quad
z_j\rightsquigarrow z_i
\text{ in }\mathcal E_{\mathrm{bar}}.
\]
This proves the claimed strict-positivity characterization.

For completeness, the two directions can also be interpreted directly in
terms of finite propagation. If no visibility-admissible directed path from
\(z_j\) to \(z_i\) exists, then no positive-rate path exists for
\(L^{\mathrm{bar}}\), and Theorem~\ref{thm:properties-reachability-preserving-propagation}
gives
$
\bigl(e^{\tau L^{\mathrm{bar}}}\bigr)_{ij}=0.
$
Thus exponentiation cannot create influence across a pair of sites that is
disconnected in the visibility graph.

If, on the other hand, there is a visibility-admissible path
\[
z_j=z_{a_0}
\longrightarrow
z_{a_1}
\longrightarrow
\cdots
\longrightarrow
z_{a_r}=z_i,
\]
with no redundant self-loops, then
$
(z_{a_\ell},z_{a_{\ell-1}})
\in
\mathcal E_{\mathrm{bar}}
$
and
$
a_\ell\neq a_{\ell-1}
$
for every \(\ell\). Hence
\[
L^{\mathrm{bar}}_{a_\ell a_{\ell-1}}
=
\exp\!\left(
-\frac{
\|\iota(z_{a_\ell})-\iota(z_{a_{\ell-1}})\|^2
}{
\varepsilon^2
}
\right)
>
0.
\]
Thus every elementary move in the path carries a strictly positive
generator rate. The reachability theorem therefore yields
$
\bigl(e^{\tau L^{\mathrm{bar}}}\bigr)_{ij}>0
$
for every \(\tau>0\).

Finally, the definition of
\(\mathcal E_{\mathrm{bar}}\) shows exactly what each elementary edge means.
If
$
(z_a,z_b)\in\mathcal E_{\mathrm{bar}},
$
then
$
\operatorname{Vis}(z_a,z_b)=1,
$
and therefore
$
[\iota(z_a),\iota(z_b)]
\cap
\mathcal O
=
\varnothing.
$
Hence every nontrivial step contributing to a propagation path is a
line-segment transfer whose segment avoids the obstacle set. Finite-time
propagation may travel around an obstacle by concatenating several such
visible transfers, but it cannot create a coupling between two sites unless
they are connected by a sequence of visibility-admissible elementary
moves. This proves the barrier-respecting interpretation.
\end{proof}

\subsection{Proof of Proposition~\ref{prop:examples-hodge-structural-meaning}}

\begin{proof}
Recall that the edge-space Hodge Laplacian is
$
\Delta^{\mathrm H}_1
=
B_1^\top B_1+B_2B_2^\top.
$
We first establish its basic spectral structure. Taking transposes gives
\[
\begin{aligned}
\left(\Delta^{\mathrm H}_1\right)^\top
&=
\left(B_1^\top B_1+B_2B_2^\top\right)^\top
\\
&=
(B_1^\top B_1)^\top
+
(B_2B_2^\top)^\top
\\
&=
B_1^\top B_1
+
B_2B_2^\top
\\
&=
\Delta^{\mathrm H}_1.
\end{aligned}
\]
Hence
$
\Delta^{\mathrm H}_1
$
is symmetric.

Let
$
x\in\mathbb R^{|E|}
$
be arbitrary. Then
\[
\begin{aligned}
x^\top\Delta^{\mathrm H}_1x
&=
x^\top
\left(
B_1^\top B_1+B_2B_2^\top
\right)x
\\
&=
x^\top B_1^\top B_1x
+
x^\top B_2B_2^\top x.
\end{aligned}
\]
The first term can be rewritten as
$
x^\top B_1^\top B_1x
=
(B_1x)^\top(B_1x)
=
\|B_1x\|_2^2,
$
while the second term satisfies
\[
\begin{aligned}
x^\top B_2B_2^\top x
&=
(B_2^\top x)^\top(B_2^\top x)
\\
&=
\|B_2^\top x\|_2^2.
\end{aligned}
\]
Consequently,
$
x^\top\Delta^{\mathrm H}_1x
=
\|B_1x\|_2^2
+
\|B_2^\top x\|_2^2.
$
Both terms on the right are nonnegative, so$
x^\top\Delta^{\mathrm H}_1x
\ge0
\qquad
\text{for every }x\in\mathbb R^{|E|}.
$
Therefore
$
\Delta^{\mathrm H}_1\succeq0.
$
Together with symmetry, this proves that
\(\Delta^{\mathrm H}_1\) is symmetric positive semidefinite.

The preceding identity also identifies its kernel. Suppose first that
$
x\in\ker(\Delta^{\mathrm H}_1).
$
Then
$
\Delta^{\mathrm H}_1x=0,
$
and hence
\[
0
=
x^\top\Delta^{\mathrm H}_1x
=
\|B_1x\|_2^2
+
\|B_2^\top x\|_2^2.
\]
Since both summands are nonnegative, their sum can vanish only if both vanish:
$
\|B_1x\|_2^2=0
\qquad\text{and}\qquad
\|B_2^\top x\|_2^2=0.
$
Thus
$
B_1x=0
\qquad\text{and}\qquad
B_2^\top x=0.
$
Therefore
$
x\in
\ker(B_1)\cap\ker(B_2^\top).
$
Conversely, if
$
B_1x=0
\qquad\text{and}\qquad
B_2^\top x=0,
$
then
\[
\begin{aligned}
\Delta^{\mathrm H}_1x
&=
B_1^\top B_1x
+
B_2B_2^\top x
\\
&=
B_1^\top 0
+
B_2 0
\\
&=
0.
\end{aligned}
\]
Hence
$
\ker(\Delta^{\mathrm H}_1)
=
\ker(B_1)\cap\ker(B_2^\top).
$
This is the harmonic edge subspace associated with the Hodge Laplacian.

Now recall that
$
A^{\mathrm H}
=
-\alpha\Delta^{\mathrm H}_1,
\qquad
\alpha>0.
$
Because \(\Delta^{\mathrm H}_1\) is symmetric,
\[
\begin{aligned}
(A^{\mathrm H})^\top
&=
\left(
-\alpha\Delta^{\mathrm H}_1
\right)^\top
\\
&=
-\alpha
\left(
\Delta^{\mathrm H}_1
\right)^\top
\\
&=
-\alpha\Delta^{\mathrm H}_1
\\
&=
A^{\mathrm H}.
\end{aligned}
\]
Thus \(A^{\mathrm H}\) is symmetric. Moreover, for every
\(x\in\mathbb R^{|E|}\),
\[
\begin{aligned}
x^\top A^{\mathrm H}x
&=
-\alpha
x^\top\Delta^{\mathrm H}_1x
\\
&=
-\alpha
\left(
\|B_1x\|_2^2
+
\|B_2^\top x\|_2^2
\right)
\\
&\le0,
\end{aligned}
\]
because \(\alpha>0\). Therefore
$
A^{\mathrm H}\preceq0.
$
Thus \(A^{\mathrm H}\) is symmetric negative semidefinite.

Since \(\alpha>0\), the scalar \(-\alpha\) is nonzero. Therefore
$
A^{\mathrm H}x=0
$
if and only if
$
-\alpha\Delta^{\mathrm H}_1x=0,
$
which is equivalent to
$
\Delta^{\mathrm H}_1x=0.
$
Consequently,
$
\ker(A^{\mathrm H})
=
\ker(\Delta^{\mathrm H}_1).
$

We next analyze the explicit residual operator
$
R_\alpha
:=
I-\alpha\Delta^{\mathrm H}_1.
$
Because \(\Delta^{\mathrm H}_1\) is real symmetric, the spectral theorem
provides an orthogonal matrix
$
U\in\mathbb R^{|E|\times|E|}
$
and nonnegative real eigenvalues
$
0\le
\lambda_1\le\lambda_2\le\cdots\le\lambda_{|E|}
$
such that
\[
\Delta^{\mathrm H}_1
=
U
\operatorname{Diag}
(\lambda_1,\ldots,\lambda_{|E|})
U^\top.
\]
In particular,
$
\lambda_{|E|}
=
\lambda_{\max}(\Delta^{\mathrm H}_1).
$
Using
$
UU^\top=U^\top U=I,$
we obtain
\[
\begin{aligned}
R_\alpha
&=
I-\alpha\Delta^{\mathrm H}_1
\\
&=
U I U^\top
-
\alpha
U
\operatorname{Diag}
(\lambda_1,\ldots,\lambda_{|E|})
U^\top
\\
&=
U
\operatorname{Diag}
\left(
1-\alpha\lambda_1,
\ldots,
1-\alpha\lambda_{|E|}
\right)
U^\top.
\end{aligned}
\]

Let
$
x\in\mathbb R^{|E|}
$
and write
$
y:=U^\top x.
$
Since \(U\) is orthogonal,
$
\|y\|_2=\|x\|_2.
$
Furthermore,
\[
\begin{aligned}
\|R_\alpha x\|_2^2
&=
\left\|
U
\operatorname{Diag}
\left(
1-\alpha\lambda_1,
\ldots,
1-\alpha\lambda_{|E|}
\right)y
\right\|_2^2
\\
&=
\left\|
\operatorname{Diag}
\left(
1-\alpha\lambda_1,
\ldots,
1-\alpha\lambda_{|E|}
\right)y
\right\|_2^2
\\
&=
\sum_{r=1}^{|E|}
(1-\alpha\lambda_r)^2y_r^2.
\end{aligned}
\]
Hence \(R_\alpha\) is nonexpansive in Euclidean norm, meaning
$
\|R_\alpha x\|_2
\le
\|x\|_2
\qquad
\text{for every }x,
$
if and only if
$
|1-\alpha\lambda_r|
\le1
\qquad
\text{for every }r.
$
Indeed, if all these inequalities hold, then
\[
\begin{aligned}
\|R_\alpha x\|_2^2
&=
\sum_{r=1}^{|E|}
(1-\alpha\lambda_r)^2y_r^2
\\
&\le
\sum_{r=1}^{|E|}y_r^2
\\
&=
\|x\|_2^2.
\end{aligned}
\]
Conversely, if
$
|1-\alpha\lambda_r|>1
$
for some \(r\), choosing \(x\) to be a unit eigenvector corresponding to
\(\lambda_r\) gives
$
\|R_\alpha x\|_2
=
|1-\alpha\lambda_r|
\|x\|_2
>
\|x\|_2,
$
so nonexpansiveness fails.

For a fixed nonnegative eigenvalue \(\lambda_r\), the inequality
$
|1-\alpha\lambda_r|
\le1
$
is equivalent to
$
-1
\le
1-\alpha\lambda_r
\le
1.
$
Subtracting \(1\) throughout gives
$
-2
\le
-\alpha\lambda_r
\le
0,
$
and multiplying by \(-1\), which reverses the inequalities, yields
$
0
\le
\alpha\lambda_r
\le
2.
$
Assume now that
$
\lambda_{\max}(\Delta^{\mathrm H}_1)>0.
$
If
$
0\le
\alpha
\le
\frac{2}{
\lambda_{\max}(\Delta^{\mathrm H}_1)
},
$
then for every \(r\),
$
0
\le
\lambda_r
\le
\lambda_{\max}(\Delta^{\mathrm H}_1),
$
and therefore
$
0
\le
\alpha\lambda_r
\le
\alpha\lambda_{\max}(\Delta^{\mathrm H}_1)
\le
2.
$
Hence
$
|1-\alpha\lambda_r|
\le1
\qquad
\text{for every }r,
$
so \(R_\alpha\) is nonexpansive.

Conversely, suppose \(R_\alpha\) is nonexpansive. Applying the preceding
condition to an eigenvector corresponding to
\(\lambda_{\max}(\Delta^{\mathrm H}_1)\) gives
$
\left|
1-\alpha\lambda_{\max}(\Delta^{\mathrm H}_1)
\right|
\le1.
$
Since
$
\lambda_{\max}(\Delta^{\mathrm H}_1)>0,
$
this is equivalent to
$
0
\le
\alpha
\lambda_{\max}(\Delta^{\mathrm H}_1)
\le
2,
$
and hence
$
0
\le
\alpha
\le
\frac{2}{
\lambda_{\max}(\Delta^{\mathrm H}_1)
}.
$
Therefore
$
I-\alpha\Delta^{\mathrm H}_1
$
is nonexpansive in Euclidean norm if and only if
$
0\le
\alpha
\le
\frac{2}{
\lambda_{\max}(\Delta^{\mathrm H}_1)
}.
$

We finally consider the exponential propagation
$
S_\tau
:=
e^{-\tau\Delta^{\mathrm H}_1},
\qquad
\tau\ge0.
$
Using the same orthogonal spectral decomposition and the power-series
definition of the matrix exponential,
\[
\begin{aligned}
S_\tau
&=
e^{
-\tau
U
\operatorname{Diag}
(\lambda_1,\ldots,\lambda_{|E|})
U^\top
}
\\
&=
U
\operatorname{Diag}
\left(
e^{-\tau\lambda_1},
\ldots,
e^{-\tau\lambda_{|E|}}
\right)
U^\top.
\end{aligned}
\]
For every \(r\),
$
\lambda_r\ge0
\qquad\text{and}\qquad
\tau\ge0,
$
so
$
-\tau\lambda_r\le0
$
and consequently
$
0<
e^{-\tau\lambda_r}
\le1.
$
Thus, for an arbitrary vector
$
x=Uy,
$
we have
\[
\begin{aligned}
\|S_\tau x\|_2^2
&=
\sum_{r=1}^{|E|}
e^{-2\tau\lambda_r}y_r^2
\\
&\le
\sum_{r=1}^{|E|}y_r^2
\\
&=
\|x\|_2^2.
\end{aligned}
\]
Therefore
$
\|e^{-\tau\Delta^{\mathrm H}_1}x\|_2
\le
\|x\|_2
\qquad
\text{for every }x
$
and every
$
\tau\ge0.
$
Equivalently,
$
\left\|
e^{-\tau\Delta^{\mathrm H}_1}
\right\|_{2\to2}
\le1
\qquad
\text{for every }\tau\ge0.
$
Thus, unlike the explicit residual map, the exponential propagation requires
no upper restriction on the propagation magnitude.

We next show preservation of the harmonic subspace. Let
$
x\in\ker(\Delta^{\mathrm H}_1).
$
Then
$
\Delta^{\mathrm H}_1x=0,
$
and therefore, for every \(n\ge1\),
$
(\Delta^{\mathrm H}_1)^nx=0.
$
Using the exponential series,
\[
\begin{aligned}
e^{-\tau\Delta^{\mathrm H}_1}x
&=
\left[
I+
\sum_{n=1}^{\infty}
\frac{(-\tau)^n
(\Delta^{\mathrm H}_1)^n}{n!}
\right]x
\\
&=
x+
\sum_{n=1}^{\infty}
\frac{(-\tau)^n}{n!}
(\Delta^{\mathrm H}_1)^nx
\\
&=
x.
\end{aligned}
\]
Hence every harmonic vector is fixed pointwise:
$
e^{-\tau\Delta^{\mathrm H}_1}x=x
\qquad
\text{for every }
x\in\ker(\Delta^{\mathrm H}_1).
$
In particular,
\[
e^{-\tau\Delta^{\mathrm H}_1}
\ker(\Delta^{\mathrm H}_1)
\subseteq
\ker(\Delta^{\mathrm H}_1).
\]

It remains to identify the long-time limit. Let
$
\mathcal I_0
:=
\{r:\lambda_r=0\}
$
and
$
\mathcal I_+
:=
\{r:\lambda_r>0\}.
$
Let \(u_1,\ldots,u_{|E|}\) denote the orthonormal columns of \(U\).
Then the orthogonal projection onto the kernel of
\(\Delta^{\mathrm H}_1\) is
$
P_{\mathrm{harm}}
:=
\sum_{r\in\mathcal I_0}
u_ru_r^\top.
$
Equivalently,
\[
P_{\mathrm{harm}}
=
U
\operatorname{Diag}
(p_1,\ldots,p_{|E|})
U^\top,
\]
where
$
p_r
=
\begin{cases}
1, & \lambda_r=0,\\
0, & \lambda_r>0.
\end{cases}
$
For every zero eigenvalue,
$
e^{-\tau\lambda_r}
=
1
\qquad
\text{for all }\tau\ge0,
$
whereas for every strictly positive eigenvalue,
$
e^{-\tau\lambda_r}
\longrightarrow
0
\qquad
\text{as }\tau\to\infty.
$
Therefore, entrywise on the spectral diagonal,
\[
\operatorname{Diag}
\left(
e^{-\tau\lambda_1},
\ldots,
e^{-\tau\lambda_{|E|}}
\right)
\longrightarrow
\operatorname{Diag}
(p_1,\ldots,p_{|E|}).
\]
Consequently,
$
e^{-\tau\Delta^{\mathrm H}_1}
\longrightarrow
P_{\mathrm{harm}}
\qquad
\text{as }\tau\to\infty.
$

In fact, this convergence holds in the induced Euclidean operator norm.
Indeed,
\[
\begin{aligned}
e^{-\tau\Delta^{\mathrm H}_1}
-
P_{\mathrm{harm}}
&=
U
\operatorname{Diag}
\left(
e^{-\tau\lambda_r}-p_r
\right)_{r=1}^{|E|}
U^\top,
\end{aligned}
\]
and hence
$
\left\|
e^{-\tau\Delta^{\mathrm H}_1}
-
P_{\mathrm{harm}}
\right\|_{2\to2}
=
\max_{r\in\mathcal I_+}
e^{-\tau\lambda_r},
$
with the right-hand side understood as \(0\) if
\(\mathcal I_+=\varnothing\). If \(\mathcal I_+\neq\varnothing\), the set of
positive eigenvalues is finite, so
$
\lambda_{\min}^{+}
:=
\min_{r\in\mathcal I_+}\lambda_r
>
0,
$
and therefore
\[
\left\|
e^{-\tau\Delta^{\mathrm H}_1}
-
P_{\mathrm{harm}}
\right\|_{2\to2}
=
e^{-\tau\lambda_{\min}^{+}}
\longrightarrow
0.
\]
If instead
$
\mathcal I_+=\varnothing,
$
then
$
\Delta^{\mathrm H}_1=0,
$
so
$
e^{-\tau\Delta^{\mathrm H}_1}
=
I
=
P_{\mathrm{harm}}
$
for every \(\tau\ge0\), and the same conclusion holds trivially.

Thus
$
e^{-\tau\Delta^{\mathrm H}_1}
$
is nonexpansive for every \(\tau\ge0\), fixes the harmonic subspace
\(\ker(\Delta^{\mathrm H}_1)\) pointwise, and converges in operator norm to
the orthogonal projection onto that subspace. This proves all claims of the
proposition.
\end{proof}

\subsection{Proof of Proposition~\ref{prop:examples-sheaf-structural-meaning}}

\begin{proof}
Recall that the degree-zero sheaf coboundary is the linear operator
$
\delta_0^{\mathcal F}
$
whose component on an oriented edge \(e:u\to v\) is
$
(\delta_0^{\mathcal F}h)_e
=
R_{ve}h_v-R_{ue}h_u,
$
and that the corresponding sheaf Laplacian is
$
\Delta^{\mathrm{sh}}
=
(\delta_0^{\mathcal F})^\top
\delta_0^{\mathcal F}.
$
All stalks are equipped with Euclidean inner products, so the transpose
appearing here is the ordinary Euclidean adjoint after choosing orthonormal
coordinates on the stalks.

We first verify symmetry. Taking the transpose gives
\[
\begin{aligned}
\left(\Delta^{\mathrm{sh}}\right)^\top
&=
\left[
(\delta_0^{\mathcal F})^\top
\delta_0^{\mathcal F}
\right]^\top
\\
&=
(\delta_0^{\mathcal F})^\top
\left[
(\delta_0^{\mathcal F})^\top
\right]^\top
\\
&=
(\delta_0^{\mathcal F})^\top
\delta_0^{\mathcal F}
\\
&=
\Delta^{\mathrm{sh}}.
\end{aligned}
\]
Hence \(\Delta^{\mathrm{sh}}\) is symmetric.

Let now
$
h=(h_v)_{v\in V}
$
be an arbitrary stacked vertex-stalk signal. Using the definition of the
sheaf Laplacian,
\[
\begin{aligned}
h^\top\Delta^{\mathrm{sh}}h
&=
h^\top
(\delta_0^{\mathcal F})^\top
\delta_0^{\mathcal F}h
\\
&=
\left(
\delta_0^{\mathcal F}h
\right)^\top
\left(
\delta_0^{\mathcal F}h
\right)
\\
&=
\left\|
\delta_0^{\mathcal F}h
\right\|_2^2.
\end{aligned}
\]
The codomain of \(\delta_0^{\mathcal F}\) is the direct sum of the edge
stalks. Since these stalks carry Euclidean inner products, the squared norm
of a stacked edge-stalk vector is the sum of the squared norms of its edge
components. Therefore
$
\left\|
\delta_0^{\mathcal F}h
\right\|_2^2
=
\sum_{e:u\to v}
\left\|
(\delta_0^{\mathcal F}h)_e
\right\|_2^2.
$
Substituting the definition of the coboundary component yields
$
\left\|
\delta_0^{\mathcal F}h
\right\|_2^2
=
\sum_{e:u\to v}
\left\|
R_{ve}h_v-R_{ue}h_u
\right\|_2^2.
$
Consequently,
$
h^\top\Delta^{\mathrm{sh}}h
=
\sum_{e:u\to v}
\left\|
R_{ve}h_v-R_{ue}h_u
\right\|_2^2.
$
Every term on the right-hand side is a squared Euclidean norm and is
therefore nonnegative. Hence
$
h^\top\Delta^{\mathrm{sh}}h
\ge0
\qquad
\text{for every }h,
$
which proves
$
\Delta^{\mathrm{sh}}\succeq0.
$
Together with the symmetry established above, this shows that
\(\Delta^{\mathrm{sh}}\) is symmetric positive semidefinite.

The same energy identity gives the kernel characterization. Suppose first
that
$
h\in\ker(\Delta^{\mathrm{sh}}).
$
Then
$
\Delta^{\mathrm{sh}}h=0,
$
and therefore
$
0
=
h^\top\Delta^{\mathrm{sh}}h.
$
Using the energy representation,
$
0
=
\left\|
\delta_0^{\mathcal F}h
\right\|_2^2.
$
A Euclidean norm vanishes if and only if its argument is the zero vector, so
$
\delta_0^{\mathcal F}h=0.
$
Hence
$
h\in\ker(\delta_0^{\mathcal F}),
$
and we have shown
$
\ker(\Delta^{\mathrm{sh}})
\subseteq
\ker(\delta_0^{\mathcal F}).
$

Conversely, suppose
$
h\in\ker(\delta_0^{\mathcal F}).
$
Then
$
\delta_0^{\mathcal F}h=0,
$
and hence
\[
\begin{aligned}
\Delta^{\mathrm{sh}}h
&=
(\delta_0^{\mathcal F})^\top
\delta_0^{\mathcal F}h
\\
&=
(\delta_0^{\mathcal F})^\top 0
\\
&=
0.
\end{aligned}
\]
Thus
$
h\in\ker(\Delta^{\mathrm{sh}}),
$
which proves the reverse inclusion. Therefore
$
\ker(\Delta^{\mathrm{sh}})
=
\ker(\delta_0^{\mathcal F}).
$

The geometric meaning of this kernel follows directly from the definition of
the coboundary. The condition
$
\delta_0^{\mathcal F}h=0
$
holds if and only if every edge component vanishes. Thus, for every oriented
edge \(e:u\to v\),
$
R_{ve}h_v-R_{ue}h_u=0,
$
or equivalently,
$
R_{ve}h_v
=
R_{ue}h_u.
$
Hence the values assigned to the two endpoints of every edge agree after
restriction to the common edge stalk. This is exactly the compatibility
condition defining a globally consistent section of the cellular sheaf.
Accordingly,
$
\ker(\Delta^{\mathrm{sh}})
=
\ker(\delta_0^{\mathcal F})
$
is the space of globally consistent sections.

We next consider the operator
$
-\alpha\Delta^{\mathrm{sh}},
\qquad
\alpha>0.
$
Since \(\Delta^{\mathrm{sh}}\) is symmetric,
\[
\begin{aligned}
\left(
-\alpha\Delta^{\mathrm{sh}}
\right)^\top
&=
-\alpha
\left(
\Delta^{\mathrm{sh}}
\right)^\top
\\
&=
-\alpha\Delta^{\mathrm{sh}},
\end{aligned}
\]
so \(-\alpha\Delta^{\mathrm{sh}}\) is symmetric. Moreover, for every \(h\),
\[
\begin{aligned}
h^\top
\left(
-\alpha\Delta^{\mathrm{sh}}
\right)
h
&=
-\alpha
h^\top\Delta^{\mathrm{sh}}h
\\
&=
-\alpha
\left\|
\delta_0^{\mathcal F}h
\right\|_2^2
\\
&=
-\alpha
\sum_{e:u\to v}
\left\|
R_{ve}h_v-R_{ue}h_u
\right\|_2^2
\\
&\le0.
\end{aligned}
\]
Therefore
$
-\alpha\Delta^{\mathrm{sh}}
\preceq0.
$

Now recall from
Definition~\ref{def:examples-sheaf-block-interaction-matrix} that
$
A^{\mathrm{sh}}
=
-\alpha\Delta^{\mathrm{sh}}
+
I_{|V|}\otimes D.
$
There is no corresponding sign-definiteness conclusion for an arbitrary
matrix \(D\). If \(D\) is not symmetric, then
$
(I_{|V|}\otimes D)^\top
=
I_{|V|}\otimes D^\top
$
need not equal
$
I_{|V|}\otimes D,
$
and therefore \(A^{\mathrm{sh}}\) need not even be symmetric. In that case
the usual Loewner notions
\(\preceq0\) and \(\succeq0\) do not apply to
\(A^{\mathrm{sh}}\) as a symmetric quadratic-form statement.

Even if \(D\) is symmetric, no fixed definiteness follows without a sign
condition on \(D\). For example, choose
$
D=\beta I_c
$
with
$
\beta>0.
$
Then
$
I_{|V|}\otimes D
=
\beta I_{|V|c},
$
and hence
\[
A^{\mathrm{sh}}
=
-\alpha\Delta^{\mathrm{sh}}
+
\beta I_{|V|c}.
\]
Because \(\Delta^{\mathrm{sh}}\) is symmetric positive semidefinite, all of
its eigenvalues are real and nonnegative. If
$
\lambda_{\max}
=
\lambda_{\max}(\Delta^{\mathrm{sh}}),
$
then choosing
$
\beta>\alpha\lambda_{\max}
$
gives, for every eigenvalue \(\lambda\) of
\(\Delta^{\mathrm{sh}}\),
$
\beta-\alpha\lambda
\ge
\beta-\alpha\lambda_{\max}
>
0.
$
Thus \(A^{\mathrm{sh}}\) is positive definite for this admissible choice of
\(D\), rather than negative semidefinite. Hence no universal definiteness
statement can hold for arbitrary \(D\).

Assume now that
$
D=D^\top\preceq0.
$
The Kronecker term is then symmetric because
\[
\begin{aligned}
(I_{|V|}\otimes D)^\top
&=
I_{|V|}^\top\otimes D^\top
\\
&=
I_{|V|}\otimes D.
\end{aligned}
\]
Since \(-\alpha\Delta^{\mathrm{sh}}\) is also symmetric, their sum satisfies
$
(A^{\mathrm{sh}})^\top
=
A^{\mathrm{sh}}.
$

To prove negative semidefiniteness, write an arbitrary stacked vertex feature
vector as
$
h
=
(h_{v_1}^\top,\ldots,h_{v_{|V|}}^\top)^\top,
\qquad
h_v\in\mathbb R^c.
$
Because \(I_{|V|}\otimes D\) is block diagonal with one copy of \(D\) on
each vertex block,
$
h^\top
(I_{|V|}\otimes D)
h
=
\sum_{v\in V}
h_v^\top D h_v.
$
The assumption
$
D\preceq0
$
gives
$
h_v^\top D h_v\le0
\qquad
\text{for every }v,
$
and hence
$
h^\top
(I_{|V|}\otimes D)
h
\le0.
$
Combining this with the sheaf-Laplacian term yields
\[
\begin{aligned}
h^\top A^{\mathrm{sh}}h
&=
-\alpha
h^\top\Delta^{\mathrm{sh}}h
+
h^\top
(I_{|V|}\otimes D)
h
\\
&=
-\alpha
\sum_{e:u\to v}
\left\|
R_{ve}h_v-R_{ue}h_u
\right\|_2^2
+
\sum_{v\in V}
h_v^\top D h_v
\\
&\le0.
\end{aligned}
\]
Therefore
$
A^{\mathrm{sh}}\preceq0.
$
Thus, whenever
$
D=D^\top\preceq0,
$
the full sheaf block correction operator is symmetric negative semidefinite.

It remains to clarify why these spectral conclusions do not imply a
coordinatewise positivity-preserving interpretation after flattening the
stalk coordinates. Positive or negative semidefiniteness constrains the
quadratic form
$
h^\top A^{\mathrm{sh}}h,
$
but it does not impose a sign on individual off-diagonal entries of the
flattened matrix.

This distinction can be seen directly from the block structure of the sheaf
Laplacian. Consider one oriented edge
$
e:u\to v.
$
The contribution of this edge to the coboundary has the two nonzero vertex
blocks
$
-R_{ue}
\qquad\text{and}\qquad
R_{ve}.
$
Consequently, the corresponding contribution to
$
(\delta_0^{\mathcal F})^\top
\delta_0^{\mathcal F}
$
contains the off-diagonal vertex blocks
$
-R_{ue}^\top R_{ve}
\qquad\text{and}\qquad
-R_{ve}^\top R_{ue}.
$
After multiplication by \(-\alpha\), the corresponding off-diagonal blocks
of
\(-\alpha\Delta^{\mathrm{sh}}\) are
$
\alpha R_{ue}^\top R_{ve}
\qquad\text{and}\qquad
\alpha R_{ve}^\top R_{ue}.
$
No assumption made above forces the entries of these products to be
nonnegative.

For example, take two vertices \(u,v\), one edge \(e:u\to v\), vertex and
edge stalks equal to \(\mathbb R^2\), and restriction maps
$
R_{ue}
=
I_2,
\qquad
R_{ve}
=
\begin{pmatrix}
1&0\\
0&-1
\end{pmatrix}.
$
Then
\[
R_{ue}^\top R_{ve}
=
\begin{pmatrix}
1&0\\
0&-1
\end{pmatrix}.
\]
Hence the \(u,v\) off-diagonal block of
\(-\alpha\Delta^{\mathrm{sh}}\) contains the entry
$
-\alpha<0
$
in its second feature coordinate. Since \(u\neq v\), this is an
off-diagonal entry of the fully flattened
\(2|V|\times2|V|\) matrix. Therefore
$
-\alpha\Delta^{\mathrm{sh}}
$
is symmetric negative semidefinite in this example but is not Metzler.

The same example also shows why negative semidefiniteness alone does not
imply positivity preservation of the exponential flow. Let
$
A=-\alpha\Delta^{\mathrm{sh}}
$
for the preceding construction, and let \((r,s)\) denote the flattened
off-diagonal coordinate for which
$
A_{rs}<0.
$
The matrix exponential has the expansion
\[
e^{tA}
=
I+tA+O(t^2)
\qquad
\text{as }t\downarrow0.
\]
Because \(r\neq s\),
$
I_{rs}=0,
$
and therefore
\[
(e^{tA})_{rs}
=
tA_{rs}+O(t^2).
\]
Since
$
A_{rs}<0,
$
the linear term is strictly negative. Hence, for all sufficiently small
\(t>0\),
$
(e^{tA})_{rs}<0.
$
Thus the exponential is not entrywise nonnegative in general, despite the
fact that \(A\) is symmetric negative semidefinite.

Accordingly, symmetry and spectral definiteness are logically distinct from
the ordered-coordinate conditions required for a block-valued Markov
generator. To invoke the positivity-preserving interpretation of
Definition~\ref{def:operators-block-valued-generator-family}, one must
separately verify the flattened conditions
$
L_{\alpha\beta}\ge0
\qquad
\text{for all }\alpha\neq\beta,
$
together with
$
L\mathbf1_{dc}=0.
$
Those properties do not follow merely from the sheaf-Laplacian Gram
structure or from negative semidefiniteness. This proves all claims of the
proposition.
\end{proof}

\subsection{Proof of Proposition~\ref{prop:appendix_projection_nonexpansive}}

\begin{proof}
Let
$
\mathcal C
:=
\mathcal A_{\mathfrak S}.
$
By Proposition~\ref{prop:operators-interaction-family-structure},
\(\mathcal C\) is a nonempty closed convex subset of the finite-dimensional
Hilbert space
\(\mathbb R^{d\times d}\) equipped with the Frobenius inner product
\[
\langle X,Y\rangle_F
:=
\operatorname{tr}(X^\top Y)
\]
and the associated norm
$
\|X\|_F
=
\sqrt{\langle X,X\rangle_F}.
$
Hence the metric projection onto \(\mathcal C\) is well defined and
single-valued. For
$
B_1,B_2\in\mathbb R^{d\times d},
$
write
$
A_1^\star
:=
\Pi_{\mathcal C}(B_1),
\qquad
A_2^\star
:=
\Pi_{\mathcal C}(B_2).
$
By definition of the metric projection,
$
A_i^\star
\in
\mathcal C
$
and
$
\|B_i-A_i^\star\|_F^2
\le
\|B_i-A\|_F^2
\qquad
\text{for every }A\in\mathcal C,
\quad
i\in\{1,2\}.
$

We first derive directly the first-order inequality needed for the
nonexpansiveness argument. Fix
$
i\in\{1,2\}
$
and an arbitrary
$
A\in\mathcal C.
$
Since \(\mathcal C\) is convex, for every
$
t\in[0,1]
$
the matrix
\[
A_i(t)
:=
(1-t)A_i^\star+tA
=
A_i^\star+t(A-A_i^\star)
\]
also belongs to \(\mathcal C\). The minimizing property of
\(A_i^\star\) therefore gives
$
\|B_i-A_i^\star\|_F^2
\le
\|B_i-A_i(t)\|_F^2.
$
Using the definition of \(A_i(t)\),
$
B_i-A_i(t)
=
B_i-A_i^\star-t(A-A_i^\star).
$
Hence
\[
\begin{aligned}
\|B_i-A_i(t)\|_F^2
&=
\left\|
B_i-A_i^\star
-
t(A-A_i^\star)
\right\|_F^2
\\
&=
\|B_i-A_i^\star\|_F^2
-
2t
\left\langle
B_i-A_i^\star,\,
A-A_i^\star
\right\rangle_F
\\
&\qquad
+
t^2
\|A-A_i^\star\|_F^2.
\end{aligned}
\]
Substituting this expansion into the projection inequality and cancelling
the common term
\(\|B_i-A_i^\star\|_F^2\) from both sides yields
\[
0
\le
-2t
\left\langle
B_i-A_i^\star,\,
A-A_i^\star
\right\rangle_F
+
t^2
\|A-A_i^\star\|_F^2.
\]
For every
$
t\in(0,1],
$
division by \(t>0\) gives
\[
2
\left\langle
B_i-A_i^\star,\,
A-A_i^\star
\right\rangle_F
\le
t
\|A-A_i^\star\|_F^2.
\]
The right-hand side converges to zero as
$
t\downarrow0.
$
Therefore
$
\left\langle
B_i-A_i^\star,\,
A-A_i^\star
\right\rangle_F
\le
0.
$
Thus, for each \(i\in\{1,2\}\),
$
\left\langle
B_i-A_i^\star,\,
A-A_i^\star
\right\rangle_F
\le0
\qquad
\text{for every }A\in\mathcal C.
$

We now apply this inequality to the two projection points. For
\(i=1\), choose
$
A=A_2^\star.
$
Since
$
A_2^\star\in\mathcal C,
$
we obtain
$
\left\langle
B_1-A_1^\star,\,
A_2^\star-A_1^\star
\right\rangle_F
\le0.
$
Equivalently,
$
\left\langle
B_1-A_1^\star,\,
A_1^\star-A_2^\star
\right\rangle_F
\ge0.
$
For \(i=2\), choose
$
A=A_1^\star.
$
This gives
$
\left\langle
B_2-A_2^\star,\,
A_1^\star-A_2^\star
\right\rangle_F
\le0.
$

Set
$
D
:=
A_1^\star-A_2^\star.
$
The two inequalities can then be written as
$
\langle B_1-A_1^\star,D\rangle_F
\ge0
$
and
$
\langle B_2-A_2^\star,D\rangle_F
\le0.
$
Subtracting the second quantity from the first therefore gives
\[
\left\langle
(B_1-A_1^\star)
-
(B_2-A_2^\star),
D
\right\rangle_F
\ge0.
\]
Since
$
(B_1-A_1^\star)
-
(B_2-A_2^\star)
=
(B_1-B_2)
-
(A_1^\star-A_2^\star),
$
and
$
A_1^\star-A_2^\star=D,
$
we obtain
$
\left\langle
(B_1-B_2)-D,\,
D
\right\rangle_F
\ge0.
$
Expanding the inner product gives
$
\langle B_1-B_2,D\rangle_F
-
\langle D,D\rangle_F
\ge0.
$
Because
$
\langle D,D\rangle_F
=
\|D\|_F^2,
$
it follows that
$
\|D\|_F^2
\le
\langle B_1-B_2,D\rangle_F.
$
Restoring the definition of \(D\), we have proved the stronger
firm-nonexpansiveness inequality
\[
\begin{aligned}
&
\left\|
\Pi_{\mathcal A_{\mathfrak S}}(B_1)
-
\Pi_{\mathcal A_{\mathfrak S}}(B_2)
\right\|_F^2
\\
&\qquad\le
\left\langle
B_1-B_2,\,
\Pi_{\mathcal A_{\mathfrak S}}(B_1)
-
\Pi_{\mathcal A_{\mathfrak S}}(B_2)
\right\rangle_F.
\end{aligned}
\]

By the Cauchy--Schwarz inequality for the Frobenius inner product,
$
\langle X,Y\rangle_F
\le
\|X\|_F\|Y\|_F.
$
Applying this with
$
X=B_1-B_2
$
and
$
Y=D
$
gives
$
\|D\|_F^2
\le
\|B_1-B_2\|_F\|D\|_F.
$
If
$
D=0,
$
then
$
\|D\|_F=0
\le
\|B_1-B_2\|_F,
$
so the desired inequality is immediate. If
$
D\neq0,
$
then
$
\|D\|_F>0,
$
and dividing both sides by
\(\|D\|_F\) yields
$
\|D\|_F
\le
\|B_1-B_2\|_F.
$
Thus in all cases,
\[
\left\|
\Pi_{\mathcal A_{\mathfrak S}}(B_1)
-
\Pi_{\mathcal A_{\mathfrak S}}(B_2)
\right\|_F
\le
\|B_1-B_2\|_F.
\]

Hence
$
\Pi_{\mathcal A_{\mathfrak S}}
:
\mathbb R^{d\times d}
\longrightarrow
\mathcal A_{\mathfrak S}
$
is \(1\)-Lipschitz with respect to the Frobenius norm. In particular, let
$
B_n\to B
$
in Frobenius norm. Then
\[
\begin{aligned}
\left\|
\Pi_{\mathcal A_{\mathfrak S}}(B_n)
-
\Pi_{\mathcal A_{\mathfrak S}}(B)
\right\|_F
&\le
\|B_n-B\|_F
\\
&\longrightarrow
0.
\end{aligned}
\]
Therefore
$
\Pi_{\mathcal A_{\mathfrak S}}(B_n)
\to
\Pi_{\mathcal A_{\mathfrak S}}(B),
$
and the metric projection
\(\Pi_{\mathcal A_{\mathfrak S}}\) is continuous.
\end{proof}

\subsection{Proof of Proposition~\ref{prop:appendix_projection_variational}}

\begin{proof}
Let
$
A^\star
=
\Pi_{\mathcal A_{\mathfrak S}}(B).
$
By Proposition~\ref{prop:operators-interaction-family-structure},
\(\mathcal A_{\mathfrak S}\) is a nonempty closed convex subset of
\(\mathbb R^{d\times d}\). By the definition of the metric projection,
\(A^\star\in\mathcal A_{\mathfrak S}\) and
$
\|B-A^\star\|_F^2
\le
\|B-C\|_F^2
\qquad
\text{for every }
C\in\mathcal A_{\mathfrak S}.
$

Fix an arbitrary
$
A\in\mathcal A_{\mathfrak S}.
$
Since \(\mathcal A_{\mathfrak S}\) is convex and both
\(A^\star\) and \(A\) belong to it, for every
$
t\in[0,1]
$
the matrix
\[
A_t
:=
(1-t)A^\star+tA
\]
also belongs to \(\mathcal A_{\mathfrak S}\). Equivalently,
$
A_t
=
A^\star+t(A-A^\star).
$
The minimizing property of \(A^\star\) can therefore be applied with
\(C=A_t\), giving
$
\|B-A^\star\|_F^2
\le
\|B-A_t\|_F^2.
$
Using the expression for \(A_t\),
$
B-A_t
=
B-A^\star-t(A-A^\star).
$
Hence
\[
\begin{aligned}
\|B-A_t\|_F^2
&=
\left\|
B-A^\star-t(A-A^\star)
\right\|_F^2
\\
&=
\left\langle
B-A^\star-t(A-A^\star),\,
B-A^\star-t(A-A^\star)
\right\rangle_F
\\
&=
\|B-A^\star\|_F^2
-
2t
\left\langle
B-A^\star,\,
A-A^\star
\right\rangle_F
\\
&\qquad
+
t^2
\|A-A^\star\|_F^2.
\end{aligned}
\]
Substituting this expansion into the projection inequality yields
\[
\begin{aligned}
\|B-A^\star\|_F^2
&\le
\|B-A^\star\|_F^2
-
2t
\left\langle
B-A^\star,\,
A-A^\star
\right\rangle_F
\\
&\qquad
+
t^2
\|A-A^\star\|_F^2.
\end{aligned}
\]
Cancelling the common term
\(\|B-A^\star\|_F^2\) from both sides gives
\[
0
\le
-2t
\left\langle
B-A^\star,\,
A-A^\star
\right\rangle_F
+
t^2
\|A-A^\star\|_F^2.
\]
For every
$
t\in(0,1],
$
we may divide by the strictly positive number \(t\), obtaining
$
0
\le
-2
\left\langle
B-A^\star,\,
A-A^\star
\right\rangle_F
+
t
\|A-A^\star\|_F^2.
$
Equivalently,
$
2
\left\langle
B-A^\star,\,
A-A^\star
\right\rangle_F
\le
t
\|A-A^\star\|_F^2.
$
The quantity
$
\|A-A^\star\|_F^2
$
is fixed and finite, so the right-hand side converges to zero as
\(t\downarrow0\). Passing to the limit gives
$
2
\left\langle
B-A^\star,\,
A-A^\star
\right\rangle_F
\le
0,
$
and therefore
\[
\boxed{
\left\langle
B-A^\star,\,
A-A^\star
\right\rangle_F
\le0
}.
\]
Since the choice of
\(A\in\mathcal A_{\mathfrak S}\) was arbitrary, the variational inequality
holds for every feasible structured operator.

Suppose now that
$
\mathcal R_+=\varnothing.
$
By Proposition~\ref{prop:operators-interaction-family-structure},
\(\mathcal A_{\mathfrak S}\) is then a linear subspace of
\(\mathbb R^{d\times d}\). In particular,
$
A^\star\in\mathcal A_{\mathfrak S}.
$
Let
$
A\in\mathcal A_{\mathfrak S}
$
be arbitrary. Since a linear subspace is closed under addition and scalar
multiplication,
$
A^\star+A
\in
\mathcal A_{\mathfrak S}
$
and
$
A^\star-A
\in
\mathcal A_{\mathfrak S}.
$

Apply the variational inequality first with the feasible matrix
$
C=A^\star+A.
$
Then
$
\left\langle
B-A^\star,\,
C-A^\star
\right\rangle_F
\le0.
$
Since
$
C-A^\star
=
(A^\star+A)-A^\star
=
A,
$
we obtain
$
\left\langle
B-A^\star,\,
A
\right\rangle_F
\le0.
$

Next apply the same variational inequality with
$
C=A^\star-A.
$
This gives
$
\left\langle
B-A^\star,\,
C-A^\star
\right\rangle_F
\le0.
$
Now
$
C-A^\star
=
(A^\star-A)-A^\star
=
-A,
$
so
$
\left\langle
B-A^\star,\,
-A
\right\rangle_F
\le0.
$
By linearity of the Frobenius inner product in its second argument,
$
-
\left\langle
B-A^\star,\,
A
\right\rangle_F
\le0,
$
which is equivalent to
$
\left\langle
B-A^\star,\,
A
\right\rangle_F
\ge0.
$
Combining the two inequalities
$
\left\langle
B-A^\star,\,
A
\right\rangle_F
\le0
$
and
$
\left\langle
B-A^\star,\,
A
\right\rangle_F
\ge0
$
forces
\[
\boxed{
\left\langle
B-A^\star,\,
A
\right\rangle_F
=
0
}.
\]
Since \(A\in\mathcal A_{\mathfrak S}\) was arbitrary,
$
\left\langle
B-A^\star,\,
A
\right\rangle_F
=
0
\qquad
\text{for every }
A\in\mathcal A_{\mathfrak S}.
$

Thus, in the unconstrained case, the residual
$
B-A^\star
$
belongs to the Frobenius-orthogonal complement of the structured linear
subspace:
$
B-A^\star
\in
\mathcal A_{\mathfrak S}^{\perp}.
$
Equivalently,
$
B
=
A^\star+(B-A^\star)
$
is the orthogonal decomposition of \(B\) into its structured component
\(A^\star\in\mathcal A_{\mathfrak S}\) and an orthogonal residual. This
proves both statements of the proposition.
\end{proof}

\subsection{Proof of Corollary~\ref{cor:appendix_pythagorean_projection}}

\begin{proof}
Let
$
A^\star
=
\Pi_{\mathcal A_{\mathfrak S}}(B),
$
and fix an arbitrary
$
A\in\mathcal A_{\mathfrak S}.
$
For notational convenience, define
$
R
:=
B-A^\star
$
and
$
D
:=
A-A^\star.
$
Then
$
A
=
A^\star+D,
$
and therefore
\[
\begin{aligned}
B-A
&=
B-(A^\star+D)
\\
&=
(B-A^\star)-D
\\
&=
R-D.
\end{aligned}
\]
Using the Frobenius inner product,
$
\langle X,Y\rangle_F
=
\operatorname{tr}(X^\top Y),
$
the squared Frobenius norm of this difference is
\[
\begin{aligned}
\|B-A\|_F^2
&=
\|R-D\|_F^2
\\
&=
\langle R-D,R-D\rangle_F
\\
&=
\langle R,R\rangle_F
-
\langle R,D\rangle_F
-
\langle D,R\rangle_F
+
\langle D,D\rangle_F.
\end{aligned}
\]
The Frobenius inner product is symmetric over real matrices, so
$
\langle D,R\rangle_F
=
\langle R,D\rangle_F.
$
Hence
$
\|B-A\|_F^2
=
\|R\|_F^2
+
\|D\|_F^2
-
2\langle R,D\rangle_F.
$
Substituting the definitions of \(R\) and \(D\) gives the exact expansion
\[
\begin{aligned}
\|B-A\|_F^2
&=
\|B-A^\star\|_F^2
+
\|A-A^\star\|_F^2
\\
&\qquad
-
2
\left\langle
B-A^\star,\,
A-A^\star
\right\rangle_F.
\end{aligned}
\]

By Proposition~\ref{prop:appendix_projection_variational}, the metric
projection satisfies
$
\left\langle
B-A^\star,\,
A-A^\star
\right\rangle_F
\le0
\qquad
\text{for every }
A\in\mathcal A_{\mathfrak S}.
$
Multiplying this inequality by \(-2\) reverses its direction and gives
$
-2
\left\langle
B-A^\star,\,
A-A^\star
\right\rangle_F
\ge0.
$
Therefore the exact expansion above implies
\[
\begin{aligned}
\|B-A\|_F^2
&=
\|B-A^\star\|_F^2
+
\|A-A^\star\|_F^2
\\
&\qquad
-
2
\left\langle
B-A^\star,\,
A-A^\star
\right\rangle_F
\\
&\ge
\|B-A^\star\|_F^2
+
\|A-A^\star\|_F^2.
\end{aligned}
\]
Thus every
$
A\in\mathcal A_{\mathfrak S}
$
satisfies the Pythagorean inequality
\[
\boxed{
\|B-A\|_F^2
\ge
\|B-A^\star\|_F^2
+
\|A-A^\star\|_F^2
}.
\]

Suppose now that
$
\mathcal R_+=\varnothing.
$
By Proposition~\ref{prop:operators-interaction-family-structure},
\(\mathcal A_{\mathfrak S}\) is then a linear subspace of
\(\mathbb R^{d\times d}\). Since
$
A\in\mathcal A_{\mathfrak S}
$
and
$
A^\star\in\mathcal A_{\mathfrak S},
$
closure under subtraction gives
$
A-A^\star
\in
\mathcal A_{\mathfrak S}.
$
Proposition~\ref{prop:appendix_projection_variational} further states that,
in this unconstrained linear-subspace case,
$
\left\langle
B-A^\star,\,
C
\right\rangle_F
=
0
\qquad
\text{for every }
C\in\mathcal A_{\mathfrak S}.
$
Applying this orthogonality relation with
$
C=A-A^\star
$
gives
$
\left\langle
B-A^\star,\,
A-A^\star
\right\rangle_F
=
0.
$
Hence the cross term in the previous norm expansion vanishes exactly:
$
-2
\left\langle
B-A^\star,\,
A-A^\star
\right\rangle_F
=
0.
$
Consequently,
\[
\boxed{
\|B-A\|_F^2
=
\|B-A^\star\|_F^2
+
\|A-A^\star\|_F^2
}
\]
for every
$
A\in\mathcal A_{\mathfrak S}
$
whenever
\(\mathcal R_+=\varnothing\).

We finally make explicit the approximation interpretation of the two terms.
Since
$
A^\star
=
\Pi_{\mathcal A_{\mathfrak S}}(B),
$
the defining property of the metric projection gives
\[
\|B-A^\star\|_F
=
\inf_{C\in\mathcal A_{\mathfrak S}}
\|B-C\|_F
=
\operatorname{dist}
\bigl(
B,\mathcal A_{\mathfrak S}
\bigr).
\]
Therefore
$
\|B-A^\star\|_F^2
=
\operatorname{dist}
\bigl(
B,\mathcal A_{\mathfrak S}
\bigr)^2
$
is the smallest squared approximation error attainable by any operator in
the prescribed structured family. In this precise sense, it is the
irreducible structured approximation error associated with the descriptor.

For any other structured operator
$
A\in\mathcal A_{\mathfrak S},
$
the quantity
$
\|A-A^\star\|_F^2
$
measures its squared Frobenius deviation from the best structured target.
Indeed, subtracting the optimal approximation error from the exact expansion
gives
\[
\begin{aligned}
\|B-A\|_F^2
-
\|B-A^\star\|_F^2
&=
\|A-A^\star\|_F^2
\\
&\qquad
-
2
\left\langle
B-A^\star,\,
A-A^\star
\right\rangle_F.
\end{aligned}
\]
Since the variational inequality implies
$
-2
\left\langle
B-A^\star,\,
A-A^\star
\right\rangle_F
\ge0,
$
we further obtain
$
\|B-A\|_F^2
-
\|B-A^\star\|_F^2
\ge
\|A-A^\star\|_F^2.
$
Thus, in the general constrained convex case, deviation from
\(A^\star\) provides a lower bound on the excess approximation error. In
the unconstrained linear-subspace case, the projection residual is
orthogonal to the structured subspace, the cross term vanishes, and this
lower bound becomes the exact Pythagorean decomposition.
\end{proof}

\subsection{Proof of Theorem~\ref{thm:appendix_coeff_projection}}

\begin{proof}
Recall that
$
k=R+d,
$
that
$
Q_r=\Phi_r
\qquad
(1\le r\le R),
$
and
$
Q_{R+i}=E_{ii}
\qquad
(1\le i\le d).
$
The associated synthesis map is
$
T:\mathbb R^k\to\mathbb R^{d\times d},
\qquad
T\alpha
=
\sum_{a=1}^k\alpha_aQ_a,
$
and the feasible coefficient set is
\[
\mathcal C_\alpha
=
\left\{
(\theta,\delta)\in\mathbb R^{R+d}
\;\middle|\;
\theta_r\ge0
\text{ for }r\in\mathcal R_+
\right\}.
\]
By construction,
$
\mathcal A_{\mathfrak S}
=
T(\mathcal C_\alpha).
$

Define
$
G_{ab}
=
\langle Q_a,Q_b\rangle_F,
\qquad
b_a(B)
=
\langle B,Q_a\rangle_F.
$
We first record the basic structure of \(G\). Since the Frobenius inner
product is symmetric,
\[
G_{ab}
=
\langle Q_a,Q_b\rangle_F
=
\langle Q_b,Q_a\rangle_F
=
G_{ba},
\]
so
$
G=G^\top.
$
Moreover, for an arbitrary coefficient vector
$
c=(c_1,\ldots,c_k)^\top\in\mathbb R^k,
$
we have
\[
\begin{aligned}
c^\top Gc
&=
\sum_{a=1}^k
\sum_{b=1}^k
c_aG_{ab}c_b
\\
&=
\sum_{a=1}^k
\sum_{b=1}^k
c_ac_b
\langle Q_a,Q_b\rangle_F
\\
&=
\left\langle
\sum_{a=1}^kc_aQ_a,\,
\sum_{b=1}^kc_bQ_b
\right\rangle_F
\\
&=
\left\|
\sum_{a=1}^kc_aQ_a
\right\|_F^2
\\
&=
\|Tc\|_F^2
\\
&\ge0.
\end{aligned}
\]
Thus
$
G\succeq0.
$
In particular, the coefficient objective
$
q_B(\alpha)
:=
\frac12\alpha^\top G\alpha
-
b(B)^\top\alpha
$
is a convex quadratic function.

We now show that this coefficient objective is exactly the operator-level
projection objective up to a constant independent of \(\alpha\). For any
$
\alpha\in\mathbb R^k,
$
we have
$
T\alpha
=
\sum_{a=1}^k\alpha_aQ_a.
$
Therefore
\[
\begin{aligned}
\frac12
\|B-T\alpha\|_F^2
&=
\frac12
\langle
B-T\alpha,\,
B-T\alpha
\rangle_F
\\
&=
\frac12\|B\|_F^2
-
\langle B,T\alpha\rangle_F
+
\frac12\|T\alpha\|_F^2.
\end{aligned}
\]
The linear term satisfies
\[
\begin{aligned}
\langle B,T\alpha\rangle_F
&=
\left\langle
B,\,
\sum_{a=1}^k\alpha_aQ_a
\right\rangle_F
\\
&=
\sum_{a=1}^k
\alpha_a
\langle B,Q_a\rangle_F
\\
&=
\sum_{a=1}^k
\alpha_ab_a(B)
\\
&=
b(B)^\top\alpha.
\end{aligned}
\]
For the quadratic term,
\[
\begin{aligned}
\|T\alpha\|_F^2
&=
\left\langle
\sum_{a=1}^k\alpha_aQ_a,\,
\sum_{b=1}^k\alpha_bQ_b
\right\rangle_F
\\
&=
\sum_{a=1}^k
\sum_{b=1}^k
\alpha_a\alpha_b
\langle Q_a,Q_b\rangle_F
\\
&=
\sum_{a=1}^k
\sum_{b=1}^k
\alpha_aG_{ab}\alpha_b
\\
&=
\alpha^\top G\alpha.
\end{aligned}
\]
Combining these identities gives
\[
\boxed{
\frac12
\|B-T\alpha\|_F^2
=
\frac12\|B\|_F^2
+
\frac12\alpha^\top G\alpha
-
b(B)^\top\alpha
}.
\]
Equivalently,
$
q_B(\alpha)
=
\frac12
\|B-T\alpha\|_F^2
-
\frac12\|B\|_F^2.
$
Since
$
\frac12\|B\|_F^2
$
does not depend on \(\alpha\), minimizing \(q_B\) over
\(\mathcal C_\alpha\) is equivalent to minimizing
$
\frac12\|B-T\alpha\|_F^2
$
over the same feasible coefficient set.

Let
$
A^\star
:=
\Pi_{\mathcal A_{\mathfrak S}}(B).
$
Because
\(\mathcal A_{\mathfrak S}\) is a nonempty closed convex subset of
\(\mathbb R^{d\times d}\),
the metric projection exists and is unique. Moreover,
$
A^\star\in\mathcal A_{\mathfrak S}.
$
Since
$
\mathcal A_{\mathfrak S}
=
T(\mathcal C_\alpha),
$
there exists at least one
$
\bar\alpha\in\mathcal C_\alpha
$
such that
$
T\bar\alpha
=
A^\star.
$
For this coefficient vector,
\[
\|B-T\bar\alpha\|_F^2
=
\|B-A^\star\|_F^2.
\]
By the defining optimality of \(A^\star\),
$
\|B-A^\star\|_F^2
\le
\|B-A\|_F^2
\qquad
\text{for every }
A\in\mathcal A_{\mathfrak S}.
$
Every
$
\alpha\in\mathcal C_\alpha
$
produces a matrix
$
T\alpha\in\mathcal A_{\mathfrak S},
$
so
$
\|B-T\bar\alpha\|_F^2
\le
\|B-T\alpha\|_F^2
\qquad
\text{for every }
\alpha\in\mathcal C_\alpha.
$
Hence \(\bar\alpha\) minimizes the coefficient problem. In particular, the
coefficient quadratic program always has at least one minimizer, even when
the Gram matrix \(G\) is singular.

Now let
$
\alpha^\star\in\mathcal C_\alpha
$
be any minimizer of the coefficient quadratic program. By the objective
identity above, it also minimizes
$
\|B-T\alpha\|_F^2
$
over
\(\alpha\in\mathcal C_\alpha\). Hence the matrix
$
T\alpha^\star
\in
\mathcal A_{\mathfrak S}
$
minimizes
$
\|B-A\|_F^2
$
over
$
A\in\mathcal A_{\mathfrak S}.
$
The operator-level minimizer is unique, so necessarily
\[
\boxed{
T\alpha^\star
=
A^\star
=
\Pi_{\mathcal A_{\mathfrak S}}(B)
}.
\]

Conversely, let
$
\alpha\in\mathcal C_\alpha
$
be any feasible coefficient vector satisfying
$
T\alpha
=
A^\star.
$
Then
$
\|B-T\alpha\|_F^2
=
\|B-A^\star\|_F^2,
$
which is the minimum possible Frobenius approximation error over the
structured family. Therefore \(\alpha\) minimizes
$
\frac12\|B-T\alpha\|_F^2,
$
and hence also minimizes
$
q_B(\alpha)
=
\frac12\alpha^\top G\alpha
-
b(B)^\top\alpha.
$
Thus the set of coefficient minimizers is exactly
\[
\left\{
\alpha\in\mathcal C_\alpha:
T\alpha
=
\Pi_{\mathcal A_{\mathfrak S}}(B)
\right\}.
\]

This equality also explains the distinction between matrix uniqueness and
coefficient uniqueness. If
$
\alpha^{(1)}
\qquad\text{and}\qquad
\alpha^{(2)}
$
are two coefficient minimizers, then both represent the same unique
projected matrix:
$
T\alpha^{(1)}
=
T\alpha^{(2)}
=
A^\star.
$
Subtracting gives
$
T
\left(
\alpha^{(1)}-\alpha^{(2)}
\right)
=
0.
$
Thus different minimizing coefficient vectors can occur only through
redundancy in the synthesis map \(T\), subject of course to both vectors
remaining feasible. The projected matrix itself remains unique regardless
of such coefficient redundancy.

Suppose now that
$
\mathcal R_+=\varnothing.
$
Then there are no sign restrictions on the coefficients, and therefore
$
\mathcal C_\alpha
=
\mathbb R^k.
$
We prove that the minimizers are exactly the solutions of
$
G\alpha=b(B).
$

Let
$
\alpha^\star\in\mathbb R^k
$
be a minimizer of \(q_B\), and let
$
z\in\mathbb R^k
$
be arbitrary. Since the problem is unconstrained,
$
\alpha^\star+tz
$
is feasible for every
$
t\in\mathbb R.
$
Using symmetry of \(G\),
\[
\begin{aligned}
q_B(\alpha^\star+tz)
&=
\frac12
(\alpha^\star+tz)^\top
G
(\alpha^\star+tz)
-
b(B)^\top
(\alpha^\star+tz)
\\
&=
\frac12
(\alpha^\star)^\top
G\alpha^\star
+
t z^\top G\alpha^\star
+
\frac{t^2}{2}z^\top Gz
\\
&\qquad
-
b(B)^\top\alpha^\star
-
t\,b(B)^\top z
\\
&=
q_B(\alpha^\star)
+
t\,z^\top
\bigl(
G\alpha^\star-b(B)
\bigr)
+
\frac{t^2}{2}
z^\top Gz.
\end{aligned}
\]
Since \(\alpha^\star\) is a minimizer,
$
q_B(\alpha^\star+tz)
-
q_B(\alpha^\star)
\ge0
$
for every \(t\in\mathbb R\). Hence
$
t\,z^\top
\bigl(
G\alpha^\star-b(B)
\bigr)
+
\frac{t^2}{2}
z^\top Gz
\ge0.
$
For \(t>0\), division by \(t\) gives
$
z^\top
\bigl(
G\alpha^\star-b(B)
\bigr)
+
\frac{t}{2}
z^\top Gz
\ge0.
$
Letting
$
t\downarrow0
$
yields
$
z^\top
\bigl(
G\alpha^\star-b(B)
\bigr)
\ge0.
$
Applying the same argument with the direction \(-z\) gives
$
-z^\top
\bigl(
G\alpha^\star-b(B)
\bigr)
\ge0,
$
or equivalently
$
z^\top
\bigl(
G\alpha^\star-b(B)
\bigr)
\le0.
$
Therefore
$
z^\top
\bigl(
G\alpha^\star-b(B)
\bigr)
=
0
$
for every
$
z\in\mathbb R^k.
$
The only vector orthogonal to every vector in \(\mathbb R^k\) is the zero
vector, so
\[
\boxed{
G\alpha^\star=b(B)
}.
\]

Conversely, suppose
$
\alpha^\star\in\mathbb R^k
$
satisfies
$
G\alpha^\star=b(B).
$
Let
$
\beta\in\mathbb R^k
$
be arbitrary and define
$
d:=\beta-\alpha^\star.
$
Then
$
\beta=\alpha^\star+d.
$
Expanding the difference of the two quadratic objectives gives
\[
\begin{aligned}
q_B(\beta)-q_B(\alpha^\star)
&=
\frac12
(\alpha^\star+d)^\top
G
(\alpha^\star+d)
-
b(B)^\top(\alpha^\star+d)
\\
&\qquad
-
\frac12
(\alpha^\star)^\top
G\alpha^\star
+
b(B)^\top\alpha^\star
\\
&=
d^\top G\alpha^\star
+
\frac12d^\top Gd
-
b(B)^\top d.
\end{aligned}
\]
Using
$
G\alpha^\star=b(B),
$
the two linear terms cancel:
\[
d^\top G\alpha^\star
-
b(B)^\top d
=
d^\top b(B)-b(B)^\top d
=
0.
\]
Hence
$
q_B(\beta)-q_B(\alpha^\star)
=
\frac12d^\top Gd.
$
Since
$
G\succeq0,
$
we have
$
d^\top Gd\ge0.
$
Therefore
$
q_B(\beta)
\ge
q_B(\alpha^\star)
\qquad
\text{for every }\beta\in\mathbb R^k.
$
Thus every solution of
$
G\alpha^\star=b(B)
$
is a global minimizer.

We have therefore proved the exact equivalence
\[
\boxed{
\alpha^\star
\text{ minimizes }q_B
\quad\Longleftrightarrow\quad
G\alpha^\star=b(B)
}
\]
when
$
\mathcal R_+=\varnothing.
$
The normal equations have at least one solution even if \(G\) is singular:
as proved above, the operator projection \(A^\star\) exists, it admits at
least one coefficient representation
\(\bar\alpha\in\mathbb R^k\), and every such representation is a minimizer;
hence it necessarily satisfies the normal equations.

Finally, assume that
$
Q_1,\ldots,Q_k
$
are linearly independent. Let
$
c\in\mathbb R^k
$
be nonzero. Then linear independence implies
$
\sum_{a=1}^kc_aQ_a
\neq0.
$
Using the Gram identity derived at the beginning of the proof,
\[
\begin{aligned}
c^\top Gc
&=
\left\|
\sum_{a=1}^kc_aQ_a
\right\|_F^2
\\
&>
0.
\end{aligned}
\]
Thus
$
G\succ0.
$

It follows that the quadratic function
$
q_B(\alpha)
=
\frac12\alpha^\top G\alpha
-
b(B)^\top\alpha
$
is strictly convex. Indeed, let
$
\alpha\neq\beta
$
and
$
t\in(0,1).
$
A direct quadratic expansion gives
\[
\begin{aligned}
&t\,q_B(\alpha)
+
(1-t)q_B(\beta)
-
q_B\bigl(t\alpha+(1-t)\beta\bigr)
\\
&\qquad=
\frac12
t(1-t)
(\alpha-\beta)^\top
G
(\alpha-\beta).
\end{aligned}
\]
Since
$
\alpha-\beta\neq0
$
and
$
G\succ0,
$
the right-hand side is strictly positive. Hence
\[
q_B\bigl(t\alpha+(1-t)\beta\bigr)
<
t\,q_B(\alpha)
+
(1-t)q_B(\beta),
\]
which is precisely strict convexity.

The feasible set
$
\mathcal C_\alpha
=
\left\{
(\theta,\delta):
\theta_r\ge0
\text{ for }r\in\mathcal R_+
\right\}
$
is convex. We have already established that a minimizer exists. If there
were two distinct minimizers
$
\alpha^{(1)}\neq\alpha^{(2)}
$
in \(\mathcal C_\alpha\), then their midpoint
$
\frac12
\alpha^{(1)}
+
\frac12
\alpha^{(2)}
$
would also lie in
\(\mathcal C_\alpha\). Strict convexity would then give
\[
\begin{aligned}
q_B\left(
\frac{
\alpha^{(1)}+\alpha^{(2)}
}{2}
\right)
&<
\frac12
q_B(\alpha^{(1)})
+
\frac12
q_B(\alpha^{(2)}),
\end{aligned}
\]
which would be strictly smaller than the common minimum value, a
contradiction. Therefore the constrained coefficient minimizer is unique.

The same argument applies in the unconstrained case
$
\mathcal C_\alpha=\mathbb R^k.
$
Equivalently, when \(G\succ0\), the normal equations have the unique
solution
$
\alpha^\star
=
G^{-1}b(B).
$
Thus linear independence of the generating matrices yields uniqueness of the
coefficient vector in both the constrained and unconstrained projection
problems, completing the proof.
\end{proof}

\subsection{Proof of Proposition~\ref{prop:appendix_kkt_constrained}}

\begin{proof}
Let
$
q_B(\alpha)
:=
\frac12\alpha^\top G\alpha
-
b(B)^\top\alpha,
\qquad
\alpha\in\mathbb R^k.
$
Recall that the feasible coefficient set is
$
\mathcal C_\alpha
=
\left\{
\alpha\in\mathbb R^k
\;\middle|\;
\alpha_r\ge0
\text{ for every }r\in\mathcal R_+
\right\}.
$
The coordinates indexed by
\(\mathcal R_+\) are therefore constrained to be nonnegative, whereas all
remaining coordinates are unrestricted.

By Theorem~\ref{thm:appendix_coeff_projection}, linear independence of
$
Q_1,\ldots,Q_k
$
implies
$
G\succ0.
$
For completeness, this follows from
$
c^\top Gc
=
\left\|
\sum_{a=1}^kc_aQ_a
\right\|_F^2
$
for every \(c\in\mathbb R^k\): if \(c\neq0\), linear independence gives
$
\sum_{a=1}^kc_aQ_a\neq0,
$
and consequently
$
c^\top Gc>0.
$
Thus \(q_B\) is a strictly convex quadratic function. In particular, if a
minimizer over the convex feasible set \(\mathcal C_\alpha\) exists, it is
unique.

We first prove necessity of the stated conditions. Let
$
\alpha^\star\in\mathcal C_\alpha
$
be the unique minimizer of \(q_B\) over
\(\mathcal C_\alpha\), and define
$
g^\star
:=
G\alpha^\star-b(B).
$
Since \(G\) is symmetric, \(g^\star\) is precisely the gradient of the
quadratic objective at \(\alpha^\star\):
$
\nabla q_B(\alpha^\star)
=
G\alpha^\star-b(B)
=
g^\star.
$

Fix a coordinate
$
a\notin\mathcal R_+.
$
This coordinate is unconstrained. Hence, for every
\(t\in\mathbb R\),
$
\alpha^\star+te_a
\in\mathcal C_\alpha,
$
where \(e_a\) denotes the \(a\)-th standard basis vector of
\(\mathbb R^k\). The optimality of \(\alpha^\star\) therefore gives
$
q_B(\alpha^\star+te_a)
-
q_B(\alpha^\star)
\ge0
\qquad
\text{for every }t\in\mathbb R.
$
We expand this difference explicitly:
\[
\begin{aligned}
q_B(\alpha^\star+te_a)
&=
\frac12
(\alpha^\star+te_a)^\top
G
(\alpha^\star+te_a)
-
b(B)^\top(\alpha^\star+te_a)
\\
&=
\frac12
(\alpha^\star)^\top G\alpha^\star
+
t\,e_a^\top G\alpha^\star
+
\frac{t^2}{2}e_a^\top Ge_a
\\
&\qquad
-
b(B)^\top\alpha^\star
-
t\,b(B)^\top e_a.
\end{aligned}
\]
Subtracting \(q_B(\alpha^\star)\) gives
$
q_B(\alpha^\star+te_a)
-
q_B(\alpha^\star)
=
t\,e_a^\top
\bigl(
G\alpha^\star-b(B)
\bigr)
+
\frac{t^2}{2}e_a^\top Ge_a.
$
By the definition of \(g^\star\),
$
q_B(\alpha^\star+te_a)
-
q_B(\alpha^\star)
=
t\,g_a^\star
+
\frac{t^2}{2}G_{aa}.
$
For \(t>0\), optimality therefore implies
$
g_a^\star
+
\frac{t}{2}G_{aa}
\ge0.
$
Letting \(t\downarrow0\) yields
$
g_a^\star\ge0.
$
Because the coordinate is unrestricted, we may also take \(t<0\). Write
\(t=-s\) with \(s>0\). Then
$
-s\,g_a^\star
+
\frac{s^2}{2}G_{aa}
\ge0.
$
Dividing by \(s>0\) gives
$
-g_a^\star
+
\frac{s}{2}G_{aa}
\ge0.
$
Letting \(s\downarrow0\) yields
$
-g_a^\star\ge0,
$
or equivalently
$
g_a^\star\le0.
$
Combining the two inequalities gives
$
g_a^\star=0
\qquad
\text{for every }a\notin\mathcal R_+.
$

Now fix a constrained coordinate
$
r\in\mathcal R_+.
$
By primal feasibility,
$
\alpha_r^\star\ge0.
$
We distinguish the two possibilities allowed by this constraint.

Suppose first that
$
\alpha_r^\star>0.
$
Then there exists
$
\varepsilon_r>0
$
such that
$
\alpha_r^\star-t\ge0
\qquad
\text{whenever }0\le t\le\varepsilon_r.
$
Hence both
$
\alpha^\star+te_r
$
and
$
\alpha^\star-te_r
$
remain feasible for all sufficiently small \(t>0\). Applying exactly the
same one-dimensional argument as above in both directions gives
$
g_r^\star\ge0
$
from the positive direction and
$
g_r^\star\le0
$
from the negative direction. Thus
$
g_r^\star=0
\qquad
\text{whenever }\alpha_r^\star>0.
$

Suppose instead that
$
\alpha_r^\star=0.
$
The negative direction is no longer feasible, but the positive direction
always is:
$
\alpha^\star+te_r
\in\mathcal C_\alpha
\qquad
\text{for every }t\ge0.
$
Optimality gives
$
q_B(\alpha^\star+te_r)
-
q_B(\alpha^\star)
\ge0
\qquad
(t\ge0).
$
As before,
\[
q_B(\alpha^\star+te_r)
-
q_B(\alpha^\star)
=
t\,g_r^\star
+
\frac{t^2}{2}G_{rr}.
\]
For \(t>0\), division by \(t\) gives
$
g_r^\star
+
\frac{t}{2}G_{rr}
\ge0.
$
Letting \(t\downarrow0\), we obtain
$
g_r^\star\ge0.
$

We can therefore define, for every
\(r\in\mathcal R_+\),
$
\lambda_r
:=
g_r^\star.
$
The preceding coordinatewise analysis shows that
$
\lambda_r\ge0
\qquad
\text{for every }r\in\mathcal R_+.
$
Moreover, if
$
\alpha_r^\star>0,
$
then
$
\lambda_r=g_r^\star=0,
$
whereas if
$
\alpha_r^\star=0,
$
the product
$
\lambda_r\alpha_r^\star
$
vanishes trivially. Hence, in all cases,
$
\lambda_r\alpha_r^\star=0
\qquad
\text{for every }r\in\mathcal R_+.
$
This is complementary slackness.

We have also proved that
$
g_a^\star=0
\qquad
\text{for }a\notin\mathcal R_+,
$
while
$
g_r^\star=\lambda_r
\qquad
\text{for }r\in\mathcal R_+.
$
Therefore the full gradient vector satisfies
$
g^\star
=
\sum_{r\in\mathcal R_+}
\lambda_re_r.
$
Recalling that
$
g^\star
=
G\alpha^\star-b(B),
$
we obtain
$
G\alpha^\star-b(B)
-
\sum_{r\in\mathcal R_+}
\lambda_re_r
=
0.
$
Thus stationarity holds. Together with the already assumed primal
feasibility
$
\alpha_r^\star\ge0,
\qquad
r\in\mathcal R_+,
$
the inequalities
$
\lambda_r\ge0,
\qquad
r\in\mathcal R_+,
$
and complementary slackness
$
\lambda_r\alpha_r^\star=0,
\qquad
r\in\mathcal R_+,
$
this proves necessity of all stated KKT conditions.

We now prove sufficiency. Suppose that
$
\alpha^\star\in\mathcal C_\alpha
$
is feasible and that there exist multipliers
$
\lambda_r\ge0,
\qquad
r\in\mathcal R_+,
$
satisfying
$
G\alpha^\star-b(B)
-
\sum_{r\in\mathcal R_+}
\lambda_re_r
=
0
$
and
$
\lambda_r\alpha_r^\star=0
\qquad
\text{for every }r\in\mathcal R_+.
$
Let
$
\alpha\in\mathcal C_\alpha
$
be any other feasible coefficient vector, and define
$
d
:=
\alpha-\alpha^\star.
$
Using symmetry of \(G\), an exact quadratic expansion gives
\[
\begin{aligned}
q_B(\alpha)
&=
q_B(\alpha^\star+d)
\\
&=
\frac12
(\alpha^\star+d)^\top
G
(\alpha^\star+d)
-
b(B)^\top(\alpha^\star+d)
\\
&=
\frac12
(\alpha^\star)^\top
G\alpha^\star
-
b(B)^\top\alpha^\star
\\
&\qquad
+
d^\top
\bigl(
G\alpha^\star-b(B)
\bigr)
+
\frac12d^\top Gd.
\end{aligned}
\]
Therefore
$
q_B(\alpha)-q_B(\alpha^\star)
=
d^\top
\bigl(
G\alpha^\star-b(B)
\bigr)
+
\frac12d^\top Gd.
$
By stationarity,
$
G\alpha^\star-b(B)
=
\sum_{r\in\mathcal R_+}
\lambda_re_r.
$
Hence
\[
\begin{aligned}
d^\top
\bigl(
G\alpha^\star-b(B)
\bigr)
&=
d^\top
\sum_{r\in\mathcal R_+}
\lambda_re_r
\\
&=
\sum_{r\in\mathcal R_+}
\lambda_r d_r
\\
&=
\sum_{r\in\mathcal R_+}
\lambda_r
\left(
\alpha_r-\alpha_r^\star
\right).
\end{aligned}
\]
Using complementary slackness,
$
\lambda_r\alpha_r^\star=0,
$
we obtain
\[
\begin{aligned}
\sum_{r\in\mathcal R_+}
\lambda_r
\left(
\alpha_r-\alpha_r^\star
\right)
&=
\sum_{r\in\mathcal R_+}
\lambda_r\alpha_r
-
\sum_{r\in\mathcal R_+}
\lambda_r\alpha_r^\star
\\
&=
\sum_{r\in\mathcal R_+}
\lambda_r\alpha_r.
\end{aligned}
\]
Because \(\alpha\) is feasible,
$
\alpha_r\ge0
\qquad
\text{for every }r\in\mathcal R_+,
$
and by dual feasibility,
$
\lambda_r\ge0.
$
Therefore
$
\sum_{r\in\mathcal R_+}
\lambda_r\alpha_r
\ge0.
$
It follows that
$
d^\top
\bigl(
G\alpha^\star-b(B)
\bigr)
\ge0.
$

Furthermore,
$
G\succ0.
$
Thus
$
d^\top Gd\ge0,
$
with equality if and only if
$
d=0.
$
Combining these estimates yields
\[
\begin{aligned}
q_B(\alpha)-q_B(\alpha^\star)
&=
\sum_{r\in\mathcal R_+}
\lambda_r\alpha_r
+
\frac12d^\top Gd
\\
&\ge0.
\end{aligned}
\]
Hence
$
q_B(\alpha)
\ge
q_B(\alpha^\star)
\qquad
\text{for every }\alpha\in\mathcal C_\alpha.
$
Therefore \(\alpha^\star\) is a global minimizer.

We can strengthen this conclusion to uniqueness directly. Suppose
$
\alpha\neq\alpha^\star.
$
Then
$
d=\alpha-\alpha^\star\neq0.
$
Since
$
G\succ0,
$
we have
$
d^\top Gd>0.
$
Consequently,
\[
\begin{aligned}
q_B(\alpha)-q_B(\alpha^\star)
&=
\sum_{r\in\mathcal R_+}
\lambda_r\alpha_r
+
\frac12d^\top Gd
\\
&>
0,
\end{aligned}
\]
because the first term is nonnegative and the second is strictly positive.
Thus every feasible vector distinct from \(\alpha^\star\) has strictly
larger objective value. Hence \(\alpha^\star\) is the unique minimizer.

This proves that a feasible vector is the unique minimizer if and only if
there exist nonnegative multipliers satisfying stationarity, primal
feasibility, dual feasibility, and complementary slackness exactly as stated.

Finally, consider the unconstrained case
$
\mathcal R_+=\varnothing.
$
There are then no multiplier variables and no inequality constraints. The
sum appearing in stationarity is an empty sum, hence equals the zero vector:
$
\sum_{r\in\varnothing}\lambda_re_r=0.
$
Therefore the stationarity condition becomes simply
$
G\alpha^\star-b(B)=0,
$
or equivalently,
$
G\alpha^\star=b(B).
$
These are precisely the normal equations in
Theorem~\ref{thm:appendix_coeff_projection}. Since
\(G\succ0\) under the present linear-independence assumption, they admit the
unique solution
$
\alpha^\star
=
G^{-1}b(B).
$
This completes the proof.
\end{proof}

\subsection{Proof of Theorem~\ref{thm:appendix_pointwise_projection}}

\begin{proof}
For notational convenience, write
$
\mathcal C
:=
\mathcal A_{\mathfrak S}
\subseteq
\mathbb R^{d\times d},
$
and let
$
P
:=
\Pi_{\mathcal C}
$
denote the Frobenius metric projection onto \(\mathcal C\).
By Proposition~\ref{prop:operators-interaction-family-structure},
\(\mathcal C\) is a nonempty closed convex subset of
\(\mathbb R^{d\times d}\). In particular, the zero matrix belongs to
\(\mathcal C\), since it is obtained in
Definition~\ref{def:operators-interaction-family} by taking
$
\theta=0,
\qquad
\delta=0.
$
Moreover, by
Proposition~\ref{prop:appendix_projection_nonexpansive},
the projection map \(P\) is \(1\)-Lipschitz with respect to the Frobenius
norm:
\[
\|P(M_1)-P(M_2)\|_F
\le
\|M_1-M_2\|_F
\qquad
\text{for all }
M_1,M_2\in\mathbb R^{d\times d}.
\]

We first establish the basic geometric properties of
\(\mathcal F_{\mathfrak S}\). Recall that
\[
\mathcal F_{\mathfrak S}
=
\left\{
A\in
L^2\!\left(
K,\mu;\mathbb R^{d\times d}
\right)
\;\middle|\;
A(x)\in\mathcal C
\text{ for }\mu\text{-almost every }x
\right\}.
\]
This set is nonempty. Indeed, consider the constant zero field
$
A_0(x):=0
\qquad
\text{for every }x\in K.
$
Since \(\mu\) is a probability measure,
\[
\int_K
\|A_0(x)\|_F^2
\,d\mu(x)
=
0,
\]
so
$
A_0
\in
L^2\!\left(
K,\mu;\mathbb R^{d\times d}
\right).
$
Because
$
0\in\mathcal C,
$
we also have
$
A_0(x)\in\mathcal C
\qquad
\text{for every }x\in K.
$
Hence
$
A_0\in\mathcal F_{\mathfrak S},
$
and therefore
$
\mathcal F_{\mathfrak S}\neq\varnothing.
$

We next prove convexity. Let
$
A_1,A_2\in\mathcal F_{\mathfrak S},
\qquad
t\in[0,1].
$
Since \(A_1,A_2\in L^2\), their convex combination
$
A_t
:=
(1-t)A_1+tA_2
$
also belongs to
$
L^2\!\left(
K,\mu;\mathbb R^{d\times d}
\right).
$
By the definition of \(\mathcal F_{\mathfrak S}\), there exist
\(\mu\)-null sets \(N_1,N_2\subseteq K\) such that
$
A_1(x)\in\mathcal C
\qquad
\text{for }x\in K\setminus N_1
$
and
$
A_2(x)\in\mathcal C
\qquad
\text{for }x\in K\setminus N_2.
$
The union
$
N:=N_1\cup N_2
$
is again a \(\mu\)-null set. For every
$
x\in K\setminus N,
$
both \(A_1(x)\) and \(A_2(x)\) belong to the convex set \(\mathcal C\).
Consequently,
\[
A_t(x)
=
(1-t)A_1(x)+tA_2(x)
\in
\mathcal C.
\]
Thus
$
A_t(x)\in\mathcal C
\qquad
\text{for }\mu\text{-almost every }x,
$
and hence
$
A_t\in\mathcal F_{\mathfrak S}.
$
Therefore \(\mathcal F_{\mathfrak S}\) is convex.

We now prove that \(\mathcal F_{\mathfrak S}\) is closed in
\(L^2(K,\mu;\mathbb R^{d\times d})\). Let
$
A_n\in\mathcal F_{\mathfrak S}
$
and suppose that
$
A_n\longrightarrow A
$
in
$
L^2\!\left(
K,\mu;\mathbb R^{d\times d}
\right).
$
We must show that
$
A\in\mathcal F_{\mathfrak S}.
$

Define the distance from a matrix \(M\in\mathbb R^{d\times d}\) to
\(\mathcal C\) by
$
d_{\mathcal C}(M)
:=
\inf_{C\in\mathcal C}
\|M-C\|_F.
$
Since \(P(M)\) is the metric projection of \(M\) onto \(\mathcal C\),
$
d_{\mathcal C}(M)
=
\|M-P(M)\|_F.
$
In particular, because \(P\) is continuous,
$
d_{\mathcal C}:
\mathbb R^{d\times d}\to\mathbb R
$
is continuous and hence Borel measurable.

For each \(n\), since
\(A_n\in\mathcal F_{\mathfrak S}\), there exists a null set \(N_n\)
such that
$
A_n(x)\in\mathcal C
\qquad
\text{for every }x\in K\setminus N_n.
$
Let
$
N:=\bigcup_{n=1}^{\infty}N_n.
$
Because this is a countable union of null sets,
$
\mu(N)=0.
$
Hence, for every \(x\in K\setminus N\) and every \(n\),
$
A_n(x)\in\mathcal C.
$
By the defining property of distance to a set,
$
d_{\mathcal C}(A(x))
\le
\|A(x)-A_n(x)\|_F
\qquad
\text{for every }x\in K\setminus N.
$
Squaring both sides gives
$
d_{\mathcal C}(A(x))^2
\le
\|A(x)-A_n(x)\|_F^2
\qquad
\text{for }\mu\text{-almost every }x.
$
Integrating over \(K\),
\[
\int_K
d_{\mathcal C}(A(x))^2
\,d\mu(x)
\le
\int_K
\|A(x)-A_n(x)\|_F^2
\,d\mu(x).
\]
The right-hand side is exactly
$
\|A-A_n\|_{L^2(K,\mu;F)}^2,
$
and by assumption
$
\|A-A_n\|_{L^2(K,\mu;F)}^2
\longrightarrow
0.
$
Therefore
\[
\int_K
d_{\mathcal C}(A(x))^2
\,d\mu(x)
=
0.
\]
The integrand is nonnegative. A nonnegative measurable function with zero
integral must vanish almost everywhere, so
$
d_{\mathcal C}(A(x))=0
\qquad
\text{for }\mu\text{-almost every }x.
$
Because \(\mathcal C\) is closed,
$
d_{\mathcal C}(M)=0
\quad\Longleftrightarrow\quad
M\in\mathcal C.
$
Consequently,
$
A(x)\in\mathcal C
\qquad
\text{for }\mu\text{-almost every }x.
$
Thus
$
A\in\mathcal F_{\mathfrak S},
$
and \(\mathcal F_{\mathfrak S}\) is closed.

Since
$
L^2\!\left(
K,\mu;\mathbb R^{d\times d}
\right)
$
is a Hilbert space under the inner product
\[
\langle A_1,A_2\rangle_{L^2}
:=
\int_K
\langle A_1(x),A_2(x)\rangle_F
\,d\mu(x),
\]
we have now established that
\(\mathcal F_{\mathfrak S}\) is a nonempty closed convex subset of the
ambient Hilbert space.

We next construct the claimed minimizer. Choose a measurable representative
of
$
B\in
L^2\!\left(
K,\mu;\mathbb R^{d\times d}
\right)
$
and define
\[
A^\star(x)
:=
P(B(x))
=
\Pi_{\mathcal A_{\mathfrak S}}(B(x)).
\]
Because \(B\) is measurable and
$
P:
\mathbb R^{d\times d}
\to
\mathbb R^{d\times d}
$
is continuous by
Proposition~\ref{prop:appendix_projection_nonexpansive},
the composition
$
x\longmapsto P(B(x))
$
is measurable. Thus \(A^\star\) is a measurable matrix-valued field.

We next verify square integrability. Since
$
0\in\mathcal C,
$
the unique projection of the zero matrix onto \(\mathcal C\) is the zero
matrix itself:
$
P(0)=0.
$
By nonexpansiveness,
\[
\begin{aligned}
\|A^\star(x)\|_F
&=
\|P(B(x))\|_F
\\
&=
\|P(B(x))-P(0)\|_F
\\
&\le
\|B(x)-0\|_F
\\
&=
\|B(x)\|_F.
\end{aligned}
\]
Squaring and integrating gives
\[
\int_K
\|A^\star(x)\|_F^2
\,d\mu(x)
\le
\int_K
\|B(x)\|_F^2
\,d\mu(x).
\]
The right-hand side is finite because \(B\in L^2\). Therefore
$
A^\star
\in
L^2\!\left(
K,\mu;\mathbb R^{d\times d}
\right).
$
Moreover, by the definition of the metric projection,
$
P(B(x))\in\mathcal C
$
for every \(x\) at which the chosen representative \(B(x)\) is defined.
In particular,
$
A^\star(x)\in\mathcal C
\qquad
\text{for }\mu\text{-almost every }x.
$
Hence
$
A^\star\in\mathcal F_{\mathfrak S}.
$

The definition is also independent of the chosen representative of \(B\)
up to almost-everywhere equality. Indeed, if \(B_1\) and \(B_2\) are two
representatives of the same \(L^2\) element, then
$
B_1(x)=B_2(x)
\qquad
\text{for }\mu\text{-almost every }x.$
Applying the single-valued map \(P\) gives
$
P(B_1(x))
=
P(B_2(x))
\qquad
\text{for }\mu\text{-almost every }x.
$
Thus the \(L^2\)-equivalence class of \(A^\star\) is well defined.

We now prove optimality. Let
$
A\in\mathcal F_{\mathfrak S}
$
be arbitrary. Then
$
A(x)\in\mathcal C
\qquad
\text{for }\mu\text{-almost every }x.
$
For every such \(x\),
$
A^\star(x)
=
P(B(x))
$
is the unique Frobenius metric projection of \(B(x)\) onto \(\mathcal C\).
Therefore
\[
\|B(x)-A^\star(x)\|_F
\le
\|B(x)-A(x)\|_F
\]
for \(\mu\)-almost every \(x\). Squaring preserves the inequality because
both sides are nonnegative:
$
\|B(x)-A^\star(x)\|_F^2
\le
\|B(x)-A(x)\|_F^2
\qquad
\text{for }\mu\text{-almost every }x.
$
Both sides are integrable. Indeed, \(B,A,A^\star\in L^2\), and hence their
pairwise differences also belong to \(L^2\). Integrating the pointwise
inequality gives
\[
\int_K
\|B(x)-A^\star(x)\|_F^2
\,d\mu(x)
\le
\int_K
\|B(x)-A(x)\|_F^2
\,d\mu(x).
\]
Since \(A\in\mathcal F_{\mathfrak S}\) was arbitrary,
\(A^\star\) is a minimizer of the field-level projection problem.

We next prove uniqueness up to \(\mu\)-almost-everywhere equality. Let
$
\widetilde A\in\mathcal F_{\mathfrak S}
$
be any other minimizer. For almost every \(x\),
$
\widetilde A(x)\in\mathcal C
$
and
$
A^\star(x)
=
P(B(x)).
$
Applying
Corollary~\ref{cor:appendix_pythagorean_projection}
pointwise with
$
B=B(x),
\qquad
A=\widetilde A(x),
\qquad
A^\star=A^\star(x),
$
gives
\[
\begin{aligned}
\|B(x)-\widetilde A(x)\|_F^2
&\ge
\|B(x)-A^\star(x)\|_F^2
\\
&\qquad
+
\|\widetilde A(x)-A^\star(x)\|_F^2
\end{aligned}
\]
for \(\mu\)-almost every \(x\). Integrating,
\[
\begin{aligned}
\int_K
\|B(x)-\widetilde A(x)\|_F^2
\,d\mu(x)
&\ge
\int_K
\|B(x)-A^\star(x)\|_F^2
\,d\mu(x)
\\
&\qquad
+
\int_K
\|\widetilde A(x)-A^\star(x)\|_F^2
\,d\mu(x).
\end{aligned}
\]
Because both
\(\widetilde A\) and \(A^\star\) are minimizers, their objective values are
equal. Hence
$
\int_K
\|\widetilde A(x)-A^\star(x)\|_F^2
\,d\mu(x)
\le0.
$
The integrand is nonnegative, so the integral is also nonnegative. Therefore
the only possibility is
$
\int_K
\|\widetilde A(x)-A^\star(x)\|_F^2
\,d\mu(x)
=
0.
$
It follows that
$
\|\widetilde A(x)-A^\star(x)\|_F^2
=
0
\qquad
\text{for }\mu\text{-almost every }x,
$
and hence
$
\widetilde A(x)
=
A^\star(x)
\qquad
\text{for }\mu\text{-almost every }x.
$
Thus the minimizer is unique as an element of the
\(L^2\) Hilbert space, equivalently unique up to
\(\mu\)-almost-everywhere equality.

It remains to establish the continuity statement. Suppose that the
\(L^2\)-equivalence class of \(B\) admits a continuous representative,
which we denote by
$
B_{\mathrm c}:K\to\mathbb R^{d\times d}.
$
Thus
$
B_{\mathrm c}(x)=B(x)
\qquad
\text{for }\mu\text{-almost every }x
$
for any chosen measurable representative \(B\) of the same \(L^2\) element.
Define
\[
A_{\mathrm c}^\star(x)
:=
P(B_{\mathrm c}(x))
=
\Pi_{\mathcal A_{\mathfrak S}}
\bigl(
B_{\mathrm c}(x)
\bigr).
\]
Since
$
B_{\mathrm c}
$
is continuous and
\(P\) is continuous by
Proposition~\ref{prop:appendix_projection_nonexpansive},
their composition
$
A_{\mathrm c}^\star
=
P\circ B_{\mathrm c}
$
is continuous on \(K\).

More explicitly, if
$
x_n\to x
\qquad
\text{in }K,
$
then continuity of \(B_{\mathrm c}\) gives
$
\|B_{\mathrm c}(x_n)-B_{\mathrm c}(x)\|_F
\longrightarrow
0.
$
By nonexpansiveness of \(P\),
\[
\begin{aligned}
\|A_{\mathrm c}^\star(x_n)-A_{\mathrm c}^\star(x)\|_F
&=
\left\|
P(B_{\mathrm c}(x_n))
-
P(B_{\mathrm c}(x))
\right\|_F
\\
&\le
\|B_{\mathrm c}(x_n)-B_{\mathrm c}(x)\|_F
\\
&\longrightarrow
0.
\end{aligned}
\]
Hence \(A_{\mathrm c}^\star\) is continuous.

Finally, because
$
B_{\mathrm c}(x)=B(x)
\qquad
\text{for }\mu\text{-almost every }x,
$
we have
$
P(B_{\mathrm c}(x))
=
P(B(x))
\qquad
\text{for }\mu\text{-almost every }x.
$
Therefore
$
A_{\mathrm c}^\star(x)
=
A^\star(x)
\qquad
\text{for }\mu\text{-almost every }x.
$
Thus \(A_{\mathrm c}^\star\) is a continuous representative of the unique
minimizing \(L^2\)-equivalence class, and it is given pointwise by
$
x
\longmapsto
\Pi_{\mathcal A_{\mathfrak S}}
\bigl(
B_{\mathrm c}(x)
\bigr).
$
This completes the proof.
\end{proof}

\subsection{Proof of Corollary~\ref{cor:appendix_integrated_residual_bound}}

\begin{proof}
Let
$
A,B
\in
L^2\!\left(
K,\mu;\mathbb R^{d\times d}
\right),
$
and choose measurable representatives of these two \(L^2\)-equivalence
classes. Define
$
M(x)
:=
B(x)-A(x).
$
Since \(A\) and \(B\) are square-integrable matrix fields,
$
M
=
B-A
\in
L^2\!\left(
K,\mu;\mathbb R^{d\times d}
\right).
$
For every \(x\) at which the chosen representatives are defined,
\[
\begin{aligned}
u_B(x)-u_A(x)
&=
B(x)x-A(x)x
\\
&=
\bigl(B(x)-A(x)\bigr)x
\\
&=
M(x)x.
\end{aligned}
\]

We first establish the elementary matrix--vector estimate used throughout
the corollary. Fix such an \(x\in K\), and write
$
M(x)
=
\bigl(M_{ij}(x)\bigr)_{i,j=1}^d.
$
The \(i\)-th coordinate of \(M(x)x\) is
$
(M(x)x)_i
=
\sum_{j=1}^d
M_{ij}(x)x_j.
$
Therefore
\[
\begin{aligned}
\|M(x)x\|_2^2
&=
\sum_{i=1}^d
\left|
(M(x)x)_i
\right|^2
\\
&=
\sum_{i=1}^d
\left|
\sum_{j=1}^d
M_{ij}(x)x_j
\right|^2.
\end{aligned}
\]
For each fixed row index \(i\), the Cauchy--Schwarz inequality in
\(\mathbb R^d\) gives
\[
\left|
\sum_{j=1}^d
M_{ij}(x)x_j
\right|^2
\le
\left(
\sum_{j=1}^d
|M_{ij}(x)|^2
\right)
\left(
\sum_{j=1}^d
|x_j|^2
\right).
\]
Since
$
\sum_{j=1}^d|x_j|^2
=
\|x\|_2^2,
$
we obtain
$
\left|
\sum_{j=1}^d
M_{ij}(x)x_j
\right|^2
\le
\left(
\sum_{j=1}^d
|M_{ij}(x)|^2
\right)
\|x\|_2^2.
$
Summing this inequality over \(i=1,\ldots,d\) yields
\[
\begin{aligned}
\|M(x)x\|_2^2
&\le
\sum_{i=1}^d
\left(
\sum_{j=1}^d
|M_{ij}(x)|^2
\right)
\|x\|_2^2
\\
&=
\left(
\sum_{i=1}^d
\sum_{j=1}^d
|M_{ij}(x)|^2
\right)
\|x\|_2^2
\\
&=
\|M(x)\|_F^2
\|x\|_2^2.
\end{aligned}
\]
Both sides are nonnegative, so taking square roots gives
$
\|M(x)x\|_2
\le
\|M(x)\|_F\|x\|_2.
$
Substituting
$
M(x)=B(x)-A(x)
$
and
$
M(x)x=u_B(x)-u_A(x)
$
gives
\[
\boxed{
\|u_B(x)-u_A(x)\|_2
\le
\|B(x)-A(x)\|_F\,\|x\|_2
}.
\]
Because \(A\) and \(B\) are \(L^2\)-equivalence classes, this statement is
understood for \(\mu\)-almost every \(x\), which proves the first claim.

We next derive the integrated estimate. Since \(K\subset\mathbb R^d\) is
compact and the map
$
x\longmapsto\|x\|_2
$
is continuous, it is bounded on \(K\). Hence
$
R_K
:=
\sup_{x\in K}\|x\|_2
<
\infty,
$
and for every \(x\in K\),
$
\|x\|_2
\le
R_K.
$
Squaring the pointwise estimate gives, for \(\mu\)-almost every \(x\),
\[
\|u_B(x)-u_A(x)\|_2^2
\le
\|B(x)-A(x)\|_F^2
\|x\|_2^2.
\]
Using
$
\|x\|_2^2
\le
R_K^2,
$
we obtain
\[
\|u_B(x)-u_A(x)\|_2^2
\le
R_K^2
\|B(x)-A(x)\|_F^2.
\]
The right-hand side is integrable because
$
B-A
\in
L^2\!\left(
K,\mu;\mathbb R^{d\times d}
\right).
$
In particular, the preceding inequality also shows that
$
u_B-u_A
\in
L^2(K,\mu;\mathbb R^d).
$
Integrating both sides over \(K\) gives
\[
\begin{aligned}
\int_K
\|u_B(x)-u_A(x)\|_2^2
\,d\mu(x)
&\le
\int_K
R_K^2
\|B(x)-A(x)\|_F^2
\,d\mu(x)
\\
&=
R_K^2
\int_K
\|B(x)-A(x)\|_F^2
\,d\mu(x).
\end{aligned}
\]
Thus
\[
\boxed{
\int_K
\|u_B(x)-u_A(x)\|_2^2
\,d\mu(x)
\le
R_K^2
\int_K
\|B(x)-A(x)\|_F^2
\,d\mu(x)
}.
\]

Finally, suppose that
$
A,B:
K\to\mathbb R^{d\times d}
$
are continuous. Then they are defined pointwise on all of \(K\), and the
algebraic argument above applies at every \(x\in K\), rather than merely
for \(\mu\)-almost every \(x\). Hence
$
\|u_B(x)-u_A(x)\|_2
\le
\|B(x)-A(x)\|_F\|x\|_2
\qquad
\text{for every }x\in K.
$
Using
$
\|x\|_2\le R_K
$
gives
$
\|u_B(x)-u_A(x)\|_2
\le
R_K
\|B(x)-A(x)\|_F
\qquad
\text{for every }x\in K.
$
Taking the supremum over \(x\in K\) on both sides yields
\[
\begin{aligned}
\sup_{x\in K}
\|u_B(x)-u_A(x)\|_2
&\le
\sup_{x\in K}
\left[
R_K
\|B(x)-A(x)\|_F
\right]
\\
&=
R_K
\sup_{x\in K}
\|B(x)-A(x)\|_F.
\end{aligned}
\]
Therefore
\[
\boxed{
\sup_{x\in K}
\|u_B(x)-u_A(x)\|_2
\le
R_K
\sup_{x\in K}
\|B(x)-A(x)\|_F
}.
\]
This proves all claims of the corollary.
\end{proof}

\subsection{Proof of Proposition~\ref{prop:appendix_projection_other_families}}

\begin{proof}
Let
$
\mathcal U
\subseteq
\mathbb R^{N\times N}
$
be an arbitrary nonempty closed convex set, and fix
$
B\in\mathbb R^{N\times N}.
$
We regard
\(\mathbb R^{N\times N}\) as a finite-dimensional Hilbert space equipped
with the Frobenius inner product
$
\langle X,Y\rangle_F
:=
\operatorname{tr}(X^\top Y)
$
and the associated norm
$
\|X\|_F
=
\sqrt{\langle X,X\rangle_F}.
$
We first prove directly that the distance from \(B\) to
\(\mathcal U\) is attained.

Define
$
d_\star
:=
\inf_{A\in\mathcal U}
\|B-A\|_F.
$
Because norms are nonnegative,
$
d_\star\ge0.
$
Since
\(\mathcal U\neq\varnothing\), choose and fix an arbitrary
$
A_0\in\mathcal U.
$
Then
$
d_\star
\le
\|B-A_0\|_F
<
\infty.
$
By the definition of the infimum, for every integer \(n\ge1\) there exists
$
A_n\in\mathcal U
$
such that
\[
d_\star
\le
\|B-A_n\|_F
<
d_\star+\frac1n.
\]
In particular,
$
\|B-A_n\|_F
<
d_\star+1
\qquad
\text{for every }n\ge1.
$
Using the triangle inequality,
\[
\begin{aligned}
\|A_n\|_F
&=
\|A_n-B+B\|_F
\\
&\le
\|A_n-B\|_F+\|B\|_F
\\
&<
d_\star+1+\|B\|_F.
\end{aligned}
\]
Thus the sequence
$
\{A_n\}_{n\ge1}
$
is bounded in the finite-dimensional vector space
\(\mathbb R^{N\times N}\).

By finite-dimensional sequential compactness of bounded sets, there exists
a subsequence
$
\{A_{n_j}\}_{j\ge1}
$
and a matrix
$
A^\star\in\mathbb R^{N\times N}
$
such that
$
A_{n_j}
\longrightarrow
A^\star
\qquad
\text{in Frobenius norm}.
$
Every member of the subsequence lies in \(\mathcal U\):
$
A_{n_j}\in\mathcal U.
$
Since \(\mathcal U\) is closed, its limit also belongs to
\(\mathcal U\). Hence
$
A^\star\in\mathcal U.
$

The Frobenius norm is continuous, so
$
\|B-A_{n_j}\|_F
\longrightarrow
\|B-A^\star\|_F.
$
On the other hand, by construction,
\[
d_\star
\le
\|B-A_{n_j}\|_F
<
d_\star+\frac1{n_j}.
\]
Since
$
n_j\longrightarrow\infty,
$
we have
$
\frac1{n_j}\longrightarrow0.
$
Therefore
$
\|B-A_{n_j}\|_F
\longrightarrow
d_\star.
$
The limit of a convergent real sequence is unique, so
$
\|B-A^\star\|_F
=
d_\star.
$
Consequently,
\[
A^\star
\in
\arg\min_{A\in\mathcal U}
\|B-A\|_F^2.
\]
Thus a metric projection exists.

We next prove uniqueness. Suppose that
$
A_1,A_2\in\mathcal U
$
are both minimizers. Then
$
\|B-A_1\|_F^2
=
\|B-A_2\|_F^2
=
d_\star^2.
$
Assume, toward a contradiction, that
$
A_1\neq A_2.
$
Since \(\mathcal U\) is convex, their midpoint
$
\overline A
:=
\frac{A_1+A_2}{2}
$
also belongs to \(\mathcal U\).

We compute its squared distance from \(B\). Set
$
X:=B-A_1,
\qquad
Y:=B-A_2.
$
Then
$
B-\overline A
=
B-\frac{A_1+A_2}{2}
=
\frac{X+Y}{2}.
$
Therefore
\[
\begin{aligned}
\|B-\overline A\|_F^2
&=
\left\|
\frac{X+Y}{2}
\right\|_F^2
\\
&=
\frac14
\|X+Y\|_F^2
\\
&=
\frac14
\left(
\|X\|_F^2
+
\|Y\|_F^2
+
2\langle X,Y\rangle_F
\right).
\end{aligned}
\]
We also have
$
\begin{aligned}
\|X-Y\|_F^2
&=
\|X\|_F^2
+
\|Y\|_F^2
-
2\langle X,Y\rangle_F.
\end{aligned}
$
Solving the latter identity for the inner-product term gives
$
2\langle X,Y\rangle_F
=
\|X\|_F^2
+
\|Y\|_F^2
-
\|X-Y\|_F^2.
$
Substituting this into the expression for
\(\|B-\overline A\|_F^2\) yields
\[
\begin{aligned}
\|B-\overline A\|_F^2
&=
\frac14
\left[
2\|X\|_F^2
+
2\|Y\|_F^2
-
\|X-Y\|_F^2
\right]
\\
&=
\frac12\|X\|_F^2
+
\frac12\|Y\|_F^2
-
\frac14\|X-Y\|_F^2.
\end{aligned}
\]
Since
$
X-Y
=
(B-A_1)-(B-A_2)
=
A_2-A_1,
$
we obtain
\[
\begin{aligned}
\|B-\overline A\|_F^2
&=
\frac12\|B-A_1\|_F^2
+
\frac12\|B-A_2\|_F^2
-
\frac14\|A_1-A_2\|_F^2
\\
&=
d_\star^2
-
\frac14\|A_1-A_2\|_F^2.
\end{aligned}
\]
Because
$
A_1\neq A_2,
$
we have
$
\|A_1-A_2\|_F^2>0,
$
and hence
$
\|B-\overline A\|_F^2
<
d_\star^2.
$
But
$
\overline A\in\mathcal U,
$
so this contradicts the definition of
\(d_\star\) as the minimum distance from \(B\) to
\(\mathcal U\). Therefore
$
A_1=A_2.
$
Thus the minimizer is unique, and we may write
$
\Pi_{\mathcal U}(B)
=
\arg\min_{A\in\mathcal U}
\|B-A\|_F^2.
$

We now verify that each of the geometry-induced families listed in the
proposition satisfies the hypotheses of this general projection principle.

By Proposition~\ref{prop:operators-generator-family-structure},
$
\mathcal L_{\mathfrak S}
$
is a closed polyhedral convex cone in
\(\mathbb R^{d\times d}\). In particular, it is closed and convex.
Moreover,
$
0\in\mathcal L_{\mathfrak S},
$
so it is nonempty. Therefore every
$
B\in\mathbb R^{d\times d}
$
has a unique Frobenius metric projection onto
\(\mathcal L_{\mathfrak S}\).

For the balanced family, Definition~\ref{def:layers-balanced-generator-family}
and the constrained-family notation give
$
\mathcal L_{\mathfrak S}^{\mathrm{bal}}
=
\mathcal L_{\mathfrak S;\mathbf1,\mathbf1}.
$
Proposition~\ref{prop:operators-feasible-compilation}, applied with$
U=\mathbf1,
\qquad
V=\mathbf1,
$
shows that this set is a closed polyhedral convex cone. Equivalently, it is
obtained from
\(\mathcal L_{\mathfrak S}\) by adding the linear constraint
$
\mathbf1^\top L=0.
$
The zero matrix satisfies all of these constraints, so
$
0\in\mathcal L_{\mathfrak S}^{\mathrm{bal}}.
$
Hence the balanced generator family is nonempty, closed, and convex and
therefore admits a unique metric projection.

By
Proposition~\ref{prop:operators-block-valued-interaction-structure},
the block-valued interaction family
$
\mathcal A^{\mathrm{blk}}_{\mathfrak S,\mathcal G}
\subseteq
\mathbb R^{dc\times dc}
$
is a finite-dimensional closed polyhedral convex cone. It contains the zero
matrix, obtained by setting
$
\beta_{rs}=0
\qquad
\text{for all }r,s
$
and
$
D_i=0
\qquad
\text{for every }i.
$
Thus
$
\mathcal A^{\mathrm{blk}}_{\mathfrak S,\mathcal G}
$
is nonempty, closed, and convex, and the general projection principle
applies in the ambient space
\(\mathbb R^{dc\times dc}\).

The block-valued generator family is
\[
\mathcal L^{\mathrm{blk}}_{\mathfrak S,\mathcal G}
=
\left\{
L\in
\mathcal A^{\mathrm{blk}}_{\mathfrak S,\mathcal G}
\;\middle|\;
L_{\alpha\beta}\ge0
\text{ for }\alpha\neq\beta,
\quad
L\mathbf1_{dc}=0
\right\}.
\]
For each pair
\(\alpha\neq\beta\), the condition
$
L_{\alpha\beta}\ge0
$
defines a closed linear half-space in
\(\mathbb R^{dc\times dc}\). The condition
$
L\mathbf1_{dc}=0
$
consists of finitely many homogeneous linear equations and therefore defines
a closed linear subspace. Consequently,
$
\mathcal L^{\mathrm{blk}}_{\mathfrak S,\mathcal G}
$
is the intersection of the closed convex cone
\(\mathcal A^{\mathrm{blk}}_{\mathfrak S,\mathcal G}\)
with finitely many closed convex sets. Hence it is closed and convex.
The zero matrix belongs to this intersection, so the family is also
nonempty. Therefore it likewise admits a unique metric projection in
\(\mathbb R^{dc\times dc}\).

This mathematical projection statement is independent of the semantic
interpretation of the coordinatewise order. However, the generator
interpretation of
\(\mathcal L^{\mathrm{blk}}_{\mathfrak S,\mathcal G}\)
as a positivity-preserving ordered-feature dynamics is appropriate only in
the regime specified in
Definition~\ref{def:operators-block-valued-generator-family}, where the
standard positive cone on the flattened feature space has the intended
meaning. For generic sign-indefinite hidden features, the block-valued
interaction family remains the natural projection target, whereas the
generator terminology should not be imposed merely from the algebraic
definition.

Finally, Proposition~\ref{prop:operators-basic-structure-multiscale-families}
shows that, when the transfer operators are fixed,
$
\mathcal A^{\mathrm{ms}},
\qquad
\widehat{\mathcal L}^{\mathrm{ms}},
\qquad
\mathcal L^{\mathrm{ms}}
$
are all finite-dimensional closed polyhedral convex cones in
\(\mathbb R^{d_0\times d_0}\). Each contains the zero matrix. Indeed, the
zero operator is obtained in
\(\mathcal A^{\mathrm{ms}}\) by taking every scale-wise interaction equal
to zero, and in
\(\widehat{\mathcal L}^{\mathrm{ms}}\) by taking every scale-wise generator
equal to zero. The additional generator constraints defining
\(\mathcal L^{\mathrm{ms}}\) are also satisfied by the zero matrix.
Therefore all three multiscale families are nonempty, closed, and convex.

Thus every family listed in the proposition satisfies the hypotheses of the
general result proved above in its corresponding finite-dimensional ambient
matrix space. Consequently, for every compatible target matrix \(B\), each
family admits a unique Frobenius metric projection.
\end{proof}

\end{document}